\PassOptionsToPackage{unicode}{hyperref}
\PassOptionsToPackage{hyphens}{url}
\PassOptionsToPackage{dvipsnames,svgnames,x11names}{xcolor}
\documentclass[
  letterpaper,
  DIV=11,
  numbers=noendperiod]{scrreprt}
  
 \DeclareUnicodeCharacter{250F}{-}
\DeclareUnicodeCharacter{2501}{-}
\DeclareUnicodeCharacter{2533}{-}
\DeclareUnicodeCharacter{2513}{-}
\DeclareUnicodeCharacter{2503}{|}
\DeclareUnicodeCharacter{2521}{|}
\DeclareUnicodeCharacter{2547}{|}
\DeclareUnicodeCharacter{2529}{|}
\DeclareUnicodeCharacter{2502}{|}
\DeclareUnicodeCharacter{2514}{|}
\DeclareUnicodeCharacter{2500}{|}
\DeclareUnicodeCharacter{2534}{|}
\DeclareUnicodeCharacter{2518}{|}

\usepackage{amsmath,amssymb}
\usepackage{iftex}
\ifPDFTeX
  \usepackage[T1]{fontenc}
  \usepackage[utf8]{inputenc}
  \usepackage{textcomp} 
\else 
  \usepackage{unicode-math}
  \defaultfontfeatures{Scale=MatchLowercase}
  \defaultfontfeatures[\rmfamily]{Ligatures=TeX,Scale=1}
\fi
\usepackage{lmodern}
\ifPDFTeX\else  
\fi
\IfFileExists{upquote.sty}{\usepackage{upquote}}{}
\IfFileExists{microtype.sty}{
  \usepackage[]{microtype}
  \UseMicrotypeSet[protrusion]{basicmath} 
}{}
\makeatletter
\@ifundefined{KOMAClassName}{
  \IfFileExists{parskip.sty}{%
    \usepackage{parskip}
  }{
    \setlength{\parindent}{0pt}
    \setlength{\parskip}{6pt plus 2pt minus 1pt}}
}{
  \KOMAoptions{parskip=half}}
\makeatother
\usepackage{xcolor}
\ifx\paragraph\undefined\else
  \let\oldparagraph\paragraph
  \renewcommand{\paragraph}[1]{\oldparagraph{#1}\mbox{}}
\fi
\ifx\subparagraph\undefined\else
  \let\oldsubparagraph\subparagraph
  \renewcommand{\subparagraph}[1]{\oldsubparagraph{#1}\mbox{}}
\fi

\usepackage{color}
\usepackage{fancyvrb}

\DefineVerbatimEnvironment{Highlighting}{Verbatim}{commandchars=\\\{\}}
\usepackage{framed}
\definecolor{shadecolor}{RGB}{241,243,245}
\newenvironment{Shaded}{\begin{snugshade}}{\end{snugshade}}

\newcommand{\BuiltInTok}[1]{\textcolor[rgb]{0.00,0.23,0.31}{#1}}
\newcommand{\CharTok}[1]{\textcolor[rgb]{0.13,0.47,0.30}{#1}}
\newcommand{\CommentTok}[1]{\textcolor[rgb]{0.37,0.37,0.37}{#1}}

\newcommand{\ControlFlowTok}[1]{\textcolor[rgb]{0.00,0.23,0.31}{#1}}
\newcommand{\DataTypeTok}[1]{\textcolor[rgb]{0.68,0.00,0.00}{#1}}
\newcommand{\DecValTok}[1]{\textcolor[rgb]{0.68,0.00,0.00}{#1}}

\newcommand{\ErrorTok}[1]{\textcolor[rgb]{0.68,0.00,0.00}{#1}}

\newcommand{\FloatTok}[1]{\textcolor[rgb]{0.68,0.00,0.00}{#1}}
\newcommand{\FunctionTok}[1]{\textcolor[rgb]{0.28,0.35,0.67}{#1}}
\newcommand{\ImportTok}[1]{\textcolor[rgb]{0.00,0.46,0.62}{#1}}

\newcommand{\KeywordTok}[1]{\textcolor[rgb]{0.00,0.23,0.31}{#1}}
\newcommand{\NormalTok}[1]{\textcolor[rgb]{0.00,0.23,0.31}{#1}}
\newcommand{\OperatorTok}[1]{\textcolor[rgb]{0.37,0.37,0.37}{#1}}
\newcommand{\OtherTok}[1]{\textcolor[rgb]{0.00,0.23,0.31}{#1}}

\newcommand{\SpecialCharTok}[1]{\textcolor[rgb]{0.37,0.37,0.37}{#1}}
\newcommand{\SpecialStringTok}[1]{\textcolor[rgb]{0.13,0.47,0.30}{#1}}
\newcommand{\StringTok}[1]{\textcolor[rgb]{0.13,0.47,0.30}{#1}}
\newcommand{\VariableTok}[1]{\textcolor[rgb]{0.07,0.07,0.07}{#1}}
\newcommand{\VerbatimStringTok}[1]{\textcolor[rgb]{0.13,0.47,0.30}{#1}}

\providecommand{\tightlist}{%
  \setlength{\itemsep}{0pt}\setlength{\parskip}{0pt}}\usepackage{longtable,booktabs,array}
\usepackage{calc} 
\usepackage{etoolbox}
\makeatletter
\patchcmd\longtable{\par}{\if@noskipsec\mbox{}\fi\par}{}{}
\makeatother
\IfFileExists{footnotehyper.sty}{\usepackage{footnotehyper}}{\usepackage{footnote}}
\makesavenoteenv{longtable}
\usepackage{graphicx}
\makeatletter
\def\maxwidth{\ifdim\Gin@nat@width>\linewidth\linewidth\else\Gin@nat@width\fi}
\def\maxheight{\ifdim\Gin@nat@height>\textheight\textheight\else\Gin@nat@height\fi}
\makeatother
\setkeys{Gin}{width=\maxwidth,height=\maxheight,keepaspectratio}
\makeatletter
\def\fps@figure{htbp}
\makeatother
\newlength{\cslhangindent}
\newlength{\csllabelwidth}
\newlength{\cslentryspacingunit} 
\newenvironment{CSLReferences}[2] 
 {
  \setlength{\parindent}{0pt}
  \ifodd #1
  \let\oldpar\par
  \def\par{\hangindent=\cslhangindent\oldpar}
  \fi
  \setlength{\parskip}{#2\cslentryspacingunit}
 }%
 {}
\usepackage{calc}

\KOMAoption{captions}{tableheading}
\makeatletter
\@ifpackageloaded{tcolorbox}{}{\usepackage[skins,breakable]{tcolorbox}}
\@ifpackageloaded{fontawesome5}{}{\usepackage{fontawesome5}}
\definecolor{quarto-callout-color}{HTML}{909090}
\definecolor{quarto-callout-note-color}{HTML}{0758E5}
\definecolor{quarto-callout-important-color}{HTML}{CC1914}
\definecolor{quarto-callout-warning-color}{HTML}{EB9113}
\definecolor{quarto-callout-tip-color}{HTML}{00A047}
\definecolor{quarto-callout-caution-color}{HTML}{FC5300}
\definecolor{quarto-callout-color-frame}{HTML}{acacac}
\definecolor{quarto-callout-note-color-frame}{HTML}{4582ec}
\definecolor{quarto-callout-important-color-frame}{HTML}{d9534f}
\definecolor{quarto-callout-warning-color-frame}{HTML}{f0ad4e}
\definecolor{quarto-callout-tip-color-frame}{HTML}{02b875}
\definecolor{quarto-callout-caution-color-frame}{HTML}{fd7e14}
\makeatother
\makeatletter
\@ifpackageloaded{bookmark}{}{\usepackage{bookmark}}
\makeatother
\makeatletter
\@ifpackageloaded{caption}{}{\usepackage{caption}}
\AtBeginDocument{%
\ifdefined\contentsname
  \renewcommand*\contentsname{Table of contents}
\else
  \newcommand\contentsname{Table of contents}
\fi
\ifdefined\listfigurename
  \renewcommand*\listfigurename{List of Figures}
\else
  \newcommand\listfigurename{List of Figures}
\fi
\ifdefined\listtablename
  \renewcommand*\listtablename{List of Tables}
\else
  \newcommand\listtablename{List of Tables}
\fi
\ifdefined\figurename
  \renewcommand*\figurename{Figure}
\else
  \newcommand\figurename{Figure}
\fi
\ifdefined\tablename
  \renewcommand*\tablename{Table}
\else
  \newcommand\tablename{Table}
\fi
}
\@ifpackageloaded{float}{}{\usepackage{float}}
\floatstyle{ruled}
\@ifundefined{c@chapter}{\newfloat{codelisting}{h}{lop}}{\newfloat{codelisting}{h}{lop}[chapter]}
\floatname{codelisting}{Listing}

\makeatother
\makeatletter
\@ifpackageloaded{caption}{}{\usepackage{caption}}
\@ifpackageloaded{subcaption}{}{\usepackage{subcaption}}
\makeatother
\makeatletter
\@ifpackageloaded{tcolorbox}{}{\usepackage[skins,breakable]{tcolorbox}}
\makeatother
\makeatletter
\@ifundefined{shadecolor}{\definecolor{shadecolor}{rgb}{.97, .97, .97}}
\makeatother
\ifLuaTeX
  \usepackage{selnolig}  
\fi
\IfFileExists{bookmark.sty}{\usepackage{bookmark}}{\usepackage{hyperref}}
\IfFileExists{xurl.sty}{\usepackage{xurl}}{} 
\hypersetup{
  pdftitle={Hyperparameter Tuning Cookbook},
  pdfauthor={Thomas Bartz-Beielstein},
  colorlinks=true,
  linkcolor={blue},
  filecolor={Maroon},
  citecolor={Blue},
  urlcolor={Blue},
  pdfcreator={LaTeX via pandoc}}

\title{Hyperparameter Tuning Cookbook}
\usepackage{etoolbox}
\makeatletter
\providecommand{\subtitle}[1]{
  \apptocmd{\@title}{\par {\large #1 \par}}{}{}
}
\makeatother
\subtitle{A guide for scikit-learn, PyTorch, river, and spotPython}
\author{Thomas Bartz-Beielstein}
\date{Jul 17, 2023}

\begin{document}
\maketitle
\ifdefined\Shaded\renewenvironment{Shaded}{\begin{tcolorbox}[interior hidden, borderline west={3pt}{0pt}{shadecolor}, sharp corners, enhanced, breakable, frame hidden, boxrule=0pt]}{\end{tcolorbox}}\fi

\renewcommand*\contentsname{Table of contents}
{
\hypersetup{linkcolor=}
\setcounter{tocdepth}{2}
\tableofcontents
}
\bookmarksetup{startatroot}

\hypertarget{preface-optimization-and-hyperparameter-tuning}{%
\chapter*{Preface: Optimization and Hyperparameter
Tuning}\label{preface-optimization-and-hyperparameter-tuning}}
\addcontentsline{toc}{chapter}{Preface: Optimization and Hyperparameter
Tuning}

\markboth{Preface: Optimization and Hyperparameter Tuning}{Preface:
Optimization and Hyperparameter Tuning}

\begin{quote}
This document provides a comprehensive guide to hyperparameter tuning
using spotPython for scikit-learn, PyTorch, and river. The first part
introduces spotPython's surrogate model-based optimization process,
while the second part focuses on hyperparameter tuning. Several case
studies are presented, including hyperparameter tuning for sklearn
models such as Support Vector Classification, Random Forests, Gradient
Boosting (XGB), and K-nearest neighbors (KNN), as well as a Hoeffding
Adaptive Tree Regressor from river. The integration of spotPython into
the PyTorch and PyTorch Lightning training workflow is also discussed.
With a hands-on approach and step-by-step explanations, this cookbook
serves as a practical starting point for anyone interested in
hyperparameter tuning with Python. Highlights include the interplay
between Tensorboard, PyTorch Lightning, spotPython, and river. This
publication is under development, with updates available on the
corresponding webpage.
\end{quote}

The goal of hyperparameter tuning is to optimize the hyperparameters in
a way that improves the performance of the machine learning or deep
learning model. Hyperparameters are parameters that are not learned
during the training process, but are set before the training process
begins. Hyperparameter tuning is an important, but often difficult and
computationally intensive task. Changing the architecture of a neural
network or the learning rate of an optimizer can have a significant
impact on the performance.

Hyperparameter tuning is referred to as ``hyperparameter optimization''
(HPO) in the literature. However, since we do not consider the
optimization, but also the understanding of the hyperparameters, we use
the term ``hyperparameter tuning'' in this book. See also the discussion
in Chapter 2 of Bartz et al. (2022), which lays the groundwork and
presents an introduction to the process of tuning Machine Learning and
Deep Learning hyperparameters and the respective methodology. Since the
key elements such as the hyperparameter tuning process and measures of
tunability and performance are presented in Bartz et al. (2022), we
refer to this chapter for details.

The simplest, but also most computationally expensive, hyperparameter
tuning approach uses manual search (or trial-and-error (Meignan et al.
2015)). Commonly encountered is simple random search, i.e., random and
repeated selection of hyperparameters for evaluation, and lattice search
(``grid search''). In addition, methods that perform directed search and
other model-free algorithms, i.e., algorithms that do not explicitly
rely on a model, e.g., evolution strategies (Bartz-Beielstein et al.
2014) or pattern search (Lewis, Torczon, and Trosset 2000) play an
important role. Also, ``hyperband'', i.e., a multi-armed bandit strategy
that dynamically allocates resources to a set of random configurations
and uses successive bisections to stop configurations with poor
performance (Li et al. 2016), is very common in hyperparameter tuning.
The most sophisticated and efficient approaches are the Bayesian
optimization and surrogate model based optimization methods, which are
based on the optimization of cost functions determined by simulations or
experiments.

We consider a surrogate optimization based hyperparameter tuning
approach that uses the Python version of the SPOT (``Sequential
Parameter Optimization Toolbox'') (Bartz-Beielstein, Lasarczyk, and
Preuss 2005), which is suitable for situations where only limited
resources are available. This may be due to limited availability and
cost of hardware, or due to the fact that confidential data may only be
processed locally, e.g., due to legal requirements. Furthermore, in our
approach, the understanding of algorithms is seen as a key tool for
enabling transparency and explainability. This can be enabled, for
example, by quantifying the contribution of machine learning and deep
learning components (nodes, layers, split decisions, activation
functions, etc.). Understanding the importance of hyperparameters and
the interactions between multiple hyperparameters plays a major role in
the interpretability and explainability of machine learning models. SPOT
provides statistical tools for understanding hyperparameters and their
interactions. Last but not least, it should be noted that the SPOT
software code is available in the open source \texttt{spotPython}
package on github\footnote{\url{https://github.com/sequential-parameter-optimization}},
allowing replicability of the results. This tutorial describes the
Python variant of SPOT, which is called \texttt{spotPython}. The R
implementation is described in Bartz et al. (2022). SPOT is an
established open source software that has been maintained for more than
15 years (Bartz-Beielstein, Lasarczyk, and Preuss 2005) (Bartz et al.
2022).

\begin{tcolorbox}[enhanced jigsaw, left=2mm, title=\textcolor{quarto-callout-important-color}{\faExclamation}\hspace{0.5em}{Important: This book is still under development.}, bottomrule=.15mm, titlerule=0mm, breakable, rightrule=.15mm, toprule=.15mm, coltitle=black, colbacktitle=quarto-callout-important-color!10!white, leftrule=.75mm, arc=.35mm, colframe=quarto-callout-important-color-frame, bottomtitle=1mm, colback=white, opacitybacktitle=0.6, toptitle=1mm, opacityback=0]

The most recent version of this book is available at
\url{https://sequential-parameter-optimization.github.io/Hyperparameter-Tuning-Cookbook/}

\end{tcolorbox}

\hypertarget{book-structure}{%
\section*{Book Structure}\label{book-structure}}
\addcontentsline{toc}{section}{Book Structure}

\markright{Book Structure}

This document is structured in two parts. The first part describes the
surrogate model based optimization process and the second part describes
the hyperparameter tuning.

The first part is structured as follows: The concept of the
hyperparameter tuning software \texttt{spotPython} is described in
Chapter~\ref{sec-spot}. This introduction is based on one-dimensional
examples. Higher-dimensional examples are presented in
Chapter~\ref{sec-multi-dim}. Chapter~\ref{sec-iso-aniso-kriging}
describes isotropic and anisotorpic kriging. How different surrogate
models from \texttt{scikit-learn} can be used as surrogates in
\texttt{spotPython} optimization runs is explained in
Chapter~\ref{sec-sklearn-surrogates}. Chapter~\ref{sec-scipy-optimizers}
describes how different optimizers from the \texttt{scipy\ optimize}
package can be used on the surrogate. The differences between the
\texttt{Kriging} implementation in \texttt{spotPython} and the
\texttt{GaussianProcessRegressor} in \texttt{scikit-learn} are explained
in Chapter~\ref{sec-gaussian-process-models}.
Chapter~\ref{sec-expected-improvement} describes the expected
improvement approach. How noisy functions can be handled is described in
Chapter~\ref{sec-noise}. Chapter~\ref{sec-ocba} demonstrates how noisy
functions can be handled with Optimal Computational Budget Allocation
(OCBA) by \texttt{Spot}.

The second part is structured as follows:
Chapter~\ref{sec-hpt-sklearn-svc} describes the hyperparameter tuning of
a
\href{https://scikit-learn.org/stable/modules/generated/sklearn.svm.SVC.html\#sklearn.svm.SVCriver\%20python}{support
vector classifier} from \texttt{scikit-learn} with \texttt{spotPython}.
Chapter~\ref{sec-river-hpt} illustrates the hyperparameter tuning of a
\href{https://riverml.xyz/0.18.0/api/tree/HoeffdingAdaptiveTreeRegressor/}{Hoeffding
Adaptive Tree Regressor} from \texttt{river} with \texttt{spotPython}.

Chapter~\ref{sec-hyperparameter-tuning-for-pytorch-14} describes the
execution of the example from the tutorial ``Hyperparameter Tuning with
Ray Tune'' (PyTorch 2023a). The integration of \texttt{spotPython} into
the \texttt{PyTorch} training workflow is described in detail in the
following sections. Section~\ref{sec-setup-14} describes the setup of
the tuners. Section~\ref{sec-data-loading-14} describes the data
loading. Section~\ref{sec-selection-of-the-algorithm-14} describes the
model to be tuned. The search space is introduced in
Section~\ref{sec-search-space-14}. Optimizers are presented in
Section~\ref{sec-optimizers-14}. How to split the data in train,
validation, and test sets is described in
Section~\ref{sec-data-splitting-14}. The selection of the loss function
and metrics is described in Section~\ref{sec-loss-functions-14}.
Section~\ref{sec-prepare-spot-call-14} describes the preparation of the
\texttt{spotPython} call. The objective function is described in
Section~\ref{sec-the-objective-function-14}. How to use results from
previous runs and default hyperparameter configurations is described in
Section~\ref{sec-default-hyperparameters}. Starting the tuner is shown
in Section~\ref{sec-call-the-hyperparameter-tuner-14}. TensorBoard can
be used to visualize the results as shown in
Section~\ref{sec-tensorboard-14}. Results are discussed and explained in
Section~\ref{sec-results-14}. Section~\ref{sec-summary} presents a
summary and an outlook for the execution of the example from the
tutorial ``Hyperparameter Tuning with Ray Tune''.

Four more examples are presented in the following sections:
Chapter~\ref{sec-hpt-random-forest-classifier} describes the
hyperparameter tuning of a
\href{https://scikit-learn.org/stable/modules/generated/sklearn.ensemble.RandomForestClassifier.html}{random
forest classifier} from \texttt{scikit-learn} with \texttt{spotPython}.
Chapter~\ref{sec-hpt-sklearn-xgb-classifier-vbdp-data} describes the
hyperparameter tuning of an
\href{https://scikit-learn.org/stable/modules/ensemble.html\#histogram-based-gradient-boosting}{XGBoost
classifier} from \texttt{scikit-learn} with \texttt{spotPython}.
Chapter~\ref{sec-hpt-sklearn-svc-vbdp-data} describes the hyperparameter
tuning of a
\href{https://scikit-learn.org/stable/modules/generated/sklearn.svm.SVC.html\#sklearn.svm.SVC}{support
vector classifier} from \texttt{scikit-learn} with \texttt{spotPython}.
Chapter~\ref{sec-hpt-sklearn-knn-classifier-vbdp-data} describes the
hyperparameter tuning of a
\href{https://scikit-learn.org/stable/modules/generated/sklearn.neighbors.KNeighborsClassifier.html}{k-nearest
neighbors classifier} from \texttt{scikit-learn} with
\texttt{spotPython}.

This part of the book is concluded with a description of the most recent
\texttt{PyTorch} hyperparameter tuning approach, which is the
integration of \texttt{spotPython} into the \texttt{PyTorch\ Lightning}
training workflow. This is described in
Chapter~\ref{sec-hyperparameter-tuning-lightning-31}. This is considered
as the most effective, efficient, and flexible way to integrate
\texttt{spotPython} into the \texttt{PyTorch} training workflow.

\begin{tcolorbox}[enhanced jigsaw, left=2mm, title=\textcolor{quarto-callout-tip-color}{\faLightbulb}\hspace{0.5em}{Hyperparameter Tuning Reference}, bottomrule=.15mm, titlerule=0mm, breakable, rightrule=.15mm, toprule=.15mm, coltitle=black, colbacktitle=quarto-callout-tip-color!10!white, leftrule=.75mm, arc=.35mm, colframe=quarto-callout-tip-color-frame, bottomtitle=1mm, colback=white, opacitybacktitle=0.6, toptitle=1mm, opacityback=0]

\begin{itemize}
\tightlist
\item
  The open access book Bartz et al. (2022) provides a comprehensive
  overview of hyperparameter tuning. It can be downloaded from
  \url{https://link.springer.com/book/10.1007/978-981-19-5170-1}.
\end{itemize}

\end{tcolorbox}

\begin{tcolorbox}[enhanced jigsaw, left=2mm, title=\textcolor{quarto-callout-note-color}{\faInfo}\hspace{0.5em}{Note}, bottomrule=.15mm, titlerule=0mm, breakable, rightrule=.15mm, toprule=.15mm, coltitle=black, colbacktitle=quarto-callout-note-color!10!white, leftrule=.75mm, arc=.35mm, colframe=quarto-callout-note-color-frame, bottomtitle=1mm, colback=white, opacitybacktitle=0.6, toptitle=1mm, opacityback=0]

The \texttt{.ipynb} notebook (Bartz-Beielstein 2023) is updated
regularly and reflects updates and changes in the \texttt{spotPython}
package. It can be downloaded from
\url{https://github.com/sequential-parameter-optimization/spotPython/blob/main/notebooks/14_spot_ray_hpt_torch_cifar10.ipynb}.

\end{tcolorbox}

\hypertarget{software-used-in-this-book}{%
\section*{Software Used in this Book}\label{software-used-in-this-book}}
\addcontentsline{toc}{section}{Software Used in this Book}

\markright{Software Used in this Book}

\href{https://github.com/sequential-parameter-optimization/spotPython}{spotPython}
(``Sequential Parameter Optimization Toolbox in Python'') is the Python
version of the well-known hyperparameter tuner SPOT, which has been
developed in the R programming environment for statistical analysis for
over a decade. The related open-access book is available here:
\href{https://link.springer.com/book/10.1007/978-981-19-5170-1}{Hyperparameter
Tuning for Machine and Deep Learning with R---A Practical Guide}.

\href{https://scikit-learn.org}{scikit-learn} is a Python module for
machine learning built on top of SciPy and is distributed under the
3-Clause BSD license. The project was started in 2007 by David
Cournapeau as a Google Summer of Code project, and since then many
volunteers have contributed.

\href{https://pytorch.org}{PyTorch} is an optimized tensor library for
deep learning using GPUs and CPUs.
\href{https://lightning.ai/docs/pytorch/latest/}{Lightning} is a
lightweight PyTorch wrapper for high-performance AI research. It allows
you to decouple the research from the engineering.

\href{https://riverml.xyz}{River} is a Python library for online machine
learning. It is designed to be used in real-world environments, where
not all data is available at once, but streaming in.

\href{https://github.com/sequential-parameter-optimization/spotRiver}{spotRiver}
provides an interface between
\href{https://github.com/sequential-parameter-optimization/spotPython}{spotPython}
and \href{https://riverml.xyz}{River}.

\part{Spot as an Optimizer}

\hypertarget{sec-spot}{%
\chapter{\texorpdfstring{Introduction to
\texttt{spotPython}}{Introduction to spotPython}}\label{sec-spot}}

Surrogate model based optimization methods are common approaches in
simulation and optimization. SPOT was developed because there is a great
need for sound statistical analysis of simulation and optimization
algorithms. SPOT includes methods for tuning based on classical
regression and analysis of variance techniques. It presents tree-based
models such as classification and regression trees and random forests as
well as Bayesian optimization (Gaussian process models, also known as
Kriging). Combinations of different meta-modeling approaches are
possible. SPOT comes with a sophisticated surrogate model based
optimization method, that can handle discrete and continuous inputs.
Furthermore, any model implemented in \texttt{scikit-learn} can be used
out-of-the-box as a surrogate in \texttt{spotPython}.

SPOT implements key techniques such as exploratory fitness landscape
analysis and sensitivity analysis. It can be used to understand the
performance of various algorithms, while simultaneously giving insights
into their algorithmic behavior.

The \texttt{spot} loop consists of the following steps:

\begin{enumerate}
\def\labelenumi{\arabic{enumi}.}
\tightlist
\item
  Init: Build initial design \(X\)
\item
  Evaluate initial design on real objective \(f\): \(y = f(X)\)
\item
  Build surrogate: \(S = S(X,y)\)
\item
  Optimize on surrogate: \(X_0 = \text{optimize}(S)\)
\item
  Evaluate on real objective: \(y_0 = f(X_0)\)
\item
  Impute (Infill) new points: \(X = X \cup X_0\), \(y = y \cup y_0\).
\item
  Got 3.
\end{enumerate}

Central Idea: Evaluation of the surrogate model \texttt{S} is much
cheaper (or / and much faster) than running the real-world experiment
\(f\). We start with a small example.

\hypertarget{example-spot-and-the-sphere-function}{%
\section{\texorpdfstring{Example: \texttt{Spot} and the Sphere
Function}{Example: Spot and the Sphere Function}}\label{example-spot-and-the-sphere-function}}

\begin{Shaded}
\begin{Highlighting}[]
\ImportTok{import}\NormalTok{ numpy }\ImportTok{as}\NormalTok{ np}
\ImportTok{from}\NormalTok{ math }\ImportTok{import}\NormalTok{ inf}
\ImportTok{from}\NormalTok{ spotPython.fun.objectivefunctions }\ImportTok{import}\NormalTok{ analytical}
\ImportTok{from}\NormalTok{ spotPython.spot }\ImportTok{import}\NormalTok{ spot}
\ImportTok{from}\NormalTok{ scipy.optimize }\ImportTok{import}\NormalTok{ shgo}
\ImportTok{from}\NormalTok{ scipy.optimize }\ImportTok{import}\NormalTok{ direct}
\ImportTok{from}\NormalTok{ scipy.optimize }\ImportTok{import}\NormalTok{ differential\_evolution}
\ImportTok{import}\NormalTok{ matplotlib.pyplot }\ImportTok{as}\NormalTok{ plt}
\end{Highlighting}
\end{Shaded}

\hypertarget{the-objective-function-sphere}{%
\subsection{The Objective Function:
Sphere}\label{the-objective-function-sphere}}

The \texttt{spotPython} package provides several classes of objective
functions. We will use an analytical objective function, i.e., a
function that can be described by a (closed) formula: \[f(x) = x^2\]

\begin{Shaded}
\begin{Highlighting}[]
\NormalTok{fun }\OperatorTok{=}\NormalTok{ analytical().fun\_sphere}
\end{Highlighting}
\end{Shaded}

We can apply the function \texttt{fun} to input values and plot the
result:

\begin{Shaded}
\begin{Highlighting}[]
\NormalTok{x }\OperatorTok{=}\NormalTok{ np.linspace(}\OperatorTok{{-}}\DecValTok{1}\NormalTok{,}\DecValTok{1}\NormalTok{,}\DecValTok{100}\NormalTok{).reshape(}\OperatorTok{{-}}\DecValTok{1}\NormalTok{,}\DecValTok{1}\NormalTok{)}
\NormalTok{y }\OperatorTok{=}\NormalTok{ fun(x)}
\NormalTok{plt.figure()}
\NormalTok{plt.plot(x, y, }\StringTok{"k"}\NormalTok{)}
\NormalTok{plt.show()}
\end{Highlighting}
\end{Shaded}

\begin{figure}[H]

{\centering \includegraphics{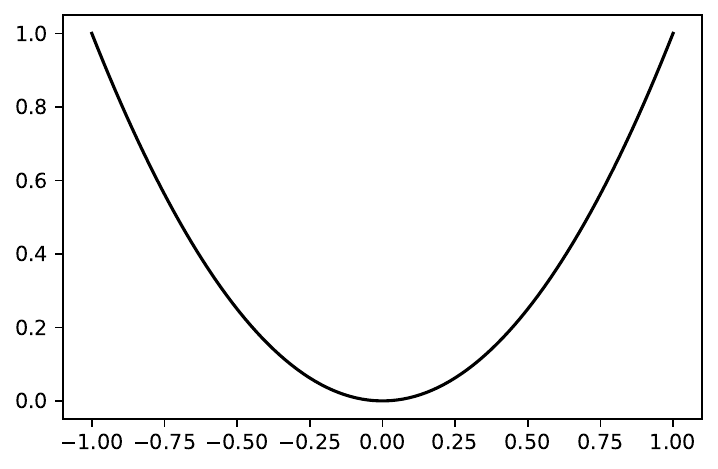}

}

\end{figure}

\begin{Shaded}
\begin{Highlighting}[]
\NormalTok{spot\_0 }\OperatorTok{=}\NormalTok{ spot.Spot(fun}\OperatorTok{=}\NormalTok{fun,}
\NormalTok{                   lower }\OperatorTok{=}\NormalTok{ np.array([}\OperatorTok{{-}}\DecValTok{1}\NormalTok{]),}
\NormalTok{                   upper }\OperatorTok{=}\NormalTok{ np.array([}\DecValTok{1}\NormalTok{]))}
\end{Highlighting}
\end{Shaded}

\begin{Shaded}
\begin{Highlighting}[]
\NormalTok{spot\_0.run()}
\end{Highlighting}
\end{Shaded}

\begin{verbatim}
spotPython tuning: 7.263311682641849e-09 [#######---] 73.33% 
\end{verbatim}

\begin{verbatim}
spotPython tuning: 7.263311682641849e-09 [########--] 80.00% 
\end{verbatim}

\begin{verbatim}
spotPython tuning: 7.263311682641849e-09 [#########-] 86.67% 
\end{verbatim}

\begin{verbatim}
spotPython tuning: 7.263311682641849e-09 [#########-] 93.33% 
\end{verbatim}

\begin{verbatim}
spotPython tuning: 3.696886711914087e-10 [##########] 100.00% Done...
\end{verbatim}

\begin{verbatim}
<spotPython.spot.spot.Spot at 0x168361ff0>
\end{verbatim}

\begin{Shaded}
\begin{Highlighting}[]
\NormalTok{spot\_0.print\_results()}
\end{Highlighting}
\end{Shaded}

\begin{verbatim}
min y: 3.696886711914087e-10
x0: 1.922728975158508e-05
\end{verbatim}

\begin{verbatim}
[['x0', 1.922728975158508e-05]]
\end{verbatim}

\begin{Shaded}
\begin{Highlighting}[]
\NormalTok{spot\_0.plot\_progress(log\_y}\OperatorTok{=}\VariableTok{True}\NormalTok{)}
\end{Highlighting}
\end{Shaded}

\begin{figure}[H]

{\centering \includegraphics{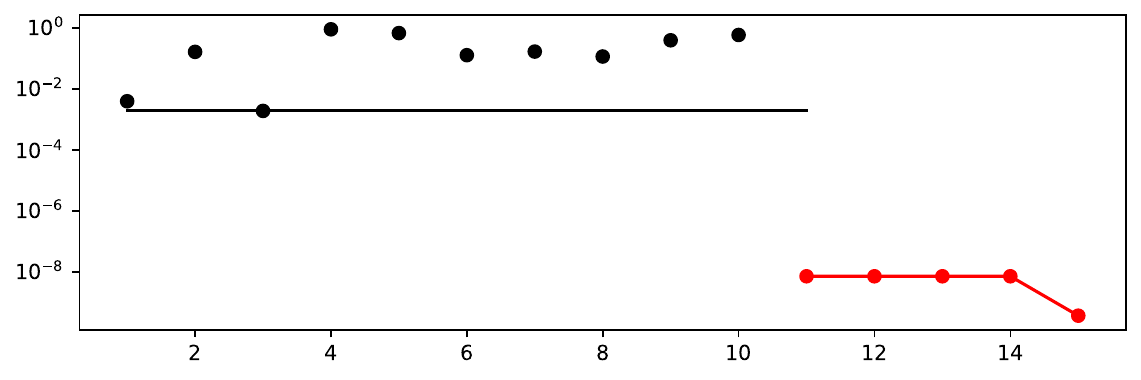}

}

\end{figure}

\begin{Shaded}
\begin{Highlighting}[]
\NormalTok{spot\_0.plot\_model()}
\end{Highlighting}
\end{Shaded}

\begin{figure}[H]

{\centering \includegraphics{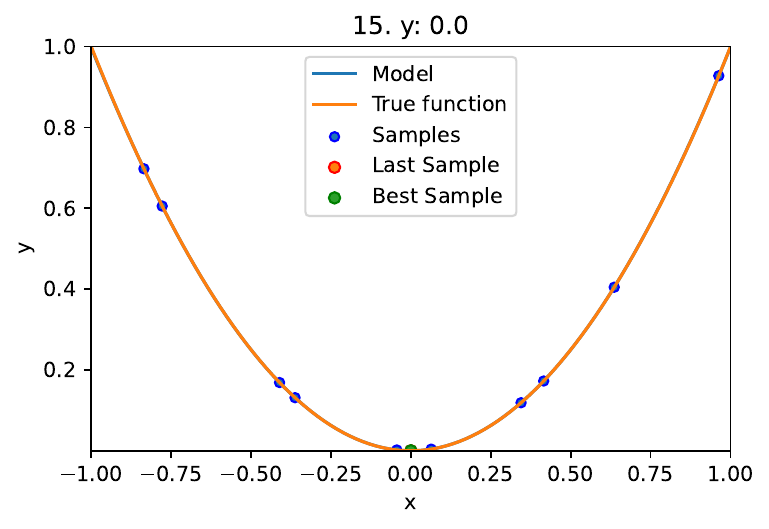}

}

\end{figure}

\hypertarget{spot-parameters-fun_evals-init_size-and-show_models}{%
\section{\texorpdfstring{\texttt{Spot} Parameters: \texttt{fun\_evals},
\texttt{init\_size} and
\texttt{show\_models}}{Spot Parameters: fun\_evals, init\_size and show\_models}}\label{spot-parameters-fun_evals-init_size-and-show_models}}

We will modify three parameters:

\begin{enumerate}
\def\labelenumi{\arabic{enumi}.}
\tightlist
\item
  The number of function evaluations (\texttt{fun\_evals})
\item
  The size of the initial design (\texttt{init\_size})
\item
  The parameter \texttt{show\_models}, which visualizes the search
  process for 1-dim functions.
\end{enumerate}

The full list of the \texttt{Spot} parameters is shown in the Help
System and in the notebook \texttt{spot\_doc.ipynb}.

\begin{Shaded}
\begin{Highlighting}[]
\NormalTok{spot\_1 }\OperatorTok{=}\NormalTok{ spot.Spot(fun}\OperatorTok{=}\NormalTok{fun,}
\NormalTok{                   lower }\OperatorTok{=}\NormalTok{ np.array([}\OperatorTok{{-}}\DecValTok{1}\NormalTok{]),}
\NormalTok{                   upper }\OperatorTok{=}\NormalTok{ np.array([}\DecValTok{2}\NormalTok{]),}
\NormalTok{                   fun\_evals}\OperatorTok{=} \DecValTok{10}\NormalTok{,}
\NormalTok{                   seed}\OperatorTok{=}\DecValTok{123}\NormalTok{,}
\NormalTok{                   show\_models}\OperatorTok{=}\VariableTok{True}\NormalTok{,}
\NormalTok{                   design\_control}\OperatorTok{=}\NormalTok{\{}\StringTok{"init\_size"}\NormalTok{: }\DecValTok{9}\NormalTok{\})}
\NormalTok{spot\_1.run()}
\end{Highlighting}
\end{Shaded}

\begin{figure}[H]

{\centering \includegraphics{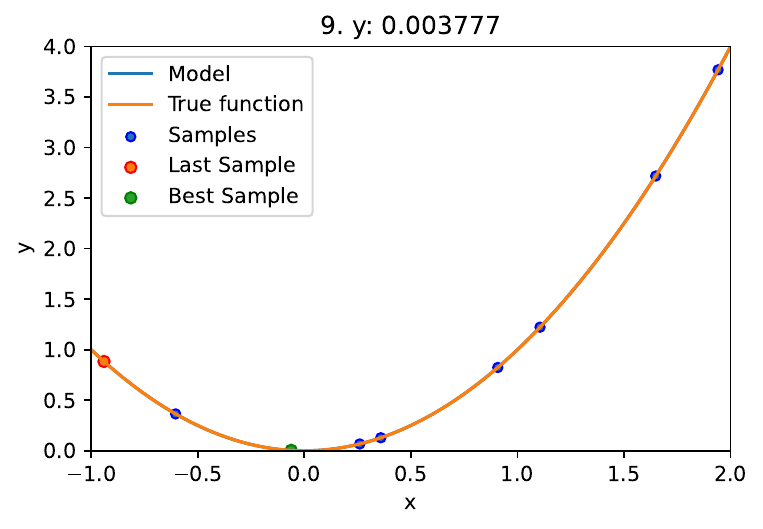}

}

\end{figure}

\begin{figure}[H]

{\centering \includegraphics{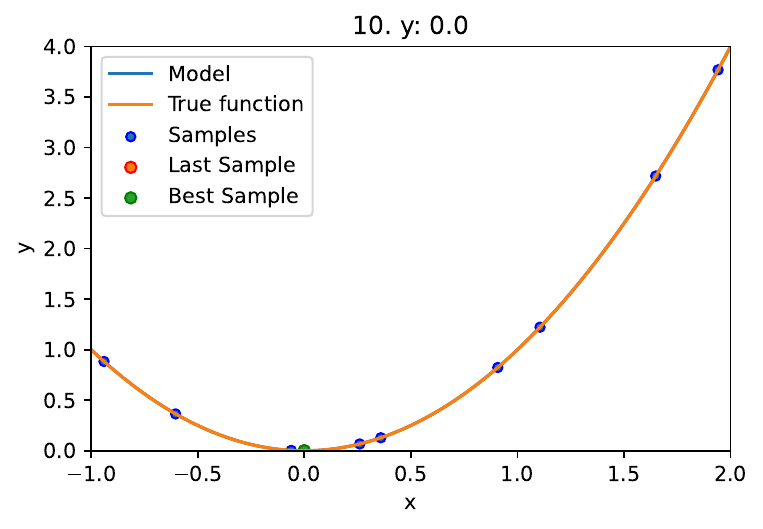}

}

\end{figure}

\begin{verbatim}
spotPython tuning: 3.6779240309761575e-07 [##########] 100.00% Done...
\end{verbatim}

\begin{verbatim}
<spotPython.spot.spot.Spot at 0x168f98910>
\end{verbatim}

\hypertarget{print-the-results}{%
\section{Print the Results}\label{print-the-results}}

\begin{Shaded}
\begin{Highlighting}[]
\NormalTok{spot\_1.print\_results()}
\end{Highlighting}
\end{Shaded}

\begin{verbatim}
min y: 3.6779240309761575e-07
x0: -0.0006064589047063418
\end{verbatim}

\begin{verbatim}
[['x0', -0.0006064589047063418]]
\end{verbatim}

\hypertarget{show-the-progress}{%
\section{Show the Progress}\label{show-the-progress}}

\begin{Shaded}
\begin{Highlighting}[]
\NormalTok{spot\_1.plot\_progress()}
\end{Highlighting}
\end{Shaded}

\begin{figure}[H]

{\centering \includegraphics{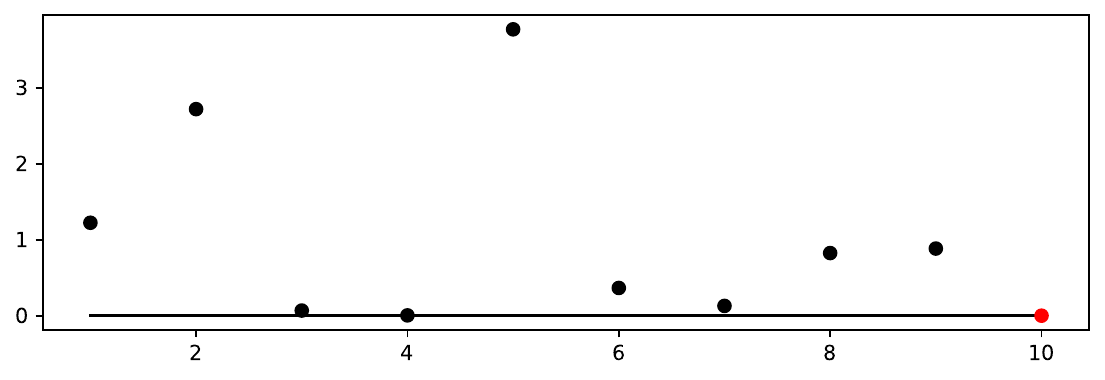}

}

\end{figure}

\hypertarget{sec-visualizing-tensorboard-01}{%
\section{Visualizing the Optimization and Hyperparameter Tuning Process
with TensorBoard}\label{sec-visualizing-tensorboard-01}}

\texttt{spotPython} supports the visualization of the hyperparameter
tuning process with TensorBoard. The following example shows how to use
TensorBoard with \texttt{spotPython}.

First, we define an ``experiment name'' to identify the hyperparameter
tuning process. The experiment name is used to create a directory for
the TensorBoard files.

\begin{Shaded}
\begin{Highlighting}[]
\ImportTok{from}\NormalTok{ spotPython.utils.}\BuiltInTok{file} \ImportTok{import}\NormalTok{ get\_experiment\_name}
\ImportTok{from}\NormalTok{ spotPython.utils.init }\ImportTok{import}\NormalTok{ fun\_control\_init}
\ImportTok{from}\NormalTok{ spotPython.utils.}\BuiltInTok{file} \ImportTok{import}\NormalTok{ get\_spot\_tensorboard\_path}

\NormalTok{PREFIX }\OperatorTok{=} \StringTok{"01"}
\NormalTok{experiment\_name }\OperatorTok{=}\NormalTok{ get\_experiment\_name(prefix}\OperatorTok{=}\NormalTok{PREFIX)}
\BuiltInTok{print}\NormalTok{(experiment\_name)}

\NormalTok{fun\_control }\OperatorTok{=}\NormalTok{ fun\_control\_init(}
\NormalTok{    spot\_tensorboard\_path}\OperatorTok{=}\NormalTok{get\_spot\_tensorboard\_path(experiment\_name))}
\end{Highlighting}
\end{Shaded}

\begin{verbatim}
01_bartz09_2023-07-17_18-00-57
\end{verbatim}

Since the \texttt{spot\_tensorboard\_path} is defined,
\texttt{spotPython} will log the optimization process in the TensorBoard
files. The TensorBoard files are stored in the directory
\texttt{spot\_tensorboard\_path}. We can pass the TensorBoard
information to the \texttt{Spot} method via the \texttt{fun\_control}
dictionary.

\begin{Shaded}
\begin{Highlighting}[]
\NormalTok{spot\_tuner }\OperatorTok{=}\NormalTok{ spot.Spot(fun}\OperatorTok{=}\NormalTok{fun,}
\NormalTok{                   lower }\OperatorTok{=}\NormalTok{ np.array([}\OperatorTok{{-}}\DecValTok{1}\NormalTok{]),}
\NormalTok{                   upper }\OperatorTok{=}\NormalTok{ np.array([}\DecValTok{2}\NormalTok{]),}
\NormalTok{                   fun\_evals}\OperatorTok{=} \DecValTok{10}\NormalTok{,}
\NormalTok{                   seed}\OperatorTok{=}\DecValTok{123}\NormalTok{,}
\NormalTok{                   show\_models}\OperatorTok{=}\VariableTok{False}\NormalTok{,}
\NormalTok{                   design\_control}\OperatorTok{=}\NormalTok{\{}\StringTok{"init\_size"}\NormalTok{: }\DecValTok{5}\NormalTok{\},}
\NormalTok{                   fun\_control}\OperatorTok{=}\NormalTok{fun\_control,)}
\NormalTok{spot\_tuner.run()}
\end{Highlighting}
\end{Shaded}

\begin{verbatim}
spotPython tuning: 2.7705924100183687e-05 [######----] 60.00% 
\end{verbatim}

\begin{verbatim}
spotPython tuning: 7.364661789374228e-07 [#######---] 70.00% 
\end{verbatim}

\begin{verbatim}
spotPython tuning: 7.364661789374228e-07 [########--] 80.00% 
\end{verbatim}

\begin{verbatim}
spotPython tuning: 3.5490065465299805e-07 [#########-] 90.00% 
\end{verbatim}

\begin{verbatim}
spotPython tuning: 7.234315072455918e-09 [##########] 100.00% Done...
\end{verbatim}

\begin{verbatim}
<spotPython.spot.spot.Spot at 0x2c439be50>
\end{verbatim}

Now we can start TensorBoard in the background. The TensorBoard process
will read the TensorBoard files and visualize the hyperparameter tuning
process. From the terminal, we can start TensorBoard with the following
command:

\begin{Shaded}
\begin{Highlighting}[]
\NormalTok{tensorboard {-}{-}logdir="./runs"}
\end{Highlighting}
\end{Shaded}

\texttt{logdir} is the directory where the TensorBoard files are stored.
In our case, the TensorBoard files are stored in the directory
\texttt{./runs}.

TensorBoard will start a web server on port 6006. We can access the
TensorBoard web server with the following URL:

\begin{Shaded}
\begin{Highlighting}[]
\NormalTok{http://localhost:6006/}
\end{Highlighting}
\end{Shaded}

The first TensorBoard visualization shows the objective function values
plotted against the wall time. The wall time is the time that has passed
since the start of the hyperparameter tuning process. The five initial
design points are shown in the upper left region of the plot. The line
visualizes the optimization process.
\includegraphics{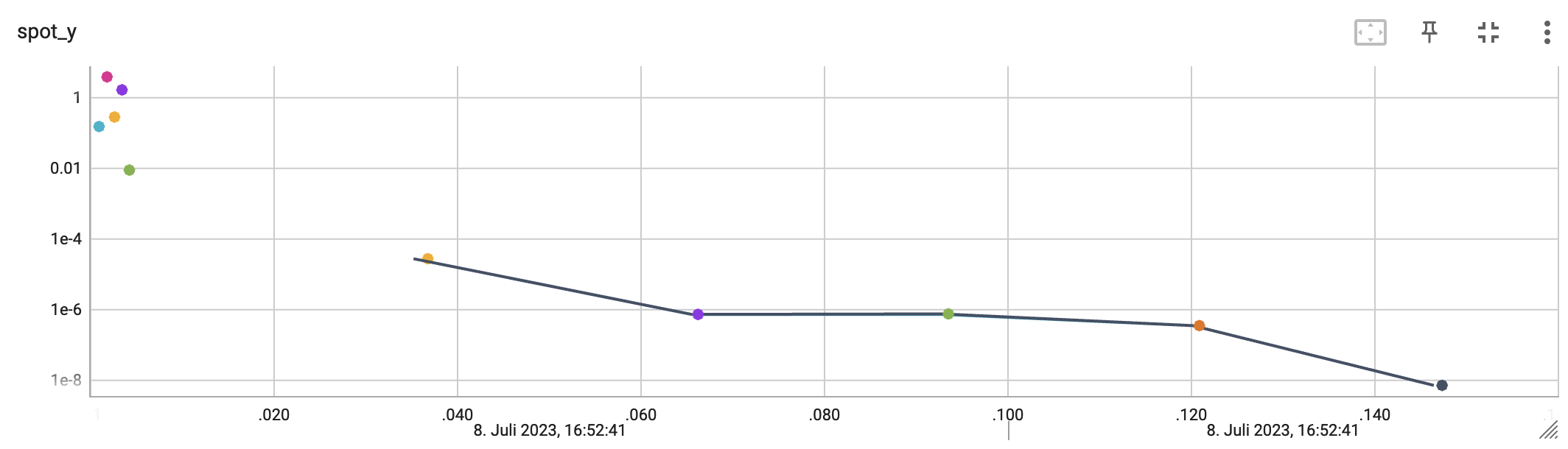}

The second TensorBoard visualization shows the input values, i.e.,
\(x_0\), plotted against the wall time.
\includegraphics{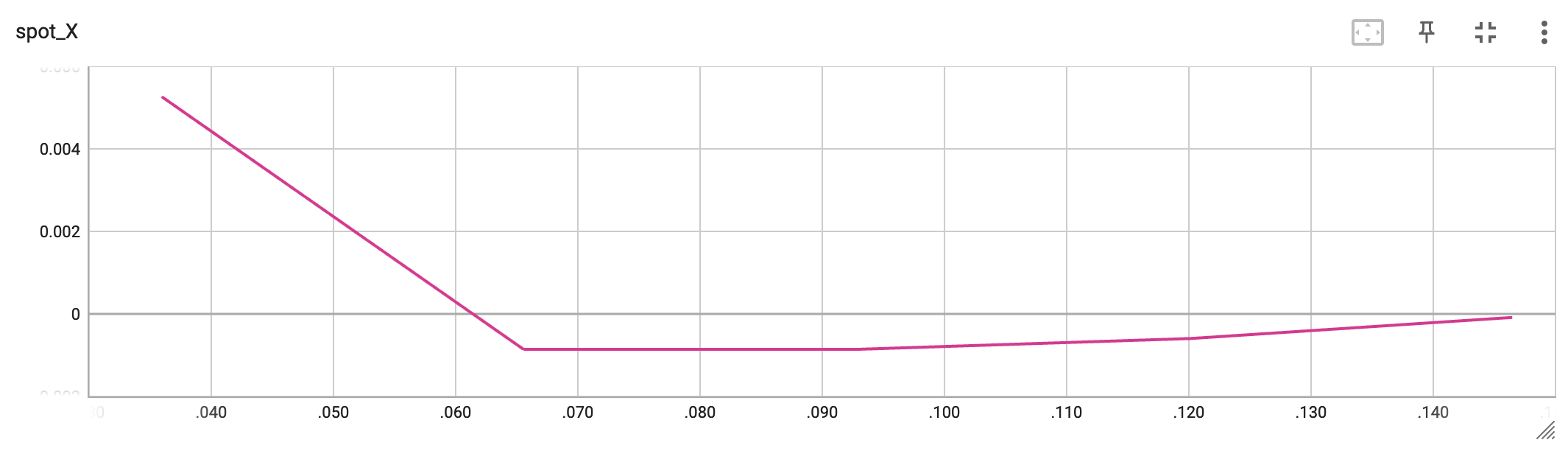}

The third TensorBoard plot illustrates how \texttt{spotPython} can be
used as a microscope for the internal mechanisms of the surrogate-based
optimization process. Here, one important parameter, the learning rate
\(\theta\) of the Kriging surrogate is plotted against the number of
optimization steps.

\begin{figure}

{\centering \includegraphics[width=0.5\textwidth,height=\textheight]{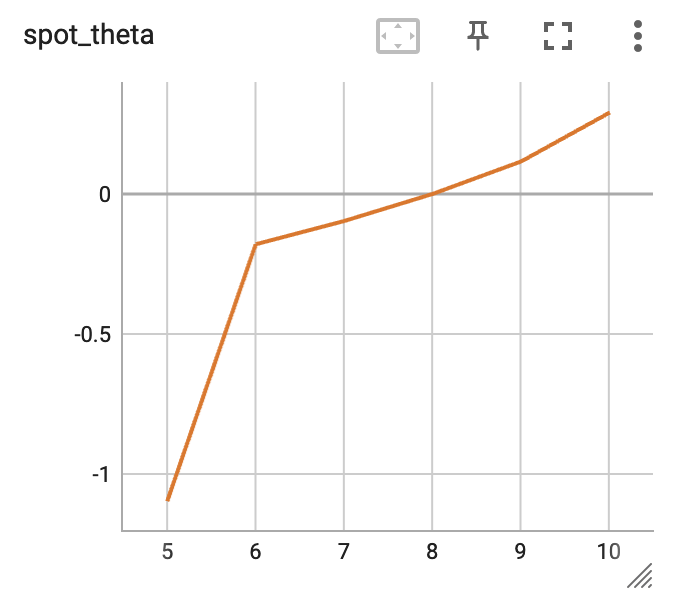}

}

\caption{TensorBoard visualization of the spotPython process.}

\end{figure}

\hypertarget{sec-multi-dim}{%
\chapter{Multi-dimensional Functions}\label{sec-multi-dim}}

This chapter illustrates how high-dimensional functions can be optimzed
and analyzed.

\hypertarget{example-spot-and-the-3-dim-sphere-function}{%
\section{\texorpdfstring{Example: \texttt{Spot} and the 3-dim Sphere
Function}{Example: Spot and the 3-dim Sphere Function}}\label{example-spot-and-the-3-dim-sphere-function}}

\begin{Shaded}
\begin{Highlighting}[]
\ImportTok{import}\NormalTok{ numpy }\ImportTok{as}\NormalTok{ np}
\ImportTok{from}\NormalTok{ spotPython.fun.objectivefunctions }\ImportTok{import}\NormalTok{ analytical}
\ImportTok{from}\NormalTok{ spotPython.spot }\ImportTok{import}\NormalTok{ spot}
\end{Highlighting}
\end{Shaded}

\hypertarget{the-objective-function-3-dim-sphere}{%
\subsection{The Objective Function: 3-dim
Sphere}\label{the-objective-function-3-dim-sphere}}

\begin{itemize}
\item
  The \texttt{spotPython} package provides several classes of objective
  functions.
\item
  We will use an analytical objective function, i.e., a function that
  can be described by a (closed) formula: \[f(x) = \sum_i^n x_i^2 \]
\item
  Here we will use \(n=3\).
\end{itemize}

\begin{Shaded}
\begin{Highlighting}[]
\NormalTok{fun }\OperatorTok{=}\NormalTok{ analytical().fun\_sphere}
\end{Highlighting}
\end{Shaded}

\begin{itemize}
\item
  The size of the \texttt{lower} bound vector determines the problem
  dimension.
\item
  Here we will use \texttt{-1.0\ *\ np.ones(3)}, i.e., a three-dim
  function.
\item
  We will use three different \texttt{theta} values (one for each
  dimension), i.e., we set

  \texttt{surrogate\_control=\{"n\_theta":\ 3\}}.
\end{itemize}

\begin{tcolorbox}[enhanced jigsaw, left=2mm, title=\textcolor{quarto-callout-note-color}{\faInfo}\hspace{0.5em}{TensorBoard}, bottomrule=.15mm, titlerule=0mm, breakable, rightrule=.15mm, toprule=.15mm, coltitle=black, colbacktitle=quarto-callout-note-color!10!white, leftrule=.75mm, arc=.35mm, colframe=quarto-callout-note-color-frame, bottomtitle=1mm, colback=white, opacitybacktitle=0.6, toptitle=1mm, opacityback=0]

Similar to the one-dimensional case, which was introduced in Section
Section~\ref{sec-visualizing-tensorboard-01}, we can use TensorBoard to
monitor the progress of the optimization. We will use the same code,
only the prefix is different:

\begin{Shaded}
\begin{Highlighting}[]
\ImportTok{from}\NormalTok{ spotPython.utils.}\BuiltInTok{file} \ImportTok{import}\NormalTok{ get\_experiment\_name}
\ImportTok{from}\NormalTok{ spotPython.utils.init }\ImportTok{import}\NormalTok{ fun\_control\_init}
\ImportTok{from}\NormalTok{ spotPython.utils.}\BuiltInTok{file} \ImportTok{import}\NormalTok{ get\_spot\_tensorboard\_path}

\NormalTok{PREFIX }\OperatorTok{=} \StringTok{"02"}
\NormalTok{experiment\_name }\OperatorTok{=}\NormalTok{ get\_experiment\_name(prefix}\OperatorTok{=}\NormalTok{PREFIX)}
\BuiltInTok{print}\NormalTok{(experiment\_name)}

\NormalTok{fun\_control }\OperatorTok{=}\NormalTok{ fun\_control\_init(}
\NormalTok{    spot\_tensorboard\_path}\OperatorTok{=}\NormalTok{get\_spot\_tensorboard\_path(experiment\_name))}
\end{Highlighting}
\end{Shaded}

\begin{verbatim}
02_bartz09_2023-07-17_18-02-37
\end{verbatim}

\end{tcolorbox}

\begin{Shaded}
\begin{Highlighting}[]
\NormalTok{spot\_3 }\OperatorTok{=}\NormalTok{ spot.Spot(fun}\OperatorTok{=}\NormalTok{fun,}
\NormalTok{                   lower }\OperatorTok{=} \OperatorTok{{-}}\FloatTok{1.0}\OperatorTok{*}\NormalTok{np.ones(}\DecValTok{3}\NormalTok{),}
\NormalTok{                   upper }\OperatorTok{=}\NormalTok{ np.ones(}\DecValTok{3}\NormalTok{),}
\NormalTok{                   var\_name}\OperatorTok{=}\NormalTok{[}\StringTok{"Pressure"}\NormalTok{, }\StringTok{"Temp"}\NormalTok{, }\StringTok{"Lambda"}\NormalTok{],}
\NormalTok{                   show\_progress}\OperatorTok{=}\VariableTok{True}\NormalTok{,}
\NormalTok{                   surrogate\_control}\OperatorTok{=}\NormalTok{\{}\StringTok{"n\_theta"}\NormalTok{: }\DecValTok{3}\NormalTok{\},}
\NormalTok{                   fun\_control}\OperatorTok{=}\NormalTok{fun\_control,)}

\NormalTok{spot\_3.run()}
\end{Highlighting}
\end{Shaded}

\begin{verbatim}
spotPython tuning: 0.03443344056467332 [#######---] 73.33% 
\end{verbatim}

\begin{verbatim}
spotPython tuning: 0.03134865993507926 [########--] 80.00% 
\end{verbatim}

\begin{verbatim}
spotPython tuning: 0.0009629342967936851 [#########-] 86.67% 
\end{verbatim}

\begin{verbatim}
spotPython tuning: 8.541951463966474e-05 [#########-] 93.33% 
\end{verbatim}

\begin{verbatim}
spotPython tuning: 6.285135731399678e-05 [##########] 100.00% Done...
\end{verbatim}

\begin{verbatim}
<spotPython.spot.spot.Spot at 0x28319da50>
\end{verbatim}

Now we can start TensorBoard in the background with the following
command:

\begin{Shaded}
\begin{Highlighting}[]
\NormalTok{tensorboard {-}{-}logdir="./runs"}
\end{Highlighting}
\end{Shaded}

We can access the TensorBoard web server with the following URL:

\begin{Shaded}
\begin{Highlighting}[]
\NormalTok{http://localhost:6006/}
\end{Highlighting}
\end{Shaded}

\hypertarget{results}{%
\subsection{Results}\label{results}}

\begin{Shaded}
\begin{Highlighting}[]
\NormalTok{spot\_3.print\_results()}
\end{Highlighting}
\end{Shaded}

\begin{verbatim}
min y: 6.285135731399678e-05
Pressure: 0.005236109709736696
Temp: 0.0019572552655686714
Lambda: 0.005621713639718905
\end{verbatim}

\begin{verbatim}
[['Pressure', 0.005236109709736696],
 ['Temp', 0.0019572552655686714],
 ['Lambda', 0.005621713639718905]]
\end{verbatim}

\begin{Shaded}
\begin{Highlighting}[]
\NormalTok{spot\_3.plot\_progress()}
\end{Highlighting}
\end{Shaded}

\begin{figure}[H]

{\centering \includegraphics{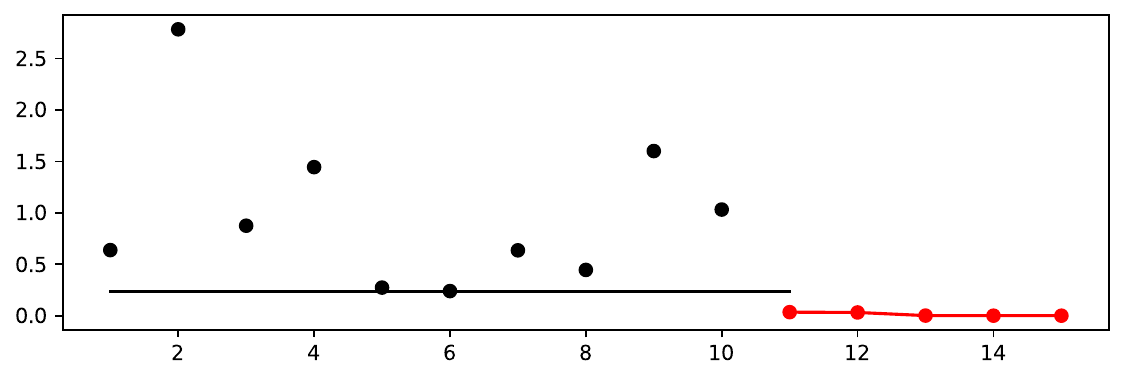}

}

\end{figure}

\hypertarget{a-contour-plot}{%
\subsection{A Contour Plot}\label{a-contour-plot}}

\begin{itemize}
\tightlist
\item
  We can select two dimensions, say \(i=0\) and \(j=1\), and generate a
  contour plot as follows.

  \begin{itemize}
  \tightlist
  \item
    Note: We have specified identical \texttt{min\_z} and
    \texttt{max\_z} values to generate comparable plots!
  \end{itemize}
\end{itemize}

\begin{Shaded}
\begin{Highlighting}[]
\NormalTok{spot\_3.plot\_contour(i}\OperatorTok{=}\DecValTok{0}\NormalTok{, j}\OperatorTok{=}\DecValTok{1}\NormalTok{, min\_z}\OperatorTok{=}\DecValTok{0}\NormalTok{, max\_z}\OperatorTok{=}\FloatTok{2.25}\NormalTok{)}
\end{Highlighting}
\end{Shaded}

\begin{figure}[H]

{\centering \includegraphics{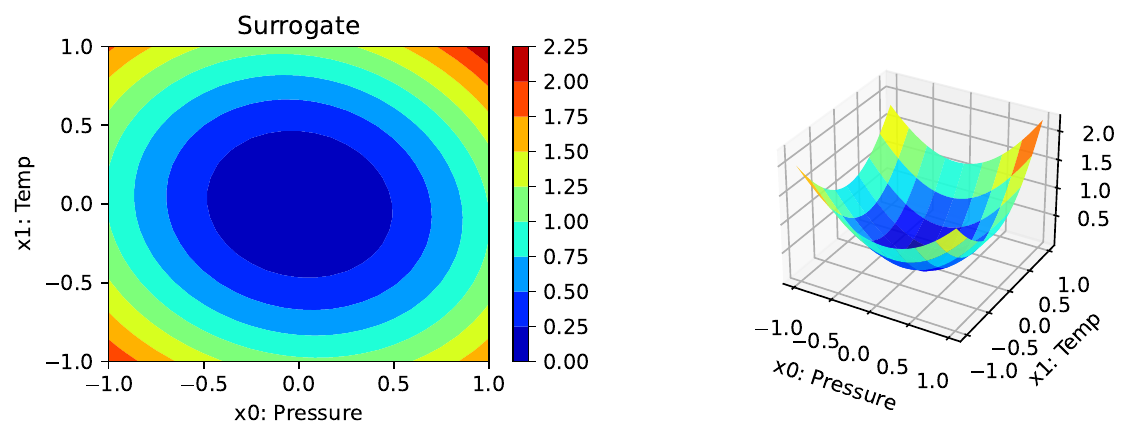}

}

\end{figure}

\begin{itemize}
\tightlist
\item
  In a similar manner, we can plot dimension \(i=0\) and \(j=2\):
\end{itemize}

\begin{Shaded}
\begin{Highlighting}[]
\NormalTok{spot\_3.plot\_contour(i}\OperatorTok{=}\DecValTok{0}\NormalTok{, j}\OperatorTok{=}\DecValTok{2}\NormalTok{, min\_z}\OperatorTok{=}\DecValTok{0}\NormalTok{, max\_z}\OperatorTok{=}\FloatTok{2.25}\NormalTok{)}
\end{Highlighting}
\end{Shaded}

\begin{figure}[H]

{\centering \includegraphics{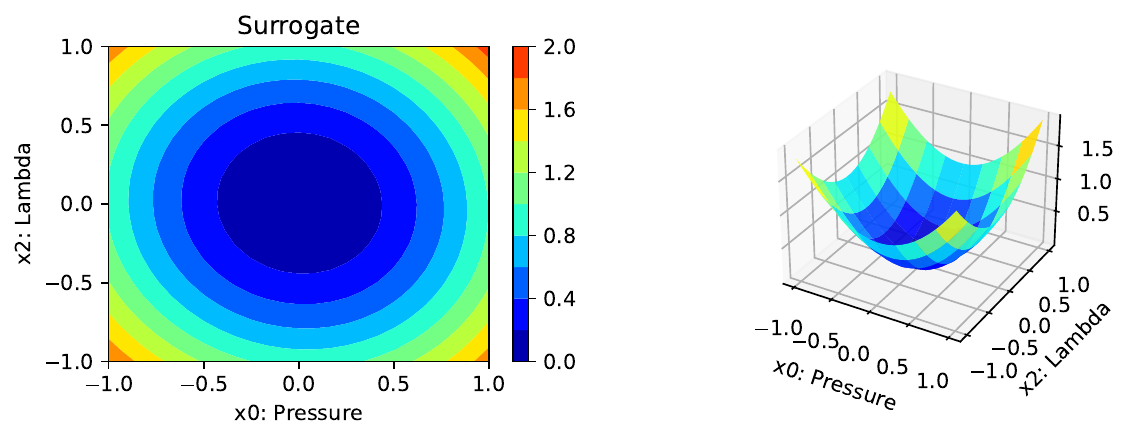}

}

\end{figure}

\begin{itemize}
\tightlist
\item
  The final combination is \(i=1\) and \(j=2\):
\end{itemize}

\begin{Shaded}
\begin{Highlighting}[]
\NormalTok{spot\_3.plot\_contour(i}\OperatorTok{=}\DecValTok{1}\NormalTok{, j}\OperatorTok{=}\DecValTok{2}\NormalTok{, min\_z}\OperatorTok{=}\DecValTok{0}\NormalTok{, max\_z}\OperatorTok{=}\FloatTok{2.25}\NormalTok{)}
\end{Highlighting}
\end{Shaded}

\begin{figure}[H]

{\centering \includegraphics{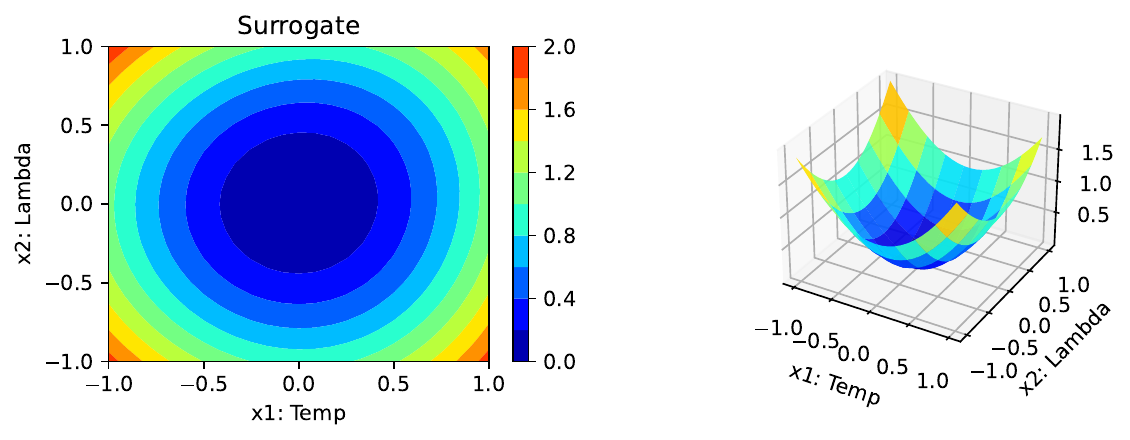}

}

\end{figure}

\begin{itemize}
\tightlist
\item
  The three plots look very similar, because the \texttt{fun\_sphere} is
  symmetric.
\item
  This can also be seen from the variable importance:
\end{itemize}

\begin{Shaded}
\begin{Highlighting}[]
\NormalTok{spot\_3.print\_importance()}
\end{Highlighting}
\end{Shaded}

\begin{verbatim}
Pressure:  99.35185545837122
Temp:  99.99999999999999
Lambda:  94.31627052007231
\end{verbatim}

\begin{verbatim}
[['Pressure', 99.35185545837122],
 ['Temp', 99.99999999999999],
 ['Lambda', 94.31627052007231]]
\end{verbatim}

\hypertarget{tensorboard-1}{%
\subsection{TensorBoard}\label{tensorboard-1}}

\begin{figure}

{\centering \includegraphics{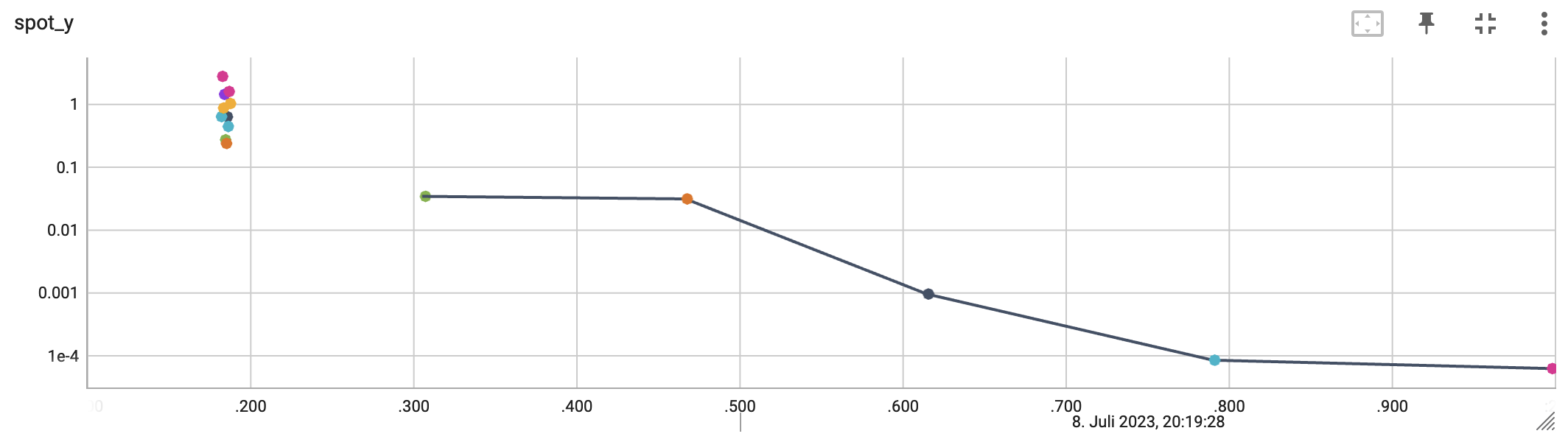}

}

\caption{TensorBoard visualization of the spotPython process. Objective
function values plotted against wall time.}

\end{figure}

The second TensorBoard visualization shows the input values, i.e.,
\(x_0, \ldots, x_2\), plotted against the wall time.
\includegraphics{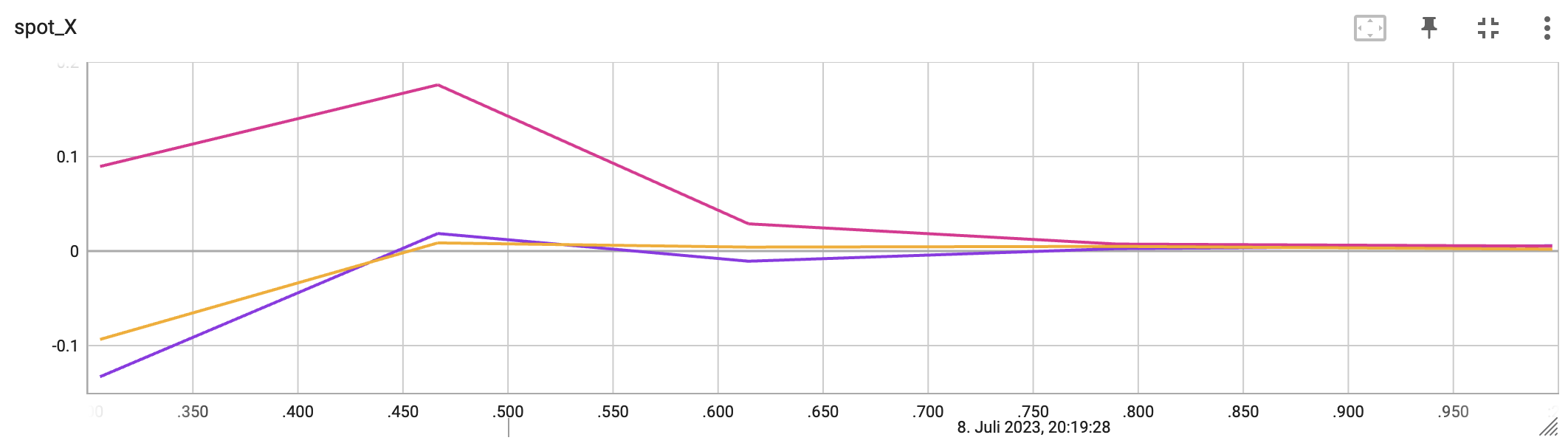}

The third TensorBoard plot illustrates how \texttt{spotPython} can be
used as a microscope for the internal mechanisms of the surrogate-based
optimization process. Here, one important parameter, the learning rate
\(\theta\) of the Kriging surrogate is plotted against the number of
optimization steps.

\begin{figure}

{\centering \includegraphics[width=1\textwidth,height=\textheight]{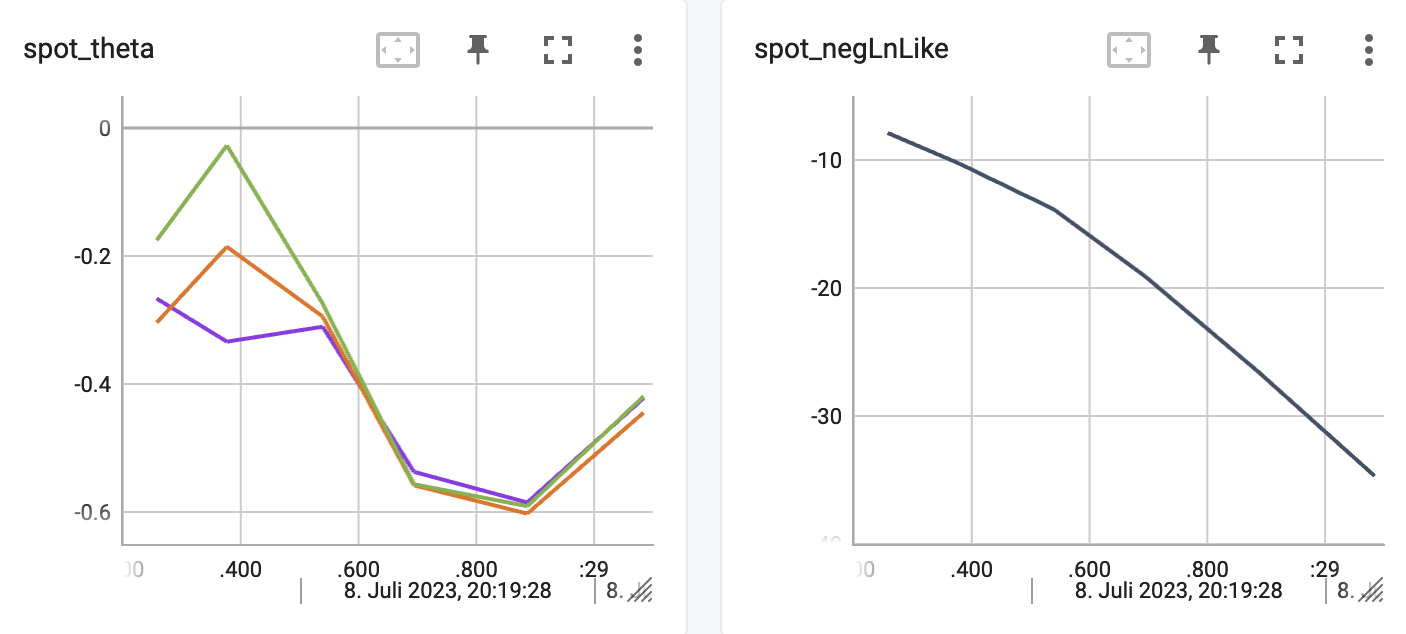}

}

\caption{TensorBoard visualization of the spotPython surrogate model.}

\end{figure}

\hypertarget{conclusion}{%
\section{Conclusion}\label{conclusion}}

Based on this quick analysis, we can conclude that all three dimensions
are equally important (as expected, because the analytical function is
known).

\hypertarget{exercises}{%
\section{Exercises}\label{exercises}}

\begin{itemize}
\tightlist
\item
  Important:

  \begin{itemize}
  \tightlist
  \item
    Results from these exercises should be added to this document, i.e.,
    you should submit an updated version of this notebook.
  \item
    Please combine your results using this notebook.
  \item
    Only one notebook from each group!
  \item
    Presentation is based on this notebook. No addtional slides are
    required!
  \item
    spotPython version \texttt{0.16.11} (or greater) is required
  \end{itemize}
\end{itemize}

\hypertarget{the-three-dimensional-fun_cubed}{%
\subsection{\texorpdfstring{The Three Dimensional
\texttt{fun\_cubed}}{The Three Dimensional fun\_cubed}}\label{the-three-dimensional-fun_cubed}}

\begin{itemize}
\tightlist
\item
  The input dimension is \texttt{3}. The search range is
  \(-1 \leq x \leq 1\) for all dimensions.
\item
  Generate contour plots
\item
  Calculate the variable importance.
\item
  Discuss the variable importance:

  \begin{itemize}
  \tightlist
  \item
    Are all variables equally important?
  \item
    If not:

    \begin{itemize}
    \tightlist
    \item
      Which is the most important variable?
    \item
      Which is the least important variable?
    \end{itemize}
  \end{itemize}
\end{itemize}

\hypertarget{the-ten-dimensional-fun_wing_wt}{%
\subsection{\texorpdfstring{The Ten Dimensional
\texttt{fun\_wing\_wt}}{The Ten Dimensional fun\_wing\_wt}}\label{the-ten-dimensional-fun_wing_wt}}

\begin{itemize}
\tightlist
\item
  The input dimension is \texttt{10}. The search range is
  \(0 \leq x \leq 1\) for all dimensions.
\item
  Calculate the variable importance.
\item
  Discuss the variable importance:

  \begin{itemize}
  \tightlist
  \item
    Are all variables equally important?
  \item
    If not:

    \begin{itemize}
    \tightlist
    \item
      Which is the most important variable?
    \item
      Which is the least important variable?
    \end{itemize}
  \item
    Generate contour plots for the three most important variables. Do
    they confirm your selection?
  \end{itemize}
\end{itemize}

\hypertarget{the-three-dimensional-fun_runge}{%
\subsection{\texorpdfstring{The Three Dimensional
\texttt{fun\_runge}}{The Three Dimensional fun\_runge}}\label{the-three-dimensional-fun_runge}}

\begin{itemize}
\tightlist
\item
  The input dimension is \texttt{3}. The search range is
  \(-5 \leq x \leq 5\) for all dimensions.
\item
  Generate contour plots
\item
  Calculate the variable importance.
\item
  Discuss the variable importance:

  \begin{itemize}
  \tightlist
  \item
    Are all variables equally important?
  \item
    If not:

    \begin{itemize}
    \tightlist
    \item
      Which is the most important variable?
    \item
      Which is the least important variable?
    \end{itemize}
  \end{itemize}
\end{itemize}

\hypertarget{the-three-dimensional-fun_linear}{%
\subsection{\texorpdfstring{The Three Dimensional
\texttt{fun\_linear}}{The Three Dimensional fun\_linear}}\label{the-three-dimensional-fun_linear}}

\begin{itemize}
\tightlist
\item
  The input dimension is \texttt{3}. The search range is
  \(-5 \leq x \leq 5\) for all dimensions.
\item
  Generate contour plots
\item
  Calculate the variable importance.
\item
  Discuss the variable importance:

  \begin{itemize}
  \tightlist
  \item
    Are all variables equally important?
  \item
    If not:

    \begin{itemize}
    \tightlist
    \item
      Which is the most important variable?
    \item
      Which is the least important variable?
    \end{itemize}
  \end{itemize}
\end{itemize}

\hypertarget{sec-iso-aniso-kriging}{%
\chapter{Isotropic and Anisotropic
Kriging}\label{sec-iso-aniso-kriging}}

This chapter illustrates the difference between isotropic and
anisotropic Kriging models. The difference is illustrated with the help
of the \texttt{spotPython} package. Isotropic Kriging models use the
same \texttt{theta} value for every dimension. Anisotropic Kriging
models use different \texttt{theta} values for each dimension.

\hypertarget{example-isotropic-spot-surrogate-and-the-2-dim-sphere-function}{%
\section{\texorpdfstring{Example: Isotropic \texttt{Spot} Surrogate and
the 2-dim Sphere
Function}{Example: Isotropic Spot Surrogate and the 2-dim Sphere Function}}\label{example-isotropic-spot-surrogate-and-the-2-dim-sphere-function}}

\begin{Shaded}
\begin{Highlighting}[]
\ImportTok{import}\NormalTok{ numpy }\ImportTok{as}\NormalTok{ np}
\ImportTok{from}\NormalTok{ math }\ImportTok{import}\NormalTok{ inf}
\ImportTok{from}\NormalTok{ spotPython.fun.objectivefunctions }\ImportTok{import}\NormalTok{ analytical}
\ImportTok{from}\NormalTok{ spotPython.spot }\ImportTok{import}\NormalTok{ spot}
\end{Highlighting}
\end{Shaded}

\hypertarget{the-objective-function-2-dim-sphere}{%
\subsection{The Objective Function: 2-dim
Sphere}\label{the-objective-function-2-dim-sphere}}

\begin{itemize}
\tightlist
\item
  The \texttt{spotPython} package provides several classes of objective
  functions.
\item
  We will use an analytical objective function, i.e., a function that
  can be described by a (closed) formula: \[f(x, y) = x^2 + y^2\]
\end{itemize}

\begin{Shaded}
\begin{Highlighting}[]
\NormalTok{fun }\OperatorTok{=}\NormalTok{ analytical().fun\_sphere}
\NormalTok{fun\_control }\OperatorTok{=}\NormalTok{ \{}\StringTok{"sigma"}\NormalTok{: }\DecValTok{0}\NormalTok{,}
               \StringTok{"seed"}\NormalTok{: }\DecValTok{123}\NormalTok{\}}
\end{Highlighting}
\end{Shaded}

\begin{itemize}
\tightlist
\item
  The size of the \texttt{lower} bound vector determines the problem
  dimension.
\item
  Here we will use \texttt{np.array({[}-1,\ -1{]})}, i.e., a two-dim
  function.
\end{itemize}

\begin{Shaded}
\begin{Highlighting}[]
\NormalTok{spot\_2 }\OperatorTok{=}\NormalTok{ spot.Spot(fun}\OperatorTok{=}\NormalTok{fun,}
\NormalTok{                   lower }\OperatorTok{=}\NormalTok{ np.array([}\OperatorTok{{-}}\DecValTok{1}\NormalTok{, }\OperatorTok{{-}}\DecValTok{1}\NormalTok{]),}
\NormalTok{                   upper }\OperatorTok{=}\NormalTok{ np.array([}\DecValTok{1}\NormalTok{, }\DecValTok{1}\NormalTok{]))}

\NormalTok{spot\_2.run()}
\end{Highlighting}
\end{Shaded}

\begin{verbatim}
spotPython tuning: 2.093282610941807e-05 [#######---] 73.33% 
\end{verbatim}

\begin{verbatim}
spotPython tuning: 2.093282610941807e-05 [########--] 80.00% 
\end{verbatim}

\begin{verbatim}
spotPython tuning: 2.093282610941807e-05 [#########-] 86.67% 
\end{verbatim}

\begin{verbatim}
spotPython tuning: 2.093282610941807e-05 [#########-] 93.33% 
\end{verbatim}

\begin{verbatim}
spotPython tuning: 2.093282610941807e-05 [##########] 100.00% Done...
\end{verbatim}

\begin{verbatim}
<spotPython.spot.spot.Spot at 0x14eb299c0>
\end{verbatim}

\hypertarget{results-1}{%
\subsection{Results}\label{results-1}}

\begin{Shaded}
\begin{Highlighting}[]
\NormalTok{spot\_2.print\_results()}
\end{Highlighting}
\end{Shaded}

\begin{verbatim}
min y: 2.093282610941807e-05
x0: 0.0016055267473267492
x1: 0.00428428640184529
\end{verbatim}

\begin{verbatim}
[['x0', 0.0016055267473267492], ['x1', 0.00428428640184529]]
\end{verbatim}

\begin{Shaded}
\begin{Highlighting}[]
\NormalTok{spot\_2.plot\_progress(log\_y}\OperatorTok{=}\VariableTok{True}\NormalTok{)}
\end{Highlighting}
\end{Shaded}

\begin{figure}[H]

{\centering \includegraphics{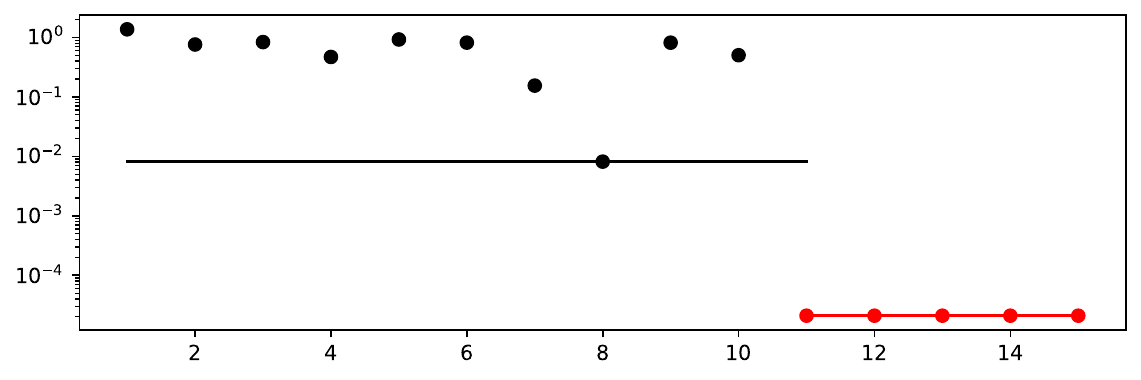}

}

\end{figure}

\hypertarget{example-with-anisotropic-kriging}{%
\section{Example With Anisotropic
Kriging}\label{example-with-anisotropic-kriging}}

\begin{itemize}
\tightlist
\item
  The default parameter setting of \texttt{spotPython}'s Kriging
  surrogate uses the same \texttt{theta} value for every dimension.
\item
  This is referred to as ``using an isotropic kernel''.
\item
  If different \texttt{theta} values are used for each dimension, then
  an anisotropic kernel is used
\item
  To enable anisotropic models in \texttt{spotPython}, the number of
  \texttt{theta} values should be larger than one.
\item
  We can use \texttt{surrogate\_control=\{"n\_theta":\ 2\}} to enable
  this behavior (\texttt{2} is the problem dimension).
\end{itemize}

\begin{tcolorbox}[enhanced jigsaw, left=2mm, title=\textcolor{quarto-callout-note-color}{\faInfo}\hspace{0.5em}{TensorBoard}, bottomrule=.15mm, titlerule=0mm, breakable, rightrule=.15mm, toprule=.15mm, coltitle=black, colbacktitle=quarto-callout-note-color!10!white, leftrule=.75mm, arc=.35mm, colframe=quarto-callout-note-color-frame, bottomtitle=1mm, colback=white, opacitybacktitle=0.6, toptitle=1mm, opacityback=0]

Similar to the one-dimensional case, which was introduced in Section
Section~\ref{sec-visualizing-tensorboard-01}, we can use TensorBoard to
monitor the progress of the optimization. We will use the same code,
only the prefix is different:

\begin{Shaded}
\begin{Highlighting}[]
\ImportTok{from}\NormalTok{ spotPython.utils.}\BuiltInTok{file} \ImportTok{import}\NormalTok{ get\_experiment\_name}
\ImportTok{from}\NormalTok{ spotPython.utils.init }\ImportTok{import}\NormalTok{ fun\_control\_init}
\ImportTok{from}\NormalTok{ spotPython.utils.}\BuiltInTok{file} \ImportTok{import}\NormalTok{ get\_spot\_tensorboard\_path}

\NormalTok{PREFIX }\OperatorTok{=} \StringTok{"03"}
\NormalTok{experiment\_name }\OperatorTok{=}\NormalTok{ get\_experiment\_name(prefix}\OperatorTok{=}\NormalTok{PREFIX)}
\BuiltInTok{print}\NormalTok{(experiment\_name)}

\NormalTok{fun\_control }\OperatorTok{=}\NormalTok{ fun\_control\_init(}
\NormalTok{    spot\_tensorboard\_path}\OperatorTok{=}\NormalTok{get\_spot\_tensorboard\_path(experiment\_name))}
\end{Highlighting}
\end{Shaded}

\begin{verbatim}
03_bartz09_2023-07-17_08-48-59
\end{verbatim}

\end{tcolorbox}

\begin{Shaded}
\begin{Highlighting}[]
\NormalTok{spot\_2\_anisotropic }\OperatorTok{=}\NormalTok{ spot.Spot(fun}\OperatorTok{=}\NormalTok{fun,}
\NormalTok{                   lower }\OperatorTok{=}\NormalTok{ np.array([}\OperatorTok{{-}}\DecValTok{1}\NormalTok{, }\OperatorTok{{-}}\DecValTok{1}\NormalTok{]),}
\NormalTok{                   upper }\OperatorTok{=}\NormalTok{ np.array([}\DecValTok{1}\NormalTok{, }\DecValTok{1}\NormalTok{]),}
\NormalTok{                   surrogate\_control}\OperatorTok{=}\NormalTok{\{}\StringTok{"n\_theta"}\NormalTok{: }\DecValTok{2}\NormalTok{\},}
\NormalTok{                   fun\_control}\OperatorTok{=}\NormalTok{fun\_control)}
\NormalTok{spot\_2\_anisotropic.run()}
\end{Highlighting}
\end{Shaded}

\begin{verbatim}
spotPython tuning: 1.991152317760403e-05 [#######---] 73.33% 
\end{verbatim}

\begin{verbatim}
spotPython tuning: 1.991152317760403e-05 [########--] 80.00% 
\end{verbatim}

\begin{verbatim}
spotPython tuning: 1.991152317760403e-05 [#########-] 86.67% 
\end{verbatim}

\begin{verbatim}
spotPython tuning: 1.991152317760403e-05 [#########-] 93.33% 
\end{verbatim}

\begin{verbatim}
spotPython tuning: 7.77061191821505e-06 [##########] 100.00% Done...
\end{verbatim}

\begin{verbatim}
<spotPython.spot.spot.Spot at 0x2a749fa60>
\end{verbatim}

\begin{itemize}
\tightlist
\item
  The search progress of the optimization with the anisotropic model can
  be visualized:
\end{itemize}

\begin{Shaded}
\begin{Highlighting}[]
\NormalTok{spot\_2\_anisotropic.plot\_progress(log\_y}\OperatorTok{=}\VariableTok{True}\NormalTok{)}
\end{Highlighting}
\end{Shaded}

\begin{figure}[H]

{\centering \includegraphics{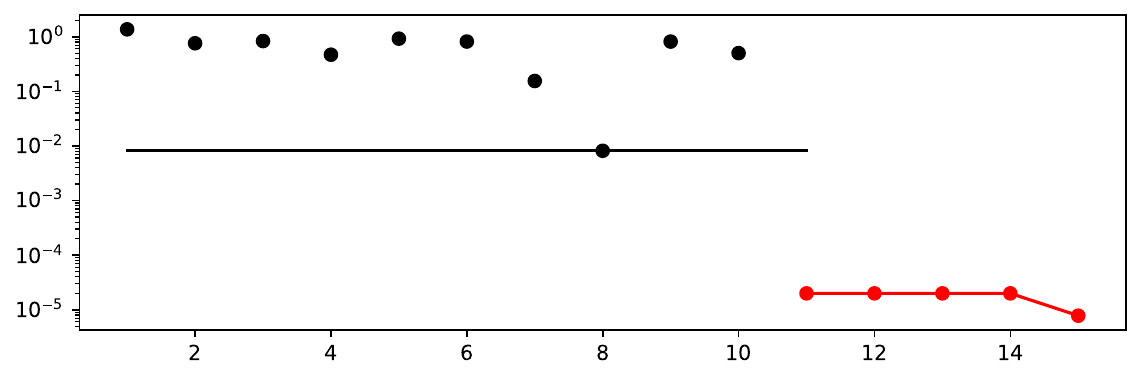}

}

\end{figure}

\begin{Shaded}
\begin{Highlighting}[]
\NormalTok{spot\_2\_anisotropic.print\_results()}
\end{Highlighting}
\end{Shaded}

\begin{verbatim}
min y: 7.77061191821505e-06
x0: -0.0024488252797500764
x1: -0.0013318658594137815
\end{verbatim}

\begin{verbatim}
[['x0', -0.0024488252797500764], ['x1', -0.0013318658594137815]]
\end{verbatim}

\begin{Shaded}
\begin{Highlighting}[]
\NormalTok{spot\_2\_anisotropic.surrogate.plot()}
\end{Highlighting}
\end{Shaded}

\begin{figure}[H]

{\centering \includegraphics{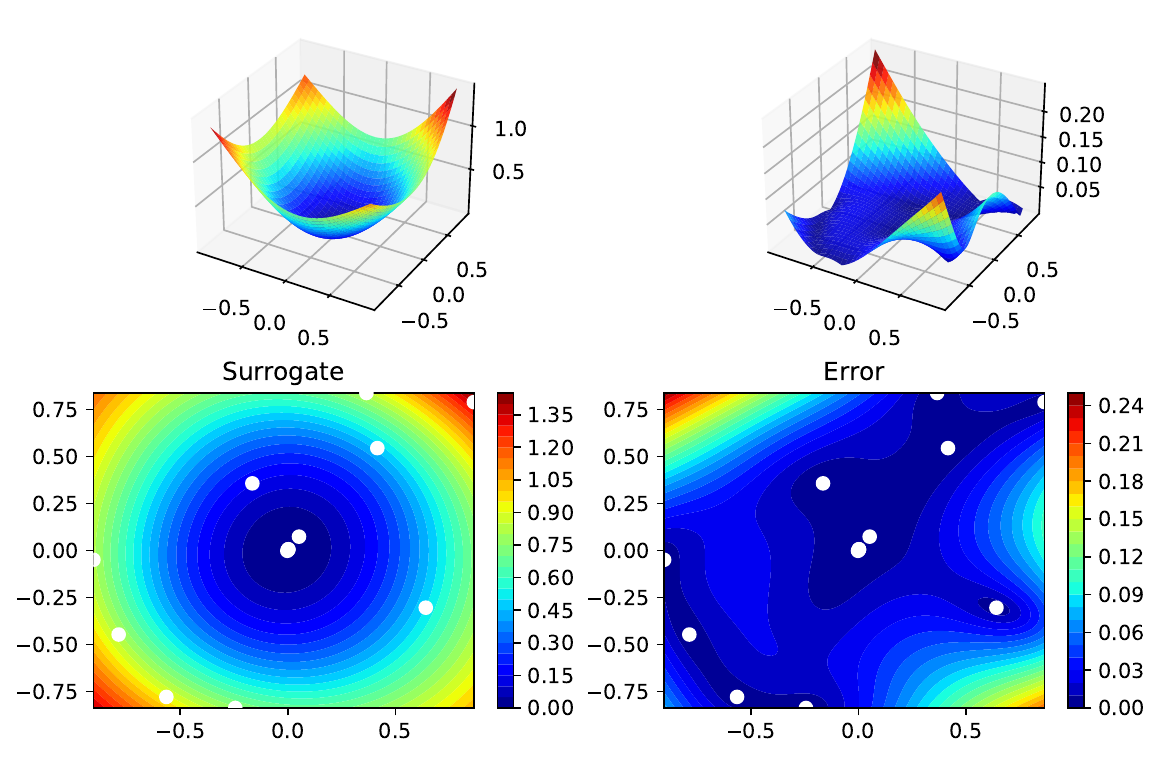}

}

\end{figure}

\hypertarget{taking-a-look-at-the-theta-values}{%
\subsection{\texorpdfstring{Taking a Look at the \texttt{theta}
Values}{Taking a Look at the theta Values}}\label{taking-a-look-at-the-theta-values}}

\hypertarget{theta-values-from-the-spot-model}{%
\subsubsection{\texorpdfstring{\texttt{theta} Values from the
\texttt{spot}
Model}{theta Values from the spot Model}}\label{theta-values-from-the-spot-model}}

\begin{itemize}
\tightlist
\item
  We can check, whether one or several \texttt{theta} values were used.
\item
  The \texttt{theta} values from the surrogate can be printed as
  follows:
\end{itemize}

\begin{Shaded}
\begin{Highlighting}[]
\NormalTok{spot\_2\_anisotropic.surrogate.theta}
\end{Highlighting}
\end{Shaded}

\begin{verbatim}
array([0.19447342, 0.30813872])
\end{verbatim}

\begin{itemize}
\tightlist
\item
  Since the surrogate from the isotropic setting was stored as
  \texttt{spot\_2}, we can also take a look at the \texttt{theta} value
  from this model:
\end{itemize}

\begin{Shaded}
\begin{Highlighting}[]
\NormalTok{spot\_2.surrogate.theta}
\end{Highlighting}
\end{Shaded}

\begin{verbatim}
array([0.26287447])
\end{verbatim}

\hypertarget{tensorboard-3}{%
\subsubsection{TensorBoard}\label{tensorboard-3}}

Now we can start TensorBoard in the background with the following
command:

\begin{Shaded}
\begin{Highlighting}[]
\NormalTok{tensorboard {-}{-}logdir="./runs"}
\end{Highlighting}
\end{Shaded}

We can access the TensorBoard web server with the following URL:

\begin{Shaded}
\begin{Highlighting}[]
\NormalTok{http://localhost:6006/}
\end{Highlighting}
\end{Shaded}

The TensorBoard plot illustrates how \texttt{spotPython} can be used as
a microscope for the internal mechanisms of the surrogate-based
optimization process. Here, one important parameter, the learning rate
\(\theta\) of the Kriging surrogate is plotted against the number of
optimization steps.

\begin{figure}

{\centering \includegraphics[width=1\textwidth,height=\textheight]{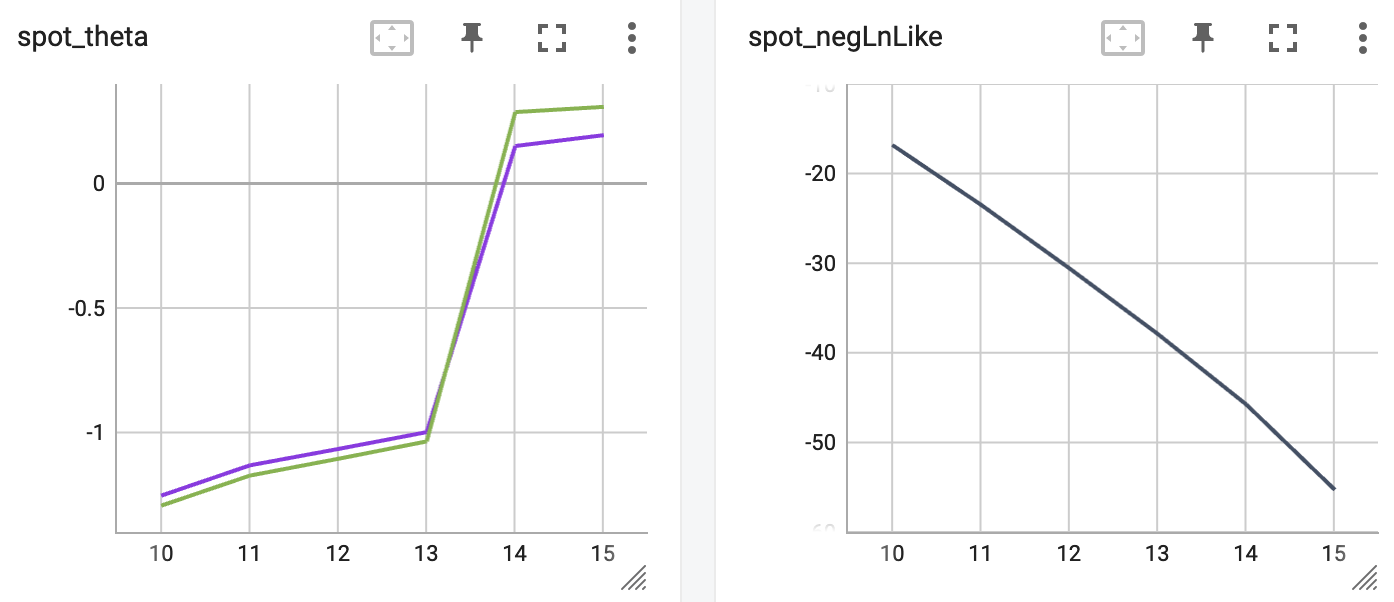}

}

\caption{TensorBoard visualization of the spotPython surrogate model.}

\end{figure}

\hypertarget{exercises-1}{%
\section{Exercises}\label{exercises-1}}

\hypertarget{fun_branin}{%
\subsection{\texorpdfstring{\texttt{fun\_branin}}{fun\_branin}}\label{fun_branin}}

\begin{itemize}
\tightlist
\item
  Describe the function.

  \begin{itemize}
  \tightlist
  \item
    The input dimension is \texttt{2}. The search range is
    \(-5 \leq x_1 \leq 10\) and \(0 \leq x_2 \leq 15\).
  \end{itemize}
\item
  Compare the results from \texttt{spotPython} run a) with isotropic and
  b) anisotropic surrogate models.
\item
  Modify the termination criterion: instead of the number of evaluations
  (which is specified via \texttt{fun\_evals}), the time should be used
  as the termination criterion. This can be done as follows
  (\texttt{max\_time=1} specifies a run time of one minute):
\end{itemize}

\begin{Shaded}
\begin{Highlighting}[]
\NormalTok{fun\_evals}\OperatorTok{=}\NormalTok{inf,}
\NormalTok{max\_time}\OperatorTok{=}\DecValTok{1}\NormalTok{,}
\end{Highlighting}
\end{Shaded}

\hypertarget{fun_sin_cos}{%
\subsection{\texorpdfstring{\texttt{fun\_sin\_cos}}{fun\_sin\_cos}}\label{fun_sin_cos}}

\begin{itemize}
\tightlist
\item
  Describe the function.

  \begin{itemize}
  \tightlist
  \item
    The input dimension is \texttt{2}. The search range is
    \(-2\pi \leq x_1 \leq 2\pi\) and \(-2\pi \leq x_2 \leq 2\pi\).
  \end{itemize}
\item
  Compare the results from \texttt{spotPython} run a) with isotropic and
  b) anisotropic surrogate models.
\item
  Modify the termination criterion (\texttt{max\_time} instead of
  \texttt{fun\_evals}) as described for \texttt{fun\_branin}.
\end{itemize}

\hypertarget{fun_runge}{%
\subsection{\texorpdfstring{\texttt{fun\_runge}}{fun\_runge}}\label{fun_runge}}

\begin{itemize}
\tightlist
\item
  Describe the function.

  \begin{itemize}
  \tightlist
  \item
    The input dimension is \texttt{2}. The search range is
    \(-5 \leq x_1 \leq 5\) and \(-5 \leq x_2 \leq 5\).
  \end{itemize}
\item
  Compare the results from \texttt{spotPython} run a) with isotropic and
  b) anisotropic surrogate models.
\item
  Modify the termination criterion (\texttt{max\_time} instead of
  \texttt{fun\_evals}) as described for \texttt{fun\_branin}.
\end{itemize}

\hypertarget{fun_wingwt}{%
\subsection{\texorpdfstring{\texttt{fun\_wingwt}}{fun\_wingwt}}\label{fun_wingwt}}

\begin{itemize}
\tightlist
\item
  Describe the function.

  \begin{itemize}
  \tightlist
  \item
    The input dimension is \texttt{10}. The search ranges are between 0
    and 1 (values are mapped internally to their natural bounds).
  \end{itemize}
\item
  Compare the results from \texttt{spotPython} run a) with isotropic and
  b) anisotropic surrogate models.
\item
  Modify the termination criterion (\texttt{max\_time} instead of
  \texttt{fun\_evals}) as described for \texttt{fun\_branin}.
\end{itemize}

\hypertarget{sec-sklearn-surrogates}{%
\chapter{\texorpdfstring{Using \texttt{sklearn} Surrogates in
\texttt{spotPython}}{Using sklearn Surrogates in spotPython}}\label{sec-sklearn-surrogates}}

Besides the internal kriging surrogate, which is used as a default py
\texttt{spotPython}, any surrogate model from \texttt{scikit-learn} can
be used as a surrogate in \texttt{spotPython}. This chapter explains how
to use \texttt{scikit-learn} surrogates in \texttt{spotPython}.

\begin{Shaded}
\begin{Highlighting}[]
\ImportTok{import}\NormalTok{ numpy }\ImportTok{as}\NormalTok{ np}
\ImportTok{from}\NormalTok{ math }\ImportTok{import}\NormalTok{ inf}
\ImportTok{from}\NormalTok{ spotPython.fun.objectivefunctions }\ImportTok{import}\NormalTok{ analytical}
\ImportTok{from}\NormalTok{ spotPython.spot }\ImportTok{import}\NormalTok{ spot}
\end{Highlighting}
\end{Shaded}

\hypertarget{example-branin-function-with-spotpythons-internal-kriging-surrogate}{%
\section{\texorpdfstring{Example: Branin Function with
\texttt{spotPython}'s Internal Kriging
Surrogate}{Example: Branin Function with spotPython's Internal Kriging Surrogate}}\label{example-branin-function-with-spotpythons-internal-kriging-surrogate}}

\hypertarget{the-objective-function-branin}{%
\subsection{The Objective Function
Branin}\label{the-objective-function-branin}}

\begin{itemize}
\item
  The \texttt{spotPython} package provides several classes of objective
  functions.
\item
  We will use an analytical objective function, i.e., a function that
  can be described by a (closed) formula.
\item
  Here we will use the Branin function:

\begin{verbatim}
  y = a * (x2 - b * x1**2 + c * x1 - r) ** 2 + s * (1 - t) * np.cos(x1) + s,
  where values of a, b, c, r, s and t are: a = 1, b = 5.1 / (4*pi**2),
  c = 5 / pi, r = 6, s = 10 and t = 1 / (8*pi).
\end{verbatim}
\item
  It has three global minima:

\begin{verbatim}
  f(x) = 0.397887 at (-pi, 12.275), (pi, 2.275), and (9.42478, 2.475).
\end{verbatim}
\end{itemize}

\begin{Shaded}
\begin{Highlighting}[]
\ImportTok{from}\NormalTok{ spotPython.fun.objectivefunctions }\ImportTok{import}\NormalTok{ analytical}
\NormalTok{lower }\OperatorTok{=}\NormalTok{ np.array([}\OperatorTok{{-}}\DecValTok{5}\NormalTok{,}\OperatorTok{{-}}\DecValTok{0}\NormalTok{])}
\NormalTok{upper }\OperatorTok{=}\NormalTok{ np.array([}\DecValTok{10}\NormalTok{,}\DecValTok{15}\NormalTok{])}
\end{Highlighting}
\end{Shaded}

\begin{Shaded}
\begin{Highlighting}[]
\NormalTok{fun }\OperatorTok{=}\NormalTok{ analytical().fun\_branin}
\end{Highlighting}
\end{Shaded}

\begin{tcolorbox}[enhanced jigsaw, left=2mm, title=\textcolor{quarto-callout-note-color}{\faInfo}\hspace{0.5em}{TensorBoard}, bottomrule=.15mm, titlerule=0mm, breakable, rightrule=.15mm, toprule=.15mm, coltitle=black, colbacktitle=quarto-callout-note-color!10!white, leftrule=.75mm, arc=.35mm, colframe=quarto-callout-note-color-frame, bottomtitle=1mm, colback=white, opacitybacktitle=0.6, toptitle=1mm, opacityback=0]

Similar to the one-dimensional case, which was introduced in Section
Section~\ref{sec-visualizing-tensorboard-01}, we can use TensorBoard to
monitor the progress of the optimization. We will use the same code,
only the prefix is different:

\begin{Shaded}
\begin{Highlighting}[]
\ImportTok{from}\NormalTok{ spotPython.utils.}\BuiltInTok{file} \ImportTok{import}\NormalTok{ get\_experiment\_name}
\ImportTok{from}\NormalTok{ spotPython.utils.init }\ImportTok{import}\NormalTok{ fun\_control\_init}
\ImportTok{from}\NormalTok{ spotPython.utils.}\BuiltInTok{file} \ImportTok{import}\NormalTok{ get\_spot\_tensorboard\_path}

\NormalTok{PREFIX }\OperatorTok{=} \StringTok{"04"}
\NormalTok{experiment\_name }\OperatorTok{=}\NormalTok{ get\_experiment\_name(prefix}\OperatorTok{=}\NormalTok{PREFIX)}
\BuiltInTok{print}\NormalTok{(experiment\_name)}

\NormalTok{fun\_control }\OperatorTok{=}\NormalTok{ fun\_control\_init(}
\NormalTok{    spot\_tensorboard\_path}\OperatorTok{=}\NormalTok{get\_spot\_tensorboard\_path(experiment\_name))}
\end{Highlighting}
\end{Shaded}

\begin{verbatim}
04_bartz09_2023-07-17_08-49-25
\end{verbatim}

\end{tcolorbox}

\hypertarget{running-the-surrogate-model-based-optimizer-spot}{%
\subsection{\texorpdfstring{Running the surrogate model based optimizer
\texttt{Spot}:}{Running the surrogate model based optimizer Spot:}}\label{running-the-surrogate-model-based-optimizer-spot}}

\begin{Shaded}
\begin{Highlighting}[]
\NormalTok{spot\_2 }\OperatorTok{=}\NormalTok{ spot.Spot(fun}\OperatorTok{=}\NormalTok{fun,}
\NormalTok{                   lower }\OperatorTok{=}\NormalTok{ lower,}
\NormalTok{                   upper }\OperatorTok{=}\NormalTok{ upper,}
\NormalTok{                   fun\_evals }\OperatorTok{=} \DecValTok{20}\NormalTok{,}
\NormalTok{                   max\_time }\OperatorTok{=}\NormalTok{ inf,}
\NormalTok{                   seed}\OperatorTok{=}\DecValTok{123}\NormalTok{,}
\NormalTok{                   design\_control}\OperatorTok{=}\NormalTok{\{}\StringTok{"init\_size"}\NormalTok{: }\DecValTok{10}\NormalTok{\},}
\NormalTok{                   fun\_control}\OperatorTok{=}\NormalTok{fun\_control)}
\end{Highlighting}
\end{Shaded}

\begin{Shaded}
\begin{Highlighting}[]
\NormalTok{spot\_2.run()}
\end{Highlighting}
\end{Shaded}

\begin{verbatim}
spotPython tuning: 3.4474628349075243 [######----] 55.00% 
\end{verbatim}

\begin{verbatim}
spotPython tuning: 3.4474628349075243 [######----] 60.00% 
\end{verbatim}

\begin{verbatim}
spotPython tuning: 3.039485786016101 [######----] 65.00% 
\end{verbatim}

\begin{verbatim}
spotPython tuning: 3.039485786016101 [#######---] 70.00% 
\end{verbatim}

\begin{verbatim}
spotPython tuning: 1.1632959357427755 [########--] 75.00% 
\end{verbatim}

\begin{verbatim}
spotPython tuning: 0.6123887750698636 [########--] 80.00% 
\end{verbatim}

\begin{verbatim}
spotPython tuning: 0.4575920097730535 [########--] 85.00% 
\end{verbatim}

\begin{verbatim}
spotPython tuning: 0.3982295132785083 [#########-] 90.00% 
\end{verbatim}

\begin{verbatim}
spotPython tuning: 0.3982295132785083 [##########] 95.00% 
\end{verbatim}

\begin{verbatim}
spotPython tuning: 0.3982295132785083 [##########] 100.00% Done...
\end{verbatim}

\begin{verbatim}
<spotPython.spot.spot.Spot at 0x13ab7cca0>
\end{verbatim}

\hypertarget{tensorboard-5}{%
\subsection{TensorBoard}\label{tensorboard-5}}

Now we can start TensorBoard in the background with the following
command:

\begin{Shaded}
\begin{Highlighting}[]
\NormalTok{tensorboard {-}{-}logdir="./runs"}
\end{Highlighting}
\end{Shaded}

We can access the TensorBoard web server with the following URL:

\begin{Shaded}
\begin{Highlighting}[]
\NormalTok{http://localhost:6006/}
\end{Highlighting}
\end{Shaded}

The TensorBoard plot illustrates how \texttt{spotPython} can be used as
a microscope for the internal mechanisms of the surrogate-based
optimization process. Here, one important parameter, the learning rate
\(\theta\) of the Kriging surrogate is plotted against the number of
optimization steps.

\begin{figure}

{\centering \includegraphics[width=1\textwidth,height=\textheight]{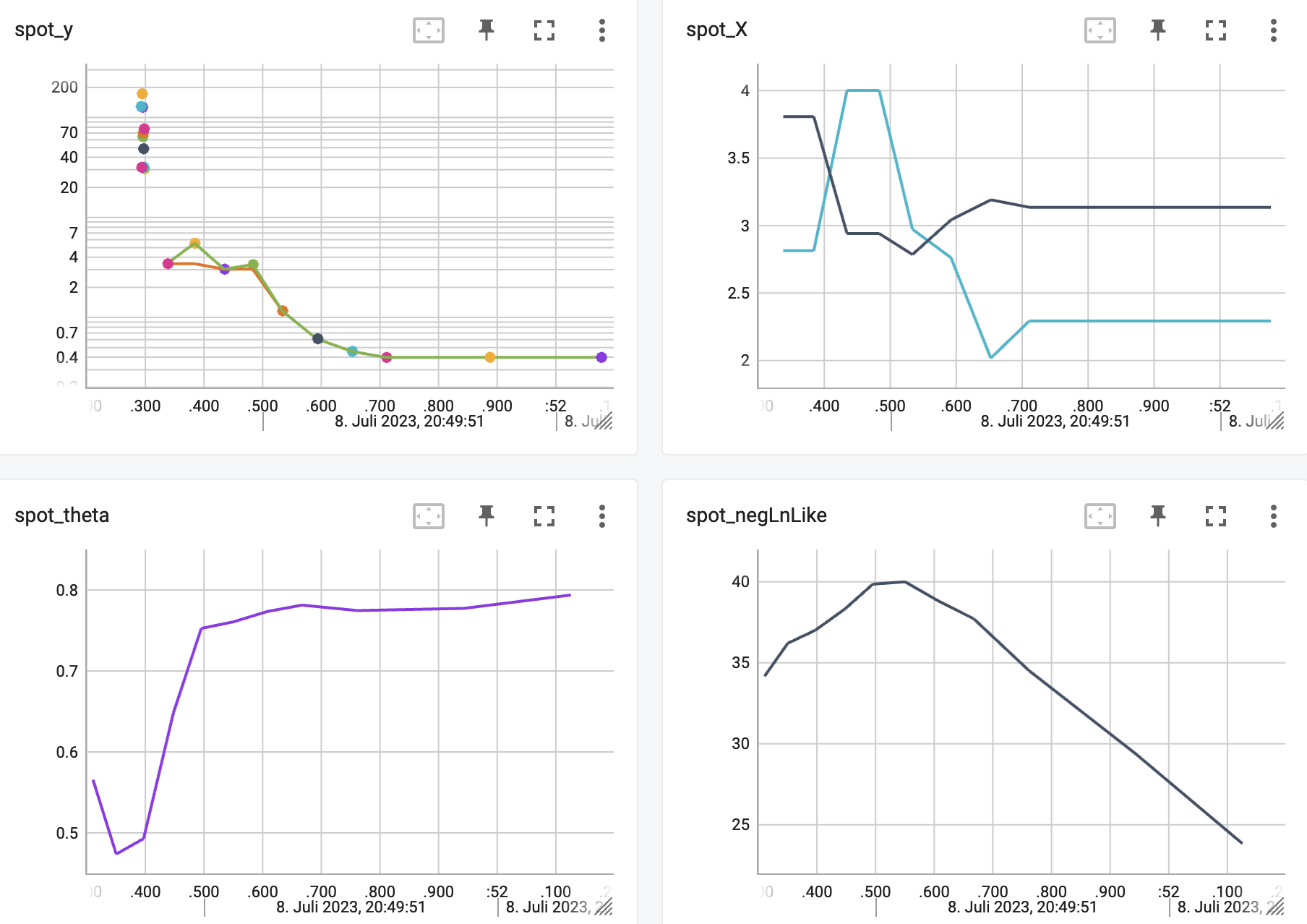}

}

\caption{TensorBoard visualization of the spotPython optimization
process and the surrogate model.}

\end{figure}

\hypertarget{print-the-results-1}{%
\subsection{Print the Results}\label{print-the-results-1}}

\begin{Shaded}
\begin{Highlighting}[]
\NormalTok{spot\_2.print\_results()}
\end{Highlighting}
\end{Shaded}

\begin{verbatim}
min y: 0.3982295132785083
x0: 3.135528626303215
x1: 2.2926027772585886
\end{verbatim}

\begin{verbatim}
[['x0', 3.135528626303215], ['x1', 2.2926027772585886]]
\end{verbatim}

\hypertarget{show-the-progress-and-the-surrogate}{%
\subsection{Show the Progress and the
Surrogate}\label{show-the-progress-and-the-surrogate}}

\begin{Shaded}
\begin{Highlighting}[]
\NormalTok{spot\_2.plot\_progress(log\_y}\OperatorTok{=}\VariableTok{True}\NormalTok{)}
\end{Highlighting}
\end{Shaded}

\begin{figure}[H]

{\centering \includegraphics{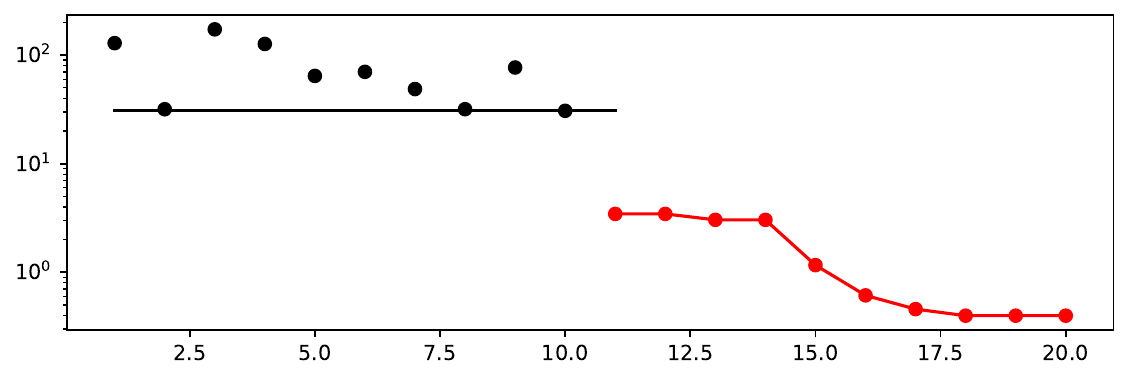}

}

\end{figure}

\begin{Shaded}
\begin{Highlighting}[]
\NormalTok{spot\_2.surrogate.plot()}
\end{Highlighting}
\end{Shaded}

\begin{figure}[H]

{\centering \includegraphics{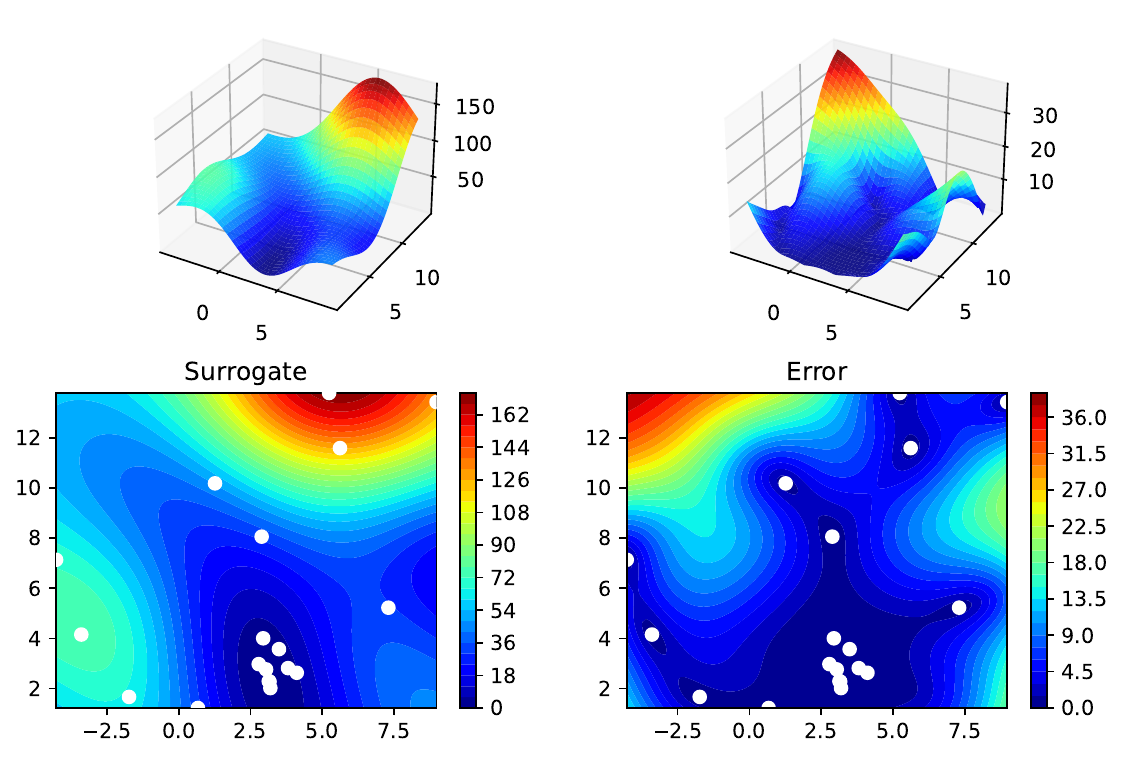}

}

\end{figure}

\hypertarget{example-using-surrogates-from-scikit-learn}{%
\section{Example: Using Surrogates From
scikit-learn}\label{example-using-surrogates-from-scikit-learn}}

\begin{itemize}
\tightlist
\item
  Default is the \texttt{spotPython} (i.e., the internal)
  \texttt{kriging} surrogate.
\item
  It can be called explicitely and passed to \texttt{Spot}.
\end{itemize}

\begin{Shaded}
\begin{Highlighting}[]
\ImportTok{from}\NormalTok{ spotPython.build.kriging }\ImportTok{import}\NormalTok{ Kriging}
\NormalTok{S\_0 }\OperatorTok{=}\NormalTok{ Kriging(name}\OperatorTok{=}\StringTok{\textquotesingle{}kriging\textquotesingle{}}\NormalTok{, seed}\OperatorTok{=}\DecValTok{123}\NormalTok{)}
\end{Highlighting}
\end{Shaded}

\begin{itemize}
\tightlist
\item
  Alternatively, models from \texttt{scikit-learn} can be selected,
  e.g., Gaussian Process, RBFs, Regression Trees, etc.
\end{itemize}

\begin{Shaded}
\begin{Highlighting}[]
\CommentTok{\# Needed for the sklearn surrogates:}
\ImportTok{from}\NormalTok{ sklearn.gaussian\_process }\ImportTok{import}\NormalTok{ GaussianProcessRegressor}
\ImportTok{from}\NormalTok{ sklearn.gaussian\_process.kernels }\ImportTok{import}\NormalTok{ RBF}
\ImportTok{from}\NormalTok{ sklearn.tree }\ImportTok{import}\NormalTok{ DecisionTreeRegressor}
\ImportTok{from}\NormalTok{ sklearn.ensemble }\ImportTok{import}\NormalTok{ RandomForestRegressor}
\ImportTok{from}\NormalTok{ sklearn }\ImportTok{import}\NormalTok{ linear\_model}
\ImportTok{from}\NormalTok{ sklearn }\ImportTok{import}\NormalTok{ tree}
\ImportTok{import}\NormalTok{ pandas }\ImportTok{as}\NormalTok{ pd}
\end{Highlighting}
\end{Shaded}

\begin{itemize}
\tightlist
\item
  Here are some additional models that might be useful later:
\end{itemize}

\begin{Shaded}
\begin{Highlighting}[]
\NormalTok{S\_Tree }\OperatorTok{=}\NormalTok{ DecisionTreeRegressor(random\_state}\OperatorTok{=}\DecValTok{0}\NormalTok{)}
\NormalTok{S\_LM }\OperatorTok{=}\NormalTok{ linear\_model.LinearRegression()}
\NormalTok{S\_Ridge }\OperatorTok{=}\NormalTok{ linear\_model.Ridge()}
\NormalTok{S\_RF }\OperatorTok{=}\NormalTok{ RandomForestRegressor(max\_depth}\OperatorTok{=}\DecValTok{2}\NormalTok{, random\_state}\OperatorTok{=}\DecValTok{0}\NormalTok{)}
\end{Highlighting}
\end{Shaded}

\hypertarget{gaussianprocessregressor-as-a-surrogate}{%
\subsection{GaussianProcessRegressor as a
Surrogate}\label{gaussianprocessregressor-as-a-surrogate}}

\begin{itemize}
\tightlist
\item
  To use a Gaussian Process model from \texttt{sklearn}, that is similar
  to \texttt{spotPython}'s \texttt{Kriging}, we can proceed as follows:
\end{itemize}

\begin{Shaded}
\begin{Highlighting}[]
\NormalTok{kernel }\OperatorTok{=} \DecValTok{1} \OperatorTok{*}\NormalTok{ RBF(length\_scale}\OperatorTok{=}\FloatTok{1.0}\NormalTok{, length\_scale\_bounds}\OperatorTok{=}\NormalTok{(}\FloatTok{1e{-}2}\NormalTok{, }\FloatTok{1e2}\NormalTok{))}
\NormalTok{S\_GP }\OperatorTok{=}\NormalTok{ GaussianProcessRegressor(kernel}\OperatorTok{=}\NormalTok{kernel, n\_restarts\_optimizer}\OperatorTok{=}\DecValTok{9}\NormalTok{)}
\end{Highlighting}
\end{Shaded}

\begin{itemize}
\item
  The scikit-learn GP model \texttt{S\_GP} is selected for \texttt{Spot}
  as follows:

  \texttt{surrogate\ =\ S\_GP}
\item
  We can check the kind of surogate model with the command
  \texttt{isinstance}:
\end{itemize}

\begin{Shaded}
\begin{Highlighting}[]
\BuiltInTok{isinstance}\NormalTok{(S\_GP, GaussianProcessRegressor)}
\end{Highlighting}
\end{Shaded}

\begin{verbatim}
True
\end{verbatim}

\begin{Shaded}
\begin{Highlighting}[]
\BuiltInTok{isinstance}\NormalTok{(S\_0, Kriging)}
\end{Highlighting}
\end{Shaded}

\begin{verbatim}
True
\end{verbatim}

\begin{itemize}
\tightlist
\item
  Similar to the \texttt{Spot} run with the internal \texttt{Kriging}
  model, we can call the run with the \texttt{scikit-learn} surrogate:
\end{itemize}

\begin{Shaded}
\begin{Highlighting}[]
\NormalTok{fun }\OperatorTok{=}\NormalTok{ analytical(seed}\OperatorTok{=}\DecValTok{123}\NormalTok{).fun\_branin}
\NormalTok{spot\_2\_GP }\OperatorTok{=}\NormalTok{ spot.Spot(fun}\OperatorTok{=}\NormalTok{fun,}
\NormalTok{                   lower }\OperatorTok{=}\NormalTok{ lower,}
\NormalTok{                   upper }\OperatorTok{=}\NormalTok{ upper,}
\NormalTok{                   fun\_evals }\OperatorTok{=} \DecValTok{20}\NormalTok{,}
\NormalTok{                   seed}\OperatorTok{=}\DecValTok{123}\NormalTok{,}
\NormalTok{                   design\_control}\OperatorTok{=}\NormalTok{\{}\StringTok{"init\_size"}\NormalTok{: }\DecValTok{10}\NormalTok{\},}
\NormalTok{                   surrogate }\OperatorTok{=}\NormalTok{ S\_GP)}
\NormalTok{spot\_2\_GP.run()}
\end{Highlighting}
\end{Shaded}

\begin{verbatim}
spotPython tuning: 18.86511402323416 [######----] 55.00% 
\end{verbatim}

\begin{verbatim}
spotPython tuning: 4.0669082302178285 [######----] 60.00% 
\end{verbatim}

\begin{verbatim}
spotPython tuning: 3.4618162795514635 [######----] 65.00% 
\end{verbatim}

\begin{verbatim}
spotPython tuning: 3.4618162795514635 [#######---] 70.00% 
\end{verbatim}

\begin{verbatim}
spotPython tuning: 1.3283163482563598 [########--] 75.00% 
\end{verbatim}

\begin{verbatim}
spotPython tuning: 0.9542592376072765 [########--] 80.00% 
\end{verbatim}

\begin{verbatim}
spotPython tuning: 0.9289433893626615 [########--] 85.00% 
\end{verbatim}

\begin{verbatim}
spotPython tuning: 0.3981201359931852 [#########-] 90.00% 
\end{verbatim}

\begin{verbatim}
spotPython tuning: 0.39799355388506363 [##########] 95.00% 
\end{verbatim}

\begin{verbatim}
spotPython tuning: 0.39799355388506363 [##########] 100.00% Done...
\end{verbatim}

\begin{verbatim}
<spotPython.spot.spot.Spot at 0x2bc4f6020>
\end{verbatim}

\begin{Shaded}
\begin{Highlighting}[]
\NormalTok{spot\_2\_GP.plot\_progress()}
\end{Highlighting}
\end{Shaded}

\begin{figure}[H]

{\centering \includegraphics{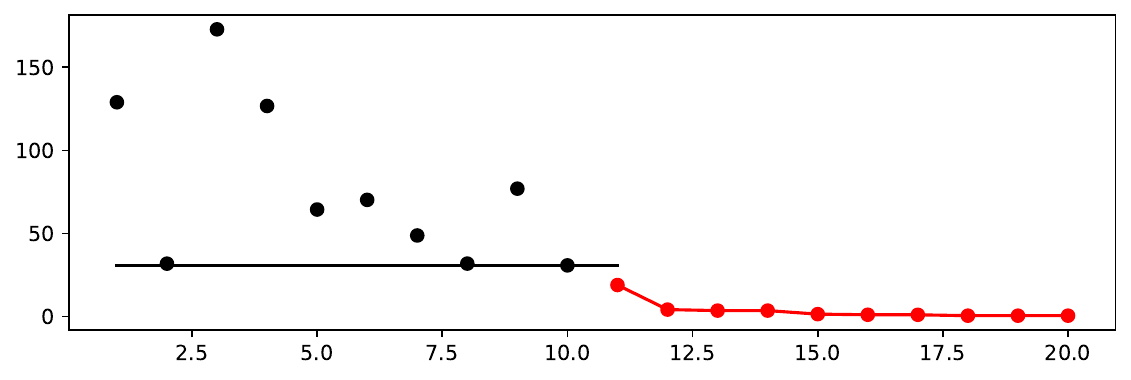}

}

\end{figure}

\begin{Shaded}
\begin{Highlighting}[]
\NormalTok{spot\_2\_GP.print\_results()}
\end{Highlighting}
\end{Shaded}

\begin{verbatim}
min y: 0.39799355388506363
x0: 3.1460470114516994
x1: 2.2748359190479013
\end{verbatim}

\begin{verbatim}
[['x0', 3.1460470114516994], ['x1', 2.2748359190479013]]
\end{verbatim}

\hypertarget{example-one-dimensional-sphere-function-with-spotpythons-kriging}{%
\section{\texorpdfstring{Example: One-dimensional Sphere Function With
\texttt{spotPython}'s
Kriging}{Example: One-dimensional Sphere Function With spotPython's Kriging}}\label{example-one-dimensional-sphere-function-with-spotpythons-kriging}}

\begin{itemize}
\tightlist
\item
  In this example, we will use an one-dimensional function, which allows
  us to visualize the optimization process.

  \begin{itemize}
  \tightlist
  \item
    \texttt{show\_models=\ True} is added to the argument list.
  \end{itemize}
\end{itemize}

\begin{Shaded}
\begin{Highlighting}[]
\ImportTok{from}\NormalTok{ spotPython.fun.objectivefunctions }\ImportTok{import}\NormalTok{ analytical}
\NormalTok{lower }\OperatorTok{=}\NormalTok{ np.array([}\OperatorTok{{-}}\DecValTok{1}\NormalTok{])}
\NormalTok{upper }\OperatorTok{=}\NormalTok{ np.array([}\DecValTok{1}\NormalTok{])}
\NormalTok{fun }\OperatorTok{=}\NormalTok{ analytical(seed}\OperatorTok{=}\DecValTok{123}\NormalTok{).fun\_sphere}
\end{Highlighting}
\end{Shaded}

\begin{Shaded}
\begin{Highlighting}[]
\NormalTok{spot\_1 }\OperatorTok{=}\NormalTok{ spot.Spot(fun}\OperatorTok{=}\NormalTok{fun,}
\NormalTok{                   lower }\OperatorTok{=}\NormalTok{ lower,}
\NormalTok{                   upper }\OperatorTok{=}\NormalTok{ upper,}
\NormalTok{                   fun\_evals }\OperatorTok{=} \DecValTok{10}\NormalTok{,}
\NormalTok{                   max\_time }\OperatorTok{=}\NormalTok{ inf,}
\NormalTok{                   seed}\OperatorTok{=}\DecValTok{123}\NormalTok{,}
\NormalTok{                   show\_models}\OperatorTok{=} \VariableTok{True}\NormalTok{,}
\NormalTok{                   tolerance\_x }\OperatorTok{=}\NormalTok{ np.sqrt(np.spacing(}\DecValTok{1}\NormalTok{)),}
\NormalTok{                   design\_control}\OperatorTok{=}\NormalTok{\{}\StringTok{"init\_size"}\NormalTok{: }\DecValTok{3}\NormalTok{\},)}
\NormalTok{spot\_1.run()}
\end{Highlighting}
\end{Shaded}

\begin{figure}[H]

{\centering \includegraphics{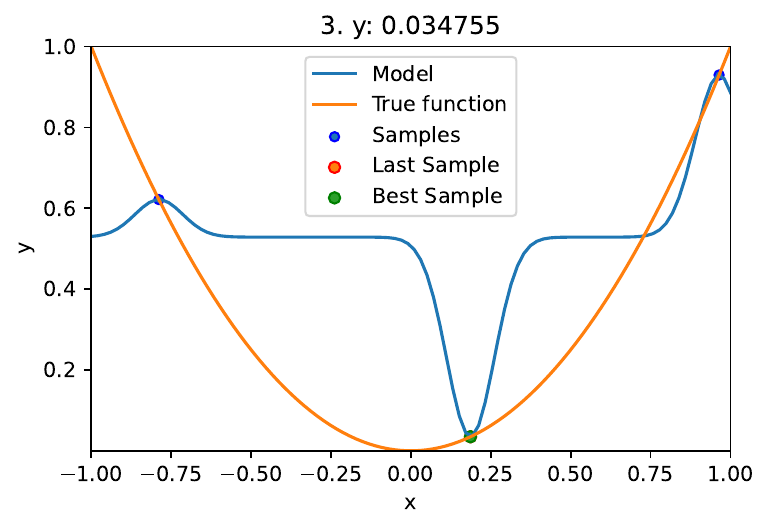}

}

\end{figure}

\begin{figure}[H]

{\centering \includegraphics{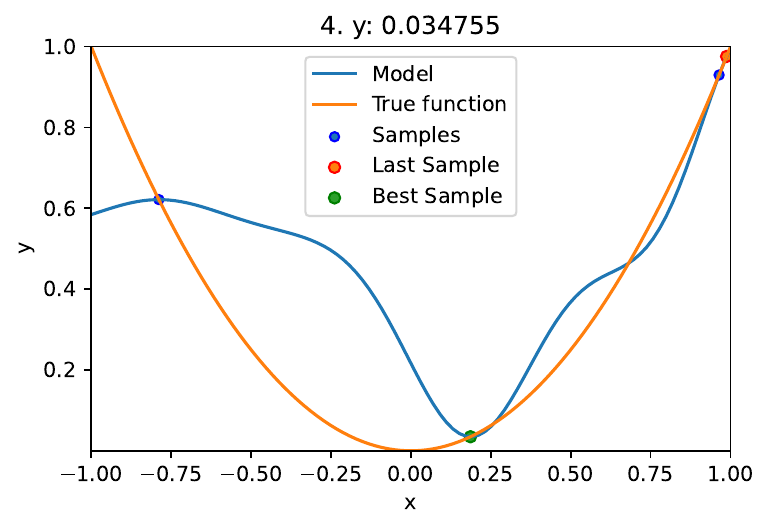}

}

\end{figure}

\begin{verbatim}
spotPython tuning: 0.03475493366922229 [####------] 40.00% 
\end{verbatim}

\begin{figure}[H]

{\centering \includegraphics{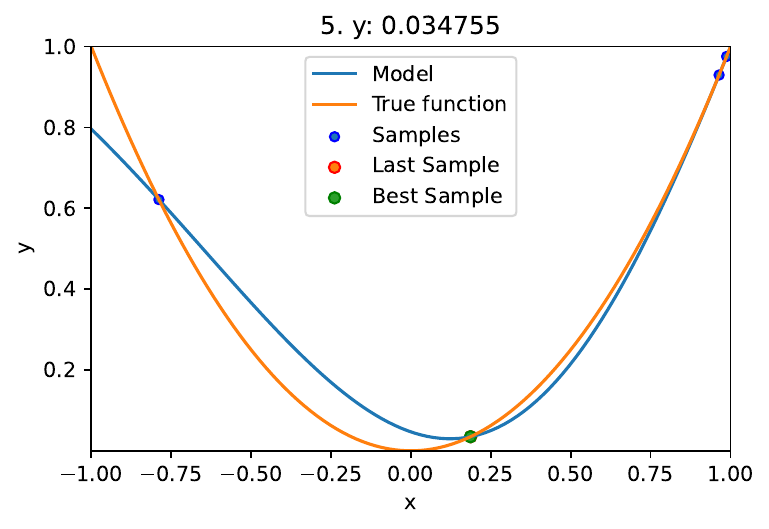}

}

\end{figure}

\begin{verbatim}
spotPython tuning: 0.03475493366922229 [#####-----] 50.00% 
\end{verbatim}

\begin{figure}[H]

{\centering \includegraphics{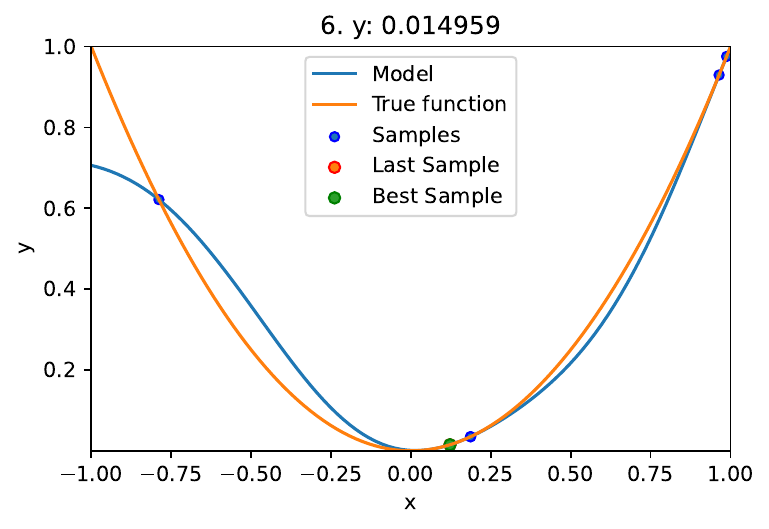}

}

\end{figure}

\begin{verbatim}
spotPython tuning: 0.014958671130600643 [######----] 60.00% 
\end{verbatim}

\begin{figure}[H]

{\centering \includegraphics{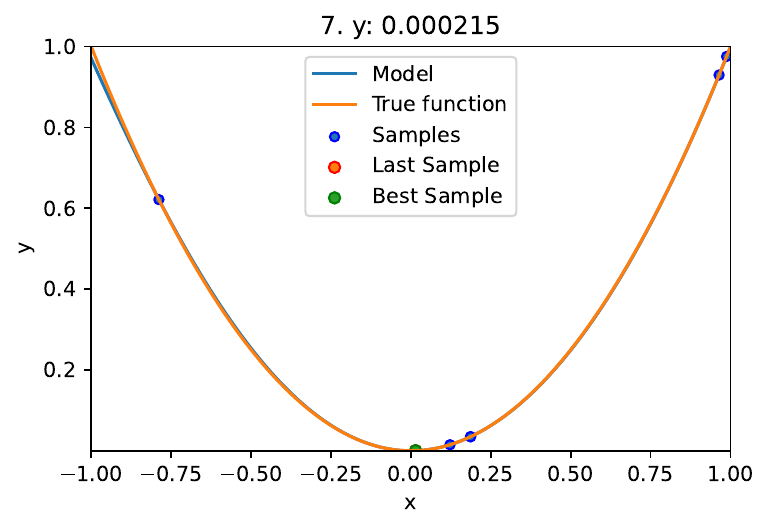}

}

\end{figure}

\begin{verbatim}
spotPython tuning: 0.0002154633036537174 [#######---] 70.00% 
\end{verbatim}

\begin{figure}[H]

{\centering \includegraphics{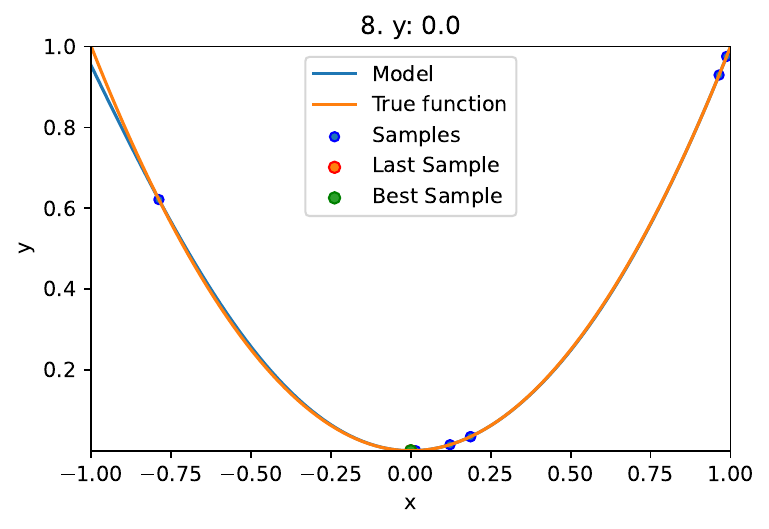}

}

\end{figure}

\begin{verbatim}
spotPython tuning: 4.41925228274096e-08 [########--] 80.00% 
\end{verbatim}

\begin{figure}[H]

{\centering \includegraphics{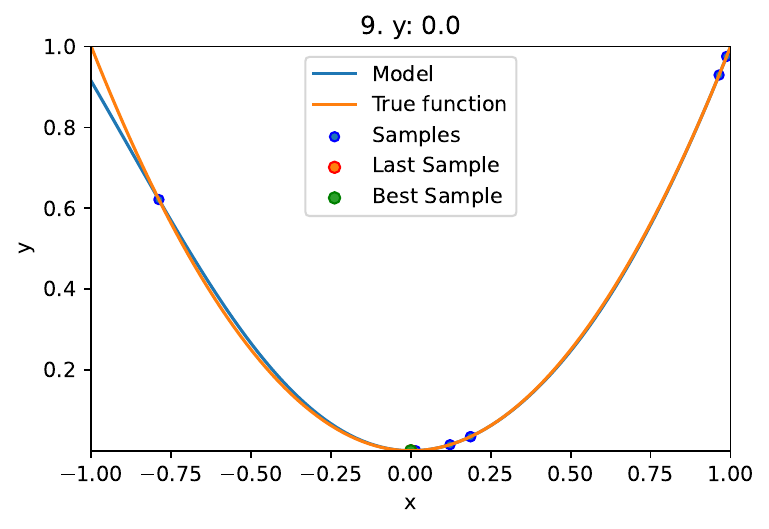}

}

\end{figure}

\begin{verbatim}
spotPython tuning: 4.41925228274096e-08 [#########-] 90.00% 
\end{verbatim}

\begin{figure}[H]

{\centering \includegraphics{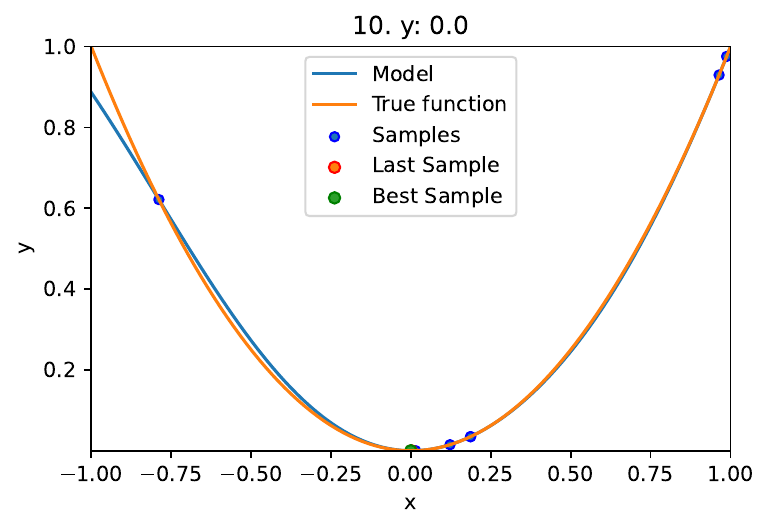}

}

\end{figure}

\begin{verbatim}
spotPython tuning: 4.41925228274096e-08 [##########] 100.00% Done...
\end{verbatim}

\begin{verbatim}
<spotPython.spot.spot.Spot at 0x2bd636b90>
\end{verbatim}

\hypertarget{results-2}{%
\subsection{Results}\label{results-2}}

\begin{Shaded}
\begin{Highlighting}[]
\NormalTok{spot\_1.print\_results()}
\end{Highlighting}
\end{Shaded}

\begin{verbatim}
min y: 4.41925228274096e-08
x0: -0.00021022017702259125
\end{verbatim}

\begin{verbatim}
[['x0', -0.00021022017702259125]]
\end{verbatim}

\begin{Shaded}
\begin{Highlighting}[]
\NormalTok{spot\_1.plot\_progress(log\_y}\OperatorTok{=}\VariableTok{True}\NormalTok{)}
\end{Highlighting}
\end{Shaded}

\begin{figure}[H]

{\centering \includegraphics{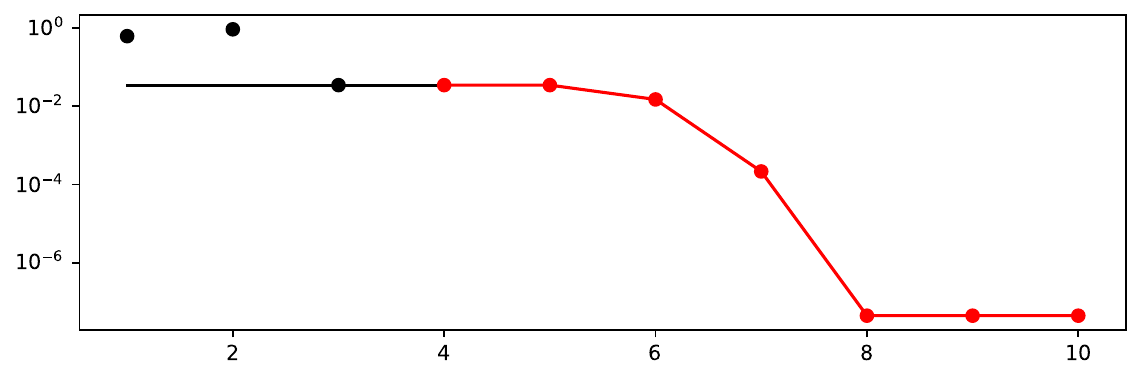}

}

\end{figure}

\begin{itemize}
\tightlist
\item
  The method \texttt{plot\_model} plots the final surrogate:
\end{itemize}

\begin{Shaded}
\begin{Highlighting}[]
\NormalTok{spot\_1.plot\_model()}
\end{Highlighting}
\end{Shaded}

\begin{figure}[H]

{\centering \includegraphics{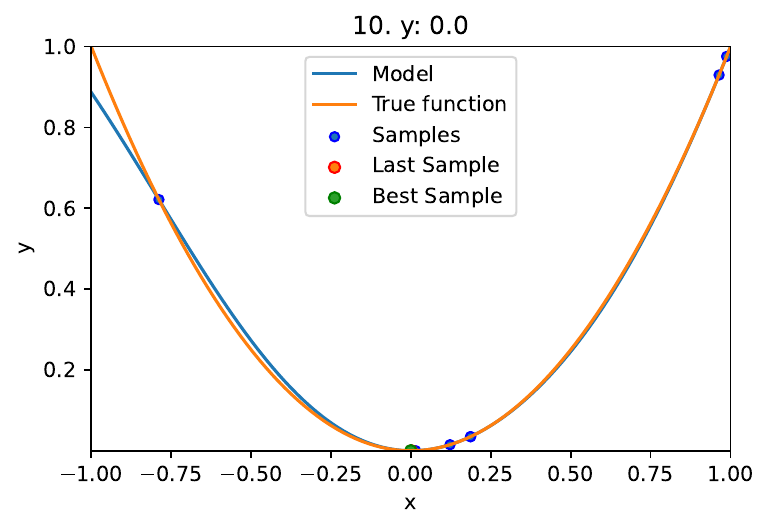}

}

\end{figure}

\hypertarget{example-sklearn-model-gaussianprocess}{%
\section{\texorpdfstring{Example: \texttt{Sklearn} Model
GaussianProcess}{Example: Sklearn Model GaussianProcess}}\label{example-sklearn-model-gaussianprocess}}

\begin{itemize}
\tightlist
\item
  This example visualizes the search process on the
  \texttt{GaussianProcessRegression} surrogate from \texttt{sklearn}.
\item
  Therefore \texttt{surrogate\ =\ S\_GP} is added to the argument list.
\end{itemize}

\begin{Shaded}
\begin{Highlighting}[]
\NormalTok{fun }\OperatorTok{=}\NormalTok{ analytical(seed}\OperatorTok{=}\DecValTok{123}\NormalTok{).fun\_sphere}
\NormalTok{spot\_1\_GP }\OperatorTok{=}\NormalTok{ spot.Spot(fun}\OperatorTok{=}\NormalTok{fun,}
\NormalTok{                   lower }\OperatorTok{=}\NormalTok{ lower,}
\NormalTok{                   upper }\OperatorTok{=}\NormalTok{ upper,}
\NormalTok{                   fun\_evals }\OperatorTok{=} \DecValTok{10}\NormalTok{,}
\NormalTok{                   max\_time }\OperatorTok{=}\NormalTok{ inf,}
\NormalTok{                   seed}\OperatorTok{=}\DecValTok{123}\NormalTok{,}
\NormalTok{                   show\_models}\OperatorTok{=} \VariableTok{True}\NormalTok{,}
\NormalTok{                   design\_control}\OperatorTok{=}\NormalTok{\{}\StringTok{"init\_size"}\NormalTok{: }\DecValTok{3}\NormalTok{\},}
\NormalTok{                   surrogate }\OperatorTok{=}\NormalTok{ S\_GP)}
\NormalTok{spot\_1\_GP.run()}
\end{Highlighting}
\end{Shaded}

\begin{figure}[H]

{\centering \includegraphics{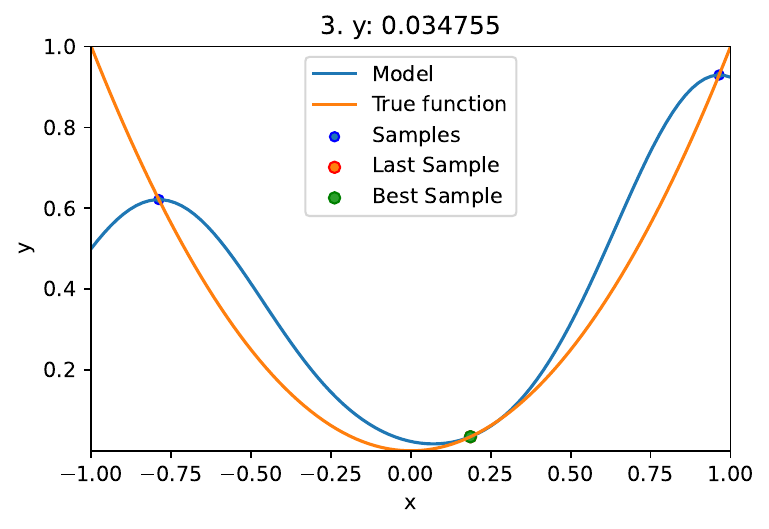}

}

\end{figure}

\begin{figure}[H]

{\centering \includegraphics{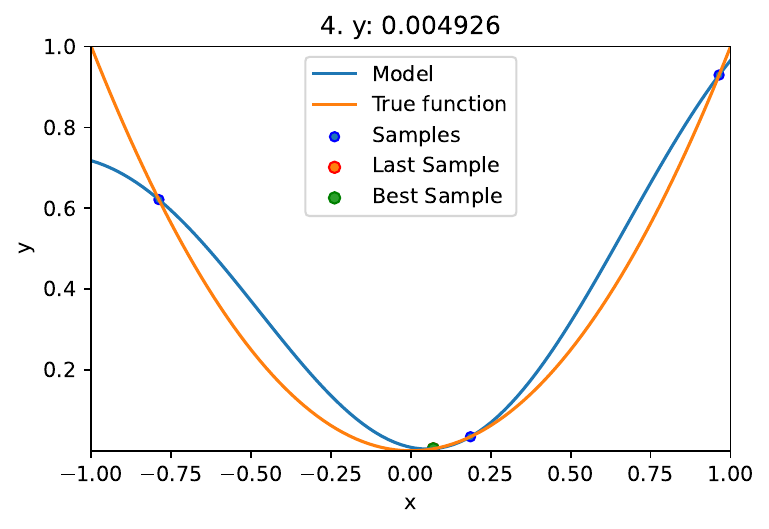}

}

\end{figure}

\begin{verbatim}
spotPython tuning: 0.004925761656816393 [####------] 40.00% 
\end{verbatim}

\begin{figure}[H]

{\centering \includegraphics{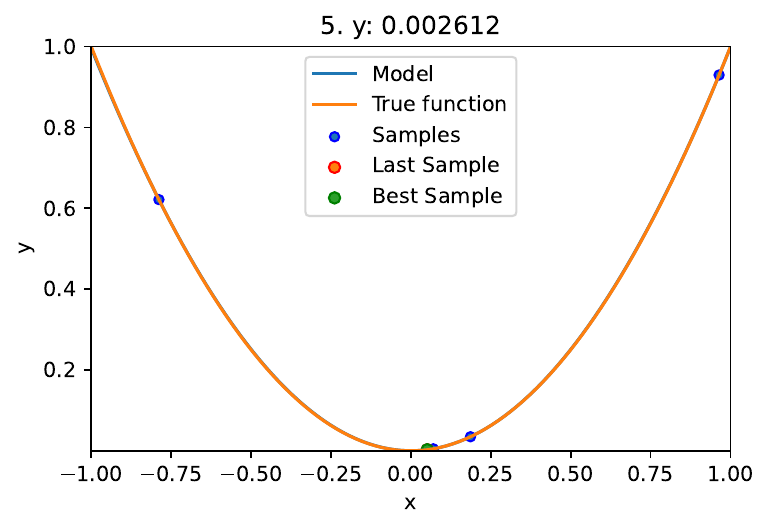}

}

\end{figure}

\begin{verbatim}
spotPython tuning: 0.0026120758453649505 [#####-----] 50.00% 
\end{verbatim}

\begin{figure}[H]

{\centering \includegraphics{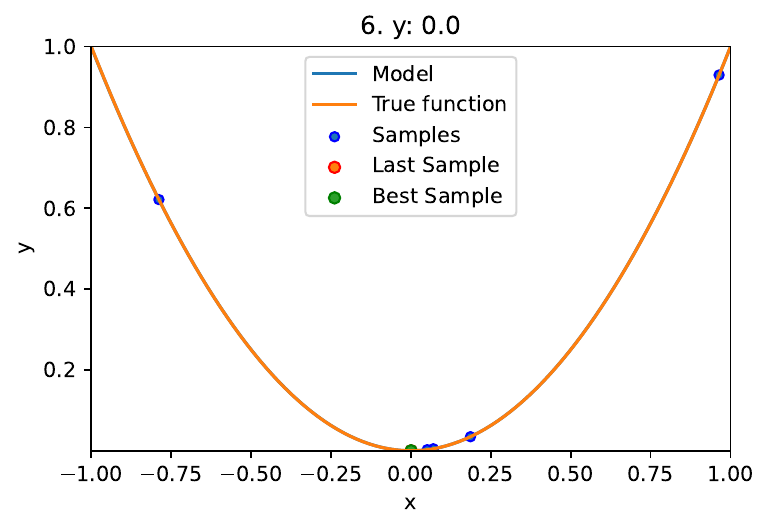}

}

\end{figure}

\begin{verbatim}
spotPython tuning: 4.492968068412204e-07 [######----] 60.00% 
\end{verbatim}

\begin{figure}[H]

{\centering \includegraphics{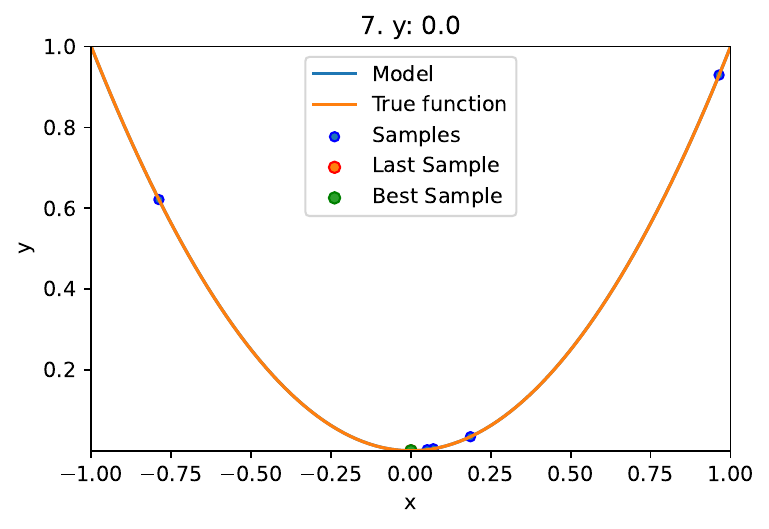}

}

\end{figure}

\begin{verbatim}
spotPython tuning: 5.520019085369139e-08 [#######---] 70.00% 
\end{verbatim}

\begin{figure}[H]

{\centering \includegraphics{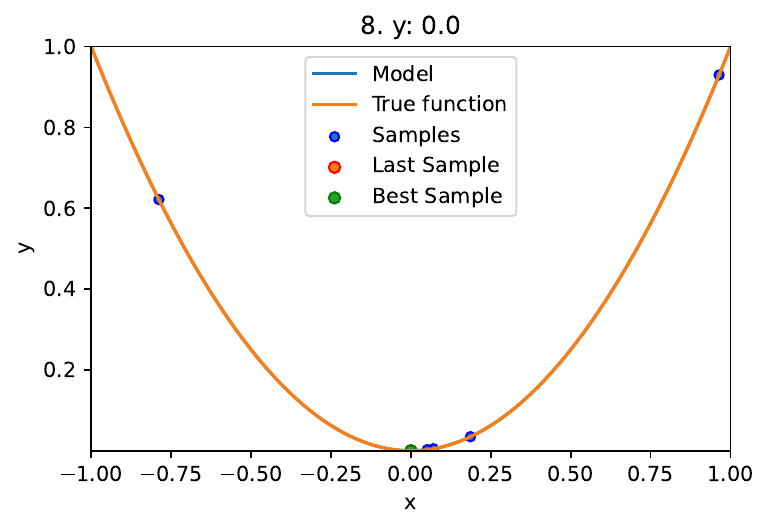}

}

\end{figure}

\begin{verbatim}
spotPython tuning: 1.8830522883506717e-08 [########--] 80.00% 
\end{verbatim}

\begin{figure}[H]

{\centering \includegraphics{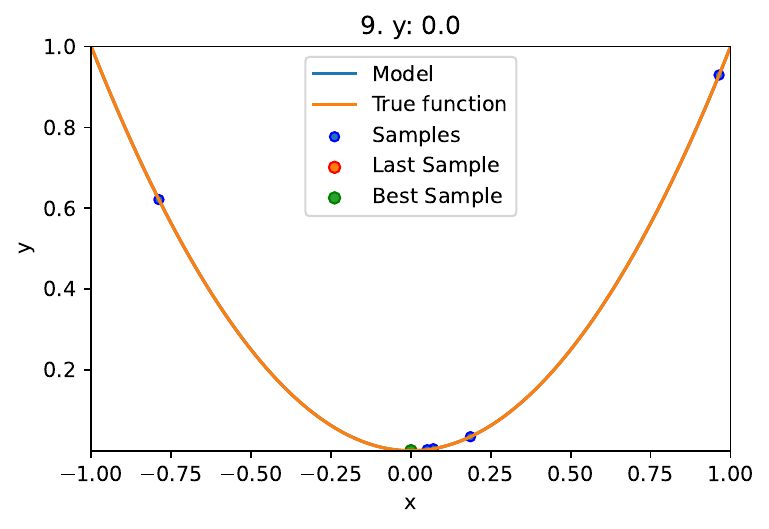}

}

\end{figure}

\begin{verbatim}
spotPython tuning: 1.2165253306918689e-08 [#########-] 90.00% 
\end{verbatim}

\begin{figure}[H]

{\centering \includegraphics{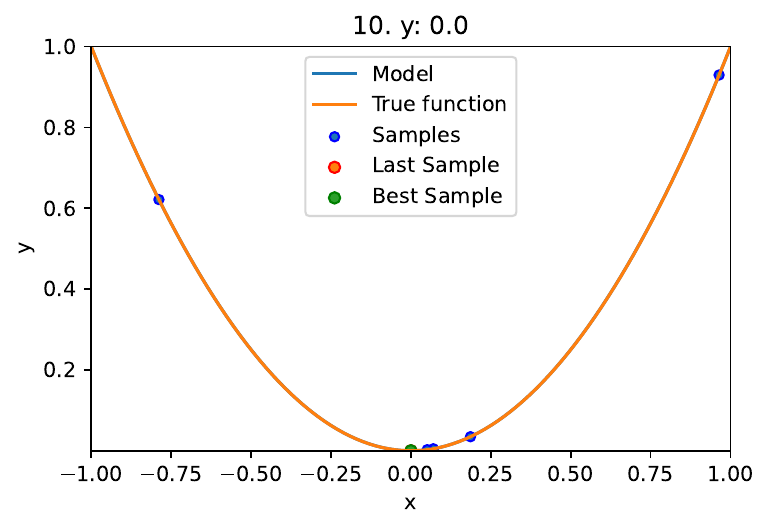}

}

\end{figure}

\begin{verbatim}
spotPython tuning: 1.0471089618292772e-08 [##########] 100.00% Done...
\end{verbatim}

\begin{verbatim}
<spotPython.spot.spot.Spot at 0x2bdbe5960>
\end{verbatim}

\begin{Shaded}
\begin{Highlighting}[]
\NormalTok{spot\_1\_GP.print\_results()}
\end{Highlighting}
\end{Shaded}

\begin{verbatim}
min y: 1.0471089618292772e-08
x0: 0.00010232834220436082
\end{verbatim}

\begin{verbatim}
[['x0', 0.00010232834220436082]]
\end{verbatim}

\begin{Shaded}
\begin{Highlighting}[]
\NormalTok{spot\_1\_GP.plot\_progress(log\_y}\OperatorTok{=}\VariableTok{True}\NormalTok{)}
\end{Highlighting}
\end{Shaded}

\begin{figure}[H]

{\centering \includegraphics{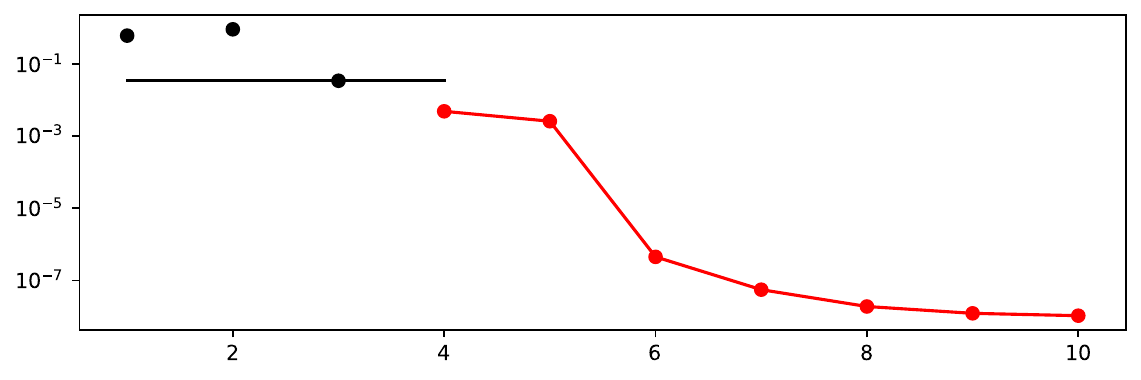}

}

\end{figure}

\begin{Shaded}
\begin{Highlighting}[]
\NormalTok{spot\_1\_GP.plot\_model()}
\end{Highlighting}
\end{Shaded}

\begin{figure}[H]

{\centering \includegraphics{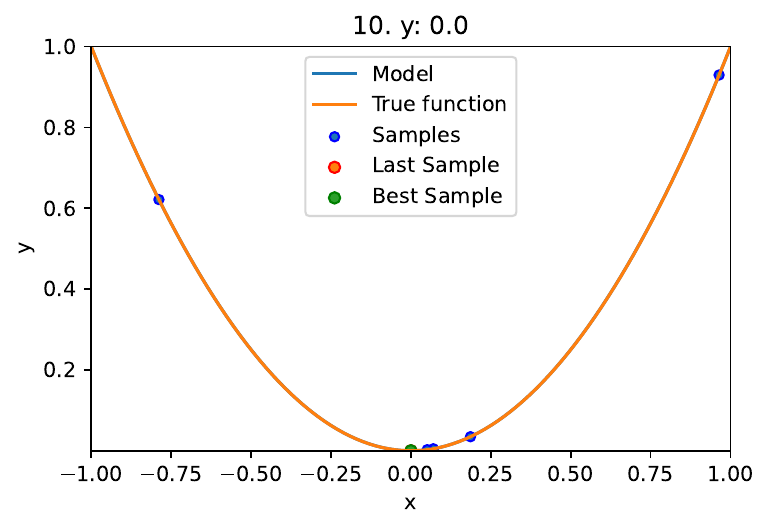}

}

\end{figure}

\hypertarget{exercises-2}{%
\section{Exercises}\label{exercises-2}}

\hypertarget{decisiontreeregressor}{%
\subsection{\texorpdfstring{\texttt{DecisionTreeRegressor}}{DecisionTreeRegressor}}\label{decisiontreeregressor}}

\begin{itemize}
\tightlist
\item
  Describe the surrogate model.
\item
  Use the surrogate as the model for optimization.
\end{itemize}

\hypertarget{randomforestregressor}{%
\subsection{\texorpdfstring{\texttt{RandomForestRegressor}}{RandomForestRegressor}}\label{randomforestregressor}}

\begin{itemize}
\tightlist
\item
  Describe the surrogate model.
\item
  Use the surrogate as the model for optimization.
\end{itemize}

\hypertarget{linear_model.linearregression}{%
\subsection{\texorpdfstring{\texttt{linear\_model.LinearRegression}}{linear\_model.LinearRegression}}\label{linear_model.linearregression}}

\begin{itemize}
\tightlist
\item
  Describe the surrogate model.
\item
  Use the surrogate as the model for optimization.
\end{itemize}

\hypertarget{linear_model.ridge}{%
\subsection{\texorpdfstring{\texttt{linear\_model.Ridge}}{linear\_model.Ridge}}\label{linear_model.ridge}}

\begin{itemize}
\tightlist
\item
  Describe the surrogate model.
\item
  Use the surrogate as the model for optimization.
\end{itemize}

\hypertarget{exercise-2}{%
\section{Exercise 2}\label{exercise-2}}

\begin{itemize}
\item
  Compare the performance of the five different surrogates on both
  objective functions:

  \begin{itemize}
  \tightlist
  \item
    \texttt{spotPython}'s internal Kriging
  \item
    \texttt{DecisionTreeRegressor}
  \item
    \texttt{RandomForestRegressor}
  \item
    \texttt{linear\_model.LinearRegression}
  \item
    \texttt{linear\_model.Ridge}
  \end{itemize}
\end{itemize}

\hypertarget{sec-scipy-optimizers}{%
\chapter{\texorpdfstring{Sequential Parameter Optimization: Using
\texttt{scipy}
Optimizers}{Sequential Parameter Optimization: Using scipy Optimizers}}\label{sec-scipy-optimizers}}

As a default optimizer, \texttt{spotPython} uses
\texttt{differential\_evolution} from the \texttt{scipy.optimize}
package. Alternatively, any other optimizer from the
\texttt{scipy.optimize} package can be used. This chapter describes how
different optimizers from the \texttt{scipy\ optimize} package can be
used on the surrogate. The optimization algorithms are available from
\url{https://docs.scipy.org/doc/scipy/reference/optimize.html}

\begin{Shaded}
\begin{Highlighting}[]
\ImportTok{import}\NormalTok{ numpy }\ImportTok{as}\NormalTok{ np}
\ImportTok{from}\NormalTok{ math }\ImportTok{import}\NormalTok{ inf}
\ImportTok{from}\NormalTok{ spotPython.fun.objectivefunctions }\ImportTok{import}\NormalTok{ analytical}
\ImportTok{from}\NormalTok{ spotPython.spot }\ImportTok{import}\NormalTok{ spot}
\ImportTok{from}\NormalTok{ scipy.optimize }\ImportTok{import}\NormalTok{ shgo}
\ImportTok{from}\NormalTok{ scipy.optimize }\ImportTok{import}\NormalTok{ direct}
\ImportTok{from}\NormalTok{ scipy.optimize }\ImportTok{import}\NormalTok{ differential\_evolution}
\ImportTok{from}\NormalTok{ scipy.optimize }\ImportTok{import}\NormalTok{ dual\_annealing}
\ImportTok{from}\NormalTok{ scipy.optimize }\ImportTok{import}\NormalTok{ basinhopping}
\end{Highlighting}
\end{Shaded}

\hypertarget{the-objective-function-branin-1}{%
\section{The Objective Function
Branin}\label{the-objective-function-branin-1}}

\begin{itemize}
\item
  The \texttt{spotPython} package provides several classes of objective
  functions.
\item
  We will use an analytical objective function, i.e., a function that
  can be described by a (closed) formula.
\item
  Here we will use the Branin function. The 2-dim Branin function is

  \[y = a * (x2 - b * x1**2 + c * x1 - r) ** 2 + s * (1 - t) * cos(x1) + s,\]
  where values of a, b, c, r, s and t are:
  \(a = 1, b = 5.1 / (4*pi**2), c = 5 / pi, r = 6, s = 10\) and
  \(t = 1 / (8*pi)\).
\item
  It has three global minima:

  \(f(x) = 0.397887\) at \((-\pi, 12.275)\), \((\pi, 2.275)\), and
  \((9.42478, 2.475)\).
\item
  Input Domain: This function is usually evaluated on the square x1 in
  {[}-5, 10{]} x x2 in {[}0, 15{]}.
\end{itemize}

\begin{Shaded}
\begin{Highlighting}[]
\ImportTok{from}\NormalTok{ spotPython.fun.objectivefunctions }\ImportTok{import}\NormalTok{ analytical}
\NormalTok{lower }\OperatorTok{=}\NormalTok{ np.array([}\OperatorTok{{-}}\DecValTok{5}\NormalTok{,}\OperatorTok{{-}}\DecValTok{0}\NormalTok{])}
\NormalTok{upper }\OperatorTok{=}\NormalTok{ np.array([}\DecValTok{10}\NormalTok{,}\DecValTok{15}\NormalTok{])}
\end{Highlighting}
\end{Shaded}

\begin{Shaded}
\begin{Highlighting}[]
\NormalTok{fun }\OperatorTok{=}\NormalTok{ analytical(seed}\OperatorTok{=}\DecValTok{123}\NormalTok{).fun\_branin}
\end{Highlighting}
\end{Shaded}

\hypertarget{the-optimizer}{%
\section{The Optimizer}\label{the-optimizer}}

\begin{itemize}
\item
  Differential Evalution from the \texttt{scikit.optimize} package, see
  \url{https://docs.scipy.org/doc/scipy/reference/generated/scipy.optimize.differential_evolution.html\#scipy.optimize.differential_evolution}
  is the default optimizer for the search on the surrogate.
\item
  Other optimiers that are available in \texttt{spotPython}:

  \begin{itemize}
  \tightlist
  \item
    \texttt{dual\_annealing}
  \item
    \texttt{direct}
  \item
    \texttt{shgo}
  \item
    \texttt{basinhopping}, see
    \url{https://docs.scipy.org/doc/scipy/reference/optimize.html\#global-optimization}.
  \end{itemize}
\item
  These can be selected as follows:

  \texttt{surrogate\_control\ =\ "model\_optimizer":\ differential\_evolution}
\item
  We will use \texttt{differential\_evolution}.
\item
  The optimizer can use \texttt{1000} evaluations. This value will be
  passed to the \texttt{differential\_evolution} method, which has the
  argument \texttt{maxiter} (int). It defines the maximum number of
  generations over which the entire differential evolution population is
  evolved, see
  \url{https://docs.scipy.org/doc/scipy/reference/generated/scipy.optimize.differential_evolution.html\#scipy.optimize.differential_evolution}
\end{itemize}

\begin{tcolorbox}[enhanced jigsaw, left=2mm, title=\textcolor{quarto-callout-note-color}{\faInfo}\hspace{0.5em}{TensorBoard}, bottomrule=.15mm, titlerule=0mm, breakable, rightrule=.15mm, toprule=.15mm, coltitle=black, colbacktitle=quarto-callout-note-color!10!white, leftrule=.75mm, arc=.35mm, colframe=quarto-callout-note-color-frame, bottomtitle=1mm, colback=white, opacitybacktitle=0.6, toptitle=1mm, opacityback=0]

Similar to the one-dimensional case, which was introduced in Section
Section~\ref{sec-visualizing-tensorboard-01}, we can use TensorBoard to
monitor the progress of the optimization. We will use the same code,
only the prefix is different:

\begin{Shaded}
\begin{Highlighting}[]
\ImportTok{from}\NormalTok{ spotPython.utils.}\BuiltInTok{file} \ImportTok{import}\NormalTok{ get\_experiment\_name}
\ImportTok{from}\NormalTok{ spotPython.utils.init }\ImportTok{import}\NormalTok{ fun\_control\_init}
\ImportTok{from}\NormalTok{ spotPython.utils.}\BuiltInTok{file} \ImportTok{import}\NormalTok{ get\_spot\_tensorboard\_path}

\NormalTok{PREFIX }\OperatorTok{=} \StringTok{"05\_DE\_"}
\NormalTok{experiment\_name }\OperatorTok{=}\NormalTok{ get\_experiment\_name(prefix}\OperatorTok{=}\NormalTok{PREFIX)}
\BuiltInTok{print}\NormalTok{(experiment\_name)}

\NormalTok{fun\_control }\OperatorTok{=}\NormalTok{ fun\_control\_init(}
\NormalTok{    spot\_tensorboard\_path}\OperatorTok{=}\NormalTok{get\_spot\_tensorboard\_path(experiment\_name))}
\end{Highlighting}
\end{Shaded}

\begin{verbatim}
05_DE__bartz09_2023-07-17_08-49-54
\end{verbatim}

\end{tcolorbox}

\begin{Shaded}
\begin{Highlighting}[]
\NormalTok{spot\_de }\OperatorTok{=}\NormalTok{ spot.Spot(fun}\OperatorTok{=}\NormalTok{fun,}
\NormalTok{                   lower }\OperatorTok{=}\NormalTok{ lower,}
\NormalTok{                   upper }\OperatorTok{=}\NormalTok{ upper,}
\NormalTok{                   fun\_evals }\OperatorTok{=} \DecValTok{20}\NormalTok{,}
\NormalTok{                   max\_time }\OperatorTok{=}\NormalTok{ inf,}
\NormalTok{                   seed}\OperatorTok{=}\DecValTok{125}\NormalTok{,}
\NormalTok{                   noise}\OperatorTok{=}\VariableTok{False}\NormalTok{,}
\NormalTok{                   show\_models}\OperatorTok{=} \VariableTok{False}\NormalTok{,}
\NormalTok{                   design\_control}\OperatorTok{=}\NormalTok{\{}\StringTok{"init\_size"}\NormalTok{: }\DecValTok{10}\NormalTok{\},}
\NormalTok{                   surrogate\_control}\OperatorTok{=}\NormalTok{\{}\StringTok{"n\_theta"}\NormalTok{: }\BuiltInTok{len}\NormalTok{(lower),}
                                      \StringTok{"model\_optimizer"}\NormalTok{: differential\_evolution,}
                                      \StringTok{"model\_fun\_evals"}\NormalTok{: }\DecValTok{1000}\NormalTok{,}
\NormalTok{                                      \},}
\NormalTok{                  fun\_control}\OperatorTok{=}\NormalTok{fun\_control)}
\NormalTok{spot\_de.run()}
\end{Highlighting}
\end{Shaded}

\begin{verbatim}
spotPython tuning: 5.213735995388665 [######----] 55.00% 
\end{verbatim}

\begin{verbatim}
spotPython tuning: 5.213735995388665 [######----] 60.00% 
\end{verbatim}

\begin{verbatim}
spotPython tuning: 2.5179080007735086 [######----] 65.00% 
\end{verbatim}

\begin{verbatim}
spotPython tuning: 1.0168713401682457 [#######---] 70.00% 
\end{verbatim}

\begin{verbatim}
spotPython tuning: 0.4160575412800043 [########--] 75.00% 
\end{verbatim}

\begin{verbatim}
spotPython tuning: 0.40966080781404557 [########--] 80.00% 
\end{verbatim}

\begin{verbatim}
spotPython tuning: 0.40966080781404557 [########--] 85.00% 
\end{verbatim}

\begin{verbatim}
spotPython tuning: 0.39989087044857285 [#########-] 90.00% 
\end{verbatim}

\begin{verbatim}
spotPython tuning: 0.3996741243343038 [##########] 95.00% 
\end{verbatim}

\begin{verbatim}
spotPython tuning: 0.39951958110619046 [##########] 100.00% Done...
\end{verbatim}

\begin{verbatim}
<spotPython.spot.spot.Spot at 0x2bcddf910>
\end{verbatim}

\hypertarget{tensorboard-7}{%
\subsection{TensorBoard}\label{tensorboard-7}}

Now we can start TensorBoard in the background with the following
command:

\begin{Shaded}
\begin{Highlighting}[]
\NormalTok{tensorboard {-}{-}logdir="./runs"}
\end{Highlighting}
\end{Shaded}

We can access the TensorBoard web server with the following URL:

\begin{Shaded}
\begin{Highlighting}[]
\NormalTok{http://localhost:6006/}
\end{Highlighting}
\end{Shaded}

The TensorBoard plot illustrates how \texttt{spotPython} can be used as
a microscope for the internal mechanisms of the surrogate-based
optimization process. Here, one important parameter, the learning rate
\(\theta\) of the Kriging surrogate is plotted against the number of
optimization steps.

\begin{figure}

{\centering \includegraphics[width=1\textwidth,height=\textheight]{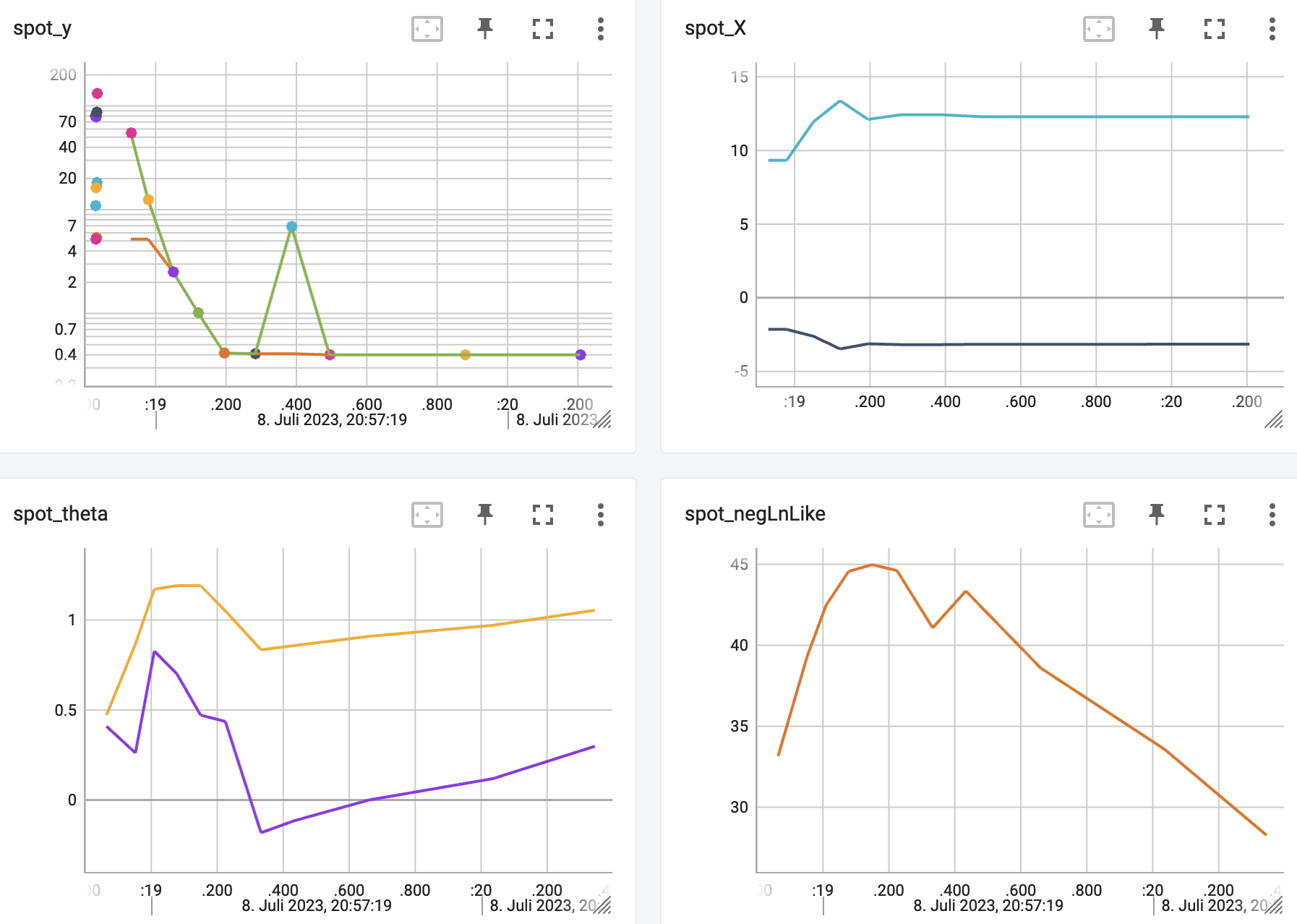}

}

\caption{TensorBoard visualization of the spotPython optimization
process and the surrogate model.}

\end{figure}

\hypertarget{print-the-results-2}{%
\section{Print the Results}\label{print-the-results-2}}

\begin{Shaded}
\begin{Highlighting}[]
\NormalTok{spot\_de.print\_results()}
\end{Highlighting}
\end{Shaded}

\begin{verbatim}
min y: 0.39951958110619046
x0: -3.1570201165683587
x1: 12.289980569430284
\end{verbatim}

\begin{verbatim}
[['x0', -3.1570201165683587], ['x1', 12.289980569430284]]
\end{verbatim}

\hypertarget{show-the-progress-1}{%
\section{Show the Progress}\label{show-the-progress-1}}

\begin{Shaded}
\begin{Highlighting}[]
\NormalTok{spot\_de.plot\_progress(log\_y}\OperatorTok{=}\VariableTok{True}\NormalTok{)}
\end{Highlighting}
\end{Shaded}

\begin{figure}[H]

{\centering \includegraphics{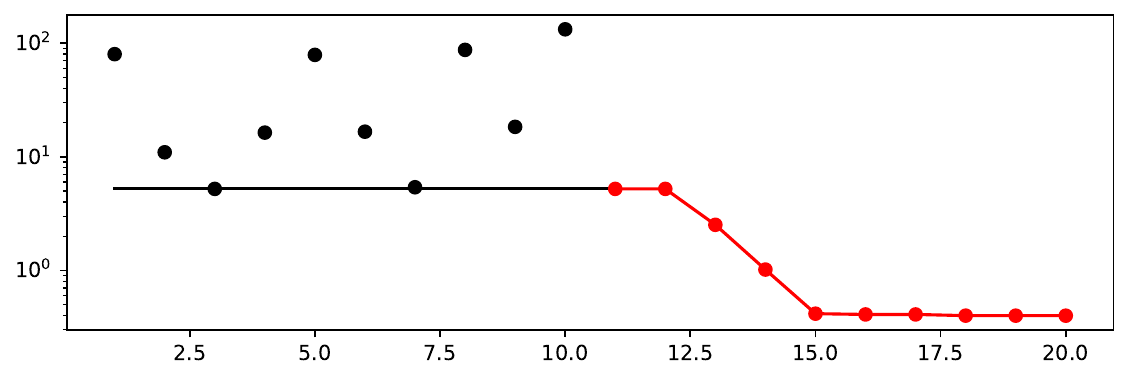}

}

\end{figure}

\begin{Shaded}
\begin{Highlighting}[]
\NormalTok{spot\_de.surrogate.plot()}
\end{Highlighting}
\end{Shaded}

\begin{figure}[H]

{\centering \includegraphics{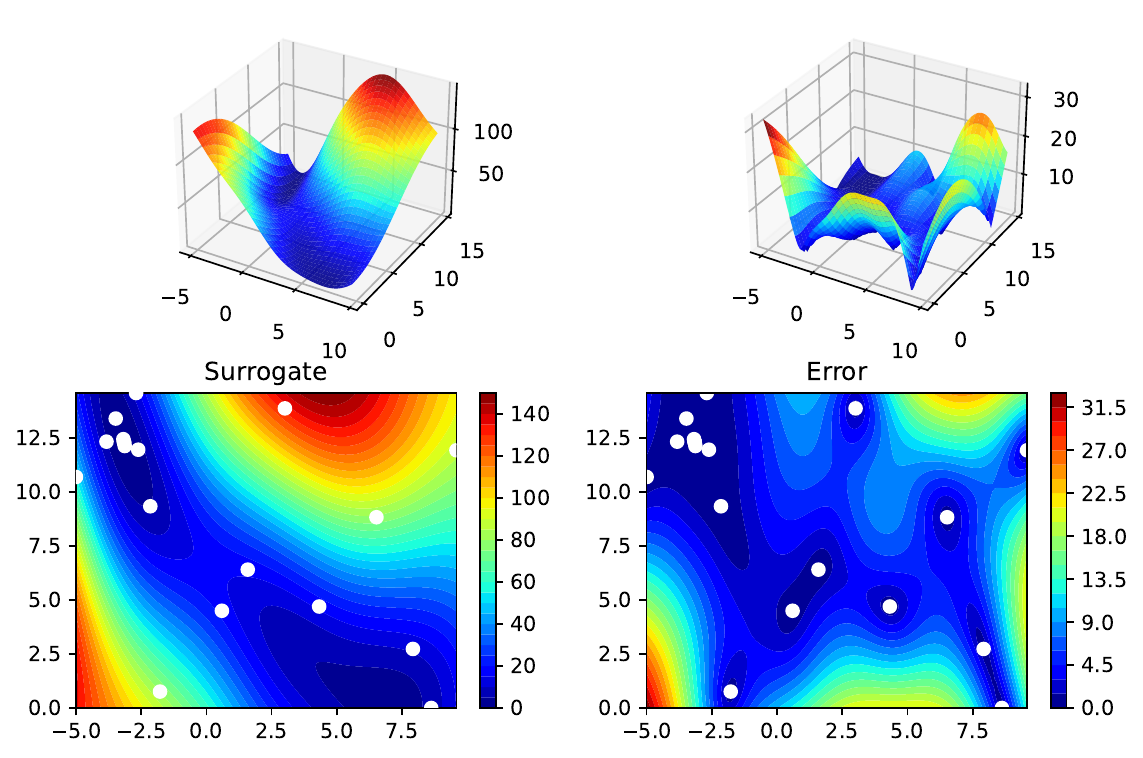}

}

\end{figure}

\hypertarget{exercises-3}{%
\section{Exercises}\label{exercises-3}}

\hypertarget{dual_annealing}{%
\subsection{\texorpdfstring{\texttt{dual\_annealing}}{dual\_annealing}}\label{dual_annealing}}

\begin{itemize}
\tightlist
\item
  Describe the optimization algorithm
\item
  Use the algorithm as an optimizer on the surrogate
\end{itemize}

\hypertarget{direct}{%
\subsection{\texorpdfstring{\texttt{direct}}{direct}}\label{direct}}

\begin{itemize}
\tightlist
\item
  Describe the optimization algorithm
\item
  Use the algorithm as an optimizer on the surrogate
\end{itemize}

\hypertarget{shgo}{%
\subsection{\texorpdfstring{\texttt{shgo}}{shgo}}\label{shgo}}

\begin{itemize}
\tightlist
\item
  Describe the optimization algorithm
\item
  Use the algorithm as an optimizer on the surrogate
\end{itemize}

\hypertarget{basinhopping}{%
\subsection{\texorpdfstring{\texttt{basinhopping}}{basinhopping}}\label{basinhopping}}

\begin{itemize}
\tightlist
\item
  Describe the optimization algorithm
\item
  Use the algorithm as an optimizer on the surrogate
\end{itemize}

\hypertarget{performance-comparison}{%
\subsection{Performance Comparison}\label{performance-comparison}}

Compare the performance and run time of the 5 different optimizers:

\begin{verbatim}
* `differential_evolution`
* `dual_annealing`
*  `direct`
* `shgo`
* `basinhopping`.
\end{verbatim}

The Branin function has three global minima:

\begin{itemize}
\tightlist
\item
  \(f(x) = 0.397887\) at

  \begin{itemize}
  \tightlist
  \item
    \((-\pi, 12.275)\),
  \item
    \((\pi, 2.275)\), and
  \item
    \((9.42478, 2.475)\).\\
  \end{itemize}
\item
  Which optima are found by the optimizers? Does the \texttt{seed}
  change this behavior?
\end{itemize}

\hypertarget{sec-gaussian-process-models}{%
\chapter{Sequential Parameter Optimization: Gaussian Process
Models}\label{sec-gaussian-process-models}}

This chapter analyzes differences between the \texttt{Kriging}
implementation in \texttt{spotPython} and the
\texttt{GaussianProcessRegressor} in \texttt{scikit-learn}.

\begin{Shaded}
\begin{Highlighting}[]
\ImportTok{import}\NormalTok{ numpy }\ImportTok{as}\NormalTok{ np}
\ImportTok{from}\NormalTok{ math }\ImportTok{import}\NormalTok{ inf}
\ImportTok{from}\NormalTok{ spotPython.fun.objectivefunctions }\ImportTok{import}\NormalTok{ analytical}
\ImportTok{from}\NormalTok{ spotPython.design.spacefilling }\ImportTok{import}\NormalTok{ spacefilling}
\ImportTok{from}\NormalTok{ spotPython.spot }\ImportTok{import}\NormalTok{ spot}
\ImportTok{from}\NormalTok{ spotPython.build.kriging }\ImportTok{import}\NormalTok{ Kriging}
\ImportTok{from}\NormalTok{ scipy.optimize }\ImportTok{import}\NormalTok{ shgo}
\ImportTok{from}\NormalTok{ scipy.optimize }\ImportTok{import}\NormalTok{ direct}
\ImportTok{from}\NormalTok{ scipy.optimize }\ImportTok{import}\NormalTok{ differential\_evolution}
\ImportTok{import}\NormalTok{ matplotlib.pyplot }\ImportTok{as}\NormalTok{ plt}
\ImportTok{import}\NormalTok{ math }\ImportTok{as}\NormalTok{ m}
\ImportTok{from}\NormalTok{ sklearn.gaussian\_process }\ImportTok{import}\NormalTok{ GaussianProcessRegressor}
\ImportTok{from}\NormalTok{ sklearn.gaussian\_process.kernels }\ImportTok{import}\NormalTok{ RBF}
\end{Highlighting}
\end{Shaded}

\hypertarget{gaussian-processes-regression-basic-introductory-scikit-learn-example}{%
\section{\texorpdfstring{Gaussian Processes Regression: Basic
Introductory \texttt{scikit-learn}
Example}{Gaussian Processes Regression: Basic Introductory scikit-learn Example}}\label{gaussian-processes-regression-basic-introductory-scikit-learn-example}}

\begin{itemize}
\item
  This is the example from
  \href{https://scikit-learn.org/stable/auto_examples/gaussian_process/plot_gpr_noisy_targets.html}{scikit-learn:
  https://scikit-learn.org/stable/auto\_examples/gaussian\_process/plot\_gpr\_noisy\_targets.html}
\item
  After fitting our model, we see that the hyperparameters of the kernel
  have been optimized.
\item
  Now, we will use our kernel to compute the mean prediction of the full
  dataset and plot the 95\% confidence interval.
\end{itemize}

\hypertarget{train-and-test-data}{%
\subsection{Train and Test Data}\label{train-and-test-data}}

\begin{Shaded}
\begin{Highlighting}[]
\NormalTok{X }\OperatorTok{=}\NormalTok{ np.linspace(start}\OperatorTok{=}\DecValTok{0}\NormalTok{, stop}\OperatorTok{=}\DecValTok{10}\NormalTok{, num}\OperatorTok{=}\DecValTok{1\_000}\NormalTok{).reshape(}\OperatorTok{{-}}\DecValTok{1}\NormalTok{, }\DecValTok{1}\NormalTok{)}
\NormalTok{y }\OperatorTok{=}\NormalTok{ np.squeeze(X }\OperatorTok{*}\NormalTok{ np.sin(X))}
\NormalTok{rng }\OperatorTok{=}\NormalTok{ np.random.RandomState(}\DecValTok{1}\NormalTok{)}
\NormalTok{training\_indices }\OperatorTok{=}\NormalTok{ rng.choice(np.arange(y.size), size}\OperatorTok{=}\DecValTok{6}\NormalTok{, replace}\OperatorTok{=}\VariableTok{False}\NormalTok{)}
\NormalTok{X\_train, y\_train }\OperatorTok{=}\NormalTok{ X[training\_indices], y[training\_indices]}
\end{Highlighting}
\end{Shaded}

\hypertarget{building-the-surrogate-with-sklearn}{%
\subsection{\texorpdfstring{Building the Surrogate With
\texttt{Sklearn}}{Building the Surrogate With Sklearn}}\label{building-the-surrogate-with-sklearn}}

\begin{itemize}
\tightlist
\item
  The model building with \texttt{sklearn} consisits of three steps:

  \begin{enumerate}
  \def\labelenumi{\arabic{enumi}.}
  \tightlist
  \item
    Instantiating the model, then
  \item
    fitting the model (using \texttt{fit}), and
  \item
    making predictions (using \texttt{predict})
  \end{enumerate}
\end{itemize}

\begin{Shaded}
\begin{Highlighting}[]
\NormalTok{kernel }\OperatorTok{=} \DecValTok{1} \OperatorTok{*}\NormalTok{ RBF(length\_scale}\OperatorTok{=}\FloatTok{1.0}\NormalTok{, length\_scale\_bounds}\OperatorTok{=}\NormalTok{(}\FloatTok{1e{-}2}\NormalTok{, }\FloatTok{1e2}\NormalTok{))}
\NormalTok{gaussian\_process }\OperatorTok{=}\NormalTok{ GaussianProcessRegressor(kernel}\OperatorTok{=}\NormalTok{kernel, n\_restarts\_optimizer}\OperatorTok{=}\DecValTok{9}\NormalTok{)}
\NormalTok{gaussian\_process.fit(X\_train, y\_train)}
\NormalTok{mean\_prediction, std\_prediction }\OperatorTok{=}\NormalTok{ gaussian\_process.predict(X, return\_std}\OperatorTok{=}\VariableTok{True}\NormalTok{)}
\end{Highlighting}
\end{Shaded}

\hypertarget{plotting-the-sklearnmodel}{%
\subsection{\texorpdfstring{Plotting the
\texttt{Sklearn}Model}{Plotting the SklearnModel}}\label{plotting-the-sklearnmodel}}

\begin{Shaded}
\begin{Highlighting}[]
\NormalTok{plt.plot(X, y, label}\OperatorTok{=}\VerbatimStringTok{r"$f(x) = x \textbackslash{}sin(x)$"}\NormalTok{, linestyle}\OperatorTok{=}\StringTok{"dotted"}\NormalTok{)}
\NormalTok{plt.scatter(X\_train, y\_train, label}\OperatorTok{=}\StringTok{"Observations"}\NormalTok{)}
\NormalTok{plt.plot(X, mean\_prediction, label}\OperatorTok{=}\StringTok{"Mean prediction"}\NormalTok{)}
\NormalTok{plt.fill\_between(}
\NormalTok{    X.ravel(),}
\NormalTok{    mean\_prediction }\OperatorTok{{-}} \FloatTok{1.96} \OperatorTok{*}\NormalTok{ std\_prediction,}
\NormalTok{    mean\_prediction }\OperatorTok{+} \FloatTok{1.96} \OperatorTok{*}\NormalTok{ std\_prediction,}
\NormalTok{    alpha}\OperatorTok{=}\FloatTok{0.5}\NormalTok{,}
\NormalTok{    label}\OperatorTok{=}\VerbatimStringTok{r"95}\SpecialCharTok{\% c}\VerbatimStringTok{onfidence interval"}\NormalTok{,}
\NormalTok{)}
\NormalTok{plt.legend()}
\NormalTok{plt.xlabel(}\StringTok{"$x$"}\NormalTok{)}
\NormalTok{plt.ylabel(}\StringTok{"$f(x)$"}\NormalTok{)}
\NormalTok{\_ }\OperatorTok{=}\NormalTok{ plt.title(}\StringTok{"sk{-}learn Version: Gaussian process regression on noise{-}free dataset"}\NormalTok{)}
\end{Highlighting}
\end{Shaded}

\begin{figure}[H]

{\centering \includegraphics{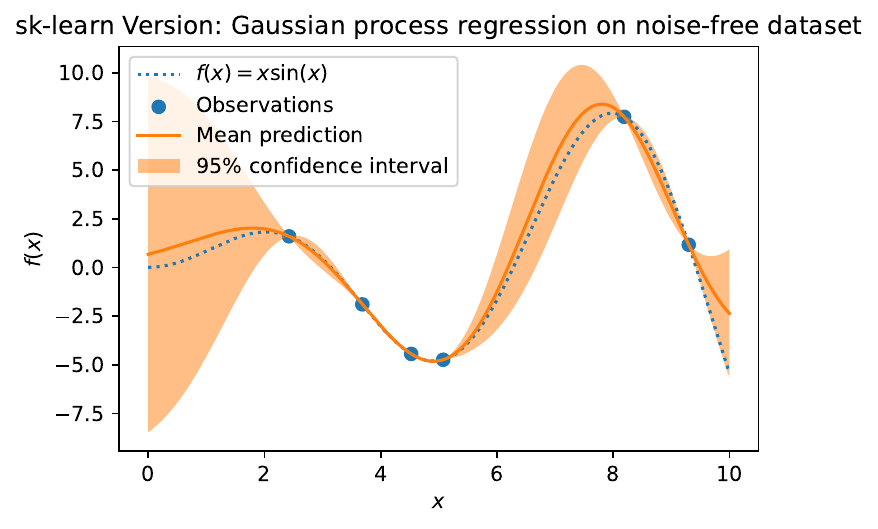}

}

\end{figure}

\hypertarget{the-spotpython-version}{%
\subsection{\texorpdfstring{The \texttt{spotPython}
Version}{The spotPython Version}}\label{the-spotpython-version}}

\begin{itemize}
\tightlist
\item
  The \texttt{spotPython} version is very similar:

  \begin{enumerate}
  \def\labelenumi{\arabic{enumi}.}
  \tightlist
  \item
    Instantiating the model, then
  \item
    fitting the model and
  \item
    making predictions (using \texttt{predict}).
  \end{enumerate}
\end{itemize}

\begin{Shaded}
\begin{Highlighting}[]
\NormalTok{S }\OperatorTok{=}\NormalTok{ Kriging(name}\OperatorTok{=}\StringTok{\textquotesingle{}kriging\textquotesingle{}}\NormalTok{,  seed}\OperatorTok{=}\DecValTok{123}\NormalTok{, log\_level}\OperatorTok{=}\DecValTok{50}\NormalTok{, cod\_type}\OperatorTok{=}\StringTok{"norm"}\NormalTok{)}
\NormalTok{S.fit(X\_train, y\_train)}
\NormalTok{S\_mean\_prediction, S\_std\_prediction, S\_ei }\OperatorTok{=}\NormalTok{ S.predict(X, return\_val}\OperatorTok{=}\StringTok{"all"}\NormalTok{)}
\end{Highlighting}
\end{Shaded}

\begin{Shaded}
\begin{Highlighting}[]
\NormalTok{plt.plot(X, y, label}\OperatorTok{=}\VerbatimStringTok{r"$f(x) = x \textbackslash{}sin(x)$"}\NormalTok{, linestyle}\OperatorTok{=}\StringTok{"dotted"}\NormalTok{)}
\NormalTok{plt.scatter(X\_train, y\_train, label}\OperatorTok{=}\StringTok{"Observations"}\NormalTok{)}
\NormalTok{plt.plot(X, S\_mean\_prediction, label}\OperatorTok{=}\StringTok{"Mean prediction"}\NormalTok{)}
\NormalTok{plt.fill\_between(}
\NormalTok{    X.ravel(),}
\NormalTok{    S\_mean\_prediction }\OperatorTok{{-}} \FloatTok{1.96} \OperatorTok{*}\NormalTok{ S\_std\_prediction,}
\NormalTok{    S\_mean\_prediction }\OperatorTok{+} \FloatTok{1.96} \OperatorTok{*}\NormalTok{ S\_std\_prediction,}
\NormalTok{    alpha}\OperatorTok{=}\FloatTok{0.5}\NormalTok{,}
\NormalTok{    label}\OperatorTok{=}\VerbatimStringTok{r"95}\SpecialCharTok{\% c}\VerbatimStringTok{onfidence interval"}\NormalTok{,}
\NormalTok{)}
\NormalTok{plt.legend()}
\NormalTok{plt.xlabel(}\StringTok{"$x$"}\NormalTok{)}
\NormalTok{plt.ylabel(}\StringTok{"$f(x)$"}\NormalTok{)}
\NormalTok{\_ }\OperatorTok{=}\NormalTok{ plt.title(}\StringTok{"spotPython Version: Gaussian process regression on noise{-}free dataset"}\NormalTok{)}
\end{Highlighting}
\end{Shaded}

\begin{figure}[H]

{\centering \includegraphics{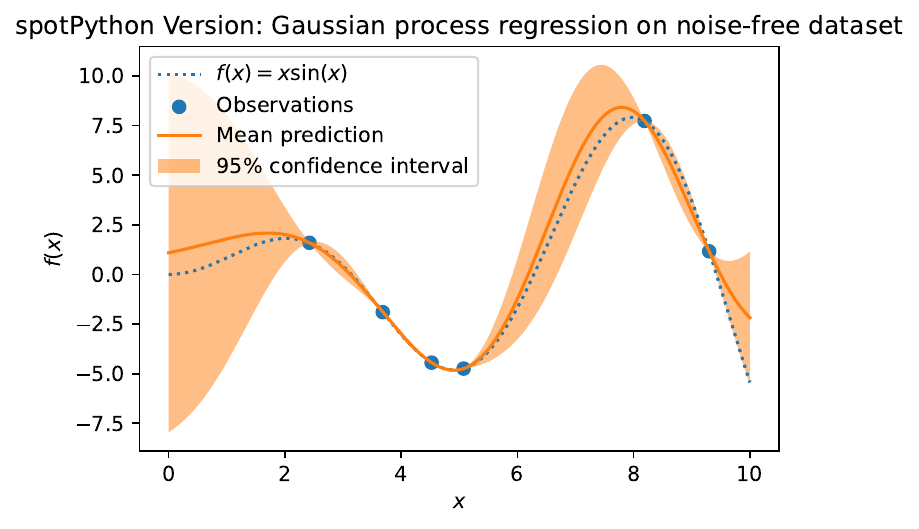}

}

\end{figure}

\hypertarget{visualizing-the-differences-between-the-spotpython-and-the-sklearn-model-fits}{%
\subsection{\texorpdfstring{Visualizing the Differences Between the
\texttt{spotPython} and the \texttt{sklearn} Model
Fits}{Visualizing the Differences Between the spotPython and the sklearn Model Fits}}\label{visualizing-the-differences-between-the-spotpython-and-the-sklearn-model-fits}}

\begin{Shaded}
\begin{Highlighting}[]
\NormalTok{plt.plot(X, y, label}\OperatorTok{=}\VerbatimStringTok{r"$f(x) = x \textbackslash{}sin(x)$"}\NormalTok{, linestyle}\OperatorTok{=}\StringTok{"dotted"}\NormalTok{)}
\NormalTok{plt.scatter(X\_train, y\_train, label}\OperatorTok{=}\StringTok{"Observations"}\NormalTok{)}
\NormalTok{plt.plot(X, S\_mean\_prediction, label}\OperatorTok{=}\StringTok{"spotPython Mean prediction"}\NormalTok{)}
\NormalTok{plt.plot(X, mean\_prediction, label}\OperatorTok{=}\StringTok{"Sklearn Mean Prediction"}\NormalTok{)}
\NormalTok{plt.legend()}
\NormalTok{plt.xlabel(}\StringTok{"$x$"}\NormalTok{)}
\NormalTok{plt.ylabel(}\StringTok{"$f(x)$"}\NormalTok{)}
\NormalTok{\_ }\OperatorTok{=}\NormalTok{ plt.title(}\StringTok{"Comparing Mean Predictions"}\NormalTok{)}
\end{Highlighting}
\end{Shaded}

\begin{figure}[H]

{\centering \includegraphics{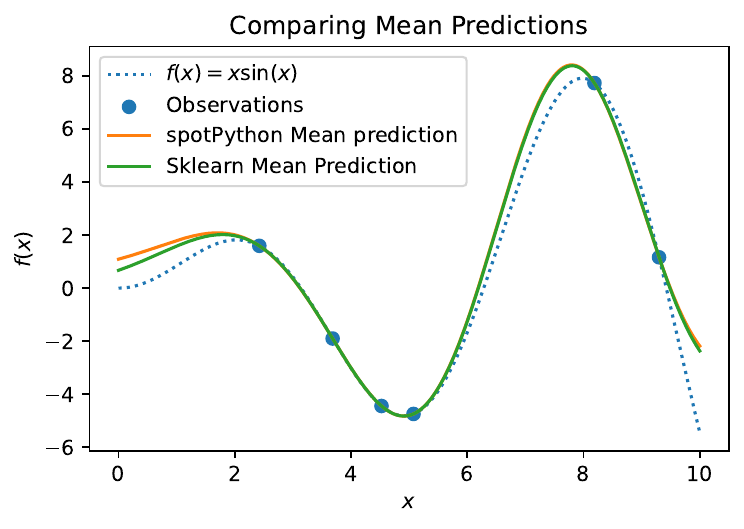}

}

\end{figure}

\hypertarget{exercises-4}{%
\section{Exercises}\label{exercises-4}}

\hypertarget{schonlau-example-function}{%
\subsection{\texorpdfstring{\texttt{Schonlau\ Example\ Function}}{Schonlau Example Function}}\label{schonlau-example-function}}

\begin{itemize}
\tightlist
\item
  The Schonlau Example Function is based on sample points only (there is
  no analytical function description available):
\end{itemize}

\begin{Shaded}
\begin{Highlighting}[]
\NormalTok{X }\OperatorTok{=}\NormalTok{ np.linspace(start}\OperatorTok{=}\DecValTok{0}\NormalTok{, stop}\OperatorTok{=}\DecValTok{13}\NormalTok{, num}\OperatorTok{=}\DecValTok{1\_000}\NormalTok{).reshape(}\OperatorTok{{-}}\DecValTok{1}\NormalTok{, }\DecValTok{1}\NormalTok{)}
\NormalTok{X\_train }\OperatorTok{=}\NormalTok{ np.array([}\FloatTok{1.}\NormalTok{, }\FloatTok{2.}\NormalTok{, }\FloatTok{3.}\NormalTok{, }\FloatTok{4.}\NormalTok{, }\FloatTok{12.}\NormalTok{]).reshape(}\OperatorTok{{-}}\DecValTok{1}\NormalTok{,}\DecValTok{1}\NormalTok{)}
\NormalTok{y\_train }\OperatorTok{=}\NormalTok{ np.array([}\FloatTok{0.}\NormalTok{, }\OperatorTok{{-}}\FloatTok{1.75}\NormalTok{, }\OperatorTok{{-}}\DecValTok{2}\NormalTok{, }\OperatorTok{{-}}\FloatTok{0.5}\NormalTok{, }\FloatTok{5.}\NormalTok{])}
\end{Highlighting}
\end{Shaded}

\begin{itemize}
\tightlist
\item
  Describe the function.
\item
  Compare the two models that were build using the \texttt{spotPython}
  and the \texttt{sklearn} surrogate.
\item
  Note: Since there is no analytical function available, you might be
  interested in adding some points and describe the effects.
\end{itemize}

\hypertarget{forrester-example-function}{%
\subsection{\texorpdfstring{\texttt{Forrester\ Example\ Function}}{Forrester Example Function}}\label{forrester-example-function}}

\begin{itemize}
\item
  The Forrester Example Function is defined as follows:

  \texttt{f(x)\ =\ (6x-\ 2)\^{}2\ sin(12x-4)\ for\ x\ in\ {[}0,1{]}.}
\item
  Data points are generated as follows:
\end{itemize}

\begin{Shaded}
\begin{Highlighting}[]
\NormalTok{X }\OperatorTok{=}\NormalTok{ np.linspace(start}\OperatorTok{={-}}\FloatTok{0.5}\NormalTok{, stop}\OperatorTok{=}\FloatTok{1.5}\NormalTok{, num}\OperatorTok{=}\DecValTok{1\_000}\NormalTok{).reshape(}\OperatorTok{{-}}\DecValTok{1}\NormalTok{, }\DecValTok{1}\NormalTok{)}
\NormalTok{X\_train }\OperatorTok{=}\NormalTok{ np.array([}\FloatTok{0.0}\NormalTok{, }\FloatTok{0.175}\NormalTok{, }\FloatTok{0.225}\NormalTok{, }\FloatTok{0.3}\NormalTok{, }\FloatTok{0.35}\NormalTok{, }\FloatTok{0.375}\NormalTok{, }\FloatTok{0.5}\NormalTok{,}\DecValTok{1}\NormalTok{]).reshape(}\OperatorTok{{-}}\DecValTok{1}\NormalTok{,}\DecValTok{1}\NormalTok{)}
\NormalTok{fun }\OperatorTok{=}\NormalTok{ analytical().fun\_forrester}
\NormalTok{fun\_control }\OperatorTok{=}\NormalTok{ \{}\StringTok{"sigma"}\NormalTok{: }\FloatTok{0.1}\NormalTok{,}
               \StringTok{"seed"}\NormalTok{: }\DecValTok{123}\NormalTok{\}}
\NormalTok{y }\OperatorTok{=}\NormalTok{ fun(X, fun\_control}\OperatorTok{=}\NormalTok{fun\_control)}
\NormalTok{y\_train }\OperatorTok{=}\NormalTok{ fun(X\_train, fun\_control}\OperatorTok{=}\NormalTok{fun\_control)}
\end{Highlighting}
\end{Shaded}

\begin{itemize}
\tightlist
\item
  Describe the function.
\item
  Compare the two models that were build using the \texttt{spotPython}
  and the \texttt{sklearn} surrogate.
\item
  Note: Modify the noise level (\texttt{"sigma"}), e.g., use a value of
  \texttt{0.2}, and compare the two models.
\end{itemize}

\begin{Shaded}
\begin{Highlighting}[]
\NormalTok{fun\_control }\OperatorTok{=}\NormalTok{ \{}\StringTok{"sigma"}\NormalTok{: }\FloatTok{0.2}\NormalTok{\}}
\end{Highlighting}
\end{Shaded}

\hypertarget{fun_runge-function-1-dim}{%
\subsection{\texorpdfstring{\texttt{fun\_runge\ Function\ (1-dim)}}{fun\_runge Function (1-dim)}}\label{fun_runge-function-1-dim}}

\begin{itemize}
\item
  The Runge function is defined as follows:

  \texttt{f(x)\ =\ 1/\ (1\ +\ sum(x\_i))\^{}2}
\item
  Data points are generated as follows:
\end{itemize}

\begin{Shaded}
\begin{Highlighting}[]
\NormalTok{gen }\OperatorTok{=}\NormalTok{ spacefilling(}\DecValTok{1}\NormalTok{)}
\NormalTok{rng }\OperatorTok{=}\NormalTok{ np.random.RandomState(}\DecValTok{1}\NormalTok{)}
\NormalTok{lower }\OperatorTok{=}\NormalTok{ np.array([}\OperatorTok{{-}}\DecValTok{10}\NormalTok{])}
\NormalTok{upper }\OperatorTok{=}\NormalTok{ np.array([}\DecValTok{10}\NormalTok{])}
\NormalTok{fun }\OperatorTok{=}\NormalTok{ analytical().fun\_runge}
\NormalTok{fun\_control }\OperatorTok{=}\NormalTok{ \{}\StringTok{"sigma"}\NormalTok{: }\FloatTok{0.025}\NormalTok{,}
               \StringTok{"seed"}\NormalTok{: }\DecValTok{123}\NormalTok{\}}
\NormalTok{X\_train }\OperatorTok{=}\NormalTok{ gen.scipy\_lhd(}\DecValTok{10}\NormalTok{, lower}\OperatorTok{=}\NormalTok{lower, upper }\OperatorTok{=}\NormalTok{ upper).reshape(}\OperatorTok{{-}}\DecValTok{1}\NormalTok{,}\DecValTok{1}\NormalTok{)}
\NormalTok{y\_train }\OperatorTok{=}\NormalTok{ fun(X, fun\_control}\OperatorTok{=}\NormalTok{fun\_control)}
\NormalTok{X }\OperatorTok{=}\NormalTok{ np.linspace(start}\OperatorTok{={-}}\DecValTok{13}\NormalTok{, stop}\OperatorTok{=}\DecValTok{13}\NormalTok{, num}\OperatorTok{=}\DecValTok{1000}\NormalTok{).reshape(}\OperatorTok{{-}}\DecValTok{1}\NormalTok{, }\DecValTok{1}\NormalTok{)}
\NormalTok{y }\OperatorTok{=}\NormalTok{ fun(X, fun\_control}\OperatorTok{=}\NormalTok{fun\_control)}
\end{Highlighting}
\end{Shaded}

\begin{itemize}
\tightlist
\item
  Describe the function.
\item
  Compare the two models that were build using the \texttt{spotPython}
  and the \texttt{sklearn} surrogate.
\item
  Note: Modify the noise level (\texttt{"sigma"}), e.g., use a value of
  \texttt{0.05}, and compare the two models.
\end{itemize}

\begin{Shaded}
\begin{Highlighting}[]
\NormalTok{fun\_control }\OperatorTok{=}\NormalTok{ \{}\StringTok{"sigma"}\NormalTok{: }\FloatTok{0.5}\NormalTok{\}}
\end{Highlighting}
\end{Shaded}

\hypertarget{fun_cubed-1-dim}{%
\subsection{\texorpdfstring{\texttt{fun\_cubed\ (1-dim)}}{fun\_cubed (1-dim)}}\label{fun_cubed-1-dim}}

\begin{itemize}
\item
  The Cubed function is defined as follows:

  \texttt{np.sum(X{[}i{]}**\ 3)}
\item
  Data points are generated as follows:
\end{itemize}

\begin{Shaded}
\begin{Highlighting}[]
\NormalTok{gen }\OperatorTok{=}\NormalTok{ spacefilling(}\DecValTok{1}\NormalTok{)}
\NormalTok{rng }\OperatorTok{=}\NormalTok{ np.random.RandomState(}\DecValTok{1}\NormalTok{)}
\NormalTok{lower }\OperatorTok{=}\NormalTok{ np.array([}\OperatorTok{{-}}\DecValTok{10}\NormalTok{])}
\NormalTok{upper }\OperatorTok{=}\NormalTok{ np.array([}\DecValTok{10}\NormalTok{])}
\NormalTok{fun }\OperatorTok{=}\NormalTok{ analytical().fun\_cubed}
\NormalTok{fun\_control }\OperatorTok{=}\NormalTok{ \{}\StringTok{"sigma"}\NormalTok{: }\FloatTok{0.025}\NormalTok{,}
               \StringTok{"seed"}\NormalTok{: }\DecValTok{123}\NormalTok{\}}
\NormalTok{X\_train }\OperatorTok{=}\NormalTok{ gen.scipy\_lhd(}\DecValTok{10}\NormalTok{, lower}\OperatorTok{=}\NormalTok{lower, upper }\OperatorTok{=}\NormalTok{ upper).reshape(}\OperatorTok{{-}}\DecValTok{1}\NormalTok{,}\DecValTok{1}\NormalTok{)}
\NormalTok{y\_train }\OperatorTok{=}\NormalTok{ fun(X, fun\_control}\OperatorTok{=}\NormalTok{fun\_control)}
\NormalTok{X }\OperatorTok{=}\NormalTok{ np.linspace(start}\OperatorTok{={-}}\DecValTok{13}\NormalTok{, stop}\OperatorTok{=}\DecValTok{13}\NormalTok{, num}\OperatorTok{=}\DecValTok{1000}\NormalTok{).reshape(}\OperatorTok{{-}}\DecValTok{1}\NormalTok{, }\DecValTok{1}\NormalTok{)}
\NormalTok{y }\OperatorTok{=}\NormalTok{ fun(X, fun\_control}\OperatorTok{=}\NormalTok{fun\_control)}
\end{Highlighting}
\end{Shaded}

\begin{itemize}
\tightlist
\item
  Describe the function.
\item
  Compare the two models that were build using the \texttt{spotPython}
  and the \texttt{sklearn} surrogate.
\item
  Note: Modify the noise level (\texttt{"sigma"}), e.g., use a value of
  \texttt{0.05}, and compare the two models.
\end{itemize}

\begin{Shaded}
\begin{Highlighting}[]
\NormalTok{fun\_control }\OperatorTok{=}\NormalTok{ \{}\StringTok{"sigma"}\NormalTok{: }\FloatTok{0.05}\NormalTok{\}}
\end{Highlighting}
\end{Shaded}

\hypertarget{the-effect-of-noise}{%
\subsection{The Effect of Noise}\label{the-effect-of-noise}}

How does the behavior of the \texttt{spotPython} fit changes when the
argument \texttt{noise} is set to \texttt{True}, i.e.,

\texttt{S\ =\ Kriging(name=\textquotesingle{}kriging\textquotesingle{},\ \ seed=123,\ n\_theta=1,\ noise=True)}

is used?

\hypertarget{sec-expected-improvement}{%
\chapter{Expected Improvement}\label{sec-expected-improvement}}

This chapter describes, analyzes, and compares different infill
criterion. An infill criterion defines how the next point \(x_{n+1}\) is
selected from the surrogate model \(S\). Expected improvement is a
popular infill criterion in Bayesian optimization.

\hypertarget{example-spot-and-the-1-dim-sphere-function}{%
\section{\texorpdfstring{Example: \texttt{Spot} and the 1-dim Sphere
Function}{Example: Spot and the 1-dim Sphere Function}}\label{example-spot-and-the-1-dim-sphere-function}}

\begin{Shaded}
\begin{Highlighting}[]
\ImportTok{import}\NormalTok{ numpy }\ImportTok{as}\NormalTok{ np}
\ImportTok{from}\NormalTok{ math }\ImportTok{import}\NormalTok{ inf}
\ImportTok{from}\NormalTok{ spotPython.fun.objectivefunctions }\ImportTok{import}\NormalTok{ analytical}
\ImportTok{from}\NormalTok{ spotPython.spot }\ImportTok{import}\NormalTok{ spot}
\ImportTok{import}\NormalTok{ matplotlib.pyplot }\ImportTok{as}\NormalTok{ plt}
\end{Highlighting}
\end{Shaded}

\hypertarget{the-objective-function-1-dim-sphere}{%
\subsection{The Objective Function: 1-dim
Sphere}\label{the-objective-function-1-dim-sphere}}

\begin{itemize}
\tightlist
\item
  The \texttt{spotPython} package provides several classes of objective
  functions.
\item
  We will use an analytical objective function, i.e., a function that
  can be described by a (closed) formula: \[f(x) = x^2 \]
\end{itemize}

\begin{Shaded}
\begin{Highlighting}[]
\NormalTok{fun }\OperatorTok{=}\NormalTok{ analytical().fun\_sphere}
\end{Highlighting}
\end{Shaded}

\begin{Shaded}
\begin{Highlighting}[]
\NormalTok{fun }\OperatorTok{=}\NormalTok{ analytical().fun\_sphere}
\end{Highlighting}
\end{Shaded}

\begin{itemize}
\tightlist
\item
  The size of the \texttt{lower} bound vector determines the problem
  dimension.
\item
  Here we will use \texttt{np.array({[}-1{]})}, i.e., a one-dim
  function.
\end{itemize}

\begin{tcolorbox}[enhanced jigsaw, left=2mm, title=\textcolor{quarto-callout-note-color}{\faInfo}\hspace{0.5em}{TensorBoard}, bottomrule=.15mm, titlerule=0mm, breakable, rightrule=.15mm, toprule=.15mm, coltitle=black, colbacktitle=quarto-callout-note-color!10!white, leftrule=.75mm, arc=.35mm, colframe=quarto-callout-note-color-frame, bottomtitle=1mm, colback=white, opacitybacktitle=0.6, toptitle=1mm, opacityback=0]

Similar to the one-dimensional case, which was introduced in Section
Section~\ref{sec-visualizing-tensorboard-01}, we can use TensorBoard to
monitor the progress of the optimization. We will use the same code,
only the prefix is different:

\begin{Shaded}
\begin{Highlighting}[]
\ImportTok{from}\NormalTok{ spotPython.utils.}\BuiltInTok{file} \ImportTok{import}\NormalTok{ get\_experiment\_name}
\ImportTok{from}\NormalTok{ spotPython.utils.init }\ImportTok{import}\NormalTok{ fun\_control\_init}
\ImportTok{from}\NormalTok{ spotPython.utils.}\BuiltInTok{file} \ImportTok{import}\NormalTok{ get\_spot\_tensorboard\_path}

\NormalTok{PREFIX }\OperatorTok{=} \StringTok{"07\_Y"}
\NormalTok{experiment\_name }\OperatorTok{=}\NormalTok{ get\_experiment\_name(prefix}\OperatorTok{=}\NormalTok{PREFIX)}
\BuiltInTok{print}\NormalTok{(experiment\_name)}

\NormalTok{fun\_control }\OperatorTok{=}\NormalTok{ fun\_control\_init(}
\NormalTok{    spot\_tensorboard\_path}\OperatorTok{=}\NormalTok{get\_spot\_tensorboard\_path(experiment\_name),}
\NormalTok{    sigma}\OperatorTok{=}\DecValTok{0}\NormalTok{,}
\NormalTok{    seed}\OperatorTok{=}\DecValTok{123}\NormalTok{,)}
\end{Highlighting}
\end{Shaded}

\begin{verbatim}
07_Y_bartz09_2023-07-17_08-50-45
\end{verbatim}

\end{tcolorbox}

\begin{Shaded}
\begin{Highlighting}[]
\NormalTok{spot\_1 }\OperatorTok{=}\NormalTok{ spot.Spot(fun}\OperatorTok{=}\NormalTok{fun,}
\NormalTok{                   fun\_evals }\OperatorTok{=} \DecValTok{25}\NormalTok{,}
\NormalTok{                   lower }\OperatorTok{=}\NormalTok{ np.array([}\OperatorTok{{-}}\DecValTok{1}\NormalTok{]),}
\NormalTok{                   upper }\OperatorTok{=}\NormalTok{ np.array([}\DecValTok{1}\NormalTok{]),}
\NormalTok{                   design\_control}\OperatorTok{=}\NormalTok{\{}\StringTok{"init\_size"}\NormalTok{: }\DecValTok{10}\NormalTok{\},}
\NormalTok{                   tolerance\_x }\OperatorTok{=}\NormalTok{ np.sqrt(np.spacing(}\DecValTok{1}\NormalTok{)),}
\NormalTok{                   fun\_control }\OperatorTok{=}\NormalTok{ fun\_control,)}

\NormalTok{spot\_1.run()}
\end{Highlighting}
\end{Shaded}

\begin{verbatim}
spotPython tuning: 7.263311682641849e-09 [####------] 44.00% 
\end{verbatim}

\begin{verbatim}
spotPython tuning: 7.263311682641849e-09 [#####-----] 48.00% 
\end{verbatim}

\begin{verbatim}
spotPython tuning: 7.263311682641849e-09 [#####-----] 52.00% 
\end{verbatim}

\begin{verbatim}
spotPython tuning: 7.263311682641849e-09 [######----] 56.00% 
\end{verbatim}

\begin{verbatim}
spotPython tuning: 3.696886711914087e-10 [######----] 60.00% 
\end{verbatim}

\begin{verbatim}
spotPython tuning: 3.696886711914087e-10 [######----] 64.00% 
\end{verbatim}

\begin{verbatim}
spotPython tuning: 3.696886711914087e-10 [#######---] 68.00% 
\end{verbatim}

\begin{verbatim}
spotPython tuning: 3.696886711914087e-10 [#######---] 72.00% 
\end{verbatim}

\begin{verbatim}
spotPython tuning: 3.696886711914087e-10 [########--] 76.00% 
\end{verbatim}

\begin{verbatim}
spotPython tuning: 3.696886711914087e-10 [########--] 80.00% 
\end{verbatim}

\begin{verbatim}
spotPython tuning: 3.696886711914087e-10 [########--] 84.00% 
\end{verbatim}

\begin{verbatim}
spotPython tuning: 3.696886711914087e-10 [#########-] 88.00% 
\end{verbatim}

\begin{verbatim}
spotPython tuning: 1.3792745942664307e-11 [#########-] 92.00% 
\end{verbatim}

\begin{verbatim}
spotPython tuning: 1.3792745942664307e-11 [##########] 96.00% 
\end{verbatim}

\begin{verbatim}
spotPython tuning: 1.3792745942664307e-11 [##########] 100.00% Done...
\end{verbatim}

\begin{verbatim}
<spotPython.spot.spot.Spot at 0x103347430>
\end{verbatim}

\hypertarget{results-3}{%
\subsection{Results}\label{results-3}}

\begin{Shaded}
\begin{Highlighting}[]
\NormalTok{spot\_1.print\_results()}
\end{Highlighting}
\end{Shaded}

\begin{verbatim}
min y: 1.3792745942664307e-11
x0: 3.7138586325632142e-06
\end{verbatim}

\begin{verbatim}
[['x0', 3.7138586325632142e-06]]
\end{verbatim}

\begin{Shaded}
\begin{Highlighting}[]
\NormalTok{spot\_1.plot\_progress(log\_y}\OperatorTok{=}\VariableTok{True}\NormalTok{)}
\end{Highlighting}
\end{Shaded}

\begin{figure}[H]

{\centering \includegraphics{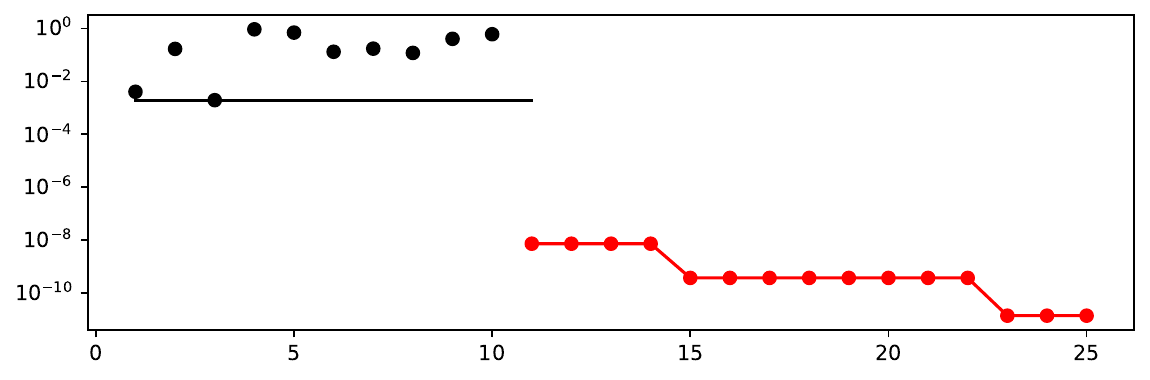}

}

\end{figure}

\begin{figure}

{\centering \includegraphics[width=1\textwidth,height=\textheight]{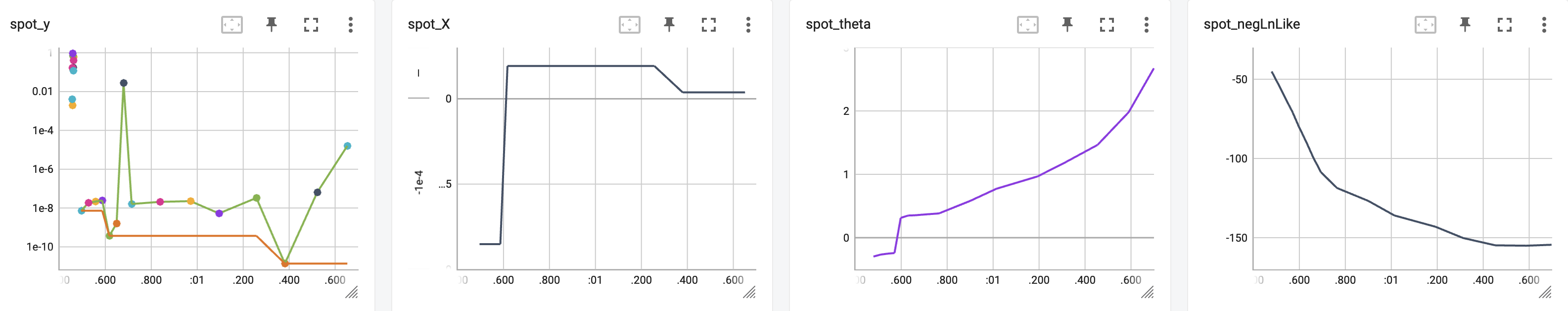}

}

\caption{TensorBoard visualization of the spotPython optimization
process and the surrogate model.}

\end{figure}

\hypertarget{same-but-with-ei-as-infill_criterion}{%
\section{Same, but with EI as
infill\_criterion}\label{same-but-with-ei-as-infill_criterion}}

\begin{Shaded}
\begin{Highlighting}[]
\NormalTok{PREFIX }\OperatorTok{=} \StringTok{"07\_EI\_ISO"}
\NormalTok{experiment\_name }\OperatorTok{=}\NormalTok{ get\_experiment\_name(prefix}\OperatorTok{=}\NormalTok{PREFIX)}
\BuiltInTok{print}\NormalTok{(experiment\_name)}
\NormalTok{fun\_control }\OperatorTok{=}\NormalTok{ fun\_control\_init(}
\NormalTok{    spot\_tensorboard\_path}\OperatorTok{=}\NormalTok{get\_spot\_tensorboard\_path(experiment\_name),}
\NormalTok{    sigma}\OperatorTok{=}\DecValTok{0}\NormalTok{,}
\NormalTok{    seed}\OperatorTok{=}\DecValTok{123}\NormalTok{,)}
\end{Highlighting}
\end{Shaded}

\begin{verbatim}
07_EI_ISO_bartz09_2023-07-17_08-50-47
\end{verbatim}

\begin{Shaded}
\begin{Highlighting}[]
\NormalTok{spot\_1\_ei }\OperatorTok{=}\NormalTok{ spot.Spot(fun}\OperatorTok{=}\NormalTok{fun,}
\NormalTok{                   lower }\OperatorTok{=}\NormalTok{ np.array([}\OperatorTok{{-}}\DecValTok{1}\NormalTok{]),}
\NormalTok{                   upper }\OperatorTok{=}\NormalTok{ np.array([}\DecValTok{1}\NormalTok{]),}
\NormalTok{                   fun\_evals }\OperatorTok{=} \DecValTok{25}\NormalTok{,}
\NormalTok{                   tolerance\_x }\OperatorTok{=}\NormalTok{ np.sqrt(np.spacing(}\DecValTok{1}\NormalTok{)),}
\NormalTok{                   infill\_criterion }\OperatorTok{=} \StringTok{"ei"}\NormalTok{,}
\NormalTok{                   design\_control}\OperatorTok{=}\NormalTok{\{}\StringTok{"init\_size"}\NormalTok{: }\DecValTok{10}\NormalTok{\},}
\NormalTok{                   fun\_control }\OperatorTok{=}\NormalTok{ fun\_control,)}
\NormalTok{spot\_1\_ei.run()}
\end{Highlighting}
\end{Shaded}

\begin{verbatim}
spotPython tuning: 1.1630341306771934e-08 [####------] 44.00% 
\end{verbatim}

\begin{verbatim}
spotPython tuning: 1.1630341306771934e-08 [#####-----] 48.00% 
\end{verbatim}

\begin{verbatim}
spotPython tuning: 1.1630341306771934e-08 [#####-----] 52.00% 
\end{verbatim}

\begin{verbatim}
spotPython tuning: 1.1630341306771934e-08 [######----] 56.00% 
\end{verbatim}

\begin{verbatim}
spotPython tuning: 2.207887258868953e-10 [######----] 60.00% 
\end{verbatim}

\begin{verbatim}
spotPython tuning: 2.207887258868953e-10 [######----] 64.00% 
\end{verbatim}

\begin{verbatim}
spotPython tuning: 2.207887258868953e-10 [#######---] 68.00% 
\end{verbatim}

\begin{verbatim}
spotPython tuning: 2.207887258868953e-10 [#######---] 72.00% 
\end{verbatim}

\begin{verbatim}
spotPython tuning: 2.207887258868953e-10 [########--] 76.00% 
\end{verbatim}

\begin{verbatim}
spotPython tuning: 2.207887258868953e-10 [########--] 80.00% 
\end{verbatim}

\begin{verbatim}
spotPython tuning: 2.207887258868953e-10 [########--] 84.00% 
\end{verbatim}

\begin{verbatim}
spotPython tuning: 2.207887258868953e-10 [#########-] 88.00% 
\end{verbatim}

\begin{verbatim}
spotPython tuning: 1.3536080613078865e-10 [#########-] 92.00% 
\end{verbatim}

\begin{verbatim}
spotPython tuning: 1.3536080613078865e-10 [##########] 96.00% 
\end{verbatim}

\begin{verbatim}
spotPython tuning: 1.3536080613078865e-10 [##########] 100.00% Done...
\end{verbatim}

\begin{verbatim}
<spotPython.spot.spot.Spot at 0x16e9e4c40>
\end{verbatim}

\begin{Shaded}
\begin{Highlighting}[]
\NormalTok{spot\_1\_ei.plot\_progress(log\_y}\OperatorTok{=}\VariableTok{True}\NormalTok{)}
\end{Highlighting}
\end{Shaded}

\begin{figure}[H]

{\centering \includegraphics{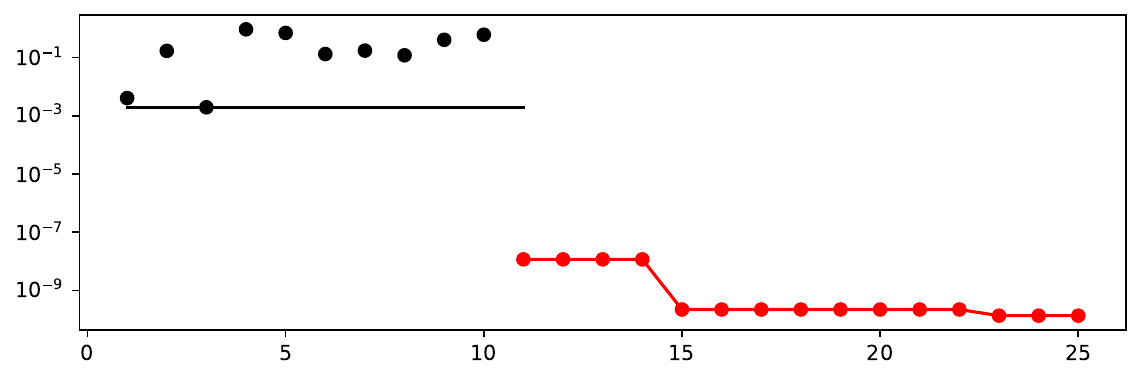}

}

\end{figure}

\begin{Shaded}
\begin{Highlighting}[]
\NormalTok{spot\_1\_ei.print\_results()}
\end{Highlighting}
\end{Shaded}

\begin{verbatim}
min y: 1.3536080613078865e-10
x0: 1.1634466301931888e-05
\end{verbatim}

\begin{verbatim}
[['x0', 1.1634466301931888e-05]]
\end{verbatim}

\begin{figure}

{\centering \includegraphics[width=1\textwidth,height=\textheight]{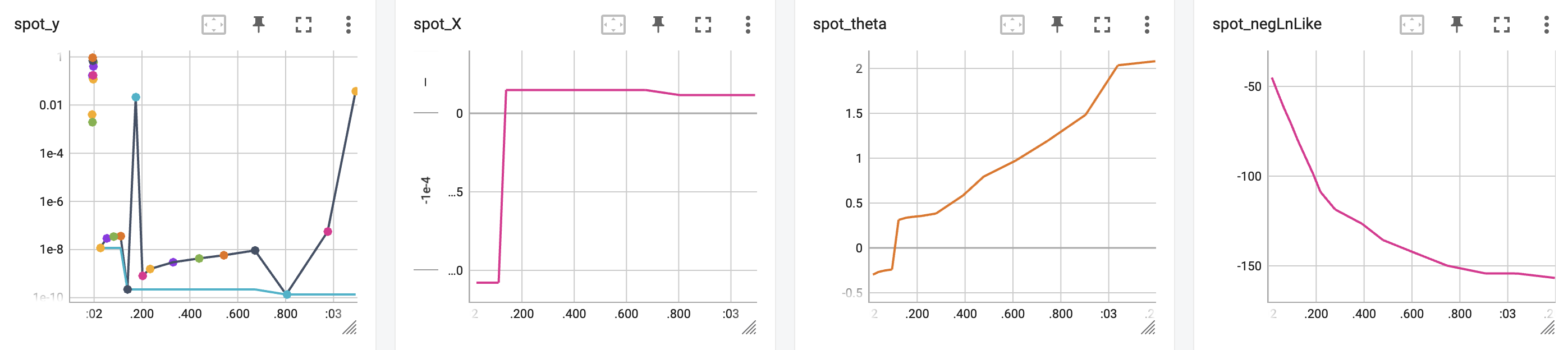}

}

\caption{TensorBoard visualization of the spotPython optimization
process and the surrogate model. Expected improvement, isotropic
Kriging.}

\end{figure}

\hypertarget{non-isotropic-kriging}{%
\section{Non-isotropic Kriging}\label{non-isotropic-kriging}}

\begin{Shaded}
\begin{Highlighting}[]
\NormalTok{PREFIX }\OperatorTok{=} \StringTok{"07\_EI\_NONISO"}
\NormalTok{experiment\_name }\OperatorTok{=}\NormalTok{ get\_experiment\_name(prefix}\OperatorTok{=}\NormalTok{PREFIX)}
\BuiltInTok{print}\NormalTok{(experiment\_name)}
\NormalTok{fun\_control }\OperatorTok{=}\NormalTok{ fun\_control\_init(}
\NormalTok{    spot\_tensorboard\_path}\OperatorTok{=}\NormalTok{get\_spot\_tensorboard\_path(experiment\_name),}
\NormalTok{    sigma}\OperatorTok{=}\DecValTok{0}\NormalTok{,}
\NormalTok{    seed}\OperatorTok{=}\DecValTok{123}\NormalTok{,)}
\end{Highlighting}
\end{Shaded}

\begin{verbatim}
07_EI_NONISO_bartz09_2023-07-17_08-50-49
\end{verbatim}

\begin{Shaded}
\begin{Highlighting}[]
\NormalTok{spot\_2\_ei\_noniso }\OperatorTok{=}\NormalTok{ spot.Spot(fun}\OperatorTok{=}\NormalTok{fun,}
\NormalTok{                   lower }\OperatorTok{=}\NormalTok{ np.array([}\OperatorTok{{-}}\DecValTok{1}\NormalTok{, }\OperatorTok{{-}}\DecValTok{1}\NormalTok{]),}
\NormalTok{                   upper }\OperatorTok{=}\NormalTok{ np.array([}\DecValTok{1}\NormalTok{, }\DecValTok{1}\NormalTok{]),}
\NormalTok{                   fun\_evals }\OperatorTok{=} \DecValTok{25}\NormalTok{,}
\NormalTok{                   tolerance\_x }\OperatorTok{=}\NormalTok{ np.sqrt(np.spacing(}\DecValTok{1}\NormalTok{)),}
\NormalTok{                   infill\_criterion }\OperatorTok{=} \StringTok{"ei"}\NormalTok{,}
\NormalTok{                   show\_models}\OperatorTok{=}\VariableTok{True}\NormalTok{,}
\NormalTok{                   design\_control}\OperatorTok{=}\NormalTok{\{}\StringTok{"init\_size"}\NormalTok{: }\DecValTok{10}\NormalTok{\},}
\NormalTok{                   surrogate\_control}\OperatorTok{=}\NormalTok{\{}\StringTok{"noise"}\NormalTok{: }\VariableTok{False}\NormalTok{,}
                                      \StringTok{"cod\_type"}\NormalTok{: }\StringTok{"norm"}\NormalTok{,}
                                      \StringTok{"min\_theta"}\NormalTok{: }\OperatorTok{{-}}\DecValTok{4}\NormalTok{,}
                                      \StringTok{"max\_theta"}\NormalTok{: }\DecValTok{3}\NormalTok{,}
                                      \StringTok{"n\_theta"}\NormalTok{: }\DecValTok{2}\NormalTok{,}
                                      \StringTok{"model\_fun\_evals"}\NormalTok{: }\DecValTok{1000}\NormalTok{,}
\NormalTok{                                      \},}
\NormalTok{                    fun\_control}\OperatorTok{=}\NormalTok{fun\_control,)}
\NormalTok{spot\_2\_ei\_noniso.run()}
\end{Highlighting}
\end{Shaded}

\begin{verbatim}
spotPython tuning: 1.754686753274553e-05 [####------] 44.00% 
\end{verbatim}

\begin{verbatim}
spotPython tuning: 1.754686753274553e-05 [#####-----] 48.00% 
\end{verbatim}

\begin{verbatim}
spotPython tuning: 1.754686753274553e-05 [#####-----] 52.00% 
\end{verbatim}

\begin{verbatim}
spotPython tuning: 1.0120806700557811e-05 [######----] 56.00% 
\end{verbatim}

\begin{verbatim}
spotPython tuning: 1.0120806700557811e-05 [######----] 60.00% 
\end{verbatim}

\begin{verbatim}
spotPython tuning: 1.8779971830281702e-07 [######----] 64.00% 
\end{verbatim}

\begin{verbatim}
spotPython tuning: 1.8779971830281702e-07 [#######---] 68.00% 
\end{verbatim}

\begin{verbatim}
spotPython tuning: 1.8779971830281702e-07 [#######---] 72.00% 
\end{verbatim}

\begin{verbatim}
spotPython tuning: 1.8779971830281702e-07 [########--] 76.00% 
\end{verbatim}

\begin{verbatim}
spotPython tuning: 1.8779971830281702e-07 [########--] 80.00% 
\end{verbatim}

\begin{verbatim}
spotPython tuning: 1.8779971830281702e-07 [########--] 84.00% 
\end{verbatim}

\begin{verbatim}
spotPython tuning: 1.8779971830281702e-07 [#########-] 88.00% 
\end{verbatim}

\begin{verbatim}
spotPython tuning: 1.8779971830281702e-07 [#########-] 92.00% 
\end{verbatim}

\begin{verbatim}
spotPython tuning: 1.8779971830281702e-07 [##########] 96.00% 
\end{verbatim}

\begin{verbatim}
spotPython tuning: 1.8779971830281702e-07 [##########] 100.00% Done...
\end{verbatim}

\begin{verbatim}
<spotPython.spot.spot.Spot at 0x16ebefee0>
\end{verbatim}

\begin{Shaded}
\begin{Highlighting}[]
\NormalTok{spot\_2\_ei\_noniso.plot\_progress(log\_y}\OperatorTok{=}\VariableTok{True}\NormalTok{)}
\end{Highlighting}
\end{Shaded}

\begin{figure}[H]

{\centering \includegraphics{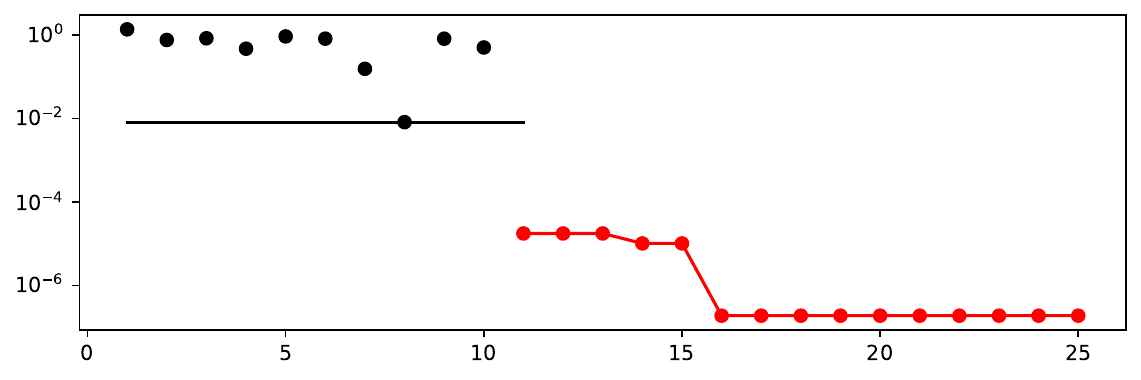}

}

\end{figure}

\begin{Shaded}
\begin{Highlighting}[]
\NormalTok{spot\_2\_ei\_noniso.print\_results()}
\end{Highlighting}
\end{Shaded}

\begin{verbatim}
min y: 1.8779971830281702e-07
x0: -0.0002783721390529846
x1: 0.0003321274913371111
\end{verbatim}

\begin{verbatim}
[['x0', -0.0002783721390529846], ['x1', 0.0003321274913371111]]
\end{verbatim}

\begin{Shaded}
\begin{Highlighting}[]
\NormalTok{spot\_2\_ei\_noniso.surrogate.plot()}
\end{Highlighting}
\end{Shaded}

\begin{figure}[H]

{\centering \includegraphics{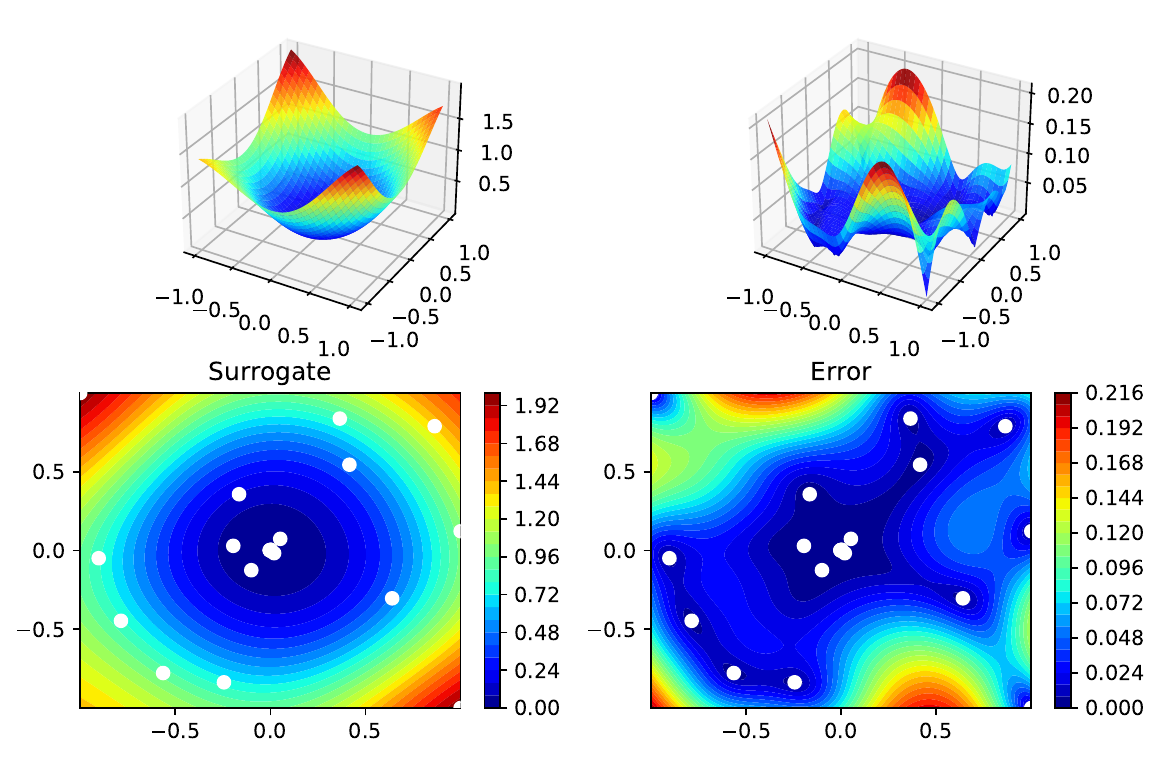}

}

\end{figure}

\begin{figure}

{\centering \includegraphics[width=1\textwidth,height=\textheight]{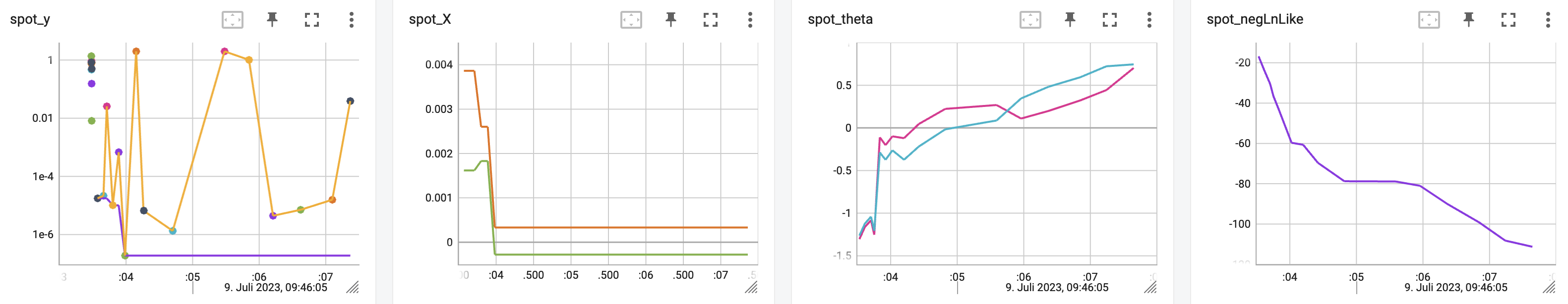}

}

\caption{TensorBoard visualization of the spotPython optimization
process and the surrogate model. Expected improvement, isotropic
Kriging.}

\end{figure}

\hypertarget{using-sklearn-surrogates}{%
\section{\texorpdfstring{Using \texttt{sklearn}
Surrogates}{Using sklearn Surrogates}}\label{using-sklearn-surrogates}}

\hypertarget{the-spot-loop}{%
\subsection{The spot Loop}\label{the-spot-loop}}

The \texttt{spot} loop consists of the following steps:

\begin{enumerate}
\def\labelenumi{\arabic{enumi}.}
\tightlist
\item
  Init: Build initial design \(X\)
\item
  Evaluate initial design on real objective \(f\): \(y = f(X)\)
\item
  Build surrogate: \(S = S(X,y)\)
\item
  Optimize on surrogate: \(X_0 = \text{optimize}(S)\)
\item
  Evaluate on real objective: \(y_0 = f(X_0)\)
\item
  Impute (Infill) new points: \(X = X \cup X_0\), \(y = y \cup y_0\).
\item
  Got 3.
\end{enumerate}

The \texttt{spot} loop is implemented in \texttt{R} as follows:

\begin{figure}

{\centering \includegraphics{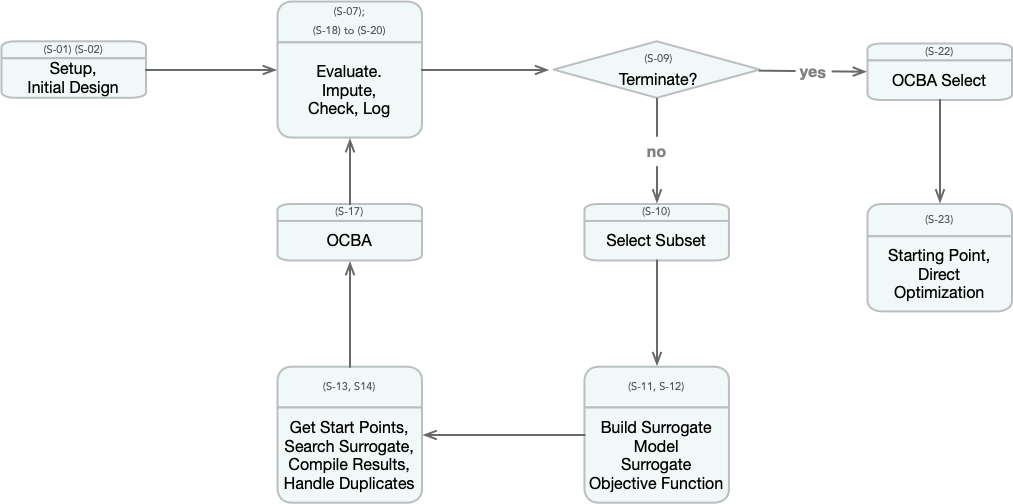}

}

\caption{Visual representation of the model based search with SPOT.
Taken from: Bartz-Beielstein, T., and Zaefferer, M. Hyperparameter
tuning approaches. In Hyperparameter Tuning for Machine and Deep
Learning with R - A Practical Guide, E. Bartz, T. Bartz-Beielstein, M.
Zaefferer, and O. Mersmann, Eds. Springer, 2022, ch.~4, pp.~67--114.}

\end{figure}

\hypertarget{spot-the-initial-model}{%
\subsection{spot: The Initial Model}\label{spot-the-initial-model}}

\hypertarget{example-modifying-the-initial-design-size}{%
\subsubsection{Example: Modifying the initial design
size}\label{example-modifying-the-initial-design-size}}

This is the ``Example: Modifying the initial design size'' from Chapter
4.5.1 in {[}bart21i{]}.

\begin{Shaded}
\begin{Highlighting}[]
\NormalTok{spot\_ei }\OperatorTok{=}\NormalTok{ spot.Spot(fun}\OperatorTok{=}\NormalTok{fun,}
\NormalTok{               lower }\OperatorTok{=}\NormalTok{ np.array([}\OperatorTok{{-}}\DecValTok{1}\NormalTok{,}\OperatorTok{{-}}\DecValTok{1}\NormalTok{]),}
\NormalTok{               upper}\OperatorTok{=}\NormalTok{ np.array([}\DecValTok{1}\NormalTok{,}\DecValTok{1}\NormalTok{]),}
\NormalTok{               design\_control}\OperatorTok{=}\NormalTok{\{}\StringTok{"init\_size"}\NormalTok{: }\DecValTok{5}\NormalTok{\})}
\NormalTok{spot\_ei.run()}
\end{Highlighting}
\end{Shaded}

\begin{verbatim}
spotPython tuning: 0.13881986540743513 [####------] 40.00% 
\end{verbatim}

\begin{verbatim}
spotPython tuning: 0.0111581443080968 [#####-----] 46.67% 
\end{verbatim}

\begin{verbatim}
spotPython tuning: 0.0010079970679825743 [#####-----] 53.33% 
\end{verbatim}

\begin{verbatim}
spotPython tuning: 0.000631621365403864 [######----] 60.00% 
\end{verbatim}

\begin{verbatim}
spotPython tuning: 0.0005883893741686826 [#######---] 66.67% 
\end{verbatim}

\begin{verbatim}
spotPython tuning: 0.00058412889636168 [#######---] 73.33% 
\end{verbatim}

\begin{verbatim}
spotPython tuning: 0.0005539414734082665 [########--] 80.00% 
\end{verbatim}

\begin{verbatim}
spotPython tuning: 0.0004401288692983916 [#########-] 86.67% 
\end{verbatim}

\begin{verbatim}
spotPython tuning: 5.8179647898944394e-05 [#########-] 93.33% 
\end{verbatim}

\begin{verbatim}
spotPython tuning: 1.7928640814182596e-05 [##########] 100.00% Done...
\end{verbatim}

\begin{verbatim}
<spotPython.spot.spot.Spot at 0x16eb6e230>
\end{verbatim}

\begin{Shaded}
\begin{Highlighting}[]
\NormalTok{spot\_ei.plot\_progress()}
\end{Highlighting}
\end{Shaded}

\begin{figure}[H]

{\centering \includegraphics{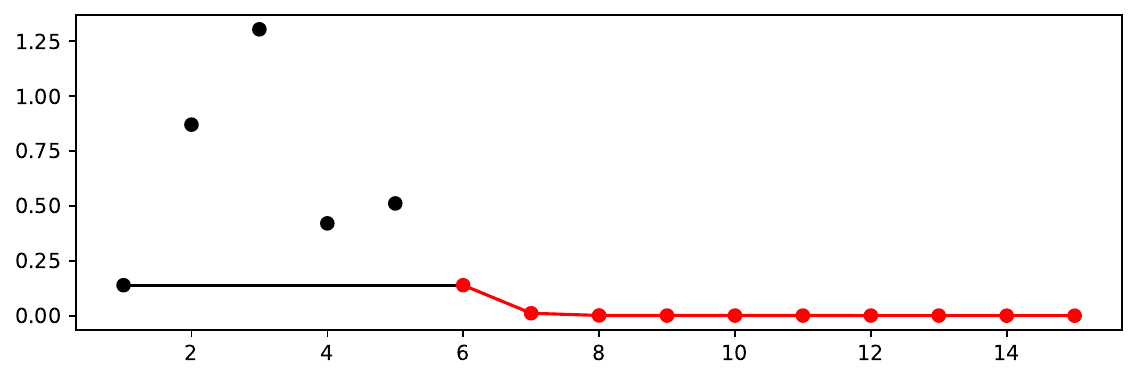}

}

\end{figure}

\begin{Shaded}
\begin{Highlighting}[]
\NormalTok{np.}\BuiltInTok{min}\NormalTok{(spot\_1.y), np.}\BuiltInTok{min}\NormalTok{(spot\_ei.y)}
\end{Highlighting}
\end{Shaded}

\begin{verbatim}
(1.3792745942664307e-11, 1.7928640814182596e-05)
\end{verbatim}

\hypertarget{init-build-initial-design}{%
\subsection{Init: Build Initial
Design}\label{init-build-initial-design}}

\begin{Shaded}
\begin{Highlighting}[]
\ImportTok{from}\NormalTok{ spotPython.design.spacefilling }\ImportTok{import}\NormalTok{ spacefilling}
\ImportTok{from}\NormalTok{ spotPython.build.kriging }\ImportTok{import}\NormalTok{ Kriging}
\ImportTok{from}\NormalTok{ spotPython.fun.objectivefunctions }\ImportTok{import}\NormalTok{ analytical}
\NormalTok{gen }\OperatorTok{=}\NormalTok{ spacefilling(}\DecValTok{2}\NormalTok{)}
\NormalTok{rng }\OperatorTok{=}\NormalTok{ np.random.RandomState(}\DecValTok{1}\NormalTok{)}
\NormalTok{lower }\OperatorTok{=}\NormalTok{ np.array([}\OperatorTok{{-}}\DecValTok{5}\NormalTok{,}\OperatorTok{{-}}\DecValTok{0}\NormalTok{])}
\NormalTok{upper }\OperatorTok{=}\NormalTok{ np.array([}\DecValTok{10}\NormalTok{,}\DecValTok{15}\NormalTok{])}
\NormalTok{fun }\OperatorTok{=}\NormalTok{ analytical().fun\_branin}

\NormalTok{X }\OperatorTok{=}\NormalTok{ gen.scipy\_lhd(}\DecValTok{10}\NormalTok{, lower}\OperatorTok{=}\NormalTok{lower, upper }\OperatorTok{=}\NormalTok{ upper)}
\BuiltInTok{print}\NormalTok{(X)}
\NormalTok{y }\OperatorTok{=}\NormalTok{ fun(X, fun\_control}\OperatorTok{=}\NormalTok{fun\_control)}
\BuiltInTok{print}\NormalTok{(y)}
\end{Highlighting}
\end{Shaded}

\begin{verbatim}
[[ 8.97647221 13.41926847]
 [ 0.66946019  1.22344228]
 [ 5.23614115 13.78185824]
 [ 5.6149825  11.5851384 ]
 [-1.72963184  1.66516096]
 [-4.26945568  7.1325531 ]
 [ 1.26363761 10.17935555]
 [ 2.88779942  8.05508969]
 [-3.39111089  4.15213772]
 [ 7.30131231  5.22275244]]
[128.95676449  31.73474356 172.89678121 126.71295908  64.34349975
  70.16178611  48.71407916  31.77322887  76.91788181  30.69410529]
\end{verbatim}

\begin{Shaded}
\begin{Highlighting}[]
\NormalTok{S }\OperatorTok{=}\NormalTok{ Kriging(name}\OperatorTok{=}\StringTok{\textquotesingle{}kriging\textquotesingle{}}\NormalTok{,  seed}\OperatorTok{=}\DecValTok{123}\NormalTok{)}
\NormalTok{S.fit(X, y)}
\NormalTok{S.plot()}
\end{Highlighting}
\end{Shaded}

\begin{figure}[H]

{\centering \includegraphics{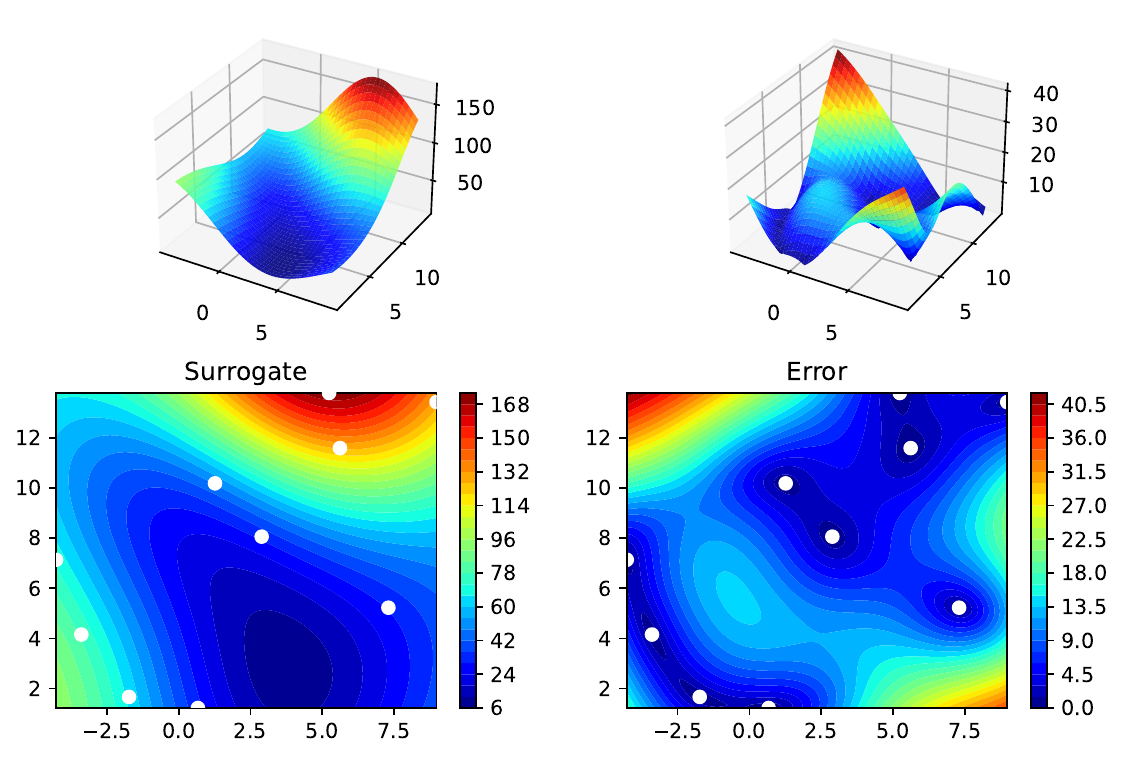}

}

\end{figure}

\begin{Shaded}
\begin{Highlighting}[]
\NormalTok{gen }\OperatorTok{=}\NormalTok{ spacefilling(}\DecValTok{2}\NormalTok{, seed}\OperatorTok{=}\DecValTok{123}\NormalTok{)}
\NormalTok{X0 }\OperatorTok{=}\NormalTok{ gen.scipy\_lhd(}\DecValTok{3}\NormalTok{)}
\NormalTok{gen }\OperatorTok{=}\NormalTok{ spacefilling(}\DecValTok{2}\NormalTok{, seed}\OperatorTok{=}\DecValTok{345}\NormalTok{)}
\NormalTok{X1 }\OperatorTok{=}\NormalTok{ gen.scipy\_lhd(}\DecValTok{3}\NormalTok{)}
\NormalTok{X2 }\OperatorTok{=}\NormalTok{ gen.scipy\_lhd(}\DecValTok{3}\NormalTok{)}
\NormalTok{gen }\OperatorTok{=}\NormalTok{ spacefilling(}\DecValTok{2}\NormalTok{, seed}\OperatorTok{=}\DecValTok{123}\NormalTok{)}
\NormalTok{X3 }\OperatorTok{=}\NormalTok{ gen.scipy\_lhd(}\DecValTok{3}\NormalTok{)}
\NormalTok{X0, X1, X2, X3}
\end{Highlighting}
\end{Shaded}

\begin{verbatim}
(array([[0.77254938, 0.31539299],
        [0.59321338, 0.93854273],
        [0.27469803, 0.3959685 ]]),
 array([[0.78373509, 0.86811887],
        [0.06692621, 0.6058029 ],
        [0.41374778, 0.00525456]]),
 array([[0.121357  , 0.69043832],
        [0.41906219, 0.32838498],
        [0.86742658, 0.52910374]]),
 array([[0.77254938, 0.31539299],
        [0.59321338, 0.93854273],
        [0.27469803, 0.3959685 ]]))
\end{verbatim}

\hypertarget{evaluate}{%
\subsection{Evaluate}\label{evaluate}}

\hypertarget{build-surrogate}{%
\subsection{Build Surrogate}\label{build-surrogate}}

\hypertarget{a-simple-predictor}{%
\subsection{A Simple Predictor}\label{a-simple-predictor}}

The code below shows how to use a simple model for prediction.

\begin{itemize}
\item
  Assume that only two (very costly) measurements are available:

  \begin{enumerate}
  \def\labelenumi{\arabic{enumi}.}
  \tightlist
  \item
    f(0) = 0.5
  \item
    f(2) = 2.5
  \end{enumerate}
\item
  We are interested in the value at \(x_0 = 1\), i.e., \(f(x_0 = 1)\),
  but cannot run an additional, third experiment.
\end{itemize}

\begin{Shaded}
\begin{Highlighting}[]
\ImportTok{from}\NormalTok{ sklearn }\ImportTok{import}\NormalTok{ linear\_model}
\NormalTok{X }\OperatorTok{=}\NormalTok{ np.array([[}\DecValTok{0}\NormalTok{], [}\DecValTok{2}\NormalTok{]])}
\NormalTok{y }\OperatorTok{=}\NormalTok{ np.array([}\FloatTok{0.5}\NormalTok{, }\FloatTok{2.5}\NormalTok{])}
\NormalTok{S\_lm }\OperatorTok{=}\NormalTok{ linear\_model.LinearRegression()}
\NormalTok{S\_lm }\OperatorTok{=}\NormalTok{ S\_lm.fit(X, y)}
\NormalTok{X0 }\OperatorTok{=}\NormalTok{ np.array([[}\DecValTok{1}\NormalTok{]])}
\NormalTok{y0 }\OperatorTok{=}\NormalTok{ S\_lm.predict(X0)}
\BuiltInTok{print}\NormalTok{(y0)}
\end{Highlighting}
\end{Shaded}

\begin{verbatim}
[1.5]
\end{verbatim}

\begin{itemize}
\tightlist
\item
  Central Idea:

  \begin{itemize}
  \tightlist
  \item
    Evaluation of the surrogate model \texttt{S\_lm} is much cheaper (or
    / and much faster) than running the real-world experiment \(f\).
  \end{itemize}
\end{itemize}

\hypertarget{gaussian-processes-regression-basic-introductory-example}{%
\section{Gaussian Processes regression: basic introductory
example}\label{gaussian-processes-regression-basic-introductory-example}}

This example was taken from
\href{https://scikit-learn.org/stable/auto_examples/gaussian_process/plot_gpr_noisy_targets.html}{scikit-learn}.
After fitting our model, we see that the hyperparameters of the kernel
have been optimized. Now, we will use our kernel to compute the mean
prediction of the full dataset and plot the 95\% confidence interval.

\begin{Shaded}
\begin{Highlighting}[]
\ImportTok{import}\NormalTok{ numpy }\ImportTok{as}\NormalTok{ np}
\ImportTok{import}\NormalTok{ matplotlib.pyplot }\ImportTok{as}\NormalTok{ plt}
\ImportTok{import}\NormalTok{ math }\ImportTok{as}\NormalTok{ m}
\ImportTok{from}\NormalTok{ sklearn.gaussian\_process }\ImportTok{import}\NormalTok{ GaussianProcessRegressor}
\ImportTok{from}\NormalTok{ sklearn.gaussian\_process.kernels }\ImportTok{import}\NormalTok{ RBF}

\NormalTok{X }\OperatorTok{=}\NormalTok{ np.linspace(start}\OperatorTok{=}\DecValTok{0}\NormalTok{, stop}\OperatorTok{=}\DecValTok{10}\NormalTok{, num}\OperatorTok{=}\DecValTok{1\_000}\NormalTok{).reshape(}\OperatorTok{{-}}\DecValTok{1}\NormalTok{, }\DecValTok{1}\NormalTok{)}
\NormalTok{y }\OperatorTok{=}\NormalTok{ np.squeeze(X }\OperatorTok{*}\NormalTok{ np.sin(X))}
\NormalTok{rng }\OperatorTok{=}\NormalTok{ np.random.RandomState(}\DecValTok{1}\NormalTok{)}
\NormalTok{training\_indices }\OperatorTok{=}\NormalTok{ rng.choice(np.arange(y.size), size}\OperatorTok{=}\DecValTok{6}\NormalTok{, replace}\OperatorTok{=}\VariableTok{False}\NormalTok{)}
\NormalTok{X\_train, y\_train }\OperatorTok{=}\NormalTok{ X[training\_indices], y[training\_indices]}

\NormalTok{kernel }\OperatorTok{=} \DecValTok{1} \OperatorTok{*}\NormalTok{ RBF(length\_scale}\OperatorTok{=}\FloatTok{1.0}\NormalTok{, length\_scale\_bounds}\OperatorTok{=}\NormalTok{(}\FloatTok{1e{-}2}\NormalTok{, }\FloatTok{1e2}\NormalTok{))}
\NormalTok{gaussian\_process }\OperatorTok{=}\NormalTok{ GaussianProcessRegressor(kernel}\OperatorTok{=}\NormalTok{kernel, n\_restarts\_optimizer}\OperatorTok{=}\DecValTok{9}\NormalTok{)}
\NormalTok{gaussian\_process.fit(X\_train, y\_train)}
\NormalTok{gaussian\_process.kernel\_}

\NormalTok{mean\_prediction, std\_prediction }\OperatorTok{=}\NormalTok{ gaussian\_process.predict(X, return\_std}\OperatorTok{=}\VariableTok{True}\NormalTok{)}

\NormalTok{plt.plot(X, y, label}\OperatorTok{=}\VerbatimStringTok{r"$f(x) = x \textbackslash{}sin(x)$"}\NormalTok{, linestyle}\OperatorTok{=}\StringTok{"dotted"}\NormalTok{)}
\NormalTok{plt.scatter(X\_train, y\_train, label}\OperatorTok{=}\StringTok{"Observations"}\NormalTok{)}
\NormalTok{plt.plot(X, mean\_prediction, label}\OperatorTok{=}\StringTok{"Mean prediction"}\NormalTok{)}
\NormalTok{plt.fill\_between(}
\NormalTok{    X.ravel(),}
\NormalTok{    mean\_prediction }\OperatorTok{{-}} \FloatTok{1.96} \OperatorTok{*}\NormalTok{ std\_prediction,}
\NormalTok{    mean\_prediction }\OperatorTok{+} \FloatTok{1.96} \OperatorTok{*}\NormalTok{ std\_prediction,}
\NormalTok{    alpha}\OperatorTok{=}\FloatTok{0.5}\NormalTok{,}
\NormalTok{    label}\OperatorTok{=}\VerbatimStringTok{r"95}\SpecialCharTok{\% c}\VerbatimStringTok{onfidence interval"}\NormalTok{,}
\NormalTok{)}
\NormalTok{plt.legend()}
\NormalTok{plt.xlabel(}\StringTok{"$x$"}\NormalTok{)}
\NormalTok{plt.ylabel(}\StringTok{"$f(x)$"}\NormalTok{)}
\NormalTok{\_ }\OperatorTok{=}\NormalTok{ plt.title(}\StringTok{"sk{-}learn Version: Gaussian process regression on noise{-}free dataset"}\NormalTok{)}
\end{Highlighting}
\end{Shaded}

\begin{figure}[H]

{\centering \includegraphics{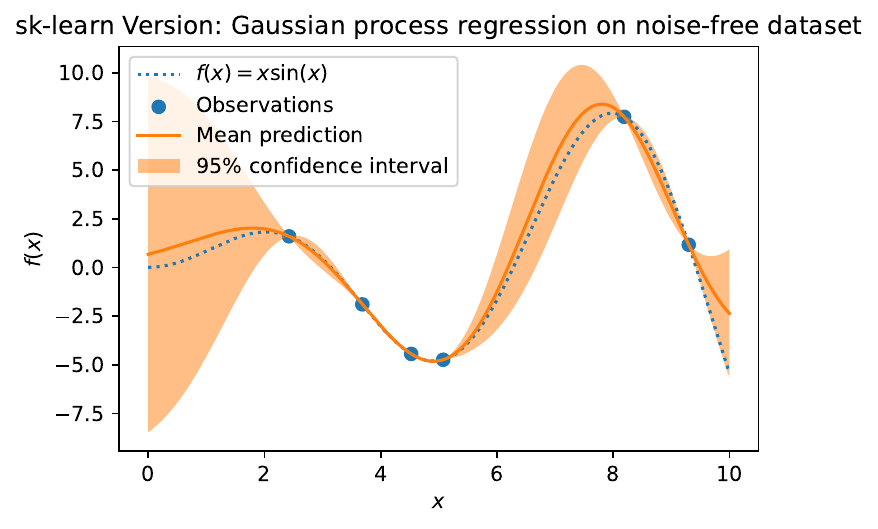}

}

\end{figure}

\begin{Shaded}
\begin{Highlighting}[]
\ImportTok{from}\NormalTok{ spotPython.build.kriging }\ImportTok{import}\NormalTok{ Kriging}
\ImportTok{import}\NormalTok{ numpy }\ImportTok{as}\NormalTok{ np}
\ImportTok{import}\NormalTok{ matplotlib.pyplot }\ImportTok{as}\NormalTok{ plt}
\NormalTok{rng }\OperatorTok{=}\NormalTok{ np.random.RandomState(}\DecValTok{1}\NormalTok{)}
\NormalTok{X }\OperatorTok{=}\NormalTok{ np.linspace(start}\OperatorTok{=}\DecValTok{0}\NormalTok{, stop}\OperatorTok{=}\DecValTok{10}\NormalTok{, num}\OperatorTok{=}\DecValTok{1\_000}\NormalTok{).reshape(}\OperatorTok{{-}}\DecValTok{1}\NormalTok{, }\DecValTok{1}\NormalTok{)}
\NormalTok{y }\OperatorTok{=}\NormalTok{ np.squeeze(X }\OperatorTok{*}\NormalTok{ np.sin(X))}
\NormalTok{training\_indices }\OperatorTok{=}\NormalTok{ rng.choice(np.arange(y.size), size}\OperatorTok{=}\DecValTok{6}\NormalTok{, replace}\OperatorTok{=}\VariableTok{False}\NormalTok{)}
\NormalTok{X\_train, y\_train }\OperatorTok{=}\NormalTok{ X[training\_indices], y[training\_indices]}


\NormalTok{S }\OperatorTok{=}\NormalTok{ Kriging(name}\OperatorTok{=}\StringTok{\textquotesingle{}kriging\textquotesingle{}}\NormalTok{,  seed}\OperatorTok{=}\DecValTok{123}\NormalTok{, log\_level}\OperatorTok{=}\DecValTok{50}\NormalTok{, cod\_type}\OperatorTok{=}\StringTok{"norm"}\NormalTok{)}
\NormalTok{S.fit(X\_train, y\_train)}

\NormalTok{mean\_prediction, std\_prediction, ei }\OperatorTok{=}\NormalTok{ S.predict(X, return\_val}\OperatorTok{=}\StringTok{"all"}\NormalTok{)}

\NormalTok{std\_prediction}

\NormalTok{plt.plot(X, y, label}\OperatorTok{=}\VerbatimStringTok{r"$f(x) = x \textbackslash{}sin(x)$"}\NormalTok{, linestyle}\OperatorTok{=}\StringTok{"dotted"}\NormalTok{)}
\NormalTok{plt.scatter(X\_train, y\_train, label}\OperatorTok{=}\StringTok{"Observations"}\NormalTok{)}
\NormalTok{plt.plot(X, mean\_prediction, label}\OperatorTok{=}\StringTok{"Mean prediction"}\NormalTok{)}
\NormalTok{plt.fill\_between(}
\NormalTok{    X.ravel(),}
\NormalTok{    mean\_prediction }\OperatorTok{{-}} \FloatTok{1.96} \OperatorTok{*}\NormalTok{ std\_prediction,}
\NormalTok{    mean\_prediction }\OperatorTok{+} \FloatTok{1.96} \OperatorTok{*}\NormalTok{ std\_prediction,}
\NormalTok{    alpha}\OperatorTok{=}\FloatTok{0.5}\NormalTok{,}
\NormalTok{    label}\OperatorTok{=}\VerbatimStringTok{r"95}\SpecialCharTok{\% c}\VerbatimStringTok{onfidence interval"}\NormalTok{,}
\NormalTok{)}
\NormalTok{plt.legend()}
\NormalTok{plt.xlabel(}\StringTok{"$x$"}\NormalTok{)}
\NormalTok{plt.ylabel(}\StringTok{"$f(x)$"}\NormalTok{)}
\NormalTok{\_ }\OperatorTok{=}\NormalTok{ plt.title(}\StringTok{"spotPython Version: Gaussian process regression on noise{-}free dataset"}\NormalTok{)}
\end{Highlighting}
\end{Shaded}

\begin{figure}[H]

{\centering \includegraphics{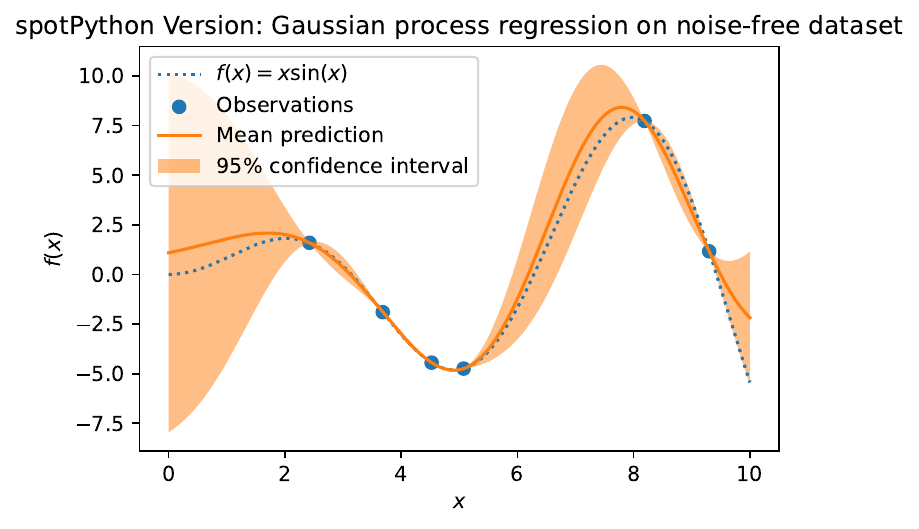}

}

\end{figure}

\hypertarget{the-surrogate-using-scikit-learn-models}{%
\section{The Surrogate: Using scikit-learn
models}\label{the-surrogate-using-scikit-learn-models}}

Default is the internal \texttt{kriging} surrogate.

\begin{Shaded}
\begin{Highlighting}[]
\NormalTok{S\_0 }\OperatorTok{=}\NormalTok{ Kriging(name}\OperatorTok{=}\StringTok{\textquotesingle{}kriging\textquotesingle{}}\NormalTok{, seed}\OperatorTok{=}\DecValTok{123}\NormalTok{)}
\end{Highlighting}
\end{Shaded}

Models from \texttt{scikit-learn} can be selected, e.g., Gaussian
Process:

\begin{Shaded}
\begin{Highlighting}[]
\CommentTok{\# Needed for the sklearn surrogates:}
\ImportTok{from}\NormalTok{ sklearn.gaussian\_process }\ImportTok{import}\NormalTok{ GaussianProcessRegressor}
\ImportTok{from}\NormalTok{ sklearn.gaussian\_process.kernels }\ImportTok{import}\NormalTok{ RBF}
\ImportTok{from}\NormalTok{ sklearn.tree }\ImportTok{import}\NormalTok{ DecisionTreeRegressor}
\ImportTok{from}\NormalTok{ sklearn.ensemble }\ImportTok{import}\NormalTok{ RandomForestRegressor}
\ImportTok{from}\NormalTok{ sklearn }\ImportTok{import}\NormalTok{ linear\_model}
\ImportTok{from}\NormalTok{ sklearn }\ImportTok{import}\NormalTok{ tree}
\ImportTok{import}\NormalTok{ pandas }\ImportTok{as}\NormalTok{ pd}
\end{Highlighting}
\end{Shaded}

\begin{Shaded}
\begin{Highlighting}[]
\NormalTok{kernel }\OperatorTok{=} \DecValTok{1} \OperatorTok{*}\NormalTok{ RBF(length\_scale}\OperatorTok{=}\FloatTok{1.0}\NormalTok{, length\_scale\_bounds}\OperatorTok{=}\NormalTok{(}\FloatTok{1e{-}2}\NormalTok{, }\FloatTok{1e2}\NormalTok{))}
\NormalTok{S\_GP }\OperatorTok{=}\NormalTok{ GaussianProcessRegressor(kernel}\OperatorTok{=}\NormalTok{kernel, n\_restarts\_optimizer}\OperatorTok{=}\DecValTok{9}\NormalTok{)}
\end{Highlighting}
\end{Shaded}

\begin{itemize}
\tightlist
\item
  and many more:
\end{itemize}

\begin{Shaded}
\begin{Highlighting}[]
\NormalTok{S\_Tree }\OperatorTok{=}\NormalTok{ DecisionTreeRegressor(random\_state}\OperatorTok{=}\DecValTok{0}\NormalTok{)}
\NormalTok{S\_LM }\OperatorTok{=}\NormalTok{ linear\_model.LinearRegression()}
\NormalTok{S\_Ridge }\OperatorTok{=}\NormalTok{ linear\_model.Ridge()}
\NormalTok{S\_RF }\OperatorTok{=}\NormalTok{ RandomForestRegressor(max\_depth}\OperatorTok{=}\DecValTok{2}\NormalTok{, random\_state}\OperatorTok{=}\DecValTok{0}\NormalTok{) }
\end{Highlighting}
\end{Shaded}

\begin{itemize}
\tightlist
\item
  The scikit-learn GP model \texttt{S\_GP} is selected.
\end{itemize}

\begin{Shaded}
\begin{Highlighting}[]
\NormalTok{S }\OperatorTok{=}\NormalTok{ S\_GP}
\end{Highlighting}
\end{Shaded}

\begin{Shaded}
\begin{Highlighting}[]
\BuiltInTok{isinstance}\NormalTok{(S, GaussianProcessRegressor)}
\end{Highlighting}
\end{Shaded}

\begin{verbatim}
True
\end{verbatim}

\begin{Shaded}
\begin{Highlighting}[]
\ImportTok{from}\NormalTok{ spotPython.fun.objectivefunctions }\ImportTok{import}\NormalTok{ analytical}
\NormalTok{fun }\OperatorTok{=}\NormalTok{ analytical().fun\_branin}
\NormalTok{lower }\OperatorTok{=}\NormalTok{ np.array([}\OperatorTok{{-}}\DecValTok{5}\NormalTok{,}\OperatorTok{{-}}\DecValTok{0}\NormalTok{])}
\NormalTok{upper }\OperatorTok{=}\NormalTok{ np.array([}\DecValTok{10}\NormalTok{,}\DecValTok{15}\NormalTok{])}
\NormalTok{design\_control}\OperatorTok{=}\NormalTok{\{}\StringTok{"init\_size"}\NormalTok{: }\DecValTok{5}\NormalTok{\}}
\NormalTok{surrogate\_control}\OperatorTok{=}\NormalTok{\{}
            \StringTok{"infill\_criterion"}\NormalTok{: }\VariableTok{None}\NormalTok{,}
            \StringTok{"n\_points"}\NormalTok{: }\DecValTok{1}\NormalTok{,}
\NormalTok{        \}}
\NormalTok{spot\_GP }\OperatorTok{=}\NormalTok{ spot.Spot(fun}\OperatorTok{=}\NormalTok{fun, lower }\OperatorTok{=}\NormalTok{ lower, upper}\OperatorTok{=}\NormalTok{ upper, surrogate}\OperatorTok{=}\NormalTok{S, }
\NormalTok{                    fun\_evals }\OperatorTok{=} \DecValTok{15}\NormalTok{, noise }\OperatorTok{=} \VariableTok{False}\NormalTok{, log\_level }\OperatorTok{=} \DecValTok{50}\NormalTok{,}
\NormalTok{                    design\_control}\OperatorTok{=}\NormalTok{design\_control,}
\NormalTok{                    surrogate\_control}\OperatorTok{=}\NormalTok{surrogate\_control)}

\NormalTok{spot\_GP.run()}
\end{Highlighting}
\end{Shaded}

\begin{verbatim}
spotPython tuning: 24.51465459019188 [####------] 40.00% 
\end{verbatim}

\begin{verbatim}
spotPython tuning: 11.003073503598229 [#####-----] 46.67% 
\end{verbatim}

\begin{verbatim}
spotPython tuning: 10.960665185123245 [#####-----] 53.33% 
\end{verbatim}

\begin{verbatim}
spotPython tuning: 10.960665185123245 [######----] 60.00% 
\end{verbatim}

\begin{verbatim}
spotPython tuning: 10.960665185123245 [#######---] 66.67% 
\end{verbatim}

\begin{verbatim}
spotPython tuning: 4.089511646427124 [#######---] 73.33% 
\end{verbatim}

\begin{verbatim}
spotPython tuning: 1.4230307255030858 [########--] 80.00% 
\end{verbatim}

\begin{verbatim}
spotPython tuning: 1.4230307255030858 [#########-] 86.67% 
\end{verbatim}

\begin{verbatim}
spotPython tuning: 1.4230307255030858 [#########-] 93.33% 
\end{verbatim}

\begin{verbatim}
spotPython tuning: 0.6949448160267053 [##########] 100.00% Done...
\end{verbatim}

\begin{verbatim}
<spotPython.spot.spot.Spot at 0x2a7d11570>
\end{verbatim}

\begin{Shaded}
\begin{Highlighting}[]
\NormalTok{spot\_GP.y}
\end{Highlighting}
\end{Shaded}

\begin{verbatim}
array([ 69.32459936, 152.38491454, 107.92560483,  24.51465459,
        76.73500031,  86.30425659,  11.0030735 ,  10.96066519,
        16.06666933,  24.08428925,   4.08951165,   1.42303073,
         1.4736037 ,  16.03577039,   0.69494482])
\end{verbatim}

\begin{Shaded}
\begin{Highlighting}[]
\NormalTok{spot\_GP.plot\_progress()}
\end{Highlighting}
\end{Shaded}

\begin{figure}[H]

{\centering \includegraphics{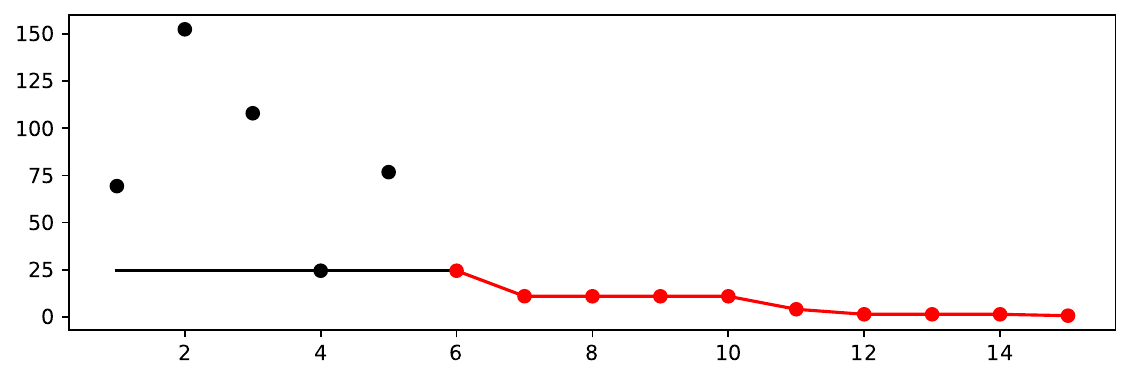}

}

\end{figure}

\begin{Shaded}
\begin{Highlighting}[]
\NormalTok{spot\_GP.print\_results()}
\end{Highlighting}
\end{Shaded}

\begin{verbatim}
min y: 0.6949448160267053
x0: 3.3575232000433637
x1: 2.3847893450472464
\end{verbatim}

\begin{verbatim}
[['x0', 3.3575232000433637], ['x1', 2.3847893450472464]]
\end{verbatim}

\hypertarget{additional-examples}{%
\section{Additional Examples}\label{additional-examples}}

\begin{Shaded}
\begin{Highlighting}[]
\CommentTok{\# Needed for the sklearn surrogates:}
\ImportTok{from}\NormalTok{ sklearn.gaussian\_process }\ImportTok{import}\NormalTok{ GaussianProcessRegressor}
\ImportTok{from}\NormalTok{ sklearn.gaussian\_process.kernels }\ImportTok{import}\NormalTok{ RBF}
\ImportTok{from}\NormalTok{ sklearn.tree }\ImportTok{import}\NormalTok{ DecisionTreeRegressor}
\ImportTok{from}\NormalTok{ sklearn.ensemble }\ImportTok{import}\NormalTok{ RandomForestRegressor}
\ImportTok{from}\NormalTok{ sklearn }\ImportTok{import}\NormalTok{ linear\_model}
\ImportTok{from}\NormalTok{ sklearn }\ImportTok{import}\NormalTok{ tree}
\ImportTok{import}\NormalTok{ pandas }\ImportTok{as}\NormalTok{ pd}
\end{Highlighting}
\end{Shaded}

\begin{Shaded}
\begin{Highlighting}[]
\NormalTok{kernel }\OperatorTok{=} \DecValTok{1} \OperatorTok{*}\NormalTok{ RBF(length\_scale}\OperatorTok{=}\FloatTok{1.0}\NormalTok{, length\_scale\_bounds}\OperatorTok{=}\NormalTok{(}\FloatTok{1e{-}2}\NormalTok{, }\FloatTok{1e2}\NormalTok{))}
\NormalTok{S\_GP }\OperatorTok{=}\NormalTok{ GaussianProcessRegressor(kernel}\OperatorTok{=}\NormalTok{kernel, n\_restarts\_optimizer}\OperatorTok{=}\DecValTok{9}\NormalTok{)}
\end{Highlighting}
\end{Shaded}

\begin{Shaded}
\begin{Highlighting}[]
\ImportTok{from}\NormalTok{ spotPython.build.kriging }\ImportTok{import}\NormalTok{ Kriging}
\ImportTok{import}\NormalTok{ numpy }\ImportTok{as}\NormalTok{ np}
\ImportTok{import}\NormalTok{ spotPython}
\ImportTok{from}\NormalTok{ spotPython.fun.objectivefunctions }\ImportTok{import}\NormalTok{ analytical}
\ImportTok{from}\NormalTok{ spotPython.spot }\ImportTok{import}\NormalTok{ spot}

\NormalTok{S\_K }\OperatorTok{=}\NormalTok{ Kriging(name}\OperatorTok{=}\StringTok{\textquotesingle{}kriging\textquotesingle{}}\NormalTok{,}
\NormalTok{              seed}\OperatorTok{=}\DecValTok{123}\NormalTok{,}
\NormalTok{              log\_level}\OperatorTok{=}\DecValTok{50}\NormalTok{,}
\NormalTok{              infill\_criterion }\OperatorTok{=} \StringTok{"y"}\NormalTok{,}
\NormalTok{              n\_theta}\OperatorTok{=}\DecValTok{1}\NormalTok{,}
\NormalTok{              noise}\OperatorTok{=}\VariableTok{False}\NormalTok{,}
\NormalTok{              cod\_type}\OperatorTok{=}\StringTok{"norm"}\NormalTok{)}
\NormalTok{fun }\OperatorTok{=}\NormalTok{ analytical().fun\_sphere}
\NormalTok{lower }\OperatorTok{=}\NormalTok{ np.array([}\OperatorTok{{-}}\DecValTok{1}\NormalTok{,}\OperatorTok{{-}}\DecValTok{1}\NormalTok{])}
\NormalTok{upper }\OperatorTok{=}\NormalTok{ np.array([}\DecValTok{1}\NormalTok{,}\DecValTok{1}\NormalTok{])}

\NormalTok{design\_control}\OperatorTok{=}\NormalTok{\{}\StringTok{"init\_size"}\NormalTok{: }\DecValTok{10}\NormalTok{\}}
\NormalTok{surrogate\_control}\OperatorTok{=}\NormalTok{\{}
            \StringTok{"n\_points"}\NormalTok{: }\DecValTok{1}\NormalTok{,}
\NormalTok{        \}}
\NormalTok{spot\_S\_K }\OperatorTok{=}\NormalTok{ spot.Spot(fun}\OperatorTok{=}\NormalTok{fun,}
\NormalTok{                     lower }\OperatorTok{=}\NormalTok{ lower,}
\NormalTok{                     upper}\OperatorTok{=}\NormalTok{ upper,}
\NormalTok{                     surrogate}\OperatorTok{=}\NormalTok{S\_K,}
\NormalTok{                     fun\_evals }\OperatorTok{=} \DecValTok{25}\NormalTok{,}
\NormalTok{                     noise }\OperatorTok{=} \VariableTok{False}\NormalTok{,}
\NormalTok{                     log\_level }\OperatorTok{=} \DecValTok{50}\NormalTok{,}
\NormalTok{                     design\_control}\OperatorTok{=}\NormalTok{design\_control,}
\NormalTok{                     surrogate\_control}\OperatorTok{=}\NormalTok{surrogate\_control)}

\NormalTok{spot\_S\_K.run()}
\end{Highlighting}
\end{Shaded}

\begin{verbatim}
spotPython tuning: 2.0398360048852566e-05 [####------] 44.00% 
\end{verbatim}

\begin{verbatim}
spotPython tuning: 2.0398360048852566e-05 [#####-----] 48.00% 
\end{verbatim}

\begin{verbatim}
spotPython tuning: 2.0398360048852566e-05 [#####-----] 52.00% 
\end{verbatim}

\begin{verbatim}
spotPython tuning: 2.0398360048852566e-05 [######----] 56.00% 
\end{verbatim}

\begin{verbatim}
spotPython tuning: 2.0398360048852566e-05 [######----] 60.00% 
\end{verbatim}

\begin{verbatim}
spotPython tuning: 1.0937897482978201e-05 [######----] 64.00% 
\end{verbatim}

\begin{verbatim}
spotPython tuning: 3.950539536972047e-06 [#######---] 68.00% 
\end{verbatim}

\begin{verbatim}
spotPython tuning: 3.2602730419203698e-06 [#######---] 72.00% 
\end{verbatim}

\begin{verbatim}
spotPython tuning: 2.4704028732017656e-06 [########--] 76.00% 
\end{verbatim}

\begin{verbatim}
spotPython tuning: 1.7687713431606244e-06 [########--] 80.00% 
\end{verbatim}

\begin{verbatim}
spotPython tuning: 1.7395335905335862e-06 [########--] 84.00% 
\end{verbatim}

\begin{verbatim}
spotPython tuning: 1.7395335905335862e-06 [#########-] 88.00% 
\end{verbatim}

\begin{verbatim}
spotPython tuning: 1.7395335905335862e-06 [#########-] 92.00% 
\end{verbatim}

\begin{verbatim}
spotPython tuning: 1.7395335905335862e-06 [##########] 96.00% 
\end{verbatim}

\begin{verbatim}
spotPython tuning: 1.7395335905335862e-06 [##########] 100.00% Done...
\end{verbatim}

\begin{verbatim}
<spotPython.spot.spot.Spot at 0x2c037a7d0>
\end{verbatim}

\begin{Shaded}
\begin{Highlighting}[]
\NormalTok{spot\_S\_K.plot\_progress(log\_y}\OperatorTok{=}\VariableTok{True}\NormalTok{)}
\end{Highlighting}
\end{Shaded}

\begin{figure}[H]

{\centering \includegraphics{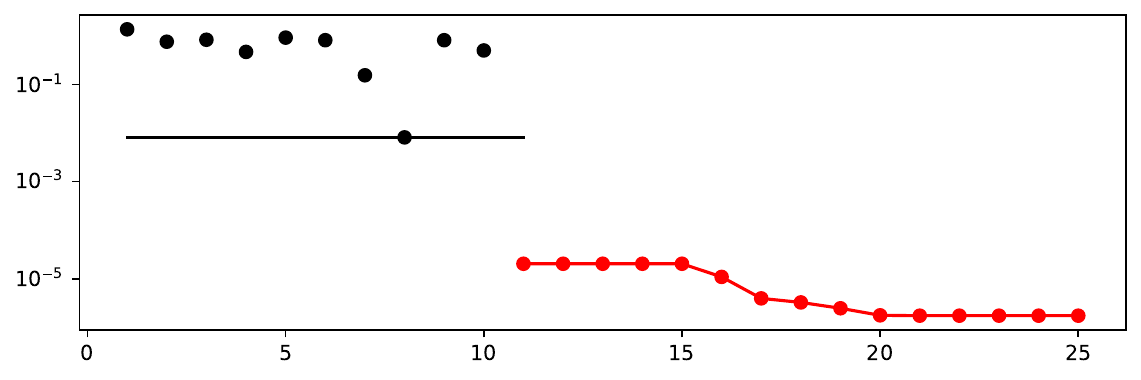}

}

\end{figure}

\begin{Shaded}
\begin{Highlighting}[]
\NormalTok{spot\_S\_K.surrogate.plot()}
\end{Highlighting}
\end{Shaded}

\begin{figure}[H]

{\centering \includegraphics{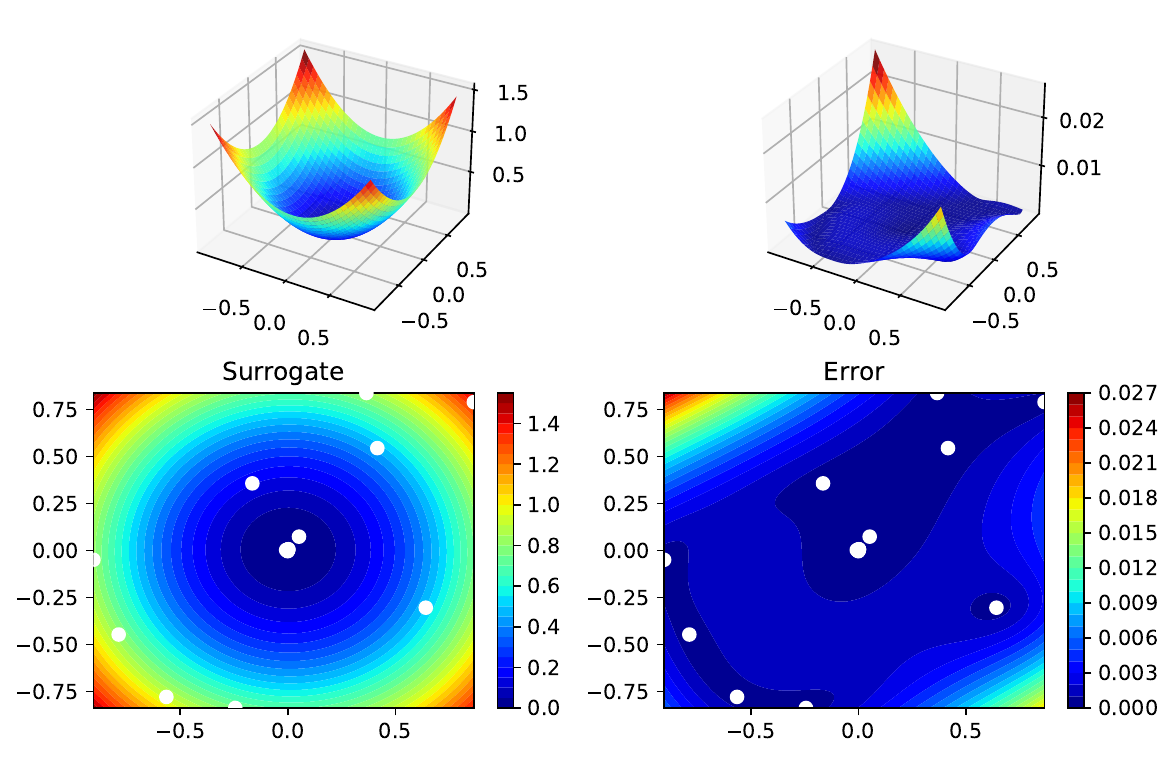}

}

\end{figure}

\begin{Shaded}
\begin{Highlighting}[]
\NormalTok{spot\_S\_K.print\_results()}
\end{Highlighting}
\end{Shaded}

\begin{verbatim}
min y: 1.7395335905335862e-06
x0: -0.0013044072412622557
x1: 0.0001950777780173277
\end{verbatim}

\begin{verbatim}
[['x0', -0.0013044072412622557], ['x1', 0.0001950777780173277]]
\end{verbatim}

\hypertarget{optimize-on-surrogate}{%
\subsection{Optimize on Surrogate}\label{optimize-on-surrogate}}

\hypertarget{evaluate-on-real-objective}{%
\subsection{Evaluate on Real
Objective}\label{evaluate-on-real-objective}}

\hypertarget{impute-infill-new-points}{%
\subsection{Impute / Infill new Points}\label{impute-infill-new-points}}

\hypertarget{tests}{%
\section{Tests}\label{tests}}

\begin{Shaded}
\begin{Highlighting}[]
\ImportTok{import}\NormalTok{ numpy }\ImportTok{as}\NormalTok{ np}
\ImportTok{from}\NormalTok{ spotPython.spot }\ImportTok{import}\NormalTok{ spot}
\ImportTok{from}\NormalTok{ spotPython.fun.objectivefunctions }\ImportTok{import}\NormalTok{ analytical}

\NormalTok{fun\_sphere }\OperatorTok{=}\NormalTok{ analytical().fun\_sphere}
\NormalTok{spot\_1 }\OperatorTok{=}\NormalTok{ spot.Spot(}
\NormalTok{    fun}\OperatorTok{=}\NormalTok{fun\_sphere,}
\NormalTok{    lower}\OperatorTok{=}\NormalTok{np.array([}\OperatorTok{{-}}\DecValTok{1}\NormalTok{, }\OperatorTok{{-}}\DecValTok{1}\NormalTok{]),}
\NormalTok{    upper}\OperatorTok{=}\NormalTok{np.array([}\DecValTok{1}\NormalTok{, }\DecValTok{1}\NormalTok{]),}
\NormalTok{    n\_points }\OperatorTok{=} \DecValTok{2}
\NormalTok{)}

\CommentTok{\# (S{-}2) Initial Design:}
\NormalTok{spot\_1.X }\OperatorTok{=}\NormalTok{ spot\_1.design.scipy\_lhd(}
\NormalTok{    spot\_1.design\_control[}\StringTok{"init\_size"}\NormalTok{], lower}\OperatorTok{=}\NormalTok{spot\_1.lower, upper}\OperatorTok{=}\NormalTok{spot\_1.upper}
\NormalTok{)}
\BuiltInTok{print}\NormalTok{(spot\_1.X)}

\CommentTok{\# (S{-}3): Eval initial design:}
\NormalTok{spot\_1.y }\OperatorTok{=}\NormalTok{ spot\_1.fun(spot\_1.X)}
\BuiltInTok{print}\NormalTok{(spot\_1.y)}

\NormalTok{spot\_1.surrogate.fit(spot\_1.X, spot\_1.y)}
\NormalTok{X0 }\OperatorTok{=}\NormalTok{ spot\_1.suggest\_new\_X()}
\BuiltInTok{print}\NormalTok{(X0)}
\ControlFlowTok{assert}\NormalTok{ X0.size }\OperatorTok{==}\NormalTok{ spot\_1.n\_points }\OperatorTok{*}\NormalTok{ spot\_1.k}
\end{Highlighting}
\end{Shaded}

\begin{verbatim}
[[ 0.86352963  0.7892358 ]
 [-0.24407197 -0.83687436]
 [ 0.36481882  0.8375811 ]
 [ 0.415331    0.54468512]
 [-0.56395091 -0.77797854]
 [-0.90259409 -0.04899292]
 [-0.16484832  0.35724741]
 [ 0.05170659  0.07401196]
 [-0.78548145 -0.44638164]
 [ 0.64017497 -0.30363301]]
[1.36857656 0.75992983 0.83463487 0.46918172 0.92329124 0.8170764
 0.15480068 0.00815134 0.81623768 0.502017  ]
[[0.00160553 0.00428429]
 [0.00160553 0.00428429]]
\end{verbatim}

\hypertarget{ei-the-famous-schonlau-example}{%
\section{EI: The Famous Schonlau
Example}\label{ei-the-famous-schonlau-example}}

\begin{Shaded}
\begin{Highlighting}[]
\NormalTok{X\_train0 }\OperatorTok{=}\NormalTok{ np.array([}\DecValTok{1}\NormalTok{, }\DecValTok{2}\NormalTok{, }\DecValTok{3}\NormalTok{, }\DecValTok{4}\NormalTok{, }\DecValTok{12}\NormalTok{]).reshape(}\OperatorTok{{-}}\DecValTok{1}\NormalTok{,}\DecValTok{1}\NormalTok{)}
\NormalTok{X\_train }\OperatorTok{=}\NormalTok{ np.linspace(start}\OperatorTok{=}\DecValTok{0}\NormalTok{, stop}\OperatorTok{=}\DecValTok{10}\NormalTok{, num}\OperatorTok{=}\DecValTok{5}\NormalTok{).reshape(}\OperatorTok{{-}}\DecValTok{1}\NormalTok{, }\DecValTok{1}\NormalTok{)}
\end{Highlighting}
\end{Shaded}

\begin{Shaded}
\begin{Highlighting}[]
\ImportTok{from}\NormalTok{ spotPython.build.kriging }\ImportTok{import}\NormalTok{ Kriging}
\ImportTok{import}\NormalTok{ numpy }\ImportTok{as}\NormalTok{ np}
\ImportTok{import}\NormalTok{ matplotlib.pyplot }\ImportTok{as}\NormalTok{ plt}

\NormalTok{X\_train }\OperatorTok{=}\NormalTok{ np.array([}\FloatTok{1.}\NormalTok{, }\FloatTok{2.}\NormalTok{, }\FloatTok{3.}\NormalTok{, }\FloatTok{4.}\NormalTok{, }\FloatTok{12.}\NormalTok{]).reshape(}\OperatorTok{{-}}\DecValTok{1}\NormalTok{,}\DecValTok{1}\NormalTok{)}
\NormalTok{y\_train }\OperatorTok{=}\NormalTok{ np.array([}\FloatTok{0.}\NormalTok{, }\OperatorTok{{-}}\FloatTok{1.75}\NormalTok{, }\OperatorTok{{-}}\DecValTok{2}\NormalTok{, }\OperatorTok{{-}}\FloatTok{0.5}\NormalTok{, }\FloatTok{5.}\NormalTok{])}

\NormalTok{S }\OperatorTok{=}\NormalTok{ Kriging(name}\OperatorTok{=}\StringTok{\textquotesingle{}kriging\textquotesingle{}}\NormalTok{,  seed}\OperatorTok{=}\DecValTok{123}\NormalTok{, log\_level}\OperatorTok{=}\DecValTok{50}\NormalTok{, n\_theta}\OperatorTok{=}\DecValTok{1}\NormalTok{, noise}\OperatorTok{=}\VariableTok{False}\NormalTok{, cod\_type}\OperatorTok{=}\StringTok{"norm"}\NormalTok{)}
\NormalTok{S.fit(X\_train, y\_train)}

\NormalTok{X }\OperatorTok{=}\NormalTok{ np.linspace(start}\OperatorTok{=}\DecValTok{0}\NormalTok{, stop}\OperatorTok{=}\DecValTok{13}\NormalTok{, num}\OperatorTok{=}\DecValTok{1000}\NormalTok{).reshape(}\OperatorTok{{-}}\DecValTok{1}\NormalTok{, }\DecValTok{1}\NormalTok{)}
\NormalTok{mean\_prediction, std\_prediction, ei }\OperatorTok{=}\NormalTok{ S.predict(X, return\_val}\OperatorTok{=}\StringTok{"all"}\NormalTok{)}

\NormalTok{plt.scatter(X\_train, y\_train, label}\OperatorTok{=}\StringTok{"Observations"}\NormalTok{)}
\NormalTok{plt.plot(X, mean\_prediction, label}\OperatorTok{=}\StringTok{"Mean prediction"}\NormalTok{)}
\ControlFlowTok{if} \VariableTok{True}\NormalTok{:}
\NormalTok{    plt.fill\_between(}
\NormalTok{        X.ravel(),}
\NormalTok{        mean\_prediction }\OperatorTok{{-}} \DecValTok{2} \OperatorTok{*}\NormalTok{ std\_prediction,}
\NormalTok{        mean\_prediction }\OperatorTok{+} \DecValTok{2} \OperatorTok{*}\NormalTok{ std\_prediction,}
\NormalTok{        alpha}\OperatorTok{=}\FloatTok{0.5}\NormalTok{,}
\NormalTok{        label}\OperatorTok{=}\VerbatimStringTok{r"95}\SpecialCharTok{\% c}\VerbatimStringTok{onfidence interval"}\NormalTok{,}
\NormalTok{    )}
\NormalTok{plt.legend()}
\NormalTok{plt.xlabel(}\StringTok{"$x$"}\NormalTok{)}
\NormalTok{plt.ylabel(}\StringTok{"$f(x)$"}\NormalTok{)}
\NormalTok{\_ }\OperatorTok{=}\NormalTok{ plt.title(}\StringTok{"Gaussian process regression on noise{-}free dataset"}\NormalTok{)}
\end{Highlighting}
\end{Shaded}

\begin{figure}[H]

{\centering \includegraphics{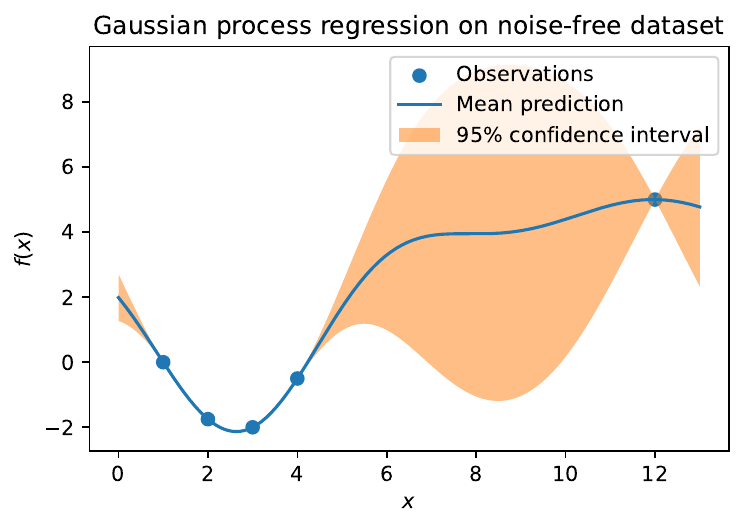}

}

\end{figure}

\begin{Shaded}
\begin{Highlighting}[]
\CommentTok{\#plt.plot(X, y, label=r"$f(x) = x \textbackslash{}sin(x)$", linestyle="dotted")}
\CommentTok{\# plt.scatter(X\_train, y\_train, label="Observations")}
\NormalTok{plt.plot(X, }\OperatorTok{{-}}\NormalTok{ei, label}\OperatorTok{=}\StringTok{"Expected Improvement"}\NormalTok{)}
\NormalTok{plt.legend()}
\NormalTok{plt.xlabel(}\StringTok{"$x$"}\NormalTok{)}
\NormalTok{plt.ylabel(}\StringTok{"$f(x)$"}\NormalTok{)}
\NormalTok{\_ }\OperatorTok{=}\NormalTok{ plt.title(}\StringTok{"Gaussian process regression on noise{-}free dataset"}\NormalTok{)}
\end{Highlighting}
\end{Shaded}

\begin{figure}[H]

{\centering \includegraphics{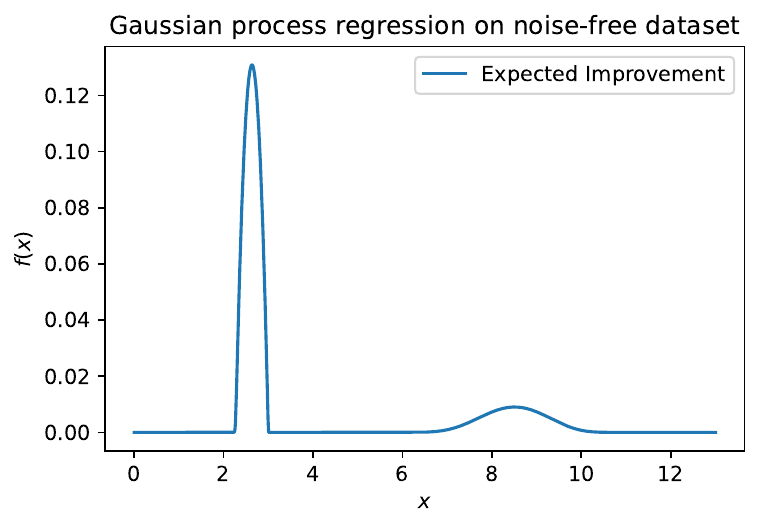}

}

\end{figure}

\begin{Shaded}
\begin{Highlighting}[]
\NormalTok{S.log}
\end{Highlighting}
\end{Shaded}

\begin{verbatim}
{'negLnLike': array([1.20788205]),
 'theta': array([1.09276]),
 'p': [],
 'Lambda': []}
\end{verbatim}

\hypertarget{ei-the-forrester-example}{%
\section{EI: The Forrester Example}\label{ei-the-forrester-example}}

\begin{Shaded}
\begin{Highlighting}[]
\ImportTok{from}\NormalTok{ spotPython.build.kriging }\ImportTok{import}\NormalTok{ Kriging}
\ImportTok{import}\NormalTok{ numpy }\ImportTok{as}\NormalTok{ np}
\ImportTok{import}\NormalTok{ matplotlib.pyplot }\ImportTok{as}\NormalTok{ plt}
\ImportTok{import}\NormalTok{ spotPython}
\ImportTok{from}\NormalTok{ spotPython.fun.objectivefunctions }\ImportTok{import}\NormalTok{ analytical}
\ImportTok{from}\NormalTok{ spotPython.spot }\ImportTok{import}\NormalTok{ spot}

\CommentTok{\# exact x locations are unknown:}
\NormalTok{X\_train }\OperatorTok{=}\NormalTok{ np.array([}\FloatTok{0.0}\NormalTok{, }\FloatTok{0.175}\NormalTok{, }\FloatTok{0.225}\NormalTok{, }\FloatTok{0.3}\NormalTok{, }\FloatTok{0.35}\NormalTok{, }\FloatTok{0.375}\NormalTok{, }\FloatTok{0.5}\NormalTok{,}\DecValTok{1}\NormalTok{]).reshape(}\OperatorTok{{-}}\DecValTok{1}\NormalTok{,}\DecValTok{1}\NormalTok{)}

\NormalTok{fun }\OperatorTok{=}\NormalTok{ analytical().fun\_forrester}
\NormalTok{fun\_control }\OperatorTok{=}\NormalTok{ fun\_control\_init(}
\NormalTok{    spot\_tensorboard\_path}\OperatorTok{=}\NormalTok{get\_spot\_tensorboard\_path(experiment\_name),}
\NormalTok{    sigma}\OperatorTok{=}\FloatTok{1.0}\NormalTok{,}
\NormalTok{    seed}\OperatorTok{=}\DecValTok{123}\NormalTok{,)}
\NormalTok{y\_train }\OperatorTok{=}\NormalTok{ fun(X\_train, fun\_control}\OperatorTok{=}\NormalTok{fun\_control)}

\NormalTok{S }\OperatorTok{=}\NormalTok{ Kriging(name}\OperatorTok{=}\StringTok{\textquotesingle{}kriging\textquotesingle{}}\NormalTok{,  seed}\OperatorTok{=}\DecValTok{123}\NormalTok{, log\_level}\OperatorTok{=}\DecValTok{50}\NormalTok{, n\_theta}\OperatorTok{=}\DecValTok{1}\NormalTok{, noise}\OperatorTok{=}\VariableTok{False}\NormalTok{, cod\_type}\OperatorTok{=}\StringTok{"norm"}\NormalTok{)}
\NormalTok{S.fit(X\_train, y\_train)}

\NormalTok{X }\OperatorTok{=}\NormalTok{ np.linspace(start}\OperatorTok{=}\DecValTok{0}\NormalTok{, stop}\OperatorTok{=}\DecValTok{1}\NormalTok{, num}\OperatorTok{=}\DecValTok{1000}\NormalTok{).reshape(}\OperatorTok{{-}}\DecValTok{1}\NormalTok{, }\DecValTok{1}\NormalTok{)}
\NormalTok{mean\_prediction, std\_prediction, ei }\OperatorTok{=}\NormalTok{ S.predict(X, return\_val}\OperatorTok{=}\StringTok{"all"}\NormalTok{)}

\NormalTok{plt.scatter(X\_train, y\_train, label}\OperatorTok{=}\StringTok{"Observations"}\NormalTok{)}
\NormalTok{plt.plot(X, mean\_prediction, label}\OperatorTok{=}\StringTok{"Mean prediction"}\NormalTok{)}
\ControlFlowTok{if} \VariableTok{True}\NormalTok{:}
\NormalTok{    plt.fill\_between(}
\NormalTok{        X.ravel(),}
\NormalTok{        mean\_prediction }\OperatorTok{{-}} \DecValTok{2} \OperatorTok{*}\NormalTok{ std\_prediction,}
\NormalTok{        mean\_prediction }\OperatorTok{+} \DecValTok{2} \OperatorTok{*}\NormalTok{ std\_prediction,}
\NormalTok{        alpha}\OperatorTok{=}\FloatTok{0.5}\NormalTok{,}
\NormalTok{        label}\OperatorTok{=}\VerbatimStringTok{r"95}\SpecialCharTok{\% c}\VerbatimStringTok{onfidence interval"}\NormalTok{,}
\NormalTok{    )}
\NormalTok{plt.legend()}
\NormalTok{plt.xlabel(}\StringTok{"$x$"}\NormalTok{)}
\NormalTok{plt.ylabel(}\StringTok{"$f(x)$"}\NormalTok{)}
\NormalTok{\_ }\OperatorTok{=}\NormalTok{ plt.title(}\StringTok{"Gaussian process regression on noise{-}free dataset"}\NormalTok{)}
\end{Highlighting}
\end{Shaded}

\begin{figure}[H]

{\centering \includegraphics{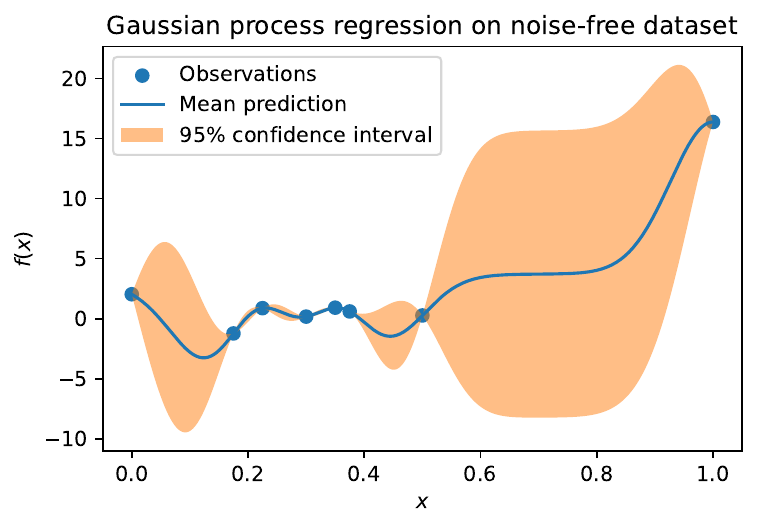}

}

\end{figure}

\begin{Shaded}
\begin{Highlighting}[]
\CommentTok{\#plt.plot(X, y, label=r"$f(x) = x \textbackslash{}sin(x)$", linestyle="dotted")}
\CommentTok{\# plt.scatter(X\_train, y\_train, label="Observations")}
\NormalTok{plt.plot(X, }\OperatorTok{{-}}\NormalTok{ei, label}\OperatorTok{=}\StringTok{"Expected Improvement"}\NormalTok{)}
\NormalTok{plt.legend()}
\NormalTok{plt.xlabel(}\StringTok{"$x$"}\NormalTok{)}
\NormalTok{plt.ylabel(}\StringTok{"$f(x)$"}\NormalTok{)}
\NormalTok{\_ }\OperatorTok{=}\NormalTok{ plt.title(}\StringTok{"Gaussian process regression on noise{-}free dataset"}\NormalTok{)}
\end{Highlighting}
\end{Shaded}

\begin{figure}[H]

{\centering \includegraphics{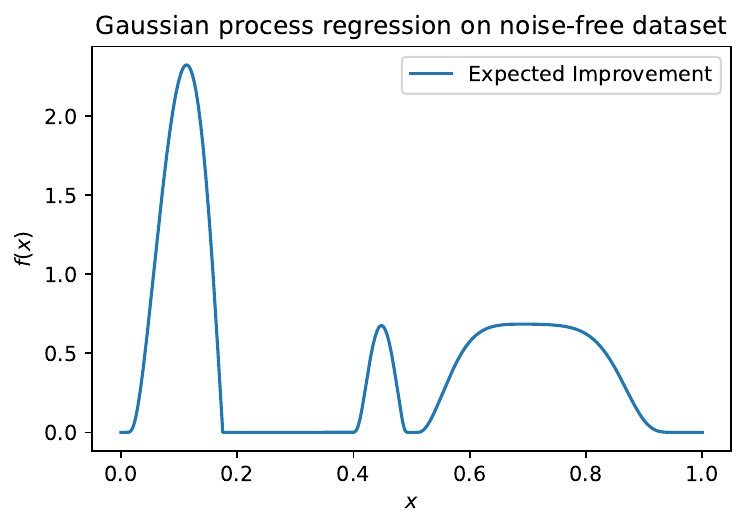}

}

\end{figure}

\hypertarget{noise}{%
\section{Noise}\label{noise}}

\begin{Shaded}
\begin{Highlighting}[]
\ImportTok{import}\NormalTok{ numpy }\ImportTok{as}\NormalTok{ np}
\ImportTok{import}\NormalTok{ spotPython}
\ImportTok{from}\NormalTok{ spotPython.fun.objectivefunctions }\ImportTok{import}\NormalTok{ analytical}
\ImportTok{from}\NormalTok{ spotPython.spot }\ImportTok{import}\NormalTok{ spot}
\ImportTok{from}\NormalTok{ spotPython.design.spacefilling }\ImportTok{import}\NormalTok{ spacefilling}
\ImportTok{from}\NormalTok{ spotPython.build.kriging }\ImportTok{import}\NormalTok{ Kriging}
\ImportTok{import}\NormalTok{ matplotlib.pyplot }\ImportTok{as}\NormalTok{ plt}

\NormalTok{gen }\OperatorTok{=}\NormalTok{ spacefilling(}\DecValTok{1}\NormalTok{)}
\NormalTok{rng }\OperatorTok{=}\NormalTok{ np.random.RandomState(}\DecValTok{1}\NormalTok{)}
\NormalTok{lower }\OperatorTok{=}\NormalTok{ np.array([}\OperatorTok{{-}}\DecValTok{10}\NormalTok{])}
\NormalTok{upper }\OperatorTok{=}\NormalTok{ np.array([}\DecValTok{10}\NormalTok{])}
\NormalTok{fun }\OperatorTok{=}\NormalTok{ analytical().fun\_sphere}
\NormalTok{fun\_control }\OperatorTok{=}\NormalTok{ fun\_control\_init(}
\NormalTok{    spot\_tensorboard\_path}\OperatorTok{=}\NormalTok{get\_spot\_tensorboard\_path(experiment\_name),}
\NormalTok{    sigma}\OperatorTok{=}\FloatTok{2.0}\NormalTok{,}
\NormalTok{    seed}\OperatorTok{=}\DecValTok{123}\NormalTok{,)}
\NormalTok{X }\OperatorTok{=}\NormalTok{ gen.scipy\_lhd(}\DecValTok{10}\NormalTok{, lower}\OperatorTok{=}\NormalTok{lower, upper }\OperatorTok{=}\NormalTok{ upper)}
\BuiltInTok{print}\NormalTok{(X)}
\NormalTok{y }\OperatorTok{=}\NormalTok{ fun(X, fun\_control}\OperatorTok{=}\NormalTok{fun\_control)}
\BuiltInTok{print}\NormalTok{(y)}
\NormalTok{y.shape}
\NormalTok{X\_train }\OperatorTok{=}\NormalTok{ X.reshape(}\OperatorTok{{-}}\DecValTok{1}\NormalTok{,}\DecValTok{1}\NormalTok{)}
\NormalTok{y\_train }\OperatorTok{=}\NormalTok{ y}

\NormalTok{S }\OperatorTok{=}\NormalTok{ Kriging(name}\OperatorTok{=}\StringTok{\textquotesingle{}kriging\textquotesingle{}}\NormalTok{,}
\NormalTok{            seed}\OperatorTok{=}\DecValTok{123}\NormalTok{,}
\NormalTok{            log\_level}\OperatorTok{=}\DecValTok{50}\NormalTok{,}
\NormalTok{            n\_theta}\OperatorTok{=}\DecValTok{1}\NormalTok{,}
\NormalTok{            noise}\OperatorTok{=}\VariableTok{False}\NormalTok{)}
\NormalTok{S.fit(X\_train, y\_train)}

\NormalTok{X\_axis }\OperatorTok{=}\NormalTok{ np.linspace(start}\OperatorTok{={-}}\DecValTok{13}\NormalTok{, stop}\OperatorTok{=}\DecValTok{13}\NormalTok{, num}\OperatorTok{=}\DecValTok{1000}\NormalTok{).reshape(}\OperatorTok{{-}}\DecValTok{1}\NormalTok{, }\DecValTok{1}\NormalTok{)}
\NormalTok{mean\_prediction, std\_prediction, ei }\OperatorTok{=}\NormalTok{ S.predict(X\_axis, return\_val}\OperatorTok{=}\StringTok{"all"}\NormalTok{)}

\CommentTok{\#plt.plot(X, y, label=r"$f(x) = x \textbackslash{}sin(x)$", linestyle="dotted")}
\NormalTok{plt.scatter(X\_train, y\_train, label}\OperatorTok{=}\StringTok{"Observations"}\NormalTok{)}
\CommentTok{\#plt.plot(X, ei, label="Expected Improvement")}
\NormalTok{plt.plot(X\_axis, mean\_prediction, label}\OperatorTok{=}\StringTok{"mue"}\NormalTok{)}
\NormalTok{plt.legend()}
\NormalTok{plt.xlabel(}\StringTok{"$x$"}\NormalTok{)}
\NormalTok{plt.ylabel(}\StringTok{"$f(x)$"}\NormalTok{)}
\NormalTok{\_ }\OperatorTok{=}\NormalTok{ plt.title(}\StringTok{"Sphere: Gaussian process regression on noisy dataset"}\NormalTok{)}
\end{Highlighting}
\end{Shaded}

\begin{verbatim}
[[ 0.63529627]
 [-4.10764204]
 [-0.44071975]
 [ 9.63125638]
 [-8.3518118 ]
 [-3.62418901]
 [ 4.15331   ]
 [ 3.4468512 ]
 [ 6.36049088]
 [-7.77978539]]
[-1.57464135 16.13714981  2.77008442 93.14904827 71.59322218 14.28895359
 15.9770567  12.96468767 39.82265329 59.88028242]
\end{verbatim}

\begin{figure}[H]

{\centering \includegraphics{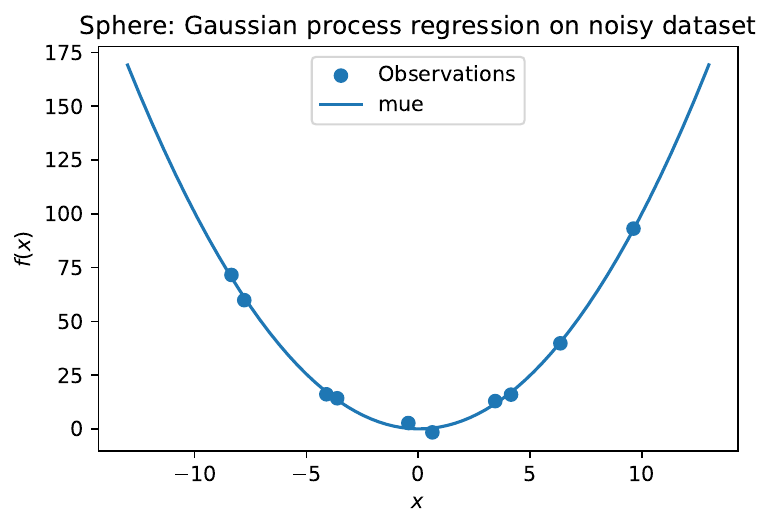}

}

\end{figure}

\begin{Shaded}
\begin{Highlighting}[]
\NormalTok{S.log}
\end{Highlighting}
\end{Shaded}

\begin{verbatim}
{'negLnLike': array([25.26601605]),
 'theta': array([-1.98024488]),
 'p': [],
 'Lambda': []}
\end{verbatim}

\begin{Shaded}
\begin{Highlighting}[]
\NormalTok{S }\OperatorTok{=}\NormalTok{ Kriging(name}\OperatorTok{=}\StringTok{\textquotesingle{}kriging\textquotesingle{}}\NormalTok{,}
\NormalTok{            seed}\OperatorTok{=}\DecValTok{123}\NormalTok{,}
\NormalTok{            log\_level}\OperatorTok{=}\DecValTok{50}\NormalTok{,}
\NormalTok{            n\_theta}\OperatorTok{=}\DecValTok{1}\NormalTok{,}
\NormalTok{            noise}\OperatorTok{=}\VariableTok{True}\NormalTok{)}
\NormalTok{S.fit(X\_train, y\_train)}

\NormalTok{X\_axis }\OperatorTok{=}\NormalTok{ np.linspace(start}\OperatorTok{={-}}\DecValTok{13}\NormalTok{, stop}\OperatorTok{=}\DecValTok{13}\NormalTok{, num}\OperatorTok{=}\DecValTok{1000}\NormalTok{).reshape(}\OperatorTok{{-}}\DecValTok{1}\NormalTok{, }\DecValTok{1}\NormalTok{)}
\NormalTok{mean\_prediction, std\_prediction, ei }\OperatorTok{=}\NormalTok{ S.predict(X\_axis, return\_val}\OperatorTok{=}\StringTok{"all"}\NormalTok{)}

\CommentTok{\#plt.plot(X, y, label=r"$f(x) = x \textbackslash{}sin(x)$", linestyle="dotted")}
\NormalTok{plt.scatter(X\_train, y\_train, label}\OperatorTok{=}\StringTok{"Observations"}\NormalTok{)}
\CommentTok{\#plt.plot(X, ei, label="Expected Improvement")}
\NormalTok{plt.plot(X\_axis, mean\_prediction, label}\OperatorTok{=}\StringTok{"mue"}\NormalTok{)}
\NormalTok{plt.legend()}
\NormalTok{plt.xlabel(}\StringTok{"$x$"}\NormalTok{)}
\NormalTok{plt.ylabel(}\StringTok{"$f(x)$"}\NormalTok{)}
\NormalTok{\_ }\OperatorTok{=}\NormalTok{ plt.title(}\StringTok{"Sphere: Gaussian process regression with nugget on noisy dataset"}\NormalTok{)}
\end{Highlighting}
\end{Shaded}

\begin{figure}[H]

{\centering \includegraphics{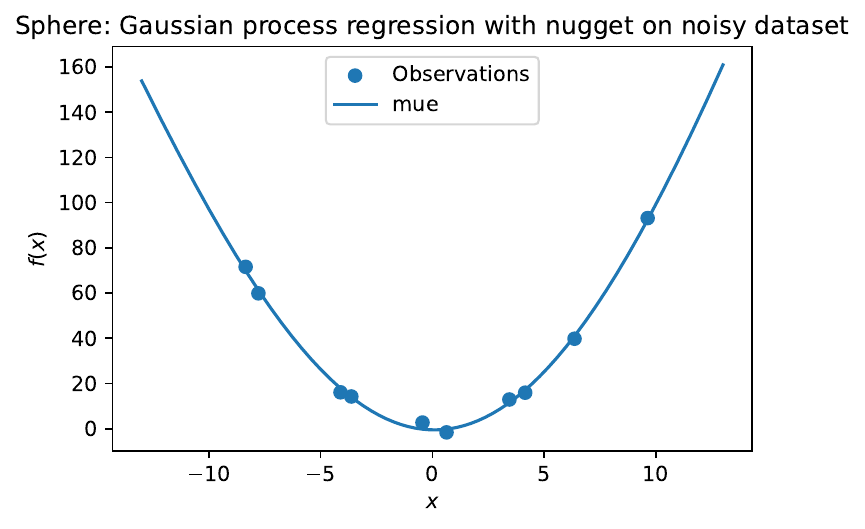}

}

\end{figure}

\begin{Shaded}
\begin{Highlighting}[]
\NormalTok{S.log}
\end{Highlighting}
\end{Shaded}

\begin{verbatim}
{'negLnLike': array([21.82530943]),
 'theta': array([-0.41935831]),
 'p': [],
 'Lambda': array([5.20850907e-05])}
\end{verbatim}

\hypertarget{cubic-function}{%
\section{Cubic Function}\label{cubic-function}}

\begin{Shaded}
\begin{Highlighting}[]
\ImportTok{import}\NormalTok{ numpy }\ImportTok{as}\NormalTok{ np}
\ImportTok{import}\NormalTok{ spotPython}
\ImportTok{from}\NormalTok{ spotPython.fun.objectivefunctions }\ImportTok{import}\NormalTok{ analytical}
\ImportTok{from}\NormalTok{ spotPython.spot }\ImportTok{import}\NormalTok{ spot}
\ImportTok{from}\NormalTok{ spotPython.design.spacefilling }\ImportTok{import}\NormalTok{ spacefilling}
\ImportTok{from}\NormalTok{ spotPython.build.kriging }\ImportTok{import}\NormalTok{ Kriging}
\ImportTok{import}\NormalTok{ matplotlib.pyplot }\ImportTok{as}\NormalTok{ plt}

\NormalTok{gen }\OperatorTok{=}\NormalTok{ spacefilling(}\DecValTok{1}\NormalTok{)}
\NormalTok{rng }\OperatorTok{=}\NormalTok{ np.random.RandomState(}\DecValTok{1}\NormalTok{)}
\NormalTok{lower }\OperatorTok{=}\NormalTok{ np.array([}\OperatorTok{{-}}\DecValTok{10}\NormalTok{])}
\NormalTok{upper }\OperatorTok{=}\NormalTok{ np.array([}\DecValTok{10}\NormalTok{])}
\NormalTok{fun }\OperatorTok{=}\NormalTok{ analytical().fun\_cubed}
\NormalTok{fun\_control }\OperatorTok{=}\NormalTok{ fun\_control\_init(}
\NormalTok{    spot\_tensorboard\_path}\OperatorTok{=}\NormalTok{get\_spot\_tensorboard\_path(experiment\_name),}
\NormalTok{    sigma}\OperatorTok{=}\FloatTok{10.0}\NormalTok{,}
\NormalTok{    seed}\OperatorTok{=}\DecValTok{123}\NormalTok{,)}

\NormalTok{X }\OperatorTok{=}\NormalTok{ gen.scipy\_lhd(}\DecValTok{10}\NormalTok{, lower}\OperatorTok{=}\NormalTok{lower, upper }\OperatorTok{=}\NormalTok{ upper)}
\BuiltInTok{print}\NormalTok{(X)}
\NormalTok{y }\OperatorTok{=}\NormalTok{ fun(X, fun\_control}\OperatorTok{=}\NormalTok{fun\_control)}
\BuiltInTok{print}\NormalTok{(y)}
\NormalTok{y.shape}
\NormalTok{X\_train }\OperatorTok{=}\NormalTok{ X.reshape(}\OperatorTok{{-}}\DecValTok{1}\NormalTok{,}\DecValTok{1}\NormalTok{)}
\NormalTok{y\_train }\OperatorTok{=}\NormalTok{ y}

\NormalTok{S }\OperatorTok{=}\NormalTok{ Kriging(name}\OperatorTok{=}\StringTok{\textquotesingle{}kriging\textquotesingle{}}\NormalTok{,  seed}\OperatorTok{=}\DecValTok{123}\NormalTok{, log\_level}\OperatorTok{=}\DecValTok{50}\NormalTok{, n\_theta}\OperatorTok{=}\DecValTok{1}\NormalTok{, noise}\OperatorTok{=}\VariableTok{False}\NormalTok{)}
\NormalTok{S.fit(X\_train, y\_train)}

\NormalTok{X\_axis }\OperatorTok{=}\NormalTok{ np.linspace(start}\OperatorTok{={-}}\DecValTok{13}\NormalTok{, stop}\OperatorTok{=}\DecValTok{13}\NormalTok{, num}\OperatorTok{=}\DecValTok{1000}\NormalTok{).reshape(}\OperatorTok{{-}}\DecValTok{1}\NormalTok{, }\DecValTok{1}\NormalTok{)}
\NormalTok{mean\_prediction, std\_prediction, ei }\OperatorTok{=}\NormalTok{ S.predict(X\_axis, return\_val}\OperatorTok{=}\StringTok{"all"}\NormalTok{)}

\NormalTok{plt.scatter(X\_train, y\_train, label}\OperatorTok{=}\StringTok{"Observations"}\NormalTok{)}
\CommentTok{\#plt.plot(X, ei, label="Expected Improvement")}
\NormalTok{plt.plot(X\_axis, mean\_prediction, label}\OperatorTok{=}\StringTok{"mue"}\NormalTok{)}
\NormalTok{plt.legend()}
\NormalTok{plt.xlabel(}\StringTok{"$x$"}\NormalTok{)}
\NormalTok{plt.ylabel(}\StringTok{"$f(x)$"}\NormalTok{)}
\NormalTok{\_ }\OperatorTok{=}\NormalTok{ plt.title(}\StringTok{"Cubed: Gaussian process regression on noisy dataset"}\NormalTok{)}
\end{Highlighting}
\end{Shaded}

\begin{verbatim}
[[ 0.63529627]
 [-4.10764204]
 [-0.44071975]
 [ 9.63125638]
 [-8.3518118 ]
 [-3.62418901]
 [ 4.15331   ]
 [ 3.4468512 ]
 [ 6.36049088]
 [-7.77978539]]
[  -9.63480707  -72.98497325   12.7936499   895.34567477 -573.35961837
  -41.83176425   65.27989461   46.37081417  254.1530734  -474.09587355]
\end{verbatim}

\begin{figure}[H]

{\centering \includegraphics{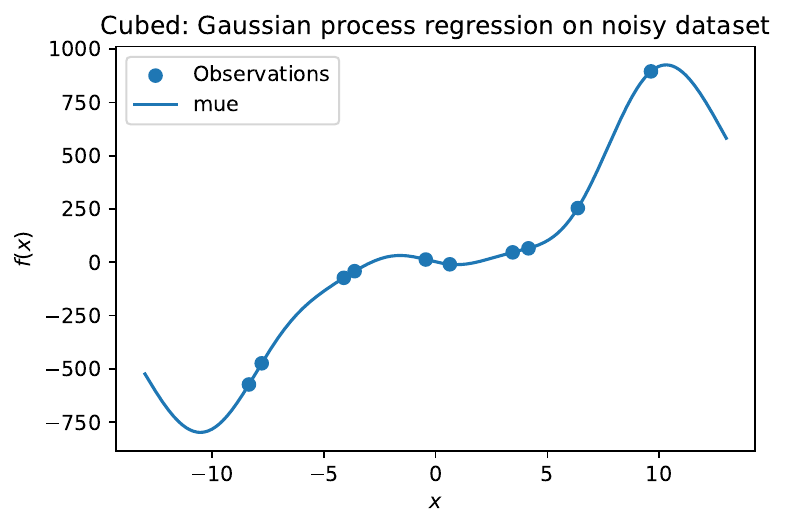}

}

\end{figure}

\begin{Shaded}
\begin{Highlighting}[]
\NormalTok{S }\OperatorTok{=}\NormalTok{ Kriging(name}\OperatorTok{=}\StringTok{\textquotesingle{}kriging\textquotesingle{}}\NormalTok{,  seed}\OperatorTok{=}\DecValTok{123}\NormalTok{, log\_level}\OperatorTok{=}\DecValTok{0}\NormalTok{, n\_theta}\OperatorTok{=}\DecValTok{1}\NormalTok{, noise}\OperatorTok{=}\VariableTok{True}\NormalTok{)}
\NormalTok{S.fit(X\_train, y\_train)}

\NormalTok{X\_axis }\OperatorTok{=}\NormalTok{ np.linspace(start}\OperatorTok{={-}}\DecValTok{13}\NormalTok{, stop}\OperatorTok{=}\DecValTok{13}\NormalTok{, num}\OperatorTok{=}\DecValTok{1000}\NormalTok{).reshape(}\OperatorTok{{-}}\DecValTok{1}\NormalTok{, }\DecValTok{1}\NormalTok{)}
\NormalTok{mean\_prediction, std\_prediction, ei }\OperatorTok{=}\NormalTok{ S.predict(X\_axis, return\_val}\OperatorTok{=}\StringTok{"all"}\NormalTok{)}

\NormalTok{plt.scatter(X\_train, y\_train, label}\OperatorTok{=}\StringTok{"Observations"}\NormalTok{)}
\CommentTok{\#plt.plot(X, ei, label="Expected Improvement")}
\NormalTok{plt.plot(X\_axis, mean\_prediction, label}\OperatorTok{=}\StringTok{"mue"}\NormalTok{)}
\NormalTok{plt.legend()}
\NormalTok{plt.xlabel(}\StringTok{"$x$"}\NormalTok{)}
\NormalTok{plt.ylabel(}\StringTok{"$f(x)$"}\NormalTok{)}
\NormalTok{\_ }\OperatorTok{=}\NormalTok{ plt.title(}\StringTok{"Cubed: Gaussian process with nugget regression on noisy dataset"}\NormalTok{)}
\end{Highlighting}
\end{Shaded}

\begin{figure}[H]

{\centering \includegraphics{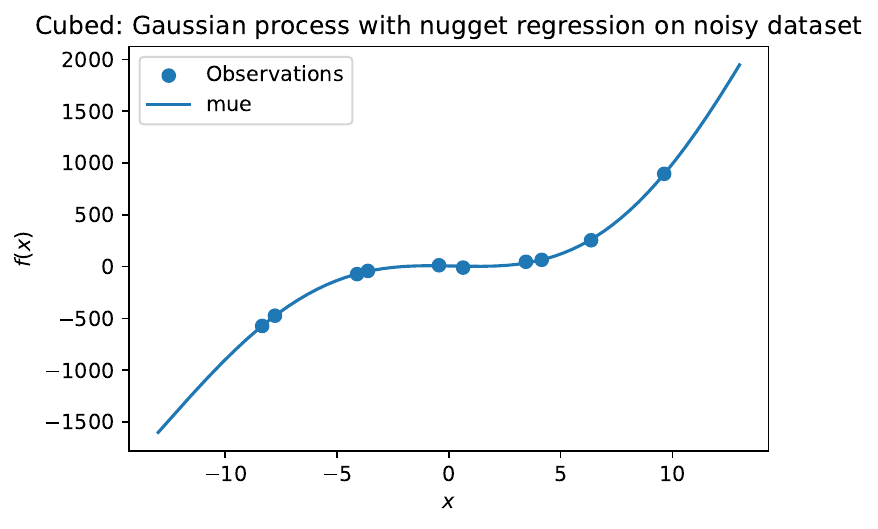}

}

\end{figure}

\begin{Shaded}
\begin{Highlighting}[]
\ImportTok{import}\NormalTok{ numpy }\ImportTok{as}\NormalTok{ np}
\ImportTok{import}\NormalTok{ spotPython}
\ImportTok{from}\NormalTok{ spotPython.fun.objectivefunctions }\ImportTok{import}\NormalTok{ analytical}
\ImportTok{from}\NormalTok{ spotPython.spot }\ImportTok{import}\NormalTok{ spot}
\ImportTok{from}\NormalTok{ spotPython.design.spacefilling }\ImportTok{import}\NormalTok{ spacefilling}
\ImportTok{from}\NormalTok{ spotPython.build.kriging }\ImportTok{import}\NormalTok{ Kriging}
\ImportTok{import}\NormalTok{ matplotlib.pyplot }\ImportTok{as}\NormalTok{ plt}

\NormalTok{gen }\OperatorTok{=}\NormalTok{ spacefilling(}\DecValTok{1}\NormalTok{)}
\NormalTok{rng }\OperatorTok{=}\NormalTok{ np.random.RandomState(}\DecValTok{1}\NormalTok{)}
\NormalTok{lower }\OperatorTok{=}\NormalTok{ np.array([}\OperatorTok{{-}}\DecValTok{10}\NormalTok{])}
\NormalTok{upper }\OperatorTok{=}\NormalTok{ np.array([}\DecValTok{10}\NormalTok{])}
\NormalTok{fun }\OperatorTok{=}\NormalTok{ analytical().fun\_runge}
\NormalTok{fun\_control }\OperatorTok{=}\NormalTok{ fun\_control\_init(}
\NormalTok{    spot\_tensorboard\_path}\OperatorTok{=}\NormalTok{get\_spot\_tensorboard\_path(experiment\_name),}
\NormalTok{    sigma}\OperatorTok{=}\FloatTok{0.25}\NormalTok{,}
\NormalTok{    seed}\OperatorTok{=}\DecValTok{123}\NormalTok{,)}

\NormalTok{X }\OperatorTok{=}\NormalTok{ gen.scipy\_lhd(}\DecValTok{10}\NormalTok{, lower}\OperatorTok{=}\NormalTok{lower, upper }\OperatorTok{=}\NormalTok{ upper)}
\BuiltInTok{print}\NormalTok{(X)}
\NormalTok{y }\OperatorTok{=}\NormalTok{ fun(X, fun\_control}\OperatorTok{=}\NormalTok{fun\_control)}
\BuiltInTok{print}\NormalTok{(y)}
\NormalTok{y.shape}
\NormalTok{X\_train }\OperatorTok{=}\NormalTok{ X.reshape(}\OperatorTok{{-}}\DecValTok{1}\NormalTok{,}\DecValTok{1}\NormalTok{)}
\NormalTok{y\_train }\OperatorTok{=}\NormalTok{ y}

\NormalTok{S }\OperatorTok{=}\NormalTok{ Kriging(name}\OperatorTok{=}\StringTok{\textquotesingle{}kriging\textquotesingle{}}\NormalTok{,  seed}\OperatorTok{=}\DecValTok{123}\NormalTok{, log\_level}\OperatorTok{=}\DecValTok{50}\NormalTok{, n\_theta}\OperatorTok{=}\DecValTok{1}\NormalTok{, noise}\OperatorTok{=}\VariableTok{False}\NormalTok{)}
\NormalTok{S.fit(X\_train, y\_train)}

\NormalTok{X\_axis }\OperatorTok{=}\NormalTok{ np.linspace(start}\OperatorTok{={-}}\DecValTok{13}\NormalTok{, stop}\OperatorTok{=}\DecValTok{13}\NormalTok{, num}\OperatorTok{=}\DecValTok{1000}\NormalTok{).reshape(}\OperatorTok{{-}}\DecValTok{1}\NormalTok{, }\DecValTok{1}\NormalTok{)}
\NormalTok{mean\_prediction, std\_prediction, ei }\OperatorTok{=}\NormalTok{ S.predict(X\_axis, return\_val}\OperatorTok{=}\StringTok{"all"}\NormalTok{)}

\NormalTok{plt.scatter(X\_train, y\_train, label}\OperatorTok{=}\StringTok{"Observations"}\NormalTok{)}
\CommentTok{\#plt.plot(X, ei, label="Expected Improvement")}
\NormalTok{plt.plot(X\_axis, mean\_prediction, label}\OperatorTok{=}\StringTok{"mue"}\NormalTok{)}
\NormalTok{plt.legend()}
\NormalTok{plt.xlabel(}\StringTok{"$x$"}\NormalTok{)}
\NormalTok{plt.ylabel(}\StringTok{"$f(x)$"}\NormalTok{)}
\NormalTok{\_ }\OperatorTok{=}\NormalTok{ plt.title(}\StringTok{"Gaussian process regression on noisy dataset"}\NormalTok{)}
\end{Highlighting}
\end{Shaded}

\begin{verbatim}
[[ 0.63529627]
 [-4.10764204]
 [-0.44071975]
 [ 9.63125638]
 [-8.3518118 ]
 [-3.62418901]
 [ 4.15331   ]
 [ 3.4468512 ]
 [ 6.36049088]
 [-7.77978539]]
[0.712453   0.05595118 0.83735691 0.0106654  0.01413372 0.07074765
 0.05479457 0.07763503 0.02412205 0.01625354]
\end{verbatim}

\begin{figure}[H]

{\centering \includegraphics{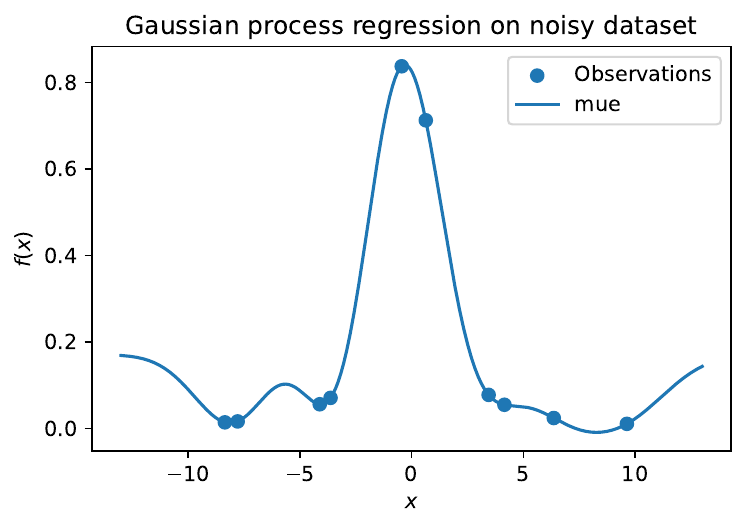}

}

\end{figure}

\begin{Shaded}
\begin{Highlighting}[]
\NormalTok{S }\OperatorTok{=}\NormalTok{ Kriging(name}\OperatorTok{=}\StringTok{\textquotesingle{}kriging\textquotesingle{}}\NormalTok{,}
\NormalTok{            seed}\OperatorTok{=}\DecValTok{123}\NormalTok{,}
\NormalTok{            log\_level}\OperatorTok{=}\DecValTok{50}\NormalTok{,}
\NormalTok{            n\_theta}\OperatorTok{=}\DecValTok{1}\NormalTok{,}
\NormalTok{            noise}\OperatorTok{=}\VariableTok{True}\NormalTok{)}
\NormalTok{S.fit(X\_train, y\_train)}

\NormalTok{X\_axis }\OperatorTok{=}\NormalTok{ np.linspace(start}\OperatorTok{={-}}\DecValTok{13}\NormalTok{, stop}\OperatorTok{=}\DecValTok{13}\NormalTok{, num}\OperatorTok{=}\DecValTok{1000}\NormalTok{).reshape(}\OperatorTok{{-}}\DecValTok{1}\NormalTok{, }\DecValTok{1}\NormalTok{)}
\NormalTok{mean\_prediction, std\_prediction, ei }\OperatorTok{=}\NormalTok{ S.predict(X\_axis, return\_val}\OperatorTok{=}\StringTok{"all"}\NormalTok{)}

\NormalTok{plt.scatter(X\_train, y\_train, label}\OperatorTok{=}\StringTok{"Observations"}\NormalTok{)}
\CommentTok{\#plt.plot(X, ei, label="Expected Improvement")}
\NormalTok{plt.plot(X\_axis, mean\_prediction, label}\OperatorTok{=}\StringTok{"mue"}\NormalTok{)}
\NormalTok{plt.legend()}
\NormalTok{plt.xlabel(}\StringTok{"$x$"}\NormalTok{)}
\NormalTok{plt.ylabel(}\StringTok{"$f(x)$"}\NormalTok{)}
\NormalTok{\_ }\OperatorTok{=}\NormalTok{ plt.title(}\StringTok{"Gaussian process regression with nugget on noisy dataset"}\NormalTok{)}
\end{Highlighting}
\end{Shaded}

\begin{figure}[H]

{\centering \includegraphics{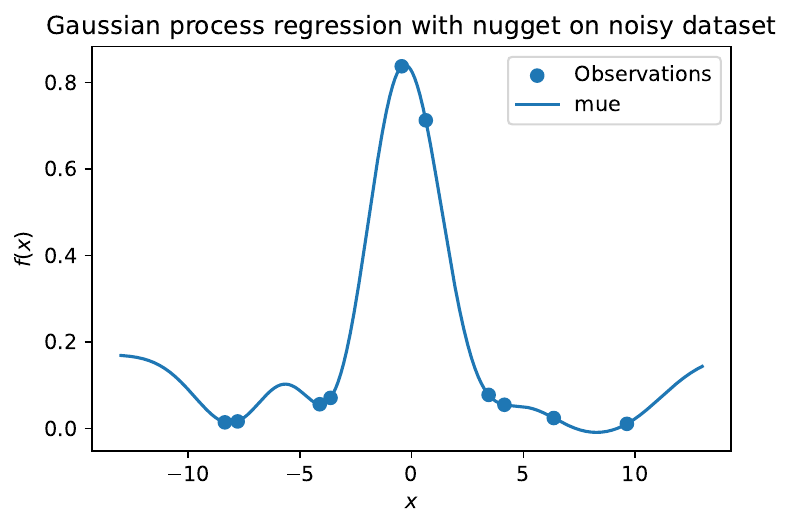}

}

\end{figure}

\hypertarget{factors}{%
\section{Factors}\label{factors}}

\begin{Shaded}
\begin{Highlighting}[]
\NormalTok{[}\StringTok{"num"}\NormalTok{] }\OperatorTok{*} \DecValTok{3}
\end{Highlighting}
\end{Shaded}

\begin{verbatim}
['num', 'num', 'num']
\end{verbatim}

\begin{Shaded}
\begin{Highlighting}[]
\ImportTok{from}\NormalTok{ spotPython.design.spacefilling }\ImportTok{import}\NormalTok{ spacefilling}
\ImportTok{from}\NormalTok{ spotPython.build.kriging }\ImportTok{import}\NormalTok{ Kriging}
\ImportTok{from}\NormalTok{ spotPython.fun.objectivefunctions }\ImportTok{import}\NormalTok{ analytical}
\ImportTok{import}\NormalTok{ numpy }\ImportTok{as}\NormalTok{ np}
\end{Highlighting}
\end{Shaded}

\begin{Shaded}
\begin{Highlighting}[]
\NormalTok{gen }\OperatorTok{=}\NormalTok{ spacefilling(}\DecValTok{2}\NormalTok{)}
\NormalTok{n }\OperatorTok{=} \DecValTok{30}
\NormalTok{rng }\OperatorTok{=}\NormalTok{ np.random.RandomState(}\DecValTok{1}\NormalTok{)}
\NormalTok{lower }\OperatorTok{=}\NormalTok{ np.array([}\OperatorTok{{-}}\DecValTok{5}\NormalTok{,}\OperatorTok{{-}}\DecValTok{0}\NormalTok{])}
\NormalTok{upper }\OperatorTok{=}\NormalTok{ np.array([}\DecValTok{10}\NormalTok{,}\DecValTok{15}\NormalTok{])}
\NormalTok{fun }\OperatorTok{=}\NormalTok{ analytical().fun\_branin\_factor}
\CommentTok{\#fun = analytical(sigma=0).fun\_sphere}

\NormalTok{X0 }\OperatorTok{=}\NormalTok{ gen.scipy\_lhd(n, lower}\OperatorTok{=}\NormalTok{lower, upper }\OperatorTok{=}\NormalTok{ upper)}
\NormalTok{X1 }\OperatorTok{=}\NormalTok{ np.random.randint(low}\OperatorTok{=}\DecValTok{1}\NormalTok{, high}\OperatorTok{=}\DecValTok{3}\NormalTok{, size}\OperatorTok{=}\NormalTok{(n,))}
\NormalTok{X }\OperatorTok{=}\NormalTok{ np.c\_[X0, X1]}
\NormalTok{y }\OperatorTok{=}\NormalTok{ fun(X)}
\NormalTok{S }\OperatorTok{=}\NormalTok{ Kriging(name}\OperatorTok{=}\StringTok{\textquotesingle{}kriging\textquotesingle{}}\NormalTok{,  seed}\OperatorTok{=}\DecValTok{123}\NormalTok{, log\_level}\OperatorTok{=}\DecValTok{50}\NormalTok{, n\_theta}\OperatorTok{=}\DecValTok{3}\NormalTok{, noise}\OperatorTok{=}\VariableTok{False}\NormalTok{, var\_type}\OperatorTok{=}\NormalTok{[}\StringTok{"num"}\NormalTok{, }\StringTok{"num"}\NormalTok{, }\StringTok{"num"}\NormalTok{])}
\NormalTok{S.fit(X, y)}
\NormalTok{Sf }\OperatorTok{=}\NormalTok{ Kriging(name}\OperatorTok{=}\StringTok{\textquotesingle{}kriging\textquotesingle{}}\NormalTok{,  seed}\OperatorTok{=}\DecValTok{123}\NormalTok{, log\_level}\OperatorTok{=}\DecValTok{50}\NormalTok{, n\_theta}\OperatorTok{=}\DecValTok{3}\NormalTok{, noise}\OperatorTok{=}\VariableTok{False}\NormalTok{, var\_type}\OperatorTok{=}\NormalTok{[}\StringTok{"num"}\NormalTok{, }\StringTok{"num"}\NormalTok{, }\StringTok{"factor"}\NormalTok{])}
\NormalTok{Sf.fit(X, y)}
\NormalTok{n }\OperatorTok{=} \DecValTok{50}
\NormalTok{X0 }\OperatorTok{=}\NormalTok{ gen.scipy\_lhd(n, lower}\OperatorTok{=}\NormalTok{lower, upper }\OperatorTok{=}\NormalTok{ upper)}
\NormalTok{X1 }\OperatorTok{=}\NormalTok{ np.random.randint(low}\OperatorTok{=}\DecValTok{1}\NormalTok{, high}\OperatorTok{=}\DecValTok{3}\NormalTok{, size}\OperatorTok{=}\NormalTok{(n,))}
\NormalTok{X }\OperatorTok{=}\NormalTok{ np.c\_[X0, X1]}
\NormalTok{y }\OperatorTok{=}\NormalTok{ fun(X)}
\NormalTok{s}\OperatorTok{=}\NormalTok{np.}\BuiltInTok{sum}\NormalTok{(np.}\BuiltInTok{abs}\NormalTok{(S.predict(X)[}\DecValTok{0}\NormalTok{] }\OperatorTok{{-}}\NormalTok{ y))}
\NormalTok{sf}\OperatorTok{=}\NormalTok{np.}\BuiltInTok{sum}\NormalTok{(np.}\BuiltInTok{abs}\NormalTok{(Sf.predict(X)[}\DecValTok{0}\NormalTok{] }\OperatorTok{{-}}\NormalTok{ y))}
\NormalTok{sf }\OperatorTok{{-}}\NormalTok{ s}
\end{Highlighting}
\end{Shaded}

\begin{verbatim}
-40.48225931963543
\end{verbatim}

\begin{Shaded}
\begin{Highlighting}[]
\CommentTok{\# vars(S)}
\end{Highlighting}
\end{Shaded}

\begin{Shaded}
\begin{Highlighting}[]
\CommentTok{\# vars(Sf)}
\end{Highlighting}
\end{Shaded}

\hypertarget{sec-noise}{%
\chapter{Hyperparameter Tuning and Noise}\label{sec-noise}}

This chapter demonstrates how noisy functions can be handled by
\texttt{Spot}.

\hypertarget{example-spot-and-the-noisy-sphere-function}{%
\section{\texorpdfstring{Example: \texttt{Spot} and the Noisy Sphere
Function}{Example: Spot and the Noisy Sphere Function}}\label{example-spot-and-the-noisy-sphere-function}}

\begin{Shaded}
\begin{Highlighting}[]
\ImportTok{import}\NormalTok{ numpy }\ImportTok{as}\NormalTok{ np}
\ImportTok{from}\NormalTok{ math }\ImportTok{import}\NormalTok{ inf}
\ImportTok{from}\NormalTok{ spotPython.fun.objectivefunctions }\ImportTok{import}\NormalTok{ analytical}
\ImportTok{from}\NormalTok{ spotPython.spot }\ImportTok{import}\NormalTok{ spot}
\ImportTok{import}\NormalTok{ matplotlib.pyplot }\ImportTok{as}\NormalTok{ plt}
\ImportTok{from}\NormalTok{ spotPython.utils.}\BuiltInTok{file} \ImportTok{import}\NormalTok{ get\_experiment\_name}
\ImportTok{from}\NormalTok{ spotPython.utils.init }\ImportTok{import}\NormalTok{ fun\_control\_init}
\ImportTok{from}\NormalTok{ spotPython.utils.}\BuiltInTok{file} \ImportTok{import}\NormalTok{ get\_spot\_tensorboard\_path}

\NormalTok{PREFIX }\OperatorTok{=} \StringTok{"08"}
\NormalTok{experiment\_name }\OperatorTok{=}\NormalTok{ get\_experiment\_name(prefix}\OperatorTok{=}\NormalTok{PREFIX)}
\BuiltInTok{print}\NormalTok{(experiment\_name)}
\end{Highlighting}
\end{Shaded}

\begin{verbatim}
08_bartz09_2023-07-17_08-51-27
\end{verbatim}

\hypertarget{the-objective-function-noisy-sphere}{%
\subsection{The Objective Function: Noisy
Sphere}\label{the-objective-function-noisy-sphere}}

\begin{itemize}
\item
  The \texttt{spotPython} package provides several classes of objective
  functions.
\item
  We will use an analytical objective function with noise, i.e., a
  function that can be described by a (closed) formula:
  \[f(x) = x^2 + \epsilon\]
\item
  Since \texttt{sigma} is set to \texttt{0.1}, noise is added to the
  function:
\end{itemize}

\begin{Shaded}
\begin{Highlighting}[]
\NormalTok{fun }\OperatorTok{=}\NormalTok{ analytical().fun\_sphere}
\NormalTok{fun\_control }\OperatorTok{=}\NormalTok{ fun\_control\_init(}
\NormalTok{    spot\_tensorboard\_path}\OperatorTok{=}\NormalTok{get\_spot\_tensorboard\_path(experiment\_name),}
\NormalTok{    sigma}\OperatorTok{=}\FloatTok{0.02}\NormalTok{,}
\NormalTok{    seed}\OperatorTok{=}\DecValTok{123}\NormalTok{,)}
\end{Highlighting}
\end{Shaded}

\begin{itemize}
\tightlist
\item
  A plot illustrates the noise:
\end{itemize}

\begin{Shaded}
\begin{Highlighting}[]
\NormalTok{x }\OperatorTok{=}\NormalTok{ np.linspace(}\OperatorTok{{-}}\DecValTok{1}\NormalTok{,}\DecValTok{1}\NormalTok{,}\DecValTok{100}\NormalTok{).reshape(}\OperatorTok{{-}}\DecValTok{1}\NormalTok{,}\DecValTok{1}\NormalTok{)}
\NormalTok{y }\OperatorTok{=}\NormalTok{ fun(x, fun\_control}\OperatorTok{=}\NormalTok{fun\_control)}
\NormalTok{plt.figure()}
\NormalTok{plt.plot(x,y, }\StringTok{"k"}\NormalTok{)}
\NormalTok{plt.show()}
\end{Highlighting}
\end{Shaded}

\begin{figure}[H]

{\centering \includegraphics{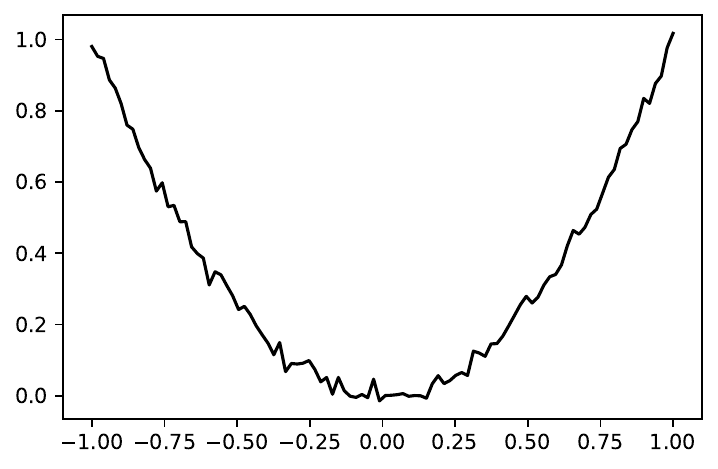}

}

\end{figure}

\texttt{Spot} is adopted as follows to cope with noisy functions:

\begin{enumerate}
\def\labelenumi{\arabic{enumi}.}
\tightlist
\item
  \texttt{fun\_repeats} is set to a value larger than 1 (here: 2)
\item
  \texttt{noise} is set to \texttt{true}. Therefore, a nugget
  (\texttt{Lambda}) term is added to the correlation matrix
\item
  \texttt{init\ size} (of the \texttt{design\_control} dictionary) is
  set to a value larger than 1 (here: 2)
\end{enumerate}

\begin{Shaded}
\begin{Highlighting}[]
\NormalTok{spot\_1\_noisy }\OperatorTok{=}\NormalTok{ spot.Spot(fun}\OperatorTok{=}\NormalTok{fun,}
\NormalTok{                   lower }\OperatorTok{=}\NormalTok{ np.array([}\OperatorTok{{-}}\DecValTok{1}\NormalTok{]),}
\NormalTok{                   upper }\OperatorTok{=}\NormalTok{ np.array([}\DecValTok{1}\NormalTok{]),}
\NormalTok{                   fun\_evals }\OperatorTok{=} \DecValTok{20}\NormalTok{,}
\NormalTok{                   fun\_repeats }\OperatorTok{=} \DecValTok{2}\NormalTok{,}
\NormalTok{                   noise }\OperatorTok{=} \VariableTok{True}\NormalTok{,}
\NormalTok{                   seed}\OperatorTok{=}\DecValTok{123}\NormalTok{,}
\NormalTok{                   show\_models}\OperatorTok{=}\VariableTok{True}\NormalTok{,}
\NormalTok{                   design\_control}\OperatorTok{=}\NormalTok{\{}\StringTok{"init\_size"}\NormalTok{: }\DecValTok{3}\NormalTok{,}
                                   \StringTok{"repeats"}\NormalTok{: }\DecValTok{2}\NormalTok{\},}
\NormalTok{                   surrogate\_control}\OperatorTok{=}\NormalTok{\{}\StringTok{"noise"}\NormalTok{: }\VariableTok{True}\NormalTok{\},}
\NormalTok{                   fun\_control}\OperatorTok{=}\NormalTok{fun\_control,)}
\end{Highlighting}
\end{Shaded}

\begin{Shaded}
\begin{Highlighting}[]
\NormalTok{spot\_1\_noisy.run()}
\end{Highlighting}
\end{Shaded}

\begin{figure}[H]

{\centering \includegraphics{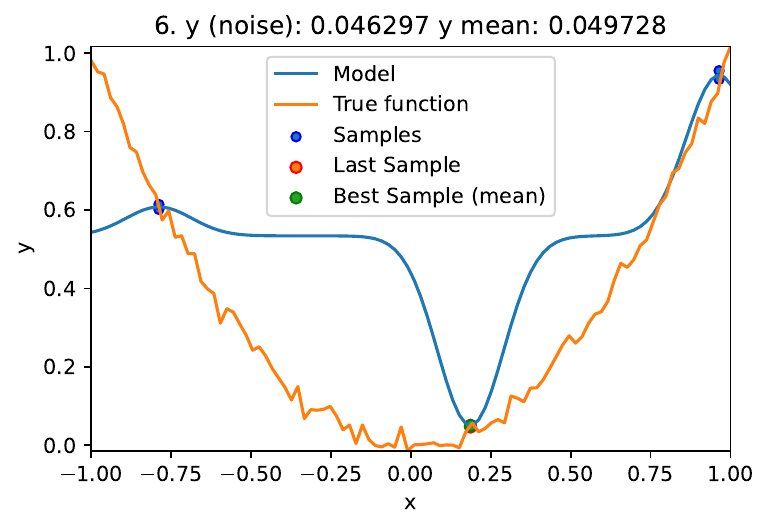}

}

\end{figure}

\begin{figure}[H]

{\centering \includegraphics{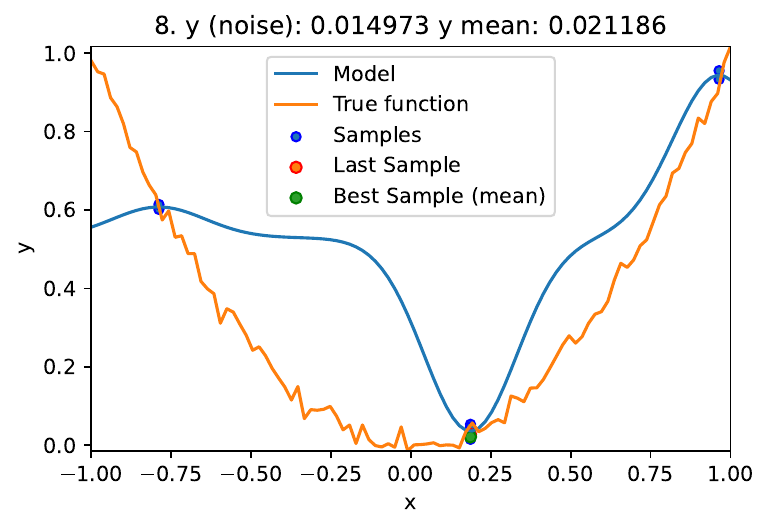}

}

\end{figure}

\begin{verbatim}
spotPython tuning: 0.01497250376669991 [####------] 40.00% 
\end{verbatim}

\begin{figure}[H]

{\centering \includegraphics{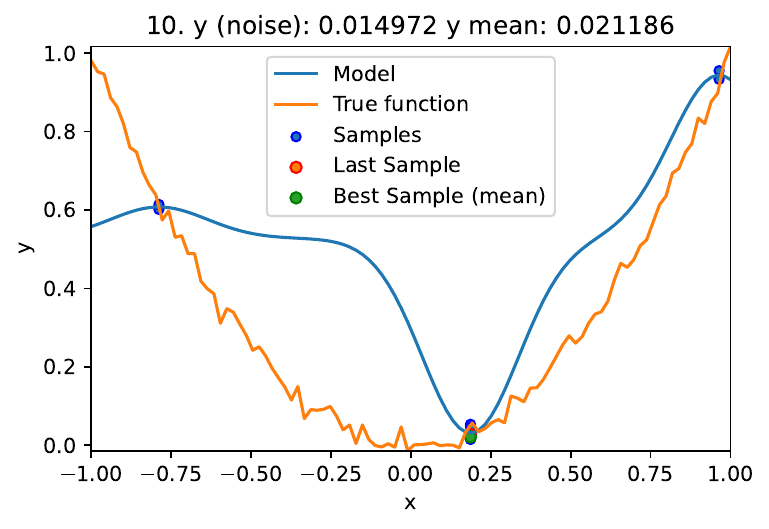}

}

\end{figure}

\begin{verbatim}
spotPython tuning: 0.01497226931667417 [#####-----] 50.00% 
\end{verbatim}

\begin{figure}[H]

{\centering \includegraphics{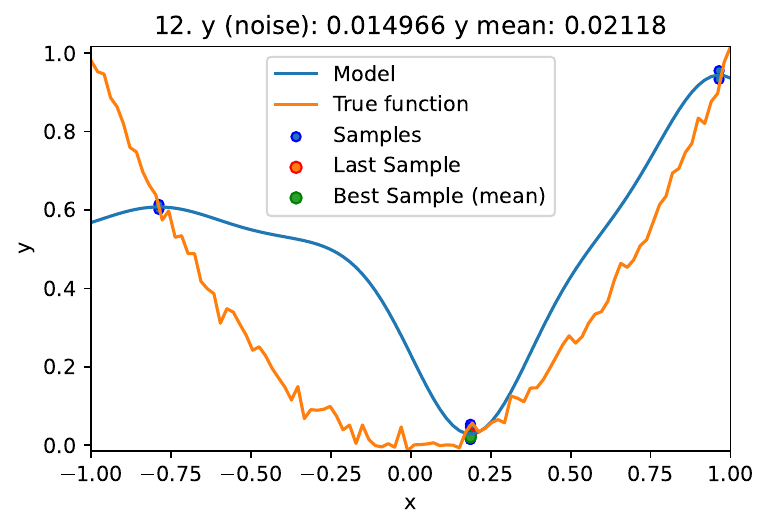}

}

\end{figure}

\begin{verbatim}
spotPython tuning: 0.01496618769080537 [######----] 60.00% 
\end{verbatim}

\begin{figure}[H]

{\centering \includegraphics{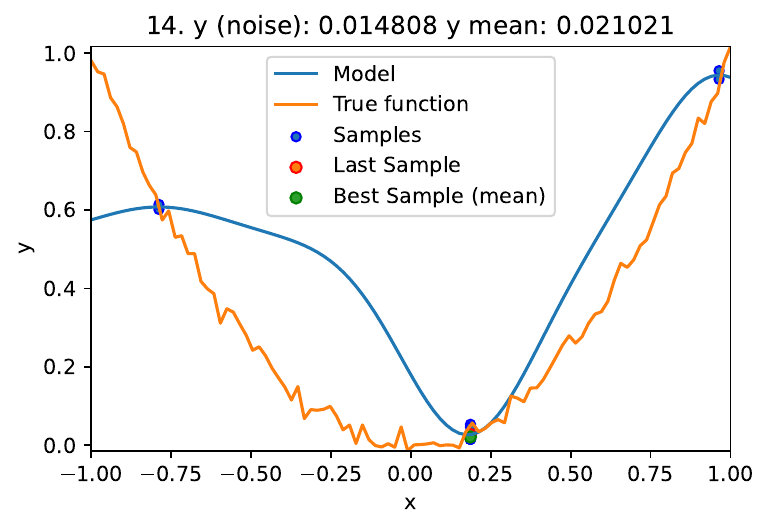}

}

\end{figure}

\begin{verbatim}
spotPython tuning: 0.014808104491512888 [#######---] 70.00% 
\end{verbatim}

\begin{figure}[H]

{\centering \includegraphics{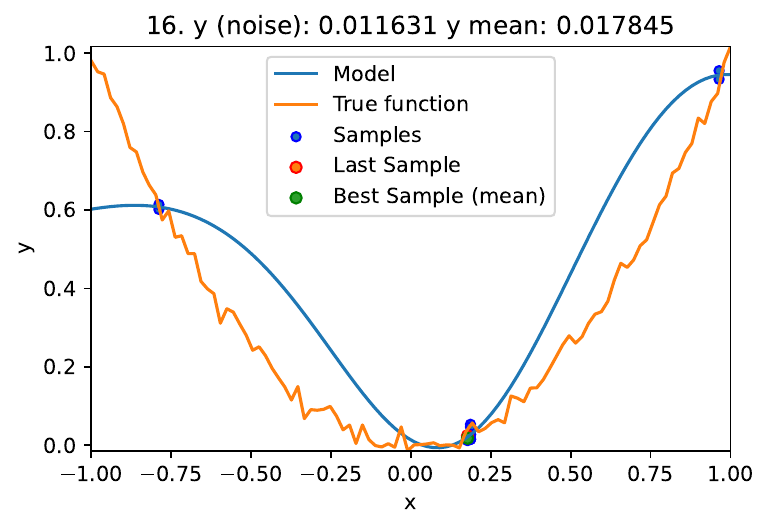}

}

\end{figure}

\begin{verbatim}
spotPython tuning: 0.011631261600357518 [########--] 80.00% 
\end{verbatim}

\begin{figure}[H]

{\centering \includegraphics{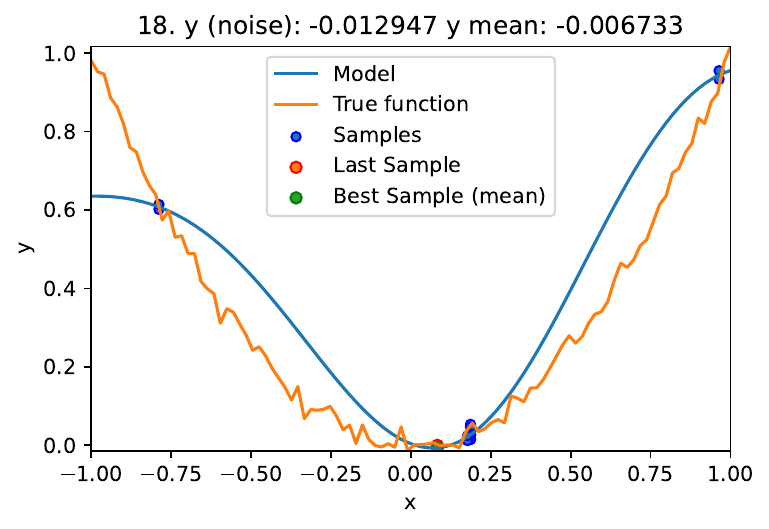}

}

\end{figure}

\begin{verbatim}
spotPython tuning: -0.012946672238374722 [#########-] 90.00% 
\end{verbatim}

\begin{figure}[H]

{\centering \includegraphics{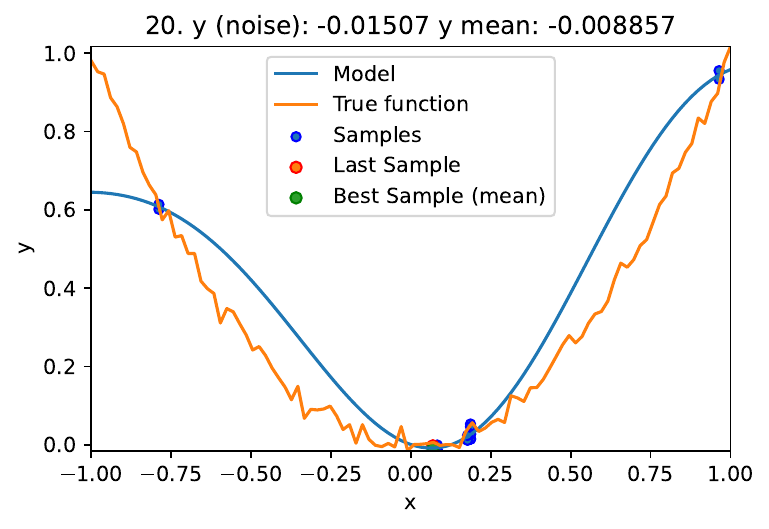}

}

\end{figure}

\begin{verbatim}
spotPython tuning: -0.015070457665271902 [##########] 100.00% Done...
\end{verbatim}

\begin{verbatim}
<spotPython.spot.spot.Spot at 0x16d2e6fe0>
\end{verbatim}

\hypertarget{print-the-results-3}{%
\section{Print the Results}\label{print-the-results-3}}

\begin{Shaded}
\begin{Highlighting}[]
\NormalTok{spot\_1\_noisy.print\_results()}
\end{Highlighting}
\end{Shaded}

\begin{verbatim}
min y: -0.015070457665271902
x0: 0.06864378589271657
min mean y: -0.008857110676472227
x0: 0.06864378589271657
\end{verbatim}

\begin{verbatim}
[['x0', 0.06864378589271657], ['x0', 0.06864378589271657]]
\end{verbatim}

\begin{Shaded}
\begin{Highlighting}[]
\NormalTok{spot\_1\_noisy.plot\_progress(log\_y}\OperatorTok{=}\VariableTok{False}\NormalTok{,}
\NormalTok{    filename}\OperatorTok{=}\StringTok{"./figures/"} \OperatorTok{+}\NormalTok{ experiment\_name}\OperatorTok{+}\StringTok{"\_progress.png"}\NormalTok{)}
\end{Highlighting}
\end{Shaded}

\begin{figure}[H]

{\centering \includegraphics{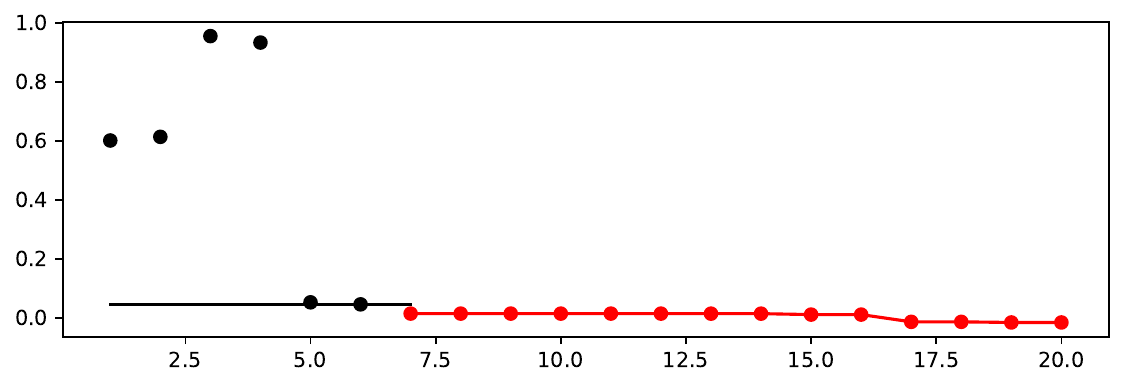}

}

\caption{Progress plot. \emph{Black} dots denote results from the
initial design. \emph{Red} dots illustrate the improvement found by the
surrogate model based optimization.}

\end{figure}

\hypertarget{noise-and-surrogates-the-nugget-effect}{%
\section{Noise and Surrogates: The Nugget
Effect}\label{noise-and-surrogates-the-nugget-effect}}

\hypertarget{the-noisy-sphere}{%
\subsection{The Noisy Sphere}\label{the-noisy-sphere}}

\hypertarget{the-data}{%
\subsubsection{The Data}\label{the-data}}

\begin{itemize}
\tightlist
\item
  We prepare some data first:
\end{itemize}

\begin{Shaded}
\begin{Highlighting}[]
\ImportTok{import}\NormalTok{ numpy }\ImportTok{as}\NormalTok{ np}
\ImportTok{import}\NormalTok{ spotPython}
\ImportTok{from}\NormalTok{ spotPython.fun.objectivefunctions }\ImportTok{import}\NormalTok{ analytical}
\ImportTok{from}\NormalTok{ spotPython.spot }\ImportTok{import}\NormalTok{ spot}
\ImportTok{from}\NormalTok{ spotPython.design.spacefilling }\ImportTok{import}\NormalTok{ spacefilling}
\ImportTok{from}\NormalTok{ spotPython.build.kriging }\ImportTok{import}\NormalTok{ Kriging}
\ImportTok{import}\NormalTok{ matplotlib.pyplot }\ImportTok{as}\NormalTok{ plt}

\NormalTok{gen }\OperatorTok{=}\NormalTok{ spacefilling(}\DecValTok{1}\NormalTok{)}
\NormalTok{rng }\OperatorTok{=}\NormalTok{ np.random.RandomState(}\DecValTok{1}\NormalTok{)}
\NormalTok{lower }\OperatorTok{=}\NormalTok{ np.array([}\OperatorTok{{-}}\DecValTok{10}\NormalTok{])}
\NormalTok{upper }\OperatorTok{=}\NormalTok{ np.array([}\DecValTok{10}\NormalTok{])}
\NormalTok{fun }\OperatorTok{=}\NormalTok{ analytical().fun\_sphere}
\NormalTok{fun\_control }\OperatorTok{=}\NormalTok{ fun\_control\_init(}
\NormalTok{    spot\_tensorboard\_path}\OperatorTok{=}\NormalTok{get\_spot\_tensorboard\_path(experiment\_name),}
\NormalTok{    sigma}\OperatorTok{=}\DecValTok{2}\NormalTok{,}
\NormalTok{    seed}\OperatorTok{=}\DecValTok{123}\NormalTok{,)}
\NormalTok{X }\OperatorTok{=}\NormalTok{ gen.scipy\_lhd(}\DecValTok{10}\NormalTok{, lower}\OperatorTok{=}\NormalTok{lower, upper }\OperatorTok{=}\NormalTok{ upper)}
\NormalTok{y }\OperatorTok{=}\NormalTok{ fun(X, fun\_control}\OperatorTok{=}\NormalTok{fun\_control)}
\NormalTok{X\_train }\OperatorTok{=}\NormalTok{ X.reshape(}\OperatorTok{{-}}\DecValTok{1}\NormalTok{,}\DecValTok{1}\NormalTok{)}
\NormalTok{y\_train }\OperatorTok{=}\NormalTok{ y}
\end{Highlighting}
\end{Shaded}

\begin{itemize}
\tightlist
\item
  A surrogate without nugget is fitted to these data:
\end{itemize}

\begin{Shaded}
\begin{Highlighting}[]
\NormalTok{S }\OperatorTok{=}\NormalTok{ Kriging(name}\OperatorTok{=}\StringTok{\textquotesingle{}kriging\textquotesingle{}}\NormalTok{,}
\NormalTok{            seed}\OperatorTok{=}\DecValTok{123}\NormalTok{,}
\NormalTok{            log\_level}\OperatorTok{=}\DecValTok{50}\NormalTok{,}
\NormalTok{            n\_theta}\OperatorTok{=}\DecValTok{1}\NormalTok{,}
\NormalTok{            noise}\OperatorTok{=}\VariableTok{False}\NormalTok{)}
\NormalTok{S.fit(X\_train, y\_train)}

\NormalTok{X\_axis }\OperatorTok{=}\NormalTok{ np.linspace(start}\OperatorTok{={-}}\DecValTok{13}\NormalTok{, stop}\OperatorTok{=}\DecValTok{13}\NormalTok{, num}\OperatorTok{=}\DecValTok{1000}\NormalTok{).reshape(}\OperatorTok{{-}}\DecValTok{1}\NormalTok{, }\DecValTok{1}\NormalTok{)}
\NormalTok{mean\_prediction, std\_prediction, ei }\OperatorTok{=}\NormalTok{ S.predict(X\_axis, return\_val}\OperatorTok{=}\StringTok{"all"}\NormalTok{)}

\NormalTok{plt.scatter(X\_train, y\_train, label}\OperatorTok{=}\StringTok{"Observations"}\NormalTok{)}
\NormalTok{plt.plot(X\_axis, mean\_prediction, label}\OperatorTok{=}\StringTok{"mue"}\NormalTok{)}
\NormalTok{plt.legend()}
\NormalTok{plt.xlabel(}\StringTok{"$x$"}\NormalTok{)}
\NormalTok{plt.ylabel(}\StringTok{"$f(x)$"}\NormalTok{)}
\NormalTok{\_ }\OperatorTok{=}\NormalTok{ plt.title(}\StringTok{"Sphere: Gaussian process regression on noisy dataset"}\NormalTok{)}
\end{Highlighting}
\end{Shaded}

\begin{figure}[H]

{\centering \includegraphics{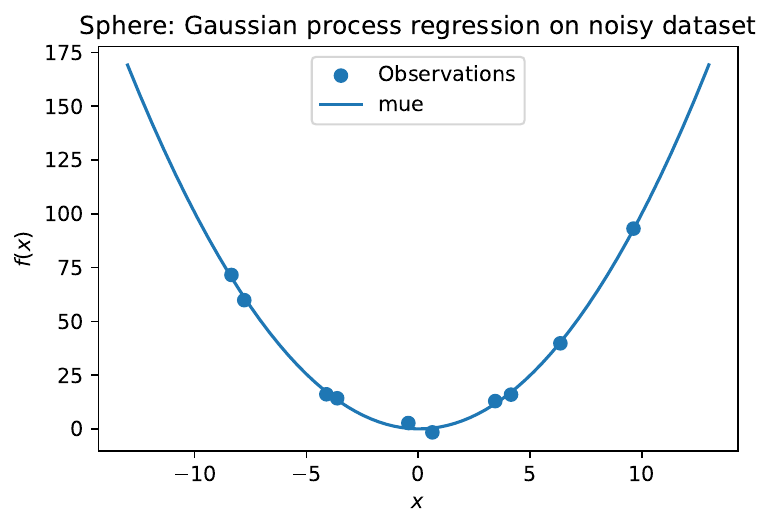}

}

\end{figure}

\begin{itemize}
\tightlist
\item
  In comparison to the surrogate without nugget, we fit a surrogate with
  nugget to the data:
\end{itemize}

\begin{Shaded}
\begin{Highlighting}[]
\NormalTok{S\_nug }\OperatorTok{=}\NormalTok{ Kriging(name}\OperatorTok{=}\StringTok{\textquotesingle{}kriging\textquotesingle{}}\NormalTok{,}
\NormalTok{            seed}\OperatorTok{=}\DecValTok{123}\NormalTok{,}
\NormalTok{            log\_level}\OperatorTok{=}\DecValTok{50}\NormalTok{,}
\NormalTok{            n\_theta}\OperatorTok{=}\DecValTok{1}\NormalTok{,}
\NormalTok{            noise}\OperatorTok{=}\VariableTok{True}\NormalTok{)}
\NormalTok{S\_nug.fit(X\_train, y\_train)}
\NormalTok{X\_axis }\OperatorTok{=}\NormalTok{ np.linspace(start}\OperatorTok{={-}}\DecValTok{13}\NormalTok{, stop}\OperatorTok{=}\DecValTok{13}\NormalTok{, num}\OperatorTok{=}\DecValTok{1000}\NormalTok{).reshape(}\OperatorTok{{-}}\DecValTok{1}\NormalTok{, }\DecValTok{1}\NormalTok{)}
\NormalTok{mean\_prediction, std\_prediction, ei }\OperatorTok{=}\NormalTok{ S\_nug.predict(X\_axis, return\_val}\OperatorTok{=}\StringTok{"all"}\NormalTok{)}
\NormalTok{plt.scatter(X\_train, y\_train, label}\OperatorTok{=}\StringTok{"Observations"}\NormalTok{)}
\NormalTok{plt.plot(X\_axis, mean\_prediction, label}\OperatorTok{=}\StringTok{"mue"}\NormalTok{)}
\NormalTok{plt.legend()}
\NormalTok{plt.xlabel(}\StringTok{"$x$"}\NormalTok{)}
\NormalTok{plt.ylabel(}\StringTok{"$f(x)$"}\NormalTok{)}
\NormalTok{\_ }\OperatorTok{=}\NormalTok{ plt.title(}\StringTok{"Sphere: Gaussian process regression with nugget on noisy dataset"}\NormalTok{)}
\end{Highlighting}
\end{Shaded}

\begin{figure}[H]

{\centering \includegraphics{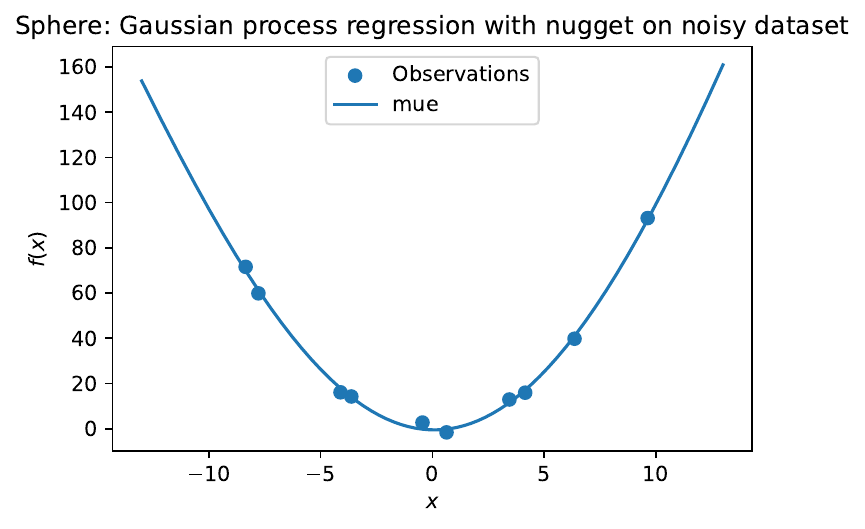}

}

\end{figure}

\begin{itemize}
\tightlist
\item
  The value of the nugget term can be extracted from the model as
  follows:
\end{itemize}

\begin{Shaded}
\begin{Highlighting}[]
\NormalTok{S.Lambda}
\end{Highlighting}
\end{Shaded}

\begin{Shaded}
\begin{Highlighting}[]
\NormalTok{S\_nug.Lambda}
\end{Highlighting}
\end{Shaded}

\begin{verbatim}
5.2085090734655785e-05
\end{verbatim}

\begin{itemize}
\tightlist
\item
  We see:

  \begin{itemize}
  \tightlist
  \item
    the first model \texttt{S} has no nugget,
  \item
    whereas the second model has a nugget value (\texttt{Lambda}) larger
    than zero.
  \end{itemize}
\end{itemize}

\hypertarget{exercises-5}{%
\section{Exercises}\label{exercises-5}}

\hypertarget{noisy-fun_cubed}{%
\subsection{\texorpdfstring{Noisy
\texttt{fun\_cubed}}{Noisy fun\_cubed}}\label{noisy-fun_cubed}}

\begin{itemize}
\tightlist
\item
  Analyse the effect of noise on the \texttt{fun\_cubed} function with
  the following settings:
\end{itemize}

\begin{Shaded}
\begin{Highlighting}[]
\NormalTok{fun }\OperatorTok{=}\NormalTok{ analytical().fun\_cubed}
\NormalTok{fun\_control }\OperatorTok{=}\NormalTok{ fun\_control\_init(}
\NormalTok{    sigma}\OperatorTok{=}\DecValTok{10}\NormalTok{,}
\NormalTok{    seed}\OperatorTok{=}\DecValTok{123}\NormalTok{,)}
\NormalTok{lower }\OperatorTok{=}\NormalTok{ np.array([}\OperatorTok{{-}}\DecValTok{10}\NormalTok{])}
\NormalTok{upper }\OperatorTok{=}\NormalTok{ np.array([}\DecValTok{10}\NormalTok{])}
\end{Highlighting}
\end{Shaded}

\hypertarget{fun_runge-1}{%
\subsection{\texorpdfstring{\texttt{fun\_runge}}{fun\_runge}}\label{fun_runge-1}}

\begin{itemize}
\tightlist
\item
  Analyse the effect of noise on the \texttt{fun\_runge} function with
  the following settings:
\end{itemize}

\begin{Shaded}
\begin{Highlighting}[]
\NormalTok{lower }\OperatorTok{=}\NormalTok{ np.array([}\OperatorTok{{-}}\DecValTok{10}\NormalTok{])}
\NormalTok{upper }\OperatorTok{=}\NormalTok{ np.array([}\DecValTok{10}\NormalTok{])}
\NormalTok{fun }\OperatorTok{=}\NormalTok{ analytical().fun\_runge}
\NormalTok{fun\_control }\OperatorTok{=}\NormalTok{ fun\_control\_init(}
\NormalTok{    sigma}\OperatorTok{=}\FloatTok{0.25}\NormalTok{,}
\NormalTok{    seed}\OperatorTok{=}\DecValTok{123}\NormalTok{,)}
\end{Highlighting}
\end{Shaded}

\hypertarget{fun_forrester}{%
\subsection{\texorpdfstring{\texttt{fun\_forrester}}{fun\_forrester}}\label{fun_forrester}}

\begin{itemize}
\tightlist
\item
  Analyse the effect of noise on the \texttt{fun\_forrester} function
  with the following settings:
\end{itemize}

\begin{Shaded}
\begin{Highlighting}[]
\NormalTok{lower }\OperatorTok{=}\NormalTok{ np.array([}\DecValTok{0}\NormalTok{])}
\NormalTok{upper }\OperatorTok{=}\NormalTok{ np.array([}\DecValTok{1}\NormalTok{])}
\NormalTok{fun }\OperatorTok{=}\NormalTok{ analytical().fun\_forrester}
\NormalTok{fun\_control }\OperatorTok{=}\NormalTok{ fun\_control\_init(}
\NormalTok{    sigma}\OperatorTok{=}\DecValTok{5}\NormalTok{,}
\NormalTok{    seed}\OperatorTok{=}\DecValTok{123}\NormalTok{,)}
\end{Highlighting}
\end{Shaded}

\hypertarget{fun_xsin}{%
\subsection{\texorpdfstring{\texttt{fun\_xsin}}{fun\_xsin}}\label{fun_xsin}}

\begin{itemize}
\tightlist
\item
  Analyse the effect of noise on the \texttt{fun\_xsin} function with
  the following settings:
\end{itemize}

\begin{Shaded}
\begin{Highlighting}[]
\NormalTok{lower }\OperatorTok{=}\NormalTok{ np.array([}\OperatorTok{{-}}\FloatTok{1.}\NormalTok{])}
\NormalTok{upper }\OperatorTok{=}\NormalTok{ np.array([}\FloatTok{1.}\NormalTok{])}
\NormalTok{fun }\OperatorTok{=}\NormalTok{ analytical().fun\_xsin}
\NormalTok{fun\_control }\OperatorTok{=}\NormalTok{ fun\_control\_init(    }
\NormalTok{    sigma}\OperatorTok{=}\FloatTok{0.5}\NormalTok{,}
\NormalTok{    seed}\OperatorTok{=}\DecValTok{123}\NormalTok{,)}
\end{Highlighting}
\end{Shaded}

\hypertarget{sec-ocba}{%
\chapter{\texorpdfstring{Handling Noise: Optimal Computational Budget
Allocation in
\texttt{Spot}}{Handling Noise: Optimal Computational Budget Allocation in Spot}}\label{sec-ocba}}

This chapter demonstrates how noisy functions can be handled with
Optimal Computational Budget Allocation (OCBA) by \texttt{Spot}.

\hypertarget{example-spot-ocba-and-the-noisy-sphere-function}{%
\section{\texorpdfstring{Example: \texttt{Spot}, OCBA, and the Noisy
Sphere
Function}{Example: Spot, OCBA, and the Noisy Sphere Function}}\label{example-spot-ocba-and-the-noisy-sphere-function}}

\begin{Shaded}
\begin{Highlighting}[]
\ImportTok{import}\NormalTok{ numpy }\ImportTok{as}\NormalTok{ np}
\ImportTok{from}\NormalTok{ math }\ImportTok{import}\NormalTok{ inf}
\ImportTok{from}\NormalTok{ spotPython.fun.objectivefunctions }\ImportTok{import}\NormalTok{ analytical}
\ImportTok{from}\NormalTok{ spotPython.spot }\ImportTok{import}\NormalTok{ spot}
\ImportTok{import}\NormalTok{ matplotlib.pyplot }\ImportTok{as}\NormalTok{ plt}
\ImportTok{from}\NormalTok{ spotPython.utils.}\BuiltInTok{file} \ImportTok{import}\NormalTok{ get\_experiment\_name}
\ImportTok{from}\NormalTok{ spotPython.utils.init }\ImportTok{import}\NormalTok{ fun\_control\_init}
\ImportTok{from}\NormalTok{ spotPython.utils.}\BuiltInTok{file} \ImportTok{import}\NormalTok{ get\_spot\_tensorboard\_path}

\NormalTok{PREFIX }\OperatorTok{=} \StringTok{"09"}
\NormalTok{experiment\_name }\OperatorTok{=}\NormalTok{ get\_experiment\_name(prefix}\OperatorTok{=}\NormalTok{PREFIX)}
\BuiltInTok{print}\NormalTok{(experiment\_name)}
\end{Highlighting}
\end{Shaded}

\begin{verbatim}
09_bartz09_2023-07-17_08-51-43
\end{verbatim}

\hypertarget{the-objective-function-noisy-sphere-1}{%
\subsection{The Objective Function: Noisy
Sphere}\label{the-objective-function-noisy-sphere-1}}

The \texttt{spotPython} package provides several classes of objective
functions. We will use an analytical objective function with noise,
i.e., a function that can be described by a (closed) formula:
\[f(x) = x^2 + \epsilon\]

Since \texttt{sigma} is set to \texttt{0.1}, noise is added to the
function:

\begin{Shaded}
\begin{Highlighting}[]
\NormalTok{fun }\OperatorTok{=}\NormalTok{ analytical().fun\_sphere}
\NormalTok{fun\_control }\OperatorTok{=}\NormalTok{ fun\_control\_init(}
\NormalTok{    spot\_tensorboard\_path}\OperatorTok{=}\NormalTok{get\_spot\_tensorboard\_path(experiment\_name),}
\NormalTok{    sigma}\OperatorTok{=}\FloatTok{0.1}\NormalTok{,}
\NormalTok{    seed}\OperatorTok{=}\DecValTok{123}\NormalTok{,)}
\end{Highlighting}
\end{Shaded}

A plot illustrates the noise:

\begin{Shaded}
\begin{Highlighting}[]
\NormalTok{x }\OperatorTok{=}\NormalTok{ np.linspace(}\OperatorTok{{-}}\DecValTok{1}\NormalTok{,}\DecValTok{1}\NormalTok{,}\DecValTok{100}\NormalTok{).reshape(}\OperatorTok{{-}}\DecValTok{1}\NormalTok{,}\DecValTok{1}\NormalTok{)}
\NormalTok{y }\OperatorTok{=}\NormalTok{ fun(x, fun\_control}\OperatorTok{=}\NormalTok{fun\_control)}
\NormalTok{plt.figure()}
\NormalTok{plt.plot(x,y, }\StringTok{"k"}\NormalTok{)}
\NormalTok{plt.show()}
\end{Highlighting}
\end{Shaded}

\begin{figure}[H]

{\centering \includegraphics{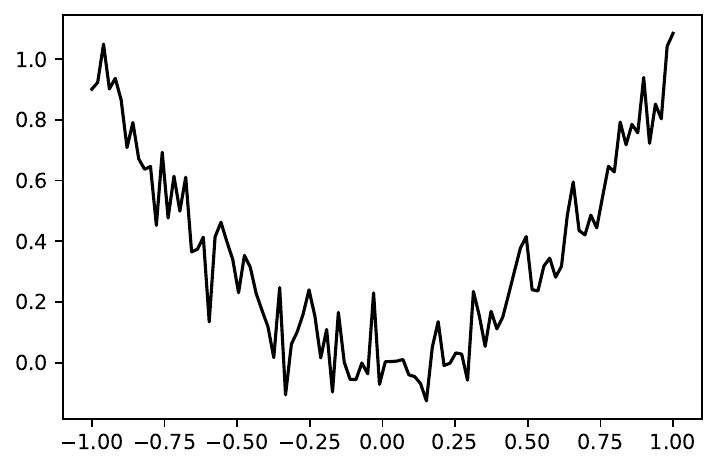}

}

\end{figure}

\texttt{Spot} is adopted as follows to cope with noisy functions:

\begin{enumerate}
\def\labelenumi{\arabic{enumi}.}
\tightlist
\item
  \texttt{fun\_repeats} is set to a value larger than 1 (here: 2)
\item
  \texttt{noise} is set to \texttt{true}. Therefore, a nugget
  (\texttt{Lambda}) term is added to the correlation matrix
\item
  \texttt{init\ size} (of the \texttt{design\_control} dictionary) is
  set to a value larger than 1 (here: 2)
\end{enumerate}

\begin{Shaded}
\begin{Highlighting}[]
\NormalTok{spot\_1\_noisy }\OperatorTok{=}\NormalTok{ spot.Spot(fun}\OperatorTok{=}\NormalTok{fun,}
\NormalTok{                   lower }\OperatorTok{=}\NormalTok{ np.array([}\OperatorTok{{-}}\DecValTok{1}\NormalTok{]),}
\NormalTok{                   upper }\OperatorTok{=}\NormalTok{ np.array([}\DecValTok{1}\NormalTok{]),}
\NormalTok{                   fun\_evals }\OperatorTok{=} \DecValTok{20}\NormalTok{,}
\NormalTok{                   fun\_repeats }\OperatorTok{=} \DecValTok{2}\NormalTok{,}
\NormalTok{                   infill\_criterion}\OperatorTok{=}\StringTok{"ei"}\NormalTok{,}
\NormalTok{                   noise }\OperatorTok{=} \VariableTok{True}\NormalTok{,}
\NormalTok{                   tolerance\_x}\OperatorTok{=}\FloatTok{0.0}\NormalTok{,}
\NormalTok{                   ocba\_delta }\OperatorTok{=} \DecValTok{1}\NormalTok{,}
\NormalTok{                   seed}\OperatorTok{=}\DecValTok{123}\NormalTok{,}
\NormalTok{                   show\_models}\OperatorTok{=}\VariableTok{True}\NormalTok{,}
\NormalTok{                   fun\_control }\OperatorTok{=}\NormalTok{ fun\_control,}
\NormalTok{                   design\_control}\OperatorTok{=}\NormalTok{\{}\StringTok{"init\_size"}\NormalTok{: }\DecValTok{3}\NormalTok{,}
                                   \StringTok{"repeats"}\NormalTok{: }\DecValTok{2}\NormalTok{\},}
\NormalTok{                   surrogate\_control}\OperatorTok{=}\NormalTok{\{}\StringTok{"noise"}\NormalTok{: }\VariableTok{True}\NormalTok{\})}
\end{Highlighting}
\end{Shaded}

\begin{Shaded}
\begin{Highlighting}[]
\NormalTok{spot\_1\_noisy.run()}
\end{Highlighting}
\end{Shaded}

\begin{figure}[H]

{\centering \includegraphics{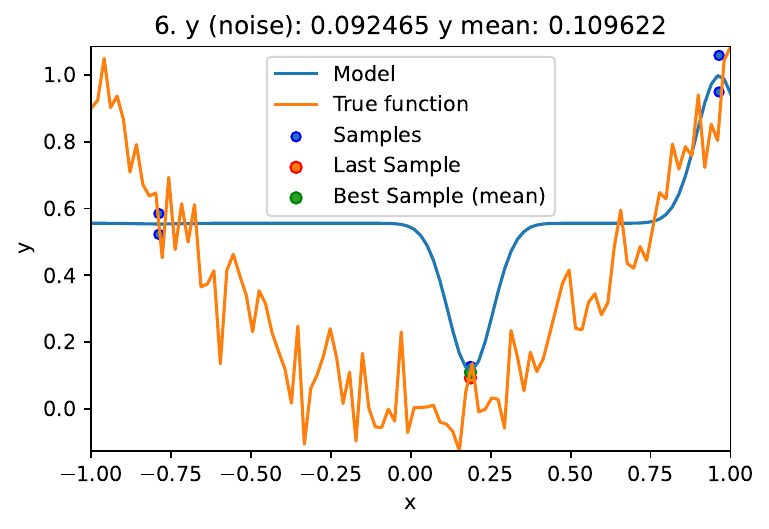}

}

\end{figure}

\begin{figure}[H]

{\centering \includegraphics{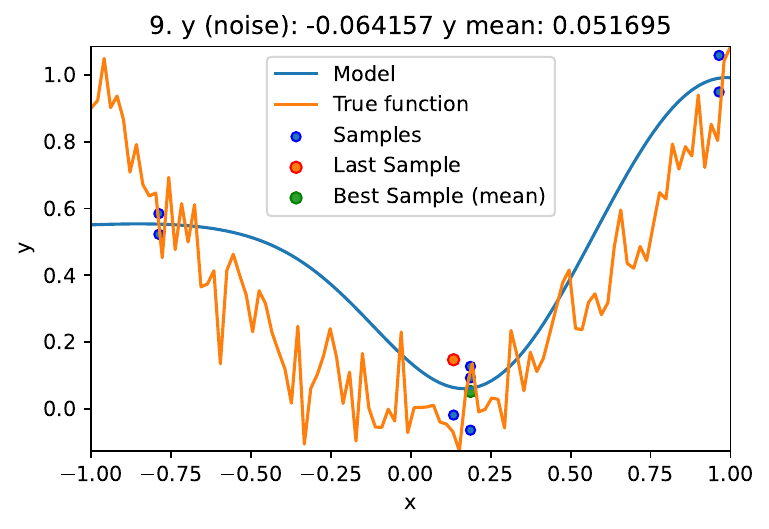}

}

\end{figure}

\begin{verbatim}
spotPython tuning: -0.0641572013655628 [####------] 45.00% 
\end{verbatim}

\begin{figure}[H]

{\centering \includegraphics{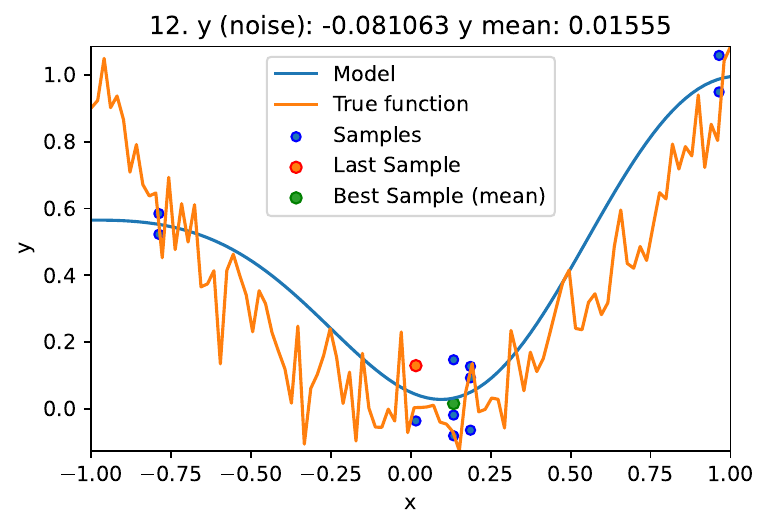}

}

\end{figure}

\begin{verbatim}
spotPython tuning: -0.08106318979661208 [######----] 60.00% 
\end{verbatim}

\begin{figure}[H]

{\centering \includegraphics{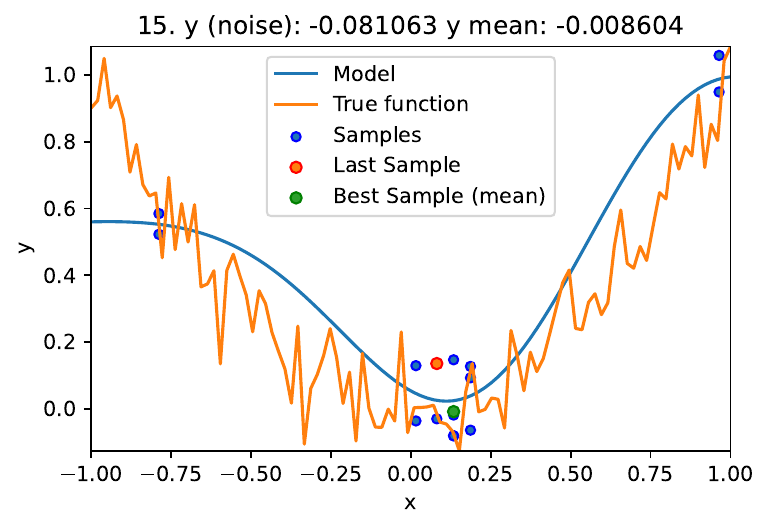}

}

\end{figure}

\begin{verbatim}
spotPython tuning: -0.08106318979661208 [########--] 75.00% 
\end{verbatim}

\begin{figure}[H]

{\centering \includegraphics{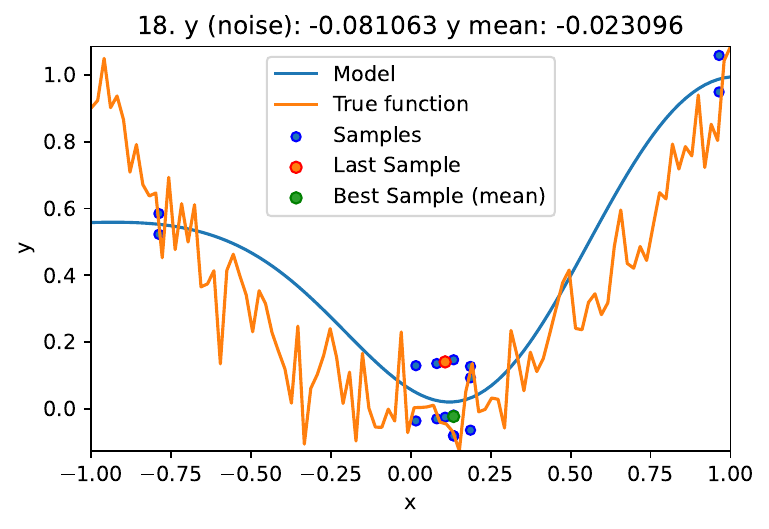}

}

\end{figure}

\begin{verbatim}
spotPython tuning: -0.08106318979661208 [#########-] 90.00% 
\end{verbatim}

\begin{figure}[H]

{\centering \includegraphics{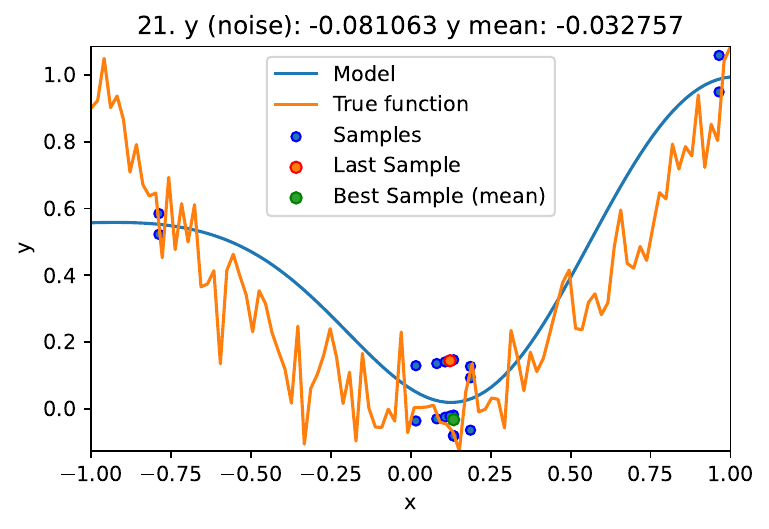}

}

\end{figure}

\begin{verbatim}
spotPython tuning: -0.08106318979661208 [##########] 100.00% Done...
\end{verbatim}

\begin{verbatim}
<spotPython.spot.spot.Spot at 0x17d973f70>
\end{verbatim}

\hypertarget{print-the-results-4}{%
\section{Print the Results}\label{print-the-results-4}}

\begin{Shaded}
\begin{Highlighting}[]
\NormalTok{spot\_1\_noisy.print\_results()}
\end{Highlighting}
\end{Shaded}

\begin{verbatim}
min y: -0.08106318979661208
x0: 0.1335999447536301
min mean y: -0.03275683462132762
x0: 0.1335999447536301
\end{verbatim}

\begin{verbatim}
[['x0', 0.1335999447536301], ['x0', 0.1335999447536301]]
\end{verbatim}

\begin{Shaded}
\begin{Highlighting}[]
\NormalTok{spot\_1\_noisy.plot\_progress(log\_y}\OperatorTok{=}\VariableTok{False}\NormalTok{)}
\end{Highlighting}
\end{Shaded}

\begin{figure}[H]

{\centering \includegraphics{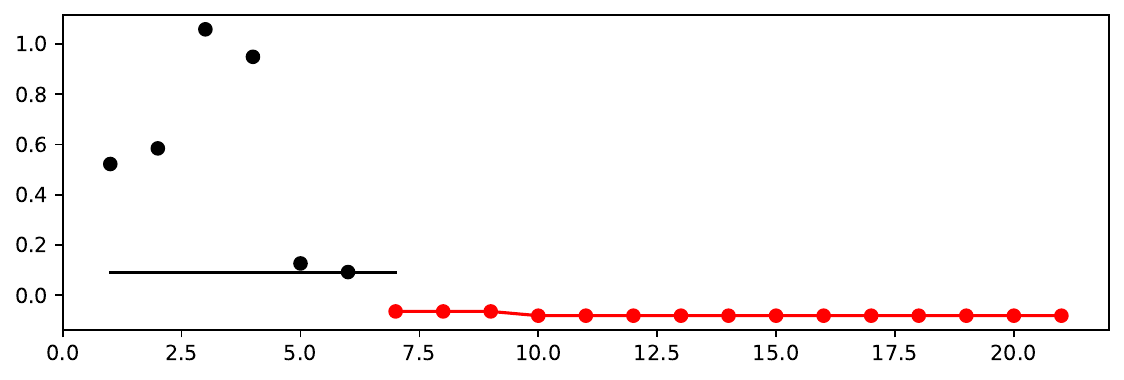}

}

\end{figure}

\hypertarget{noise-and-surrogates-the-nugget-effect-1}{%
\section{Noise and Surrogates: The Nugget
Effect}\label{noise-and-surrogates-the-nugget-effect-1}}

\hypertarget{the-noisy-sphere-1}{%
\subsection{The Noisy Sphere}\label{the-noisy-sphere-1}}

\hypertarget{the-data-1}{%
\subsubsection{The Data}\label{the-data-1}}

We prepare some data first:

\begin{Shaded}
\begin{Highlighting}[]
\ImportTok{import}\NormalTok{ numpy }\ImportTok{as}\NormalTok{ np}
\ImportTok{import}\NormalTok{ spotPython}
\ImportTok{from}\NormalTok{ spotPython.fun.objectivefunctions }\ImportTok{import}\NormalTok{ analytical}
\ImportTok{from}\NormalTok{ spotPython.spot }\ImportTok{import}\NormalTok{ spot}
\ImportTok{from}\NormalTok{ spotPython.design.spacefilling }\ImportTok{import}\NormalTok{ spacefilling}
\ImportTok{from}\NormalTok{ spotPython.build.kriging }\ImportTok{import}\NormalTok{ Kriging}
\ImportTok{import}\NormalTok{ matplotlib.pyplot }\ImportTok{as}\NormalTok{ plt}

\NormalTok{gen }\OperatorTok{=}\NormalTok{ spacefilling(}\DecValTok{1}\NormalTok{)}
\NormalTok{rng }\OperatorTok{=}\NormalTok{ np.random.RandomState(}\DecValTok{1}\NormalTok{)}
\NormalTok{lower }\OperatorTok{=}\NormalTok{ np.array([}\OperatorTok{{-}}\DecValTok{10}\NormalTok{])}
\NormalTok{upper }\OperatorTok{=}\NormalTok{ np.array([}\DecValTok{10}\NormalTok{])}
\NormalTok{fun }\OperatorTok{=}\NormalTok{ analytical().fun\_sphere}
\NormalTok{fun\_control }\OperatorTok{=}\NormalTok{ fun\_control\_init(    }
\NormalTok{    sigma}\OperatorTok{=}\DecValTok{2}\NormalTok{,}
\NormalTok{    seed}\OperatorTok{=}\DecValTok{125}\NormalTok{)}
\NormalTok{X }\OperatorTok{=}\NormalTok{ gen.scipy\_lhd(}\DecValTok{10}\NormalTok{, lower}\OperatorTok{=}\NormalTok{lower, upper }\OperatorTok{=}\NormalTok{ upper)}
\NormalTok{y }\OperatorTok{=}\NormalTok{ fun(X, fun\_control}\OperatorTok{=}\NormalTok{fun\_control)}
\NormalTok{X\_train }\OperatorTok{=}\NormalTok{ X.reshape(}\OperatorTok{{-}}\DecValTok{1}\NormalTok{,}\DecValTok{1}\NormalTok{)}
\NormalTok{y\_train }\OperatorTok{=}\NormalTok{ y}
\end{Highlighting}
\end{Shaded}

A surrogate without nugget is fitted to these data:

\begin{Shaded}
\begin{Highlighting}[]
\NormalTok{S }\OperatorTok{=}\NormalTok{ Kriging(name}\OperatorTok{=}\StringTok{\textquotesingle{}kriging\textquotesingle{}}\NormalTok{,}
\NormalTok{            seed}\OperatorTok{=}\DecValTok{123}\NormalTok{,}
\NormalTok{            log\_level}\OperatorTok{=}\DecValTok{50}\NormalTok{,}
\NormalTok{            n\_theta}\OperatorTok{=}\DecValTok{1}\NormalTok{,}
\NormalTok{            noise}\OperatorTok{=}\VariableTok{False}\NormalTok{)}
\NormalTok{S.fit(X\_train, y\_train)}

\NormalTok{X\_axis }\OperatorTok{=}\NormalTok{ np.linspace(start}\OperatorTok{={-}}\DecValTok{13}\NormalTok{, stop}\OperatorTok{=}\DecValTok{13}\NormalTok{, num}\OperatorTok{=}\DecValTok{1000}\NormalTok{).reshape(}\OperatorTok{{-}}\DecValTok{1}\NormalTok{, }\DecValTok{1}\NormalTok{)}
\NormalTok{mean\_prediction, std\_prediction, ei }\OperatorTok{=}\NormalTok{ S.predict(X\_axis, return\_val}\OperatorTok{=}\StringTok{"all"}\NormalTok{)}

\NormalTok{plt.scatter(X\_train, y\_train, label}\OperatorTok{=}\StringTok{"Observations"}\NormalTok{)}
\NormalTok{plt.plot(X\_axis, mean\_prediction, label}\OperatorTok{=}\StringTok{"mue"}\NormalTok{)}
\NormalTok{plt.legend()}
\NormalTok{plt.xlabel(}\StringTok{"$x$"}\NormalTok{)}
\NormalTok{plt.ylabel(}\StringTok{"$f(x)$"}\NormalTok{)}
\NormalTok{\_ }\OperatorTok{=}\NormalTok{ plt.title(}\StringTok{"Sphere: Gaussian process regression on noisy dataset"}\NormalTok{)}
\end{Highlighting}
\end{Shaded}

\begin{figure}[H]

{\centering \includegraphics{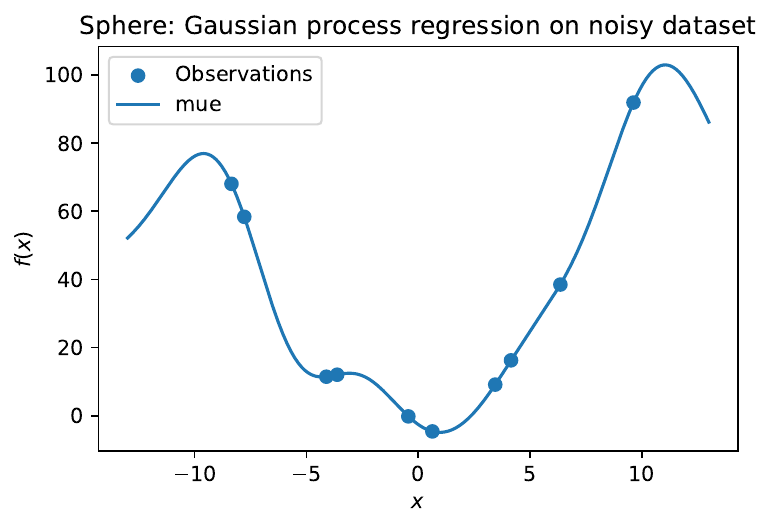}

}

\end{figure}

In comparison to the surrogate without nugget, we fit a surrogate with
nugget to the data:

\begin{Shaded}
\begin{Highlighting}[]
\NormalTok{S\_nug }\OperatorTok{=}\NormalTok{ Kriging(name}\OperatorTok{=}\StringTok{\textquotesingle{}kriging\textquotesingle{}}\NormalTok{,}
\NormalTok{            seed}\OperatorTok{=}\DecValTok{123}\NormalTok{,}
\NormalTok{            log\_level}\OperatorTok{=}\DecValTok{50}\NormalTok{,}
\NormalTok{            n\_theta}\OperatorTok{=}\DecValTok{1}\NormalTok{,}
\NormalTok{            noise}\OperatorTok{=}\VariableTok{True}\NormalTok{)}
\NormalTok{S\_nug.fit(X\_train, y\_train)}
\NormalTok{X\_axis }\OperatorTok{=}\NormalTok{ np.linspace(start}\OperatorTok{={-}}\DecValTok{13}\NormalTok{, stop}\OperatorTok{=}\DecValTok{13}\NormalTok{, num}\OperatorTok{=}\DecValTok{1000}\NormalTok{).reshape(}\OperatorTok{{-}}\DecValTok{1}\NormalTok{, }\DecValTok{1}\NormalTok{)}
\NormalTok{mean\_prediction, std\_prediction, ei }\OperatorTok{=}\NormalTok{ S\_nug.predict(X\_axis, return\_val}\OperatorTok{=}\StringTok{"all"}\NormalTok{)}
\NormalTok{plt.scatter(X\_train, y\_train, label}\OperatorTok{=}\StringTok{"Observations"}\NormalTok{)}
\NormalTok{plt.plot(X\_axis, mean\_prediction, label}\OperatorTok{=}\StringTok{"mue"}\NormalTok{)}
\NormalTok{plt.legend()}
\NormalTok{plt.xlabel(}\StringTok{"$x$"}\NormalTok{)}
\NormalTok{plt.ylabel(}\StringTok{"$f(x)$"}\NormalTok{)}
\NormalTok{\_ }\OperatorTok{=}\NormalTok{ plt.title(}\StringTok{"Sphere: Gaussian process regression with nugget on noisy dataset"}\NormalTok{)}
\end{Highlighting}
\end{Shaded}

\begin{figure}[H]

{\centering \includegraphics{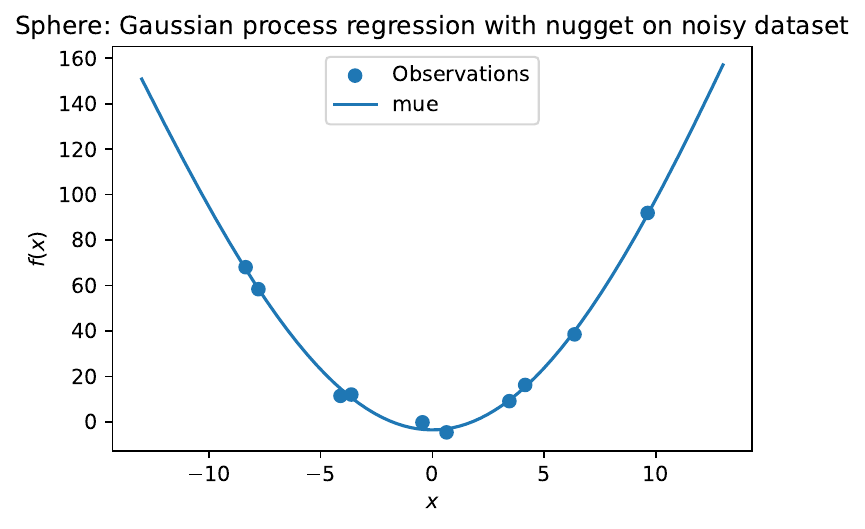}

}

\end{figure}

The value of the nugget term can be extracted from the model as follows:

\begin{Shaded}
\begin{Highlighting}[]
\NormalTok{S.Lambda}
\end{Highlighting}
\end{Shaded}

\begin{Shaded}
\begin{Highlighting}[]
\NormalTok{S\_nug.Lambda}
\end{Highlighting}
\end{Shaded}

\begin{verbatim}
9.088150066416743e-05
\end{verbatim}

We see:

\begin{itemize}
\tightlist
\item
  the first model \texttt{S} has no nugget,
\item
  whereas the second model has a nugget value (\texttt{Lambda}) larger
  than zero.
\end{itemize}

\hypertarget{exercises-6}{%
\section{Exercises}\label{exercises-6}}

\hypertarget{noisy-fun_cubed-1}{%
\subsection{\texorpdfstring{Noisy
\texttt{fun\_cubed}}{Noisy fun\_cubed}}\label{noisy-fun_cubed-1}}

Analyse the effect of noise on the \texttt{fun\_cubed} function with the
following settings:

\begin{Shaded}
\begin{Highlighting}[]
\NormalTok{fun }\OperatorTok{=}\NormalTok{ analytical().fun\_cubed}
\NormalTok{fun\_control }\OperatorTok{=}\NormalTok{ fun\_control\_init(    }
\NormalTok{    sigma}\OperatorTok{=}\DecValTok{10}\NormalTok{,}
\NormalTok{    seed}\OperatorTok{=}\DecValTok{123}\NormalTok{)}
\NormalTok{lower }\OperatorTok{=}\NormalTok{ np.array([}\OperatorTok{{-}}\DecValTok{10}\NormalTok{])}
\NormalTok{upper }\OperatorTok{=}\NormalTok{ np.array([}\DecValTok{10}\NormalTok{])}
\end{Highlighting}
\end{Shaded}

\hypertarget{fun_runge-2}{%
\subsection{\texorpdfstring{\texttt{fun\_runge}}{fun\_runge}}\label{fun_runge-2}}

Analyse the effect of noise on the \texttt{fun\_runge} function with the
following settings:

\begin{Shaded}
\begin{Highlighting}[]
\NormalTok{lower }\OperatorTok{=}\NormalTok{ np.array([}\OperatorTok{{-}}\DecValTok{10}\NormalTok{])}
\NormalTok{upper }\OperatorTok{=}\NormalTok{ np.array([}\DecValTok{10}\NormalTok{])}
\NormalTok{fun }\OperatorTok{=}\NormalTok{ analytical().fun\_runge}
\NormalTok{fun\_control }\OperatorTok{=}\NormalTok{ fun\_control\_init(    }
\NormalTok{    sigma}\OperatorTok{=}\FloatTok{0.25}\NormalTok{,}
\NormalTok{    seed}\OperatorTok{=}\DecValTok{123}\NormalTok{)}
\end{Highlighting}
\end{Shaded}

\hypertarget{fun_forrester-1}{%
\subsection{\texorpdfstring{\texttt{fun\_forrester}}{fun\_forrester}}\label{fun_forrester-1}}

Analyse the effect of noise on the \texttt{fun\_forrester} function with
the following settings:

\begin{Shaded}
\begin{Highlighting}[]
\NormalTok{lower }\OperatorTok{=}\NormalTok{ np.array([}\DecValTok{0}\NormalTok{])}
\NormalTok{upper }\OperatorTok{=}\NormalTok{ np.array([}\DecValTok{1}\NormalTok{])}
\NormalTok{fun }\OperatorTok{=}\NormalTok{ analytical().fun\_forrester}
\NormalTok{fun\_control }\OperatorTok{=}\NormalTok{ \{}\StringTok{"sigma"}\NormalTok{: }\DecValTok{5}\NormalTok{,}
               \StringTok{"seed"}\NormalTok{: }\DecValTok{123}\NormalTok{\}}
\end{Highlighting}
\end{Shaded}

\hypertarget{fun_xsin-1}{%
\subsection{\texorpdfstring{\texttt{fun\_xsin}}{fun\_xsin}}\label{fun_xsin-1}}

Analyse the effect of noise on the \texttt{fun\_xsin} function with the
following settings:

\begin{Shaded}
\begin{Highlighting}[]
\NormalTok{lower }\OperatorTok{=}\NormalTok{ np.array([}\OperatorTok{{-}}\FloatTok{1.}\NormalTok{])}
\NormalTok{upper }\OperatorTok{=}\NormalTok{ np.array([}\FloatTok{1.}\NormalTok{])}
\NormalTok{fun }\OperatorTok{=}\NormalTok{ analytical().fun\_xsin}
\NormalTok{fun\_control }\OperatorTok{=}\NormalTok{ fun\_control\_init(    }
\NormalTok{    sigma}\OperatorTok{=}\FloatTok{0.5}\NormalTok{,}
\NormalTok{    seed}\OperatorTok{=}\DecValTok{123}\NormalTok{)}
\end{Highlighting}
\end{Shaded}

\part{Hyperparameter Tuning}

\hypertarget{sec-hpt-sklearn-svc}{%
\chapter{HPT: sklearn SVC on Moons Data}\label{sec-hpt-sklearn-svc}}

This chapter is a tutorial for the Hyperparameter Tuning (HPT) of a
\texttt{sklearn} SVC model on the Moons dataset.

\hypertarget{sec-setup-10}{%
\section{Step 1: Setup}\label{sec-setup-10}}

Before we consider the detailed experimental setup, we select the
parameters that affect run time, initial design size and the device that
is used.

\begin{tcolorbox}[enhanced jigsaw, left=2mm, title=\textcolor{quarto-callout-caution-color}{\faFire}\hspace{0.5em}{Caution: Run time and initial design size should be increased for real
experiments}, bottomrule=.15mm, titlerule=0mm, breakable, rightrule=.15mm, toprule=.15mm, coltitle=black, colbacktitle=quarto-callout-caution-color!10!white, leftrule=.75mm, arc=.35mm, colframe=quarto-callout-caution-color-frame, bottomtitle=1mm, colback=white, opacitybacktitle=0.6, toptitle=1mm, opacityback=0]

\begin{itemize}
\tightlist
\item
  MAX\_TIME is set to one minute for demonstration purposes. For real
  experiments, this should be increased to at least 1 hour.
\item
  INIT\_SIZE is set to 5 for demonstration purposes. For real
  experiments, this should be increased to at least 10.
\end{itemize}

\end{tcolorbox}

\begin{Shaded}
\begin{Highlighting}[]
\NormalTok{MAX\_TIME }\OperatorTok{=} \DecValTok{1}
\NormalTok{INIT\_SIZE }\OperatorTok{=} \DecValTok{10}
\NormalTok{PREFIX }\OperatorTok{=} \StringTok{"10"}
\end{Highlighting}
\end{Shaded}

\hypertarget{step-2-initialization-of-the-empty-fun_control-dictionary}{%
\section{\texorpdfstring{Step 2: Initialization of the Empty
\texttt{fun\_control}
Dictionary}{Step 2: Initialization of the Empty fun\_control Dictionary}}\label{step-2-initialization-of-the-empty-fun_control-dictionary}}

The \texttt{fun\_control} dictionary is the central data structure that
is used to control the optimization process. It is initialized as
follows:

\begin{Shaded}
\begin{Highlighting}[]
\ImportTok{from}\NormalTok{ spotPython.utils.init }\ImportTok{import}\NormalTok{ fun\_control\_init}
\ImportTok{from}\NormalTok{ spotPython.utils.}\BuiltInTok{file} \ImportTok{import}\NormalTok{ get\_experiment\_name, get\_spot\_tensorboard\_path}
\ImportTok{from}\NormalTok{ spotPython.utils.device }\ImportTok{import}\NormalTok{ getDevice}

\NormalTok{experiment\_name }\OperatorTok{=}\NormalTok{ get\_experiment\_name(prefix}\OperatorTok{=}\NormalTok{PREFIX)}

\NormalTok{fun\_control }\OperatorTok{=}\NormalTok{ fun\_control\_init(}
\NormalTok{    task}\OperatorTok{=}\StringTok{"classification"}\NormalTok{,}
\NormalTok{    spot\_tensorboard\_path}\OperatorTok{=}\NormalTok{get\_spot\_tensorboard\_path(experiment\_name),}
\NormalTok{    TENSORBOARD\_CLEAN}\OperatorTok{=}\VariableTok{True}\NormalTok{)}
\end{Highlighting}
\end{Shaded}

\hypertarget{sec-data-loading-10}{%
\section{Step 3: SKlearn Load Data
(Classification)}\label{sec-data-loading-10}}

Randomly generate classification data.

\begin{Shaded}
\begin{Highlighting}[]
\ImportTok{import}\NormalTok{ pandas }\ImportTok{as}\NormalTok{ pd}
\ImportTok{import}\NormalTok{ numpy }\ImportTok{as}\NormalTok{ np}
\ImportTok{from}\NormalTok{ sklearn.model\_selection }\ImportTok{import}\NormalTok{ train\_test\_split}
\ImportTok{from}\NormalTok{ sklearn.datasets }\ImportTok{import}\NormalTok{ make\_moons, make\_circles, make\_classification}
\NormalTok{n\_features }\OperatorTok{=} \DecValTok{2}
\NormalTok{n\_samples }\OperatorTok{=} \DecValTok{500}
\NormalTok{target\_column }\OperatorTok{=} \StringTok{"y"}
\NormalTok{ds }\OperatorTok{=}\NormalTok{  make\_moons(n\_samples, noise}\OperatorTok{=}\FloatTok{0.5}\NormalTok{, random\_state}\OperatorTok{=}\DecValTok{0}\NormalTok{)}
\NormalTok{X, y }\OperatorTok{=}\NormalTok{ ds}
\NormalTok{X\_train, X\_test, y\_train, y\_test }\OperatorTok{=}\NormalTok{ train\_test\_split(}
\NormalTok{    X, y, test\_size}\OperatorTok{=}\FloatTok{0.3}\NormalTok{, random\_state}\OperatorTok{=}\DecValTok{42}
\NormalTok{)}
\NormalTok{train }\OperatorTok{=}\NormalTok{ pd.DataFrame(np.hstack((X\_train, y\_train.reshape(}\OperatorTok{{-}}\DecValTok{1}\NormalTok{, }\DecValTok{1}\NormalTok{))))}
\NormalTok{test }\OperatorTok{=}\NormalTok{ pd.DataFrame(np.hstack((X\_test, y\_test.reshape(}\OperatorTok{{-}}\DecValTok{1}\NormalTok{, }\DecValTok{1}\NormalTok{))))}
\NormalTok{train.columns }\OperatorTok{=}\NormalTok{ [}\SpecialStringTok{f"x}\SpecialCharTok{\{}\NormalTok{i}\SpecialCharTok{\}}\SpecialStringTok{"} \ControlFlowTok{for}\NormalTok{ i }\KeywordTok{in} \BuiltInTok{range}\NormalTok{(}\DecValTok{1}\NormalTok{, n\_features}\OperatorTok{+}\DecValTok{1}\NormalTok{)] }\OperatorTok{+}\NormalTok{ [target\_column]}
\NormalTok{test.columns }\OperatorTok{=}\NormalTok{ [}\SpecialStringTok{f"x}\SpecialCharTok{\{}\NormalTok{i}\SpecialCharTok{\}}\SpecialStringTok{"} \ControlFlowTok{for}\NormalTok{ i }\KeywordTok{in} \BuiltInTok{range}\NormalTok{(}\DecValTok{1}\NormalTok{, n\_features}\OperatorTok{+}\DecValTok{1}\NormalTok{)] }\OperatorTok{+}\NormalTok{ [target\_column]}
\NormalTok{train.head()}
\end{Highlighting}
\end{Shaded}

\begin{longtable}[]{@{}llll@{}}
\toprule\noalign{}
& x1 & x2 & y \\
\midrule\noalign{}
\endhead
\bottomrule\noalign{}
\endlastfoot
0 & 1.960101 & 0.383172 & 0.0 \\
1 & 2.354420 & -0.536942 & 1.0 \\
2 & 1.682186 & -0.332108 & 0.0 \\
3 & 1.856507 & 0.687220 & 1.0 \\
4 & 1.925524 & 0.427413 & 1.0 \\
\end{longtable}

\begin{Shaded}
\begin{Highlighting}[]
\ImportTok{import}\NormalTok{ matplotlib.pyplot }\ImportTok{as}\NormalTok{ plt}
\ImportTok{from}\NormalTok{ matplotlib.colors }\ImportTok{import}\NormalTok{ ListedColormap}

\NormalTok{x\_min, x\_max }\OperatorTok{=}\NormalTok{ X[:, }\DecValTok{0}\NormalTok{].}\BuiltInTok{min}\NormalTok{() }\OperatorTok{{-}} \FloatTok{0.5}\NormalTok{, X[:, }\DecValTok{0}\NormalTok{].}\BuiltInTok{max}\NormalTok{() }\OperatorTok{+} \FloatTok{0.5}
\NormalTok{y\_min, y\_max }\OperatorTok{=}\NormalTok{ X[:, }\DecValTok{1}\NormalTok{].}\BuiltInTok{min}\NormalTok{() }\OperatorTok{{-}} \FloatTok{0.5}\NormalTok{, X[:, }\DecValTok{1}\NormalTok{].}\BuiltInTok{max}\NormalTok{() }\OperatorTok{+} \FloatTok{0.5}
\NormalTok{cm }\OperatorTok{=}\NormalTok{ plt.cm.RdBu}
\NormalTok{cm\_bright }\OperatorTok{=}\NormalTok{ ListedColormap([}\StringTok{"\#FF0000"}\NormalTok{, }\StringTok{"\#0000FF"}\NormalTok{])}
\NormalTok{ax }\OperatorTok{=}\NormalTok{ plt.subplot(}\DecValTok{1}\NormalTok{, }\DecValTok{1}\NormalTok{, }\DecValTok{1}\NormalTok{)}
\NormalTok{ax.set\_title(}\StringTok{"Input data"}\NormalTok{)}
\CommentTok{\# Plot the training points}
\NormalTok{ax.scatter(X\_train[:, }\DecValTok{0}\NormalTok{], X\_train[:, }\DecValTok{1}\NormalTok{], c}\OperatorTok{=}\NormalTok{y\_train, cmap}\OperatorTok{=}\NormalTok{cm\_bright, edgecolors}\OperatorTok{=}\StringTok{"k"}\NormalTok{)}
\CommentTok{\# Plot the testing points}
\NormalTok{ax.scatter(}
\NormalTok{    X\_test[:, }\DecValTok{0}\NormalTok{], X\_test[:, }\DecValTok{1}\NormalTok{], c}\OperatorTok{=}\NormalTok{y\_test, cmap}\OperatorTok{=}\NormalTok{cm\_bright, alpha}\OperatorTok{=}\FloatTok{0.6}\NormalTok{, edgecolors}\OperatorTok{=}\StringTok{"k"}
\NormalTok{)}
\NormalTok{ax.set\_xlim(x\_min, x\_max)}
\NormalTok{ax.set\_ylim(y\_min, y\_max)}
\NormalTok{ax.set\_xticks(())}
\NormalTok{ax.set\_yticks(())}
\NormalTok{plt.tight\_layout()}
\NormalTok{plt.show()}
\end{Highlighting}
\end{Shaded}

\begin{figure}[H]

{\centering \includegraphics{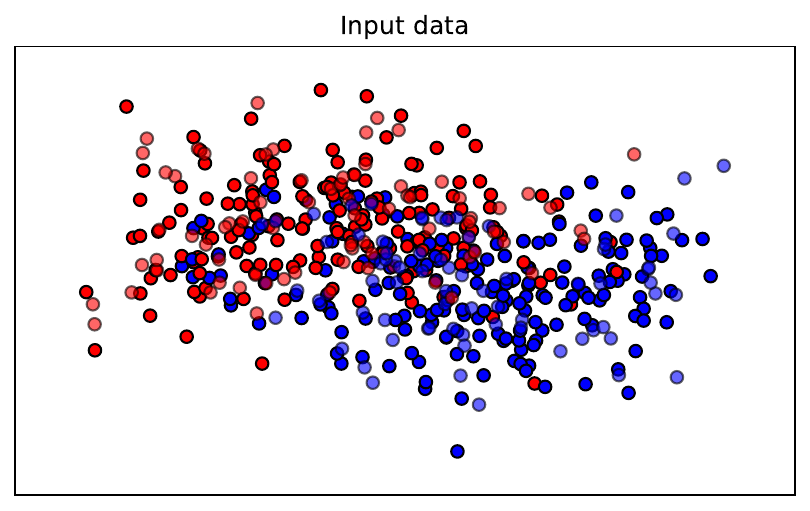}

}

\end{figure}

\begin{Shaded}
\begin{Highlighting}[]
\NormalTok{n\_samples }\OperatorTok{=} \BuiltInTok{len}\NormalTok{(train)}
\CommentTok{\# add the dataset to the fun\_control}
\NormalTok{fun\_control.update(\{}\StringTok{"data"}\NormalTok{: }\VariableTok{None}\NormalTok{, }\CommentTok{\# dataset,}
               \StringTok{"train"}\NormalTok{: train,}
               \StringTok{"test"}\NormalTok{: test,}
               \StringTok{"n\_samples"}\NormalTok{: n\_samples,}
               \StringTok{"target\_column"}\NormalTok{: target\_column\})}
\end{Highlighting}
\end{Shaded}

\hypertarget{sec-specification-of-preprocessing-model-10}{%
\section{Step 4: Specification of the Preprocessing
Model}\label{sec-specification-of-preprocessing-model-10}}

Data preprocesssing can be very simple, e.g., you can ignore it. Then
you would choose the \texttt{prep\_model} ``None'':

\begin{Shaded}
\begin{Highlighting}[]
\NormalTok{prep\_model }\OperatorTok{=} \VariableTok{None}
\NormalTok{fun\_control.update(\{}\StringTok{"prep\_model"}\NormalTok{: prep\_model\})}
\end{Highlighting}
\end{Shaded}

A default approach for numerical data is the \texttt{StandardScaler}
(mean 0, variance 1). This can be selected as follows:

\begin{Shaded}
\begin{Highlighting}[]
\ImportTok{from}\NormalTok{ sklearn.preprocessing }\ImportTok{import}\NormalTok{ StandardScaler}
\NormalTok{prep\_model }\OperatorTok{=}\NormalTok{ StandardScaler()}
\NormalTok{fun\_control.update(\{}\StringTok{"prep\_model"}\NormalTok{: prep\_model\})}
\end{Highlighting}
\end{Shaded}

Even more complicated pre-processing steps are possible, e.g., the
follwing pipeline:

\begin{Shaded}
\begin{Highlighting}[]
\NormalTok{categorical\_columns = []}
\NormalTok{one\_hot\_encoder = OneHotEncoder(handle\_unknown="ignore", sparse\_output=False)}
\NormalTok{prep\_model = ColumnTransformer(}
\NormalTok{         transformers=[}
\NormalTok{             ("categorical", one\_hot\_encoder, categorical\_columns),}
\NormalTok{         ],}
\NormalTok{         remainder=StandardScaler(),}
\NormalTok{     )}
\end{Highlighting}
\end{Shaded}

\hypertarget{step-5-select-model-algorithm-and-core_model_hyper_dict}{%
\section{\texorpdfstring{Step 5: Select Model (\texttt{algorithm}) and
\texttt{core\_model\_hyper\_dict}}{Step 5: Select Model (algorithm) and core\_model\_hyper\_dict}}\label{step-5-select-model-algorithm-and-core_model_hyper_dict}}

The selection of the algorithm (ML model) that should be tuned is done
by specifying the its name from the \texttt{sklearn} implementation. For
example, the \texttt{SVC} support vector machine classifier is selected
as follows:

\begin{Shaded}
\begin{Highlighting}[]
\ImportTok{from}\NormalTok{ spotPython.hyperparameters.values }\ImportTok{import}\NormalTok{ add\_core\_model\_to\_fun\_control}
\ImportTok{from}\NormalTok{ spotPython.data.sklearn\_hyper\_dict }\ImportTok{import}\NormalTok{ SklearnHyperDict}
\ImportTok{from}\NormalTok{ sklearn.svm }\ImportTok{import}\NormalTok{ SVC}
\NormalTok{add\_core\_model\_to\_fun\_control(core\_model}\OperatorTok{=}\NormalTok{SVC,}
\NormalTok{                              fun\_control}\OperatorTok{=}\NormalTok{fun\_control,}
\NormalTok{                              hyper\_dict}\OperatorTok{=}\NormalTok{SklearnHyperDict,}
\NormalTok{                              filename}\OperatorTok{=}\VariableTok{None}\NormalTok{)}
\end{Highlighting}
\end{Shaded}

Now \texttt{fun\_control} has the information from the JSON file. The
corresponding entries for the \texttt{core\_model} class are shown
below.

\begin{Shaded}
\begin{Highlighting}[]
\NormalTok{fun\_control[}\StringTok{\textquotesingle{}core\_model\_hyper\_dict\textquotesingle{}}\NormalTok{]}
\end{Highlighting}
\end{Shaded}

\begin{verbatim}
{'C': {'type': 'float',
  'default': 1.0,
  'transform': 'None',
  'lower': 0.1,
  'upper': 10.0},
 'kernel': {'levels': ['linear', 'poly', 'rbf', 'sigmoid'],
  'type': 'factor',
  'default': 'rbf',
  'transform': 'None',
  'core_model_parameter_type': 'str',
  'lower': 0,
  'upper': 3},
 'degree': {'type': 'int',
  'default': 3,
  'transform': 'None',
  'lower': 3,
  'upper': 3},
 'gamma': {'levels': ['scale', 'auto'],
  'type': 'factor',
  'default': 'scale',
  'transform': 'None',
  'core_model_parameter_type': 'str',
  'lower': 0,
  'upper': 1},
 'coef0': {'type': 'float',
  'default': 0.0,
  'transform': 'None',
  'lower': 0.0,
  'upper': 0.0},
 'shrinking': {'levels': [0, 1],
  'type': 'factor',
  'default': 0,
  'transform': 'None',
  'core_model_parameter_type': 'bool',
  'lower': 0,
  'upper': 1},
 'probability': {'levels': [0, 1],
  'type': 'factor',
  'default': 0,
  'transform': 'None',
  'core_model_parameter_type': 'bool',
  'lower': 0,
  'upper': 1},
 'tol': {'type': 'float',
  'default': 0.001,
  'transform': 'None',
  'lower': 0.0001,
  'upper': 0.01},
 'cache_size': {'type': 'float',
  'default': 200,
  'transform': 'None',
  'lower': 100,
  'upper': 400},
 'break_ties': {'levels': [0, 1],
  'type': 'factor',
  'default': 0,
  'transform': 'None',
  'core_model_parameter_type': 'bool',
  'lower': 0,
  'upper': 1}}
\end{verbatim}

\begin{tcolorbox}[enhanced jigsaw, left=2mm, title=\textcolor{quarto-callout-note-color}{\faInfo}\hspace{0.5em}{\texttt{sklearn\ Model} Selection}, bottomrule=.15mm, titlerule=0mm, breakable, rightrule=.15mm, toprule=.15mm, coltitle=black, colbacktitle=quarto-callout-note-color!10!white, leftrule=.75mm, arc=.35mm, colframe=quarto-callout-note-color-frame, bottomtitle=1mm, colback=white, opacitybacktitle=0.6, toptitle=1mm, opacityback=0]

The following \texttt{sklearn} models are supported by default:

\begin{itemize}
\tightlist
\item
  RidgeCV
\item
  RandomForestClassifier
\item
  SVC
\item
  LogisticRegression
\item
  KNeighborsClassifier
\item
  GradientBoostingClassifier
\item
  GradientBoostingRegressor
\item
  ElasticNet
\end{itemize}

They can be imported as follows:

\begin{Shaded}
\begin{Highlighting}[]
\NormalTok{from sklearn.linear\_model import RidgeCV}
\NormalTok{from sklearn.ensemble import RandomForestClassifier}
\NormalTok{from sklearn.svm import SVC}
\NormalTok{from sklearn.linear\_model import LogisticRegression}
\NormalTok{from sklearn.neighbors import KNeighborsClassifier}
\NormalTok{from sklearn.ensemble import GradientBoostingClassifier}
\NormalTok{from sklearn.ensemble import GradientBoostingRegressor}
\NormalTok{from sklearn.linear\_model import ElasticNet}
\end{Highlighting}
\end{Shaded}

\end{tcolorbox}

\hypertarget{step-6-modify-hyper_dict-hyperparameters-for-the-selected-algorithm-aka-core_model}{%
\section{\texorpdfstring{Step 6: Modify \texttt{hyper\_dict}
Hyperparameters for the Selected Algorithm aka
\texttt{core\_model}}{Step 6: Modify hyper\_dict Hyperparameters for the Selected Algorithm aka core\_model}}\label{step-6-modify-hyper_dict-hyperparameters-for-the-selected-algorithm-aka-core_model}}

\texttt{spotPython} provides functions for modifying the
hyperparameters, their bounds and factors as well as for activating and
de-activating hyperparameters without re-compilation of the Python
source code. These functions were described in
Section~\ref{sec-modification-of-hyperparameters-14}.

\hypertarget{modify-hyperparameter-of-type-numeric-and-integer-boolean}{%
\subsection{Modify hyperparameter of type numeric and integer
(boolean)}\label{modify-hyperparameter-of-type-numeric-and-integer-boolean}}

Numeric and boolean values can be modified using the
\texttt{modify\_hyper\_parameter\_bounds} method.

\begin{tcolorbox}[enhanced jigsaw, left=2mm, title=\textcolor{quarto-callout-note-color}{\faInfo}\hspace{0.5em}{\texttt{sklearn\ Model} Hyperparameters}, bottomrule=.15mm, titlerule=0mm, breakable, rightrule=.15mm, toprule=.15mm, coltitle=black, colbacktitle=quarto-callout-note-color!10!white, leftrule=.75mm, arc=.35mm, colframe=quarto-callout-note-color-frame, bottomtitle=1mm, colback=white, opacitybacktitle=0.6, toptitle=1mm, opacityback=0]

The hyperparameters of the \texttt{sklearn} \texttt{SVC} model are
described in the
\href{https://scikit-learn.org/stable/modules/generated/sklearn.svm.SVC.html}{sklearn
documentation}.

\end{tcolorbox}

\begin{itemize}
\tightlist
\item
  For example, to change the \texttt{tol} hyperparameter of the
  \texttt{SVC} model to the interval {[}1e-5, 1e-3{]}, the following
  code can be used:
\end{itemize}

\begin{Shaded}
\begin{Highlighting}[]
\ImportTok{from}\NormalTok{ spotPython.hyperparameters.values }\ImportTok{import}\NormalTok{ modify\_hyper\_parameter\_bounds}
\NormalTok{modify\_hyper\_parameter\_bounds(fun\_control, }\StringTok{"tol"}\NormalTok{, bounds}\OperatorTok{=}\NormalTok{[}\FloatTok{1e{-}5}\NormalTok{, }\FloatTok{1e{-}3}\NormalTok{])}
\NormalTok{modify\_hyper\_parameter\_bounds(fun\_control, }\StringTok{"probability"}\NormalTok{, bounds}\OperatorTok{=}\NormalTok{[}\DecValTok{0}\NormalTok{, }\DecValTok{0}\NormalTok{])}
\NormalTok{fun\_control[}\StringTok{"core\_model\_hyper\_dict"}\NormalTok{][}\StringTok{"tol"}\NormalTok{]}
\end{Highlighting}
\end{Shaded}

\begin{verbatim}
{'type': 'float',
 'default': 0.001,
 'transform': 'None',
 'lower': 1e-05,
 'upper': 0.001}
\end{verbatim}

\hypertarget{modify-hyperparameter-of-type-factor}{%
\subsection{Modify hyperparameter of type
factor}\label{modify-hyperparameter-of-type-factor}}

Factors can be modified with the
\texttt{modify\_hyper\_parameter\_levels} function. For example, to
exclude the \texttt{sigmoid} kernel from the tuning, the \texttt{kernel}
hyperparameter of the \texttt{SVC} model can be modified as follows:

\begin{Shaded}
\begin{Highlighting}[]
\ImportTok{from}\NormalTok{ spotPython.hyperparameters.values }\ImportTok{import}\NormalTok{ modify\_hyper\_parameter\_levels}
\NormalTok{modify\_hyper\_parameter\_levels(fun\_control, }\StringTok{"kernel"}\NormalTok{, [}\StringTok{"poly"}\NormalTok{, }\StringTok{"rbf"}\NormalTok{])}
\NormalTok{fun\_control[}\StringTok{"core\_model\_hyper\_dict"}\NormalTok{][}\StringTok{"kernel"}\NormalTok{]}
\end{Highlighting}
\end{Shaded}

\begin{verbatim}
{'levels': ['poly', 'rbf'],
 'type': 'factor',
 'default': 'rbf',
 'transform': 'None',
 'core_model_parameter_type': 'str',
 'lower': 0,
 'upper': 1}
\end{verbatim}

\hypertarget{sec-optimizers-10}{%
\subsection{Optimizers}\label{sec-optimizers-10}}

Optimizers are described in Section~\ref{sec-optimizers-14}.

\hypertarget{step-7-selection-of-the-objective-loss-function}{%
\section{Step 7: Selection of the Objective (Loss)
Function}\label{step-7-selection-of-the-objective-loss-function}}

There are two metrics:

\begin{enumerate}
\def\labelenumi{\arabic{enumi}.}
\tightlist
\item
  \texttt{metric\_river} is used for the river based evaluation via
  \texttt{eval\_oml\_iter\_progressive}.
\item
  \texttt{metric\_sklearn} is used for the sklearn based evaluation.
\end{enumerate}

\begin{Shaded}
\begin{Highlighting}[]
\ImportTok{from}\NormalTok{ sklearn.metrics }\ImportTok{import}\NormalTok{ mean\_absolute\_error, accuracy\_score, roc\_curve, roc\_auc\_score, log\_loss, mean\_squared\_error}
\NormalTok{fun\_control.update(\{}
               \StringTok{"metric\_sklearn"}\NormalTok{: log\_loss,}
               \StringTok{"weights"}\NormalTok{: }\FloatTok{1.0}\NormalTok{,}
\NormalTok{               \})}
\end{Highlighting}
\end{Shaded}

\begin{tcolorbox}[enhanced jigsaw, left=2mm, title=\textcolor{quarto-callout-warning-color}{\faExclamationTriangle}\hspace{0.5em}{\texttt{metric\_sklearn}: Minimization and Maximization}, bottomrule=.15mm, titlerule=0mm, breakable, rightrule=.15mm, toprule=.15mm, coltitle=black, colbacktitle=quarto-callout-warning-color!10!white, leftrule=.75mm, arc=.35mm, colframe=quarto-callout-warning-color-frame, bottomtitle=1mm, colback=white, opacitybacktitle=0.6, toptitle=1mm, opacityback=0]

\begin{itemize}
\tightlist
\item
  Because the \texttt{metric\_sklearn} is used for the sklearn based
  evaluation, it is important to know whether the metric should be
  minimized or maximized.
\item
  The \texttt{weights} parameter is used to indicate whether the metric
  should be minimized or maximized.
\item
  If \texttt{weights} is set to \texttt{-1.0}, the metric is maximized.
\item
  If \texttt{weights} is set to \texttt{1.0}, the metric is minimized,
  e.g., \texttt{weights\ =\ 1.0} for \texttt{mean\_absolute\_error}, or
  \texttt{weights\ =\ -1.0} for \texttt{roc\_auc\_score}.
\end{itemize}

\end{tcolorbox}

\hypertarget{predict-classes-or-class-probabilities}{%
\subsection{Predict Classes or Class
Probabilities}\label{predict-classes-or-class-probabilities}}

If the key \texttt{"predict\_proba"} is set to \texttt{True}, the class
probabilities are predicted. \texttt{False} is the default, i.e., the
classes are predicted.

\begin{Shaded}
\begin{Highlighting}[]
\NormalTok{fun\_control.update(\{}
               \StringTok{"predict\_proba"}\NormalTok{: }\VariableTok{False}\NormalTok{,}
\NormalTok{               \})}
\end{Highlighting}
\end{Shaded}

\hypertarget{step-8-calling-the-spot-function}{%
\section{Step 8: Calling the SPOT
Function}\label{step-8-calling-the-spot-function}}

\hypertarget{sec-prepare-spot-call-10}{%
\subsection{Preparing the SPOT Call}\label{sec-prepare-spot-call-10}}

The following code passes the information about the parameter ranges and
bounds to \texttt{spot}.

\begin{Shaded}
\begin{Highlighting}[]
\CommentTok{\# extract the variable types, names, and bounds}
\ImportTok{from}\NormalTok{ spotPython.hyperparameters.values }\ImportTok{import}\NormalTok{ (    }
\NormalTok{    get\_var\_name,}
\NormalTok{    get\_var\_type,}
\NormalTok{    get\_bound\_values}
\NormalTok{    )}
\NormalTok{var\_type }\OperatorTok{=}\NormalTok{ get\_var\_type(fun\_control)}
\NormalTok{var\_name }\OperatorTok{=}\NormalTok{ get\_var\_name(fun\_control)}
\NormalTok{lower }\OperatorTok{=}\NormalTok{ get\_bound\_values(fun\_control, }\StringTok{"lower"}\NormalTok{)}
\NormalTok{upper }\OperatorTok{=}\NormalTok{ get\_bound\_values(fun\_control, }\StringTok{"upper"}\NormalTok{)}
\end{Highlighting}
\end{Shaded}

\begin{Shaded}
\begin{Highlighting}[]
\ImportTok{from}\NormalTok{ spotPython.utils.eda }\ImportTok{import}\NormalTok{ gen\_design\_table}
\BuiltInTok{print}\NormalTok{(gen\_design\_table(fun\_control))}
\end{Highlighting}
\end{Shaded}

\begin{verbatim}
| name        | type   | default   |   lower |   upper | transform   |
|-------------|--------|-----------|---------|---------|-------------|
| C           | float  | 1.0       |   0.1   |  10     | None        |
| kernel      | factor | rbf       |   0     |   1     | None        |
| degree      | int    | 3         |   3     |   3     | None        |
| gamma       | factor | scale     |   0     |   1     | None        |
| coef0       | float  | 0.0       |   0     |   0     | None        |
| shrinking   | factor | 0         |   0     |   1     | None        |
| probability | factor | 0         |   0     |   0     | None        |
| tol         | float  | 0.001     |   1e-05 |   0.001 | None        |
| cache_size  | float  | 200.0     | 100     | 400     | None        |
| break_ties  | factor | 0         |   0     |   1     | None        |
\end{verbatim}

\hypertarget{sec-the-objective-function-10}{%
\subsection{The Objective
Function}\label{sec-the-objective-function-10}}

The objective function is selected next. It implements an interface from
\texttt{sklearn}'s training, validation, and testing methods to
\texttt{spotPython}.

\begin{Shaded}
\begin{Highlighting}[]
\ImportTok{from}\NormalTok{ spotPython.fun.hypersklearn }\ImportTok{import}\NormalTok{ HyperSklearn}
\NormalTok{fun }\OperatorTok{=}\NormalTok{ HyperSklearn().fun\_sklearn}
\end{Highlighting}
\end{Shaded}

\begin{Shaded}
\begin{Highlighting}[]
\ImportTok{from}\NormalTok{ spotPython.hyperparameters.values }\ImportTok{import}\NormalTok{ get\_default\_hyperparameters\_as\_array}
\CommentTok{\# X\_start = get\_default\_hyperparameters\_as\_array(fun\_control)}
\end{Highlighting}
\end{Shaded}

\hypertarget{run-the-spot-optimizer}{%
\subsection{\texorpdfstring{Run the \texttt{Spot}
Optimizer}{Run the Spot Optimizer}}\label{run-the-spot-optimizer}}

\begin{itemize}
\tightlist
\item
  Run SPOT for approx. x mins (\texttt{max\_time}).
\item
  Note: the run takes longer, because the evaluation time of initial
  design (here: \texttt{initi\_size}, 20 points) is not considered.
\end{itemize}

\hypertarget{sec-call-the-hyperparameter-tuner-10}{%
\subsection{Starting the Hyperparameter
Tuning}\label{sec-call-the-hyperparameter-tuner-10}}

\begin{Shaded}
\begin{Highlighting}[]
\ImportTok{import}\NormalTok{ numpy }\ImportTok{as}\NormalTok{ np}
\ImportTok{from}\NormalTok{ spotPython.spot }\ImportTok{import}\NormalTok{ spot}
\ImportTok{from}\NormalTok{ math }\ImportTok{import}\NormalTok{ inf}
\NormalTok{spot\_tuner }\OperatorTok{=}\NormalTok{ spot.Spot(fun}\OperatorTok{=}\NormalTok{fun,}
\NormalTok{                   lower }\OperatorTok{=}\NormalTok{ lower,}
\NormalTok{                   upper }\OperatorTok{=}\NormalTok{ upper,}
\NormalTok{                   fun\_evals }\OperatorTok{=}\NormalTok{ inf,}
\NormalTok{                   fun\_repeats }\OperatorTok{=} \DecValTok{1}\NormalTok{,}
\NormalTok{                   max\_time }\OperatorTok{=}\NormalTok{ MAX\_TIME,}
\NormalTok{                   noise }\OperatorTok{=} \VariableTok{False}\NormalTok{,}
\NormalTok{                   tolerance\_x }\OperatorTok{=}\NormalTok{ np.sqrt(np.spacing(}\DecValTok{1}\NormalTok{)),}
\NormalTok{                   var\_type }\OperatorTok{=}\NormalTok{ var\_type,}
\NormalTok{                   var\_name }\OperatorTok{=}\NormalTok{ var\_name,}
\NormalTok{                   infill\_criterion }\OperatorTok{=} \StringTok{"y"}\NormalTok{,}
\NormalTok{                   n\_points }\OperatorTok{=} \DecValTok{1}\NormalTok{,}
\NormalTok{                   seed}\OperatorTok{=}\DecValTok{123}\NormalTok{,}
\NormalTok{                   log\_level }\OperatorTok{=} \DecValTok{50}\NormalTok{,}
\NormalTok{                   show\_models}\OperatorTok{=} \VariableTok{False}\NormalTok{,}
\NormalTok{                   show\_progress}\OperatorTok{=} \VariableTok{True}\NormalTok{,}
\NormalTok{                   fun\_control }\OperatorTok{=}\NormalTok{ fun\_control,}
\NormalTok{                   design\_control}\OperatorTok{=}\NormalTok{\{}\StringTok{"init\_size"}\NormalTok{: INIT\_SIZE,}
                                   \StringTok{"repeats"}\NormalTok{: }\DecValTok{1}\NormalTok{\},}
\NormalTok{                   surrogate\_control}\OperatorTok{=}\NormalTok{\{}\StringTok{"noise"}\NormalTok{: }\VariableTok{True}\NormalTok{,}
                                      \StringTok{"cod\_type"}\NormalTok{: }\StringTok{"norm"}\NormalTok{,}
                                      \StringTok{"min\_theta"}\NormalTok{: }\OperatorTok{{-}}\DecValTok{4}\NormalTok{,}
                                      \StringTok{"max\_theta"}\NormalTok{: }\DecValTok{3}\NormalTok{,}
                                      \StringTok{"n\_theta"}\NormalTok{: }\BuiltInTok{len}\NormalTok{(var\_name),}
                                      \StringTok{"model\_fun\_evals"}\NormalTok{: }\DecValTok{10\_000}\NormalTok{,}
                                      \StringTok{"log\_level"}\NormalTok{: }\DecValTok{50}
\NormalTok{                                      \})}
\NormalTok{spot\_tuner.run()}
\end{Highlighting}
\end{Shaded}

\begin{verbatim}
spotPython tuning: 5.734217584632275 [----------] 1.53% 
\end{verbatim}

\begin{verbatim}
spotPython tuning: 5.734217584632275 [----------] 3.47% 
\end{verbatim}

\begin{verbatim}
spotPython tuning: 5.734217584632275 [#---------] 5.57% 
\end{verbatim}

\begin{verbatim}
spotPython tuning: 5.734217584632275 [#---------] 7.52% 
\end{verbatim}

\begin{verbatim}
spotPython tuning: 5.734217584632275 [#---------] 9.29% 
\end{verbatim}

\begin{verbatim}
spotPython tuning: 5.734217584632275 [#---------] 11.16% 
\end{verbatim}

\begin{verbatim}
spotPython tuning: 5.734217584632275 [#---------] 13.14% 
\end{verbatim}

\begin{verbatim}
spotPython tuning: 5.734217584632275 [##--------] 21.26% 
\end{verbatim}

\begin{verbatim}
spotPython tuning: 5.734217584632275 [###-------] 29.05% 
\end{verbatim}

\begin{verbatim}
spotPython tuning: 5.734217584632275 [####------] 38.06% 
\end{verbatim}

\begin{verbatim}
spotPython tuning: 5.734217584632275 [#####-----] 46.45% 
\end{verbatim}

\begin{verbatim}
spotPython tuning: 5.734217584632275 [######----] 56.03% 
\end{verbatim}

\begin{verbatim}
spotPython tuning: 5.734217584632275 [#######---] 65.53% 
\end{verbatim}

\begin{verbatim}
spotPython tuning: 5.734217584632275 [#######---] 73.32% 
\end{verbatim}

\begin{verbatim}
spotPython tuning: 5.734217584632275 [#########-] 85.10% 
\end{verbatim}

\begin{verbatim}
spotPython tuning: 5.734217584632275 [#########-] 93.14% 
\end{verbatim}

\begin{verbatim}
spotPython tuning: 5.734217584632275 [##########] 99.32% 
\end{verbatim}

\begin{verbatim}
spotPython tuning: 5.734217584632275 [##########] 100.00% Done...
\end{verbatim}

\begin{verbatim}
<spotPython.spot.spot.Spot at 0x297b03f70>
\end{verbatim}

\hypertarget{sec-results-tuning-10}{%
\section{Step 9: Results}\label{sec-results-tuning-10}}

\begin{Shaded}
\begin{Highlighting}[]
\ImportTok{from}\NormalTok{ spotPython.utils.}\BuiltInTok{file} \ImportTok{import}\NormalTok{ save\_pickle}
\NormalTok{save\_pickle(spot\_tuner, experiment\_name)}
\end{Highlighting}
\end{Shaded}

\begin{Shaded}
\begin{Highlighting}[]
\ImportTok{from}\NormalTok{ spotPython.utils.}\BuiltInTok{file} \ImportTok{import}\NormalTok{ load\_pickle}
\NormalTok{spot\_tuner }\OperatorTok{=}\NormalTok{ load\_pickle(experiment\_name)}
\end{Highlighting}
\end{Shaded}

\begin{itemize}
\tightlist
\item
  Show the Progress of the hyperparameter tuning:
\end{itemize}

After the hyperparameter tuning run is finished, the progress of the
hyperparameter tuning can be visualized.

\begin{Shaded}
\begin{Highlighting}[]
\NormalTok{spot\_tuner.plot\_progress(log\_y}\OperatorTok{=}\VariableTok{False}\NormalTok{,}
\NormalTok{    filename}\OperatorTok{=}\StringTok{"./figures/"} \OperatorTok{+}\NormalTok{ experiment\_name}\OperatorTok{+}\StringTok{"\_progress.png"}\NormalTok{)}
\end{Highlighting}
\end{Shaded}

\begin{figure}[H]

{\centering \includegraphics{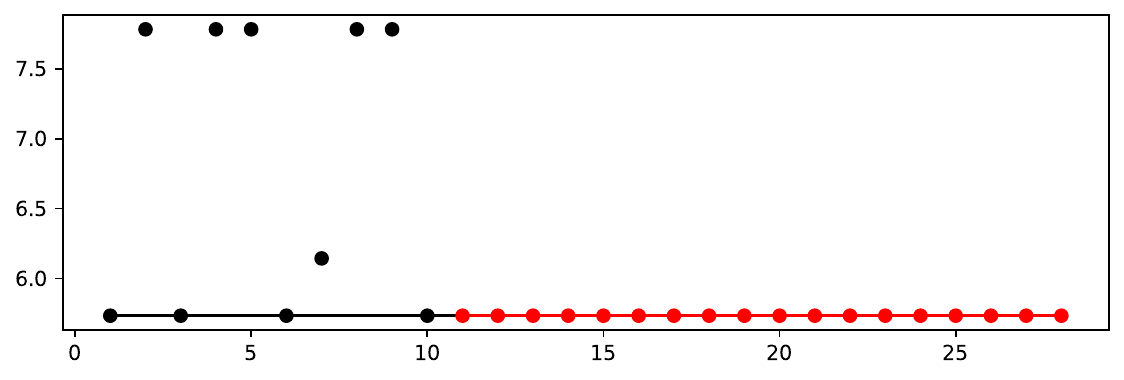}

}

\caption{Progress plot. \emph{Black} dots denote results from the
initial design. \emph{Red} dots illustrate the improvement found by the
surrogate model based optimization.}

\end{figure}

\begin{itemize}
\tightlist
\item
  Print the results
\end{itemize}

\begin{Shaded}
\begin{Highlighting}[]
\BuiltInTok{print}\NormalTok{(gen\_design\_table(fun\_control}\OperatorTok{=}\NormalTok{fun\_control,}
\NormalTok{    spot}\OperatorTok{=}\NormalTok{spot\_tuner))}
\end{Highlighting}
\end{Shaded}

\begin{verbatim}
| name        | type   | default   |   lower |   upper |                tuned | transform   |   importance | stars   |
|-------------|--------|-----------|---------|---------|----------------------|-------------|--------------|---------|
| C           | float  | 1.0       |     0.1 |    10.0 |    2.394471655384338 | None        |         6.79 | *       |
| kernel      | factor | rbf       |     0.0 |     1.0 |                  1.0 | None        |       100.00 | ***     |
| degree      | int    | 3         |     3.0 |     3.0 |                  3.0 | None        |         0.00 |         |
| gamma       | factor | scale     |     0.0 |     1.0 |                  0.0 | None        |         0.00 |         |
| coef0       | float  | 0.0       |     0.0 |     0.0 |                  0.0 | None        |         0.00 |         |
| shrinking   | factor | 0         |     0.0 |     1.0 |                  0.0 | None        |         0.00 |         |
| probability | factor | 0         |     0.0 |     0.0 |                  0.0 | None        |         0.00 |         |
| tol         | float  | 0.001     |   1e-05 |   0.001 | 0.000982585315792582 | None        |         0.00 |         |
| cache_size  | float  | 200.0     |   100.0 |   400.0 |    375.6371648003268 | None        |         0.00 |         |
| break_ties  | factor | 0         |     0.0 |     1.0 |                  0.0 | None        |         0.00 |         |
\end{verbatim}

\hypertarget{show-variable-importance}{%
\subsection{Show variable importance}\label{show-variable-importance}}

\begin{Shaded}
\begin{Highlighting}[]
\NormalTok{spot\_tuner.plot\_importance(threshold}\OperatorTok{=}\FloatTok{0.025}\NormalTok{, filename}\OperatorTok{=}\StringTok{"./figures/"} \OperatorTok{+}\NormalTok{ experiment\_name}\OperatorTok{+}\StringTok{"\_importance.png"}\NormalTok{)}
\end{Highlighting}
\end{Shaded}

\begin{figure}[H]

{\centering \includegraphics{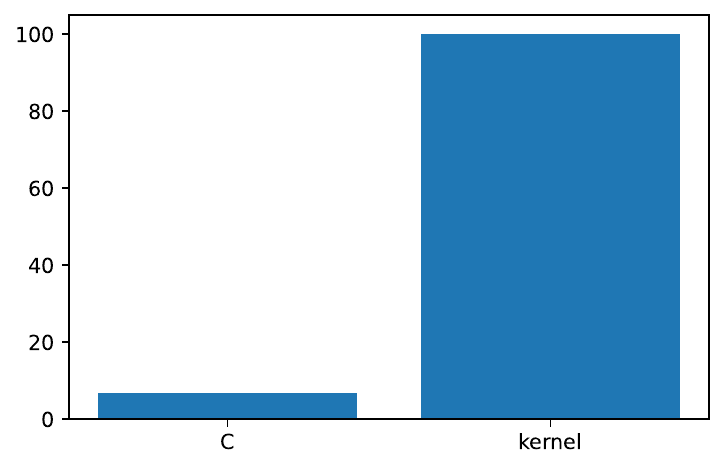}

}

\caption{Variable importance plot, threshold 0.025.}

\end{figure}

\hypertarget{get-default-hyperparameters}{%
\subsection{Get Default
Hyperparameters}\label{get-default-hyperparameters}}

\begin{Shaded}
\begin{Highlighting}[]
\ImportTok{from}\NormalTok{ spotPython.hyperparameters.values }\ImportTok{import}\NormalTok{ get\_default\_values, transform\_hyper\_parameter\_values}
\NormalTok{values\_default }\OperatorTok{=}\NormalTok{ get\_default\_values(fun\_control)}
\NormalTok{values\_default }\OperatorTok{=}\NormalTok{ transform\_hyper\_parameter\_values(fun\_control}\OperatorTok{=}\NormalTok{fun\_control, hyper\_parameter\_values}\OperatorTok{=}\NormalTok{values\_default)}
\NormalTok{values\_default}
\end{Highlighting}
\end{Shaded}

\begin{verbatim}
{'C': 1.0,
 'kernel': 'rbf',
 'degree': 3,
 'gamma': 'scale',
 'coef0': 0.0,
 'shrinking': 0,
 'probability': 0,
 'tol': 0.001,
 'cache_size': 200.0,
 'break_ties': 0}
\end{verbatim}

\begin{Shaded}
\begin{Highlighting}[]
\ImportTok{from}\NormalTok{ sklearn.pipeline }\ImportTok{import}\NormalTok{ make\_pipeline}
\NormalTok{model\_default }\OperatorTok{=}\NormalTok{ make\_pipeline(fun\_control[}\StringTok{"prep\_model"}\NormalTok{], fun\_control[}\StringTok{"core\_model"}\NormalTok{](}\OperatorTok{**}\NormalTok{values\_default))}
\NormalTok{model\_default}
\end{Highlighting}
\end{Shaded}

\begin{verbatim}
Pipeline(steps=[('standardscaler', StandardScaler()),
                ('svc',
                 SVC(break_ties=0, cache_size=200.0, probability=0,
                     shrinking=0))])
\end{verbatim}

\hypertarget{get-spot-results}{%
\subsection{Get SPOT Results}\label{get-spot-results}}

\begin{Shaded}
\begin{Highlighting}[]
\NormalTok{X }\OperatorTok{=}\NormalTok{ spot\_tuner.to\_all\_dim(spot\_tuner.min\_X.reshape(}\DecValTok{1}\NormalTok{,}\OperatorTok{{-}}\DecValTok{1}\NormalTok{))}
\BuiltInTok{print}\NormalTok{(X)}
\end{Highlighting}
\end{Shaded}

\begin{verbatim}
[[2.39447166e+00 1.00000000e+00 3.00000000e+00 0.00000000e+00
  0.00000000e+00 0.00000000e+00 0.00000000e+00 9.82585316e-04
  3.75637165e+02 0.00000000e+00]]
\end{verbatim}

\begin{Shaded}
\begin{Highlighting}[]
\ImportTok{from}\NormalTok{ spotPython.hyperparameters.values }\ImportTok{import}\NormalTok{ assign\_values, return\_conf\_list\_from\_var\_dict}
\NormalTok{v\_dict }\OperatorTok{=}\NormalTok{ assign\_values(X, fun\_control[}\StringTok{"var\_name"}\NormalTok{])}
\NormalTok{return\_conf\_list\_from\_var\_dict(var\_dict}\OperatorTok{=}\NormalTok{v\_dict, fun\_control}\OperatorTok{=}\NormalTok{fun\_control)}
\end{Highlighting}
\end{Shaded}

\begin{verbatim}
[{'C': 2.394471655384338,
  'kernel': 'rbf',
  'degree': 3,
  'gamma': 'scale',
  'coef0': 0.0,
  'shrinking': 0,
  'probability': 0,
  'tol': 0.000982585315792582,
  'cache_size': 375.6371648003268,
  'break_ties': 0}]
\end{verbatim}

\begin{Shaded}
\begin{Highlighting}[]
\ImportTok{from}\NormalTok{ spotPython.hyperparameters.values }\ImportTok{import}\NormalTok{ get\_one\_sklearn\_model\_from\_X}
\NormalTok{model\_spot }\OperatorTok{=}\NormalTok{ get\_one\_sklearn\_model\_from\_X(X, fun\_control)}
\NormalTok{model\_spot}
\end{Highlighting}
\end{Shaded}

\begin{verbatim}
Pipeline(steps=[('standardscaler', StandardScaler()),
                ('svc',
                 SVC(C=2.394471655384338, break_ties=0,
                     cache_size=375.6371648003268, probability=0, shrinking=0,
                     tol=0.000982585315792582))])
\end{verbatim}

\hypertarget{plot-compare-predictions}{%
\subsection{Plot: Compare Predictions}\label{plot-compare-predictions}}

\begin{Shaded}
\begin{Highlighting}[]
\ImportTok{from}\NormalTok{ spotPython.plot.validation }\ImportTok{import}\NormalTok{ plot\_roc}
\NormalTok{plot\_roc([model\_default, model\_spot], fun\_control, model\_names}\OperatorTok{=}\NormalTok{[}\StringTok{"Default"}\NormalTok{, }\StringTok{"Spot"}\NormalTok{])}
\end{Highlighting}
\end{Shaded}

\begin{figure}[H]

{\centering \includegraphics{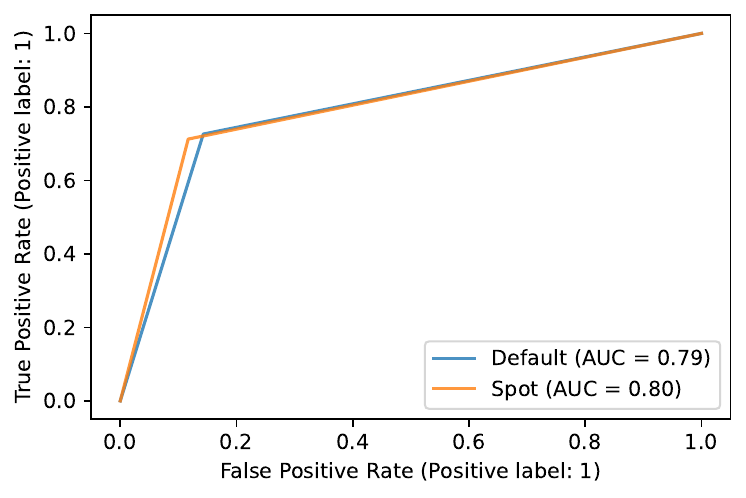}

}

\end{figure}

\begin{Shaded}
\begin{Highlighting}[]
\ImportTok{from}\NormalTok{ spotPython.plot.validation }\ImportTok{import}\NormalTok{ plot\_confusion\_matrix}
\NormalTok{plot\_confusion\_matrix(model\_default, fun\_control, title }\OperatorTok{=} \StringTok{"Default"}\NormalTok{)}
\end{Highlighting}
\end{Shaded}

\begin{figure}[H]

{\centering \includegraphics{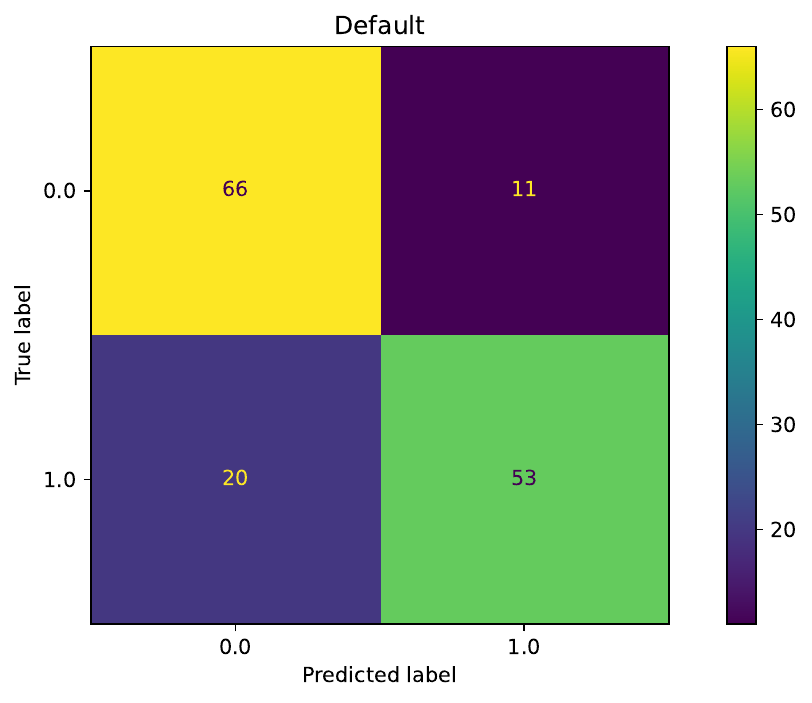}

}

\end{figure}

\begin{Shaded}
\begin{Highlighting}[]
\NormalTok{plot\_confusion\_matrix(model\_spot, fun\_control, title}\OperatorTok{=}\StringTok{"SPOT"}\NormalTok{)}
\end{Highlighting}
\end{Shaded}

\begin{figure}[H]

{\centering \includegraphics{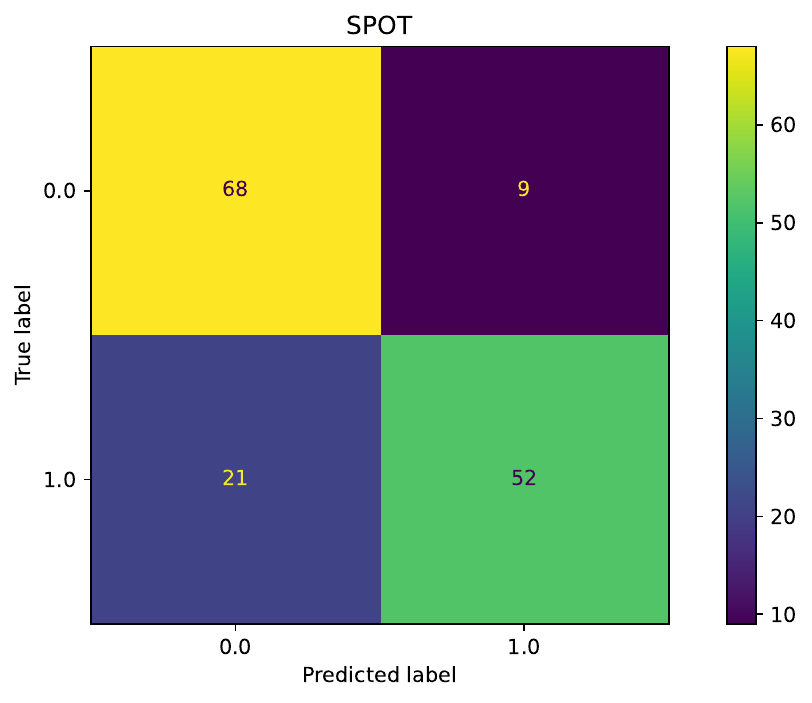}

}

\end{figure}

\begin{Shaded}
\begin{Highlighting}[]
\BuiltInTok{min}\NormalTok{(spot\_tuner.y), }\BuiltInTok{max}\NormalTok{(spot\_tuner.y)}
\end{Highlighting}
\end{Shaded}

\begin{verbatim}
(5.734217584632275, 7.782152436286657)
\end{verbatim}

\hypertarget{detailed-hyperparameter-plots}{%
\subsection{Detailed Hyperparameter
Plots}\label{detailed-hyperparameter-plots}}

\begin{Shaded}
\begin{Highlighting}[]
\NormalTok{filename }\OperatorTok{=} \StringTok{"./figures/"} \OperatorTok{+}\NormalTok{ experiment\_name}
\NormalTok{spot\_tuner.plot\_important\_hyperparameter\_contour(filename}\OperatorTok{=}\NormalTok{filename)}
\end{Highlighting}
\end{Shaded}

\begin{verbatim}
C:  6.78742297418671
kernel:  100.0
\end{verbatim}

\begin{figure}[H]

{\centering \includegraphics{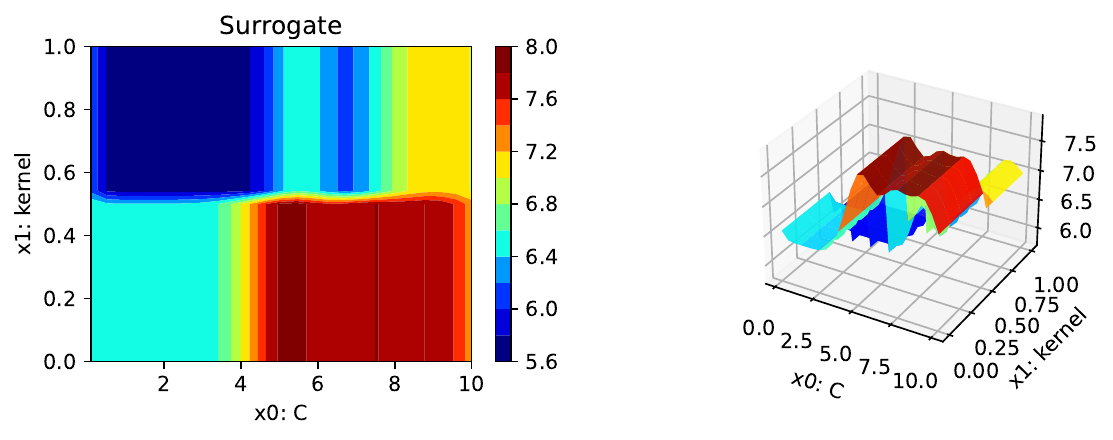}

}

\end{figure}

\hypertarget{parallel-coordinates-plot}{%
\subsection{Parallel Coordinates Plot}\label{parallel-coordinates-plot}}

\begin{Shaded}
\begin{Highlighting}[]
\NormalTok{spot\_tuner.parallel\_plot()}
\end{Highlighting}
\end{Shaded}

\begin{verbatim}
Unable to display output for mime type(s): text/html
\end{verbatim}

\begin{verbatim}
Unable to display output for mime type(s): text/html
\end{verbatim}

\hypertarget{plot-all-combinations-of-hyperparameters}{%
\subsection{Plot all Combinations of
Hyperparameters}\label{plot-all-combinations-of-hyperparameters}}

\begin{itemize}
\tightlist
\item
  Warning: this may take a while.
\end{itemize}

\begin{Shaded}
\begin{Highlighting}[]
\NormalTok{PLOT\_ALL }\OperatorTok{=} \VariableTok{False}
\ControlFlowTok{if}\NormalTok{ PLOT\_ALL:}
\NormalTok{    n }\OperatorTok{=}\NormalTok{ spot\_tuner.k}
    \ControlFlowTok{for}\NormalTok{ i }\KeywordTok{in} \BuiltInTok{range}\NormalTok{(n}\OperatorTok{{-}}\DecValTok{1}\NormalTok{):}
        \ControlFlowTok{for}\NormalTok{ j }\KeywordTok{in} \BuiltInTok{range}\NormalTok{(i}\OperatorTok{+}\DecValTok{1}\NormalTok{, n):}
\NormalTok{            spot\_tuner.plot\_contour(i}\OperatorTok{=}\NormalTok{i, j}\OperatorTok{=}\NormalTok{j, min\_z}\OperatorTok{=}\NormalTok{min\_z, max\_z }\OperatorTok{=}\NormalTok{ max\_z)}
\end{Highlighting}
\end{Shaded}

\hypertarget{sec-river-hpt}{%
\chapter{\texorpdfstring{\texttt{river} Hyperparameter Tuning: Hoeffding
Adaptive Tree Regressor with Friedman Drift
Data}{river Hyperparameter Tuning: Hoeffding Adaptive Tree Regressor with Friedman Drift Data}}\label{sec-river-hpt}}

This chapter demonstrates hyperparameter tuning for \texttt{river}'s
\texttt{Hoeffding\ Adaptive\ Tree\ Regressor} with the Friedman drift
data set
\href{https://riverml.xyz/0.18.0/api/datasets/synth/FriedmanDrift/}{{[}SOURCE{]}}.
The \texttt{Hoeffding\ Adaptive\ Tree\ Regressor} is a decision tree
that uses the Hoeffding bound to limit the number of splits evaluated at
each node. The \texttt{Hoeffding\ Adaptive\ Tree\ Regressor} is a
regression tree, i.e., it predicts a real value for each sample. The
\texttt{Hoeffding\ Adaptive\ Tree\ Regressor} is a drift aware model,
i.e., it can handle concept drifts.

\hypertarget{sec-setup-13}{%
\section{Setup}\label{sec-setup-13}}

Before we consider the detailed experimental setup, we select the
parameters that affect run time, initial design size, size of the data
set, and the experiment name.

\begin{itemize}
\tightlist
\item
  \texttt{MAX\_TIME}: The maximum run time in seconds for the
  hyperparameter tuning process.
\item
  \texttt{INIT\_SIZE}: The initial design size for the hyperparameter
  tuning process.
\item
  \texttt{PREFIX}: The prefix for the experiment name.
\item
  \texttt{K}: The factor that determines the number of samples in the
  data set.
\end{itemize}

\begin{tcolorbox}[enhanced jigsaw, left=2mm, title=\textcolor{quarto-callout-caution-color}{\faFire}\hspace{0.5em}{Caution: Run time and initial design size should be increased for real
experiments}, bottomrule=.15mm, titlerule=0mm, breakable, rightrule=.15mm, toprule=.15mm, coltitle=black, colbacktitle=quarto-callout-caution-color!10!white, leftrule=.75mm, arc=.35mm, colframe=quarto-callout-caution-color-frame, bottomtitle=1mm, colback=white, opacitybacktitle=0.6, toptitle=1mm, opacityback=0]

\begin{itemize}
\tightlist
\item
  \texttt{MAX\_TIME} is set to one minute for demonstration purposes.
  For real experiments, this should be increased to at least 1 hour.
\item
  \texttt{INIT\_SIZE} is set to 5 for demonstration purposes. For real
  experiments, this should be increased to at least 10.
\item
  \texttt{K} is the multiplier for the number of samples. If it is set
  to 1, then \texttt{100\_000}samples are taken. It is set to 0.1 for
  demonstration purposes. For real experiments, this should be increased
  to at least 1.
\end{itemize}

\end{tcolorbox}

\begin{Shaded}
\begin{Highlighting}[]
\NormalTok{MAX\_TIME }\OperatorTok{=} \DecValTok{1}
\NormalTok{INIT\_SIZE }\OperatorTok{=} \DecValTok{5}
\NormalTok{PREFIX}\OperatorTok{=}\StringTok{"10{-}river"}
\NormalTok{K }\OperatorTok{=} \FloatTok{.1}
\end{Highlighting}
\end{Shaded}

\begin{Shaded}
\begin{Highlighting}[]
\ImportTok{import}\NormalTok{ os}
\ImportTok{from}\NormalTok{ spotPython.utils.}\BuiltInTok{file} \ImportTok{import}\NormalTok{ get\_experiment\_name}
\NormalTok{experiment\_name }\OperatorTok{=}\NormalTok{ get\_experiment\_name(prefix}\OperatorTok{=}\NormalTok{PREFIX)}
\BuiltInTok{print}\NormalTok{(experiment\_name)}
\end{Highlighting}
\end{Shaded}

\begin{verbatim}
10-river_bartz09_2023-07-17_08-56-52
\end{verbatim}

\begin{itemize}
\tightlist
\item
  This notebook exemplifies hyperparameter tuning with SPOT (spotPython
  and spotRiver).
\item
  The hyperparameter software SPOT was developed in R (statistical
  programming language), see Open Access book ``Hyperparameter Tuning
  for Machine and Deep Learning with R - A Practical Guide'', available
  here: \url{https://link.springer.com/book/10.1007/978-981-19-5170-1}.
\item
  This notebook demonstrates hyperparameter tuning for \texttt{river}.
  It is based on the notebook ``Incremental decision trees in river: the
  Hoeffding Tree case'', see:
  \url{https://riverml.xyz/0.15.0/recipes/on-hoeffding-trees/\#42-regression-tree-splitters}.
\item
  Here we will use the river \texttt{HTR} and \texttt{HATR} functions as
  in ``Incremental decision trees in river: the Hoeffding Tree case'',
  see:
  \url{https://riverml.xyz/0.15.0/recipes/on-hoeffding-trees/\#42-regression-tree-splitters}.
\end{itemize}

\hypertarget{initialization-of-the-fun_control-dictionary}{%
\section{\texorpdfstring{Initialization of the \texttt{fun\_control}
Dictionary}{Initialization of the fun\_control Dictionary}}\label{initialization-of-the-fun_control-dictionary}}

\texttt{spotPython} supports the visualization of the hyperparameter
tuning process with TensorBoard. The following example shows how to use
TensorBoard with \texttt{spotPython}.

First, we define an ``experiment name'' to identify the hyperparameter
tuning process. The experiment name is also used to create a directory
for the TensorBoard files.

\begin{Shaded}
\begin{Highlighting}[]
\ImportTok{from}\NormalTok{ spotPython.utils.init }\ImportTok{import}\NormalTok{ fun\_control\_init}
\ImportTok{from}\NormalTok{ spotPython.utils.}\BuiltInTok{file} \ImportTok{import}\NormalTok{ get\_spot\_tensorboard\_path}

\NormalTok{experiment\_name }\OperatorTok{=}\NormalTok{ get\_experiment\_name(prefix}\OperatorTok{=}\NormalTok{PREFIX)}
\NormalTok{fun\_control }\OperatorTok{=}\NormalTok{ fun\_control\_init(}
\NormalTok{    spot\_tensorboard\_path}\OperatorTok{=}\NormalTok{get\_spot\_tensorboard\_path(experiment\_name),}
\NormalTok{    TENSORBOARD\_CLEAN}\OperatorTok{=}\VariableTok{True}\NormalTok{)}
\end{Highlighting}
\end{Shaded}

\begin{tcolorbox}[enhanced jigsaw, left=2mm, title=\textcolor{quarto-callout-tip-color}{\faLightbulb}\hspace{0.5em}{Tip: TensorBoard}, bottomrule=.15mm, titlerule=0mm, breakable, rightrule=.15mm, toprule=.15mm, coltitle=black, colbacktitle=quarto-callout-tip-color!10!white, leftrule=.75mm, arc=.35mm, colframe=quarto-callout-tip-color-frame, bottomtitle=1mm, colback=white, opacitybacktitle=0.6, toptitle=1mm, opacityback=0]

\begin{itemize}
\tightlist
\item
  Since the \texttt{spot\_tensorboard\_path} argument is not
  \texttt{None}, which is the default, \texttt{spotPython} will log the
  optimization process in the TensorBoard folder.
\item
  Section~\ref{sec-tensorboard-10} describes how to start TensorBoard
  and access the TensorBoard dashboard.
\item
  The \texttt{TENSORBOARD\_CLEAN} argument is set to \texttt{True} to
  archive the TensorBoard folder if it already exists. This is useful if
  you want to start a hyperparameter tuning process from scratch. If you
  want to continue a hyperparameter tuning process, set
  \texttt{TENSORBOARD\_CLEAN} to \texttt{False}. Then the TensorBoard
  folder will not be archived and the old and new TensorBoard files will
  shown in the TensorBoard dashboard.
\end{itemize}

\end{tcolorbox}

\hypertarget{load-data-the-friedman-drift-data}{%
\section{Load Data: The Friedman Drift
Data}\label{load-data-the-friedman-drift-data}}

We will use the Friedman synthetic dataset with concept drifts
\href{https://riverml.xyz/0.18.0/api/datasets/synth/FriedmanDrift/}{{[}SOURCE{]}}.
Each observation is composed of ten features. Each feature value is
sampled uniformly in {[}0, 1{]}. Only the first five features are
relevant. The target is defined by different functions depending on the
type of the drift. Global Recurring Abrupt drift will be used, i.e., the
concept drift appears over the whole instance space. There are two
points of concept drift. At the second point of drift the old concept
reoccurs.

The following parameters are used to generate and handle the data set:

\begin{itemize}
\tightlist
\item
  horizon: The prediction horizon in hours.
\item
  n\_samples: The number of samples in the data set.
\item
  p\_1: The position of the first concept drift.
\item
  p\_2: The position of the second concept drift.
\item
  position: The position of the concept drifts.
\item
  n\_train: The number of samples used for training.
\end{itemize}

\begin{Shaded}
\begin{Highlighting}[]
\NormalTok{horizon }\OperatorTok{=} \DecValTok{7}\OperatorTok{*}\DecValTok{24}
\NormalTok{n\_samples }\OperatorTok{=} \BuiltInTok{int}\NormalTok{(K}\OperatorTok{*}\DecValTok{100\_000}\NormalTok{)}
\NormalTok{p\_1 }\OperatorTok{=} \BuiltInTok{int}\NormalTok{(K}\OperatorTok{*}\DecValTok{25\_000}\NormalTok{)}
\NormalTok{p\_2 }\OperatorTok{=} \BuiltInTok{int}\NormalTok{(K}\OperatorTok{*}\DecValTok{50\_000}\NormalTok{)}
\NormalTok{position}\OperatorTok{=}\NormalTok{(p\_1, p\_2)}
\NormalTok{n\_train }\OperatorTok{=} \DecValTok{1\_000}
\end{Highlighting}
\end{Shaded}

\begin{Shaded}
\begin{Highlighting}[]
\ImportTok{from}\NormalTok{ river.datasets }\ImportTok{import}\NormalTok{ synth}
\ImportTok{import}\NormalTok{ pandas }\ImportTok{as}\NormalTok{ pd}
\NormalTok{dataset }\OperatorTok{=}\NormalTok{ synth.FriedmanDrift(}
\NormalTok{   drift\_type}\OperatorTok{=}\StringTok{\textquotesingle{}gra\textquotesingle{}}\NormalTok{,}
\NormalTok{   position}\OperatorTok{=}\NormalTok{position,}
\NormalTok{   seed}\OperatorTok{=}\DecValTok{123}
\NormalTok{)}
\end{Highlighting}
\end{Shaded}

\begin{itemize}
\tightlist
\item
  We will use \texttt{spotRiver}'s \texttt{convert\_to\_df} function
  \href{https://github.com/sequential-parameter-optimization/spotRiver/blob/main/src/spotRiver/utils/data_conversion.py}{{[}SOURCE{]}}
  to convert the \texttt{river} data set to a \texttt{pandas} data
  frame.
\end{itemize}

\begin{Shaded}
\begin{Highlighting}[]
\ImportTok{from}\NormalTok{ spotRiver.utils.data\_conversion }\ImportTok{import}\NormalTok{ convert\_to\_df}
\NormalTok{target\_column }\OperatorTok{=} \StringTok{"y"}
\NormalTok{df }\OperatorTok{=}\NormalTok{ convert\_to\_df(dataset, target\_column}\OperatorTok{=}\NormalTok{target\_column, n\_total}\OperatorTok{=}\NormalTok{n\_samples)}
\end{Highlighting}
\end{Shaded}

\begin{itemize}
\tightlist
\item
  Add column names x1 until x10 to the first 10 columns of the dataframe
  and the column name y to the last column of the dataframe.
\item
  Then split the data frame into a training and test data set. The train
  and test data sets are stored in the \texttt{fun\_control} dictionary.
\end{itemize}

\begin{Shaded}
\begin{Highlighting}[]
\NormalTok{df.columns }\OperatorTok{=}\NormalTok{ [}\SpecialStringTok{f"x}\SpecialCharTok{\{}\NormalTok{i}\SpecialCharTok{\}}\SpecialStringTok{"} \ControlFlowTok{for}\NormalTok{ i }\KeywordTok{in} \BuiltInTok{range}\NormalTok{(}\DecValTok{1}\NormalTok{, }\DecValTok{11}\NormalTok{)] }\OperatorTok{+}\NormalTok{ [}\StringTok{"y"}\NormalTok{]}
\NormalTok{fun\_control.update(\{}\StringTok{"train"}\NormalTok{:  df[:n\_train],}
                    \StringTok{"test"}\NormalTok{:  df[n\_train:],}
                    \StringTok{"n\_samples"}\NormalTok{: n\_samples,}
                    \StringTok{"target\_column"}\NormalTok{: target\_column\})}
\end{Highlighting}
\end{Shaded}

\hypertarget{specification-of-the-preprocessing-model}{%
\section{Specification of the Preprocessing
Model}\label{specification-of-the-preprocessing-model}}

\begin{itemize}
\tightlist
\item
  We use the \texttt{StandardScaler}
  \href{https://riverml.xyz/dev/api/preprocessing/StandardScaler/}{{[}SOURCE{]}}
  from \texttt{river} as the preprocessing model. The
  \texttt{StandardScaler} is used to standardize the data set, i.e., it
  has zero mean and unit variance.
\end{itemize}

\begin{Shaded}
\begin{Highlighting}[]
\ImportTok{from}\NormalTok{ river }\ImportTok{import}\NormalTok{ preprocessing}
\NormalTok{prep\_model }\OperatorTok{=}\NormalTok{ preprocessing.StandardScaler()}
\NormalTok{fun\_control.update(\{}\StringTok{"prep\_model"}\NormalTok{: prep\_model\})}
\end{Highlighting}
\end{Shaded}

\hypertarget{selectselect-model-algorithm-and-core_model_hyper_dict}{%
\section{\texorpdfstring{SelectSelect Model (\texttt{algorithm}) and
\texttt{core\_model\_hyper\_dict}}{SelectSelect Model (algorithm) and core\_model\_hyper\_dict}}\label{selectselect-model-algorithm-and-core_model_hyper_dict}}

\texttt{spotPython} hyperparameter tuning approach uses two components:

\begin{enumerate}
\def\labelenumi{\arabic{enumi}.}
\tightlist
\item
  a model (class) and
\item
  an associated hyperparameter dictionary.
\end{enumerate}

Here, the \texttt{river} model class
\texttt{HoeffdingAdaptiveTreeRegressor}
\href{https://riverml.xyz/dev/api/tree/HoeffdingAdaptiveTreeRegressor/}{{[}SOURCE{]}}
is selected.

The corresponding hyperparameters are loaded from the associated
dictionary, which is stored as a JSON file
\href{https://github.com/sequential-parameter-optimization/spotRiver/blob/main/src/spotRiver/data/river_hyper_dict.json}{{[}SOURCE{]}}.
The JSON file contains hyperparameter type information, names, and
bounds.

The method \texttt{add\_core\_model\_to\_fun\_control} adds the model
and the hyperparameter dictionary to the \texttt{fun\_control}
dictionary.

Alternatively, you can load a local hyper\_dict. Simply set
\texttt{river\_hyper\_dict.json} as the filename. If \texttt{filename}is
set to \texttt{None}, which is the default, the hyper\_dict
\href{https://github.com/sequential-parameter-optimization/spotRiver/blob/main/src/spotRiver/data/river_hyper_dict.json}{{[}SOURCE{]}}
is loaded from the \texttt{spotRiver} package.

\begin{Shaded}
\begin{Highlighting}[]
\ImportTok{from}\NormalTok{ river.tree }\ImportTok{import}\NormalTok{ HoeffdingAdaptiveTreeRegressor}
\ImportTok{from}\NormalTok{ spotRiver.data.river\_hyper\_dict }\ImportTok{import}\NormalTok{ RiverHyperDict}
\ImportTok{from}\NormalTok{ spotPython.hyperparameters.values }\ImportTok{import}\NormalTok{ add\_core\_model\_to\_fun\_control}
\NormalTok{add\_core\_model\_to\_fun\_control(core\_model}\OperatorTok{=}\NormalTok{HoeffdingAdaptiveTreeRegressor,}
\NormalTok{                              fun\_control}\OperatorTok{=}\NormalTok{fun\_control,}
\NormalTok{                              hyper\_dict}\OperatorTok{=}\NormalTok{RiverHyperDict,}
\NormalTok{                              filename}\OperatorTok{=}\VariableTok{None}\NormalTok{)}
\end{Highlighting}
\end{Shaded}

\hypertarget{modify-hyper_dict-hyperparameters-for-the-selected-algorithm-aka-core_model}{%
\section{\texorpdfstring{Modify \texttt{hyper\_dict} Hyperparameters for
the Selected Algorithm aka
\texttt{core\_model}}{Modify hyper\_dict Hyperparameters for the Selected Algorithm aka core\_model}}\label{modify-hyper_dict-hyperparameters-for-the-selected-algorithm-aka-core_model}}

After the \texttt{core\_model} and the \texttt{core\_model\_hyper\_dict}
are added to the \texttt{fun\_control} dictionary, the hyperparameter
tuning can be started. However, in some settings, the user wants to
modify the hyperparameters of the \texttt{core\_model\_hyper\_dict}.
This can be done with the \texttt{modify\_hyper\_parameter\_bounds} and
\texttt{modify\_hyper\_parameter\_levels} functions
\href{https://github.com/sequential-parameter-optimization/spotPython/blob/main/src/spotPython/hyperparameters/values.py}{{[}SOURCE{]}}.

The following code shows how hyperparameter of type numeric and integer
(boolean) can be modified. The \texttt{modify\_hyper\_parameter\_bounds}
function is used to modify the bounds of the hyperparameter
\texttt{delta} and \texttt{merit\_preprune}. Similar option exists for
the \texttt{modify\_hyper\_parameter\_levels} function to modify the
levels of categorical hyperparameters.

\begin{Shaded}
\begin{Highlighting}[]
\ImportTok{from}\NormalTok{ spotPython.hyperparameters.values }\ImportTok{import}\NormalTok{ modify\_hyper\_parameter\_bounds}
\NormalTok{modify\_hyper\_parameter\_bounds(fun\_control, }\StringTok{"delta"}\NormalTok{, bounds}\OperatorTok{=}\NormalTok{[}\FloatTok{1e{-}10}\NormalTok{, }\FloatTok{1e{-}6}\NormalTok{])}
\NormalTok{modify\_hyper\_parameter\_bounds(fun\_control, }\StringTok{"merit\_preprune"}\NormalTok{, [}\DecValTok{0}\NormalTok{, }\DecValTok{0}\NormalTok{])}
\end{Highlighting}
\end{Shaded}

\begin{tcolorbox}[enhanced jigsaw, left=2mm, title=\textcolor{quarto-callout-note-color}{\faInfo}\hspace{0.5em}{Note: Active and Inactive Hyperparameters}, bottomrule=.15mm, titlerule=0mm, breakable, rightrule=.15mm, toprule=.15mm, coltitle=black, colbacktitle=quarto-callout-note-color!10!white, leftrule=.75mm, arc=.35mm, colframe=quarto-callout-note-color-frame, bottomtitle=1mm, colback=white, opacitybacktitle=0.6, toptitle=1mm, opacityback=0]

Hyperparameters can be excluded from the tuning procedure by selecting
identical values for the lower and upper bounds. For example, the
hyperparameter \texttt{merit\_preprune} is excluded from the tuning
procedure by setting the bounds to \texttt{{[}0,\ 0{]}}.

\end{tcolorbox}

\texttt{spotPython}'s method \texttt{gen\_design\_table} summarizes the
experimental design that is used for the hyperparameter tuning:

\begin{Shaded}
\begin{Highlighting}[]
\ImportTok{from}\NormalTok{ spotPython.utils.eda }\ImportTok{import}\NormalTok{ gen\_design\_table}
\BuiltInTok{print}\NormalTok{(gen\_design\_table(fun\_control))}
\end{Highlighting}
\end{Shaded}

\begin{verbatim}
| name                   | type   | default          |      lower |    upper | transform             |
|------------------------|--------|------------------|------------|----------|-----------------------|
| grace_period           | int    | 200              |     10     | 1000     | None                  |
| max_depth              | int    | 20               |      2     |   20     | transform_power_2_int |
| delta                  | float  | 1e-07            |      1e-10 |    1e-06 | None                  |
| tau                    | float  | 0.05             |      0.01  |    0.1   | None                  |
| leaf_prediction        | factor | mean             |      0     |    2     | None                  |
| leaf_model             | factor | LinearRegression |      0     |    2     | None                  |
| model_selector_decay   | float  | 0.95             |      0.9   |    0.99  | None                  |
| splitter               | factor | EBSTSplitter     |      0     |    2     | None                  |
| min_samples_split      | int    | 5                |      2     |   10     | None                  |
| bootstrap_sampling     | factor | 0                |      0     |    1     | None                  |
| drift_window_threshold | int    | 300              |    100     |  500     | None                  |
| switch_significance    | float  | 0.05             |      0.01  |    0.1   | None                  |
| binary_split           | factor | 0                |      0     |    1     | None                  |
| max_size               | float  | 500.0            |    100     | 1000     | None                  |
| memory_estimate_period | int    | 1000000          | 100000     |    1e+06 | None                  |
| stop_mem_management    | factor | 0                |      0     |    1     | None                  |
| remove_poor_attrs      | factor | 0                |      0     |    1     | None                  |
| merit_preprune         | factor | 0                |      0     |    0     | None                  |
\end{verbatim}

\hypertarget{selection-of-the-objective-function}{%
\section{Selection of the Objective
Function}\label{selection-of-the-objective-function}}

The \texttt{metric\_sklearn} is used for the sklearn based evaluation
via \texttt{eval\_oml\_horizon}
\href{https://github.com/sequential-parameter-optimization/spotRiver/blob/main/src/spotRiver/evaluation/eval_bml.py}{{[}SOURCE{]}}.
Here we use the \texttt{mean\_absolute\_error}
\href{https://scikit-learn.org/stable/modules/generated/sklearn.metrics.mean_absolute_error.html}{{[}SOURCE{]}}
as the objective function.

\begin{tcolorbox}[enhanced jigsaw, left=2mm, title=\textcolor{quarto-callout-note-color}{\faInfo}\hspace{0.5em}{Note: Additional metrics}, bottomrule=.15mm, titlerule=0mm, breakable, rightrule=.15mm, toprule=.15mm, coltitle=black, colbacktitle=quarto-callout-note-color!10!white, leftrule=.75mm, arc=.35mm, colframe=quarto-callout-note-color-frame, bottomtitle=1mm, colback=white, opacitybacktitle=0.6, toptitle=1mm, opacityback=0]

\texttt{spotRiver} also supports additional metrics. For example, the
\texttt{metric\_river} is used for the river based evaluation via
\texttt{eval\_oml\_iter\_progressive}
\href{https://github.com/sequential-parameter-optimization/spotRiver/blob/main/src/spotRiver/evaluation/eval_oml.py}{{[}SOURCE{]}}.
The \texttt{metric\_river} is implemented to simulate the behaviour of
the ``original'' \texttt{river} metrics.

\end{tcolorbox}

\texttt{spotRiver} provides information about the model' s score
(metric), memory, and time. The hyperparamter tuner requires a single
objective. Therefore, a weighted sum of the metric, memory, and time is
computed. The weights are defined in the \texttt{weights} array.

\begin{tcolorbox}[enhanced jigsaw, left=2mm, title=\textcolor{quarto-callout-note-color}{\faInfo}\hspace{0.5em}{Note: Weights}, bottomrule=.15mm, titlerule=0mm, breakable, rightrule=.15mm, toprule=.15mm, coltitle=black, colbacktitle=quarto-callout-note-color!10!white, leftrule=.75mm, arc=.35mm, colframe=quarto-callout-note-color-frame, bottomtitle=1mm, colback=white, opacitybacktitle=0.6, toptitle=1mm, opacityback=0]

The \texttt{weights} provide a flexible way to define specific
requirements, e.g., if the memory is more important than the time, the
weight for the memory can be increased.

\end{tcolorbox}

The \texttt{oml\_grace\_period} defines the number of observations that
are used for the initial training of the model. The \texttt{step}
defines the iteration number at which to yield results. This only takes
into account the predictions, and not the training steps. The
\texttt{weight\_coeff} defines a multiplier for the results: results are
multiplied by (step/n\_steps)**weight\_coeff, where n\_steps is the
total number of iterations. Results from the beginning have a lower
weight than results from the end if weight\_coeff \textgreater{} 1. If
weight\_coeff == 0, all results have equal weight. Note, that the
\texttt{weight\_coeff} is only used internally for the tuner and does
not affect the results that are used for the evaluation or comparisons.

\begin{Shaded}
\begin{Highlighting}[]
\ImportTok{import}\NormalTok{ numpy }\ImportTok{as}\NormalTok{ np}
\ImportTok{from}\NormalTok{ sklearn.metrics }\ImportTok{import}\NormalTok{ mean\_absolute\_error}

\NormalTok{weights }\OperatorTok{=}\NormalTok{ np.array([}\DecValTok{1}\NormalTok{, }\DecValTok{1}\OperatorTok{/}\DecValTok{1000}\NormalTok{, }\DecValTok{1}\OperatorTok{/}\DecValTok{1000}\NormalTok{])}\OperatorTok{*}\FloatTok{10\_000.0}
\NormalTok{oml\_grace\_period }\OperatorTok{=} \DecValTok{2}
\NormalTok{step }\OperatorTok{=} \DecValTok{100}
\NormalTok{weight\_coeff }\OperatorTok{=} \FloatTok{1.0}

\NormalTok{fun\_control.update(\{}
               \StringTok{"horizon"}\NormalTok{: horizon,}
               \StringTok{"oml\_grace\_period"}\NormalTok{: oml\_grace\_period,}
               \StringTok{"weights"}\NormalTok{: weights,}
               \StringTok{"step"}\NormalTok{: step,}
               \StringTok{"weight\_coeff"}\NormalTok{: weight\_coeff,}
               \StringTok{"metric\_sklearn"}\NormalTok{: mean\_absolute\_error}
\NormalTok{               \})}
\end{Highlighting}
\end{Shaded}

\hypertarget{calling-the-spot-function}{%
\section{Calling the SPOT Function}\label{calling-the-spot-function}}

\hypertarget{prepare-the-spot-parameters}{%
\subsection{Prepare the SPOT
Parameters}\label{prepare-the-spot-parameters}}

The hyperparameter tuning configuration is stored in the
\texttt{fun\_control} dictionary. Since \texttt{Spot} can be used as an
optimization algorithm with a similar interface as optimization
algorithms from \texttt{scipy.optimize}
\href{https://docs.scipy.org/doc/scipy/reference/optimize.html\#module-scipy.optimize}{{[}LINK{]}},
the bounds and variable types have to be specified explicitely. The
\texttt{get\_var\_type}, \texttt{get\_var\_name}, and
\texttt{get\_bound\_values} functions
\href{https://github.com/sequential-parameter-optimization/spotPython/blob/main/src/spotPython/hyperparameters/values.py}{{[}SOURCE{]}}
implement the required functionality.

\begin{itemize}
\tightlist
\item
  Get types and variable names as well as lower and upper bounds for the
  hyperparameters, so that they can be passed to the \texttt{Spot}
  function.
\end{itemize}

\begin{Shaded}
\begin{Highlighting}[]
\ImportTok{from}\NormalTok{ spotPython.hyperparameters.values }\ImportTok{import}\NormalTok{ (}
\NormalTok{    get\_var\_type,}
\NormalTok{    get\_var\_name,}
\NormalTok{    get\_bound\_values}
\NormalTok{    )}
\NormalTok{var\_type }\OperatorTok{=}\NormalTok{ get\_var\_type(fun\_control)}
\NormalTok{var\_name }\OperatorTok{=}\NormalTok{ get\_var\_name(fun\_control)}
\NormalTok{lower }\OperatorTok{=}\NormalTok{ get\_bound\_values(fun\_control, }\StringTok{"lower"}\NormalTok{)}
\NormalTok{upper }\OperatorTok{=}\NormalTok{ get\_bound\_values(fun\_control, }\StringTok{"upper"}\NormalTok{)}
\end{Highlighting}
\end{Shaded}

\hypertarget{sec-the-objective-function-13}{%
\subsection{The Objective
Function}\label{sec-the-objective-function-13}}

The objective function \texttt{fun\_oml\_horizon}
\href{https://github.com/sequential-parameter-optimization/spotRiver/blob/main/src/spotRiver/fun/hyperriver.py}{{[}SOURCE{]}}
is selected next.

\begin{Shaded}
\begin{Highlighting}[]
\ImportTok{from}\NormalTok{ spotRiver.fun.hyperriver }\ImportTok{import}\NormalTok{ HyperRiver}
\NormalTok{fun }\OperatorTok{=}\NormalTok{ HyperRiver().fun\_oml\_horizon}
\end{Highlighting}
\end{Shaded}

The following code snippet shows how to get the default hyperparameters
as an array, so that they can be passed to the \texttt{Spot} function.

\begin{Shaded}
\begin{Highlighting}[]
\ImportTok{from}\NormalTok{ spotPython.hyperparameters.values }\ImportTok{import}\NormalTok{ get\_default\_hyperparameters\_as\_array}
\NormalTok{X\_start }\OperatorTok{=}\NormalTok{ get\_default\_hyperparameters\_as\_array(fun\_control)}
\end{Highlighting}
\end{Shaded}

\hypertarget{run-the-spot-optimizer-1}{%
\subsection{\texorpdfstring{Run the \texttt{Spot}
Optimizer}{Run the Spot Optimizer}}\label{run-the-spot-optimizer-1}}

The class \texttt{Spot}
\href{https://github.com/sequential-parameter-optimization/spotPython/blob/main/src/spotPython/spot/spot.py}{{[}SOURCE{]}}
is the hyperparameter tuning workhorse. It is initialized with the
following parameters:

\begin{itemize}
\tightlist
\item
  \texttt{fun}: the objective function
\item
  \texttt{lower}: lower bounds of the hyperparameters
\item
  \texttt{upper}: upper bounds of the hyperparameters
\item
  \texttt{fun\_evals}: number of function evaluations
\item
  \texttt{max\_time}: maximum time in seconds
\item
  \texttt{tolerance\_x}: tolerance for the hyperparameters
\item
  \texttt{var\_type}: variable types of the hyperparameters
\item
  \texttt{var\_name}: variable names of the hyperparameters
\item
  \texttt{show\_progress}: show progress bar
\item
  \texttt{fun\_control}: dictionary with control parameters for the
  objective function
\item
  \texttt{design\_control}: dictionary with control parameters for the
  initial design
\item
  \texttt{surrogate\_control}: dictionary with control parameters for
  the surrogate model
\end{itemize}

\begin{tcolorbox}[enhanced jigsaw, left=2mm, title=\textcolor{quarto-callout-note-color}{\faInfo}\hspace{0.5em}{Note: Total run time}, bottomrule=.15mm, titlerule=0mm, breakable, rightrule=.15mm, toprule=.15mm, coltitle=black, colbacktitle=quarto-callout-note-color!10!white, leftrule=.75mm, arc=.35mm, colframe=quarto-callout-note-color-frame, bottomtitle=1mm, colback=white, opacitybacktitle=0.6, toptitle=1mm, opacityback=0]

The total run time may exceed the specified \texttt{max\_time}, because
the initial design (here: \texttt{init\_size} = INIT\_SIZE as specified
above) is always evaluated, even if this takes longer than
\texttt{max\_time}.

\end{tcolorbox}

\begin{Shaded}
\begin{Highlighting}[]
\ImportTok{from}\NormalTok{ spotPython.spot }\ImportTok{import}\NormalTok{ spot}
\ImportTok{from}\NormalTok{ math }\ImportTok{import}\NormalTok{ inf}
\NormalTok{spot\_tuner }\OperatorTok{=}\NormalTok{ spot.Spot(fun}\OperatorTok{=}\NormalTok{fun,}
\NormalTok{                   lower }\OperatorTok{=}\NormalTok{ lower,}
\NormalTok{                   upper }\OperatorTok{=}\NormalTok{ upper,}
\NormalTok{                   fun\_evals }\OperatorTok{=}\NormalTok{ inf,}
\NormalTok{                   max\_time }\OperatorTok{=}\NormalTok{ MAX\_TIME,}
\NormalTok{                   tolerance\_x }\OperatorTok{=}\NormalTok{ np.sqrt(np.spacing(}\DecValTok{1}\NormalTok{)),}
\NormalTok{                   var\_type }\OperatorTok{=}\NormalTok{ var\_type,}
\NormalTok{                   var\_name }\OperatorTok{=}\NormalTok{ var\_name,}
\NormalTok{                   show\_progress}\OperatorTok{=} \VariableTok{True}\NormalTok{,}
\NormalTok{                   fun\_control }\OperatorTok{=}\NormalTok{ fun\_control,}
\NormalTok{                   design\_control}\OperatorTok{=}\NormalTok{\{}\StringTok{"init\_size"}\NormalTok{: INIT\_SIZE\},}
\NormalTok{                   surrogate\_control}\OperatorTok{=}\NormalTok{\{}\StringTok{"noise"}\NormalTok{: }\VariableTok{False}\NormalTok{,}
                                      \StringTok{"cod\_type"}\NormalTok{: }\StringTok{"norm"}\NormalTok{,}
                                      \StringTok{"min\_theta"}\NormalTok{: }\OperatorTok{{-}}\DecValTok{4}\NormalTok{,}
                                      \StringTok{"max\_theta"}\NormalTok{: }\DecValTok{3}\NormalTok{,}
                                      \StringTok{"n\_theta"}\NormalTok{: }\BuiltInTok{len}\NormalTok{(var\_name),}
                                      \StringTok{"model\_fun\_evals"}\NormalTok{: }\DecValTok{10\_000}\NormalTok{\})}
\NormalTok{spot\_tuner.run(X\_start}\OperatorTok{=}\NormalTok{X\_start)}
\end{Highlighting}
\end{Shaded}

\begin{verbatim}
spotPython tuning: 2.1954027176053987 [##--------] 18.25% 
\end{verbatim}

\begin{verbatim}
spotPython tuning: 2.1954027176053987 [###-------] 34.32% 
\end{verbatim}

\begin{verbatim}
spotPython tuning: 2.1954027176053987 [#####-----] 48.89% 
\end{verbatim}

\begin{verbatim}
spotPython tuning: 2.1558528518089006 [######----] 63.67% 
\end{verbatim}

\begin{verbatim}
spotPython tuning: 2.1189652804422368 [########--] 75.61% 
\end{verbatim}

\begin{verbatim}
spotPython tuning: 2.1189652804422368 [########--] 84.84% 
\end{verbatim}

\begin{verbatim}
spotPython tuning: 2.1189652804422368 [##########] 100.00% Done...
\end{verbatim}

\begin{verbatim}
<spotPython.spot.spot.Spot at 0x2a258b730>
\end{verbatim}

\hypertarget{sec-tensorboard-10}{%
\subsection{TensorBoard}\label{sec-tensorboard-10}}

Now we can start TensorBoard in the background with the following
command, where \texttt{./runs} is the default directory for the
TensorBoard log files:

\begin{Shaded}
\begin{Highlighting}[]
\NormalTok{tensorboard {-}{-}logdir="./runs"}
\end{Highlighting}
\end{Shaded}

\begin{tcolorbox}[enhanced jigsaw, left=2mm, title=\textcolor{quarto-callout-tip-color}{\faLightbulb}\hspace{0.5em}{Tip: TENSORBOARD\_PATH}, bottomrule=.15mm, titlerule=0mm, breakable, rightrule=.15mm, toprule=.15mm, coltitle=black, colbacktitle=quarto-callout-tip-color!10!white, leftrule=.75mm, arc=.35mm, colframe=quarto-callout-tip-color-frame, bottomtitle=1mm, colback=white, opacitybacktitle=0.6, toptitle=1mm, opacityback=0]

The TensorBoard path can be printed with the following command:

\begin{Shaded}
\begin{Highlighting}[]
\ImportTok{from}\NormalTok{ spotPython.utils.}\BuiltInTok{file} \ImportTok{import}\NormalTok{ get\_tensorboard\_path}
\NormalTok{get\_tensorboard\_path(fun\_control)}
\end{Highlighting}
\end{Shaded}

\begin{verbatim}
'runs/'
\end{verbatim}

\end{tcolorbox}

We can access the TensorBoard web server with the following URL:

\begin{Shaded}
\begin{Highlighting}[]
\NormalTok{http://localhost:6006/}
\end{Highlighting}
\end{Shaded}

The TensorBoard plot illustrates how \texttt{spotPython} can be used as
a microscope for the internal mechanisms of the surrogate-based
optimization process. Here, one important parameter, the learning rate
\(\theta\) of the Kriging surrogate
\href{https://github.com/sequential-parameter-optimization/spotPython/blob/main/src/spotPython/build/kriging.py}{{[}SOURCE{]}}
is plotted against the number of optimization steps.

\begin{figure}

{\centering \includegraphics[width=1\textwidth,height=\textheight]{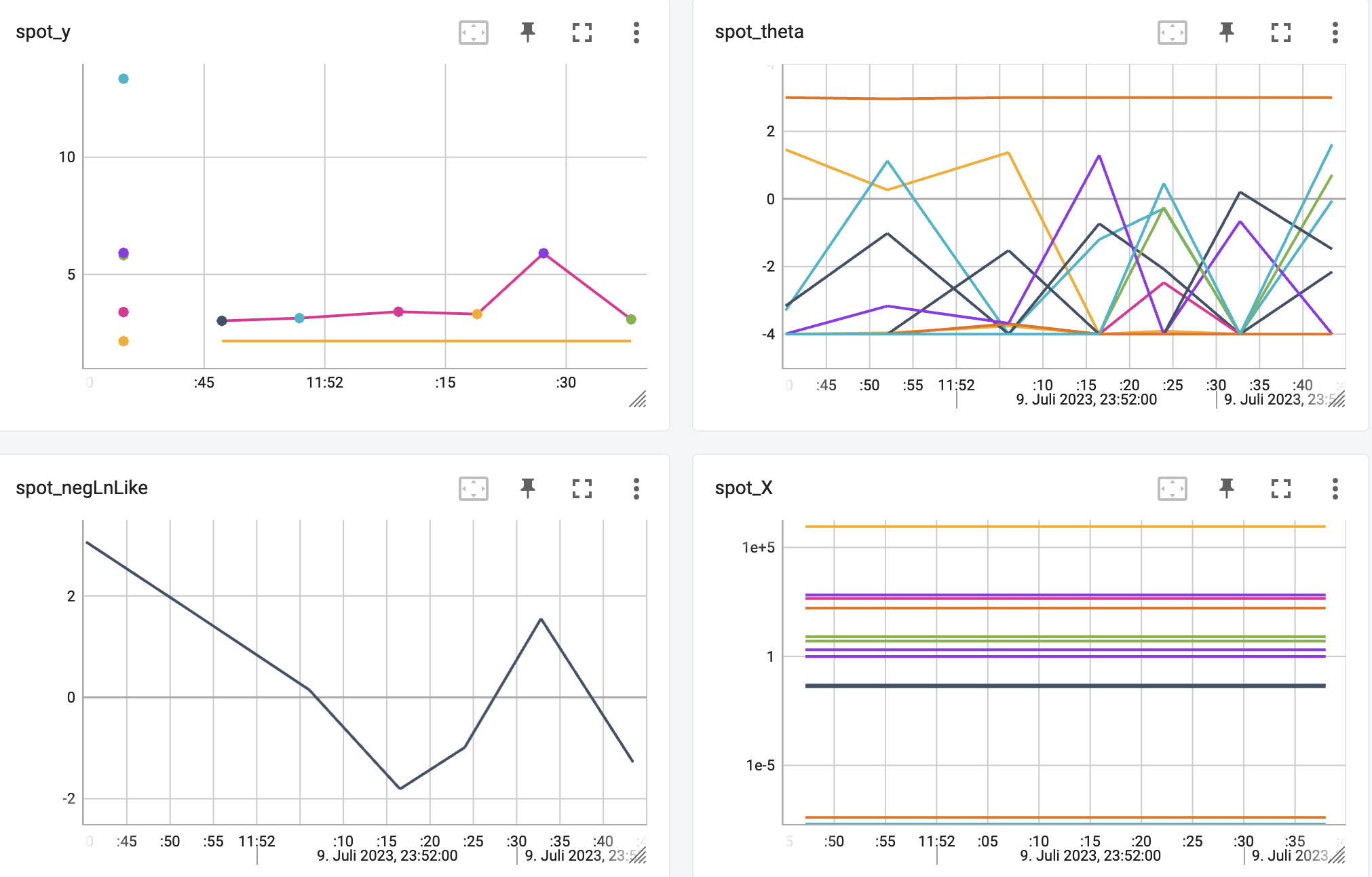}

}

\caption{TensorBoard visualization of the spotPython optimization
process and the surrogate model.}

\end{figure}

\hypertarget{results-4}{%
\subsection{Results}\label{results-4}}

After the hyperparameter tuning run is finished, the results can be
saved and reloaded with the following commands:

\begin{Shaded}
\begin{Highlighting}[]
\ImportTok{from}\NormalTok{ spotPython.utils.}\BuiltInTok{file} \ImportTok{import}\NormalTok{ save\_pickle}
\NormalTok{save\_pickle(spot\_tuner, experiment\_name)}
\end{Highlighting}
\end{Shaded}

\begin{Shaded}
\begin{Highlighting}[]
\ImportTok{from}\NormalTok{ spotPython.utils.}\BuiltInTok{file} \ImportTok{import}\NormalTok{ load\_pickle}
\NormalTok{spot\_tuner }\OperatorTok{=}\NormalTok{ load\_pickle(experiment\_name)}
\end{Highlighting}
\end{Shaded}

After the hyperparameter tuning run is finished, the progress of the
hyperparameter tuning can be visualized. The black points represent the
performace values (score or metric) of hyperparameter configurations
from the initial design, whereas the red points represents the
hyperparameter configurations found by the surrogate model based
optimization.

\begin{Shaded}
\begin{Highlighting}[]
\NormalTok{spot\_tuner.plot\_progress(log\_y}\OperatorTok{=}\VariableTok{True}\NormalTok{, filename}\OperatorTok{=}\StringTok{"./figures/"} \OperatorTok{+}\NormalTok{ experiment\_name}\OperatorTok{+}\StringTok{"\_progress.pdf"}\NormalTok{)}
\end{Highlighting}
\end{Shaded}

\begin{figure}[H]

{\centering \includegraphics{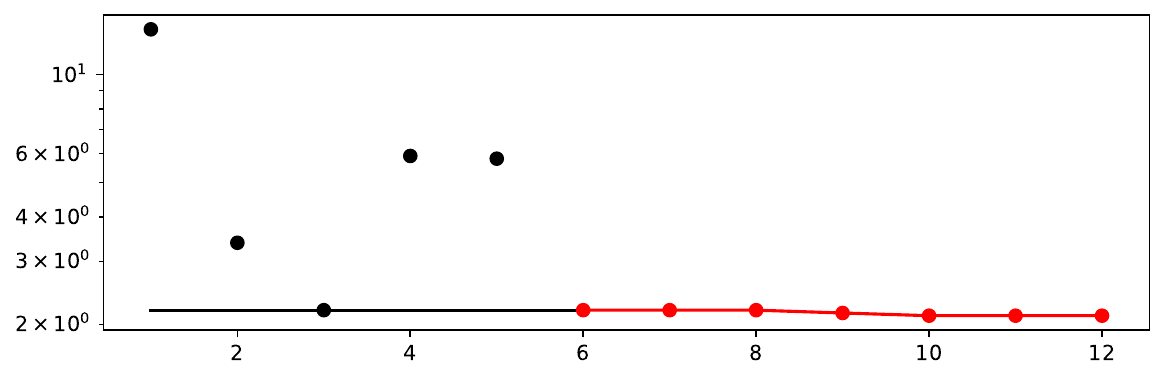}

}

\end{figure}

Results can also be printed in tabular form.

\begin{Shaded}
\begin{Highlighting}[]
\BuiltInTok{print}\NormalTok{(gen\_design\_table(fun\_control}\OperatorTok{=}\NormalTok{fun\_control, spot}\OperatorTok{=}\NormalTok{spot\_tuner))}
\end{Highlighting}
\end{Shaded}

\begin{verbatim}
| name                   | type   | default          |    lower |     upper |              tuned | transform             |   importance | stars   |
|------------------------|--------|------------------|----------|-----------|--------------------|-----------------------|--------------|---------|
| grace_period           | int    | 200              |     10.0 |    1000.0 |              531.0 | None                  |         0.00 |         |
| max_depth              | int    | 20               |      2.0 |      20.0 |                2.0 | transform_power_2_int |         0.00 |         |
| delta                  | float  | 1e-07            |    1e-10 |     1e-06 |              1e-06 | None                  |         0.00 |         |
| tau                    | float  | 0.05             |     0.01 |       0.1 |               0.01 | None                  |         0.00 |         |
| leaf_prediction        | factor | mean             |      0.0 |       2.0 |                1.0 | None                  |         0.26 | .       |
| leaf_model             | factor | LinearRegression |      0.0 |       2.0 |                0.0 | None                  |         0.77 | .       |
| model_selector_decay   | float  | 0.95             |      0.9 |      0.99 |                0.9 | None                  |         0.00 |         |
| splitter               | factor | EBSTSplitter     |      0.0 |       2.0 |                2.0 | None                  |       100.00 | ***     |
| min_samples_split      | int    | 5                |      2.0 |      10.0 |                3.0 | None                  |         0.00 |         |
| bootstrap_sampling     | factor | 0                |      0.0 |       1.0 |                0.0 | None                  |         0.00 |         |
| drift_window_threshold | int    | 300              |    100.0 |     500.0 |              219.0 | None                  |         0.00 |         |
| switch_significance    | float  | 0.05             |     0.01 |       0.1 |               0.01 | None                  |         0.00 |         |
| binary_split           | factor | 0                |      0.0 |       1.0 |                0.0 | None                  |         0.00 |         |
| max_size               | float  | 500.0            |    100.0 |    1000.0 | 117.02444411023869 | None                  |         0.00 |         |
| memory_estimate_period | int    | 1000000          | 100000.0 | 1000000.0 |           382498.0 | None                  |         0.00 |         |
| stop_mem_management    | factor | 0                |      0.0 |       1.0 |                1.0 | None                  |         0.14 | .       |
| remove_poor_attrs      | factor | 0                |      0.0 |       1.0 |                1.0 | None                  |         0.00 |         |
| merit_preprune         | factor | 0                |      0.0 |       0.0 |                0.0 | None                  |         0.00 |         |
\end{verbatim}

A histogram can be used to visualize the most important hyperparameters.

\begin{Shaded}
\begin{Highlighting}[]
\NormalTok{spot\_tuner.plot\_importance(threshold}\OperatorTok{=}\FloatTok{0.0025}\NormalTok{, filename}\OperatorTok{=}\StringTok{"./figures/"} \OperatorTok{+}\NormalTok{ experiment\_name}\OperatorTok{+}\StringTok{"\_importance.pdf"}\NormalTok{)}
\end{Highlighting}
\end{Shaded}

\begin{figure}[H]

{\centering \includegraphics{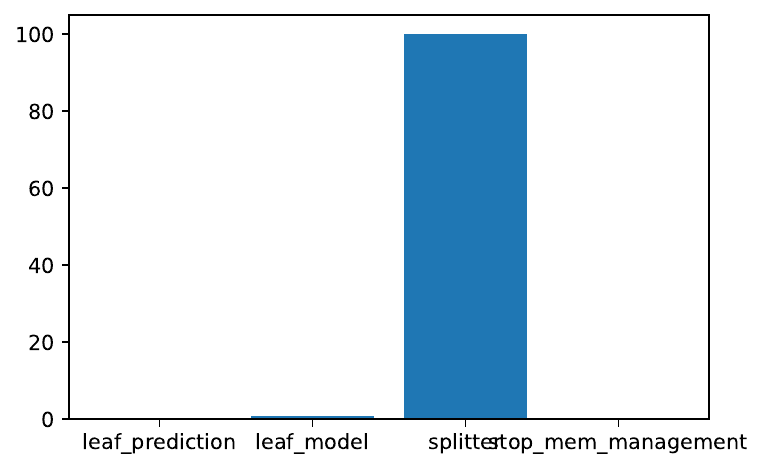}

}

\end{figure}

\hypertarget{the-larger-data-set}{%
\section{The Larger Data Set}\label{the-larger-data-set}}

After the hyperparamter were tuned on a small data set, we can now apply
the hyperparameter configuration to a larger data set. The following
code snippet shows how to generate the larger data set.

\begin{tcolorbox}[enhanced jigsaw, left=2mm, title=\textcolor{quarto-callout-caution-color}{\faFire}\hspace{0.5em}{Caution: Increased Friedman-Drift Data Set}, bottomrule=.15mm, titlerule=0mm, breakable, rightrule=.15mm, toprule=.15mm, coltitle=black, colbacktitle=quarto-callout-caution-color!10!white, leftrule=.75mm, arc=.35mm, colframe=quarto-callout-caution-color-frame, bottomtitle=1mm, colback=white, opacitybacktitle=0.6, toptitle=1mm, opacityback=0]

\begin{itemize}
\tightlist
\item
  The Friedman-Drift Data Set is increased by a factor of two to show
  the transferability of the hyperparameter tuning results.
\item
  Larger values of \texttt{K} lead to a longer run time.
\end{itemize}

\end{tcolorbox}

\begin{Shaded}
\begin{Highlighting}[]
\NormalTok{K }\OperatorTok{=} \FloatTok{0.2}
\NormalTok{n\_samples }\OperatorTok{=} \BuiltInTok{int}\NormalTok{(K}\OperatorTok{*}\DecValTok{100\_000}\NormalTok{)}
\NormalTok{p\_1 }\OperatorTok{=} \BuiltInTok{int}\NormalTok{(K}\OperatorTok{*}\DecValTok{25\_000}\NormalTok{)}
\NormalTok{p\_2 }\OperatorTok{=} \BuiltInTok{int}\NormalTok{(K}\OperatorTok{*}\DecValTok{50\_000}\NormalTok{)}
\NormalTok{position}\OperatorTok{=}\NormalTok{(p\_1, p\_2)}
\end{Highlighting}
\end{Shaded}

\begin{Shaded}
\begin{Highlighting}[]
\NormalTok{dataset }\OperatorTok{=}\NormalTok{ synth.FriedmanDrift(}
\NormalTok{   drift\_type}\OperatorTok{=}\StringTok{\textquotesingle{}gra\textquotesingle{}}\NormalTok{,}
\NormalTok{   position}\OperatorTok{=}\NormalTok{position,}
\NormalTok{   seed}\OperatorTok{=}\DecValTok{123}
\NormalTok{)}
\end{Highlighting}
\end{Shaded}

The larger data set is converted to a Pandas data frame and passed to
the \texttt{fun\_control} dictionary.

\begin{Shaded}
\begin{Highlighting}[]
\NormalTok{df }\OperatorTok{=}\NormalTok{ convert\_to\_df(dataset, target\_column}\OperatorTok{=}\NormalTok{target\_column, n\_total}\OperatorTok{=}\NormalTok{n\_samples)}
\NormalTok{df.columns }\OperatorTok{=}\NormalTok{ [}\SpecialStringTok{f"x}\SpecialCharTok{\{}\NormalTok{i}\SpecialCharTok{\}}\SpecialStringTok{"} \ControlFlowTok{for}\NormalTok{ i }\KeywordTok{in} \BuiltInTok{range}\NormalTok{(}\DecValTok{1}\NormalTok{, }\DecValTok{11}\NormalTok{)] }\OperatorTok{+}\NormalTok{ [}\StringTok{"y"}\NormalTok{]}
\NormalTok{fun\_control.update(\{}\StringTok{"train"}\NormalTok{: df[:n\_train],}
                    \StringTok{"test"}\NormalTok{: df[n\_train:],}
                    \StringTok{"n\_samples"}\NormalTok{: n\_samples,}
                    \StringTok{"target\_column"}\NormalTok{: target\_column\})}
\end{Highlighting}
\end{Shaded}

\hypertarget{get-default-hyperparameters-1}{%
\section{Get Default
Hyperparameters}\label{get-default-hyperparameters-1}}

The default hyperparameters, whihc will be used for a comparion with the
tuned hyperparameters, can be obtained with the following commands:

\begin{Shaded}
\begin{Highlighting}[]
\ImportTok{from}\NormalTok{ spotPython.hyperparameters.values }\ImportTok{import}\NormalTok{ get\_one\_core\_model\_from\_X}
\ImportTok{from}\NormalTok{ spotPython.hyperparameters.values }\ImportTok{import}\NormalTok{ get\_default\_hyperparameters\_as\_array}
\NormalTok{X\_start }\OperatorTok{=}\NormalTok{ get\_default\_hyperparameters\_as\_array(fun\_control)}
\NormalTok{model\_default }\OperatorTok{=}\NormalTok{ get\_one\_core\_model\_from\_X(X\_start, fun\_control)}
\end{Highlighting}
\end{Shaded}

\begin{tcolorbox}[enhanced jigsaw, left=2mm, title=\textcolor{quarto-callout-note-color}{\faInfo}\hspace{0.5em}{Note: \texttt{spotPython} tunes numpy arrays}, bottomrule=.15mm, titlerule=0mm, breakable, rightrule=.15mm, toprule=.15mm, coltitle=black, colbacktitle=quarto-callout-note-color!10!white, leftrule=.75mm, arc=.35mm, colframe=quarto-callout-note-color-frame, bottomtitle=1mm, colback=white, opacitybacktitle=0.6, toptitle=1mm, opacityback=0]

\begin{itemize}
\tightlist
\item
  \texttt{spotPython} tunes numpy arrays, i.e., the hyperparameters are
  stored in a numpy array.
\end{itemize}

\end{tcolorbox}

The model with the default hyperparameters can be trained and evaluated
with the following commands:

\begin{Shaded}
\begin{Highlighting}[]
\ImportTok{from}\NormalTok{ spotRiver.evaluation.eval\_bml }\ImportTok{import}\NormalTok{ eval\_oml\_horizon}

\NormalTok{df\_eval\_default, df\_true\_default }\OperatorTok{=}\NormalTok{ eval\_oml\_horizon(}
\NormalTok{                    model}\OperatorTok{=}\NormalTok{model\_default,}
\NormalTok{                    train}\OperatorTok{=}\NormalTok{fun\_control[}\StringTok{"train"}\NormalTok{],}
\NormalTok{                    test}\OperatorTok{=}\NormalTok{fun\_control[}\StringTok{"test"}\NormalTok{],}
\NormalTok{                    target\_column}\OperatorTok{=}\NormalTok{fun\_control[}\StringTok{"target\_column"}\NormalTok{],}
\NormalTok{                    horizon}\OperatorTok{=}\NormalTok{fun\_control[}\StringTok{"horizon"}\NormalTok{],}
\NormalTok{                    oml\_grace\_period}\OperatorTok{=}\NormalTok{fun\_control[}\StringTok{"oml\_grace\_period"}\NormalTok{],}
\NormalTok{                    metric}\OperatorTok{=}\NormalTok{fun\_control[}\StringTok{"metric\_sklearn"}\NormalTok{],}
\NormalTok{                )}
\end{Highlighting}
\end{Shaded}

The three performance criteria, i.e., scaoe (metric), runtime, and
memory consumption, can be visualized with the following commands:

\begin{Shaded}
\begin{Highlighting}[]
\ImportTok{from}\NormalTok{ spotRiver.evaluation.eval\_bml }\ImportTok{import}\NormalTok{ plot\_bml\_oml\_horizon\_metrics, plot\_bml\_oml\_horizon\_predictions}
\NormalTok{df\_labels}\OperatorTok{=}\NormalTok{[}\StringTok{"default"}\NormalTok{]}
\NormalTok{plot\_bml\_oml\_horizon\_metrics(df\_eval }\OperatorTok{=}\NormalTok{ [df\_eval\_default], log\_y}\OperatorTok{=}\VariableTok{False}\NormalTok{, df\_labels}\OperatorTok{=}\NormalTok{df\_labels, metric}\OperatorTok{=}\NormalTok{fun\_control[}\StringTok{"metric\_sklearn"}\NormalTok{])}
\end{Highlighting}
\end{Shaded}

\begin{figure}[H]

{\centering \includegraphics{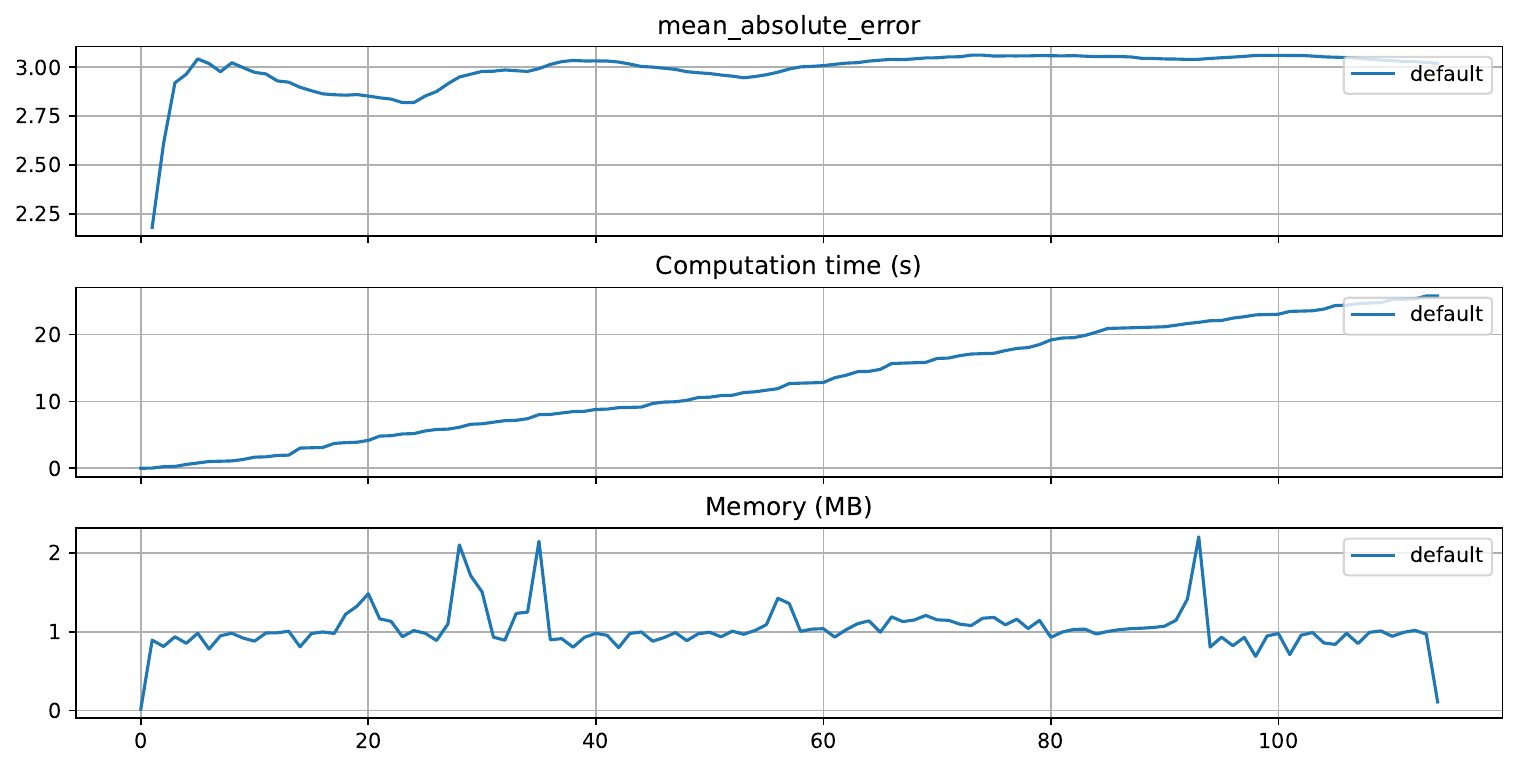}

}

\end{figure}

\hypertarget{show-predictions}{%
\subsection{Show Predictions}\label{show-predictions}}

\begin{itemize}
\tightlist
\item
  Select a subset of the data set for the visualization of the
  predictions:

  \begin{itemize}
  \tightlist
  \item
    We use the mean, \(m\), of the data set as the center of the
    visualization.
  \item
    We use 100 data points, i.e., \(m \pm 50\) as the visualization
    window.
  \end{itemize}
\end{itemize}

\begin{Shaded}
\begin{Highlighting}[]
\NormalTok{m }\OperatorTok{=}\NormalTok{ fun\_control[}\StringTok{"test"}\NormalTok{].shape[}\DecValTok{0}\NormalTok{]}
\NormalTok{a }\OperatorTok{=} \BuiltInTok{int}\NormalTok{(m}\OperatorTok{/}\DecValTok{2}\NormalTok{)}\OperatorTok{{-}}\DecValTok{50}
\NormalTok{b }\OperatorTok{=} \BuiltInTok{int}\NormalTok{(m}\OperatorTok{/}\DecValTok{2}\NormalTok{)}
\end{Highlighting}
\end{Shaded}

\begin{Shaded}
\begin{Highlighting}[]
\NormalTok{plot\_bml\_oml\_horizon\_predictions(df\_true }\OperatorTok{=}\NormalTok{ [df\_true\_default[a:b]], target\_column}\OperatorTok{=}\NormalTok{target\_column,  df\_labels}\OperatorTok{=}\NormalTok{df\_labels)}
\end{Highlighting}
\end{Shaded}

\begin{figure}[H]

{\centering \includegraphics{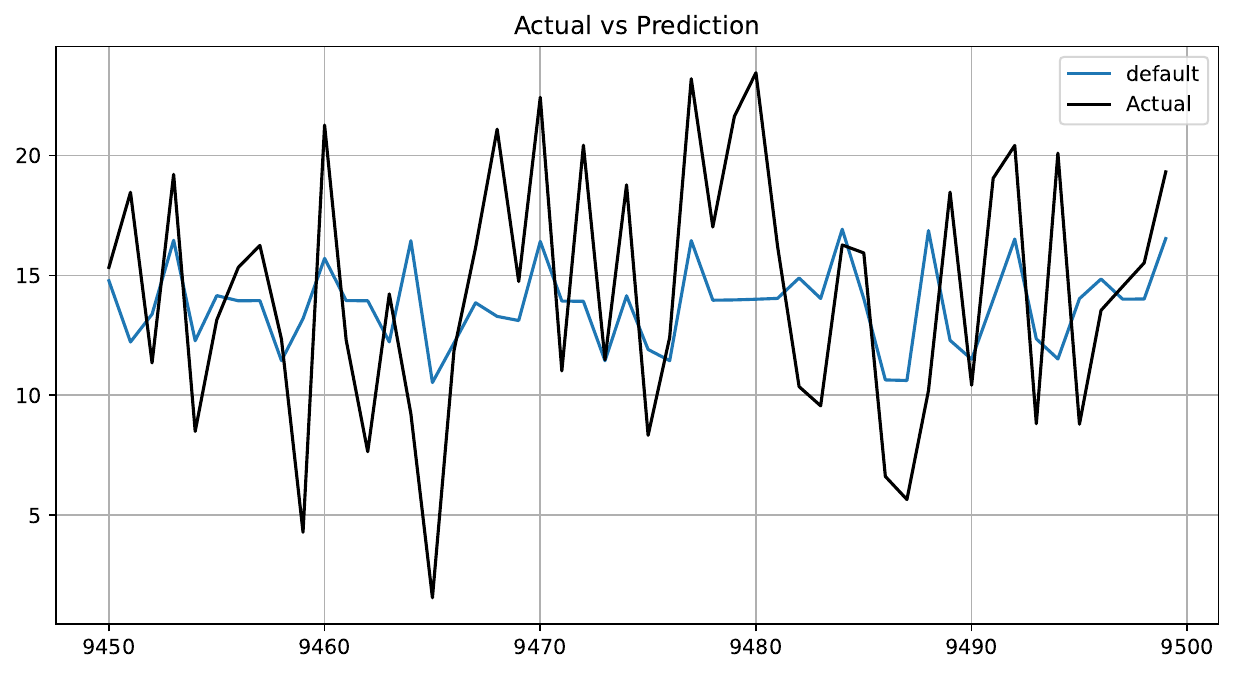}

}

\end{figure}

\hypertarget{get-spot-results-1}{%
\section{Get SPOT Results}\label{get-spot-results-1}}

In a similar way, we can obtain the hyperparameters found by
\texttt{spotPython}.

\begin{Shaded}
\begin{Highlighting}[]
\ImportTok{from}\NormalTok{ spotPython.hyperparameters.values }\ImportTok{import}\NormalTok{ get\_one\_core\_model\_from\_X}
\NormalTok{X }\OperatorTok{=}\NormalTok{ spot\_tuner.to\_all\_dim(spot\_tuner.min\_X.reshape(}\DecValTok{1}\NormalTok{,}\OperatorTok{{-}}\DecValTok{1}\NormalTok{))}
\NormalTok{model\_spot }\OperatorTok{=}\NormalTok{ get\_one\_core\_model\_from\_X(X, fun\_control)}
\end{Highlighting}
\end{Shaded}

\begin{Shaded}
\begin{Highlighting}[]
\NormalTok{df\_eval\_spot, df\_true\_spot }\OperatorTok{=}\NormalTok{ eval\_oml\_horizon(}
\NormalTok{                    model}\OperatorTok{=}\NormalTok{model\_spot,}
\NormalTok{                    train}\OperatorTok{=}\NormalTok{fun\_control[}\StringTok{"train"}\NormalTok{],}
\NormalTok{                    test}\OperatorTok{=}\NormalTok{fun\_control[}\StringTok{"test"}\NormalTok{],}
\NormalTok{                    target\_column}\OperatorTok{=}\NormalTok{fun\_control[}\StringTok{"target\_column"}\NormalTok{],}
\NormalTok{                    horizon}\OperatorTok{=}\NormalTok{fun\_control[}\StringTok{"horizon"}\NormalTok{],}
\NormalTok{                    oml\_grace\_period}\OperatorTok{=}\NormalTok{fun\_control[}\StringTok{"oml\_grace\_period"}\NormalTok{],}
\NormalTok{                    metric}\OperatorTok{=}\NormalTok{fun\_control[}\StringTok{"metric\_sklearn"}\NormalTok{],}
\NormalTok{                )}
\end{Highlighting}
\end{Shaded}

\begin{Shaded}
\begin{Highlighting}[]
\NormalTok{df\_labels}\OperatorTok{=}\NormalTok{[}\StringTok{"default"}\NormalTok{, }\StringTok{"spot"}\NormalTok{]}
\NormalTok{plot\_bml\_oml\_horizon\_metrics(df\_eval }\OperatorTok{=}\NormalTok{ [df\_eval\_default, df\_eval\_spot], log\_y}\OperatorTok{=}\VariableTok{False}\NormalTok{, df\_labels}\OperatorTok{=}\NormalTok{df\_labels, metric}\OperatorTok{=}\NormalTok{fun\_control[}\StringTok{"metric\_sklearn"}\NormalTok{], filename}\OperatorTok{=}\StringTok{"./figures/"} \OperatorTok{+}\NormalTok{ experiment\_name}\OperatorTok{+}\StringTok{"\_metrics.pdf"}\NormalTok{)}
\end{Highlighting}
\end{Shaded}

\begin{figure}[H]

{\centering \includegraphics{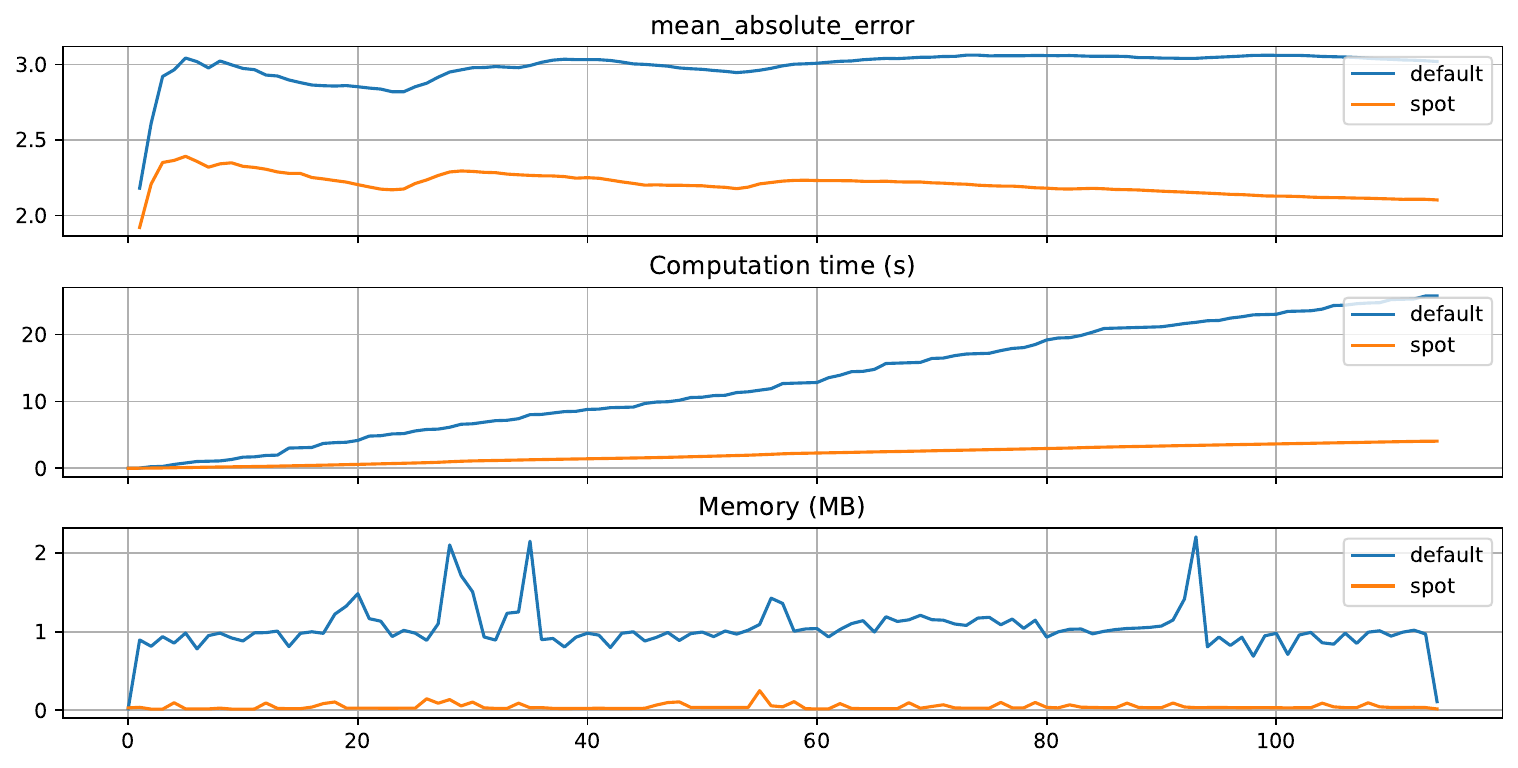}

}

\end{figure}

\begin{Shaded}
\begin{Highlighting}[]
\NormalTok{plot\_bml\_oml\_horizon\_predictions(df\_true }\OperatorTok{=}\NormalTok{ [df\_true\_default[a:b], df\_true\_spot[a:b]], target\_column}\OperatorTok{=}\NormalTok{target\_column,  df\_labels}\OperatorTok{=}\NormalTok{df\_labels, filename}\OperatorTok{=}\StringTok{"./figures/"} \OperatorTok{+}\NormalTok{ experiment\_name}\OperatorTok{+}\StringTok{"\_predictions.pdf"}\NormalTok{)}
\end{Highlighting}
\end{Shaded}

\begin{figure}[H]

{\centering \includegraphics{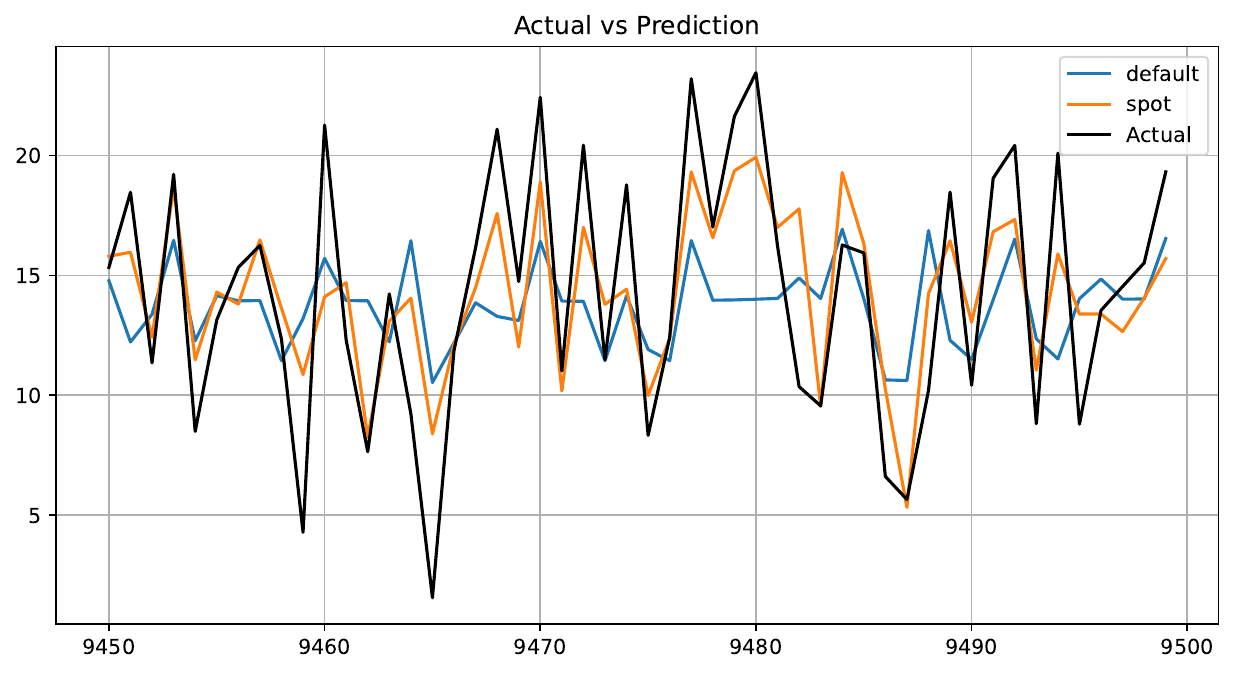}

}

\end{figure}

\begin{Shaded}
\begin{Highlighting}[]
\ImportTok{from}\NormalTok{ spotPython.plot.validation }\ImportTok{import}\NormalTok{ plot\_actual\_vs\_predicted}
\NormalTok{plot\_actual\_vs\_predicted(y\_test}\OperatorTok{=}\NormalTok{df\_true\_default[}\StringTok{"y"}\NormalTok{], y\_pred}\OperatorTok{=}\NormalTok{df\_true\_default[}\StringTok{"Prediction"}\NormalTok{], title}\OperatorTok{=}\StringTok{"Default"}\NormalTok{)}
\NormalTok{plot\_actual\_vs\_predicted(y\_test}\OperatorTok{=}\NormalTok{df\_true\_spot[}\StringTok{"y"}\NormalTok{], y\_pred}\OperatorTok{=}\NormalTok{df\_true\_spot[}\StringTok{"Prediction"}\NormalTok{], title}\OperatorTok{=}\StringTok{"SPOT"}\NormalTok{)}
\end{Highlighting}
\end{Shaded}

\begin{figure}[H]

{\centering \includegraphics{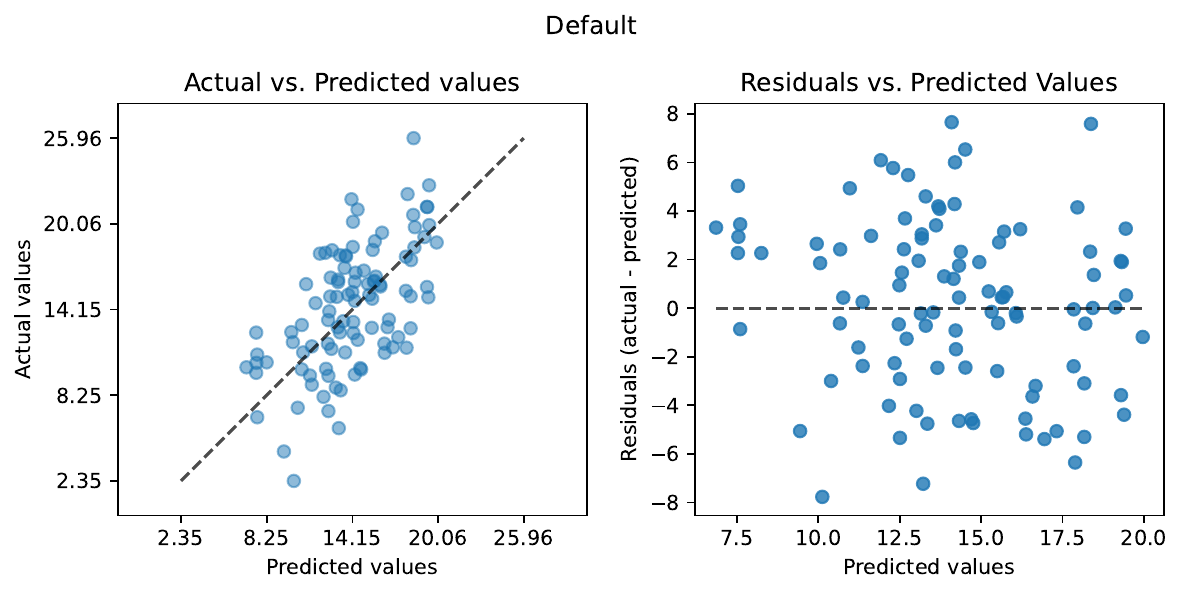}

}

\end{figure}

\begin{figure}[H]

{\centering \includegraphics{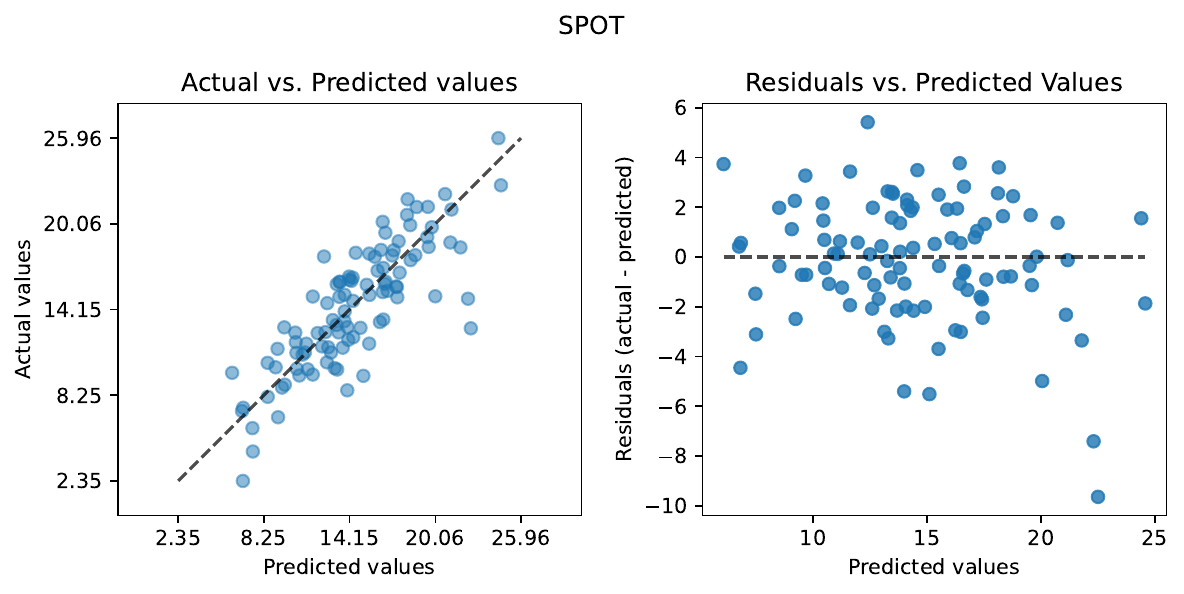}

}

\end{figure}

\hypertarget{visualize-regression-trees}{%
\section{Visualize Regression Trees}\label{visualize-regression-trees}}

\begin{Shaded}
\begin{Highlighting}[]
\NormalTok{dataset\_f }\OperatorTok{=}\NormalTok{ dataset.take(n\_samples)}
\ControlFlowTok{for}\NormalTok{ x, y }\KeywordTok{in}\NormalTok{ dataset\_f:}
\NormalTok{    model\_default.learn\_one(x, y)}
\end{Highlighting}
\end{Shaded}

\begin{tcolorbox}[enhanced jigsaw, left=2mm, title=\textcolor{quarto-callout-caution-color}{\faFire}\hspace{0.5em}{Caution: Large Trees}, bottomrule=.15mm, titlerule=0mm, breakable, rightrule=.15mm, toprule=.15mm, coltitle=black, colbacktitle=quarto-callout-caution-color!10!white, leftrule=.75mm, arc=.35mm, colframe=quarto-callout-caution-color-frame, bottomtitle=1mm, colback=white, opacitybacktitle=0.6, toptitle=1mm, opacityback=0]

\begin{itemize}
\tightlist
\item
  Since the trees are large, the visualization is suppressed by default.
\item
  To visualize the trees, uncomment the following line.
\end{itemize}

\end{tcolorbox}

\begin{Shaded}
\begin{Highlighting}[]
\CommentTok{\# model\_default.draw()}
\end{Highlighting}
\end{Shaded}

\begin{Shaded}
\begin{Highlighting}[]
\NormalTok{model\_default.summary}
\end{Highlighting}
\end{Shaded}

\begin{verbatim}
{'n_nodes': 35,
 'n_branches': 17,
 'n_leaves': 18,
 'n_active_leaves': 96,
 'n_inactive_leaves': 0,
 'height': 6,
 'total_observed_weight': 39002.0,
 'n_alternate_trees': 21,
 'n_pruned_alternate_trees': 6,
 'n_switch_alternate_trees': 2}
\end{verbatim}

\hypertarget{spot-model}{%
\subsection{Spot Model}\label{spot-model}}

\begin{Shaded}
\begin{Highlighting}[]
\NormalTok{dataset\_f }\OperatorTok{=}\NormalTok{ dataset.take(n\_samples)}
\ControlFlowTok{for}\NormalTok{ x, y }\KeywordTok{in}\NormalTok{ dataset\_f:}
\NormalTok{    model\_spot.learn\_one(x, y)}
\end{Highlighting}
\end{Shaded}

\begin{tcolorbox}[enhanced jigsaw, left=2mm, title=\textcolor{quarto-callout-caution-color}{\faFire}\hspace{0.5em}{Caution: Large Trees}, bottomrule=.15mm, titlerule=0mm, breakable, rightrule=.15mm, toprule=.15mm, coltitle=black, colbacktitle=quarto-callout-caution-color!10!white, leftrule=.75mm, arc=.35mm, colframe=quarto-callout-caution-color-frame, bottomtitle=1mm, colback=white, opacitybacktitle=0.6, toptitle=1mm, opacityback=0]

\begin{itemize}
\tightlist
\item
  Since the trees are large, the visualization is suppressed by default.
\item
  To visualize the trees, uncomment the following line.
\end{itemize}

\end{tcolorbox}

\begin{Shaded}
\begin{Highlighting}[]
\CommentTok{\# model\_spot.draw()}
\end{Highlighting}
\end{Shaded}

\begin{Shaded}
\begin{Highlighting}[]
\NormalTok{model\_spot.summary}
\end{Highlighting}
\end{Shaded}

\begin{verbatim}
{'n_nodes': 21,
 'n_branches': 10,
 'n_leaves': 11,
 'n_active_leaves': -3919,
 'n_inactive_leaves': 3956,
 'height': 5,
 'total_observed_weight': 39002.0,
 'n_alternate_trees': 21,
 'n_pruned_alternate_trees': 5,
 'n_switch_alternate_trees': 1}
\end{verbatim}

\begin{Shaded}
\begin{Highlighting}[]
\ImportTok{from}\NormalTok{ spotPython.utils.eda }\ImportTok{import}\NormalTok{ compare\_two\_tree\_models}
\BuiltInTok{print}\NormalTok{(compare\_two\_tree\_models(model\_default, model\_spot))}
\end{Highlighting}
\end{Shaded}

\begin{verbatim}
| Parameter                |   Default |   Spot |
|--------------------------|-----------|--------|
| n_nodes                  |        35 |     21 |
| n_branches               |        17 |     10 |
| n_leaves                 |        18 |     11 |
| n_active_leaves          |        96 |  -3919 |
| n_inactive_leaves        |         0 |   3956 |
| height                   |         6 |      5 |
| total_observed_weight    |     39002 |  39002 |
| n_alternate_trees        |        21 |     21 |
| n_pruned_alternate_trees |         6 |      5 |
| n_switch_alternate_trees |         2 |      1 |
\end{verbatim}

\hypertarget{detailed-hyperparameter-plots-1}{%
\section{Detailed Hyperparameter
Plots}\label{detailed-hyperparameter-plots-1}}

\begin{Shaded}
\begin{Highlighting}[]
\NormalTok{filename }\OperatorTok{=} \StringTok{"./figures/"} \OperatorTok{+}\NormalTok{ experiment\_name}
\NormalTok{spot\_tuner.plot\_important\_hyperparameter\_contour(filename}\OperatorTok{=}\NormalTok{filename)}
\end{Highlighting}
\end{Shaded}

\begin{verbatim}
leaf_prediction:  0.25541913168507413
leaf_model:  0.771487672697361
splitter:  100.0
stop_mem_management:  0.14201240670317347
\end{verbatim}

\begin{figure}[H]

{\centering \includegraphics{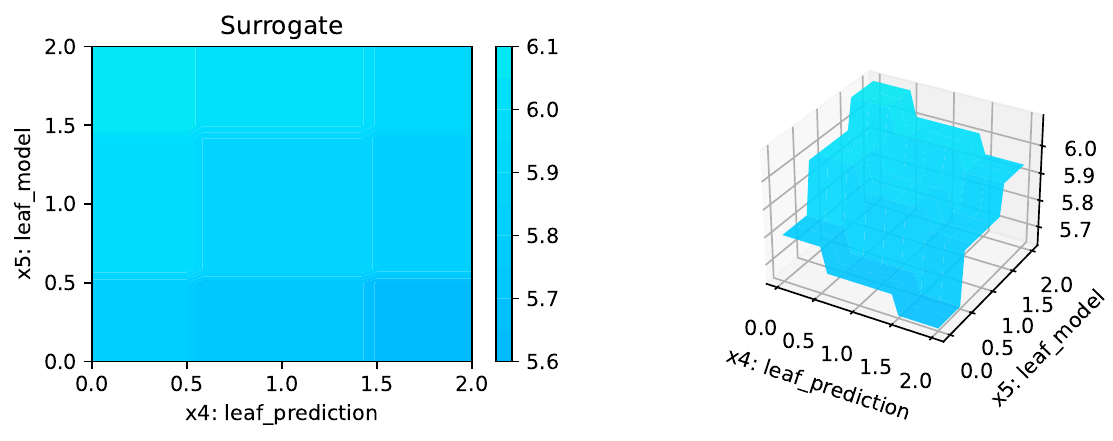}

}

\end{figure}

\begin{figure}[H]

{\centering \includegraphics{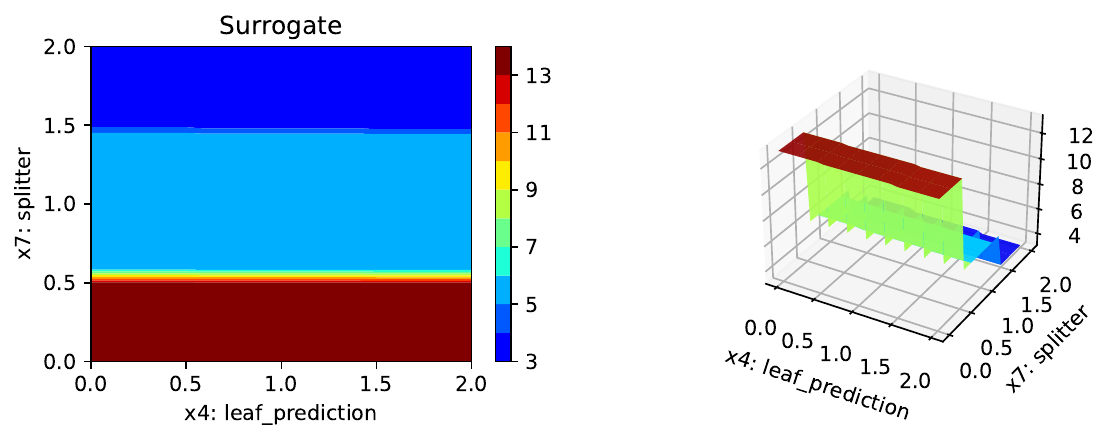}

}

\end{figure}

\begin{figure}[H]

{\centering \includegraphics{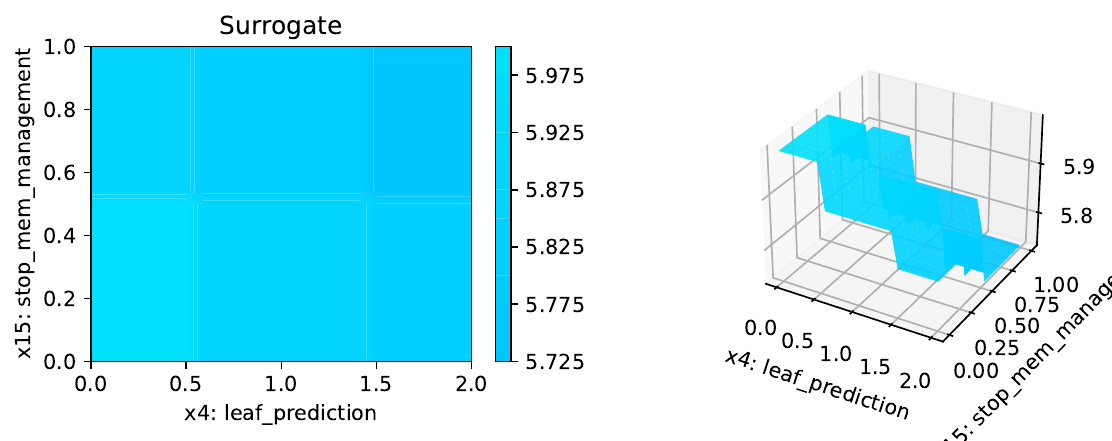}

}

\end{figure}

\begin{figure}[H]

{\centering \includegraphics{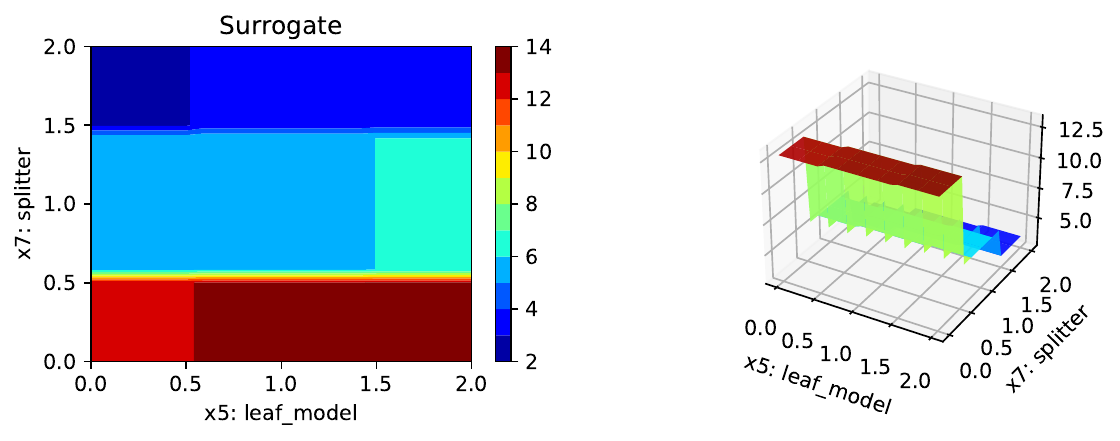}

}

\end{figure}

\begin{figure}[H]

{\centering \includegraphics{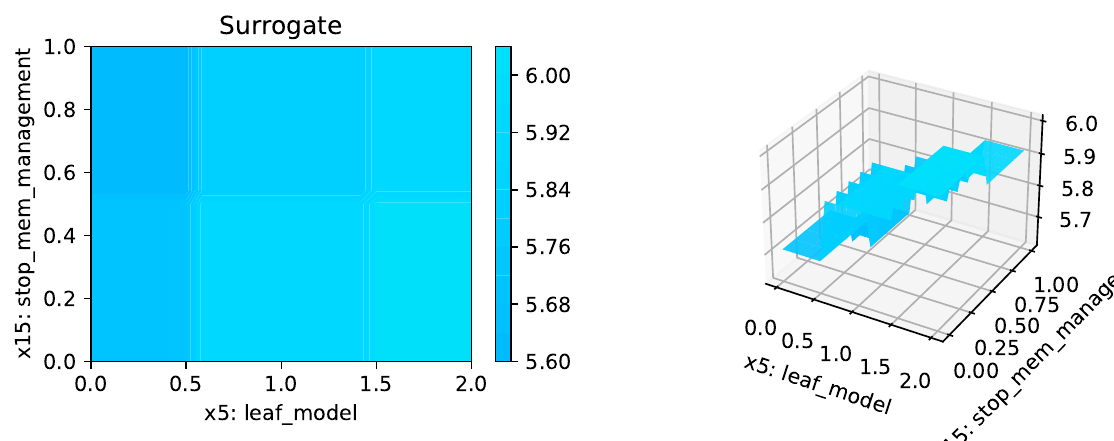}

}

\end{figure}

\begin{figure}[H]

{\centering \includegraphics{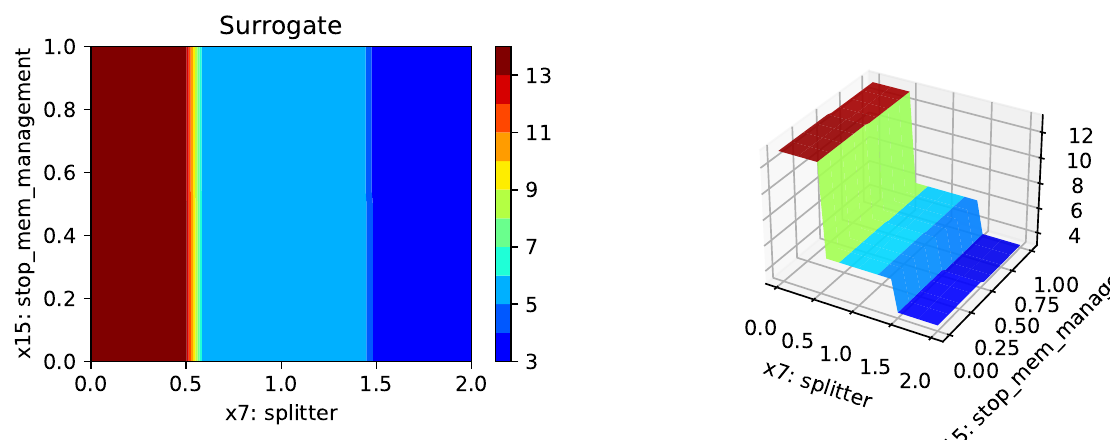}

}

\end{figure}

\hypertarget{parallel-coordinates-plots}{%
\section{Parallel Coordinates Plots}\label{parallel-coordinates-plots}}

\begin{Shaded}
\begin{Highlighting}[]
\NormalTok{spot\_tuner.parallel\_plot()}
\end{Highlighting}
\end{Shaded}

\begin{verbatim}
Unable to display output for mime type(s): text/html
\end{verbatim}

\begin{verbatim}
Unable to display output for mime type(s): text/html
\end{verbatim}

\hypertarget{plot-all-combinations-of-hyperparameters-1}{%
\section{Plot all Combinations of
Hyperparameters}\label{plot-all-combinations-of-hyperparameters-1}}

\begin{itemize}
\tightlist
\item
  Warning: this may take a while.
\end{itemize}

\begin{Shaded}
\begin{Highlighting}[]
\NormalTok{PLOT\_ALL }\OperatorTok{=} \VariableTok{False}
\ControlFlowTok{if}\NormalTok{ PLOT\_ALL:}
\NormalTok{    n }\OperatorTok{=}\NormalTok{ spot\_tuner.k}
    \ControlFlowTok{for}\NormalTok{ i }\KeywordTok{in} \BuiltInTok{range}\NormalTok{(n}\OperatorTok{{-}}\DecValTok{1}\NormalTok{):}
        \ControlFlowTok{for}\NormalTok{ j }\KeywordTok{in} \BuiltInTok{range}\NormalTok{(i}\OperatorTok{+}\DecValTok{1}\NormalTok{, n):}
\NormalTok{            spot\_tuner.plot\_contour(i}\OperatorTok{=}\NormalTok{i, j}\OperatorTok{=}\NormalTok{j, min\_z}\OperatorTok{=}\NormalTok{min\_z, max\_z }\OperatorTok{=}\NormalTok{ max\_z)}
\end{Highlighting}
\end{Shaded}

\hypertarget{sec-hyperparameter-tuning-for-pytorch-14}{%
\chapter{\texorpdfstring{HPT: PyTorch With \texttt{spotPython} and Ray
Tune on
CIFAR10}{HPT: PyTorch With spotPython and Ray Tune on CIFAR10}}\label{sec-hyperparameter-tuning-for-pytorch-14}}

In this tutorial, we will show how \texttt{spotPython} can be integrated
into the \texttt{PyTorch} training workflow. It is based on the tutorial
``Hyperparameter Tuning with Ray Tune'' from the \texttt{PyTorch}
documentation (PyTorch 2023a), which is an extension of the tutorial
``Training a Classifier'' (PyTorch 2023b) for training a CIFAR10 image
classifier.

\begin{tcolorbox}[enhanced jigsaw, left=2mm, title=\textcolor{quarto-callout-note-color}{\faInfo}\hspace{0.5em}{Note: PyTorch and Lightning}, bottomrule=.15mm, titlerule=0mm, breakable, rightrule=.15mm, toprule=.15mm, coltitle=black, colbacktitle=quarto-callout-note-color!10!white, leftrule=.75mm, arc=.35mm, colframe=quarto-callout-note-color-frame, bottomtitle=1mm, colback=white, opacitybacktitle=0.6, toptitle=1mm, opacityback=0]

Instead of using the \texttt{PyTorch} interface directly as explained in
this chapter, we recommend using the \texttt{PyTorch\ Lightning}
interface. The \texttt{PyTorch\ Lightning} interface is explained in
Chapter~\ref{sec-hyperparameter-tuning-lightning-31}

\end{tcolorbox}

A typical hyperparameter tuning process with \texttt{spotPython}
consists of the following steps:

\begin{enumerate}
\def\labelenumi{\arabic{enumi}.}
\tightlist
\item
  Loading the data (training and test datasets), see
  Section~\ref{sec-data-loading-14}.
\item
  Specification of the preprocessing model, see
  Section~\ref{sec-specification-of-preprocessing-model-14}. This model
  is called \texttt{prep\_model} (``preparation'' or pre-processing).
  The information required for the hyperparameter tuning is stored in
  the dictionary \texttt{fun\_control}. Thus, the information needed for
  the execution of the hyperparameter tuning is available in a readable
  form.
\item
  Selection of the machine learning or deep learning model to be tuned,
  see Section~\ref{sec-selection-of-the-algorithm-14}. This is called
  the \texttt{core\_model}. Once the \texttt{core\_model} is defined,
  then the associated hyperparameters are stored in the
  \texttt{fun\_control} dictionary. First, the hyperparameters of the
  \texttt{core\_model} are initialized with the default values of the
  \texttt{core\_model}. As default values we use the default values
  contained in the \texttt{spotPython} package for the algorithms of the
  \texttt{torch} package.
\item
  Modification of the default values for the hyperparameters used in
  \texttt{core\_model}, see
  Section~\ref{sec-modification-of-default-values}. This step is
  optional.

  \begin{enumerate}
  \def\labelenumii{\arabic{enumii}.}
  \tightlist
  \item
    numeric parameters are modified by changing the bounds.
  \item
    categorical parameters are modified by changing the categories
    (``levels'').
  \end{enumerate}
\item
  Selection of target function (loss function) for the optimizer, see
  Section~\ref{sec-loss-functions-14}.
\item
  Calling SPOT with the corresponding parameters, see
  Section~\ref{sec-call-the-hyperparameter-tuner-14}. The results are
  stored in a dictionary and are available for further analysis.
\item
  Presentation, visualization and interpretation of the results, see
  Section~\ref{sec-results-14}.
\end{enumerate}

\texttt{spotPython} can be installed via pip\footnote{Alternatively, the
  source code can be downloaded from gitHub:
  \url{https://github.com/sequential-parameter-optimization/spotPython}.}.

\begin{Shaded}
\begin{Highlighting}[]
\NormalTok{!pip install spotPython}
\end{Highlighting}
\end{Shaded}

\begin{itemize}
\tightlist
\item
  Uncomment the following lines if you want to for (re-)installation the
  latest version of \texttt{spotPython} from gitHub.
\end{itemize}

\begin{Shaded}
\begin{Highlighting}[]
\CommentTok{\# import sys}
\CommentTok{\# !\{sys.executable\} {-}m pip install {-}{-}upgrade build}
\CommentTok{\# !\{sys.executable\} {-}m pip install {-}{-}upgrade {-}{-}force{-}reinstall spotPython}
\end{Highlighting}
\end{Shaded}

Results that refer to the \texttt{Ray\ Tune} package are taken from
\url{https://PyTorch.org/tutorials/beginner/hyperparameter_tuning_tutorial.html}\footnote{We
  were not able to install \texttt{Ray\ Tune} on our system. Therefore,
  we used the results from the \texttt{PyTorch} tutorial.}.

\hypertarget{sec-setup-14}{%
\section{Step 1: Setup}\label{sec-setup-14}}

Before we consider the detailed experimental setup, we select the
parameters that affect run time, initial design size and the device that
is used.

\begin{tcolorbox}[enhanced jigsaw, left=2mm, title=\textcolor{quarto-callout-caution-color}{\faFire}\hspace{0.5em}{Caution: Run time and initial design size should be increased for real
experiments}, bottomrule=.15mm, titlerule=0mm, breakable, rightrule=.15mm, toprule=.15mm, coltitle=black, colbacktitle=quarto-callout-caution-color!10!white, leftrule=.75mm, arc=.35mm, colframe=quarto-callout-caution-color-frame, bottomtitle=1mm, colback=white, opacitybacktitle=0.6, toptitle=1mm, opacityback=0]

\begin{itemize}
\tightlist
\item
  MAX\_TIME is set to one minute for demonstration purposes. For real
  experiments, this should be increased to at least 1 hour.
\item
  INIT\_SIZE is set to 5 for demonstration purposes. For real
  experiments, this should be increased to at least 10.
\end{itemize}

\end{tcolorbox}

\begin{tcolorbox}[enhanced jigsaw, left=2mm, title=\textcolor{quarto-callout-note-color}{\faInfo}\hspace{0.5em}{Note: Device selection}, bottomrule=.15mm, titlerule=0mm, breakable, rightrule=.15mm, toprule=.15mm, coltitle=black, colbacktitle=quarto-callout-note-color!10!white, leftrule=.75mm, arc=.35mm, colframe=quarto-callout-note-color-frame, bottomtitle=1mm, colback=white, opacitybacktitle=0.6, toptitle=1mm, opacityback=0]

\begin{itemize}
\tightlist
\item
  The device can be selected by setting the variable \texttt{DEVICE}.
\item
  Since we are using a simple neural net, the setting \texttt{"cpu"} is
  preferred (on Mac).
\item
  If you have a GPU, you can use \texttt{"cuda:0"} instead.
\item
  If DEVICE is set to \texttt{"auto"} or \texttt{None},
  \texttt{spotPython} will automatically select the device.

  \begin{itemize}
  \tightlist
  \item
    This might result in \texttt{"mps"} on Macs, which is not the best
    choice for simple neural nets.
  \end{itemize}
\end{itemize}

\end{tcolorbox}

\begin{Shaded}
\begin{Highlighting}[]
\NormalTok{MAX\_TIME }\OperatorTok{=} \DecValTok{1}
\NormalTok{INIT\_SIZE }\OperatorTok{=} \DecValTok{5}
\NormalTok{DEVICE }\OperatorTok{=} \StringTok{"auto"} \CommentTok{\# "cpu"}
\NormalTok{PREFIX }\OperatorTok{=} \StringTok{"14{-}torch"}
\end{Highlighting}
\end{Shaded}

\begin{Shaded}
\begin{Highlighting}[]
\ImportTok{from}\NormalTok{ spotPython.utils.device }\ImportTok{import}\NormalTok{ getDevice}
\NormalTok{DEVICE }\OperatorTok{=}\NormalTok{ getDevice(DEVICE)}
\BuiltInTok{print}\NormalTok{(DEVICE)}
\end{Highlighting}
\end{Shaded}

\begin{verbatim}
mps
\end{verbatim}

\begin{Shaded}
\begin{Highlighting}[]
\ImportTok{import}\NormalTok{ warnings}
\NormalTok{warnings.filterwarnings(}\StringTok{"ignore"}\NormalTok{)}
\end{Highlighting}
\end{Shaded}

\hypertarget{sec-initialization-fun-control-14}{%
\section{\texorpdfstring{Step 2: Initialization of the
\texttt{fun\_control}
Dictionary}{Step 2: Initialization of the fun\_control Dictionary}}\label{sec-initialization-fun-control-14}}

\texttt{spotPython} uses a Python dictionary for storing the information
required for the hyperparameter tuning process. This dictionary is
called \texttt{fun\_control} and is initialized with the function
\texttt{fun\_control\_init}. The function \texttt{fun\_control\_init}
returns a skeleton dictionary. The dictionary is filled with the
required information for the hyperparameter tuning process. It stores
the hyperparameter tuning settings, e.g., the deep learning network
architecture that should be tuned, the classification (or regression)
problem, and the data that is used for the tuning. The dictionary is
used as an input for the SPOT function.

\begin{Shaded}
\begin{Highlighting}[]
\ImportTok{from}\NormalTok{ spotPython.utils.init }\ImportTok{import}\NormalTok{ fun\_control\_init}
\ImportTok{from}\NormalTok{ spotPython.utils.}\BuiltInTok{file} \ImportTok{import}\NormalTok{ get\_experiment\_name, get\_spot\_tensorboard\_path}
\ImportTok{from}\NormalTok{ spotPython.utils.device }\ImportTok{import}\NormalTok{ getDevice}

\NormalTok{experiment\_name }\OperatorTok{=}\NormalTok{ get\_experiment\_name(prefix}\OperatorTok{=}\NormalTok{PREFIX)}

\NormalTok{fun\_control }\OperatorTok{=}\NormalTok{ fun\_control\_init(}
\NormalTok{    task}\OperatorTok{=}\StringTok{"classification"}\NormalTok{,}
\NormalTok{    spot\_tensorboard\_path}\OperatorTok{=}\NormalTok{get\_spot\_tensorboard\_path(experiment\_name),}
\NormalTok{    device}\OperatorTok{=}\NormalTok{DEVICE,)}
\end{Highlighting}
\end{Shaded}

\hypertarget{sec-data-loading-14}{%
\section{Step 3: PyTorch Data Loading}\label{sec-data-loading-14}}

The data loading process is implemented in the same manner as described
in the Section ``Data loaders'' in PyTorch (2023a). The data loaders are
wrapped into the function \texttt{load\_data\_cifar10} which is
identical to the function \texttt{load\_data} in PyTorch (2023a). A
global data directory is used, which allows sharing the data directory
between different trials. The method \texttt{load\_data\_cifar10} is
part of the \texttt{spotPython} package and can be imported from
\texttt{spotPython.data.torchdata}.

In the following step, the test and train data are added to the
dictionary \texttt{fun\_control}.

\begin{Shaded}
\begin{Highlighting}[]
\ImportTok{from}\NormalTok{ spotPython.data.torchdata }\ImportTok{import}\NormalTok{ load\_data\_cifar10}
\NormalTok{train, test }\OperatorTok{=}\NormalTok{ load\_data\_cifar10()}
\NormalTok{n\_samples }\OperatorTok{=} \BuiltInTok{len}\NormalTok{(train)}
\CommentTok{\# add the dataset to the fun\_control}
\NormalTok{fun\_control.update(\{}
    \StringTok{"train"}\NormalTok{: train,}
    \StringTok{"test"}\NormalTok{: test,}
    \StringTok{"n\_samples"}\NormalTok{: n\_samples\})}
\end{Highlighting}
\end{Shaded}

\begin{verbatim}
Files already downloaded and verified
\end{verbatim}

\begin{verbatim}
Files already downloaded and verified
\end{verbatim}

\hypertarget{sec-specification-of-preprocessing-model-14}{%
\section{Step 4: Specification of the Preprocessing
Model}\label{sec-specification-of-preprocessing-model-14}}

After the training and test data are specified and added to the
\texttt{fun\_control} dictionary, \texttt{spotPython} allows the
specification of a data preprocessing pipeline, e.g., for the scaling of
the data or for the one-hot encoding of categorical variables. The
preprocessing model is called \texttt{prep\_model} (``preparation'' or
pre-processing) and includes steps that are not subject to the
hyperparameter tuning process. The preprocessing model is specified in
the \texttt{fun\_control} dictionary. The preprocessing model can be
implemented as a \texttt{sklearn} pipeline. The following code shows a
typical preprocessing pipeline:

\begin{Shaded}
\begin{Highlighting}[]
\NormalTok{categorical\_columns = ["cities", "colors"]}
\NormalTok{one\_hot\_encoder = OneHotEncoder(handle\_unknown="ignore",}
\NormalTok{                                    sparse\_output=False)}
\NormalTok{prep\_model = ColumnTransformer(}
\NormalTok{        transformers=[}
\NormalTok{             ("categorical", one\_hot\_encoder, categorical\_columns),}
\NormalTok{         ],}
\NormalTok{         remainder=StandardScaler(),}
\NormalTok{     )}
\end{Highlighting}
\end{Shaded}

Because the Ray Tune (\texttt{ray{[}tune{]}}) hyperparameter tuning as
described in PyTorch (2023a) does not use a preprocessing model, the
preprocessing model is set to \texttt{None} here.

\begin{Shaded}
\begin{Highlighting}[]
\NormalTok{prep\_model }\OperatorTok{=} \VariableTok{None}
\NormalTok{fun\_control.update(\{}\StringTok{"prep\_model"}\NormalTok{: prep\_model\})}
\end{Highlighting}
\end{Shaded}

\hypertarget{sec-selection-of-the-algorithm-14}{%
\section{\texorpdfstring{Step 5: Select Model (\texttt{algorithm}) and
\texttt{core\_model\_hyper\_dict}}{Step 5: Select Model (algorithm) and core\_model\_hyper\_dict}}\label{sec-selection-of-the-algorithm-14}}

The same neural network model as implemented in the section
``Configurable neural network'' of the \texttt{PyTorch} tutorial
(PyTorch 2023a) is used here. We will show the implementation from
PyTorch (2023a) in Section~\ref{sec-implementation-with-raytune} first,
before the extended implementation with \texttt{spotPython} is shown in
Section~\ref{sec-implementation-with-spotpython-14}.

\hypertarget{sec-implementation-with-raytune}{%
\subsubsection{Implementing a Configurable Neural Network With Ray
Tune}\label{sec-implementation-with-raytune}}

We used the same hyperparameters that are implemented as configurable in
the \texttt{PyTorch} tutorial. We specify the layer sizes, namely
\texttt{l1} and \texttt{l2}, of the fully connected layers:

\begin{Shaded}
\begin{Highlighting}[]
\NormalTok{class Net(nn.Module):}
\NormalTok{    def \_\_init\_\_(self, l1=120, l2=84):}
\NormalTok{        super(Net, self).\_\_init\_\_()}
\NormalTok{        self.conv1 = nn.Conv2d(3, 6, 5)}
\NormalTok{        self.pool = nn.MaxPool2d(2, 2)}
\NormalTok{        self.conv2 = nn.Conv2d(6, 16, 5)}
\NormalTok{        self.fc1 = nn.Linear(16 * 5 * 5, l1)}
\NormalTok{        self.fc2 = nn.Linear(l1, l2)}
\NormalTok{        self.fc3 = nn.Linear(l2, 10)}

\NormalTok{    def forward(self, x):}
\NormalTok{        x = self.pool(F.relu(self.conv1(x)))}
\NormalTok{        x = self.pool(F.relu(self.conv2(x)))}
\NormalTok{        x = x.view({-}1, 16 * 5 * 5)}
\NormalTok{        x = F.relu(self.fc1(x))}
\NormalTok{        x = F.relu(self.fc2(x))}
\NormalTok{        x = self.fc3(x)}
\NormalTok{        return x}
\end{Highlighting}
\end{Shaded}

The learning rate, i.e., \texttt{lr}, of the optimizer is made
configurable, too:

\begin{Shaded}
\begin{Highlighting}[]
\NormalTok{optimizer = optim.SGD(net.parameters(), lr=config["lr"], momentum=0.9)}
\end{Highlighting}
\end{Shaded}

\hypertarget{sec-implementation-with-spotpython-14}{%
\subsubsection{Implementing a Configurable Neural Network With
spotPython}\label{sec-implementation-with-spotpython-14}}

\texttt{spotPython} implements a class which is similar to the class
described in the \texttt{PyTorch} tutorial. The class is called
\texttt{Net\_CIFAR10} and is implemented in the file
\texttt{netcifar10.py}.

\begin{Shaded}
\begin{Highlighting}[]
\NormalTok{from torch import nn}
\NormalTok{import torch.nn.functional as F}
\NormalTok{import spotPython.torch.netcore as netcore}


\NormalTok{class Net\_CIFAR10(netcore.Net\_Core):}
\NormalTok{    def \_\_init\_\_(self, l1, l2, lr\_mult, batch\_size, epochs, k\_folds, patience,}
\NormalTok{    optimizer, sgd\_momentum):}
\NormalTok{        super(Net\_CIFAR10, self).\_\_init\_\_(}
\NormalTok{            lr\_mult=lr\_mult,}
\NormalTok{            batch\_size=batch\_size,}
\NormalTok{            epochs=epochs,}
\NormalTok{            k\_folds=k\_folds,}
\NormalTok{            patience=patience,}
\NormalTok{            optimizer=optimizer,}
\NormalTok{            sgd\_momentum=sgd\_momentum,}
\NormalTok{        )}
\NormalTok{        self.conv1 = nn.Conv2d(3, 6, 5)}
\NormalTok{        self.pool = nn.MaxPool2d(2, 2)}
\NormalTok{        self.conv2 = nn.Conv2d(6, 16, 5)}
\NormalTok{        self.fc1 = nn.Linear(16 * 5 * 5, l1)}
\NormalTok{        self.fc2 = nn.Linear(l1, l2)}
\NormalTok{        self.fc3 = nn.Linear(l2, 10)}

\NormalTok{    def forward(self, x):}
\NormalTok{        x = self.pool(F.relu(self.conv1(x)))}
\NormalTok{        x = self.pool(F.relu(self.conv2(x)))}
\NormalTok{        x = x.view({-}1, 16 * 5 * 5)}
\NormalTok{        x = F.relu(self.fc1(x))}
\NormalTok{        x = F.relu(self.fc2(x))}
\NormalTok{        x = self.fc3(x)}
\NormalTok{        return x}
\end{Highlighting}
\end{Shaded}

\hypertarget{sec-the-netcore-class-14}{%
\subsection{\texorpdfstring{The \texttt{Net\_Core}
class}{The Net\_Core class}}\label{sec-the-netcore-class-14}}

\texttt{Net\_CIFAR10} inherits from the class \texttt{Net\_Core} which
is implemented in the file \texttt{netcore.py}. It implements the
additional attributes that are common to all neural network models. The
\texttt{Net\_Core} class is implemented in the file \texttt{netcore.py}.
It implements hyperparameters as attributes, that are not used by the
\texttt{core\_model}, e.g.:

\begin{itemize}
\tightlist
\item
  optimizer (\texttt{optimizer}),
\item
  learning rate (\texttt{lr}),
\item
  batch size (\texttt{batch\_size}),
\item
  epochs (\texttt{epochs}),
\item
  k\_folds (\texttt{k\_folds}), and
\item
  early stopping criterion ``patience'' (\texttt{patience}).
\end{itemize}

Users can add further attributes to the class. The class
\texttt{Net\_Core} is shown below.

\begin{Shaded}
\begin{Highlighting}[]
\NormalTok{from torch import nn}


\NormalTok{class Net\_Core(nn.Module):}
\NormalTok{    def \_\_init\_\_(self, lr\_mult, batch\_size, epochs, k\_folds, patience,}
\NormalTok{        optimizer, sgd\_momentum):}
\NormalTok{        super(Net\_Core, self).\_\_init\_\_()}
\NormalTok{        self.lr\_mult = lr\_mult}
\NormalTok{        self.batch\_size = batch\_size}
\NormalTok{        self.epochs = epochs}
\NormalTok{        self.k\_folds = k\_folds}
\NormalTok{        self.patience = patience}
\NormalTok{        self.optimizer = optimizer}
\NormalTok{        self.sgd\_momentum = sgd\_momentum}
\end{Highlighting}
\end{Shaded}

\hypertarget{sec-comparison}{%
\subsection{Comparison of the Approach Described in the PyTorch Tutorial
With spotPython}\label{sec-comparison}}

Comparing the class \texttt{Net} from the \texttt{PyTorch} tutorial and
the class \texttt{Net\_CIFAR10} from \texttt{spotPython}, we see that
the class \texttt{Net\_CIFAR10} has additional attributes and does not
inherit from \texttt{nn} directly. It adds an additional class,
\texttt{Net\_core}, that takes care of additional attributes that are
common to all neural network models, e.g., the learning rate multiplier
\texttt{lr\_mult} or the batch size \texttt{batch\_size}.

\texttt{spotPython}'s \texttt{core\_model} implements an instance of the
\texttt{Net\_CIFAR10} class. In addition to the basic neural network
model, the \texttt{core\_model} can use these additional attributes.
\texttt{spotPython} provides methods for handling these additional
attributes to guarantee 100\% compatibility with the \texttt{PyTorch}
classes. The method \texttt{add\_core\_model\_to\_fun\_control} adds the
hyperparameters and additional attributes to the \texttt{fun\_control}
dictionary. The method is shown below.

\begin{Shaded}
\begin{Highlighting}[]
\ImportTok{from}\NormalTok{ spotPython.torch.netcifar10 }\ImportTok{import}\NormalTok{ Net\_CIFAR10}
\ImportTok{from}\NormalTok{ spotPython.data.torch\_hyper\_dict }\ImportTok{import}\NormalTok{ TorchHyperDict}
\ImportTok{from}\NormalTok{ spotPython.hyperparameters.values }\ImportTok{import}\NormalTok{ add\_core\_model\_to\_fun\_control}
\NormalTok{core\_model }\OperatorTok{=}\NormalTok{ Net\_CIFAR10}
\NormalTok{add\_core\_model\_to\_fun\_control(core\_model}\OperatorTok{=}\NormalTok{core\_model,}
\NormalTok{                              fun\_control}\OperatorTok{=}\NormalTok{fun\_control,}
\NormalTok{                              hyper\_dict}\OperatorTok{=}\NormalTok{TorchHyperDict,}
\NormalTok{                              filename}\OperatorTok{=}\VariableTok{None}\NormalTok{)}
\end{Highlighting}
\end{Shaded}

\hypertarget{sec-search-space-14}{%
\subsection{The Search Space:
Hyperparameters}\label{sec-search-space-14}}

In Section~\ref{sec-configuring-the-search-space-with-ray-tune}, we
first describe how to configure the search space with
\texttt{ray{[}tune{]}} (as shown in PyTorch (2023a)) and then how to
configure the search space with \texttt{spotPython} in -14.

\hypertarget{sec-configuring-the-search-space-with-ray-tune}{%
\subsection{Configuring the Search Space With Ray
Tune}\label{sec-configuring-the-search-space-with-ray-tune}}

Ray Tune's search space can be configured as follows (PyTorch 2023a):

\begin{Shaded}
\begin{Highlighting}[]
\NormalTok{config = \{}
\NormalTok{    "l1": tune.sample\_from(lambda \_: 2**np.random.randint(2, 9)),}
\NormalTok{    "l2": tune.sample\_from(lambda \_: 2**np.random.randint(2, 9)),}
\NormalTok{    "lr": tune.loguniform(1e{-}4, 1e{-}1),}
\NormalTok{    "batch\_size": tune.choice([2, 4, 8, 16])}
\NormalTok{\}}
\end{Highlighting}
\end{Shaded}

The \texttt{tune.sample\_from()} function enables the user to define
sample methods to obtain hyperparameters. In this example, the
\texttt{l1} and \texttt{l2} parameters should be powers of 2 between 4
and 256, so either 4, 8, 16, 32, 64, 128, or 256. The \texttt{lr}
(learning rate) should be uniformly sampled between 0.0001 and 0.1.
Lastly, the batch size is a choice between 2, 4, 8, and 16.

At each trial, \texttt{ray{[}tune{]}} will randomly sample a combination
of parameters from these search spaces. It will then train a number of
models in parallel and find the best performing one among these.
\texttt{ray{[}tune{]}} uses the \texttt{ASHAScheduler} which will
terminate bad performing trials early.

\hypertarget{sec-configuring-search-space-spotpython-14}{%
\subsection{Configuring the Search Space With
spotPython}\label{sec-configuring-search-space-spotpython-14}}

\hypertarget{the-hyper_dict-hyperparameters-for-the-selected-algorithm}{%
\subsubsection{\texorpdfstring{The \texttt{hyper\_dict} Hyperparameters
for the Selected
Algorithm}{The hyper\_dict Hyperparameters for the Selected Algorithm}}\label{the-hyper_dict-hyperparameters-for-the-selected-algorithm}}

\texttt{spotPython} uses \texttt{JSON} files for the specification of
the hyperparameters. Users can specify their individual \texttt{JSON}
files, or they can use the \texttt{JSON} files provided by
\texttt{spotPython}. The \texttt{JSON} file for the \texttt{core\_model}
is called \texttt{torch\_hyper\_dict.json}.

In contrast to \texttt{ray{[}tune{]}}, \texttt{spotPython} can handle
numerical, boolean, and categorical hyperparameters. They can be
specified in the \texttt{JSON} file in a similar way as the numerical
hyperparameters as shown below. Each entry in the \texttt{JSON} file
represents one hyperparameter with the following structure:
\texttt{type}, \texttt{default}, \texttt{transform}, \texttt{lower}, and
\texttt{upper}.

\begin{Shaded}
\begin{Highlighting}[]
\ErrorTok{"factor\_hyperparameter":} \FunctionTok{\{}
    \DataTypeTok{"levels"}\FunctionTok{:} \OtherTok{[}\StringTok{"A"}\OtherTok{,} \StringTok{"B"}\OtherTok{,} \StringTok{"C"}\OtherTok{]}\FunctionTok{,}
    \DataTypeTok{"type"}\FunctionTok{:} \StringTok{"factor"}\FunctionTok{,}
    \DataTypeTok{"default"}\FunctionTok{:} \StringTok{"B"}\FunctionTok{,}
    \DataTypeTok{"transform"}\FunctionTok{:} \StringTok{"None"}\FunctionTok{,}
    \DataTypeTok{"core\_model\_parameter\_type"}\FunctionTok{:} \StringTok{"str"}\FunctionTok{,}
    \DataTypeTok{"lower"}\FunctionTok{:} \DecValTok{0}\FunctionTok{,}
    \DataTypeTok{"upper"}\FunctionTok{:} \DecValTok{2}\FunctionTok{\}}\ErrorTok{,}
\end{Highlighting}
\end{Shaded}

The corresponding entries for the core\_model` class are shown below.

\begin{Shaded}
\begin{Highlighting}[]
\NormalTok{fun\_control[}\StringTok{\textquotesingle{}core\_model\_hyper\_dict\textquotesingle{}}\NormalTok{]}
\end{Highlighting}
\end{Shaded}

\begin{verbatim}
{'l1': {'type': 'int',
  'default': 5,
  'transform': 'transform_power_2_int',
  'lower': 2,
  'upper': 9},
 'l2': {'type': 'int',
  'default': 5,
  'transform': 'transform_power_2_int',
  'lower': 2,
  'upper': 9},
 'lr_mult': {'type': 'float',
  'default': 1.0,
  'transform': 'None',
  'lower': 0.1,
  'upper': 10.0},
 'batch_size': {'type': 'int',
  'default': 4,
  'transform': 'transform_power_2_int',
  'lower': 1,
  'upper': 4},
 'epochs': {'type': 'int',
  'default': 3,
  'transform': 'transform_power_2_int',
  'lower': 3,
  'upper': 4},
 'k_folds': {'type': 'int',
  'default': 1,
  'transform': 'None',
  'lower': 1,
  'upper': 1},
 'patience': {'type': 'int',
  'default': 5,
  'transform': 'None',
  'lower': 2,
  'upper': 10},
 'optimizer': {'levels': ['Adadelta',
   'Adagrad',
   'Adam',
   'AdamW',
   'SparseAdam',
   'Adamax',
   'ASGD',
   'NAdam',
   'RAdam',
   'RMSprop',
   'Rprop',
   'SGD'],
  'type': 'factor',
  'default': 'SGD',
  'transform': 'None',
  'class_name': 'torch.optim',
  'core_model_parameter_type': 'str',
  'lower': 0,
  'upper': 12},
 'sgd_momentum': {'type': 'float',
  'default': 0.0,
  'transform': 'None',
  'lower': 0.0,
  'upper': 1.0}}
\end{verbatim}

\hypertarget{sec-modification-of-hyperparameters-14}{%
\section{\texorpdfstring{Step 6: Modify \texttt{hyper\_dict}
Hyperparameters for the Selected Algorithm aka
\texttt{core\_model}}{Step 6: Modify hyper\_dict Hyperparameters for the Selected Algorithm aka core\_model}}\label{sec-modification-of-hyperparameters-14}}

Ray tune (PyTorch 2023a) does not provide a way to change the specified
hyperparameters without re-compilation. However, \texttt{spotPython}
provides functions for modifying the hyperparameters, their bounds and
factors as well as for activating and de-activating hyperparameters
without re-compilation of the Python source code. These functions are
described in the following.

\hypertarget{sec-modification-of-default-values}{%
\subsubsection{\texorpdfstring{Modify \texttt{hyper\_dict}
Hyperparameters for the Selected Algorithm aka
\texttt{core\_model}}{Modify hyper\_dict Hyperparameters for the Selected Algorithm aka core\_model}}\label{sec-modification-of-default-values}}

After specifying the model, the corresponding hyperparameters, their
types and bounds are loaded from the \texttt{JSON} file
\texttt{torch\_hyper\_dict.json}. After loading, the user can modify the
hyperparameters, e.g., the bounds. \texttt{spotPython} provides a simple
rule for de-activating hyperparameters: If the lower and the upper bound
are set to identical values, the hyperparameter is de-activated. This is
useful for the hyperparameter tuning, because it allows to specify a
hyperparameter in the \texttt{JSON} file, but to de-activate it in the
\texttt{fun\_control} dictionary. This is done in the next step.

\hypertarget{modify-hyperparameters-of-type-numeric-and-integer-boolean}{%
\subsubsection{Modify Hyperparameters of Type numeric and integer
(boolean)}\label{modify-hyperparameters-of-type-numeric-and-integer-boolean}}

Since the hyperparameter \texttt{k\_folds} is not used in the
\texttt{PyTorch} tutorial, it is de-activated here by setting the lower
and upper bound to the same value. Note, \texttt{k\_folds} is of type
``integer''.

\begin{Shaded}
\begin{Highlighting}[]
\ImportTok{from}\NormalTok{ spotPython.hyperparameters.values }\ImportTok{import}\NormalTok{ modify\_hyper\_parameter\_bounds}
\NormalTok{modify\_hyper\_parameter\_bounds(fun\_control, }
    \StringTok{"batch\_size"}\NormalTok{, bounds}\OperatorTok{=}\NormalTok{[}\DecValTok{1}\NormalTok{, }\DecValTok{5}\NormalTok{])}
\NormalTok{modify\_hyper\_parameter\_bounds(fun\_control, }
    \StringTok{"k\_folds"}\NormalTok{, bounds}\OperatorTok{=}\NormalTok{[}\DecValTok{0}\NormalTok{, }\DecValTok{0}\NormalTok{])}
\NormalTok{modify\_hyper\_parameter\_bounds(fun\_control, }
    \StringTok{"patience"}\NormalTok{, bounds}\OperatorTok{=}\NormalTok{[}\DecValTok{3}\NormalTok{, }\DecValTok{3}\NormalTok{])}
\end{Highlighting}
\end{Shaded}

\hypertarget{modify-hyperparameter-of-type-factor-1}{%
\subsubsection{Modify Hyperparameter of Type
factor}\label{modify-hyperparameter-of-type-factor-1}}

In a similar manner as for the numerical hyperparameters, the
categorical hyperparameters can be modified. New configurations can be
chosen by adding or deleting levels. For example, the hyperparameter
\texttt{optimizer} can be re-configured as follows:

In the following setting, two optimizers (\texttt{"SGD"} and
\texttt{"Adam"}) will be compared during the \texttt{spotPython}
hyperparameter tuning. The hyperparameter \texttt{optimizer} is active.

\begin{Shaded}
\begin{Highlighting}[]
\ImportTok{from}\NormalTok{ spotPython.hyperparameters.values }\ImportTok{import}\NormalTok{ modify\_hyper\_parameter\_levels}
\NormalTok{modify\_hyper\_parameter\_levels(fun\_control,}
     \StringTok{"optimizer"}\NormalTok{, [}\StringTok{"SGD"}\NormalTok{, }\StringTok{"Adam"}\NormalTok{])}
\end{Highlighting}
\end{Shaded}

The hyperparameter \texttt{optimizer} can be de-activated by choosing
only one value (level), here: \texttt{"SGD"}.

\begin{Shaded}
\begin{Highlighting}[]
\NormalTok{modify\_hyper\_parameter\_levels(fun\_control, }\StringTok{"optimizer"}\NormalTok{, [}\StringTok{"SGD"}\NormalTok{])}
\end{Highlighting}
\end{Shaded}

As discussed in Section~\ref{sec-optimizers-14}, there are some issues
with the LBFGS optimizer. Therefore, the usage of the LBFGS optimizer is
not deactivated in \texttt{spotPython} by default. However, the LBFGS
optimizer can be activated by adding it to the list of optimizers.
\texttt{Rprop} was removed, because it does perform very poorly (as some
pre-tests have shown). However, it can also be activated by adding it to
the list of optimizers. Since \texttt{SparseAdam} does not support dense
gradients, \texttt{Adam} was used instead. Therefore, there are 10
default optimizers:

\begin{Shaded}
\begin{Highlighting}[]
\NormalTok{modify\_hyper\_parameter\_levels(fun\_control, }\StringTok{"optimizer"}\NormalTok{,}
\NormalTok{    [}\StringTok{"Adadelta"}\NormalTok{, }\StringTok{"Adagrad"}\NormalTok{, }\StringTok{"Adam"}\NormalTok{, }\StringTok{"AdamW"}\NormalTok{, }\StringTok{"Adamax"}\NormalTok{, }\StringTok{"ASGD"}\NormalTok{, }
    \StringTok{"NAdam"}\NormalTok{, }\StringTok{"RAdam"}\NormalTok{, }\StringTok{"RMSprop"}\NormalTok{, }\StringTok{"SGD"}\NormalTok{])}
\end{Highlighting}
\end{Shaded}

\hypertarget{sec-optimizers-14}{%
\subsection{Optimizers}\label{sec-optimizers-14}}

Table~\ref{tbl-optimizers} shows some of the optimizers available in
\texttt{PyTorch}:

\(a\) denotes (0.9,0.999), \(b\) (0.5,1.2), and \(c\) (1e-6, 50),
respectively. \(R\) denotes \texttt{required,\ but\ unspecified}. ``m''
denotes \texttt{momentum}, ``w\_d'' \texttt{weight\_decay}, ``d''
\texttt{dampening}, ``n'' \texttt{nesterov}, ``r'' \texttt{rho},
``l\_s'' \texttt{learning\ rate\ for\ scaling\ delta}, ``l\_d''
\texttt{lr\_decay}, ``b'' \texttt{betas}, ``l'' \texttt{lambd}, ``a''
\texttt{alpha}, ``m\_d'' for \texttt{momentum\_decay}, ``e''
\texttt{etas}, and ``s\_s'' for \texttt{step\_sizes}.

\hypertarget{tbl-optimizers}{}
\begin{longtable}[]{@{}
  >{\raggedright\arraybackslash}p{(\columnwidth - 28\tabcolsep) * \real{0.2051}}
  >{\raggedright\arraybackslash}p{(\columnwidth - 28\tabcolsep) * \real{0.0769}}
  >{\raggedright\arraybackslash}p{(\columnwidth - 28\tabcolsep) * \real{0.0513}}
  >{\raggedright\arraybackslash}p{(\columnwidth - 28\tabcolsep) * \real{0.0769}}
  >{\raggedright\arraybackslash}p{(\columnwidth - 28\tabcolsep) * \real{0.0513}}
  >{\raggedright\arraybackslash}p{(\columnwidth - 28\tabcolsep) * \real{0.0513}}
  >{\raggedright\arraybackslash}p{(\columnwidth - 28\tabcolsep) * \real{0.0513}}
  >{\raggedright\arraybackslash}p{(\columnwidth - 28\tabcolsep) * \real{0.0513}}
  >{\raggedright\arraybackslash}p{(\columnwidth - 28\tabcolsep) * \real{0.0513}}
  >{\raggedright\arraybackslash}p{(\columnwidth - 28\tabcolsep) * \real{0.0513}}
  >{\raggedright\arraybackslash}p{(\columnwidth - 28\tabcolsep) * \real{0.0769}}
  >{\raggedright\arraybackslash}p{(\columnwidth - 28\tabcolsep) * \real{0.0513}}
  >{\raggedright\arraybackslash}p{(\columnwidth - 28\tabcolsep) * \real{0.0513}}
  >{\raggedright\arraybackslash}p{(\columnwidth - 28\tabcolsep) * \real{0.0513}}
  >{\raggedright\arraybackslash}p{(\columnwidth - 28\tabcolsep) * \real{0.0513}}@{}}
\caption{\label{tbl-optimizers}Optimizers available in PyTorch
(selection). The default values are shown in the table.}\tabularnewline
\toprule\noalign{}
\begin{minipage}[b]{\linewidth}\raggedright
Optimizer
\end{minipage} & \begin{minipage}[b]{\linewidth}\raggedright
lr
\end{minipage} & \begin{minipage}[b]{\linewidth}\raggedright
m
\end{minipage} & \begin{minipage}[b]{\linewidth}\raggedright
w\_d
\end{minipage} & \begin{minipage}[b]{\linewidth}\raggedright
d
\end{minipage} & \begin{minipage}[b]{\linewidth}\raggedright
n
\end{minipage} & \begin{minipage}[b]{\linewidth}\raggedright
r
\end{minipage} & \begin{minipage}[b]{\linewidth}\raggedright
l\_s
\end{minipage} & \begin{minipage}[b]{\linewidth}\raggedright
l\_d
\end{minipage} & \begin{minipage}[b]{\linewidth}\raggedright
b
\end{minipage} & \begin{minipage}[b]{\linewidth}\raggedright
l
\end{minipage} & \begin{minipage}[b]{\linewidth}\raggedright
a
\end{minipage} & \begin{minipage}[b]{\linewidth}\raggedright
m\_d
\end{minipage} & \begin{minipage}[b]{\linewidth}\raggedright
e
\end{minipage} & \begin{minipage}[b]{\linewidth}\raggedright
s\_s
\end{minipage} \\
\midrule\noalign{}
\endfirsthead
\toprule\noalign{}
\begin{minipage}[b]{\linewidth}\raggedright
Optimizer
\end{minipage} & \begin{minipage}[b]{\linewidth}\raggedright
lr
\end{minipage} & \begin{minipage}[b]{\linewidth}\raggedright
m
\end{minipage} & \begin{minipage}[b]{\linewidth}\raggedright
w\_d
\end{minipage} & \begin{minipage}[b]{\linewidth}\raggedright
d
\end{minipage} & \begin{minipage}[b]{\linewidth}\raggedright
n
\end{minipage} & \begin{minipage}[b]{\linewidth}\raggedright
r
\end{minipage} & \begin{minipage}[b]{\linewidth}\raggedright
l\_s
\end{minipage} & \begin{minipage}[b]{\linewidth}\raggedright
l\_d
\end{minipage} & \begin{minipage}[b]{\linewidth}\raggedright
b
\end{minipage} & \begin{minipage}[b]{\linewidth}\raggedright
l
\end{minipage} & \begin{minipage}[b]{\linewidth}\raggedright
a
\end{minipage} & \begin{minipage}[b]{\linewidth}\raggedright
m\_d
\end{minipage} & \begin{minipage}[b]{\linewidth}\raggedright
e
\end{minipage} & \begin{minipage}[b]{\linewidth}\raggedright
s\_s
\end{minipage} \\
\midrule\noalign{}
\endhead
\bottomrule\noalign{}
\endlastfoot
Adadelta & - & - & 0. & - & - & 0.9 & 1. & - & - & - & - & - & - & - \\
Adagrad & 1e-2 & - & 0. & - & - & - & - & 0. & - & - & - & - & - & - \\
Adam & 1e-3 & - & 0. & - & - & - & - & - & \(a\) & - & - & - & - & - \\
AdamW & 1e-3 & - & 1e-2 & - & - & - & - & - & \(a\) & - & - & - & - &
- \\
SparseAdam & 1e-3 & - & - & - & - & - & - & - & \(a\) & - & - & - & - &
- \\
Adamax & 2e-3 & - & 0. & - & - & - & - & - & \(a\) & - & - & - & - &
- \\
ASGD & 1e-2 & .9 & 0. & - & F & - & - & - & - & 1e-4 & .75 & - & - &
- \\
LBFGS & 1. & - & - & - & - & - & - & - & - & - & - & - & - & - \\
NAdam & 2e-3 & - & 0. & - & - & - & - & - & \(a\) & - & - & 0 & - & - \\
RAdam & 1e-3 & - & 0. & - & - & - & - & - & \(a\) & - & - & - & - & - \\
RMSprop & 1e-2 & 0. & 0. & - & - & - & - & - & \(a\) & - & - & - & - &
- \\
Rprop & 1e-2 & - & - & - & - & - & - & - & - & - & \(b\) & \(c\) & - &
- \\
SGD & \(R\) & 0. & 0. & 0. & F & - & - & - & - & - & - & - & - & - \\
\end{longtable}

\texttt{spotPython} implements an \texttt{optimization} handler that
maps the optimizer names to the corresponding \texttt{PyTorch}
optimizers.

\begin{tcolorbox}[enhanced jigsaw, left=2mm, title=\textcolor{quarto-callout-note-color}{\faInfo}\hspace{0.5em}{A note on LBFGS}, bottomrule=.15mm, titlerule=0mm, breakable, rightrule=.15mm, toprule=.15mm, coltitle=black, colbacktitle=quarto-callout-note-color!10!white, leftrule=.75mm, arc=.35mm, colframe=quarto-callout-note-color-frame, bottomtitle=1mm, colback=white, opacitybacktitle=0.6, toptitle=1mm, opacityback=0]

We recommend deactivating \texttt{PyTorch}'s LBFGS optimizer, because it
does not perform very well. The \texttt{PyTorch} documentation, see
\url{https://pytorch.org/docs/stable/generated/torch.optim.LBFGS.html\#torch.optim.LBFGS},
states:

\begin{quote}
This is a very memory intensive optimizer (it requires additional
\texttt{param\_bytes\ *\ (history\_size\ +\ 1)} bytes). If it doesn't
fit in memory try reducing the history size, or use a different
algorithm.
\end{quote}

Furthermore, the LBFGS optimizer is not compatible with the
\texttt{PyTorch} tutorial. The reason is that the LBFGS optimizer
requires the \texttt{closure} function, which is not implemented in the
\texttt{PyTorch} tutorial. Therefore, the \texttt{LBFGS} optimizer is
recommended here. Since there are ten optimizers in the portfolio, it is
not recommended tuning the hyperparameters that effect one single
optimizer only.

\end{tcolorbox}

\begin{tcolorbox}[enhanced jigsaw, left=2mm, title=\textcolor{quarto-callout-note-color}{\faInfo}\hspace{0.5em}{A note on the learning rate}, bottomrule=.15mm, titlerule=0mm, breakable, rightrule=.15mm, toprule=.15mm, coltitle=black, colbacktitle=quarto-callout-note-color!10!white, leftrule=.75mm, arc=.35mm, colframe=quarto-callout-note-color-frame, bottomtitle=1mm, colback=white, opacitybacktitle=0.6, toptitle=1mm, opacityback=0]

\texttt{spotPython} provides a multiplier for the default learning
rates, \texttt{lr\_mult}, because optimizers use different learning
rates. Using a multiplier for the learning rates might enable a
simultaneous tuning of the learning rates for all optimizers. However,
this is not recommended, because the learning rates are not comparable
across optimizers. Therefore, we recommend fixing the learning rate for
all optimizers if multiple optimizers are used. This can be done by
setting the lower and upper bounds of the learning rate multiplier to
the same value as shown below.

Thus, the learning rate, which affects the \texttt{SGD} optimizer, will
be set to a fixed value. We choose the default value of \texttt{1e-3}
for the learning rate, because it is used in other \texttt{PyTorch}
examples (it is also the default value used by \texttt{spotPython} as
defined in the \texttt{optimizer\_handler()} method). We recommend
tuning the learning rate later, when a reduced set of optimizers is
fixed. Here, we will demonstrate how to select in a screening phase the
optimizers that should be used for the hyperparameter tuning.

\end{tcolorbox}

For the same reason, we will fix the \texttt{sgd\_momentum} to
\texttt{0.9}.

\begin{Shaded}
\begin{Highlighting}[]
\NormalTok{modify\_hyper\_parameter\_bounds(fun\_control,}
    \StringTok{"lr\_mult"}\NormalTok{, bounds}\OperatorTok{=}\NormalTok{[}\FloatTok{1.0}\NormalTok{, }\FloatTok{1.0}\NormalTok{])}
\NormalTok{modify\_hyper\_parameter\_bounds(fun\_control,}
    \StringTok{"sgd\_momentum"}\NormalTok{, bounds}\OperatorTok{=}\NormalTok{[}\FloatTok{0.9}\NormalTok{, }\FloatTok{0.9}\NormalTok{])}
\end{Highlighting}
\end{Shaded}

\hypertarget{step-7-selection-of-the-objective-loss-function-1}{%
\section{Step 7: Selection of the Objective (Loss)
Function}\label{step-7-selection-of-the-objective-loss-function-1}}

\hypertarget{sec-data-splitting-14}{%
\subsection{Evaluation: Data Splitting}\label{sec-data-splitting-14}}

The evaluation procedure requires the specification of the way how the
data is split into a train and a test set and the loss function (and a
metric). As a default, \texttt{spotPython} provides a standard hold-out
data split and cross validation.

\hypertarget{hold-out-data-split}{%
\subsection{Hold-out Data Split}\label{hold-out-data-split}}

If a hold-out data split is used, the data will be partitioned into a
training, a validation, and a test data set. The split depends on the
setting of the \texttt{eval} parameter. If \texttt{eval} is set to
\texttt{train\_hold\_out}, one data set, usually the original training
data set, is split into a new training and a validation data set. The
training data set is used for training the model. The validation data
set is used for the evaluation of the hyperparameter configuration and
early stopping to prevent overfitting. In this case, the original test
data set is not used.

\begin{tcolorbox}[enhanced jigsaw, left=2mm, title=\textcolor{quarto-callout-note-color}{\faInfo}\hspace{0.5em}{Note}, bottomrule=.15mm, titlerule=0mm, breakable, rightrule=.15mm, toprule=.15mm, coltitle=black, colbacktitle=quarto-callout-note-color!10!white, leftrule=.75mm, arc=.35mm, colframe=quarto-callout-note-color-frame, bottomtitle=1mm, colback=white, opacitybacktitle=0.6, toptitle=1mm, opacityback=0]

\texttt{spotPython} returns the hyperparameters of the machine learning
and deep learning models, e.g., number of layers, learning rate, or
optimizer, but not the model weights. Therefore, after the SPOT run is
finished, the corresponding model with the optimized architecture has to
be trained again with the best hyperparameter configuration. The
training is performed on the training data set. The test data set is
used for the final evaluation of the model.

Summarizing, the following splits are performed in the hold-out setting:

\begin{enumerate}
\def\labelenumi{\arabic{enumi}.}
\tightlist
\item
  Run \texttt{spotPython} with \texttt{eval} set to
  \texttt{train\_hold\_out} to determine the best hyperparameter
  configuration.
\item
  Train the model with the best hyperparameter configuration
  (``architecture'') on the training data set:
  \texttt{train\_tuned(model\_spot,\ train,\ "model\_spot.pt")}.
\item
  Test the model on the test data:
  \texttt{test\_tuned(model\_spot,\ test,\ "model\_spot.pt")}
\end{enumerate}

These steps will be exemplified in the following sections.

\end{tcolorbox}

In addition to this \texttt{hold-out} setting, \texttt{spotPython}
provides another hold-out setting, where an explicit test data is
specified by the user that will be used as the validation set. To choose
this option, the \texttt{eval} parameter is set to
\texttt{test\_hold\_out}. In this case, the training data set is used
for the model training. Then, the explicitly defined test data set is
used for the evaluation of the hyperparameter configuration (the
validation).

\hypertarget{cross-validation}{%
\subsection{Cross-Validation}\label{cross-validation}}

The cross validation setting is used by setting the \texttt{eval}
parameter to \texttt{train\_cv} or \texttt{test\_cv}. In both cases, the
data set is split into \(k\) folds. The model is trained on \(k-1\)
folds and evaluated on the remaining fold. This is repeated \(k\) times,
so that each fold is used exactly once for evaluation. The final
evaluation is performed on the test data set. The cross validation
setting is useful for small data sets, because it allows to use all data
for training and evaluation. However, it is computationally expensive,
because the model has to be trained \(k\) times.

\begin{tcolorbox}[enhanced jigsaw, left=2mm, title=\textcolor{quarto-callout-note-color}{\faInfo}\hspace{0.5em}{Note}, bottomrule=.15mm, titlerule=0mm, breakable, rightrule=.15mm, toprule=.15mm, coltitle=black, colbacktitle=quarto-callout-note-color!10!white, leftrule=.75mm, arc=.35mm, colframe=quarto-callout-note-color-frame, bottomtitle=1mm, colback=white, opacitybacktitle=0.6, toptitle=1mm, opacityback=0]

Combinations of the above settings are possible, e.g., cross validation
can be used for training and hold-out for evaluation or \emph{vice
versa}. Also, cross validation can be used for training and testing.
Because cross validation is not used in the \texttt{PyTorch} tutorial
(PyTorch 2023a), it is not considered further here.

\end{tcolorbox}

\hypertarget{overview-of-the-evaluation-settings}{%
\subsection{Overview of the Evaluation
Settings}\label{overview-of-the-evaluation-settings}}

\hypertarget{settings-for-the-hyperparameter-tuning}{%
\subsubsection{Settings for the Hyperparameter
Tuning}\label{settings-for-the-hyperparameter-tuning}}

An overview of the training evaluations is shown in
Table~\ref{tbl-eval-settings}. \texttt{"train\_cv"} and
\texttt{"test\_cv"} use \texttt{sklearn.model\_selection.KFold()}
internally. More details on the data splitting are provided in
Section~\ref{sec-detailed-data-splitting} (in the Appendix).

\hypertarget{tbl-eval-settings}{}
\begin{longtable}[]{@{}
  >{\raggedright\arraybackslash}p{(\columnwidth - 8\tabcolsep) * \real{0.1429}}
  >{\centering\arraybackslash}p{(\columnwidth - 8\tabcolsep) * \real{0.1429}}
  >{\centering\arraybackslash}p{(\columnwidth - 8\tabcolsep) * \real{0.1429}}
  >{\raggedright\arraybackslash}p{(\columnwidth - 8\tabcolsep) * \real{0.2857}}
  >{\raggedright\arraybackslash}p{(\columnwidth - 8\tabcolsep) * \real{0.2857}}@{}}
\caption{\label{tbl-eval-settings}Overview of the evaluation
settings.}\tabularnewline
\toprule\noalign{}
\begin{minipage}[b]{\linewidth}\raggedright
\texttt{eval}
\end{minipage} & \begin{minipage}[b]{\linewidth}\centering
\texttt{train}
\end{minipage} & \begin{minipage}[b]{\linewidth}\centering
\texttt{test}
\end{minipage} & \begin{minipage}[b]{\linewidth}\raggedright
function
\end{minipage} & \begin{minipage}[b]{\linewidth}\raggedright
comment
\end{minipage} \\
\midrule\noalign{}
\endfirsthead
\toprule\noalign{}
\begin{minipage}[b]{\linewidth}\raggedright
\texttt{eval}
\end{minipage} & \begin{minipage}[b]{\linewidth}\centering
\texttt{train}
\end{minipage} & \begin{minipage}[b]{\linewidth}\centering
\texttt{test}
\end{minipage} & \begin{minipage}[b]{\linewidth}\raggedright
function
\end{minipage} & \begin{minipage}[b]{\linewidth}\raggedright
comment
\end{minipage} \\
\midrule\noalign{}
\endhead
\bottomrule\noalign{}
\endlastfoot
\texttt{"train\_hold\_out"} & \(\checkmark\) & &
\texttt{train\_one\_epoch()}, \texttt{validate\_one\_epoch()} for early
stopping & splits the \texttt{train} data set internally \\
\texttt{"test\_hold\_out"} & \(\checkmark\) & \(\checkmark\) &
\texttt{train\_one\_epoch()}, \texttt{validate\_one\_epoch()} for early
stopping & use the \texttt{test\ data\ set} for
\texttt{validate\_one\_epoch()} \\
\texttt{"train\_cv"} & \(\checkmark\) & &
\texttt{evaluate\_cv(net,\ train)} & CV using the \texttt{train} data
set \\
\texttt{"test\_cv"} & & \(\checkmark\) &
\texttt{evaluate\_cv(net,\ test)} & CV using the \texttt{test} data set
. Identical to \texttt{"train\_cv"}, uses only test data. \\
\end{longtable}

\hypertarget{settings-for-the-final-evaluation-of-the-tuned-architecture}{%
\subsubsection{Settings for the Final Evaluation of the Tuned
Architecture}\label{settings-for-the-final-evaluation-of-the-tuned-architecture}}

\hypertarget{training-of-the-tuned-architecture}{%
\paragraph{Training of the Tuned
Architecture}\label{training-of-the-tuned-architecture}}

\texttt{train\_tuned(model,\ train)}: train the model with the best
hyperparameter configuration (or simply the default) on the training
data set. It splits the \texttt{train}data into new \texttt{train} and
\texttt{validation} sets using
\texttt{create\_train\_val\_data\_loaders()}, which calls
\texttt{torch.utils.data.random\_split()} internally. Currently, 60\% of
the data is used for training and 40\% for validation. The
\texttt{train} data is used for training the model with
\texttt{train\_hold\_out()}. The \texttt{validation} data is used for
early stopping using \texttt{validate\_fold\_or\_hold\_out()} on the
\texttt{validation} data set.

\hypertarget{testing-of-the-tuned-architecture}{%
\paragraph{Testing of the Tuned
Architecture}\label{testing-of-the-tuned-architecture}}

\texttt{test\_tuned(model,\ test)}: test the model on the test data set.
No data splitting is performed. The (trained) model is evaluated using
the \texttt{validate\_fold\_or\_hold\_out()} function. Note: During
training, \texttt{"shuffle"} is set to \texttt{True}, whereas during
testing, \texttt{"shuffle"} is set to \texttt{False}.

Section~\ref{sec-final-model-evaluation} describes the final evaluation
of the tuned architecture.

\begin{Shaded}
\begin{Highlighting}[]
\NormalTok{fun\_control.update(\{}
    \StringTok{"eval"}\NormalTok{: }\StringTok{"train\_hold\_out"}\NormalTok{,}
    \StringTok{"path"}\NormalTok{: }\StringTok{"torch\_model.pt"}\NormalTok{,}
    \StringTok{"shuffle"}\NormalTok{: }\VariableTok{True}\NormalTok{\})}
\end{Highlighting}
\end{Shaded}

\hypertarget{sec-loss-functions-14}{%
\subsection{Evaluation: Loss Functions and
Metrics}\label{sec-loss-functions-14}}

The key \texttt{"loss\_function"} specifies the loss function which is
used during the optimization. There are several different loss functions
under \texttt{PyTorch}'s \texttt{nn} package. For example, a simple loss
is \texttt{MSELoss}, which computes the mean-squared error between the
output and the target. In this tutorial we will use
\texttt{CrossEntropyLoss}, because it is also used in the
\texttt{PyTorch} tutorial.

\begin{Shaded}
\begin{Highlighting}[]
\ImportTok{from}\NormalTok{ torch.nn }\ImportTok{import}\NormalTok{ CrossEntropyLoss}
\NormalTok{loss\_function }\OperatorTok{=}\NormalTok{ CrossEntropyLoss()}
\NormalTok{fun\_control.update(\{}\StringTok{"loss\_function"}\NormalTok{: loss\_function\})}
\end{Highlighting}
\end{Shaded}

In addition to the loss functions, \texttt{spotPython} provides access
to a large number of metrics.

\begin{itemize}
\tightlist
\item
  The key \texttt{"metric\_sklearn"} is used for metrics that follow the
  \texttt{scikit-learn} conventions.
\item
  The key \texttt{"river\_metric"} is used for the river based
  evaluation (Montiel et al. 2021) via
  \texttt{eval\_oml\_iter\_progressive}, and
\item
  the key \texttt{"metric\_torch"} is used for the metrics from
  \texttt{TorchMetrics}.
\end{itemize}

\texttt{TorchMetrics} is a collection of more than 90 PyTorch metrics,
see \url{https://torchmetrics.readthedocs.io/en/latest/}. Because the
\texttt{PyTorch} tutorial uses the accuracy as metric, we use the same
metric here. Currently, accuracy is computed in the tutorial's example
code. We will use \texttt{TorchMetrics} instead, because it offers more
flexibilty, e.g., it can be used for regression and classification.
Furthermore, \texttt{TorchMetrics} offers the following advantages:

\begin{verbatim}
* A standardized interface to increase reproducibility
* Reduces Boilerplate
* Distributed-training compatible
* Rigorously tested
* Automatic accumulation over batches
* Automatic synchronization between multiple devices
\end{verbatim}

Therefore, we set

\begin{Shaded}
\begin{Highlighting}[]
\ImportTok{import}\NormalTok{ torchmetrics}
\NormalTok{metric\_torch }\OperatorTok{=}\NormalTok{ torchmetrics.Accuracy(task}\OperatorTok{=}\StringTok{"multiclass"}\NormalTok{, num\_classes}\OperatorTok{=}\DecValTok{10}\NormalTok{).to(fun\_control[}\StringTok{"device"}\NormalTok{])}
\NormalTok{fun\_control.update(\{}\StringTok{"metric\_torch"}\NormalTok{: metric\_torch\})}
\end{Highlighting}
\end{Shaded}

\hypertarget{step-8-calling-the-spot-function-1}{%
\section{Step 8: Calling the SPOT
Function}\label{step-8-calling-the-spot-function-1}}

\hypertarget{sec-prepare-spot-call-14}{%
\subsection{Preparing the SPOT Call}\label{sec-prepare-spot-call-14}}

The following code passes the information about the parameter ranges and
bounds to \texttt{spot}.

\begin{Shaded}
\begin{Highlighting}[]
\ImportTok{from}\NormalTok{ spotPython.hyperparameters.values }\ImportTok{import}\NormalTok{ (}
\NormalTok{    get\_var\_type,}
\NormalTok{    get\_var\_name,}
\NormalTok{    get\_bound\_values}
\NormalTok{    )}
\NormalTok{var\_type }\OperatorTok{=}\NormalTok{ get\_var\_type(fun\_control)}
\NormalTok{var\_name }\OperatorTok{=}\NormalTok{ get\_var\_name(fun\_control)}

\NormalTok{lower }\OperatorTok{=}\NormalTok{ get\_bound\_values(fun\_control, }\StringTok{"lower"}\NormalTok{)}
\NormalTok{upper }\OperatorTok{=}\NormalTok{ get\_bound\_values(fun\_control, }\StringTok{"upper"}\NormalTok{)}
\end{Highlighting}
\end{Shaded}

Now, the dictionary \texttt{fun\_control} contains all information
needed for the hyperparameter tuning. Before the hyperparameter tuning
is started, it is recommended to take a look at the experimental design.
The method \texttt{gen\_design\_table} generates a design table as
follows:

\begin{Shaded}
\begin{Highlighting}[]
\ImportTok{from}\NormalTok{ spotPython.utils.eda }\ImportTok{import}\NormalTok{ gen\_design\_table}
\BuiltInTok{print}\NormalTok{(gen\_design\_table(fun\_control))}
\end{Highlighting}
\end{Shaded}

\begin{verbatim}
| name         | type   | default   |   lower |   upper | transform             |
|--------------|--------|-----------|---------|---------|-----------------------|
| l1           | int    | 5         |     2   |     9   | transform_power_2_int |
| l2           | int    | 5         |     2   |     9   | transform_power_2_int |
| lr_mult      | float  | 1.0       |     1   |     1   | None                  |
| batch_size   | int    | 4         |     1   |     5   | transform_power_2_int |
| epochs       | int    | 3         |     3   |     4   | transform_power_2_int |
| k_folds      | int    | 1         |     0   |     0   | None                  |
| patience     | int    | 5         |     3   |     3   | None                  |
| optimizer    | factor | SGD       |     0   |     9   | None                  |
| sgd_momentum | float  | 0.0       |     0.9 |     0.9 | None                  |
\end{verbatim}

This allows to check if all information is available and if the
information is correct. \textbf{?@tbl-design} shows the experimental
design for the hyperparameter tuning. The table shows the
hyperparameters, their types, default values, lower and upper bounds,
and the transformation function. The transformation function is used to
transform the hyperparameter values from the unit hypercube to the
original domain. The transformation function is applied to the
hyperparameter values before the evaluation of the objective function.
Hyperparameter transformations are shown in the column ``transform'',
e.g., the \texttt{l1} default is \texttt{5}, which results in the value
\(2^5 = 32\) for the network, because the transformation
\texttt{transform\_power\_2\_int} was selected in the \texttt{JSON}
file. The default value of the \texttt{batch\_size} is set to
\texttt{4}, which results in a batch size of \(2^4 = 16\).

\hypertarget{sec-the-objective-function-14}{%
\subsection{\texorpdfstring{The Objective Function
\texttt{fun\_torch}}{The Objective Function fun\_torch}}\label{sec-the-objective-function-14}}

The objective function \texttt{fun\_torch} is selected next. It
implements an interface from \texttt{PyTorch}'s training, validation,
and testing methods to \texttt{spotPython}.

\begin{Shaded}
\begin{Highlighting}[]
\ImportTok{from}\NormalTok{ spotPython.fun.hypertorch }\ImportTok{import}\NormalTok{ HyperTorch}
\NormalTok{fun }\OperatorTok{=}\NormalTok{ HyperTorch().fun\_torch}
\end{Highlighting}
\end{Shaded}

\hypertarget{sec-default-hyperparameters}{%
\subsection{Using Default Hyperparameters or Results from Previous
Runs}\label{sec-default-hyperparameters}}

We add the default setting to the initial design:

\begin{Shaded}
\begin{Highlighting}[]
\ImportTok{from}\NormalTok{ spotPython.hyperparameters.values }\ImportTok{import}\NormalTok{ get\_default\_hyperparameters\_as\_array}
\NormalTok{X\_start }\OperatorTok{=}\NormalTok{ get\_default\_hyperparameters\_as\_array(fun\_control)}
\end{Highlighting}
\end{Shaded}

\hypertarget{sec-call-the-hyperparameter-tuner-14}{%
\subsection{Starting the Hyperparameter
Tuning}\label{sec-call-the-hyperparameter-tuner-14}}

The \texttt{spotPython} hyperparameter tuning is started by calling the
\texttt{Spot} function. Here, we will run the tuner for approximately 30
minutes (\texttt{max\_time}). Note: the initial design is always
evaluated in the \texttt{spotPython} run. As a consequence, the run may
take longer than specified by \texttt{max\_time}, because the evaluation
time of initial design (here: \texttt{init\_size}, 10 points) is
performed independently of \texttt{max\_time}. During the run, results
from the training is shown. These results can be visualized with
Tensorboard as will be shown in Section~\ref{sec-tensorboard-14}.

\begin{Shaded}
\begin{Highlighting}[]
\ImportTok{from}\NormalTok{ spotPython.spot }\ImportTok{import}\NormalTok{ spot}
\ImportTok{from}\NormalTok{ math }\ImportTok{import}\NormalTok{ inf}
\ImportTok{import}\NormalTok{ numpy }\ImportTok{as}\NormalTok{ np}
\NormalTok{spot\_tuner }\OperatorTok{=}\NormalTok{ spot.Spot(fun}\OperatorTok{=}\NormalTok{fun,}
\NormalTok{                   lower }\OperatorTok{=}\NormalTok{ lower,}
\NormalTok{                   upper }\OperatorTok{=}\NormalTok{ upper,}
\NormalTok{                   fun\_evals }\OperatorTok{=}\NormalTok{ inf,}
\NormalTok{                   max\_time }\OperatorTok{=}\NormalTok{ MAX\_TIME,}
\NormalTok{                   tolerance\_x }\OperatorTok{=}\NormalTok{ np.sqrt(np.spacing(}\DecValTok{1}\NormalTok{)),}
\NormalTok{                   var\_type }\OperatorTok{=}\NormalTok{ var\_type,}
\NormalTok{                   var\_name }\OperatorTok{=}\NormalTok{ var\_name,}
\NormalTok{                   show\_progress}\OperatorTok{=} \VariableTok{True}\NormalTok{,}
\NormalTok{                   fun\_control }\OperatorTok{=}\NormalTok{ fun\_control,}
\NormalTok{                   design\_control}\OperatorTok{=}\NormalTok{\{}\StringTok{"init\_size"}\NormalTok{: INIT\_SIZE\},}
\NormalTok{                   surrogate\_control}\OperatorTok{=}\NormalTok{\{}\StringTok{"noise"}\NormalTok{: }\VariableTok{True}\NormalTok{,}
                                      \StringTok{"cod\_type"}\NormalTok{: }\StringTok{"norm"}\NormalTok{,}
                                      \StringTok{"min\_theta"}\NormalTok{: }\OperatorTok{{-}}\DecValTok{4}\NormalTok{,}
                                      \StringTok{"max\_theta"}\NormalTok{: }\DecValTok{3}\NormalTok{,}
                                      \StringTok{"n\_theta"}\NormalTok{: }\BuiltInTok{len}\NormalTok{(var\_name),}
                                      \StringTok{"model\_fun\_evals"}\NormalTok{: }\DecValTok{10\_000}
\NormalTok{                                      \})}
\NormalTok{spot\_tuner.run(X\_start}\OperatorTok{=}\NormalTok{X\_start)}
\end{Highlighting}
\end{Shaded}

\begin{verbatim}

config: {'l1': 128, 'l2': 8, 'lr_mult': 1.0, 'batch_size': 32, 'epochs': 16, 'k_folds': 0, 'patience': 3, 'optimizer': 'AdamW', 'sgd_momentum': 0.9}
Epoch: 1 | 
\end{verbatim}

\begin{verbatim}
MulticlassAccuracy: 0.3889499902725220 | Loss: 1.6403590366363525 | Acc: 0.3889500000000000.
Epoch: 2 | 
\end{verbatim}

\begin{verbatim}
MulticlassAccuracy: 0.4578999876976013 | Loss: 1.4816969134330749 | Acc: 0.4579000000000000.
Epoch: 3 | 
\end{verbatim}

\begin{verbatim}
MulticlassAccuracy: 0.4945999979972839 | Loss: 1.3767625138282775 | Acc: 0.4946000000000000.
Epoch: 4 | 
\end{verbatim}

\begin{verbatim}
MulticlassAccuracy: 0.5118499994277954 | Loss: 1.3446329971313478 | Acc: 0.5118500000000000.
Epoch: 5 | 
\end{verbatim}

\begin{verbatim}
MulticlassAccuracy: 0.5447499752044678 | Loss: 1.2767737101554870 | Acc: 0.5447500000000000.
Epoch: 6 | 
\end{verbatim}

\begin{verbatim}
MulticlassAccuracy: 0.5664499998092651 | Loss: 1.2234437763214112 | Acc: 0.5664500000000000.
Epoch: 7 | 
\end{verbatim}

\begin{verbatim}
MulticlassAccuracy: 0.5648499727249146 | Loss: 1.2325385323524476 | Acc: 0.5648500000000000.
Epoch: 8 | 
\end{verbatim}

\begin{verbatim}
MulticlassAccuracy: 0.5896499752998352 | Loss: 1.1611093239784240 | Acc: 0.5896500000000000.
Epoch: 9 | 
\end{verbatim}

\begin{verbatim}
MulticlassAccuracy: 0.6015999913215637 | Loss: 1.1370150957107543 | Acc: 0.6016000000000000.
Epoch: 10 | 
\end{verbatim}

\begin{verbatim}
MulticlassAccuracy: 0.6074000000953674 | Loss: 1.1378371593475343 | Acc: 0.6074000000000001.
Epoch: 11 | 
\end{verbatim}

\begin{verbatim}
MulticlassAccuracy: 0.6036999821662903 | Loss: 1.1592556796073914 | Acc: 0.6037000000000000.
Epoch: 12 | 
\end{verbatim}

\begin{verbatim}
MulticlassAccuracy: 0.5997499823570251 | Loss: 1.1987680685997009 | Acc: 0.5997500000000000.
Early stopping at epoch 11
Returned to Spot: Validation loss: 1.1987680685997009

config: {'l1': 16, 'l2': 16, 'lr_mult': 1.0, 'batch_size': 8, 'epochs': 8, 'k_folds': 0, 'patience': 3, 'optimizer': 'NAdam', 'sgd_momentum': 0.9}
Epoch: 1 | 
\end{verbatim}

\begin{verbatim}
MulticlassAccuracy: 0.3920499980449677 | Loss: 1.6102165319681168 | Acc: 0.3920500000000000.
Epoch: 2 | 
\end{verbatim}

\begin{verbatim}
MulticlassAccuracy: 0.4390000104904175 | Loss: 1.5077767979741097 | Acc: 0.4390000000000000.
Epoch: 3 | 
\end{verbatim}

\begin{verbatim}
MulticlassAccuracy: 0.4700999855995178 | Loss: 1.4581756867766380 | Acc: 0.4701000000000000.
Epoch: 4 | 
\end{verbatim}

\begin{verbatim}
MulticlassAccuracy: 0.4981499910354614 | Loss: 1.3969129746913911 | Acc: 0.4981500000000000.
Epoch: 5 | 
\end{verbatim}

\begin{verbatim}
MulticlassAccuracy: 0.5059000253677368 | Loss: 1.3693460956692696 | Acc: 0.5059000000000000.
Epoch: 6 | 
\end{verbatim}

\begin{verbatim}
MulticlassAccuracy: 0.5133500099182129 | Loss: 1.3540988440275192 | Acc: 0.5133500000000000.
Epoch: 7 | 
\end{verbatim}

\begin{verbatim}
MulticlassAccuracy: 0.5081499814987183 | Loss: 1.3817692994177342 | Acc: 0.5081500000000000.
Epoch: 8 | 
\end{verbatim}

\begin{verbatim}
MulticlassAccuracy: 0.5159500241279602 | Loss: 1.3653468480706215 | Acc: 0.5159500000000000.
Returned to Spot: Validation loss: 1.3653468480706215

config: {'l1': 256, 'l2': 128, 'lr_mult': 1.0, 'batch_size': 2, 'epochs': 16, 'k_folds': 0, 'patience': 3, 'optimizer': 'RMSprop', 'sgd_momentum': 0.9}
Epoch: 1 | 
\end{verbatim}

\begin{verbatim}
MulticlassAccuracy: 0.0958499982953072 | Loss: 2.3086834851264952 | Acc: 0.0958500000000000.
Epoch: 2 | 
\end{verbatim}

\begin{verbatim}
MulticlassAccuracy: 0.0987000018358231 | Loss: 2.3107500833988190 | Acc: 0.0987000000000000.
Epoch: 3 | 
\end{verbatim}

\begin{verbatim}
MulticlassAccuracy: 0.0958499982953072 | Loss: 2.3054559610605239 | Acc: 0.0958500000000000.
Epoch: 4 | 
\end{verbatim}

\begin{verbatim}
MulticlassAccuracy: 0.1013000011444092 | Loss: 2.3091404678583145 | Acc: 0.1013000000000000.
Epoch: 5 | 
\end{verbatim}

\begin{verbatim}
MulticlassAccuracy: 0.0958499982953072 | Loss: 2.3109533527135850 | Acc: 0.0958500000000000.
Epoch: 6 | 
\end{verbatim}

\begin{verbatim}
MulticlassAccuracy: 0.0987000018358231 | Loss: 2.3080133529186249 | Acc: 0.0987000000000000.
Early stopping at epoch 5
Returned to Spot: Validation loss: 2.308013352918625

config: {'l1': 8, 'l2': 32, 'lr_mult': 1.0, 'batch_size': 4, 'epochs': 8, 'k_folds': 0, 'patience': 3, 'optimizer': 'Adamax', 'sgd_momentum': 0.9}
Epoch: 1 | 
\end{verbatim}

\begin{verbatim}
MulticlassAccuracy: 0.3910000026226044 | Loss: 1.6194829273104667 | Acc: 0.3910000000000000.
Epoch: 2 | 
\end{verbatim}

\begin{verbatim}
MulticlassAccuracy: 0.4532499909400940 | Loss: 1.5181912495672703 | Acc: 0.4532500000000000.
Epoch: 3 | 
\end{verbatim}

\begin{verbatim}
MulticlassAccuracy: 0.5023999810218811 | Loss: 1.3594324642419815 | Acc: 0.5024000000000000.
Epoch: 4 | 
\end{verbatim}

\begin{verbatim}
MulticlassAccuracy: 0.5066999793052673 | Loss: 1.3639220094040037 | Acc: 0.5067000000000000.
Epoch: 5 | 
\end{verbatim}

\begin{verbatim}
MulticlassAccuracy: 0.5313000082969666 | Loss: 1.3084210138827563 | Acc: 0.5313000000000000.
Epoch: 6 | 
\end{verbatim}

\begin{verbatim}
MulticlassAccuracy: 0.5376499891281128 | Loss: 1.3020537653062492 | Acc: 0.5376500000000000.
Epoch: 7 | 
\end{verbatim}

\begin{verbatim}
MulticlassAccuracy: 0.5404999852180481 | Loss: 1.2979997927054763 | Acc: 0.5405000000000000.
Epoch: 8 | 
\end{verbatim}

\begin{verbatim}
MulticlassAccuracy: 0.5505999922752380 | Loss: 1.2794678398683668 | Acc: 0.5506000000000000.
Returned to Spot: Validation loss: 1.2794678398683668

config: {'l1': 64, 'l2': 512, 'lr_mult': 1.0, 'batch_size': 16, 'epochs': 16, 'k_folds': 0, 'patience': 3, 'optimizer': 'Adagrad', 'sgd_momentum': 0.9}
Epoch: 1 | 
\end{verbatim}

\begin{verbatim}
MulticlassAccuracy: 0.4688499867916107 | Loss: 1.4396714681148528 | Acc: 0.4688500000000000.
Epoch: 2 | 
\end{verbatim}

\begin{verbatim}
MulticlassAccuracy: 0.4978500008583069 | Loss: 1.3743870592117309 | Acc: 0.4978500000000000.
Epoch: 3 | 
\end{verbatim}

\begin{verbatim}
MulticlassAccuracy: 0.5149000287055969 | Loss: 1.3301207626819611 | Acc: 0.5149000000000000.
Epoch: 4 | 
\end{verbatim}

\begin{verbatim}
MulticlassAccuracy: 0.5352500081062317 | Loss: 1.2803554334163665 | Acc: 0.5352500000000000.
Epoch: 5 | 
\end{verbatim}

\begin{verbatim}
MulticlassAccuracy: 0.5407999753952026 | Loss: 1.2673199267387389 | Acc: 0.5407999999999999.
Epoch: 6 | 
\end{verbatim}

\begin{verbatim}
MulticlassAccuracy: 0.5474500060081482 | Loss: 1.2426155496835709 | Acc: 0.5474500000000000.
Epoch: 7 | 
\end{verbatim}

\begin{verbatim}
MulticlassAccuracy: 0.5532000064849854 | Loss: 1.2252585200309754 | Acc: 0.5532000000000000.
Epoch: 8 | 
\end{verbatim}

\begin{verbatim}
MulticlassAccuracy: 0.5598499774932861 | Loss: 1.2217366221427917 | Acc: 0.5598500000000000.
Epoch: 9 | 
\end{verbatim}

\begin{verbatim}
MulticlassAccuracy: 0.5702000260353088 | Loss: 1.2027698907375335 | Acc: 0.5702000000000000.
Epoch: 10 | 
\end{verbatim}

\begin{verbatim}
MulticlassAccuracy: 0.5695499777793884 | Loss: 1.1946598905563355 | Acc: 0.5695500000000000.
Epoch: 11 | 
\end{verbatim}

\begin{verbatim}
MulticlassAccuracy: 0.5720999836921692 | Loss: 1.1931119963169099 | Acc: 0.5721000000000001.
Epoch: 12 | 
\end{verbatim}

\begin{verbatim}
MulticlassAccuracy: 0.5777500271797180 | Loss: 1.1757407437086105 | Acc: 0.5777500000000000.
Epoch: 13 | 
\end{verbatim}

\begin{verbatim}
MulticlassAccuracy: 0.5833500027656555 | Loss: 1.1655059050798415 | Acc: 0.5833500000000000.
Epoch: 14 | 
\end{verbatim}

\begin{verbatim}
MulticlassAccuracy: 0.5854499936103821 | Loss: 1.1665637883186339 | Acc: 0.5854500000000000.
Epoch: 15 | 
\end{verbatim}

\begin{verbatim}
MulticlassAccuracy: 0.5885499715805054 | Loss: 1.1581050729990006 | Acc: 0.5885500000000000.
Epoch: 16 | 
\end{verbatim}

\begin{verbatim}
MulticlassAccuracy: 0.5877500176429749 | Loss: 1.1598053013563157 | Acc: 0.5877500000000000.
Returned to Spot: Validation loss: 1.1598053013563157
\end{verbatim}

\begin{verbatim}

config: {'l1': 64, 'l2': 256, 'lr_mult': 1.0, 'batch_size': 16, 'epochs': 16, 'k_folds': 0, 'patience': 3, 'optimizer': 'Adagrad', 'sgd_momentum': 0.9}
Epoch: 1 | 
\end{verbatim}

\begin{verbatim}
MulticlassAccuracy: 0.4435999989509583 | Loss: 1.5161994444847107 | Acc: 0.4436000000000000.
Epoch: 2 | 
\end{verbatim}

\begin{verbatim}
MulticlassAccuracy: 0.4676499962806702 | Loss: 1.4507200250148773 | Acc: 0.4676500000000000.
Epoch: 3 | 
\end{verbatim}

\begin{verbatim}
MulticlassAccuracy: 0.4885500073432922 | Loss: 1.4064176963806152 | Acc: 0.4885500000000000.
Epoch: 4 | 
\end{verbatim}

\begin{verbatim}
MulticlassAccuracy: 0.4984500110149384 | Loss: 1.3765785826206207 | Acc: 0.4984500000000000.
Epoch: 5 | 
\end{verbatim}

\begin{verbatim}
MulticlassAccuracy: 0.5091999769210815 | Loss: 1.3492139563083649 | Acc: 0.5092000000000000.
Epoch: 6 | 
\end{verbatim}

\begin{verbatim}
MulticlassAccuracy: 0.5235000252723694 | Loss: 1.3260424315452575 | Acc: 0.5235000000000000.
Epoch: 7 | 
\end{verbatim}

\begin{verbatim}
MulticlassAccuracy: 0.5347999930381775 | Loss: 1.2992566047668457 | Acc: 0.5348000000000001.
Epoch: 8 | 
\end{verbatim}

\begin{verbatim}
MulticlassAccuracy: 0.5384500026702881 | Loss: 1.2924042490005494 | Acc: 0.5384500000000000.
Epoch: 9 | 
\end{verbatim}

\begin{verbatim}
MulticlassAccuracy: 0.5433999896049500 | Loss: 1.2770100817918777 | Acc: 0.5434000000000000.
Epoch: 10 | 
\end{verbatim}

\begin{verbatim}
MulticlassAccuracy: 0.5457999706268311 | Loss: 1.2646812784671784 | Acc: 0.5458000000000000.
Epoch: 11 | 
\end{verbatim}

\begin{verbatim}
MulticlassAccuracy: 0.5486000180244446 | Loss: 1.2627830792903900 | Acc: 0.5486000000000000.
Epoch: 12 | 
\end{verbatim}

\begin{verbatim}
MulticlassAccuracy: 0.5608000159263611 | Loss: 1.2396654787063599 | Acc: 0.5608000000000000.
Epoch: 13 | 
\end{verbatim}

\begin{verbatim}
MulticlassAccuracy: 0.5554000139236450 | Loss: 1.2407209475994110 | Acc: 0.5554000000000000.
Epoch: 14 | 
\end{verbatim}

\begin{verbatim}
MulticlassAccuracy: 0.5677000284194946 | Loss: 1.2263578844547272 | Acc: 0.5677000000000000.
Epoch: 15 | 
\end{verbatim}

\begin{verbatim}
MulticlassAccuracy: 0.5665000081062317 | Loss: 1.2272662802696228 | Acc: 0.5665000000000000.
Epoch: 16 | 
\end{verbatim}

\begin{verbatim}
MulticlassAccuracy: 0.5688999891281128 | Loss: 1.2138021411895752 | Acc: 0.5689000000000000.
Returned to Spot: Validation loss: 1.2138021411895752
spotPython tuning: 1.1598053013563157 [##########] 100.00% Done...
\end{verbatim}

\begin{verbatim}
<spotPython.spot.spot.Spot at 0x1553f78e0>
\end{verbatim}

\hypertarget{sec-tensorboard-14}{%
\section{Step 9: Tensorboard}\label{sec-tensorboard-14}}

The textual output shown in the console (or code cell) can be visualized
with Tensorboard.

\hypertarget{tensorboard-start-tensorboard}{%
\subsection{Tensorboard: Start
Tensorboard}\label{tensorboard-start-tensorboard}}

Start TensorBoard through the command line to visualize data you logged.
Specify the root log directory as used in
\texttt{fun\_control\ =\ fun\_control\_init(task="regression",\ tensorboard\_path="runs/24\_spot\_torch\_regression")}
as the \texttt{tensorboard\_path}. The argument logdir points to
directory where TensorBoard will look to find event files that it can
display. TensorBoard will recursively walk the directory structure
rooted at logdir, looking for .\emph{tfevents.} files.

\begin{Shaded}
\begin{Highlighting}[]
\NormalTok{tensorboard {-}{-}logdir=runs}
\end{Highlighting}
\end{Shaded}

Go to the URL it provides or to \url{http://localhost:6006/}. The
following figures show some screenshots of Tensorboard.

\begin{figure}

{\centering \includegraphics{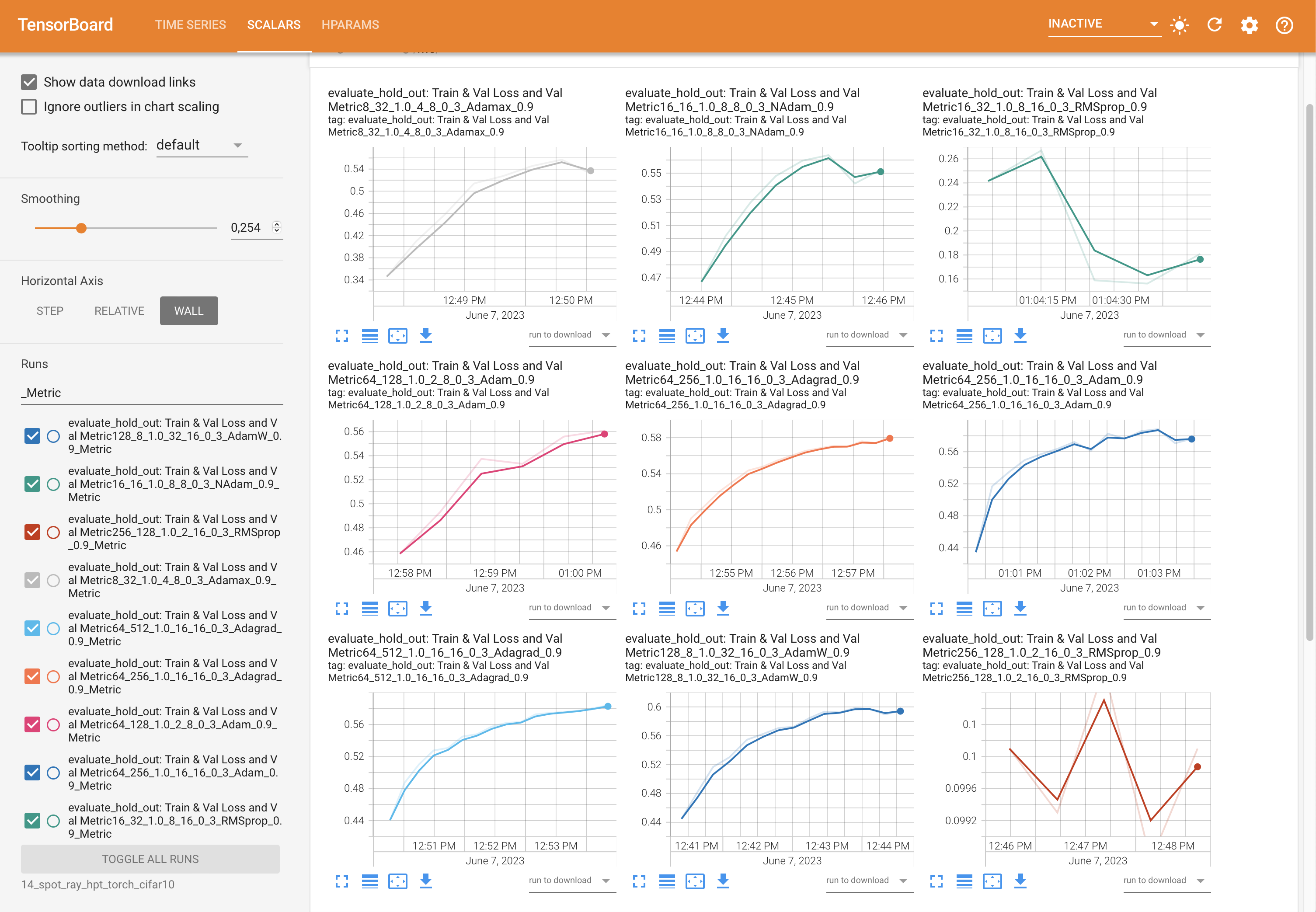}

}

\caption{\label{fig-tensorboard_0}Tensorboard}

\end{figure}

\begin{figure}

{\centering \includegraphics{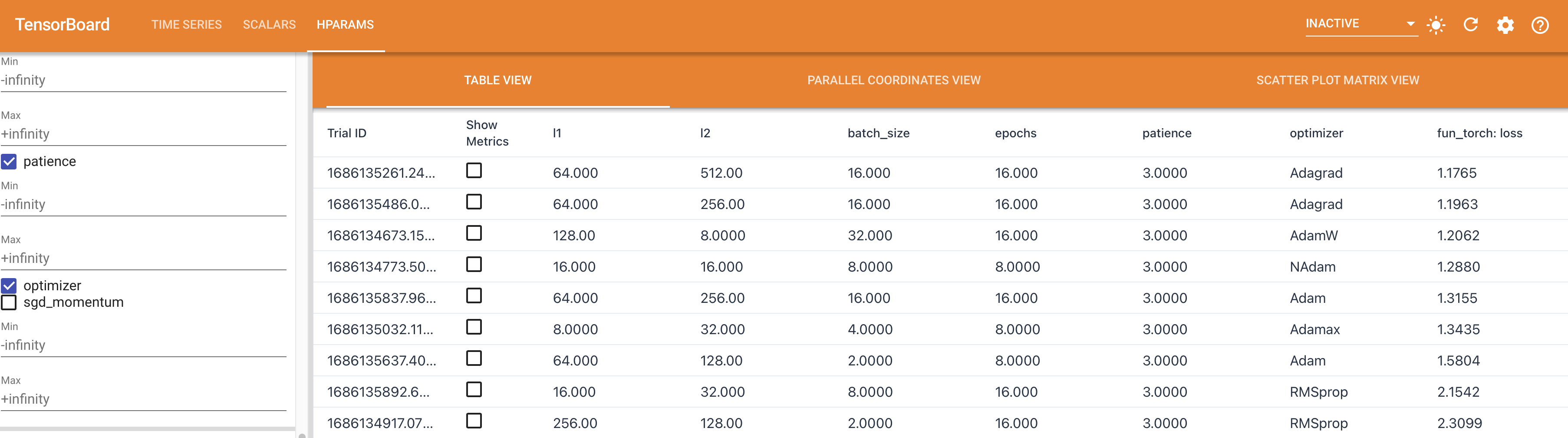}

}

\caption{\label{fig-tensorboard_hdparams}Tensorboard}

\end{figure}

\hypertarget{sec-saving-the-state-of-the-notebook}{%
\subsection{Saving the State of the
Notebook}\label{sec-saving-the-state-of-the-notebook}}

The state of the notebook can be saved and reloaded as follows:

\begin{Shaded}
\begin{Highlighting}[]
\ImportTok{import}\NormalTok{ pickle}
\NormalTok{SAVE }\OperatorTok{=} \VariableTok{False}
\NormalTok{LOAD }\OperatorTok{=} \VariableTok{False}

\ControlFlowTok{if}\NormalTok{ SAVE:}
\NormalTok{    result\_file\_name }\OperatorTok{=} \StringTok{"res\_"} \OperatorTok{+}\NormalTok{ experiment\_name }\OperatorTok{+} \StringTok{".pkl"}
    \ControlFlowTok{with} \BuiltInTok{open}\NormalTok{(result\_file\_name, }\StringTok{\textquotesingle{}wb\textquotesingle{}}\NormalTok{) }\ImportTok{as}\NormalTok{ f:}
\NormalTok{        pickle.dump(spot\_tuner, f)}

\ControlFlowTok{if}\NormalTok{ LOAD:}
\NormalTok{    result\_file\_name }\OperatorTok{=} \StringTok{"add\_the\_name\_of\_the\_result\_file\_here.pkl"}
    \ControlFlowTok{with} \BuiltInTok{open}\NormalTok{(result\_file\_name, }\StringTok{\textquotesingle{}rb\textquotesingle{}}\NormalTok{) }\ImportTok{as}\NormalTok{ f:}
\NormalTok{        spot\_tuner }\OperatorTok{=}\NormalTok{  pickle.load(f)}
\end{Highlighting}
\end{Shaded}

\hypertarget{sec-results-14}{%
\section{Step 10: Results}\label{sec-results-14}}

After the hyperparameter tuning run is finished, the progress of the
hyperparameter tuning can be visualized. The following code generates
the progress plot from \textbf{?@fig-progress}.

\begin{Shaded}
\begin{Highlighting}[]
\NormalTok{spot\_tuner.plot\_progress(log\_y}\OperatorTok{=}\VariableTok{False}\NormalTok{, }
\NormalTok{    filename}\OperatorTok{=}\StringTok{"./figures/"} \OperatorTok{+}\NormalTok{ experiment\_name}\OperatorTok{+}\StringTok{"\_progress.png"}\NormalTok{)}
\end{Highlighting}
\end{Shaded}

\begin{figure}[H]

{\centering \includegraphics{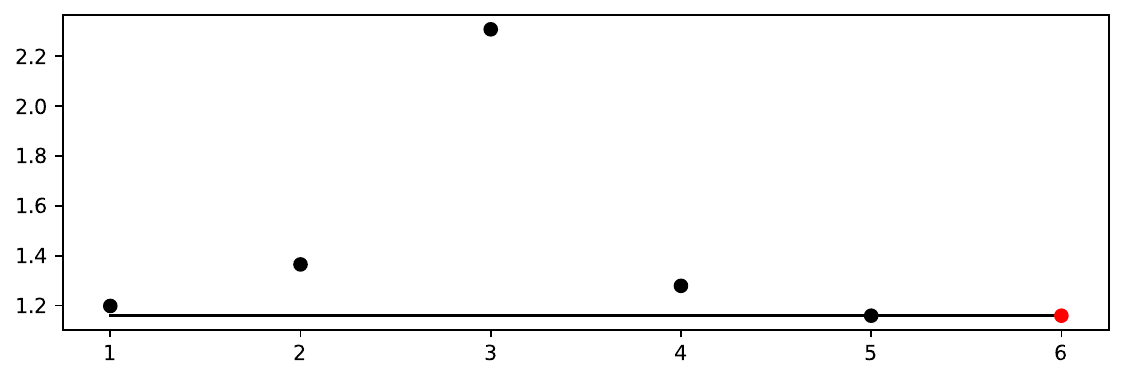}

}

\caption{Progress plot. \emph{Black} dots denote results from the
initial design. \emph{Red} dots illustrate the improvement found by the
surrogate model based optimization.}

\end{figure}

\textbf{?@fig-progress} shows a typical behaviour that can be observed
in many hyperparameter studies (Bartz et al. 2022): the largest
improvement is obtained during the evaluation of the initial design. The
surrogate model based optimization-optimization with the surrogate
refines the results. \textbf{?@fig-progress} also illustrates one major
difference between \texttt{ray{[}tune{]}} as used in PyTorch (2023a) and
\texttt{spotPython}: the \texttt{ray{[}tune{]}} uses a random search and
will generate results similar to the \emph{black} dots, whereas
\texttt{spotPython} uses a surrogate model based optimization and
presents results represented by \emph{red} dots in
\textbf{?@fig-progress}. The surrogate model based optimization is
considered to be more efficient than a random search, because the
surrogate model guides the search towards promising regions in the
hyperparameter space.

In addition to the improved (``optimized'') hyperparameter values,
\texttt{spotPython} allows a statistical analysis, e.g., a sensitivity
analysis, of the results. We can print the results of the hyperparameter
tuning, see \textbf{?@tbl-results}. The table shows the hyperparameters,
their types, default values, lower and upper bounds, and the
transformation function. The column ``tuned'' shows the tuned values.
The column ``importance'' shows the importance of the hyperparameters.
The column ``stars'' shows the importance of the hyperparameters in
stars. The importance is computed by the SPOT software.

\begin{Shaded}
\begin{Highlighting}[]
\ImportTok{from}\NormalTok{ spotPython.utils.eda }\ImportTok{import}\NormalTok{ gen\_design\_table}
\BuiltInTok{print}\NormalTok{(gen\_design\_table(fun\_control}\OperatorTok{=}\NormalTok{fun\_control, spot}\OperatorTok{=}\NormalTok{spot\_tuner))}
\end{Highlighting}
\end{Shaded}

\begin{verbatim}
| name         | type   | default   |   lower |   upper |   tuned | transform             |   importance | stars   |
|--------------|--------|-----------|---------|---------|---------|-----------------------|--------------|---------|
| l1           | int    | 5         |     2.0 |     9.0 |     6.0 | transform_power_2_int |         0.10 | .       |
| l2           | int    | 5         |     2.0 |     9.0 |     9.0 | transform_power_2_int |         0.00 |         |
| lr_mult      | float  | 1.0       |     1.0 |     1.0 |     1.0 | None                  |         0.00 |         |
| batch_size   | int    | 4         |     1.0 |     5.0 |     4.0 | transform_power_2_int |        16.86 | *       |
| epochs       | int    | 3         |     3.0 |     4.0 |     4.0 | transform_power_2_int |       100.00 | ***     |
| k_folds      | int    | 1         |     0.0 |     0.0 |     0.0 | None                  |         0.00 |         |
| patience     | int    | 5         |     3.0 |     3.0 |     3.0 | None                  |         0.00 |         |
| optimizer    | factor | SGD       |     0.0 |     9.0 |     1.0 | None                  |         3.49 | *       |
| sgd_momentum | float  | 0.0       |     0.9 |     0.9 |     0.9 | None                  |         0.00 |         |
\end{verbatim}

To visualize the most important hyperparameters, \texttt{spotPython}
provides the function \texttt{plot\_importance}. The following code
generates the importance plot from \textbf{?@fig-importance}.

\begin{Shaded}
\begin{Highlighting}[]
\NormalTok{spot\_tuner.plot\_importance(threshold}\OperatorTok{=}\FloatTok{0.025}\NormalTok{,}
\NormalTok{    filename}\OperatorTok{=}\StringTok{"./figures/"} \OperatorTok{+}\NormalTok{ experiment\_name}\OperatorTok{+}\StringTok{"\_importance.png"}\NormalTok{)}
\end{Highlighting}
\end{Shaded}

\begin{figure}[H]

{\centering \includegraphics{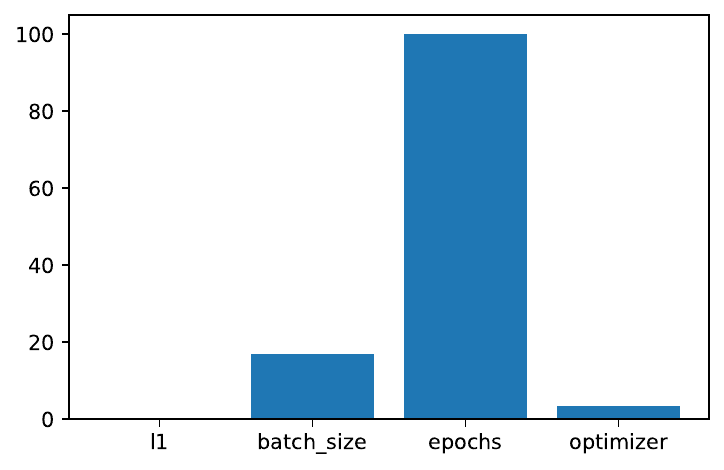}

}

\caption{Variable importance plot, threshold 0.025.}

\end{figure}

\hypertarget{sec-get-spot-results-14}{%
\subsection{Get the Tuned Architecture (SPOT
Results)}\label{sec-get-spot-results-14}}

The architecture of the \texttt{spotPython} model can be obtained as
follows. First, the numerical representation of the hyperparameters are
obtained, i.e., the numpy array \texttt{X} is generated. This array is
then used to generate the model \texttt{model\_spot} by the function
\texttt{get\_one\_core\_model\_from\_X}. The model \texttt{model\_spot}
has the following architecture:

\begin{Shaded}
\begin{Highlighting}[]
\ImportTok{from}\NormalTok{ spotPython.hyperparameters.values }\ImportTok{import}\NormalTok{ get\_one\_core\_model\_from\_X}
\NormalTok{X }\OperatorTok{=}\NormalTok{ spot\_tuner.to\_all\_dim(spot\_tuner.min\_X.reshape(}\DecValTok{1}\NormalTok{,}\OperatorTok{{-}}\DecValTok{1}\NormalTok{))}
\NormalTok{model\_spot }\OperatorTok{=}\NormalTok{ get\_one\_core\_model\_from\_X(X, fun\_control)}
\NormalTok{model\_spot}
\end{Highlighting}
\end{Shaded}

\begin{verbatim}
Net_CIFAR10(
  (conv1): Conv2d(3, 6, kernel_size=(5, 5), stride=(1, 1))
  (pool): MaxPool2d(kernel_size=2, stride=2, padding=0, dilation=1, ceil_mode=False)
  (conv2): Conv2d(6, 16, kernel_size=(5, 5), stride=(1, 1))
  (fc1): Linear(in_features=400, out_features=64, bias=True)
  (fc2): Linear(in_features=64, out_features=512, bias=True)
  (fc3): Linear(in_features=512, out_features=10, bias=True)
)
\end{verbatim}

\hypertarget{get-default-hyperparameters-2}{%
\subsection{Get Default
Hyperparameters}\label{get-default-hyperparameters-2}}

In a similar manner as in Section~\ref{sec-get-spot-results-14}, the
default hyperparameters can be obtained.

\begin{Shaded}
\begin{Highlighting}[]
\CommentTok{\# fun\_control was modified, we generate a new one with the original }
\CommentTok{\# default hyperparameters}
\ImportTok{from}\NormalTok{ spotPython.hyperparameters.values }\ImportTok{import}\NormalTok{ get\_one\_core\_model\_from\_X}
\ImportTok{from}\NormalTok{ spotPython.hyperparameters.values }\ImportTok{import}\NormalTok{ get\_default\_hyperparameters\_as\_array}
\NormalTok{X\_start }\OperatorTok{=}\NormalTok{ get\_default\_hyperparameters\_as\_array(fun\_control)}
\NormalTok{model\_default }\OperatorTok{=}\NormalTok{ get\_one\_core\_model\_from\_X(X\_start, fun\_control)}
\NormalTok{model\_default}
\end{Highlighting}
\end{Shaded}

\begin{verbatim}
Net_CIFAR10(
  (conv1): Conv2d(3, 6, kernel_size=(5, 5), stride=(1, 1))
  (pool): MaxPool2d(kernel_size=2, stride=2, padding=0, dilation=1, ceil_mode=False)
  (conv2): Conv2d(6, 16, kernel_size=(5, 5), stride=(1, 1))
  (fc1): Linear(in_features=400, out_features=32, bias=True)
  (fc2): Linear(in_features=32, out_features=32, bias=True)
  (fc3): Linear(in_features=32, out_features=10, bias=True)
)
\end{verbatim}

\hypertarget{evaluation-of-the-default-architecture}{%
\subsection{Evaluation of the Default
Architecture}\label{evaluation-of-the-default-architecture}}

The method \texttt{train\_tuned} takes a model architecture without
trained weights and trains this model with the train data. The train
data is split into train and validation data. The validation data is
used for early stopping. The trained model weights are saved as a
dictionary.

This evaluation is similar to the final evaluation in PyTorch (2023a).

\begin{Shaded}
\begin{Highlighting}[]
\ImportTok{from}\NormalTok{ spotPython.torch.traintest }\ImportTok{import}\NormalTok{ (}
\NormalTok{    train\_tuned,}
\NormalTok{    test\_tuned,}
\NormalTok{    )}
\NormalTok{train\_tuned(net}\OperatorTok{=}\NormalTok{model\_default, train\_dataset}\OperatorTok{=}\NormalTok{train, shuffle}\OperatorTok{=}\VariableTok{True}\NormalTok{,}
\NormalTok{        loss\_function}\OperatorTok{=}\NormalTok{fun\_control[}\StringTok{"loss\_function"}\NormalTok{],}
\NormalTok{        metric}\OperatorTok{=}\NormalTok{fun\_control[}\StringTok{"metric\_torch"}\NormalTok{],}
\NormalTok{        device }\OperatorTok{=}\NormalTok{ fun\_control[}\StringTok{"device"}\NormalTok{], show\_batch\_interval}\OperatorTok{=}\DecValTok{1\_000\_000}\NormalTok{,}
\NormalTok{        path}\OperatorTok{=}\VariableTok{None}\NormalTok{,}
\NormalTok{        task}\OperatorTok{=}\NormalTok{fun\_control[}\StringTok{"task"}\NormalTok{],)}

\NormalTok{test\_tuned(net}\OperatorTok{=}\NormalTok{model\_default, test\_dataset}\OperatorTok{=}\NormalTok{test, }
\NormalTok{        loss\_function}\OperatorTok{=}\NormalTok{fun\_control[}\StringTok{"loss\_function"}\NormalTok{],}
\NormalTok{        metric}\OperatorTok{=}\NormalTok{fun\_control[}\StringTok{"metric\_torch"}\NormalTok{],}
\NormalTok{        shuffle}\OperatorTok{=}\VariableTok{False}\NormalTok{, }
\NormalTok{        device }\OperatorTok{=}\NormalTok{ fun\_control[}\StringTok{"device"}\NormalTok{],}
\NormalTok{        task}\OperatorTok{=}\NormalTok{fun\_control[}\StringTok{"task"}\NormalTok{],)        }
\end{Highlighting}
\end{Shaded}

\begin{verbatim}
Epoch: 1 | 
\end{verbatim}

\begin{verbatim}
MulticlassAccuracy: 0.1013000011444092 | Loss: 2.2993141119003297 | Acc: 0.1013000000000000.
Epoch: 2 | 
\end{verbatim}

\begin{verbatim}
MulticlassAccuracy: 0.1157499998807907 | Loss: 2.2862341335296632 | Acc: 0.1157500000000000.
Epoch: 3 | 
\end{verbatim}

\begin{verbatim}
MulticlassAccuracy: 0.1534000039100647 | Loss: 2.2558263620376588 | Acc: 0.1534000000000000.
Epoch: 4 | 
\end{verbatim}

\begin{verbatim}
MulticlassAccuracy: 0.2099500000476837 | Loss: 2.2096788969039918 | Acc: 0.2099500000000000.
Epoch: 5 | 
\end{verbatim}

\begin{verbatim}
MulticlassAccuracy: 0.2171999961137772 | Loss: 2.1583650140762329 | Acc: 0.2172000000000000.
Epoch: 6 | 
\end{verbatim}

\begin{verbatim}
MulticlassAccuracy: 0.2302500009536743 | Loss: 2.1003214435577391 | Acc: 0.2302500000000000.
Epoch: 7 | 
\end{verbatim}

\begin{verbatim}
MulticlassAccuracy: 0.2409500032663345 | Loss: 2.0469134126663207 | Acc: 0.2409500000000000.
Epoch: 8 | 
\end{verbatim}

\begin{verbatim}
MulticlassAccuracy: 0.2525500059127808 | Loss: 2.0065110932350159 | Acc: 0.2525500000000000.
Returned to Spot: Validation loss: 2.006511093235016
\end{verbatim}

\begin{verbatim}
MulticlassAccuracy: 0.2576000094413757 | Loss: 2.0048375873565676 | Acc: 0.2576000000000000.
Final evaluation: Validation loss: 2.0048375873565676
Final evaluation: Validation metric: 0.25760000944137573
----------------------------------------------
\end{verbatim}

\begin{verbatim}
(2.0048375873565676, nan, tensor(0.2576, device='mps:0'))
\end{verbatim}

\hypertarget{evaluation-of-the-tuned-architecture}{%
\subsection{Evaluation of the Tuned
Architecture}\label{evaluation-of-the-tuned-architecture}}

The following code trains the model \texttt{model\_spot}.

If \texttt{path} is set to a filename, e.g.,
\texttt{path\ =\ "model\_spot\_trained.pt"}, the weights of the trained
model will be saved to this file.

If \texttt{path} is set to a filename, e.g.,
\texttt{path\ =\ "model\_spot\_trained.pt"}, the weights of the trained
model will be loaded from this file.

\begin{Shaded}
\begin{Highlighting}[]
\NormalTok{train\_tuned(net}\OperatorTok{=}\NormalTok{model\_spot, train\_dataset}\OperatorTok{=}\NormalTok{train,}
\NormalTok{        loss\_function}\OperatorTok{=}\NormalTok{fun\_control[}\StringTok{"loss\_function"}\NormalTok{],}
\NormalTok{        metric}\OperatorTok{=}\NormalTok{fun\_control[}\StringTok{"metric\_torch"}\NormalTok{],}
\NormalTok{        shuffle}\OperatorTok{=}\VariableTok{True}\NormalTok{,}
\NormalTok{        device }\OperatorTok{=}\NormalTok{ fun\_control[}\StringTok{"device"}\NormalTok{],}
\NormalTok{        path}\OperatorTok{=}\VariableTok{None}\NormalTok{,}
\NormalTok{        task}\OperatorTok{=}\NormalTok{fun\_control[}\StringTok{"task"}\NormalTok{],)}
\NormalTok{test\_tuned(net}\OperatorTok{=}\NormalTok{model\_spot, test\_dataset}\OperatorTok{=}\NormalTok{test,}
\NormalTok{            shuffle}\OperatorTok{=}\VariableTok{False}\NormalTok{,}
\NormalTok{            loss\_function}\OperatorTok{=}\NormalTok{fun\_control[}\StringTok{"loss\_function"}\NormalTok{],}
\NormalTok{            metric}\OperatorTok{=}\NormalTok{fun\_control[}\StringTok{"metric\_torch"}\NormalTok{],}
\NormalTok{            device }\OperatorTok{=}\NormalTok{ fun\_control[}\StringTok{"device"}\NormalTok{],}
\NormalTok{            task}\OperatorTok{=}\NormalTok{fun\_control[}\StringTok{"task"}\NormalTok{],)}
\end{Highlighting}
\end{Shaded}

\begin{verbatim}
Epoch: 1 | 
\end{verbatim}

\begin{verbatim}
MulticlassAccuracy: 0.4553500115871429 | Loss: 1.4807784632682801 | Acc: 0.4553500000000000.
Epoch: 2 | 
\end{verbatim}

\begin{verbatim}
MulticlassAccuracy: 0.4986500144004822 | Loss: 1.3824706964015960 | Acc: 0.4986500000000000.
Epoch: 3 | 
\end{verbatim}

\begin{verbatim}
MulticlassAccuracy: 0.5169000029563904 | Loss: 1.3429181780815125 | Acc: 0.5169000000000000.
Epoch: 4 | 
\end{verbatim}

\begin{verbatim}
MulticlassAccuracy: 0.5296999812126160 | Loss: 1.3132050466537475 | Acc: 0.5296999999999999.
Epoch: 5 | 
\end{verbatim}

\begin{verbatim}
MulticlassAccuracy: 0.5366500020027161 | Loss: 1.2941528817415238 | Acc: 0.5366500000000000.
Epoch: 6 | 
\end{verbatim}

\begin{verbatim}
MulticlassAccuracy: 0.5393499732017517 | Loss: 1.2878079622745513 | Acc: 0.5393500000000000.
Epoch: 7 | 
\end{verbatim}

\begin{verbatim}
MulticlassAccuracy: 0.5490499734878540 | Loss: 1.2646987820148468 | Acc: 0.5490500000000000.
Epoch: 8 | 
\end{verbatim}

\begin{verbatim}
MulticlassAccuracy: 0.5544499754905701 | Loss: 1.2544260616302489 | Acc: 0.5544500000000000.
Epoch: 9 | 
\end{verbatim}

\begin{verbatim}
MulticlassAccuracy: 0.5620999932289124 | Loss: 1.2338094377756119 | Acc: 0.5621000000000000.
Epoch: 10 | 
\end{verbatim}

\begin{verbatim}
MulticlassAccuracy: 0.5637500286102295 | Loss: 1.2312240300893784 | Acc: 0.5637500000000000.
Epoch: 11 | 
\end{verbatim}

\begin{verbatim}
MulticlassAccuracy: 0.5688999891281128 | Loss: 1.2254522174358369 | Acc: 0.5689000000000000.
Epoch: 12 | 
\end{verbatim}

\begin{verbatim}
MulticlassAccuracy: 0.5709999799728394 | Loss: 1.2168787500381471 | Acc: 0.5710000000000000.
Epoch: 13 | 
\end{verbatim}

\begin{verbatim}
MulticlassAccuracy: 0.5732499957084656 | Loss: 1.2131404494524003 | Acc: 0.5732500000000000.
Epoch: 14 | 
\end{verbatim}

\begin{verbatim}
MulticlassAccuracy: 0.5752500295639038 | Loss: 1.2019421347618102 | Acc: 0.5752500000000000.
Epoch: 15 | 
\end{verbatim}

\begin{verbatim}
MulticlassAccuracy: 0.5807499885559082 | Loss: 1.1982519413948058 | Acc: 0.5807500000000000.
Epoch: 16 | 
\end{verbatim}

\begin{verbatim}
MulticlassAccuracy: 0.5807999968528748 | Loss: 1.1949795161724091 | Acc: 0.5808000000000000.
Returned to Spot: Validation loss: 1.194979516172409
\end{verbatim}

\begin{verbatim}
MulticlassAccuracy: 0.5852000117301941 | Loss: 1.2008806512832642 | Acc: 0.5852000000000001.
Final evaluation: Validation loss: 1.2008806512832642
Final evaluation: Validation metric: 0.5852000117301941
----------------------------------------------
\end{verbatim}

\begin{verbatim}
(1.2008806512832642, nan, tensor(0.5852, device='mps:0'))
\end{verbatim}

\hypertarget{detailed-hyperparameter-plots-2}{%
\subsection{Detailed Hyperparameter
Plots}\label{detailed-hyperparameter-plots-2}}

The contour plots in this section visualize the interactions of the
three most important hyperparameters. Since some of these
hyperparameters take fatorial or integer values, sometimes step-like
fitness landcapes (or response surfaces) are generated. SPOT draws the
interactions of the main hyperparameters by default. It is also possible
to visualize all interactions.

\begin{Shaded}
\begin{Highlighting}[]
\NormalTok{filename }\OperatorTok{=} \StringTok{"./figures/"} \OperatorTok{+}\NormalTok{ experiment\_name}
\NormalTok{spot\_tuner.plot\_important\_hyperparameter\_contour(filename}\OperatorTok{=}\NormalTok{filename)}
\end{Highlighting}
\end{Shaded}

\begin{verbatim}
l1:  0.10134443931754378
batch_size:  16.862145330943314
epochs:  100.0
optimizer:  3.487626907795692
\end{verbatim}

\begin{figure}[H]

{\centering \includegraphics{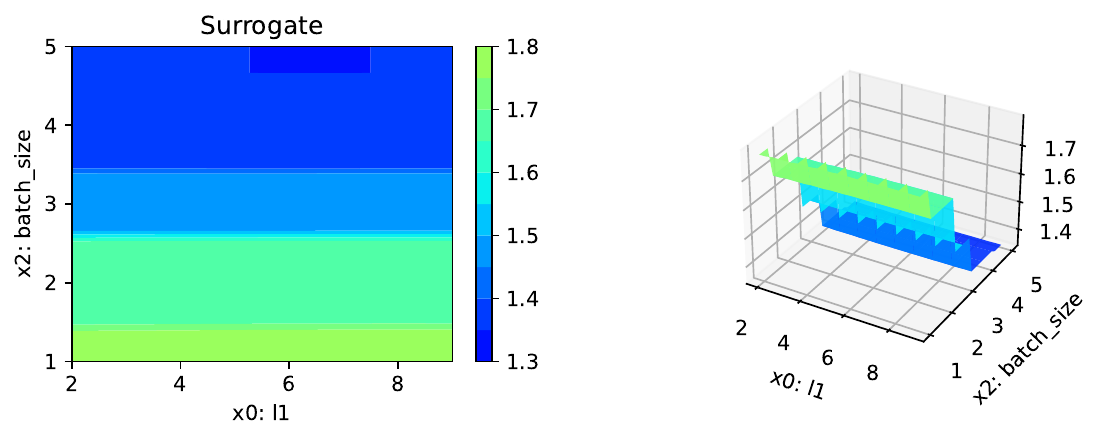}

}

\caption{Contour plots.}

\end{figure}

\begin{figure}[H]

{\centering \includegraphics{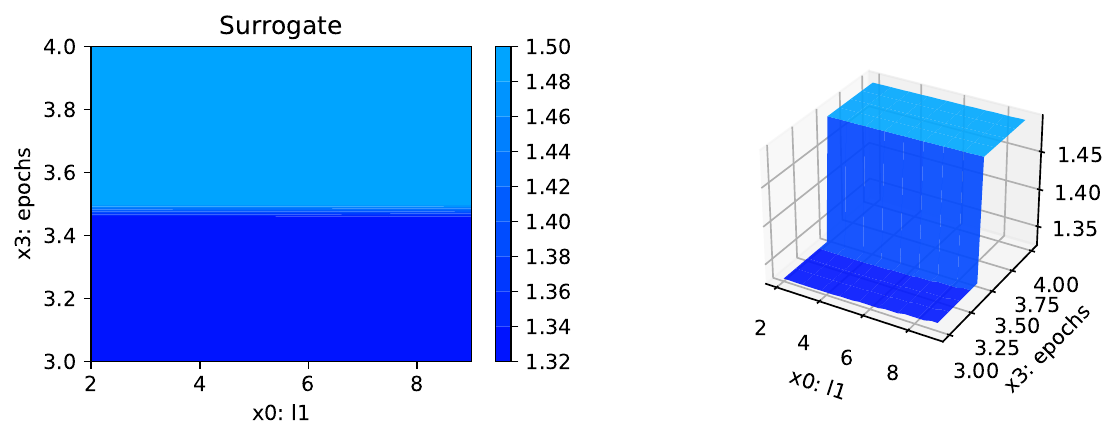}

}

\end{figure}

\begin{figure}[H]

{\centering \includegraphics{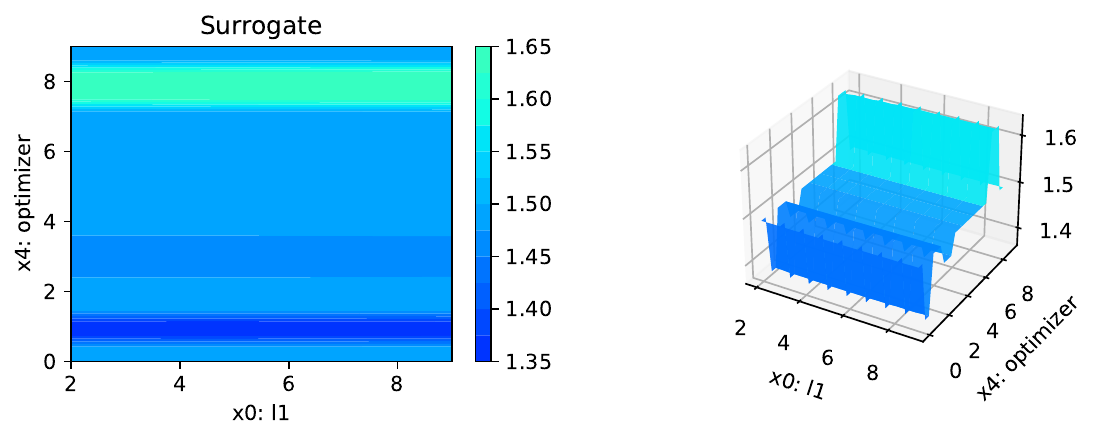}

}

\end{figure}

\begin{figure}[H]

{\centering \includegraphics{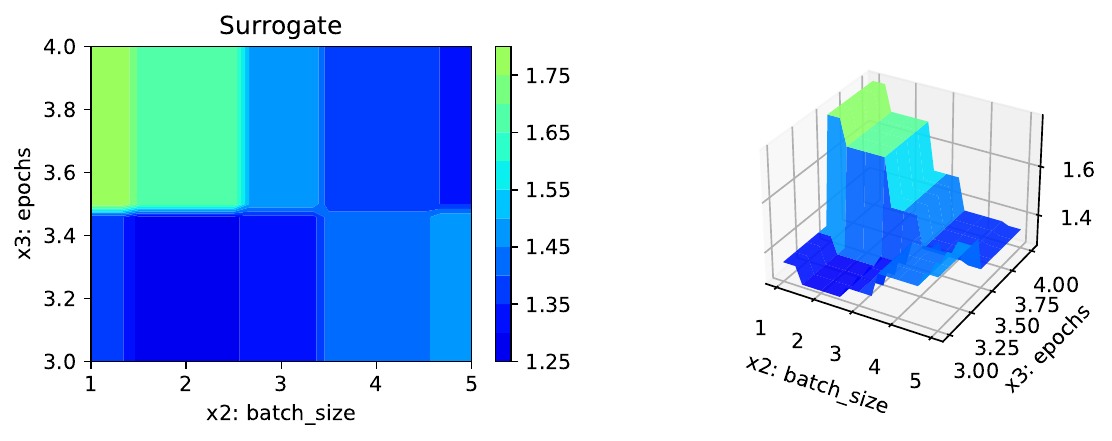}

}

\end{figure}

\begin{figure}[H]

{\centering \includegraphics{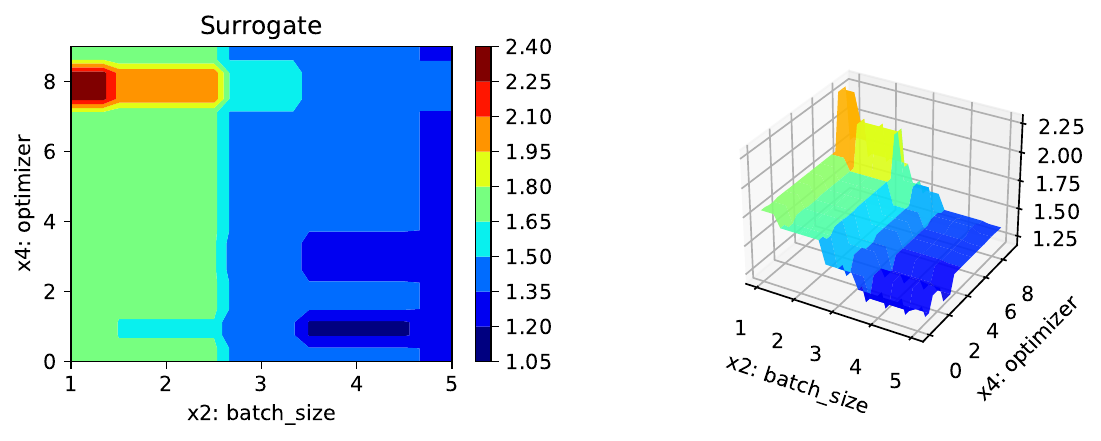}

}

\end{figure}

\begin{figure}[H]

{\centering \includegraphics{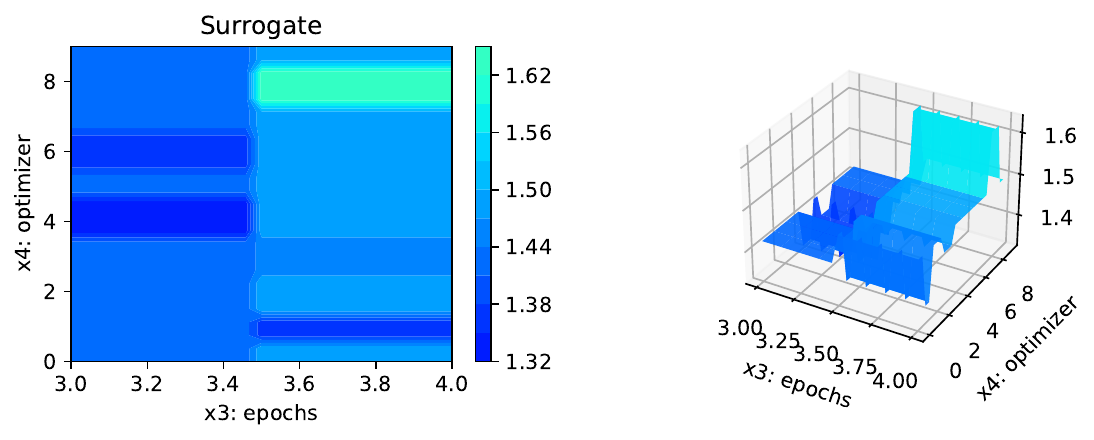}

}

\end{figure}

The figures (\textbf{?@fig-contour}) show the contour plots of the loss
as a function of the hyperparameters. These plots are very helpful for
benchmark studies and for understanding neural networks.
\texttt{spotPython} provides additional tools for a visual inspection of
the results and give valuable insights into the hyperparameter tuning
process. This is especially useful for model explainability,
transparency, and trustworthiness. In addition to the contour plots,
\textbf{?@fig-parallel} shows the parallel plot of the hyperparameters.

\begin{Shaded}
\begin{Highlighting}[]
\NormalTok{spot\_tuner.parallel\_plot()}
\end{Highlighting}
\end{Shaded}

\begin{verbatim}
Unable to display output for mime type(s): text/html
\end{verbatim}

Parallel coordinates plots

\begin{verbatim}
Unable to display output for mime type(s): text/html
\end{verbatim}

\hypertarget{sec-summary}{%
\section{Summary and Outlook}\label{sec-summary}}

This tutorial presents the hyperparameter tuning open source software
\texttt{spotPython} for \texttt{PyTorch}. To show its basic features, a
comparison with the ``official'' \texttt{PyTorch} hyperparameter tuning
tutorial (PyTorch 2023a) is presented. Some of the advantages of
\texttt{spotPython} are:

\begin{itemize}
\tightlist
\item
  Numerical and categorical hyperparameters.
\item
  Powerful surrogate models.
\item
  Flexible approach and easy to use.
\item
  Simple JSON files for the specification of the hyperparameters.
\item
  Extension of default and user specified network classes.
\item
  Noise handling techniques.
\item
  Interaction with \texttt{tensorboard}.
\end{itemize}

Currently, only rudimentary parallel and distributed neural network
training is possible, but these capabilities will be extended in the
future. The next version of \texttt{spotPython} will also include a more
detailed documentation and more examples.

\begin{tcolorbox}[enhanced jigsaw, left=2mm, title=\textcolor{quarto-callout-important-color}{\faExclamation}\hspace{0.5em}{Important}, bottomrule=.15mm, titlerule=0mm, breakable, rightrule=.15mm, toprule=.15mm, coltitle=black, colbacktitle=quarto-callout-important-color!10!white, leftrule=.75mm, arc=.35mm, colframe=quarto-callout-important-color-frame, bottomtitle=1mm, colback=white, opacitybacktitle=0.6, toptitle=1mm, opacityback=0]

Important: This tutorial does not present a complete benchmarking study
(Bartz-Beielstein et al. 2020). The results are only preliminary and
highly dependent on the local configuration (hard- and software). Our
goal is to provide a first impression of the performance of the
hyperparameter tuning package \texttt{spotPython}. To demonstrate its
capabilities, a quick comparison with \texttt{ray{[}tune{]}} was
performed. \texttt{ray{[}tune{]}} was chosen, because it is presented as
``an industry standard tool for distributed hyperparameter tuning.'' The
results should be interpreted with care.

\end{tcolorbox}

\hypertarget{sec-appendix}{%
\section{Appendix}\label{sec-appendix}}

\hypertarget{sample-output-from-ray-tunes-run}{%
\subsection{Sample Output From Ray Tune's
Run}\label{sample-output-from-ray-tunes-run}}

The output from \texttt{ray{[}tune{]}} could look like this (PyTorch
2023b):

\begin{Shaded}
\begin{Highlighting}[]
\NormalTok{Number of trials: 10 (10 TERMINATED)}
\NormalTok{{-}{-}{-}{-}{-}{-}+{-}{-}{-}{-}{-}{-}+{-}{-}{-}{-}{-}{-}{-}{-}{-}{-}{-}{-}{-}+{-}{-}{-}{-}{-}{-}{-}{-}{-}{-}{-}{-}{-}{-}+{-}{-}{-}{-}{-}{-}{-}{-}{-}+{-}{-}{-}{-}{-}{-}{-}{-}{-}{-}{-}{-}+{-}{-}{-}{-}{-}{-}{-}{-}{-}{-}{-}{-}{-}{-}{-}{-}{-}{-}{-}{-}+}
\NormalTok{|   l1 |   l2 |          lr |   batch\_size |    loss |   accuracy | training\_iteration |}
\NormalTok{+{-}{-}{-}{-}{-}{-}+{-}{-}{-}{-}{-}{-}+{-}{-}{-}{-}{-}{-}{-}{-}{-}{-}{-}{-}{-}+{-}{-}{-}{-}{-}{-}{-}{-}{-}{-}{-}{-}{-}{-}+{-}{-}{-}{-}{-}{-}{-}{-}{-}+{-}{-}{-}{-}{-}{-}{-}{-}{-}{-}{-}{-}+{-}{-}{-}{-}{-}{-}{-}{-}{-}{-}{-}{-}{-}{-}{-}{-}{-}{-}{-}{-}|}
\NormalTok{|   64 |    4 | 0.00011629  |            2 | 1.87273 |     0.244  |                  2 |}
\NormalTok{|   32 |   64 | 0.000339763 |            8 | 1.23603 |     0.567  |                  8 |}
\NormalTok{|    8 |   16 | 0.00276249  |           16 | 1.1815  |     0.5836 |                 10 |}
\NormalTok{|    4 |   64 | 0.000648721 |            4 | 1.31131 |     0.5224 |                  8 |}
\NormalTok{|   32 |   16 | 0.000340753 |            8 | 1.26454 |     0.5444 |                  8 |}
\NormalTok{|    8 |    4 | 0.000699775 |            8 | 1.99594 |     0.1983 |                  2 |}
\NormalTok{|  256 |    8 | 0.0839654   |           16 | 2.3119  |     0.0993 |                  1 |}
\NormalTok{|   16 |  128 | 0.0758154   |           16 | 2.33575 |     0.1327 |                  1 |}
\NormalTok{|   16 |    8 | 0.0763312   |           16 | 2.31129 |     0.1042 |                  4 |}
\NormalTok{|  128 |   16 | 0.000124903 |            4 | 2.26917 |     0.1945 |                  1 |}
\NormalTok{+{-}{-}{-}{-}{-}+{-}{-}{-}{-}{-}{-}+{-}{-}{-}{-}{-}{-}+{-}{-}{-}{-}{-}{-}{-}{-}{-}{-}{-}{-}{-}+{-}{-}{-}{-}{-}{-}{-}{-}{-}{-}{-}{-}{-}{-}+{-}{-}{-}{-}{-}{-}{-}{-}{-}+{-}{-}{-}{-}{-}{-}{-}{-}{-}{-}{-}{-}+{-}{-}{-}{-}{-}{-}{-}{-}{-}{-}{-}{-}{-}{-}{-}{-}{-}{-}{-}{-}+}
\NormalTok{Best trial config: \{\textquotesingle{}l1\textquotesingle{}: 8, \textquotesingle{}l2\textquotesingle{}: 16, \textquotesingle{}lr\textquotesingle{}: 0.00276249, \textquotesingle{}batch\_size\textquotesingle{}: 16, \textquotesingle{}data\_dir\textquotesingle{}: \textquotesingle{}...\textquotesingle{}\}}
\NormalTok{Best trial final validation loss: 1.181501}
\NormalTok{Best trial final validation accuracy: 0.5836}
\NormalTok{Best trial test set accuracy: 0.5806}
\end{Highlighting}
\end{Shaded}

\hypertarget{sec-hpt-random-forest-classifier}{%
\chapter{HPT: sklearn RandomForestClassifier VBDP
Data}\label{sec-hpt-random-forest-classifier}}

This chapter describes the hyperparameter tuning of a
\texttt{RandomForestClassifier} on the Vector Borne Disease Prediction
(VBDP) data set.

\begin{tcolorbox}[enhanced jigsaw, left=2mm, title=\textcolor{quarto-callout-important-color}{\faExclamation}\hspace{0.5em}{Vector Borne Disease Prediction Data Set}, bottomrule=.15mm, titlerule=0mm, breakable, rightrule=.15mm, toprule=.15mm, coltitle=black, colbacktitle=quarto-callout-important-color!10!white, leftrule=.75mm, arc=.35mm, colframe=quarto-callout-important-color-frame, bottomtitle=1mm, colback=white, opacitybacktitle=0.6, toptitle=1mm, opacityback=0]

This chapter uses the Vector Borne Disease Prediction data set from
Kaggle. It is a categorical dataset for eleven Vector Borne Diseases
with associated symptoms.

\begin{quote}
The person who associated a work with this deed has dedicated the work
to the public domain by waiving all of his or her rights to the work
worldwide under copyright law, including all related and neighboring
rights, to the extent allowed by law.You can copy, modify, distribute
and perform the work, even for commercial purposes, all without asking
permission. See Other Information below, see
\url{https://creativecommons.org/publicdomain/zero/1.0/}.
\end{quote}

The data set is available at:
\url{https://www.kaggle.com/datasets/richardbernat/vector-borne-disease-prediction},

The data should be downloaded and stored in the \texttt{data/VBDP}
subfolder. The data set is not available as a part of the
\texttt{spotPython} package.

\end{tcolorbox}

\hypertarget{sec-setup-16}{%
\section{Step 1: Setup}\label{sec-setup-16}}

Before we consider the detailed experimental setup, we select the
parameters that affect run time and the initial design size.

\begin{Shaded}
\begin{Highlighting}[]
\NormalTok{MAX\_TIME }\OperatorTok{=} \DecValTok{1}
\NormalTok{INIT\_SIZE }\OperatorTok{=} \DecValTok{5}
\NormalTok{ORIGINAL }\OperatorTok{=} \VariableTok{True}
\NormalTok{PREFIX }\OperatorTok{=} \StringTok{"16"}
\end{Highlighting}
\end{Shaded}

\begin{Shaded}
\begin{Highlighting}[]
\ImportTok{import}\NormalTok{ warnings}
\NormalTok{warnings.filterwarnings(}\StringTok{"ignore"}\NormalTok{)}
\end{Highlighting}
\end{Shaded}

\hypertarget{step-2-initialization-of-the-empty-fun_control-dictionary-1}{%
\section{\texorpdfstring{Step 2: Initialization of the Empty
\texttt{fun\_control}
Dictionary}{Step 2: Initialization of the Empty fun\_control Dictionary}}\label{step-2-initialization-of-the-empty-fun_control-dictionary-1}}

\begin{Shaded}
\begin{Highlighting}[]
\ImportTok{from}\NormalTok{ spotPython.utils.init }\ImportTok{import}\NormalTok{ fun\_control\_init}
\ImportTok{from}\NormalTok{ spotPython.utils.}\BuiltInTok{file} \ImportTok{import}\NormalTok{ get\_experiment\_name, get\_spot\_tensorboard\_path}
\ImportTok{from}\NormalTok{ spotPython.utils.device }\ImportTok{import}\NormalTok{ getDevice}

\NormalTok{experiment\_name }\OperatorTok{=}\NormalTok{ get\_experiment\_name(prefix}\OperatorTok{=}\NormalTok{PREFIX)}

\NormalTok{fun\_control }\OperatorTok{=}\NormalTok{ fun\_control\_init(}
\NormalTok{    task}\OperatorTok{=}\StringTok{"classification"}\NormalTok{,}
\NormalTok{    spot\_tensorboard\_path}\OperatorTok{=}\NormalTok{get\_spot\_tensorboard\_path(experiment\_name))}
\end{Highlighting}
\end{Shaded}

\hypertarget{step-3-pytorch-data-loading}{%
\section{Step 3: PyTorch Data
Loading}\label{step-3-pytorch-data-loading}}

\hypertarget{load-data-classification-vbdp}{%
\subsection{Load Data: Classification
VBDP}\label{load-data-classification-vbdp}}

\begin{Shaded}
\begin{Highlighting}[]
\ImportTok{import}\NormalTok{ pandas }\ImportTok{as}\NormalTok{ pd}
\ControlFlowTok{if}\NormalTok{ ORIGINAL }\OperatorTok{==} \VariableTok{True}\NormalTok{:}
\NormalTok{    train\_df }\OperatorTok{=}\NormalTok{ pd.read\_csv(}\StringTok{\textquotesingle{}./data/VBDP/trainn.csv\textquotesingle{}}\NormalTok{)}
\NormalTok{    test\_df }\OperatorTok{=}\NormalTok{ pd.read\_csv(}\StringTok{\textquotesingle{}./data/VBDP/testt.csv\textquotesingle{}}\NormalTok{)}
\ControlFlowTok{else}\NormalTok{:}
\NormalTok{    train\_df }\OperatorTok{=}\NormalTok{ pd.read\_csv(}\StringTok{\textquotesingle{}./data/VBDP/train.csv\textquotesingle{}}\NormalTok{)}
    \CommentTok{\# remove the id column}
\NormalTok{    train\_df }\OperatorTok{=}\NormalTok{ train\_df.drop(columns}\OperatorTok{=}\NormalTok{[}\StringTok{\textquotesingle{}id\textquotesingle{}}\NormalTok{])}
\end{Highlighting}
\end{Shaded}

\begin{Shaded}
\begin{Highlighting}[]
\ImportTok{from}\NormalTok{ sklearn.preprocessing }\ImportTok{import}\NormalTok{ OrdinalEncoder}
\NormalTok{n\_samples }\OperatorTok{=}\NormalTok{ train\_df.shape[}\DecValTok{0}\NormalTok{]}
\NormalTok{n\_features }\OperatorTok{=}\NormalTok{ train\_df.shape[}\DecValTok{1}\NormalTok{] }\OperatorTok{{-}} \DecValTok{1}
\NormalTok{target\_column }\OperatorTok{=} \StringTok{"prognosis"}
\CommentTok{\# Encoder our prognosis labels as integers for easier decoding later}
\NormalTok{enc }\OperatorTok{=}\NormalTok{ OrdinalEncoder()}
\NormalTok{train\_df[target\_column] }\OperatorTok{=}\NormalTok{ enc.fit\_transform(train\_df[[target\_column]])}
\NormalTok{train\_df.columns }\OperatorTok{=}\NormalTok{ [}\SpecialStringTok{f"x}\SpecialCharTok{\{}\NormalTok{i}\SpecialCharTok{\}}\SpecialStringTok{"} \ControlFlowTok{for}\NormalTok{ i }\KeywordTok{in} \BuiltInTok{range}\NormalTok{(}\DecValTok{1}\NormalTok{, n\_features}\OperatorTok{+}\DecValTok{1}\NormalTok{)] }\OperatorTok{+}\NormalTok{ [target\_column]}
\BuiltInTok{print}\NormalTok{(train\_df.shape)}
\NormalTok{train\_df.head()}
\end{Highlighting}
\end{Shaded}

\begin{verbatim}
(252, 65)
\end{verbatim}

\begin{longtable}[]{@{}llllllllllllllllllllll@{}}
\toprule\noalign{}
& x1 & x2 & x3 & x4 & x5 & x6 & x7 & x8 & x9 & x10 & ... & x56 & x57 &
x58 & x59 & x60 & x61 & x62 & x63 & x64 & prognosis \\
\midrule\noalign{}
\endhead
\bottomrule\noalign{}
\endlastfoot
0 & 0 & 1 & 1 & 1 & 1 & 0 & 1 & 0 & 0 & 0 & ... & 0 & 0 & 0 & 0 & 0 & 0
& 0 & 0 & 0 & 0.0 \\
1 & 1 & 1 & 1 & 1 & 1 & 0 & 1 & 1 & 1 & 0 & ... & 0 & 0 & 0 & 0 & 0 & 0
& 0 & 0 & 0 & 0.0 \\
2 & 0 & 1 & 0 & 1 & 0 & 0 & 1 & 1 & 0 & 0 & ... & 0 & 0 & 0 & 0 & 0 & 0
& 0 & 0 & 0 & 0.0 \\
3 & 0 & 0 & 0 & 0 & 0 & 1 & 1 & 1 & 0 & 0 & ... & 0 & 0 & 0 & 0 & 0 & 0
& 0 & 0 & 0 & 0.0 \\
4 & 1 & 0 & 0 & 0 & 1 & 1 & 1 & 1 & 0 & 0 & ... & 0 & 0 & 0 & 0 & 0 & 0
& 0 & 0 & 0 & 0.0 \\
\end{longtable}

The full data set \texttt{train\_df} 64 features. The target column is
labeled as \texttt{prognosis}.

\hypertarget{holdout-train-and-test-data}{%
\subsection{Holdout Train and Test
Data}\label{holdout-train-and-test-data}}

We split out a hold-out test set (25\% of the data) so we can calculate
an example MAP@K

\begin{Shaded}
\begin{Highlighting}[]
\ImportTok{import}\NormalTok{ numpy }\ImportTok{as}\NormalTok{ np}
\ImportTok{from}\NormalTok{ sklearn.model\_selection }\ImportTok{import}\NormalTok{ train\_test\_split}
\NormalTok{X\_train, X\_test, y\_train, y\_test }\OperatorTok{=}\NormalTok{ train\_test\_split(train\_df.drop(target\_column, axis}\OperatorTok{=}\DecValTok{1}\NormalTok{), train\_df[target\_column],}
\NormalTok{                                                    random\_state}\OperatorTok{=}\DecValTok{42}\NormalTok{,}
\NormalTok{                                                    test\_size}\OperatorTok{=}\FloatTok{0.25}\NormalTok{,}
\NormalTok{                                                    stratify}\OperatorTok{=}\NormalTok{train\_df[target\_column])}
\NormalTok{train }\OperatorTok{=}\NormalTok{ pd.DataFrame(np.hstack((X\_train, np.array(y\_train).reshape(}\OperatorTok{{-}}\DecValTok{1}\NormalTok{, }\DecValTok{1}\NormalTok{))))}
\NormalTok{test }\OperatorTok{=}\NormalTok{ pd.DataFrame(np.hstack((X\_test, np.array(y\_test).reshape(}\OperatorTok{{-}}\DecValTok{1}\NormalTok{, }\DecValTok{1}\NormalTok{))))}
\NormalTok{train.columns }\OperatorTok{=}\NormalTok{ [}\SpecialStringTok{f"x}\SpecialCharTok{\{}\NormalTok{i}\SpecialCharTok{\}}\SpecialStringTok{"} \ControlFlowTok{for}\NormalTok{ i }\KeywordTok{in} \BuiltInTok{range}\NormalTok{(}\DecValTok{1}\NormalTok{, n\_features}\OperatorTok{+}\DecValTok{1}\NormalTok{)] }\OperatorTok{+}\NormalTok{ [target\_column]}
\NormalTok{test.columns }\OperatorTok{=}\NormalTok{ [}\SpecialStringTok{f"x}\SpecialCharTok{\{}\NormalTok{i}\SpecialCharTok{\}}\SpecialStringTok{"} \ControlFlowTok{for}\NormalTok{ i }\KeywordTok{in} \BuiltInTok{range}\NormalTok{(}\DecValTok{1}\NormalTok{, n\_features}\OperatorTok{+}\DecValTok{1}\NormalTok{)] }\OperatorTok{+}\NormalTok{ [target\_column]}
\BuiltInTok{print}\NormalTok{(train.shape)}
\BuiltInTok{print}\NormalTok{(test.shape)}
\NormalTok{train.head()}
\end{Highlighting}
\end{Shaded}

\begin{verbatim}
(189, 65)
(63, 65)
\end{verbatim}

\begin{longtable}[]{@{}llllllllllllllllllllll@{}}
\toprule\noalign{}
& x1 & x2 & x3 & x4 & x5 & x6 & x7 & x8 & x9 & x10 & ... & x56 & x57 &
x58 & x59 & x60 & x61 & x62 & x63 & x64 & prognosis \\
\midrule\noalign{}
\endhead
\bottomrule\noalign{}
\endlastfoot
0 & 1.0 & 0.0 & 0.0 & 1.0 & 0.0 & 1.0 & 0.0 & 0.0 & 0.0 & 1.0 & ... &
0.0 & 0.0 & 0.0 & 0.0 & 1.0 & 1.0 & 1.0 & 0.0 & 0.0 & 7.0 \\
1 & 1.0 & 0.0 & 1.0 & 1.0 & 1.0 & 1.0 & 1.0 & 0.0 & 1.0 & 1.0 & ... &
0.0 & 1.0 & 1.0 & 1.0 & 1.0 & 0.0 & 1.0 & 1.0 & 1.0 & 3.0 \\
2 & 0.0 & 0.0 & 1.0 & 0.0 & 1.0 & 0.0 & 0.0 & 0.0 & 0.0 & 0.0 & ... &
0.0 & 0.0 & 0.0 & 0.0 & 0.0 & 0.0 & 0.0 & 0.0 & 0.0 & 10.0 \\
3 & 1.0 & 1.0 & 1.0 & 1.0 & 1.0 & 1.0 & 0.0 & 0.0 & 1.0 & 1.0 & ... &
1.0 & 0.0 & 1.0 & 1.0 & 1.0 & 0.0 & 0.0 & 1.0 & 1.0 & 3.0 \\
4 & 1.0 & 1.0 & 1.0 & 0.0 & 1.0 & 1.0 & 0.0 & 1.0 & 1.0 & 0.0 & ... &
0.0 & 0.0 & 0.0 & 0.0 & 0.0 & 0.0 & 0.0 & 0.0 & 0.0 & 8.0 \\
\end{longtable}

\begin{Shaded}
\begin{Highlighting}[]
\CommentTok{\# add the dataset to the fun\_control}
\NormalTok{fun\_control.update(\{}\StringTok{"data"}\NormalTok{: train\_df, }\CommentTok{\# full dataset,}
               \StringTok{"train"}\NormalTok{: train,}
               \StringTok{"test"}\NormalTok{: test,}
               \StringTok{"n\_samples"}\NormalTok{: n\_samples,}
               \StringTok{"target\_column"}\NormalTok{: target\_column\})}
\end{Highlighting}
\end{Shaded}

\hypertarget{sec-specification-of-preprocessing-model-16}{%
\section{Step 4: Specification of the Preprocessing
Model}\label{sec-specification-of-preprocessing-model-16}}

Data preprocesssing can be very simple, e.g., you can ignore it. Then
you would choose the \texttt{prep\_model} ``None'':

\begin{Shaded}
\begin{Highlighting}[]
\NormalTok{prep\_model }\OperatorTok{=} \VariableTok{None}
\NormalTok{fun\_control.update(\{}\StringTok{"prep\_model"}\NormalTok{: prep\_model\})}
\end{Highlighting}
\end{Shaded}

A default approach for numerical data is the \texttt{StandardScaler}
(mean 0, variance 1). This can be selected as follows:

\begin{Shaded}
\begin{Highlighting}[]
\CommentTok{\# prep\_model = StandardScaler()}
\CommentTok{\# fun\_control.update(\{"prep\_model": prep\_model\})}
\end{Highlighting}
\end{Shaded}

Even more complicated pre-processing steps are possible, e.g., the
follwing pipeline:

\begin{Shaded}
\begin{Highlighting}[]
\CommentTok{\# categorical\_columns = []}
\CommentTok{\# one\_hot\_encoder = OneHotEncoder(handle\_unknown="ignore", sparse\_output=False)}
\CommentTok{\# prep\_model = ColumnTransformer(}
\CommentTok{\#         transformers=[}
\CommentTok{\#             ("categorical", one\_hot\_encoder, categorical\_columns),}
\CommentTok{\#         ],}
\CommentTok{\#         remainder=StandardScaler(),}
\CommentTok{\#     )}
\end{Highlighting}
\end{Shaded}

\hypertarget{step-5-select-model-algorithm-and-core_model_hyper_dict-1}{%
\section{\texorpdfstring{Step 5: Select Model (\texttt{algorithm}) and
\texttt{core\_model\_hyper\_dict}}{Step 5: Select Model (algorithm) and core\_model\_hyper\_dict}}\label{step-5-select-model-algorithm-and-core_model_hyper_dict-1}}

The selection of the algorithm (ML model) that should be tuned is done
by specifying the its name from the \texttt{sklearn} implementation. For
example, the \texttt{SVC} support vector machine classifier is selected
as follows:

\texttt{add\_core\_model\_to\_fun\_control(SVC,\ fun\_control,\ SklearnHyperDict)}

Other core\_models are, e.g.,:

\begin{itemize}
\tightlist
\item
  RidgeCV
\item
  GradientBoostingRegressor
\item
  ElasticNet
\item
  RandomForestClassifier
\item
  LogisticRegression
\item
  KNeighborsClassifier
\item
  RandomForestClassifier
\item
  GradientBoostingClassifier
\item
  HistGradientBoostingClassifier
\end{itemize}

We will use the \texttt{RandomForestClassifier} classifier in this
example.

\begin{Shaded}
\begin{Highlighting}[]
\ImportTok{from}\NormalTok{ sklearn.linear\_model }\ImportTok{import}\NormalTok{ RidgeCV}
\ImportTok{from}\NormalTok{ sklearn.ensemble }\ImportTok{import}\NormalTok{ RandomForestClassifier}
\ImportTok{from}\NormalTok{ sklearn.svm }\ImportTok{import}\NormalTok{ SVC}
\ImportTok{from}\NormalTok{ sklearn.linear\_model }\ImportTok{import}\NormalTok{ LogisticRegression}
\ImportTok{from}\NormalTok{ sklearn.neighbors }\ImportTok{import}\NormalTok{ KNeighborsClassifier}
\ImportTok{from}\NormalTok{ sklearn.ensemble }\ImportTok{import}\NormalTok{ GradientBoostingClassifier}
\ImportTok{from}\NormalTok{ sklearn.ensemble }\ImportTok{import}\NormalTok{ GradientBoostingRegressor}
\ImportTok{from}\NormalTok{ sklearn.linear\_model }\ImportTok{import}\NormalTok{ ElasticNet}
\ImportTok{from}\NormalTok{ spotPython.hyperparameters.values }\ImportTok{import}\NormalTok{ add\_core\_model\_to\_fun\_control}
\ImportTok{from}\NormalTok{ spotPython.data.sklearn\_hyper\_dict }\ImportTok{import}\NormalTok{ SklearnHyperDict}
\ImportTok{from}\NormalTok{ spotPython.fun.hypersklearn }\ImportTok{import}\NormalTok{ HyperSklearn}
\end{Highlighting}
\end{Shaded}

\begin{Shaded}
\begin{Highlighting}[]
\CommentTok{\# core\_model  = RidgeCV}
\CommentTok{\# core\_model = GradientBoostingRegressor}
\CommentTok{\# core\_model = ElasticNet}
\NormalTok{core\_model }\OperatorTok{=}\NormalTok{ RandomForestClassifier}
\CommentTok{\# core\_model = SVC}
\CommentTok{\# core\_model = LogisticRegression}
\CommentTok{\# core\_model = KNeighborsClassifier}
\CommentTok{\# core\_model = GradientBoostingClassifier}
\NormalTok{add\_core\_model\_to\_fun\_control(core\_model}\OperatorTok{=}\NormalTok{core\_model,}
\NormalTok{                              fun\_control}\OperatorTok{=}\NormalTok{fun\_control,}
\NormalTok{                              hyper\_dict}\OperatorTok{=}\NormalTok{SklearnHyperDict,}
\NormalTok{                              filename}\OperatorTok{=}\VariableTok{None}\NormalTok{)}
\end{Highlighting}
\end{Shaded}

Now \texttt{fun\_control} has the information from the JSON file. The
available hyperparameters are:

\begin{Shaded}
\begin{Highlighting}[]
\BuiltInTok{print}\NormalTok{(}\OperatorTok{*}\NormalTok{fun\_control[}\StringTok{"core\_model\_hyper\_dict"}\NormalTok{].keys(), sep}\OperatorTok{=}\StringTok{"}\CharTok{\textbackslash{}n}\StringTok{"}\NormalTok{)}
\end{Highlighting}
\end{Shaded}

\begin{verbatim}
n_estimators
criterion
max_depth
min_samples_split
min_samples_leaf
min_weight_fraction_leaf
max_features
max_leaf_nodes
min_impurity_decrease
bootstrap
oob_score
\end{verbatim}

\hypertarget{step-6-modify-hyper_dict-hyperparameters-for-the-selected-algorithm-aka-core_model-1}{%
\section{\texorpdfstring{Step 6: Modify \texttt{hyper\_dict}
Hyperparameters for the Selected Algorithm aka
\texttt{core\_model}}{Step 6: Modify hyper\_dict Hyperparameters for the Selected Algorithm aka core\_model}}\label{step-6-modify-hyper_dict-hyperparameters-for-the-selected-algorithm-aka-core_model-1}}

\hypertarget{modify-hyperparameter-of-type-numeric-and-integer-boolean-1}{%
\subsection{Modify hyperparameter of type numeric and integer
(boolean)}\label{modify-hyperparameter-of-type-numeric-and-integer-boolean-1}}

Numeric and boolean values can be modified using the
\texttt{modify\_hyper\_parameter\_bounds} method. For example, to change
the \texttt{tol} hyperparameter of the \texttt{SVC} model to the
interval {[}1e-3, 1e-2{]}, the following code can be used:

\texttt{modify\_hyper\_parameter\_bounds(fun\_control,\ "tol",\ bounds={[}1e-3,\ 1e-2{]})}

\begin{Shaded}
\begin{Highlighting}[]
\ImportTok{from}\NormalTok{ spotPython.hyperparameters.values }\ImportTok{import}\NormalTok{ modify\_hyper\_parameter\_bounds}
\CommentTok{\# modify\_hyper\_parameter\_bounds(fun\_control, "tol", bounds=[1e{-}3, 1e{-}2])}
\end{Highlighting}
\end{Shaded}

\hypertarget{modify-hyperparameter-of-type-factor-2}{%
\subsection{Modify hyperparameter of type
factor}\label{modify-hyperparameter-of-type-factor-2}}

\texttt{spotPython} provides functions for modifying the
hyperparameters, their bounds and factors as well as for activating and
de-activating hyperparameters without re-compilation of the Python
source code. These functions were described in
Section~\ref{sec-modification-of-hyperparameters-14}.

Factors can be modified with the
\texttt{modify\_hyper\_parameter\_levels} function. For example, to
exclude the \texttt{sigmoid} kernel from the tuning, the \texttt{kernel}
hyperparameter of the \texttt{SVC} model can be modified as follows:

\texttt{modify\_hyper\_parameter\_levels(fun\_control,\ "kernel",\ {[}"linear",\ "rbf"{]})}

The new setting can be controlled via:

\texttt{fun\_control{[}"core\_model\_hyper\_dict"{]}{[}"kernel"{]}}

\begin{Shaded}
\begin{Highlighting}[]
\ImportTok{from}\NormalTok{ spotPython.hyperparameters.values }\ImportTok{import}\NormalTok{ modify\_hyper\_parameter\_levels}
\CommentTok{\# XGBoost:}
\CommentTok{\# modify\_hyper\_parameter\_levels(fun\_control, "loss", ["log\_loss"])}
\end{Highlighting}
\end{Shaded}

\begin{tcolorbox}[enhanced jigsaw, left=2mm, title=\textcolor{quarto-callout-note-color}{\faInfo}\hspace{0.5em}{Note: RandomForestClassifier and Out-of-bag Estimation}, bottomrule=.15mm, titlerule=0mm, breakable, rightrule=.15mm, toprule=.15mm, coltitle=black, colbacktitle=quarto-callout-note-color!10!white, leftrule=.75mm, arc=.35mm, colframe=quarto-callout-note-color-frame, bottomtitle=1mm, colback=white, opacitybacktitle=0.6, toptitle=1mm, opacityback=0]

Since \texttt{oob\_score} requires the \texttt{bootstrap} hyperparameter
to \texttt{True}, we set the \texttt{oob\_score} parameter to
\texttt{False}. The \texttt{oob\_score} is later discussed in
Section~\ref{sec-oob-score-16}.

\end{tcolorbox}

\begin{Shaded}
\begin{Highlighting}[]
\NormalTok{modify\_hyper\_parameter\_bounds(fun\_control, }\StringTok{"bootstrap"}\NormalTok{, bounds}\OperatorTok{=}\NormalTok{[}\DecValTok{0}\NormalTok{, }\DecValTok{1}\NormalTok{])}
\NormalTok{modify\_hyper\_parameter\_bounds(fun\_control, }\StringTok{"oob\_score"}\NormalTok{, bounds}\OperatorTok{=}\NormalTok{[}\DecValTok{0}\NormalTok{, }\DecValTok{0}\NormalTok{])}
\end{Highlighting}
\end{Shaded}

\hypertarget{sec-optimizers-16}{%
\subsection{Optimizers}\label{sec-optimizers-16}}

Optimizers are described in Section~\ref{sec-optimizers-14}.

\hypertarget{selection-of-the-objective-metric-and-loss-functions}{%
\subsection{Selection of the Objective: Metric and Loss
Functions}\label{selection-of-the-objective-metric-and-loss-functions}}

\begin{itemize}
\tightlist
\item
  Machine learning models are optimized with respect to a metric, for
  example, the \texttt{accuracy} function.
\item
  Deep learning, e.g., neural networks are optimized with respect to a
  loss function, for example, the \texttt{cross\_entropy} function and
  evaluated with respect to a metric, for example, the \texttt{accuracy}
  function.
\end{itemize}

\hypertarget{step-7-selection-of-the-objective-loss-function-2}{%
\section{Step 7: Selection of the Objective (Loss)
Function}\label{step-7-selection-of-the-objective-loss-function-2}}

The loss function, that is usually used in deep learning for optimizing
the weights of the net, is stored in the \texttt{fun\_control}
dictionary as \texttt{"loss\_function"}.

\hypertarget{metric-function}{%
\subsection{Metric Function}\label{metric-function}}

There are two different types of metrics in \texttt{spotPython}:

\begin{enumerate}
\def\labelenumi{\arabic{enumi}.}
\tightlist
\item
  \texttt{"metric\_river"} is used for the river based evaluation via
  \texttt{eval\_oml\_iter\_progressive}.
\item
  \texttt{"metric\_sklearn"} is used for the sklearn based evaluation.
\end{enumerate}

We will consider multi-class classification metrics, e.g.,
\texttt{mapk\_score} and \texttt{top\_k\_accuracy\_score}.

\begin{tcolorbox}[enhanced jigsaw, left=2mm, title=\textcolor{quarto-callout-note-color}{\faInfo}\hspace{0.5em}{Predict Probabilities}, bottomrule=.15mm, titlerule=0mm, breakable, rightrule=.15mm, toprule=.15mm, coltitle=black, colbacktitle=quarto-callout-note-color!10!white, leftrule=.75mm, arc=.35mm, colframe=quarto-callout-note-color-frame, bottomtitle=1mm, colback=white, opacitybacktitle=0.6, toptitle=1mm, opacityback=0]

In this multi-class classification example the machine learning
algorithm should return the probabilities of the specific classes
(\texttt{"predict\_proba"}) instead of the predicted values.

\end{tcolorbox}

We set \texttt{"predict\_proba"} to \texttt{True} in the
\texttt{fun\_control} dictionary.

\hypertarget{the-mapk-metric}{%
\subsubsection{The MAPK Metric}\label{the-mapk-metric}}

To select the MAPK metric, the following two entries can be added to the
\texttt{fun\_control} dictionary:

\texttt{"metric\_sklearn":\ mapk\_score"}

\texttt{"metric\_params":\ \{"k":\ 3\}}.

\hypertarget{other-metrics}{%
\subsubsection{Other Metrics}\label{other-metrics}}

Alternatively, other metrics for multi-class classification can be used,
e.g.,: * top\_k\_accuracy\_score or * roc\_auc\_score

The metric \texttt{roc\_auc\_score} requires the parameter
\texttt{"multi\_class"}, e.g.,

\texttt{"multi\_class":\ "ovr"}.

This is set in the \texttt{fun\_control} dictionary.

\begin{tcolorbox}[enhanced jigsaw, left=2mm, title=\textcolor{quarto-callout-note-color}{\faInfo}\hspace{0.5em}{Weights}, bottomrule=.15mm, titlerule=0mm, breakable, rightrule=.15mm, toprule=.15mm, coltitle=black, colbacktitle=quarto-callout-note-color!10!white, leftrule=.75mm, arc=.35mm, colframe=quarto-callout-note-color-frame, bottomtitle=1mm, colback=white, opacitybacktitle=0.6, toptitle=1mm, opacityback=0]

\texttt{spotPython} performs a minimization, therefore, metrics that
should be maximized have to be multiplied by -1. This is done by setting
\texttt{"weights"} to \texttt{-1}.

\end{tcolorbox}

\begin{itemize}
\tightlist
\item
  The complete setup for the metric in our example is:
\end{itemize}

\begin{Shaded}
\begin{Highlighting}[]
\ImportTok{from}\NormalTok{ spotPython.utils.metrics }\ImportTok{import}\NormalTok{ mapk\_score}
\NormalTok{fun\_control.update(\{}
               \StringTok{"weights"}\NormalTok{: }\OperatorTok{{-}}\DecValTok{1}\NormalTok{,}
               \StringTok{"metric\_sklearn"}\NormalTok{: mapk\_score,}
               \StringTok{"predict\_proba"}\NormalTok{: }\VariableTok{True}\NormalTok{,}
               \StringTok{"metric\_params"}\NormalTok{: \{}\StringTok{"k"}\NormalTok{: }\DecValTok{3}\NormalTok{\},}
\NormalTok{               \})}
\end{Highlighting}
\end{Shaded}

\hypertarget{evaluation-on-hold-out-data}{%
\subsection{Evaluation on Hold-out
Data}\label{evaluation-on-hold-out-data}}

\begin{itemize}
\tightlist
\item
  The default method for computing the performance is
  \texttt{"eval\_holdout"}.
\item
  Alternatively, cross-validation can be used for every machine learning
  model.
\item
  Specifically for RandomForests, the OOB-score can be used.
\end{itemize}

\begin{Shaded}
\begin{Highlighting}[]
\NormalTok{fun\_control.update(\{}
    \StringTok{"eval"}\NormalTok{: }\StringTok{"train\_hold\_out"}\NormalTok{,}
\NormalTok{\})}
\end{Highlighting}
\end{Shaded}

\hypertarget{sec-oob-score-16}{%
\subsection{OOB Score}\label{sec-oob-score-16}}

Using the OOB-Score is a very efficient way to estimate the performance
of a random forest classifier. The OOB-Score is calculated on the
training data and does not require a hold-out test set. If the OOB-Score
is used, the key ``eval'' in the \texttt{fun\_control} dictionary should
be set to \texttt{"oob\_score"} as shown below.

\begin{tcolorbox}[enhanced jigsaw, left=2mm, title=\textcolor{quarto-callout-note-color}{\faInfo}\hspace{0.5em}{OOB-Score}, bottomrule=.15mm, titlerule=0mm, breakable, rightrule=.15mm, toprule=.15mm, coltitle=black, colbacktitle=quarto-callout-note-color!10!white, leftrule=.75mm, arc=.35mm, colframe=quarto-callout-note-color-frame, bottomtitle=1mm, colback=white, opacitybacktitle=0.6, toptitle=1mm, opacityback=0]

In addition to setting the key \texttt{"eval"} in the
\texttt{fun\_control} dictionary to \texttt{"oob\_score"}, the keys
\texttt{"oob\_score"} and \texttt{"bootstrap"} have to be set to
\texttt{True}, because the OOB-Score requires the bootstrap method.

\end{tcolorbox}

\begin{itemize}
\tightlist
\item
  Uncomment the following lines to use the OOB-Score:
\end{itemize}

\begin{Shaded}
\begin{Highlighting}[]
\NormalTok{fun\_control.update(\{}
    \StringTok{"eval"}\NormalTok{: }\StringTok{"eval\_oob\_score"}\NormalTok{,}
\NormalTok{\})}
\NormalTok{modify\_hyper\_parameter\_bounds(fun\_control, }\StringTok{"bootstrap"}\NormalTok{, bounds}\OperatorTok{=}\NormalTok{[}\DecValTok{1}\NormalTok{, }\DecValTok{1}\NormalTok{])}
\NormalTok{modify\_hyper\_parameter\_bounds(fun\_control, }\StringTok{"oob\_score"}\NormalTok{, bounds}\OperatorTok{=}\NormalTok{[}\DecValTok{1}\NormalTok{, }\DecValTok{1}\NormalTok{])}
\end{Highlighting}
\end{Shaded}

\hypertarget{cross-validation-1}{%
\subsubsection{Cross Validation}\label{cross-validation-1}}

Instead of using the OOB-score, the classical cross validation can be
used. The number of folds is set by the key \texttt{"k\_folds"}. For
example, to use 5-fold cross validation, the key \texttt{"k\_folds"} is
set to \texttt{5}. Uncomment the following line to use cross validation:

\begin{Shaded}
\begin{Highlighting}[]
\CommentTok{\# fun\_control.update(\{}
\CommentTok{\#      "eval": "train\_cv",}
\CommentTok{\#      "k\_folds": 10,}
\CommentTok{\# \})}
\end{Highlighting}
\end{Shaded}

\hypertarget{step-8-calling-the-spot-function-2}{%
\section{Step 8: Calling the SPOT
Function}\label{step-8-calling-the-spot-function-2}}

\hypertarget{sec-prepare-spot-call-16}{%
\subsection{Preparing the SPOT Call}\label{sec-prepare-spot-call-16}}

\begin{itemize}
\tightlist
\item
  Get types and variable names as well as lower and upper bounds for the
  hyperparameters.
\end{itemize}

\begin{Shaded}
\begin{Highlighting}[]
\CommentTok{\# extract the variable types, names, and bounds}
\ImportTok{from}\NormalTok{ spotPython.hyperparameters.values }\ImportTok{import}\NormalTok{ (get\_bound\_values,}
\NormalTok{    get\_var\_name,}
\NormalTok{    get\_var\_type,)}
\NormalTok{var\_type }\OperatorTok{=}\NormalTok{ get\_var\_type(fun\_control)}
\NormalTok{var\_name }\OperatorTok{=}\NormalTok{ get\_var\_name(fun\_control)}
\NormalTok{lower }\OperatorTok{=}\NormalTok{ get\_bound\_values(fun\_control, }\StringTok{"lower"}\NormalTok{)}
\NormalTok{upper }\OperatorTok{=}\NormalTok{ get\_bound\_values(fun\_control, }\StringTok{"upper"}\NormalTok{)}
\end{Highlighting}
\end{Shaded}

\begin{Shaded}
\begin{Highlighting}[]
\ImportTok{from}\NormalTok{ spotPython.utils.eda }\ImportTok{import}\NormalTok{ gen\_design\_table}
\BuiltInTok{print}\NormalTok{(gen\_design\_table(fun\_control))}
\end{Highlighting}
\end{Shaded}

\begin{verbatim}
| name                     | type   | default   |   lower |   upper | transform              |
|--------------------------|--------|-----------|---------|---------|------------------------|
| n_estimators             | int    | 7         |       5 |   10    | transform_power_2_int  |
| criterion                | factor | gini      |       0 |    2    | None                   |
| max_depth                | int    | 10        |       1 |   20    | transform_power_2_int  |
| min_samples_split        | int    | 2         |       2 |  100    | None                   |
| min_samples_leaf         | int    | 1         |       1 |   25    | None                   |
| min_weight_fraction_leaf | float  | 0.0       |       0 |    0.01 | None                   |
| max_features             | factor | sqrt      |       0 |    1    | transform_none_to_None |
| max_leaf_nodes           | int    | 10        |       7 |   12    | transform_power_2_int  |
| min_impurity_decrease    | float  | 0.0       |       0 |    0.01 | None                   |
| bootstrap                | factor | 1         |       1 |    1    | None                   |
| oob_score                | factor | 0         |       1 |    1    | None                   |
\end{verbatim}

\hypertarget{sec-the-objective-function-16}{%
\subsection{The Objective
Function}\label{sec-the-objective-function-16}}

The objective function is selected next. It implements an interface from
\texttt{sklearn}'s training, validation, and testing methods to
\texttt{spotPython}.

\begin{Shaded}
\begin{Highlighting}[]
\ImportTok{from}\NormalTok{ spotPython.fun.hypersklearn }\ImportTok{import}\NormalTok{ HyperSklearn}
\NormalTok{fun }\OperatorTok{=}\NormalTok{ HyperSklearn().fun\_sklearn}
\end{Highlighting}
\end{Shaded}

\hypertarget{run-the-spot-optimizer-2}{%
\subsection{\texorpdfstring{Run the \texttt{Spot}
Optimizer}{Run the Spot Optimizer}}\label{run-the-spot-optimizer-2}}

\begin{itemize}
\tightlist
\item
  Run SPOT for approx. x mins (\texttt{max\_time}).
\item
  Note: the run takes longer, because the evaluation time of initial
  design (here: \texttt{initi\_size}, 20 points) is not considered.
\end{itemize}

\begin{Shaded}
\begin{Highlighting}[]
\ImportTok{from}\NormalTok{ spotPython.hyperparameters.values }\ImportTok{import}\NormalTok{ get\_default\_hyperparameters\_as\_array}
\NormalTok{X\_start }\OperatorTok{=}\NormalTok{ get\_default\_hyperparameters\_as\_array(fun\_control)}
\NormalTok{X\_start}
\end{Highlighting}
\end{Shaded}

\begin{verbatim}
array([[ 7.,  0., 10.,  2.,  1.,  0.,  0., 10.,  0.,  1.,  0.]])
\end{verbatim}

\begin{Shaded}
\begin{Highlighting}[]
\ImportTok{import}\NormalTok{ numpy }\ImportTok{as}\NormalTok{ np}
\ImportTok{from}\NormalTok{ spotPython.spot }\ImportTok{import}\NormalTok{ spot}
\ImportTok{from}\NormalTok{ math }\ImportTok{import}\NormalTok{ inf}
\NormalTok{spot\_tuner }\OperatorTok{=}\NormalTok{ spot.Spot(fun}\OperatorTok{=}\NormalTok{fun,}
\NormalTok{                   lower }\OperatorTok{=}\NormalTok{ lower,}
\NormalTok{                   upper }\OperatorTok{=}\NormalTok{ upper,}
\NormalTok{                   fun\_evals }\OperatorTok{=}\NormalTok{ inf,}
\NormalTok{                   fun\_repeats }\OperatorTok{=} \DecValTok{1}\NormalTok{,}
\NormalTok{                   max\_time }\OperatorTok{=}\NormalTok{ MAX\_TIME,}
\NormalTok{                   noise }\OperatorTok{=} \VariableTok{False}\NormalTok{,}
\NormalTok{                   tolerance\_x }\OperatorTok{=}\NormalTok{ np.sqrt(np.spacing(}\DecValTok{1}\NormalTok{)),}
\NormalTok{                   var\_type }\OperatorTok{=}\NormalTok{ var\_type,}
\NormalTok{                   var\_name }\OperatorTok{=}\NormalTok{ var\_name,}
\NormalTok{                   infill\_criterion }\OperatorTok{=} \StringTok{"y"}\NormalTok{,}
\NormalTok{                   n\_points }\OperatorTok{=} \DecValTok{1}\NormalTok{,}
\NormalTok{                   seed}\OperatorTok{=}\DecValTok{123}\NormalTok{,}
\NormalTok{                   log\_level }\OperatorTok{=} \DecValTok{50}\NormalTok{,}
\NormalTok{                   show\_models}\OperatorTok{=} \VariableTok{False}\NormalTok{,}
\NormalTok{                   show\_progress}\OperatorTok{=} \VariableTok{True}\NormalTok{,}
\NormalTok{                   fun\_control }\OperatorTok{=}\NormalTok{ fun\_control,}
\NormalTok{                   design\_control}\OperatorTok{=}\NormalTok{\{}\StringTok{"init\_size"}\NormalTok{: INIT\_SIZE,}
                                   \StringTok{"repeats"}\NormalTok{: }\DecValTok{1}\NormalTok{\},}
\NormalTok{                   surrogate\_control}\OperatorTok{=}\NormalTok{\{}\StringTok{"noise"}\NormalTok{: }\VariableTok{True}\NormalTok{,}
                                      \StringTok{"cod\_type"}\NormalTok{: }\StringTok{"norm"}\NormalTok{,}
                                      \StringTok{"min\_theta"}\NormalTok{: }\OperatorTok{{-}}\DecValTok{4}\NormalTok{,}
                                      \StringTok{"max\_theta"}\NormalTok{: }\DecValTok{3}\NormalTok{,}
                                      \StringTok{"n\_theta"}\NormalTok{: }\BuiltInTok{len}\NormalTok{(var\_name),}
                                      \StringTok{"model\_fun\_evals"}\NormalTok{: }\DecValTok{10\_000}\NormalTok{,}
                                      \StringTok{"log\_level"}\NormalTok{: }\DecValTok{50}
\NormalTok{                                      \})}
\NormalTok{spot\_tuner.run(X\_start}\OperatorTok{=}\NormalTok{X\_start)}
\end{Highlighting}
\end{Shaded}

\begin{verbatim}
spotPython tuning: -0.8544973544973545 [----------] 1.76% 
\end{verbatim}

\begin{verbatim}
spotPython tuning: -0.8544973544973545 [----------] 3.80% 
\end{verbatim}

\begin{verbatim}
spotPython tuning: -0.8641975308641975 [#---------] 6.92% 
\end{verbatim}

\begin{verbatim}
spotPython tuning: -0.8641975308641975 [#---------] 9.12% 
\end{verbatim}

\begin{verbatim}
spotPython tuning: -0.8641975308641975 [#---------] 10.69% 
\end{verbatim}

\begin{verbatim}
spotPython tuning: -0.8641975308641975 [#---------] 12.42% 
\end{verbatim}

\begin{verbatim}
spotPython tuning: -0.8659611992945327 [#---------] 14.82% 
\end{verbatim}

\begin{verbatim}
spotPython tuning: -0.8686067019400352 [##--------] 17.34% 
\end{verbatim}

\begin{verbatim}
spotPython tuning: -0.8712522045855379 [##--------] 19.92% 
\end{verbatim}

\begin{verbatim}
spotPython tuning: -0.8712522045855379 [##--------] 22.59% 
\end{verbatim}

\begin{verbatim}
spotPython tuning: -0.8712522045855379 [###-------] 25.38% 
\end{verbatim}

\begin{verbatim}
spotPython tuning: -0.8712522045855379 [###-------] 27.49% 
\end{verbatim}

\begin{verbatim}
spotPython tuning: -0.8712522045855379 [###-------] 33.54% 
\end{verbatim}

\begin{verbatim}
spotPython tuning: -0.8712522045855379 [####------] 40.15% 
\end{verbatim}

\begin{verbatim}
spotPython tuning: -0.879188712522046 [#####-----] 46.73% 
\end{verbatim}

\begin{verbatim}
spotPython tuning: -0.879188712522046 [#####-----] 52.89% 
\end{verbatim}

\begin{verbatim}
spotPython tuning: -0.879188712522046 [######----] 58.32% 
\end{verbatim}

\begin{verbatim}
spotPython tuning: -0.879188712522046 [######----] 63.55% 
\end{verbatim}

\begin{verbatim}
spotPython tuning: -0.879188712522046 [#######---] 69.01% 
\end{verbatim}

\begin{verbatim}
spotPython tuning: -0.879188712522046 [########--] 76.03% 
\end{verbatim}

\begin{verbatim}
spotPython tuning: -0.879188712522046 [########--] 82.55% 
\end{verbatim}

\begin{verbatim}
spotPython tuning: -0.879188712522046 [#########-] 88.23% 
\end{verbatim}

\begin{verbatim}
spotPython tuning: -0.879188712522046 [#########-] 94.82% 
\end{verbatim}

\begin{verbatim}
spotPython tuning: -0.879188712522046 [##########] 100.00% Done...
\end{verbatim}

\begin{verbatim}
<spotPython.spot.spot.Spot at 0x17fd5ace0>
\end{verbatim}

\hypertarget{sec-tensorboard-16}{%
\section{Step 9: Tensorboard}\label{sec-tensorboard-16}}

The textual output shown in the console (or code cell) can be visualized
with Tensorboard as described in Section~\ref{sec-tensorboard-14}, see
also the description in the documentation:
\href{https://sequential-parameter-optimization.github.io/spotPython/14_spot_ray_hpt_torch_cifar10.html\#sec-tensorboard-14}{Tensorboard.}

\hypertarget{sec-results-tuning-16}{%
\section{Step 10: Results}\label{sec-results-tuning-16}}

After the hyperparameter tuning run is finished, the progress of the
hyperparameter tuning can be visualized. The following code generates
the progress plot from \textbf{?@fig-progress}.

\begin{Shaded}
\begin{Highlighting}[]
\NormalTok{spot\_tuner.plot\_progress(log\_y}\OperatorTok{=}\VariableTok{False}\NormalTok{,}
\NormalTok{    filename}\OperatorTok{=}\StringTok{"./figures/"} \OperatorTok{+}\NormalTok{ experiment\_name}\OperatorTok{+}\StringTok{"\_progress.png"}\NormalTok{)}
\end{Highlighting}
\end{Shaded}

\begin{figure}[H]

{\centering \includegraphics{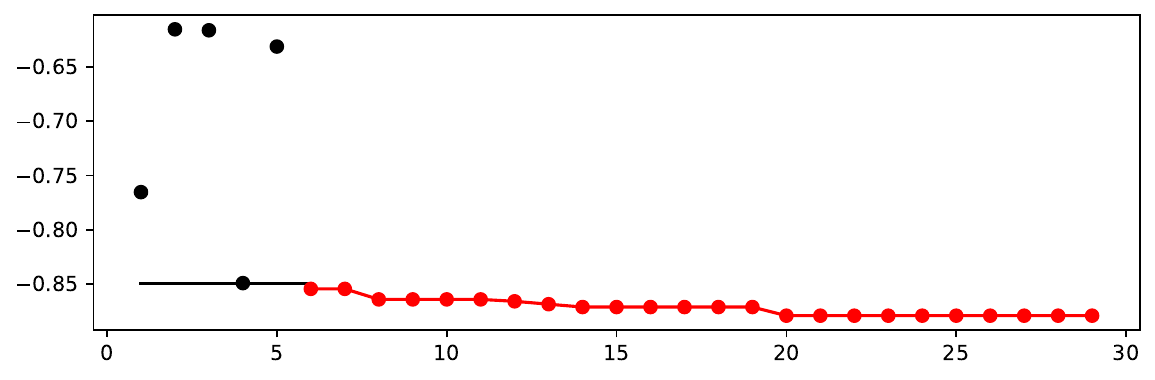}

}

\caption{Progress plot. \emph{Black} dots denote results from the
initial design. \emph{Red} dots illustrate the improvement found by the
surrogate model based optimization.}

\end{figure}

\begin{itemize}
\tightlist
\item
  Print the results
\end{itemize}

\begin{Shaded}
\begin{Highlighting}[]
\BuiltInTok{print}\NormalTok{(gen\_design\_table(fun\_control}\OperatorTok{=}\NormalTok{fun\_control,}
\NormalTok{    spot}\OperatorTok{=}\NormalTok{spot\_tuner))}
\end{Highlighting}
\end{Shaded}

\begin{verbatim}
| name                     | type   | default   |   lower |   upper |                tuned | transform              |   importance | stars   |
|--------------------------|--------|-----------|---------|---------|----------------------|------------------------|--------------|---------|
| n_estimators             | int    | 7         |     5.0 |    10.0 |                 10.0 | transform_power_2_int  |         0.13 | .       |
| criterion                | factor | gini      |     0.0 |     2.0 |                  1.0 | None                   |         0.00 |         |
| max_depth                | int    | 10        |     1.0 |    20.0 |                  4.0 | transform_power_2_int  |         0.24 | .       |
| min_samples_split        | int    | 2         |     2.0 |   100.0 |                  2.0 | None                   |         1.35 | *       |
| min_samples_leaf         | int    | 1         |     1.0 |    25.0 |                  1.0 | None                   |         0.19 | .       |
| min_weight_fraction_leaf | float  | 0.0       |     0.0 |    0.01 | 0.003002545876925399 | None                   |         0.00 |         |
| max_features             | factor | sqrt      |     0.0 |     1.0 |                  0.0 | transform_none_to_None |       100.00 | ***     |
| max_leaf_nodes           | int    | 10        |     7.0 |    12.0 |                 10.0 | transform_power_2_int  |         0.00 |         |
| min_impurity_decrease    | float  | 0.0       |     0.0 |    0.01 | 0.005762348549695934 | None                   |         0.00 |         |
| bootstrap                | factor | 1         |     1.0 |     1.0 |                  1.0 | None                   |         0.00 |         |
| oob_score                | factor | 0         |     1.0 |     1.0 |                  1.0 | None                   |         0.00 |         |
\end{verbatim}

\hypertarget{show-variable-importance-1}{%
\subsection{Show variable importance}\label{show-variable-importance-1}}

\begin{Shaded}
\begin{Highlighting}[]
\NormalTok{spot\_tuner.plot\_importance(threshold}\OperatorTok{=}\FloatTok{0.025}\NormalTok{, filename}\OperatorTok{=}\StringTok{"./figures/"} \OperatorTok{+}\NormalTok{ experiment\_name}\OperatorTok{+}\StringTok{"\_importance.png"}\NormalTok{)}
\end{Highlighting}
\end{Shaded}

\begin{figure}[H]

{\centering \includegraphics{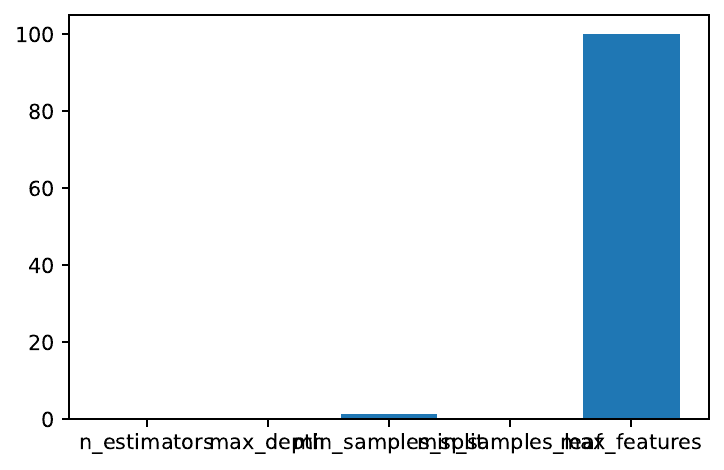}

}

\caption{Variable importance plot, threshold 0.025.}

\end{figure}

\hypertarget{get-default-hyperparameters-3}{%
\subsection{Get Default
Hyperparameters}\label{get-default-hyperparameters-3}}

\begin{Shaded}
\begin{Highlighting}[]
\ImportTok{from}\NormalTok{ spotPython.hyperparameters.values }\ImportTok{import}\NormalTok{ get\_default\_values, transform\_hyper\_parameter\_values}
\NormalTok{values\_default }\OperatorTok{=}\NormalTok{ get\_default\_values(fun\_control)}
\NormalTok{values\_default }\OperatorTok{=}\NormalTok{ transform\_hyper\_parameter\_values(fun\_control}\OperatorTok{=}\NormalTok{fun\_control, hyper\_parameter\_values}\OperatorTok{=}\NormalTok{values\_default)}
\NormalTok{values\_default}
\end{Highlighting}
\end{Shaded}

\begin{verbatim}
{'n_estimators': 128,
 'criterion': 'gini',
 'max_depth': 1024,
 'min_samples_split': 2,
 'min_samples_leaf': 1,
 'min_weight_fraction_leaf': 0.0,
 'max_features': 'sqrt',
 'max_leaf_nodes': 1024,
 'min_impurity_decrease': 0.0,
 'bootstrap': 1,
 'oob_score': 0}
\end{verbatim}

\begin{Shaded}
\begin{Highlighting}[]
\ImportTok{from}\NormalTok{ sklearn.pipeline }\ImportTok{import}\NormalTok{ make\_pipeline}
\NormalTok{model\_default }\OperatorTok{=}\NormalTok{ make\_pipeline(fun\_control[}\StringTok{"prep\_model"}\NormalTok{], fun\_control[}\StringTok{"core\_model"}\NormalTok{](}\OperatorTok{**}\NormalTok{values\_default))}
\NormalTok{model\_default}
\end{Highlighting}
\end{Shaded}

\begin{verbatim}
Pipeline(steps=[('nonetype', None),
                ('randomforestclassifier',
                 RandomForestClassifier(bootstrap=1, max_depth=1024,
                                        max_leaf_nodes=1024, n_estimators=128,
                                        oob_score=0))])
\end{verbatim}

\hypertarget{get-spot-results-2}{%
\subsection{Get SPOT Results}\label{get-spot-results-2}}

\begin{Shaded}
\begin{Highlighting}[]
\NormalTok{X }\OperatorTok{=}\NormalTok{ spot\_tuner.to\_all\_dim(spot\_tuner.min\_X.reshape(}\DecValTok{1}\NormalTok{,}\OperatorTok{{-}}\DecValTok{1}\NormalTok{))}
\BuiltInTok{print}\NormalTok{(X)}
\end{Highlighting}
\end{Shaded}

\begin{verbatim}
[[1.00000000e+01 1.00000000e+00 4.00000000e+00 2.00000000e+00
  1.00000000e+00 3.00254588e-03 0.00000000e+00 1.00000000e+01
  5.76234855e-03 1.00000000e+00 1.00000000e+00]]
\end{verbatim}

\begin{Shaded}
\begin{Highlighting}[]
\ImportTok{from}\NormalTok{ spotPython.hyperparameters.values }\ImportTok{import}\NormalTok{ assign\_values, return\_conf\_list\_from\_var\_dict}
\NormalTok{v\_dict }\OperatorTok{=}\NormalTok{ assign\_values(X, fun\_control[}\StringTok{"var\_name"}\NormalTok{])}
\NormalTok{return\_conf\_list\_from\_var\_dict(var\_dict}\OperatorTok{=}\NormalTok{v\_dict, fun\_control}\OperatorTok{=}\NormalTok{fun\_control)}
\end{Highlighting}
\end{Shaded}

\begin{verbatim}
[{'n_estimators': 1024,
  'criterion': 'entropy',
  'max_depth': 16,
  'min_samples_split': 2,
  'min_samples_leaf': 1,
  'min_weight_fraction_leaf': 0.003002545876925399,
  'max_features': 'sqrt',
  'max_leaf_nodes': 1024,
  'min_impurity_decrease': 0.005762348549695934,
  'bootstrap': 1,
  'oob_score': 1}]
\end{verbatim}

\begin{Shaded}
\begin{Highlighting}[]
\ImportTok{from}\NormalTok{ spotPython.hyperparameters.values }\ImportTok{import}\NormalTok{ get\_one\_sklearn\_model\_from\_X}
\NormalTok{model\_spot }\OperatorTok{=}\NormalTok{ get\_one\_sklearn\_model\_from\_X(X, fun\_control)}
\NormalTok{model\_spot}
\end{Highlighting}
\end{Shaded}

\begin{verbatim}
RandomForestClassifier(bootstrap=1, criterion='entropy', max_depth=16,
                       max_leaf_nodes=1024,
                       min_impurity_decrease=0.005762348549695934,
                       min_weight_fraction_leaf=0.003002545876925399,
                       n_estimators=1024, oob_score=1)
\end{verbatim}

\hypertarget{evaluate-spot-results}{%
\subsection{Evaluate SPOT Results}\label{evaluate-spot-results}}

\begin{itemize}
\tightlist
\item
  Fetch the data.
\end{itemize}

\begin{Shaded}
\begin{Highlighting}[]
\ImportTok{from}\NormalTok{ spotPython.utils.convert }\ImportTok{import}\NormalTok{ get\_Xy\_from\_df}
\NormalTok{X\_train, y\_train }\OperatorTok{=}\NormalTok{ get\_Xy\_from\_df(fun\_control[}\StringTok{"train"}\NormalTok{], fun\_control[}\StringTok{"target\_column"}\NormalTok{])}
\NormalTok{X\_test, y\_test }\OperatorTok{=}\NormalTok{ get\_Xy\_from\_df(fun\_control[}\StringTok{"test"}\NormalTok{], fun\_control[}\StringTok{"target\_column"}\NormalTok{])}
\NormalTok{X\_test.shape, y\_test.shape}
\end{Highlighting}
\end{Shaded}

\begin{verbatim}
((63, 64), (63,))
\end{verbatim}

\begin{itemize}
\tightlist
\item
  Fit the model with the tuned hyperparameters. This gives one result:
\end{itemize}

\begin{Shaded}
\begin{Highlighting}[]
\NormalTok{model\_spot.fit(X\_train, y\_train)}
\NormalTok{y\_pred }\OperatorTok{=}\NormalTok{ model\_spot.predict\_proba(X\_test)}
\NormalTok{res }\OperatorTok{=}\NormalTok{ mapk\_score(y\_true}\OperatorTok{=}\NormalTok{y\_test, y\_pred}\OperatorTok{=}\NormalTok{y\_pred, k}\OperatorTok{=}\DecValTok{3}\NormalTok{)}
\NormalTok{res}
\end{Highlighting}
\end{Shaded}

\begin{verbatim}
0.8465608465608465
\end{verbatim}

\begin{Shaded}
\begin{Highlighting}[]
\KeywordTok{def}\NormalTok{ repeated\_eval(n, model):}
\NormalTok{    res\_values }\OperatorTok{=}\NormalTok{ []}
    \ControlFlowTok{for}\NormalTok{ i }\KeywordTok{in} \BuiltInTok{range}\NormalTok{(n):}
\NormalTok{        model.fit(X\_train, y\_train)}
\NormalTok{        y\_pred }\OperatorTok{=}\NormalTok{ model.predict\_proba(X\_test)}
\NormalTok{        res }\OperatorTok{=}\NormalTok{ mapk\_score(y\_true}\OperatorTok{=}\NormalTok{y\_test, y\_pred}\OperatorTok{=}\NormalTok{y\_pred, k}\OperatorTok{=}\DecValTok{3}\NormalTok{)}
\NormalTok{        res\_values.append(res)}
\NormalTok{    mean\_res }\OperatorTok{=}\NormalTok{ np.mean(res\_values)}
    \BuiltInTok{print}\NormalTok{(}\SpecialStringTok{f"mean\_res: }\SpecialCharTok{\{}\NormalTok{mean\_res}\SpecialCharTok{\}}\SpecialStringTok{"}\NormalTok{)}
\NormalTok{    std\_res }\OperatorTok{=}\NormalTok{ np.std(res\_values)}
    \BuiltInTok{print}\NormalTok{(}\SpecialStringTok{f"std\_res: }\SpecialCharTok{\{}\NormalTok{std\_res}\SpecialCharTok{\}}\SpecialStringTok{"}\NormalTok{)}
\NormalTok{    min\_res }\OperatorTok{=}\NormalTok{ np.}\BuiltInTok{min}\NormalTok{(res\_values)}
    \BuiltInTok{print}\NormalTok{(}\SpecialStringTok{f"min\_res: }\SpecialCharTok{\{}\NormalTok{min\_res}\SpecialCharTok{\}}\SpecialStringTok{"}\NormalTok{)}
\NormalTok{    max\_res }\OperatorTok{=}\NormalTok{ np.}\BuiltInTok{max}\NormalTok{(res\_values)}
    \BuiltInTok{print}\NormalTok{(}\SpecialStringTok{f"max\_res: }\SpecialCharTok{\{}\NormalTok{max\_res}\SpecialCharTok{\}}\SpecialStringTok{"}\NormalTok{)}
\NormalTok{    median\_res }\OperatorTok{=}\NormalTok{ np.median(res\_values)}
    \BuiltInTok{print}\NormalTok{(}\SpecialStringTok{f"median\_res: }\SpecialCharTok{\{}\NormalTok{median\_res}\SpecialCharTok{\}}\SpecialStringTok{"}\NormalTok{)}
    \ControlFlowTok{return}\NormalTok{ mean\_res, std\_res, min\_res, max\_res, median\_res}
\end{Highlighting}
\end{Shaded}

\hypertarget{handling-non-deterministic-results}{%
\subsection{Handling Non-deterministic
Results}\label{handling-non-deterministic-results}}

\begin{itemize}
\tightlist
\item
  Because the model is non-determinstic, we perform \(n=30\) runs and
  calculate the mean and standard deviation of the performance metric.
\end{itemize}

\begin{Shaded}
\begin{Highlighting}[]
\NormalTok{\_ }\OperatorTok{=}\NormalTok{ repeated\_eval(}\DecValTok{30}\NormalTok{, model\_spot)}
\end{Highlighting}
\end{Shaded}

\begin{verbatim}
mean_res: 0.8524691358024691
std_res: 0.006887909906588555
min_res: 0.8386243386243385
max_res: 0.8703703703703703
median_res: 0.8544973544973544
\end{verbatim}

\hypertarget{evalution-of-the-default-hyperparameters}{%
\subsection{Evalution of the Default
Hyperparameters}\label{evalution-of-the-default-hyperparameters}}

\begin{Shaded}
\begin{Highlighting}[]
\NormalTok{model\_default.fit(X\_train, y\_train)[}\StringTok{"randomforestclassifier"}\NormalTok{]}
\end{Highlighting}
\end{Shaded}

\begin{verbatim}
RandomForestClassifier(bootstrap=1, max_depth=1024, max_leaf_nodes=1024,
                       n_estimators=128, oob_score=0)
\end{verbatim}

\begin{itemize}
\tightlist
\item
  One evaluation of the default hyperparameters is performed on the
  hold-out test set.
\end{itemize}

\begin{Shaded}
\begin{Highlighting}[]
\NormalTok{y\_pred }\OperatorTok{=}\NormalTok{ model\_default.predict\_proba(X\_test)}
\NormalTok{mapk\_score(y\_true}\OperatorTok{=}\NormalTok{y\_test, y\_pred}\OperatorTok{=}\NormalTok{y\_pred, k}\OperatorTok{=}\DecValTok{3}\NormalTok{)}
\end{Highlighting}
\end{Shaded}

\begin{verbatim}
0.8597883597883599
\end{verbatim}

Since one single evaluation is not meaningful, we perform, similar to
the evaluation of the SPOT results, \(n=30\) runs of the default setting
and and calculate the mean and standard deviation of the performance
metric.

\begin{Shaded}
\begin{Highlighting}[]
\NormalTok{\_ }\OperatorTok{=}\NormalTok{ repeated\_eval(}\DecValTok{30}\NormalTok{, model\_default)}
\end{Highlighting}
\end{Shaded}

\begin{verbatim}
mean_res: 0.8509700176366846
std_res: 0.011538220270755644
min_res: 0.828042328042328
max_res: 0.8703703703703703
median_res: 0.8505291005291006
\end{verbatim}

\hypertarget{plot-compare-predictions-1}{%
\subsection{Plot: Compare
Predictions}\label{plot-compare-predictions-1}}

\begin{Shaded}
\begin{Highlighting}[]
\ImportTok{from}\NormalTok{ spotPython.plot.validation }\ImportTok{import}\NormalTok{ plot\_confusion\_matrix}
\NormalTok{plot\_confusion\_matrix(model\_default, fun\_control, title }\OperatorTok{=} \StringTok{"Default"}\NormalTok{)}
\end{Highlighting}
\end{Shaded}

\begin{figure}[H]

{\centering \includegraphics{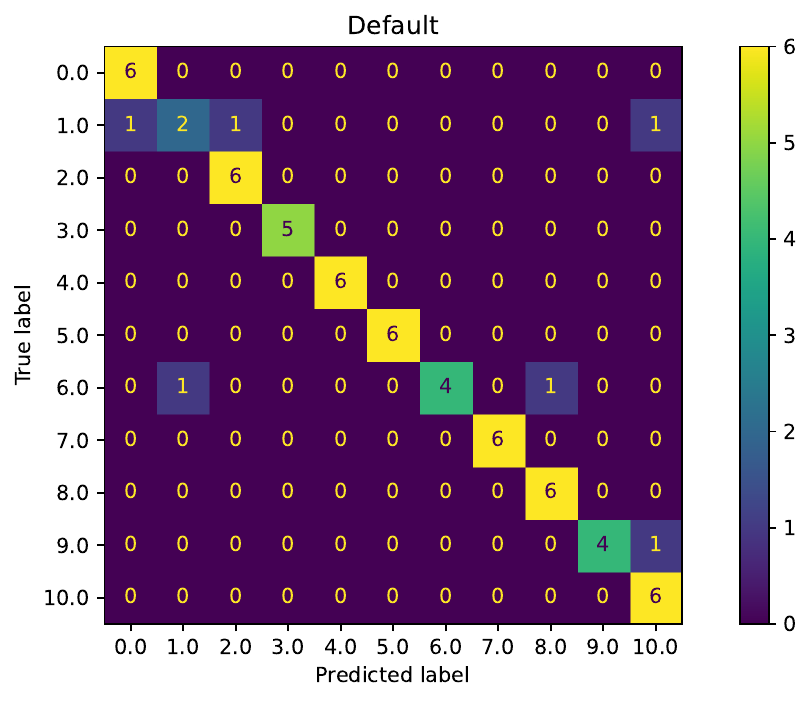}

}

\end{figure}

\begin{Shaded}
\begin{Highlighting}[]
\NormalTok{plot\_confusion\_matrix(model\_spot, fun\_control, title}\OperatorTok{=}\StringTok{"SPOT"}\NormalTok{)}
\end{Highlighting}
\end{Shaded}

\begin{figure}[H]

{\centering \includegraphics{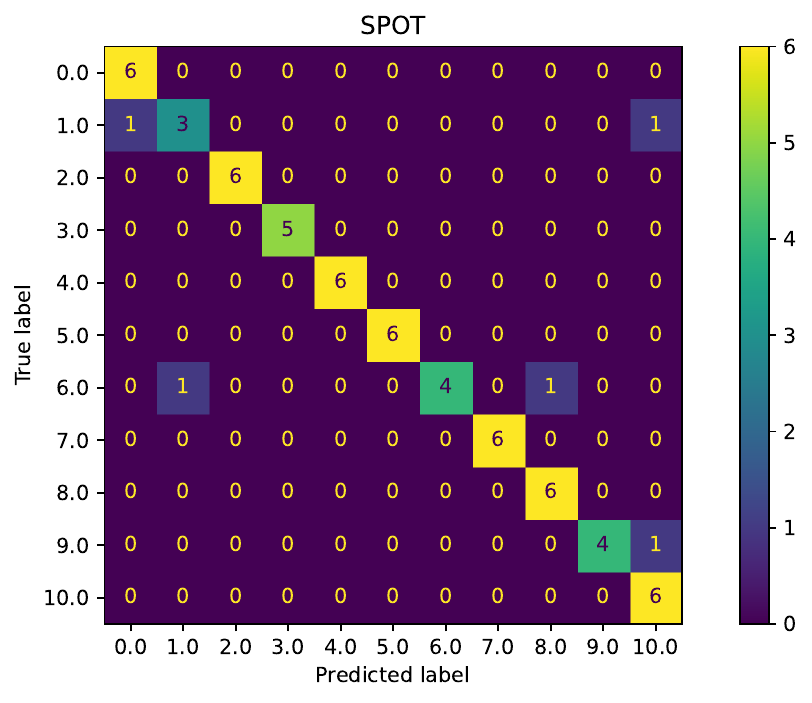}

}

\end{figure}

\begin{Shaded}
\begin{Highlighting}[]
\BuiltInTok{min}\NormalTok{(spot\_tuner.y), }\BuiltInTok{max}\NormalTok{(spot\_tuner.y)}
\end{Highlighting}
\end{Shaded}

\begin{verbatim}
(-0.879188712522046, -0.6155202821869489)
\end{verbatim}

\hypertarget{cross-validated-evaluations}{%
\subsection{Cross-validated
Evaluations}\label{cross-validated-evaluations}}

\begin{Shaded}
\begin{Highlighting}[]
\ImportTok{from}\NormalTok{ spotPython.sklearn.traintest }\ImportTok{import}\NormalTok{ evaluate\_cv}
\NormalTok{fun\_control.update(\{}
     \StringTok{"eval"}\NormalTok{: }\StringTok{"train\_cv"}\NormalTok{,}
     \StringTok{"k\_folds"}\NormalTok{: }\DecValTok{10}\NormalTok{,}
\NormalTok{\})}
\NormalTok{evaluate\_cv(model}\OperatorTok{=}\NormalTok{model\_spot, fun\_control}\OperatorTok{=}\NormalTok{fun\_control, verbose}\OperatorTok{=}\DecValTok{0}\NormalTok{)}
\end{Highlighting}
\end{Shaded}

\begin{verbatim}
(0.8746588693957115, None)
\end{verbatim}

\begin{Shaded}
\begin{Highlighting}[]
\NormalTok{fun\_control.update(\{}
     \StringTok{"eval"}\NormalTok{: }\StringTok{"test\_cv"}\NormalTok{,}
     \StringTok{"k\_folds"}\NormalTok{: }\DecValTok{10}\NormalTok{,}
\NormalTok{\})}
\NormalTok{evaluate\_cv(model}\OperatorTok{=}\NormalTok{model\_spot, fun\_control}\OperatorTok{=}\NormalTok{fun\_control, verbose}\OperatorTok{=}\DecValTok{0}\NormalTok{)}
\end{Highlighting}
\end{Shaded}

\begin{verbatim}
Error in fun_sklearn(). Call to evaluate_cv failed. err=ValueError('n_splits=10 cannot be greater than the number of members in each class.'), type(err)=<class 'ValueError'>
\end{verbatim}

\begin{verbatim}
(nan, None)
\end{verbatim}

\begin{itemize}
\tightlist
\item
  This is the evaluation that will be used in the comparison:
\end{itemize}

\begin{Shaded}
\begin{Highlighting}[]
\NormalTok{fun\_control.update(\{}
     \StringTok{"eval"}\NormalTok{: }\StringTok{"data\_cv"}\NormalTok{,}
     \StringTok{"k\_folds"}\NormalTok{: }\DecValTok{10}\NormalTok{,}
\NormalTok{\})}
\NormalTok{evaluate\_cv(model}\OperatorTok{=}\NormalTok{model\_spot, fun\_control}\OperatorTok{=}\NormalTok{fun\_control, verbose}\OperatorTok{=}\DecValTok{0}\NormalTok{)}
\end{Highlighting}
\end{Shaded}

\begin{verbatim}
(0.879974358974359, None)
\end{verbatim}

\hypertarget{detailed-hyperparameter-plots-3}{%
\subsection{Detailed Hyperparameter
Plots}\label{detailed-hyperparameter-plots-3}}

\begin{Shaded}
\begin{Highlighting}[]
\NormalTok{filename }\OperatorTok{=} \StringTok{"./figures/"} \OperatorTok{+}\NormalTok{ experiment\_name}
\NormalTok{spot\_tuner.plot\_important\_hyperparameter\_contour(filename}\OperatorTok{=}\NormalTok{filename)}
\end{Highlighting}
\end{Shaded}

\begin{verbatim}
n_estimators:  0.13008809627543882
max_depth:  0.2405297237337157
min_samples_split:  1.346605405278484
min_samples_leaf:  0.19308393515083044
max_features:  100.0
\end{verbatim}

\begin{figure}[H]

{\centering \includegraphics{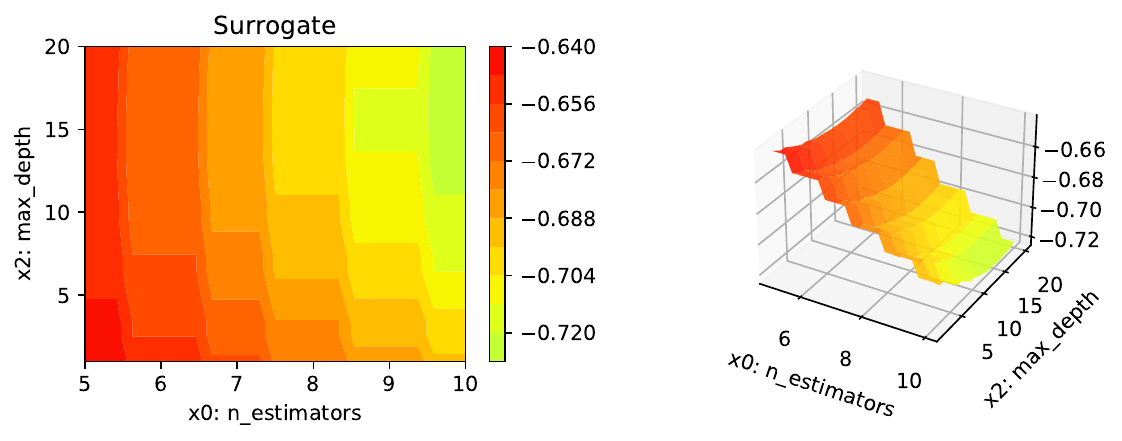}

}

\end{figure}

\begin{figure}[H]

{\centering \includegraphics{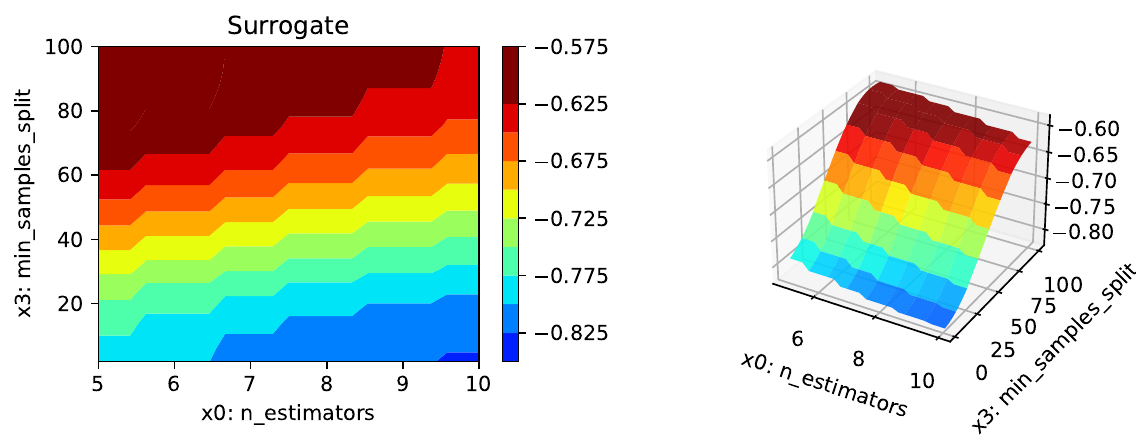}

}

\end{figure}

\begin{figure}[H]

{\centering \includegraphics{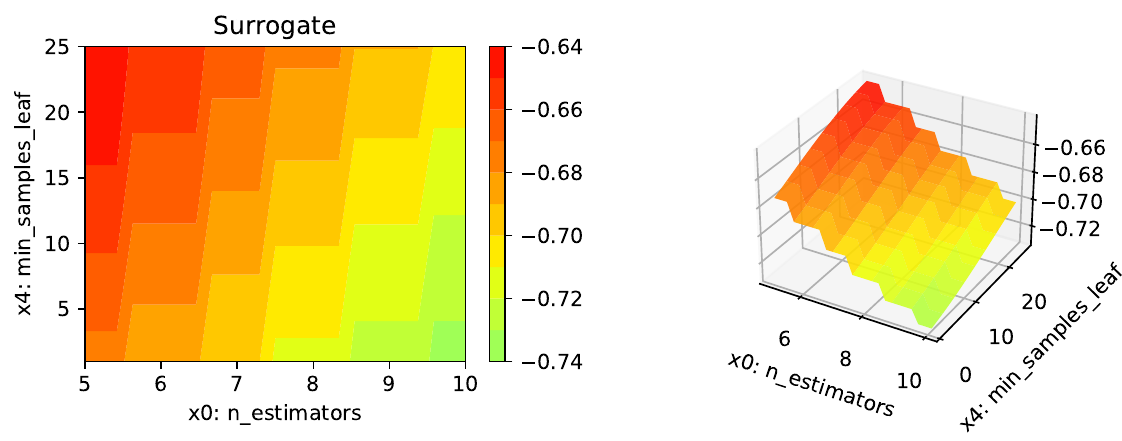}

}

\end{figure}

\begin{figure}[H]

{\centering \includegraphics{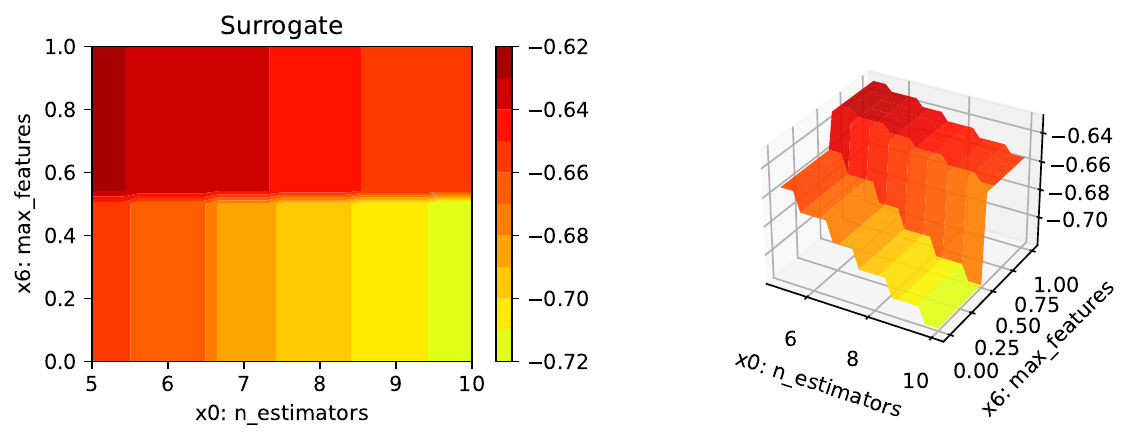}

}

\end{figure}

\begin{figure}[H]

{\centering \includegraphics{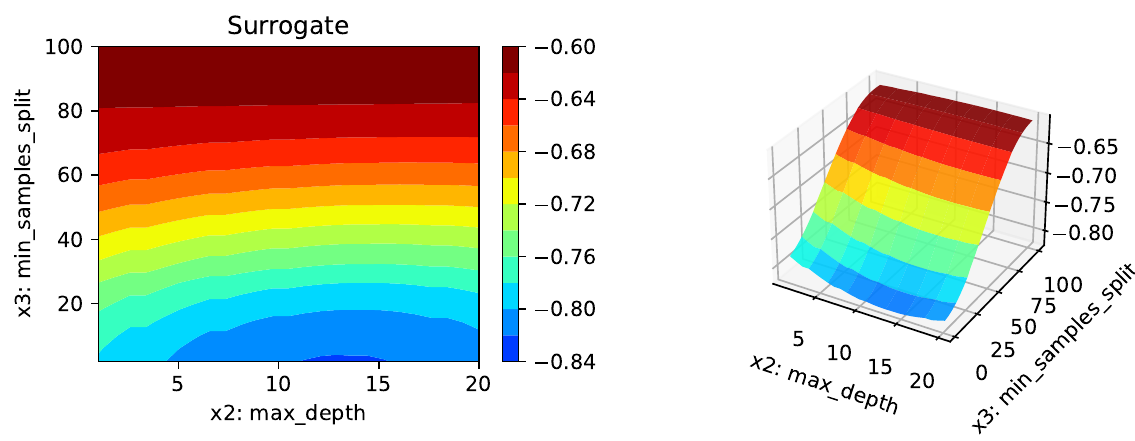}

}

\end{figure}

\begin{figure}[H]

{\centering \includegraphics{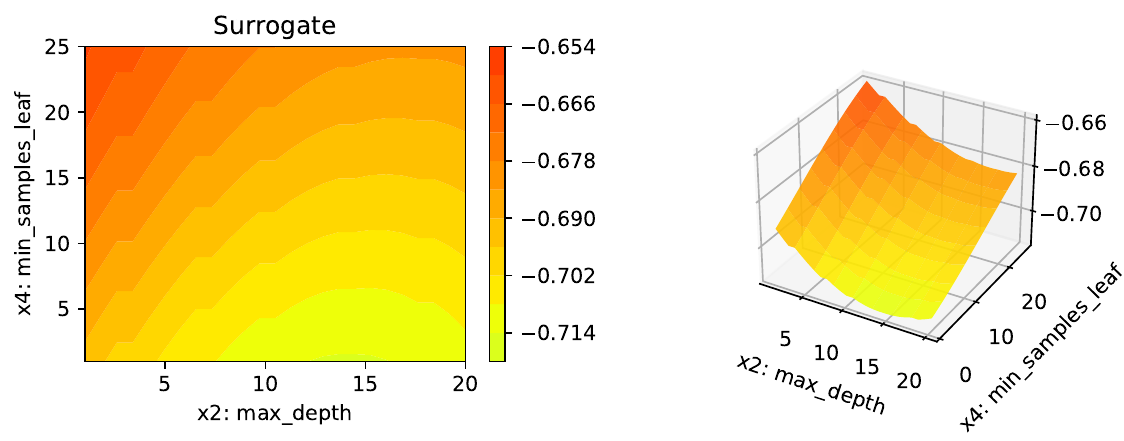}

}

\end{figure}

\begin{figure}[H]

{\centering \includegraphics{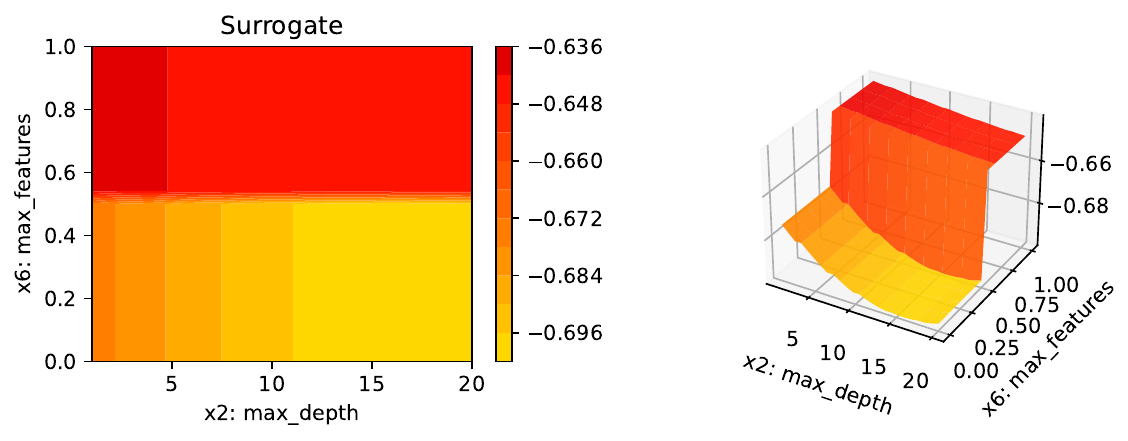}

}

\end{figure}

\begin{figure}[H]

{\centering \includegraphics{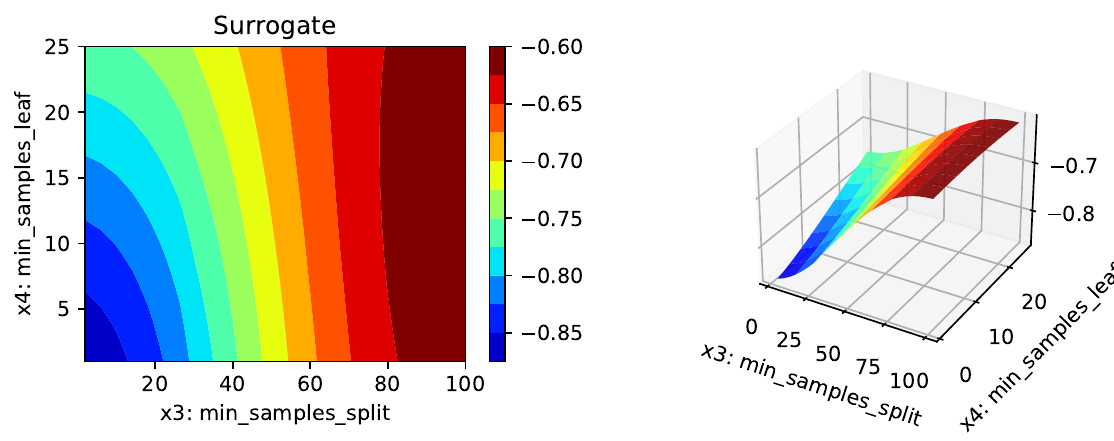}

}

\end{figure}

\begin{figure}[H]

{\centering \includegraphics{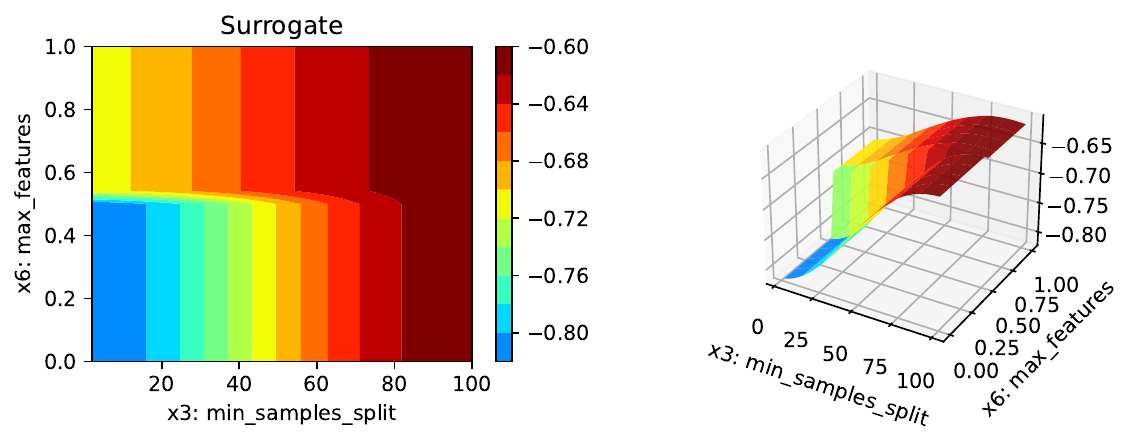}

}

\end{figure}

\begin{figure}[H]

{\centering \includegraphics{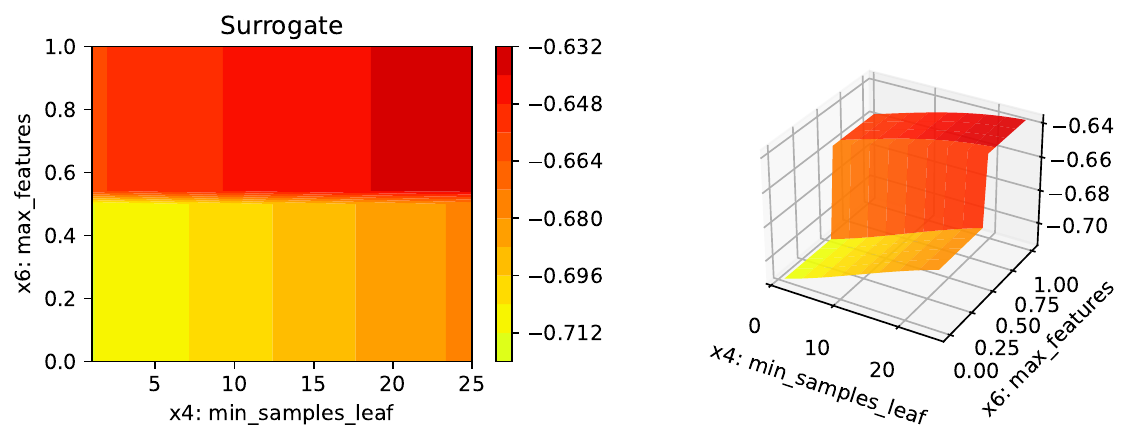}

}

\end{figure}

\hypertarget{parallel-coordinates-plot-1}{%
\subsection{Parallel Coordinates
Plot}\label{parallel-coordinates-plot-1}}

\begin{Shaded}
\begin{Highlighting}[]
\NormalTok{spot\_tuner.parallel\_plot()}
\end{Highlighting}
\end{Shaded}

\begin{verbatim}
Unable to display output for mime type(s): text/html
\end{verbatim}

\begin{verbatim}
Unable to display output for mime type(s): text/html
\end{verbatim}

\hypertarget{plot-all-combinations-of-hyperparameters-2}{%
\subsection{Plot all Combinations of
Hyperparameters}\label{plot-all-combinations-of-hyperparameters-2}}

\begin{itemize}
\tightlist
\item
  Warning: this may take a while.
\end{itemize}

\begin{Shaded}
\begin{Highlighting}[]
\NormalTok{PLOT\_ALL }\OperatorTok{=} \VariableTok{False}
\ControlFlowTok{if}\NormalTok{ PLOT\_ALL:}
\NormalTok{    n }\OperatorTok{=}\NormalTok{ spot\_tuner.k}
    \ControlFlowTok{for}\NormalTok{ i }\KeywordTok{in} \BuiltInTok{range}\NormalTok{(n}\OperatorTok{{-}}\DecValTok{1}\NormalTok{):}
        \ControlFlowTok{for}\NormalTok{ j }\KeywordTok{in} \BuiltInTok{range}\NormalTok{(i}\OperatorTok{+}\DecValTok{1}\NormalTok{, n):}
\NormalTok{            spot\_tuner.plot\_contour(i}\OperatorTok{=}\NormalTok{i, j}\OperatorTok{=}\NormalTok{j, min\_z}\OperatorTok{=}\NormalTok{min\_z, max\_z }\OperatorTok{=}\NormalTok{ max\_z)}
\end{Highlighting}
\end{Shaded}

\hypertarget{sec-hpt-sklearn-xgb-classifier-vbdp-data}{%
\chapter{HPT: sklearn XGB Classifier VBDP
Data}\label{sec-hpt-sklearn-xgb-classifier-vbdp-data}}

This chapter describes the hyperparameter tuning of a
\texttt{HistGradientBoostingClassifier} on the Vector Borne Disease
Prediction (VBDP) data set.

\begin{tcolorbox}[enhanced jigsaw, left=2mm, title=\textcolor{quarto-callout-important-color}{\faExclamation}\hspace{0.5em}{Vector Borne Disease Prediction Data Set}, bottomrule=.15mm, titlerule=0mm, breakable, rightrule=.15mm, toprule=.15mm, coltitle=black, colbacktitle=quarto-callout-important-color!10!white, leftrule=.75mm, arc=.35mm, colframe=quarto-callout-important-color-frame, bottomtitle=1mm, colback=white, opacitybacktitle=0.6, toptitle=1mm, opacityback=0]

This chapter uses the Vector Borne Disease Prediction data set from
Kaggle. It is a categorical dataset for eleven Vector Borne Diseases
with associated symptoms.

\begin{quote}
The person who associated a work with this deed has dedicated the work
to the public domain by waiving all of his or her rights to the work
worldwide under copyright law, including all related and neighboring
rights, to the extent allowed by law.You can copy, modify, distribute
and perform the work, even for commercial purposes, all without asking
permission. See Other Information below, see
\url{https://creativecommons.org/publicdomain/zero/1.0/}.
\end{quote}

The data set is available at:
\url{https://www.kaggle.com/datasets/richardbernat/vector-borne-disease-prediction},

The data should be downloaded and stored in the \texttt{data/VBDP}
subfolder. The data set is not available as a part of the
\texttt{spotPython} package.

\end{tcolorbox}

\hypertarget{sec-setup-17}{%
\section{Step 1: Setup}\label{sec-setup-17}}

Before we consider the detailed experimental setup, we select the
parameters that affect run time and the initial design size.

\begin{Shaded}
\begin{Highlighting}[]
\NormalTok{MAX\_TIME }\OperatorTok{=} \DecValTok{1}
\NormalTok{INIT\_SIZE }\OperatorTok{=} \DecValTok{5}
\NormalTok{ORIGINAL }\OperatorTok{=} \VariableTok{True}
\NormalTok{PREFIX }\OperatorTok{=} \StringTok{"17"}
\end{Highlighting}
\end{Shaded}

\begin{Shaded}
\begin{Highlighting}[]
\ImportTok{import}\NormalTok{ warnings}
\NormalTok{warnings.filterwarnings(}\StringTok{"ignore"}\NormalTok{)}
\end{Highlighting}
\end{Shaded}

\hypertarget{step-2-initialization-of-the-empty-fun_control-dictionary-2}{%
\section{\texorpdfstring{Step 2: Initialization of the Empty
\texttt{fun\_control}
Dictionary}{Step 2: Initialization of the Empty fun\_control Dictionary}}\label{step-2-initialization-of-the-empty-fun_control-dictionary-2}}

\begin{Shaded}
\begin{Highlighting}[]
\ImportTok{from}\NormalTok{ spotPython.utils.init }\ImportTok{import}\NormalTok{ fun\_control\_init}
\ImportTok{from}\NormalTok{ spotPython.utils.}\BuiltInTok{file} \ImportTok{import}\NormalTok{ get\_experiment\_name, get\_spot\_tensorboard\_path}
\ImportTok{from}\NormalTok{ spotPython.utils.device }\ImportTok{import}\NormalTok{ getDevice}

\NormalTok{experiment\_name }\OperatorTok{=}\NormalTok{ get\_experiment\_name(prefix}\OperatorTok{=}\NormalTok{PREFIX)}

\NormalTok{fun\_control }\OperatorTok{=}\NormalTok{ fun\_control\_init(}
\NormalTok{    task}\OperatorTok{=}\StringTok{"classification"}\NormalTok{,}
\NormalTok{    spot\_tensorboard\_path}\OperatorTok{=}\NormalTok{get\_spot\_tensorboard\_path(experiment\_name))}
\end{Highlighting}
\end{Shaded}

\hypertarget{sec-data-loading-17}{%
\section{Step 3: PyTorch Data Loading}\label{sec-data-loading-17}}

\hypertarget{load-data-classification-vbdp-1}{%
\subsection{1. Load Data: Classification
VBDP}\label{load-data-classification-vbdp-1}}

\begin{Shaded}
\begin{Highlighting}[]
\ImportTok{import}\NormalTok{ pandas }\ImportTok{as}\NormalTok{ pd}
\ControlFlowTok{if}\NormalTok{ ORIGINAL }\OperatorTok{==} \VariableTok{True}\NormalTok{:}
\NormalTok{    train\_df }\OperatorTok{=}\NormalTok{ pd.read\_csv(}\StringTok{\textquotesingle{}./data/VBDP/trainn.csv\textquotesingle{}}\NormalTok{)}
\NormalTok{    test\_df }\OperatorTok{=}\NormalTok{ pd.read\_csv(}\StringTok{\textquotesingle{}./data/VBDP/testt.csv\textquotesingle{}}\NormalTok{)}
\ControlFlowTok{else}\NormalTok{:}
\NormalTok{    train\_df }\OperatorTok{=}\NormalTok{ pd.read\_csv(}\StringTok{\textquotesingle{}./data/VBDP/train.csv\textquotesingle{}}\NormalTok{)}
    \CommentTok{\# remove the id column}
\NormalTok{    train\_df }\OperatorTok{=}\NormalTok{ train\_df.drop(columns}\OperatorTok{=}\NormalTok{[}\StringTok{\textquotesingle{}id\textquotesingle{}}\NormalTok{])}
\end{Highlighting}
\end{Shaded}

\begin{Shaded}
\begin{Highlighting}[]
\ImportTok{from}\NormalTok{ sklearn.preprocessing }\ImportTok{import}\NormalTok{ OrdinalEncoder}
\NormalTok{n\_samples }\OperatorTok{=}\NormalTok{ train\_df.shape[}\DecValTok{0}\NormalTok{]}
\NormalTok{n\_features }\OperatorTok{=}\NormalTok{ train\_df.shape[}\DecValTok{1}\NormalTok{] }\OperatorTok{{-}} \DecValTok{1}
\NormalTok{target\_column }\OperatorTok{=} \StringTok{"prognosis"}
\CommentTok{\# Encoder our prognosis labels as integers for easier decoding later}
\NormalTok{enc }\OperatorTok{=}\NormalTok{ OrdinalEncoder()}
\NormalTok{train\_df[target\_column] }\OperatorTok{=}\NormalTok{ enc.fit\_transform(train\_df[[target\_column]])}
\NormalTok{train\_df.columns }\OperatorTok{=}\NormalTok{ [}\SpecialStringTok{f"x}\SpecialCharTok{\{}\NormalTok{i}\SpecialCharTok{\}}\SpecialStringTok{"} \ControlFlowTok{for}\NormalTok{ i }\KeywordTok{in} \BuiltInTok{range}\NormalTok{(}\DecValTok{1}\NormalTok{, n\_features}\OperatorTok{+}\DecValTok{1}\NormalTok{)] }\OperatorTok{+}\NormalTok{ [target\_column]}
\BuiltInTok{print}\NormalTok{(train\_df.shape)}
\NormalTok{train\_df.head()}
\end{Highlighting}
\end{Shaded}

\begin{verbatim}
(252, 65)
\end{verbatim}

\begin{longtable}[]{@{}llllllllllllllllllllll@{}}
\toprule\noalign{}
& x1 & x2 & x3 & x4 & x5 & x6 & x7 & x8 & x9 & x10 & ... & x56 & x57 &
x58 & x59 & x60 & x61 & x62 & x63 & x64 & prognosis \\
\midrule\noalign{}
\endhead
\bottomrule\noalign{}
\endlastfoot
0 & 0 & 1 & 1 & 1 & 1 & 0 & 1 & 0 & 0 & 0 & ... & 0 & 0 & 0 & 0 & 0 & 0
& 0 & 0 & 0 & 0.0 \\
1 & 1 & 1 & 1 & 1 & 1 & 0 & 1 & 1 & 1 & 0 & ... & 0 & 0 & 0 & 0 & 0 & 0
& 0 & 0 & 0 & 0.0 \\
2 & 0 & 1 & 0 & 1 & 0 & 0 & 1 & 1 & 0 & 0 & ... & 0 & 0 & 0 & 0 & 0 & 0
& 0 & 0 & 0 & 0.0 \\
3 & 0 & 0 & 0 & 0 & 0 & 1 & 1 & 1 & 0 & 0 & ... & 0 & 0 & 0 & 0 & 0 & 0
& 0 & 0 & 0 & 0.0 \\
4 & 1 & 0 & 0 & 0 & 1 & 1 & 1 & 1 & 0 & 0 & ... & 0 & 0 & 0 & 0 & 0 & 0
& 0 & 0 & 0 & 0.0 \\
\end{longtable}

The full data set \texttt{train\_df} 64 features. The target column is
labeled as \texttt{prognosis}.

\hypertarget{holdout-train-and-test-data-1}{%
\subsection{Holdout Train and Test
Data}\label{holdout-train-and-test-data-1}}

We split out a hold-out test set (25\% of the data) so we can calculate
an example MAP@K

\begin{Shaded}
\begin{Highlighting}[]
\ImportTok{import}\NormalTok{ numpy }\ImportTok{as}\NormalTok{ np}
\ImportTok{from}\NormalTok{ sklearn.model\_selection }\ImportTok{import}\NormalTok{ train\_test\_split}
\NormalTok{X\_train, X\_test, y\_train, y\_test }\OperatorTok{=}\NormalTok{ train\_test\_split(train\_df.drop(target\_column, axis}\OperatorTok{=}\DecValTok{1}\NormalTok{), train\_df[target\_column],}
\NormalTok{                                                    random\_state}\OperatorTok{=}\DecValTok{42}\NormalTok{,}
\NormalTok{                                                    test\_size}\OperatorTok{=}\FloatTok{0.25}\NormalTok{,}
\NormalTok{                                                    stratify}\OperatorTok{=}\NormalTok{train\_df[target\_column])}
\NormalTok{train }\OperatorTok{=}\NormalTok{ pd.DataFrame(np.hstack((X\_train, np.array(y\_train).reshape(}\OperatorTok{{-}}\DecValTok{1}\NormalTok{, }\DecValTok{1}\NormalTok{))))}
\NormalTok{test }\OperatorTok{=}\NormalTok{ pd.DataFrame(np.hstack((X\_test, np.array(y\_test).reshape(}\OperatorTok{{-}}\DecValTok{1}\NormalTok{, }\DecValTok{1}\NormalTok{))))}
\NormalTok{train.columns }\OperatorTok{=}\NormalTok{ [}\SpecialStringTok{f"x}\SpecialCharTok{\{}\NormalTok{i}\SpecialCharTok{\}}\SpecialStringTok{"} \ControlFlowTok{for}\NormalTok{ i }\KeywordTok{in} \BuiltInTok{range}\NormalTok{(}\DecValTok{1}\NormalTok{, n\_features}\OperatorTok{+}\DecValTok{1}\NormalTok{)] }\OperatorTok{+}\NormalTok{ [target\_column]}
\NormalTok{test.columns }\OperatorTok{=}\NormalTok{ [}\SpecialStringTok{f"x}\SpecialCharTok{\{}\NormalTok{i}\SpecialCharTok{\}}\SpecialStringTok{"} \ControlFlowTok{for}\NormalTok{ i }\KeywordTok{in} \BuiltInTok{range}\NormalTok{(}\DecValTok{1}\NormalTok{, n\_features}\OperatorTok{+}\DecValTok{1}\NormalTok{)] }\OperatorTok{+}\NormalTok{ [target\_column]}
\BuiltInTok{print}\NormalTok{(train.shape)}
\BuiltInTok{print}\NormalTok{(test.shape)}
\NormalTok{train.head()}
\end{Highlighting}
\end{Shaded}

\begin{verbatim}
(189, 65)
(63, 65)
\end{verbatim}

\begin{longtable}[]{@{}llllllllllllllllllllll@{}}
\toprule\noalign{}
& x1 & x2 & x3 & x4 & x5 & x6 & x7 & x8 & x9 & x10 & ... & x56 & x57 &
x58 & x59 & x60 & x61 & x62 & x63 & x64 & prognosis \\
\midrule\noalign{}
\endhead
\bottomrule\noalign{}
\endlastfoot
0 & 1.0 & 0.0 & 0.0 & 1.0 & 0.0 & 1.0 & 0.0 & 0.0 & 0.0 & 1.0 & ... &
0.0 & 0.0 & 0.0 & 0.0 & 1.0 & 1.0 & 1.0 & 0.0 & 0.0 & 7.0 \\
1 & 1.0 & 0.0 & 1.0 & 1.0 & 1.0 & 1.0 & 1.0 & 0.0 & 1.0 & 1.0 & ... &
0.0 & 1.0 & 1.0 & 1.0 & 1.0 & 0.0 & 1.0 & 1.0 & 1.0 & 3.0 \\
2 & 0.0 & 0.0 & 1.0 & 0.0 & 1.0 & 0.0 & 0.0 & 0.0 & 0.0 & 0.0 & ... &
0.0 & 0.0 & 0.0 & 0.0 & 0.0 & 0.0 & 0.0 & 0.0 & 0.0 & 10.0 \\
3 & 1.0 & 1.0 & 1.0 & 1.0 & 1.0 & 1.0 & 0.0 & 0.0 & 1.0 & 1.0 & ... &
1.0 & 0.0 & 1.0 & 1.0 & 1.0 & 0.0 & 0.0 & 1.0 & 1.0 & 3.0 \\
4 & 1.0 & 1.0 & 1.0 & 0.0 & 1.0 & 1.0 & 0.0 & 1.0 & 1.0 & 0.0 & ... &
0.0 & 0.0 & 0.0 & 0.0 & 0.0 & 0.0 & 0.0 & 0.0 & 0.0 & 8.0 \\
\end{longtable}

\begin{Shaded}
\begin{Highlighting}[]
\CommentTok{\# add the dataset to the fun\_control}
\NormalTok{fun\_control.update(\{}\StringTok{"data"}\NormalTok{: train\_df, }\CommentTok{\# full dataset,}
               \StringTok{"train"}\NormalTok{: train,}
               \StringTok{"test"}\NormalTok{: test,}
               \StringTok{"n\_samples"}\NormalTok{: n\_samples,}
               \StringTok{"target\_column"}\NormalTok{: target\_column\})}
\end{Highlighting}
\end{Shaded}

\hypertarget{sec-specification-of-preprocessing-model-17}{%
\section{Step 4: Specification of the Preprocessing
Model}\label{sec-specification-of-preprocessing-model-17}}

Data preprocesssing can be very simple, e.g., you can ignore it. Then
you would choose the \texttt{prep\_model} ``None'':

\begin{Shaded}
\begin{Highlighting}[]
\NormalTok{prep\_model }\OperatorTok{=} \VariableTok{None}
\NormalTok{fun\_control.update(\{}\StringTok{"prep\_model"}\NormalTok{: prep\_model\})}
\end{Highlighting}
\end{Shaded}

A default approach for numerical data is the \texttt{StandardScaler}
(mean 0, variance 1). This can be selected as follows:

\begin{Shaded}
\begin{Highlighting}[]
\CommentTok{\# prep\_model = StandardScaler()}
\CommentTok{\# fun\_control.update(\{"prep\_model": prep\_model\})}
\end{Highlighting}
\end{Shaded}

Even more complicated pre-processing steps are possible, e.g., the
follwing pipeline:

\begin{Shaded}
\begin{Highlighting}[]
\CommentTok{\# categorical\_columns = []}
\CommentTok{\# one\_hot\_encoder = OneHotEncoder(handle\_unknown="ignore", sparse\_output=False)}
\CommentTok{\# prep\_model = ColumnTransformer(}
\CommentTok{\#         transformers=[}
\CommentTok{\#             ("categorical", one\_hot\_encoder, categorical\_columns),}
\CommentTok{\#         ],}
\CommentTok{\#         remainder=StandardScaler(),}
\CommentTok{\#     )}
\end{Highlighting}
\end{Shaded}

\hypertarget{step-5-select-model-algorithm-and-core_model_hyper_dict-2}{%
\section{\texorpdfstring{Step 5: Select Model (\texttt{algorithm}) and
\texttt{core\_model\_hyper\_dict}}{Step 5: Select Model (algorithm) and core\_model\_hyper\_dict}}\label{step-5-select-model-algorithm-and-core_model_hyper_dict-2}}

The selection of the algorithm (ML model) that should be tuned is done
by specifying the its name from the \texttt{sklearn} implementation. For
example, the \texttt{SVC} support vector machine classifier is selected
as follows:

\texttt{add\_core\_model\_to\_fun\_control(SVC,\ fun\_control,\ SklearnHyperDict)}

Other core\_models are, e.g.,:

\begin{itemize}
\tightlist
\item
  RidgeCV
\item
  GradientBoostingRegressor
\item
  ElasticNet
\item
  RandomForestClassifier
\item
  LogisticRegression
\item
  KNeighborsClassifier
\item
  RandomForestClassifier
\item
  GradientBoostingClassifier
\item
  HistGradientBoostingClassifier
\end{itemize}

We will use the \texttt{RandomForestClassifier} classifier in this
example.

\begin{Shaded}
\begin{Highlighting}[]
\ImportTok{from}\NormalTok{ sklearn.linear\_model }\ImportTok{import}\NormalTok{ RidgeCV}
\ImportTok{from}\NormalTok{ sklearn.ensemble }\ImportTok{import}\NormalTok{ RandomForestClassifier}
\ImportTok{from}\NormalTok{ sklearn.svm }\ImportTok{import}\NormalTok{ SVC}
\ImportTok{from}\NormalTok{ sklearn.linear\_model }\ImportTok{import}\NormalTok{ LogisticRegression}
\ImportTok{from}\NormalTok{ sklearn.neighbors }\ImportTok{import}\NormalTok{ KNeighborsClassifier}
\ImportTok{from}\NormalTok{ sklearn.ensemble }\ImportTok{import}\NormalTok{ GradientBoostingClassifier}
\ImportTok{from}\NormalTok{ sklearn.ensemble }\ImportTok{import}\NormalTok{ GradientBoostingRegressor}
\ImportTok{from}\NormalTok{ sklearn.ensemble }\ImportTok{import}\NormalTok{ HistGradientBoostingClassifier}
\ImportTok{from}\NormalTok{ sklearn.linear\_model }\ImportTok{import}\NormalTok{ ElasticNet}
\ImportTok{from}\NormalTok{ spotPython.hyperparameters.values }\ImportTok{import}\NormalTok{ add\_core\_model\_to\_fun\_control}
\ImportTok{from}\NormalTok{ spotPython.data.sklearn\_hyper\_dict }\ImportTok{import}\NormalTok{ SklearnHyperDict}
\ImportTok{from}\NormalTok{ spotPython.fun.hypersklearn }\ImportTok{import}\NormalTok{ HyperSklearn}
\end{Highlighting}
\end{Shaded}

\begin{Shaded}
\begin{Highlighting}[]
\CommentTok{\# core\_model  = RidgeCV}
\CommentTok{\# core\_model = GradientBoostingRegressor}
\CommentTok{\# core\_model = ElasticNet}
\NormalTok{core\_model }\OperatorTok{=}\NormalTok{ RandomForestClassifier}
\CommentTok{\# core\_model = SVC}
\CommentTok{\# core\_model = LogisticRegression}
\CommentTok{\# core\_model = KNeighborsClassifier}
\CommentTok{\# core\_model = GradientBoostingClassifier}
\NormalTok{core\_model }\OperatorTok{=}\NormalTok{ HistGradientBoostingClassifier}
\NormalTok{add\_core\_model\_to\_fun\_control(core\_model}\OperatorTok{=}\NormalTok{core\_model,}
\NormalTok{                              fun\_control}\OperatorTok{=}\NormalTok{fun\_control,}
\NormalTok{                              hyper\_dict}\OperatorTok{=}\NormalTok{SklearnHyperDict,}
\NormalTok{                              filename}\OperatorTok{=}\VariableTok{None}\NormalTok{)}
\end{Highlighting}
\end{Shaded}

Now \texttt{fun\_control} has the information from the JSON file. The
available hyperparameters are:

\begin{Shaded}
\begin{Highlighting}[]
\BuiltInTok{print}\NormalTok{(}\OperatorTok{*}\NormalTok{fun\_control[}\StringTok{"core\_model\_hyper\_dict"}\NormalTok{].keys(), sep}\OperatorTok{=}\StringTok{"}\CharTok{\textbackslash{}n}\StringTok{"}\NormalTok{)}
\end{Highlighting}
\end{Shaded}

\begin{verbatim}
loss
learning_rate
max_iter
max_leaf_nodes
max_depth
min_samples_leaf
l2_regularization
max_bins
early_stopping
n_iter_no_change
tol
\end{verbatim}

\hypertarget{step-6-modify-hyper_dict-hyperparameters-for-the-selected-algorithm-aka-core_model-2}{%
\section{\texorpdfstring{Step 6: Modify \texttt{hyper\_dict}
Hyperparameters for the Selected Algorithm aka
\texttt{core\_model}}{Step 6: Modify hyper\_dict Hyperparameters for the Selected Algorithm aka core\_model}}\label{step-6-modify-hyper_dict-hyperparameters-for-the-selected-algorithm-aka-core_model-2}}

\hypertarget{modify-hyperparameter-of-type-numeric-and-integer-boolean-2}{%
\subsection{Modify hyperparameter of type numeric and integer
(boolean)}\label{modify-hyperparameter-of-type-numeric-and-integer-boolean-2}}

Numeric and boolean values can be modified using the
\texttt{modify\_hyper\_parameter\_bounds} method. For example, to change
the \texttt{tol} hyperparameter of the \texttt{SVC} model to the
interval {[}1e-3, 1e-2{]}, the following code can be used:

\texttt{modify\_hyper\_parameter\_bounds(fun\_control,\ "tol",\ bounds={[}1e-3,\ 1e-2{]})}

\begin{Shaded}
\begin{Highlighting}[]
\ImportTok{from}\NormalTok{ spotPython.hyperparameters.values }\ImportTok{import}\NormalTok{ modify\_hyper\_parameter\_bounds}
\CommentTok{\# modify\_hyper\_parameter\_bounds(fun\_control, "tol", bounds=[1e{-}3, 1e{-}2])}
\CommentTok{\# modify\_hyper\_parameter\_bounds(fun\_control, "min\_samples\_split", bounds=[3, 20])}
\CommentTok{\# modify\_hyper\_parameter\_bounds(fun\_control, "dual", bounds=[0, 0])}
\CommentTok{\# modify\_hyper\_parameter\_bounds(fun\_control, "probability", bounds=[1, 1])}
\CommentTok{\# fun\_control["core\_model\_hyper\_dict"]["tol"]}
\CommentTok{\# modify\_hyper\_parameter\_bounds(fun\_control, "min\_samples\_leaf", bounds=[1, 25])}
\CommentTok{\# modify\_hyper\_parameter\_bounds(fun\_control, "n\_estimators", bounds=[5, 10])}
\end{Highlighting}
\end{Shaded}

\hypertarget{modify-hyperparameter-of-type-factor-3}{%
\subsection{Modify hyperparameter of type
factor}\label{modify-hyperparameter-of-type-factor-3}}

\texttt{spotPython} provides functions for modifying the
hyperparameters, their bounds and factors as well as for activating and
de-activating hyperparameters without re-compilation of the Python
source code. These functions were described in
Section~\ref{sec-modification-of-hyperparameters-14}.

Factors can be modified with the
\texttt{modify\_hyper\_parameter\_levels} function. For example, to
exclude the \texttt{sigmoid} kernel from the tuning, the \texttt{kernel}
hyperparameter of the \texttt{SVC} model can be modified as follows:

\texttt{modify\_hyper\_parameter\_levels(fun\_control,\ "kernel",\ {[}"linear",\ "rbf"{]})}

The new setting can be controlled via:

\texttt{fun\_control{[}"core\_model\_hyper\_dict"{]}{[}"kernel"{]}}

\begin{Shaded}
\begin{Highlighting}[]
\ImportTok{from}\NormalTok{ spotPython.hyperparameters.values }\ImportTok{import}\NormalTok{ modify\_hyper\_parameter\_levels}
\CommentTok{\# XGBoost:}
\NormalTok{modify\_hyper\_parameter\_levels(fun\_control, }\StringTok{"loss"}\NormalTok{, [}\StringTok{"log\_loss"}\NormalTok{])}
\end{Highlighting}
\end{Shaded}

\hypertarget{sec-optimizers-17}{%
\subsection{Optimizers}\label{sec-optimizers-17}}

Optimizers are described in Section~\ref{sec-optimizers-14}.

\hypertarget{step-7-selection-of-the-objective-loss-function-3}{%
\section{Step 7: Selection of the Objective (Loss)
Function}\label{step-7-selection-of-the-objective-loss-function-3}}

\hypertarget{evaluation}{%
\subsection{Evaluation}\label{evaluation}}

The evaluation procedure requires the specification of two elements:

\begin{enumerate}
\def\labelenumi{\arabic{enumi}.}
\tightlist
\item
  the way how the data is split into a train and a test set and
\item
  the loss function (and a metric).
\end{enumerate}

\hypertarget{selection-of-the-objective-metric-and-loss-functions-1}{%
\subsection{Selection of the Objective: Metric and Loss
Functions}\label{selection-of-the-objective-metric-and-loss-functions-1}}

\begin{itemize}
\tightlist
\item
  Machine learning models are optimized with respect to a metric, for
  example, the \texttt{accuracy} function.
\item
  Deep learning, e.g., neural networks are optimized with respect to a
  loss function, for example, the \texttt{cross\_entropy} function and
  evaluated with respect to a metric, for example, the \texttt{accuracy}
  function.
\end{itemize}

\hypertarget{loss-function}{%
\subsection{Loss Function}\label{loss-function}}

The loss function, that is usually used in deep learning for optimizing
the weights of the net, is stored in the \texttt{fun\_control}
dictionary as \texttt{"loss\_function"}.

\hypertarget{metric-function-1}{%
\subsection{Metric Function}\label{metric-function-1}}

There are two different types of metrics in \texttt{spotPython}:

\begin{enumerate}
\def\labelenumi{\arabic{enumi}.}
\tightlist
\item
  \texttt{"metric\_river"} is used for the river based evaluation via
  \texttt{eval\_oml\_iter\_progressive}.
\item
  \texttt{"metric\_sklearn"} is used for the sklearn based evaluation.
\end{enumerate}

We will consider multi-class classification metrics, e.g.,
\texttt{mapk\_score} and \texttt{top\_k\_accuracy\_score}.

\begin{tcolorbox}[enhanced jigsaw, left=2mm, title=\textcolor{quarto-callout-note-color}{\faInfo}\hspace{0.5em}{Predict Probabilities}, bottomrule=.15mm, titlerule=0mm, breakable, rightrule=.15mm, toprule=.15mm, coltitle=black, colbacktitle=quarto-callout-note-color!10!white, leftrule=.75mm, arc=.35mm, colframe=quarto-callout-note-color-frame, bottomtitle=1mm, colback=white, opacitybacktitle=0.6, toptitle=1mm, opacityback=0]

In this multi-class classification example the machine learning
algorithm should return the probabilities of the specific classes
(\texttt{"predict\_proba"}) instead of the predicted values.

\end{tcolorbox}

We set \texttt{"predict\_proba"} to \texttt{True} in the
\texttt{fun\_control} dictionary.

\hypertarget{the-mapk-metric-1}{%
\subsubsection{The MAPK Metric}\label{the-mapk-metric-1}}

To select the MAPK metric, the following two entries can be added to the
\texttt{fun\_control} dictionary:

\texttt{"metric\_sklearn":\ mapk\_score"}

\texttt{"metric\_params":\ \{"k":\ 3\}}.

\hypertarget{other-metrics-1}{%
\subsubsection{Other Metrics}\label{other-metrics-1}}

Alternatively, other metrics for multi-class classification can be used,
e.g.,: * top\_k\_accuracy\_score or * roc\_auc\_score

The metric \texttt{roc\_auc\_score} requires the parameter
\texttt{"multi\_class"}, e.g.,

\texttt{"multi\_class":\ "ovr"}.

This is set in the \texttt{fun\_control} dictionary.

\begin{tcolorbox}[enhanced jigsaw, left=2mm, title=\textcolor{quarto-callout-note-color}{\faInfo}\hspace{0.5em}{Weights}, bottomrule=.15mm, titlerule=0mm, breakable, rightrule=.15mm, toprule=.15mm, coltitle=black, colbacktitle=quarto-callout-note-color!10!white, leftrule=.75mm, arc=.35mm, colframe=quarto-callout-note-color-frame, bottomtitle=1mm, colback=white, opacitybacktitle=0.6, toptitle=1mm, opacityback=0]

\texttt{spotPython} performs a minimization, therefore, metrics that
should be maximized have to be multiplied by -1. This is done by setting
\texttt{"weights"} to \texttt{-1}.

\end{tcolorbox}

\begin{itemize}
\tightlist
\item
  The complete setup for the metric in our example is:
\end{itemize}

\begin{Shaded}
\begin{Highlighting}[]
\ImportTok{from}\NormalTok{ spotPython.utils.metrics }\ImportTok{import}\NormalTok{ mapk\_score}
\NormalTok{fun\_control.update(\{}
               \StringTok{"weights"}\NormalTok{: }\OperatorTok{{-}}\DecValTok{1}\NormalTok{,}
               \StringTok{"metric\_sklearn"}\NormalTok{: mapk\_score,}
               \StringTok{"predict\_proba"}\NormalTok{: }\VariableTok{True}\NormalTok{,}
               \StringTok{"metric\_params"}\NormalTok{: \{}\StringTok{"k"}\NormalTok{: }\DecValTok{3}\NormalTok{\},}
\NormalTok{               \})}
\end{Highlighting}
\end{Shaded}

\hypertarget{evaluation-on-hold-out-data-1}{%
\subsection{Evaluation on Hold-out
Data}\label{evaluation-on-hold-out-data-1}}

\begin{itemize}
\tightlist
\item
  The default method for computing the performance is
  \texttt{"eval\_holdout"}.
\item
  Alternatively, cross-validation can be used for every machine learning
  model.
\item
  Specifically for RandomForests, the OOB-score can be used.
\end{itemize}

\begin{Shaded}
\begin{Highlighting}[]
\NormalTok{fun\_control.update(\{}
    \StringTok{"eval"}\NormalTok{: }\StringTok{"train\_hold\_out"}\NormalTok{,}
\NormalTok{\})}
\end{Highlighting}
\end{Shaded}

\hypertarget{cross-validation-2}{%
\subsubsection{Cross Validation}\label{cross-validation-2}}

Instead of using the OOB-score, the classical cross validation can be
used. The number of folds is set by the key \texttt{"k\_folds"}. For
example, to use 5-fold cross validation, the key \texttt{"k\_folds"} is
set to \texttt{5}. Uncomment the following line to use cross validation:

\begin{Shaded}
\begin{Highlighting}[]
\CommentTok{\# fun\_control.update(\{}
\CommentTok{\#      "eval": "train\_cv",}
\CommentTok{\#      "k\_folds": 10,}
\CommentTok{\# \})}
\end{Highlighting}
\end{Shaded}

\hypertarget{step-8-calling-the-spot-function-3}{%
\section{Step 8: Calling the SPOT
Function}\label{step-8-calling-the-spot-function-3}}

\hypertarget{sec-prepare-spot-call-17}{%
\subsection{Preparing the SPOT Call}\label{sec-prepare-spot-call-17}}

\begin{itemize}
\tightlist
\item
  Get types and variable names as well as lower and upper bounds for the
  hyperparameters.
\end{itemize}

\begin{Shaded}
\begin{Highlighting}[]
\CommentTok{\# extract the variable types, names, and bounds}
\ImportTok{from}\NormalTok{ spotPython.hyperparameters.values }\ImportTok{import}\NormalTok{ (get\_bound\_values,}
\NormalTok{    get\_var\_name,}
\NormalTok{    get\_var\_type,)}
\NormalTok{var\_type }\OperatorTok{=}\NormalTok{ get\_var\_type(fun\_control)}
\NormalTok{var\_name }\OperatorTok{=}\NormalTok{ get\_var\_name(fun\_control)}
\NormalTok{lower }\OperatorTok{=}\NormalTok{ get\_bound\_values(fun\_control, }\StringTok{"lower"}\NormalTok{)}
\NormalTok{upper }\OperatorTok{=}\NormalTok{ get\_bound\_values(fun\_control, }\StringTok{"upper"}\NormalTok{)}
\end{Highlighting}
\end{Shaded}

\begin{Shaded}
\begin{Highlighting}[]
\ImportTok{from}\NormalTok{ spotPython.utils.eda }\ImportTok{import}\NormalTok{ gen\_design\_table}
\BuiltInTok{print}\NormalTok{(gen\_design\_table(fun\_control))}
\end{Highlighting}
\end{Shaded}

\begin{verbatim}
| name              | type   | default   |   lower |   upper | transform             |
|-------------------|--------|-----------|---------|---------|-----------------------|
| loss              | factor | log_loss  |   0     |   0     | None                  |
| learning_rate     | float  | -1.0      |  -5     |   0     | transform_power_10    |
| max_iter          | int    | 7         |   3     |  10     | transform_power_2_int |
| max_leaf_nodes    | int    | 5         |   1     |  12     | transform_power_2_int |
| max_depth         | int    | 2         |   1     |  20     | transform_power_2_int |
| min_samples_leaf  | int    | 4         |   2     |  10     | transform_power_2_int |
| l2_regularization | float  | 0.0       |   0     |  10     | None                  |
| max_bins          | int    | 255       | 127     | 255     | None                  |
| early_stopping    | factor | 1         |   0     |   1     | None                  |
| n_iter_no_change  | int    | 10        |   5     |  20     | None                  |
| tol               | float  | 0.0001    |   1e-05 |   0.001 | None                  |
\end{verbatim}

\hypertarget{sec-the-objective-function-17}{%
\subsection{The Objective
Function}\label{sec-the-objective-function-17}}

The objective function is selected next. It implements an interface from
\texttt{sklearn}'s training, validation, and testing methods to
\texttt{spotPython}.

\begin{Shaded}
\begin{Highlighting}[]
\ImportTok{from}\NormalTok{ spotPython.fun.hypersklearn }\ImportTok{import}\NormalTok{ HyperSklearn}
\NormalTok{fun }\OperatorTok{=}\NormalTok{ HyperSklearn().fun\_sklearn}
\end{Highlighting}
\end{Shaded}

\hypertarget{run-the-spot-optimizer-3}{%
\subsection{\texorpdfstring{Run the \texttt{Spot}
Optimizer}{Run the Spot Optimizer}}\label{run-the-spot-optimizer-3}}

\begin{itemize}
\tightlist
\item
  Run SPOT for approx. x mins (\texttt{max\_time}).
\item
  Note: the run takes longer, because the evaluation time of initial
  design (here: \texttt{initi\_size}, 20 points) is not considered.
\end{itemize}

\begin{Shaded}
\begin{Highlighting}[]
\ImportTok{from}\NormalTok{ spotPython.hyperparameters.values }\ImportTok{import}\NormalTok{ get\_default\_hyperparameters\_as\_array}
\NormalTok{X\_start }\OperatorTok{=}\NormalTok{ get\_default\_hyperparameters\_as\_array(fun\_control)}
\NormalTok{X\_start}
\end{Highlighting}
\end{Shaded}

\begin{verbatim}
array([[ 0.00e+00, -1.00e+00,  7.00e+00,  5.00e+00,  2.00e+00,  4.00e+00,
         0.00e+00,  2.55e+02,  1.00e+00,  1.00e+01,  1.00e-04]])
\end{verbatim}

\begin{Shaded}
\begin{Highlighting}[]
\ImportTok{import}\NormalTok{ numpy }\ImportTok{as}\NormalTok{ np}
\ImportTok{from}\NormalTok{ spotPython.spot }\ImportTok{import}\NormalTok{ spot}
\ImportTok{from}\NormalTok{ math }\ImportTok{import}\NormalTok{ inf}
\NormalTok{spot\_tuner }\OperatorTok{=}\NormalTok{ spot.Spot(fun}\OperatorTok{=}\NormalTok{fun,}
\NormalTok{                   lower }\OperatorTok{=}\NormalTok{ lower,}
\NormalTok{                   upper }\OperatorTok{=}\NormalTok{ upper,}
\NormalTok{                   fun\_evals }\OperatorTok{=}\NormalTok{ inf,}
\NormalTok{                   fun\_repeats }\OperatorTok{=} \DecValTok{1}\NormalTok{,}
\NormalTok{                   max\_time }\OperatorTok{=}\NormalTok{ MAX\_TIME,}
\NormalTok{                   noise }\OperatorTok{=} \VariableTok{False}\NormalTok{,}
\NormalTok{                   tolerance\_x }\OperatorTok{=}\NormalTok{ np.sqrt(np.spacing(}\DecValTok{1}\NormalTok{)),}
\NormalTok{                   var\_type }\OperatorTok{=}\NormalTok{ var\_type,}
\NormalTok{                   var\_name }\OperatorTok{=}\NormalTok{ var\_name,}
\NormalTok{                   infill\_criterion }\OperatorTok{=} \StringTok{"y"}\NormalTok{,}
\NormalTok{                   n\_points }\OperatorTok{=} \DecValTok{1}\NormalTok{,}
\NormalTok{                   seed}\OperatorTok{=}\DecValTok{123}\NormalTok{,}
\NormalTok{                   log\_level }\OperatorTok{=} \DecValTok{50}\NormalTok{,}
\NormalTok{                   show\_models}\OperatorTok{=} \VariableTok{False}\NormalTok{,}
\NormalTok{                   show\_progress}\OperatorTok{=} \VariableTok{True}\NormalTok{,}
\NormalTok{                   fun\_control }\OperatorTok{=}\NormalTok{ fun\_control,}
\NormalTok{                   design\_control}\OperatorTok{=}\NormalTok{\{}\StringTok{"init\_size"}\NormalTok{: INIT\_SIZE,}
                                   \StringTok{"repeats"}\NormalTok{: }\DecValTok{1}\NormalTok{\},}
\NormalTok{                   surrogate\_control}\OperatorTok{=}\NormalTok{\{}\StringTok{"noise"}\NormalTok{: }\VariableTok{True}\NormalTok{,}
                                      \StringTok{"cod\_type"}\NormalTok{: }\StringTok{"norm"}\NormalTok{,}
                                      \StringTok{"min\_theta"}\NormalTok{: }\OperatorTok{{-}}\DecValTok{4}\NormalTok{,}
                                      \StringTok{"max\_theta"}\NormalTok{: }\DecValTok{3}\NormalTok{,}
                                      \StringTok{"n\_theta"}\NormalTok{: }\BuiltInTok{len}\NormalTok{(var\_name),}
                                      \StringTok{"model\_fun\_evals"}\NormalTok{: }\DecValTok{10\_000}\NormalTok{,}
                                      \StringTok{"log\_level"}\NormalTok{: }\DecValTok{50}
\NormalTok{                                      \})}
\NormalTok{spot\_tuner.run(X\_start}\OperatorTok{=}\NormalTok{X\_start)}
\end{Highlighting}
\end{Shaded}

\begin{verbatim}
spotPython tuning: -0.84375 [#---------] 5.21% 
\end{verbatim}

\begin{verbatim}
spotPython tuning: -0.84375 [#---------] 9.72% 
\end{verbatim}

\begin{verbatim}
spotPython tuning: -0.84375 [#---------] 12.01% 
\end{verbatim}

\begin{verbatim}
spotPython tuning: -0.84375 [##--------] 15.54% 
\end{verbatim}

\begin{verbatim}
spotPython tuning: -0.84375 [###-------] 26.95% 
\end{verbatim}

\begin{verbatim}
spotPython tuning: -0.84375 [###-------] 32.56% 
\end{verbatim}

\begin{verbatim}
spotPython tuning: -0.84375 [####------] 36.91% 
\end{verbatim}

\begin{verbatim}
spotPython tuning: -0.84375 [####------] 39.93% 
\end{verbatim}

\begin{verbatim}
spotPython tuning: -0.84375 [#####-----] 45.14% 
\end{verbatim}

\begin{verbatim}
spotPython tuning: -0.84375 [#####-----] 47.63% 
\end{verbatim}

\begin{verbatim}
spotPython tuning: -0.8680555555555557 [######----] 57.62% 
\end{verbatim}

\begin{verbatim}
spotPython tuning: -0.8680555555555557 [######----] 63.53% 
\end{verbatim}

\begin{verbatim}
spotPython tuning: -0.8680555555555557 [########--] 79.92% 
\end{verbatim}

\begin{verbatim}
spotPython tuning: -0.8680555555555557 [#########-] 91.80% 
\end{verbatim}

\begin{verbatim}
spotPython tuning: -0.8680555555555557 [##########] 100.00% Done...
\end{verbatim}

\begin{verbatim}
<spotPython.spot.spot.Spot at 0x2c01e7eb0>
\end{verbatim}

\hypertarget{sec-tensorboard-17}{%
\section{Step 9: Tensorboard}\label{sec-tensorboard-17}}

The textual output shown in the console (or code cell) can be visualized
with Tensorboard as described in Section~\ref{sec-tensorboard-14}, see
also the description in the documentation:
\href{https://sequential-parameter-optimization.github.io/spotPython/14_spot_ray_hpt_torch_cifar10.html\#sec-tensorboard-14}{Tensorboard.}

\hypertarget{sec-results-tuning-17}{%
\section{Step 10: Results}\label{sec-results-tuning-17}}

After the hyperparameter tuning run is finished, the progress of the
hyperparameter tuning can be visualized. The following code generates
the progress plot from \textbf{?@fig-progress}.

\begin{Shaded}
\begin{Highlighting}[]
\NormalTok{spot\_tuner.plot\_progress(log\_y}\OperatorTok{=}\VariableTok{False}\NormalTok{,}
\NormalTok{    filename}\OperatorTok{=}\StringTok{"./figures/"} \OperatorTok{+}\NormalTok{ experiment\_name}\OperatorTok{+}\StringTok{"\_progress.png"}\NormalTok{)}
\end{Highlighting}
\end{Shaded}

\begin{figure}[H]

{\centering \includegraphics{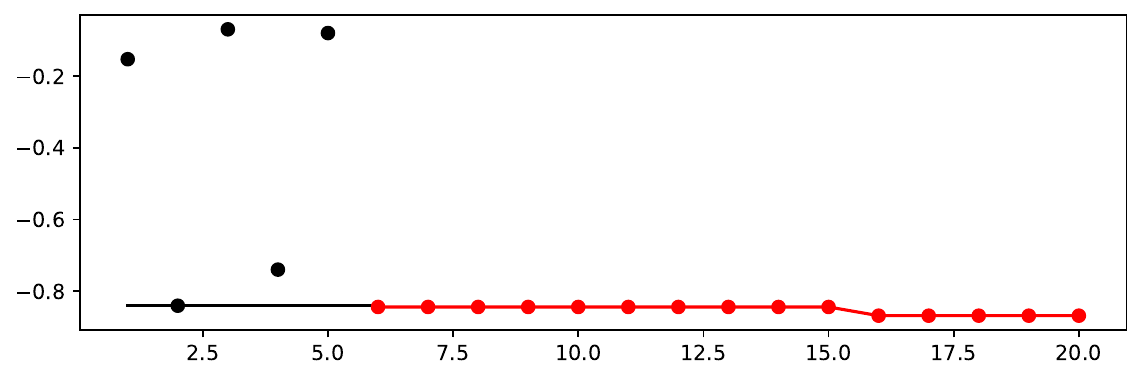}

}

\caption{Progress plot. \emph{Black} dots denote results from the
initial design. \emph{Red} dots illustrate the improvement found by the
surrogate model based optimization.}

\end{figure}

\begin{itemize}
\tightlist
\item
  Print the results
\end{itemize}

\begin{Shaded}
\begin{Highlighting}[]
\BuiltInTok{print}\NormalTok{(gen\_design\_table(fun\_control}\OperatorTok{=}\NormalTok{fun\_control,}
\NormalTok{    spot}\OperatorTok{=}\NormalTok{spot\_tuner))}
\end{Highlighting}
\end{Shaded}

\begin{verbatim}
| name              | type   | default   |   lower |   upper |               tuned | transform             |   importance | stars   |
|-------------------|--------|-----------|---------|---------|---------------------|-----------------------|--------------|---------|
| loss              | factor | log_loss  |     0.0 |     0.0 |                 0.0 | None                  |         0.00 |         |
| learning_rate     | float  | -1.0      |    -5.0 |     0.0 | -0.3668375393724054 | transform_power_10    |         0.36 | .       |
| max_iter          | int    | 7         |     3.0 |    10.0 |                 8.0 | transform_power_2_int |         0.13 | .       |
| max_leaf_nodes    | int    | 5         |     1.0 |    12.0 |                 6.0 | transform_power_2_int |         0.00 |         |
| max_depth         | int    | 2         |     1.0 |    20.0 |                17.0 | transform_power_2_int |         0.00 |         |
| min_samples_leaf  | int    | 4         |     2.0 |    10.0 |                 2.0 | transform_power_2_int |         1.61 | *       |
| l2_regularization | float  | 0.0       |     0.0 |    10.0 |                10.0 | None                  |         0.00 |         |
| max_bins          | int    | 255       |   127.0 |   255.0 |               140.0 | None                  |         0.00 |         |
| early_stopping    | factor | 1         |     0.0 |     1.0 |                 1.0 | None                  |       100.00 | ***     |
| n_iter_no_change  | int    | 10        |     5.0 |    20.0 |                 8.0 | None                  |         0.00 |         |
| tol               | float  | 0.0001    |   1e-05 |   0.001 |               0.001 | None                  |         0.00 |         |
\end{verbatim}

\hypertarget{show-variable-importance-2}{%
\subsection{Show variable importance}\label{show-variable-importance-2}}

\begin{Shaded}
\begin{Highlighting}[]
\NormalTok{spot\_tuner.plot\_importance(threshold}\OperatorTok{=}\FloatTok{0.025}\NormalTok{, filename}\OperatorTok{=}\StringTok{"./figures/"} \OperatorTok{+}\NormalTok{ experiment\_name}\OperatorTok{+}\StringTok{"\_importance.png"}\NormalTok{)}
\end{Highlighting}
\end{Shaded}

\begin{figure}[H]

{\centering \includegraphics{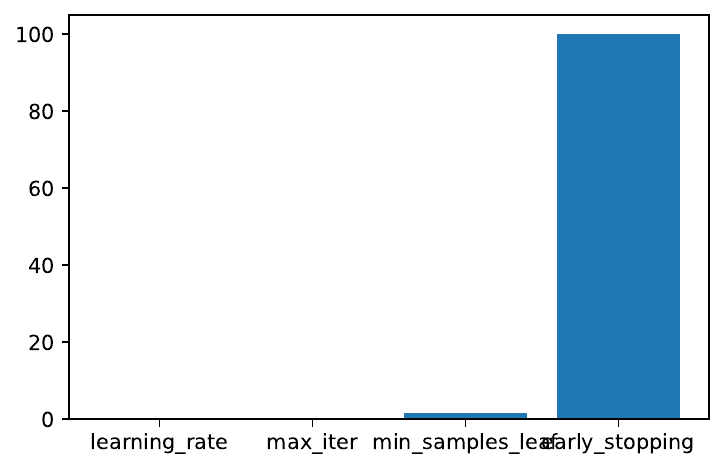}

}

\caption{Variable importance plot, threshold 0.025.}

\end{figure}

\hypertarget{get-default-hyperparameters-4}{%
\subsection{Get Default
Hyperparameters}\label{get-default-hyperparameters-4}}

\begin{Shaded}
\begin{Highlighting}[]
\ImportTok{from}\NormalTok{ spotPython.hyperparameters.values }\ImportTok{import}\NormalTok{ get\_default\_values, transform\_hyper\_parameter\_values}
\NormalTok{values\_default }\OperatorTok{=}\NormalTok{ get\_default\_values(fun\_control)}
\NormalTok{values\_default }\OperatorTok{=}\NormalTok{ transform\_hyper\_parameter\_values(fun\_control}\OperatorTok{=}\NormalTok{fun\_control, hyper\_parameter\_values}\OperatorTok{=}\NormalTok{values\_default)}
\NormalTok{values\_default}
\end{Highlighting}
\end{Shaded}

\begin{verbatim}
{'loss': 'log_loss',
 'learning_rate': 0.1,
 'max_iter': 128,
 'max_leaf_nodes': 32,
 'max_depth': 4,
 'min_samples_leaf': 16,
 'l2_regularization': 0.0,
 'max_bins': 255,
 'early_stopping': 1,
 'n_iter_no_change': 10,
 'tol': 0.0001}
\end{verbatim}

\begin{Shaded}
\begin{Highlighting}[]
\ImportTok{from}\NormalTok{ sklearn.pipeline }\ImportTok{import}\NormalTok{ make\_pipeline}
\NormalTok{model\_default }\OperatorTok{=}\NormalTok{ make\_pipeline(fun\_control[}\StringTok{"prep\_model"}\NormalTok{], fun\_control[}\StringTok{"core\_model"}\NormalTok{](}\OperatorTok{**}\NormalTok{values\_default))}
\NormalTok{model\_default}
\end{Highlighting}
\end{Shaded}

\begin{verbatim}
Pipeline(steps=[('nonetype', None),
                ('histgradientboostingclassifier',
                 HistGradientBoostingClassifier(early_stopping=1, max_depth=4,
                                                max_iter=128, max_leaf_nodes=32,
                                                min_samples_leaf=16,
                                                tol=0.0001))])
\end{verbatim}

\hypertarget{get-spot-results-3}{%
\subsection{Get SPOT Results}\label{get-spot-results-3}}

\begin{Shaded}
\begin{Highlighting}[]
\NormalTok{X }\OperatorTok{=}\NormalTok{ spot\_tuner.to\_all\_dim(spot\_tuner.min\_X.reshape(}\DecValTok{1}\NormalTok{,}\OperatorTok{{-}}\DecValTok{1}\NormalTok{))}
\BuiltInTok{print}\NormalTok{(X)}
\end{Highlighting}
\end{Shaded}

\begin{verbatim}
[[ 0.00000000e+00 -3.66837539e-01  8.00000000e+00  6.00000000e+00
   1.70000000e+01  2.00000000e+00  1.00000000e+01  1.40000000e+02
   1.00000000e+00  8.00000000e+00  1.00000000e-03]]
\end{verbatim}

\begin{Shaded}
\begin{Highlighting}[]
\ImportTok{from}\NormalTok{ spotPython.hyperparameters.values }\ImportTok{import}\NormalTok{ assign\_values, return\_conf\_list\_from\_var\_dict}
\NormalTok{v\_dict }\OperatorTok{=}\NormalTok{ assign\_values(X, fun\_control[}\StringTok{"var\_name"}\NormalTok{])}
\NormalTok{return\_conf\_list\_from\_var\_dict(var\_dict}\OperatorTok{=}\NormalTok{v\_dict, fun\_control}\OperatorTok{=}\NormalTok{fun\_control)}
\end{Highlighting}
\end{Shaded}

\begin{verbatim}
[{'loss': 'log_loss',
  'learning_rate': 0.429697137559405,
  'max_iter': 256,
  'max_leaf_nodes': 64,
  'max_depth': 131072,
  'min_samples_leaf': 4,
  'l2_regularization': 10.0,
  'max_bins': 140,
  'early_stopping': 1,
  'n_iter_no_change': 8,
  'tol': 0.001}]
\end{verbatim}

\begin{Shaded}
\begin{Highlighting}[]
\ImportTok{from}\NormalTok{ spotPython.hyperparameters.values }\ImportTok{import}\NormalTok{ get\_one\_sklearn\_model\_from\_X}
\NormalTok{model\_spot }\OperatorTok{=}\NormalTok{ get\_one\_sklearn\_model\_from\_X(X, fun\_control)}
\NormalTok{model\_spot}
\end{Highlighting}
\end{Shaded}

\begin{verbatim}
HistGradientBoostingClassifier(early_stopping=1, l2_regularization=10.0,
                               learning_rate=0.429697137559405, max_bins=140,
                               max_depth=131072, max_iter=256,
                               max_leaf_nodes=64, min_samples_leaf=4,
                               n_iter_no_change=8, tol=0.001)
\end{verbatim}

\hypertarget{evaluate-spot-results-1}{%
\subsection{Evaluate SPOT Results}\label{evaluate-spot-results-1}}

\begin{itemize}
\tightlist
\item
  Fetch the data.
\end{itemize}

\begin{Shaded}
\begin{Highlighting}[]
\ImportTok{from}\NormalTok{ spotPython.utils.convert }\ImportTok{import}\NormalTok{ get\_Xy\_from\_df}
\NormalTok{X\_train, y\_train }\OperatorTok{=}\NormalTok{ get\_Xy\_from\_df(fun\_control[}\StringTok{"train"}\NormalTok{], fun\_control[}\StringTok{"target\_column"}\NormalTok{])}
\NormalTok{X\_test, y\_test }\OperatorTok{=}\NormalTok{ get\_Xy\_from\_df(fun\_control[}\StringTok{"test"}\NormalTok{], fun\_control[}\StringTok{"target\_column"}\NormalTok{])}
\NormalTok{X\_test.shape, y\_test.shape}
\end{Highlighting}
\end{Shaded}

\begin{verbatim}
((63, 64), (63,))
\end{verbatim}

\begin{itemize}
\tightlist
\item
  Fit the model with the tuned hyperparameters. This gives one result:
\end{itemize}

\begin{Shaded}
\begin{Highlighting}[]
\NormalTok{model\_spot.fit(X\_train, y\_train)}
\NormalTok{y\_pred }\OperatorTok{=}\NormalTok{ model\_spot.predict\_proba(X\_test)}
\NormalTok{res }\OperatorTok{=}\NormalTok{ mapk\_score(y\_true}\OperatorTok{=}\NormalTok{y\_test, y\_pred}\OperatorTok{=}\NormalTok{y\_pred, k}\OperatorTok{=}\DecValTok{3}\NormalTok{)}
\NormalTok{res}
\end{Highlighting}
\end{Shaded}

\begin{verbatim}
0.7910052910052912
\end{verbatim}

\begin{Shaded}
\begin{Highlighting}[]
\KeywordTok{def}\NormalTok{ repeated\_eval(n, model):}
\NormalTok{    res\_values }\OperatorTok{=}\NormalTok{ []}
    \ControlFlowTok{for}\NormalTok{ i }\KeywordTok{in} \BuiltInTok{range}\NormalTok{(n):}
\NormalTok{        model.fit(X\_train, y\_train)}
\NormalTok{        y\_pred }\OperatorTok{=}\NormalTok{ model.predict\_proba(X\_test)}
\NormalTok{        res }\OperatorTok{=}\NormalTok{ mapk\_score(y\_true}\OperatorTok{=}\NormalTok{y\_test, y\_pred}\OperatorTok{=}\NormalTok{y\_pred, k}\OperatorTok{=}\DecValTok{3}\NormalTok{)}
\NormalTok{        res\_values.append(res)}
\NormalTok{    mean\_res }\OperatorTok{=}\NormalTok{ np.mean(res\_values)}
    \BuiltInTok{print}\NormalTok{(}\SpecialStringTok{f"mean\_res: }\SpecialCharTok{\{}\NormalTok{mean\_res}\SpecialCharTok{\}}\SpecialStringTok{"}\NormalTok{)}
\NormalTok{    std\_res }\OperatorTok{=}\NormalTok{ np.std(res\_values)}
    \BuiltInTok{print}\NormalTok{(}\SpecialStringTok{f"std\_res: }\SpecialCharTok{\{}\NormalTok{std\_res}\SpecialCharTok{\}}\SpecialStringTok{"}\NormalTok{)}
\NormalTok{    min\_res }\OperatorTok{=}\NormalTok{ np.}\BuiltInTok{min}\NormalTok{(res\_values)}
    \BuiltInTok{print}\NormalTok{(}\SpecialStringTok{f"min\_res: }\SpecialCharTok{\{}\NormalTok{min\_res}\SpecialCharTok{\}}\SpecialStringTok{"}\NormalTok{)}
\NormalTok{    max\_res }\OperatorTok{=}\NormalTok{ np.}\BuiltInTok{max}\NormalTok{(res\_values)}
    \BuiltInTok{print}\NormalTok{(}\SpecialStringTok{f"max\_res: }\SpecialCharTok{\{}\NormalTok{max\_res}\SpecialCharTok{\}}\SpecialStringTok{"}\NormalTok{)}
\NormalTok{    median\_res }\OperatorTok{=}\NormalTok{ np.median(res\_values)}
    \BuiltInTok{print}\NormalTok{(}\SpecialStringTok{f"median\_res: }\SpecialCharTok{\{}\NormalTok{median\_res}\SpecialCharTok{\}}\SpecialStringTok{"}\NormalTok{)}
    \ControlFlowTok{return}\NormalTok{ mean\_res, std\_res, min\_res, max\_res, median\_res}
\end{Highlighting}
\end{Shaded}

\hypertarget{handling-non-deterministic-results-1}{%
\subsection{Handling Non-deterministic
Results}\label{handling-non-deterministic-results-1}}

\begin{itemize}
\tightlist
\item
  Because the model is non-determinstic, we perform \(n=30\) runs and
  calculate the mean and standard deviation of the performance metric.
\end{itemize}

\begin{Shaded}
\begin{Highlighting}[]
\NormalTok{\_ }\OperatorTok{=}\NormalTok{ repeated\_eval(}\DecValTok{30}\NormalTok{, model\_spot)}
\end{Highlighting}
\end{Shaded}

\begin{verbatim}
mean_res: 0.7858906525573192
std_res: 0.01388041789767267
min_res: 0.7566137566137567
max_res: 0.8201058201058202
median_res: 0.7830687830687831
\end{verbatim}

\hypertarget{evalution-of-the-default-hyperparameters-1}{%
\subsection{Evalution of the Default
Hyperparameters}\label{evalution-of-the-default-hyperparameters-1}}

\begin{Shaded}
\begin{Highlighting}[]
\NormalTok{model\_default.fit(X\_train, y\_train)[}\StringTok{"histgradientboostingclassifier"}\NormalTok{]}
\end{Highlighting}
\end{Shaded}

\begin{verbatim}
HistGradientBoostingClassifier(early_stopping=1, max_depth=4, max_iter=128,
                               max_leaf_nodes=32, min_samples_leaf=16,
                               tol=0.0001)
\end{verbatim}

\begin{itemize}
\tightlist
\item
  One evaluation of the default hyperparameters is performed on the
  hold-out test set.
\end{itemize}

\begin{Shaded}
\begin{Highlighting}[]
\NormalTok{y\_pred }\OperatorTok{=}\NormalTok{ model\_default.predict\_proba(X\_test)}
\NormalTok{mapk\_score(y\_true}\OperatorTok{=}\NormalTok{y\_test, y\_pred}\OperatorTok{=}\NormalTok{y\_pred, k}\OperatorTok{=}\DecValTok{3}\NormalTok{)}
\end{Highlighting}
\end{Shaded}

\begin{verbatim}
0.7592592592592592
\end{verbatim}

Since one single evaluation is not meaningful, we perform, similar to
the evaluation of the SPOT results, \(n=30\) runs of the default setting
and and calculate the mean and standard deviation of the performance
metric.

\begin{Shaded}
\begin{Highlighting}[]
\NormalTok{\_ }\OperatorTok{=}\NormalTok{ repeated\_eval(}\DecValTok{30}\NormalTok{, model\_default)}
\end{Highlighting}
\end{Shaded}

\begin{verbatim}
mean_res: 0.7952380952380952
std_res: 0.013824280735284227
min_res: 0.7671957671957672
max_res: 0.8253968253968255
median_res: 0.7962962962962963
\end{verbatim}

\hypertarget{plot-compare-predictions-2}{%
\subsection{Plot: Compare
Predictions}\label{plot-compare-predictions-2}}

\begin{Shaded}
\begin{Highlighting}[]
\ImportTok{from}\NormalTok{ spotPython.plot.validation }\ImportTok{import}\NormalTok{ plot\_confusion\_matrix}
\NormalTok{plot\_confusion\_matrix(model\_default, fun\_control, title }\OperatorTok{=} \StringTok{"Default"}\NormalTok{)}
\end{Highlighting}
\end{Shaded}

\begin{figure}[H]

{\centering \includegraphics{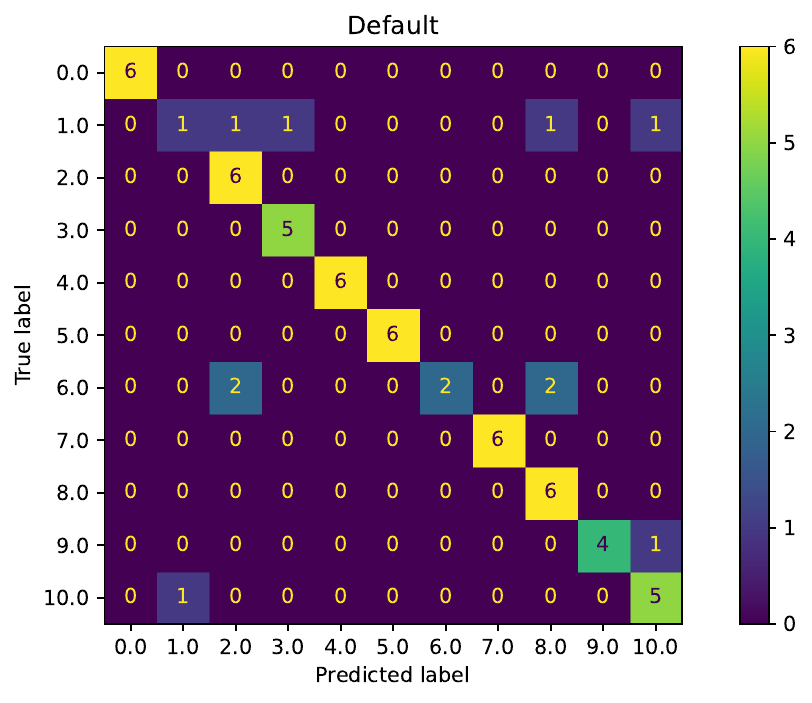}

}

\end{figure}

\begin{Shaded}
\begin{Highlighting}[]
\NormalTok{plot\_confusion\_matrix(model\_spot, fun\_control, title}\OperatorTok{=}\StringTok{"SPOT"}\NormalTok{)}
\end{Highlighting}
\end{Shaded}

\begin{figure}[H]

{\centering \includegraphics{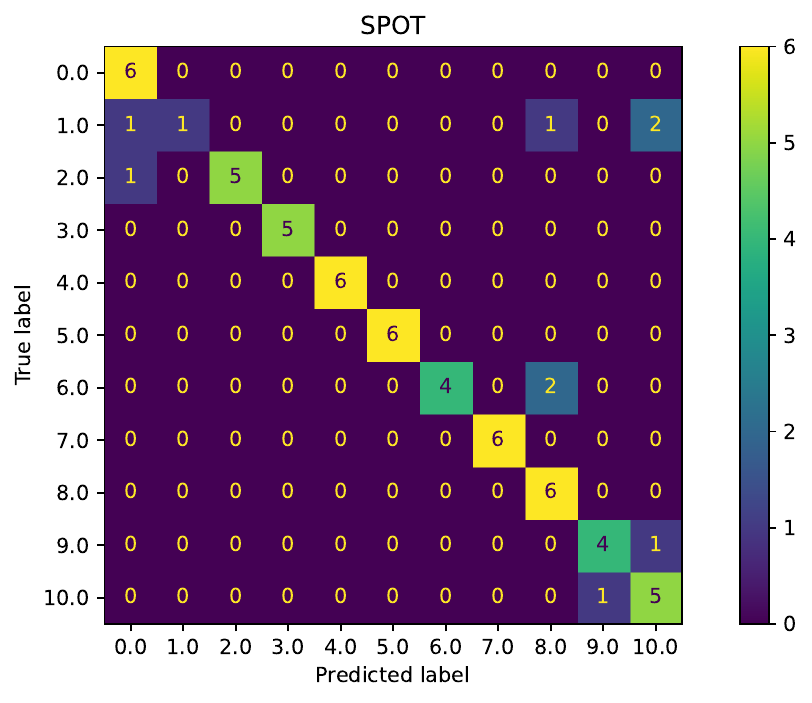}

}

\end{figure}

\begin{Shaded}
\begin{Highlighting}[]
\BuiltInTok{min}\NormalTok{(spot\_tuner.y), }\BuiltInTok{max}\NormalTok{(spot\_tuner.y)}
\end{Highlighting}
\end{Shaded}

\begin{verbatim}
(-0.8680555555555557, -0.06944444444444443)
\end{verbatim}

\hypertarget{cross-validated-evaluations-1}{%
\subsection{Cross-validated
Evaluations}\label{cross-validated-evaluations-1}}

\begin{Shaded}
\begin{Highlighting}[]
\ImportTok{from}\NormalTok{ spotPython.sklearn.traintest }\ImportTok{import}\NormalTok{ evaluate\_cv}
\NormalTok{fun\_control.update(\{}
     \StringTok{"eval"}\NormalTok{: }\StringTok{"train\_cv"}\NormalTok{,}
     \StringTok{"k\_folds"}\NormalTok{: }\DecValTok{10}\NormalTok{,}
\NormalTok{\})}
\NormalTok{evaluate\_cv(model}\OperatorTok{=}\NormalTok{model\_spot, fun\_control}\OperatorTok{=}\NormalTok{fun\_control, verbose}\OperatorTok{=}\DecValTok{0}\NormalTok{)}
\end{Highlighting}
\end{Shaded}

\begin{verbatim}
(0.8021442495126706, None)
\end{verbatim}

\begin{Shaded}
\begin{Highlighting}[]
\NormalTok{fun\_control.update(\{}
     \StringTok{"eval"}\NormalTok{: }\StringTok{"test\_cv"}\NormalTok{,}
     \StringTok{"k\_folds"}\NormalTok{: }\DecValTok{10}\NormalTok{,}
\NormalTok{\})}
\NormalTok{evaluate\_cv(model}\OperatorTok{=}\NormalTok{model\_spot, fun\_control}\OperatorTok{=}\NormalTok{fun\_control, verbose}\OperatorTok{=}\DecValTok{0}\NormalTok{)}
\end{Highlighting}
\end{Shaded}

\begin{verbatim}
Error in fun_sklearn(). Call to evaluate_cv failed. err=ValueError('n_splits=10 cannot be greater than the number of members in each class.'), type(err)=<class 'ValueError'>
\end{verbatim}

\begin{verbatim}
(nan, None)
\end{verbatim}

\begin{itemize}
\tightlist
\item
  This is the evaluation that will be used in the comparison:
\end{itemize}

\begin{Shaded}
\begin{Highlighting}[]
\NormalTok{fun\_control.update(\{}
     \StringTok{"eval"}\NormalTok{: }\StringTok{"data\_cv"}\NormalTok{,}
     \StringTok{"k\_folds"}\NormalTok{: }\DecValTok{10}\NormalTok{,}
\NormalTok{\})}
\NormalTok{evaluate\_cv(model}\OperatorTok{=}\NormalTok{model\_spot, fun\_control}\OperatorTok{=}\NormalTok{fun\_control, verbose}\OperatorTok{=}\DecValTok{0}\NormalTok{)}
\end{Highlighting}
\end{Shaded}

\begin{verbatim}
(0.8348974358974359, None)
\end{verbatim}

\hypertarget{detailed-hyperparameter-plots-4}{%
\subsection{Detailed Hyperparameter
Plots}\label{detailed-hyperparameter-plots-4}}

\begin{Shaded}
\begin{Highlighting}[]
\NormalTok{filename }\OperatorTok{=} \StringTok{"./figures/"} \OperatorTok{+}\NormalTok{ experiment\_name}
\NormalTok{spot\_tuner.plot\_important\_hyperparameter\_contour(filename}\OperatorTok{=}\NormalTok{filename)}
\end{Highlighting}
\end{Shaded}

\begin{verbatim}
learning_rate:  0.3567677848399139
max_iter:  0.13022107198445454
min_samples_leaf:  1.6076317023925468
early_stopping:  100.0
\end{verbatim}

\begin{figure}[H]

{\centering \includegraphics{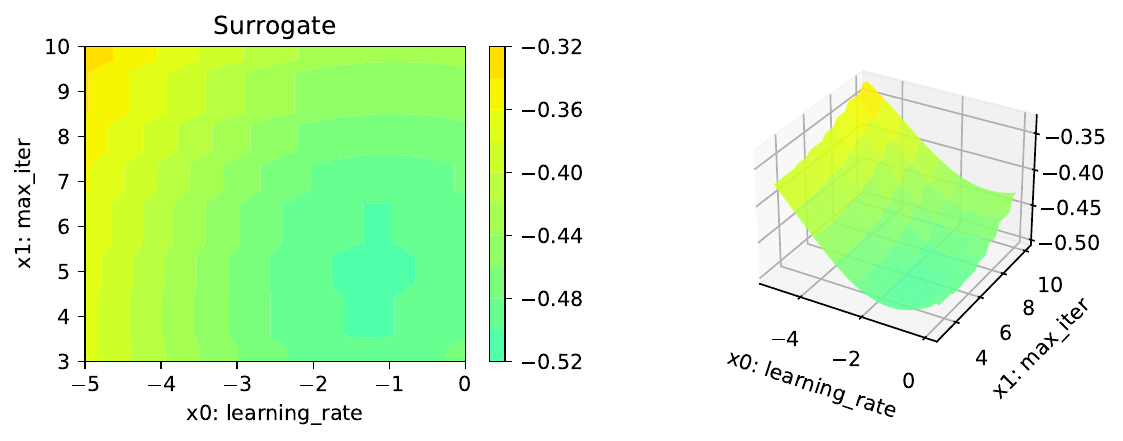}

}

\end{figure}

\begin{figure}[H]

{\centering \includegraphics{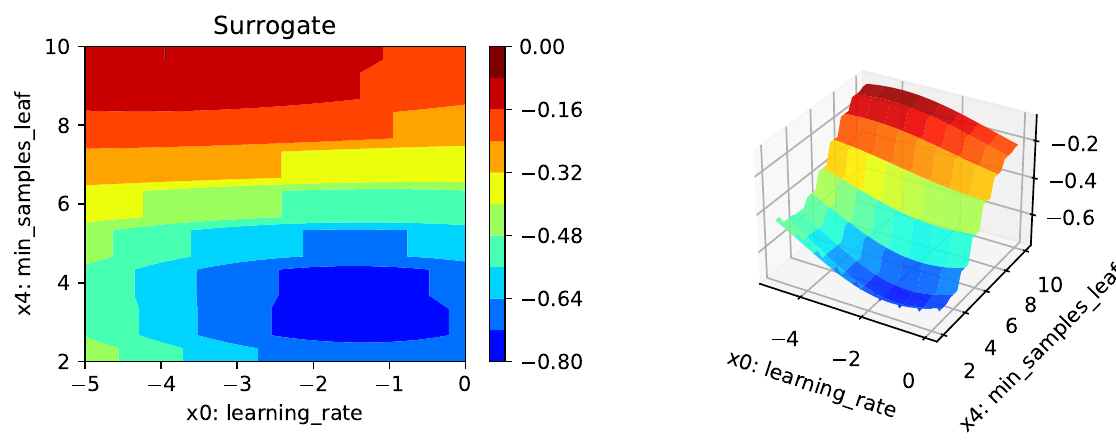}

}

\end{figure}

\begin{figure}[H]

{\centering \includegraphics{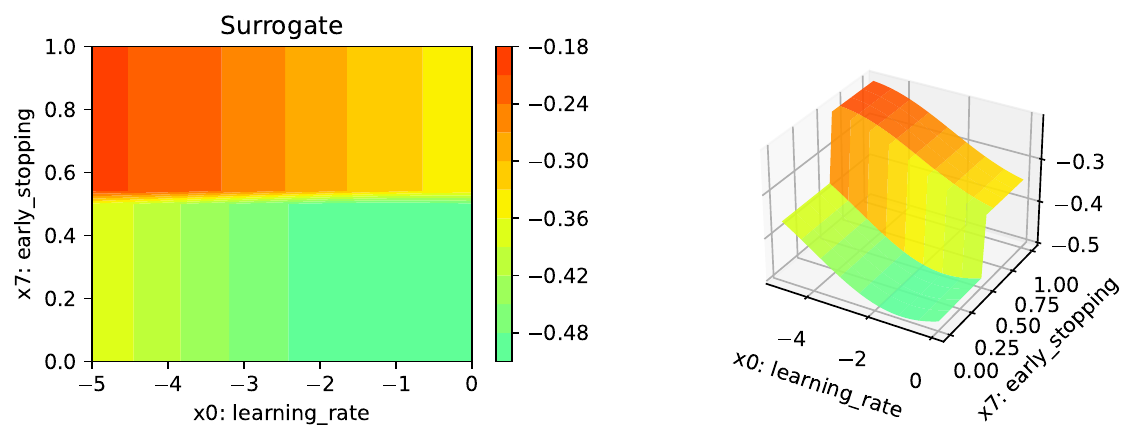}

}

\end{figure}

\begin{figure}[H]

{\centering \includegraphics{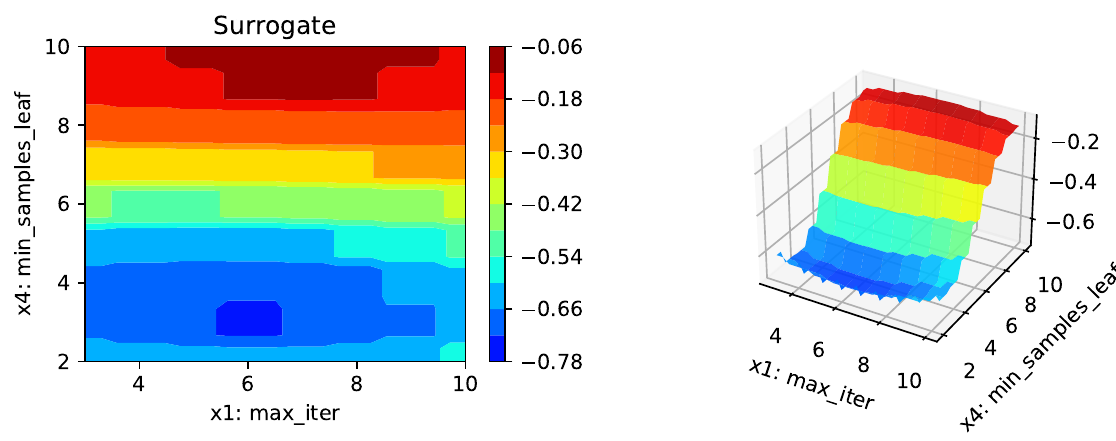}

}

\end{figure}

\begin{figure}[H]

{\centering \includegraphics{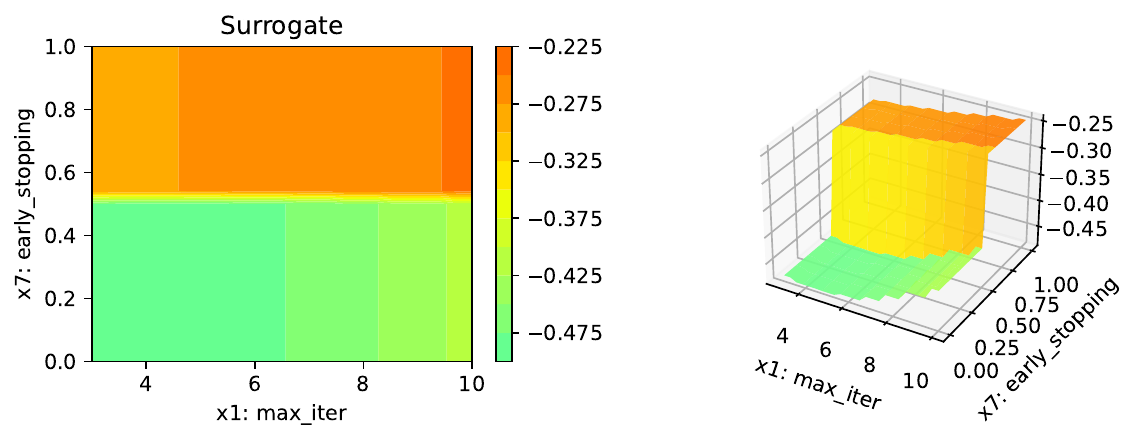}

}

\end{figure}

\begin{figure}[H]

{\centering \includegraphics{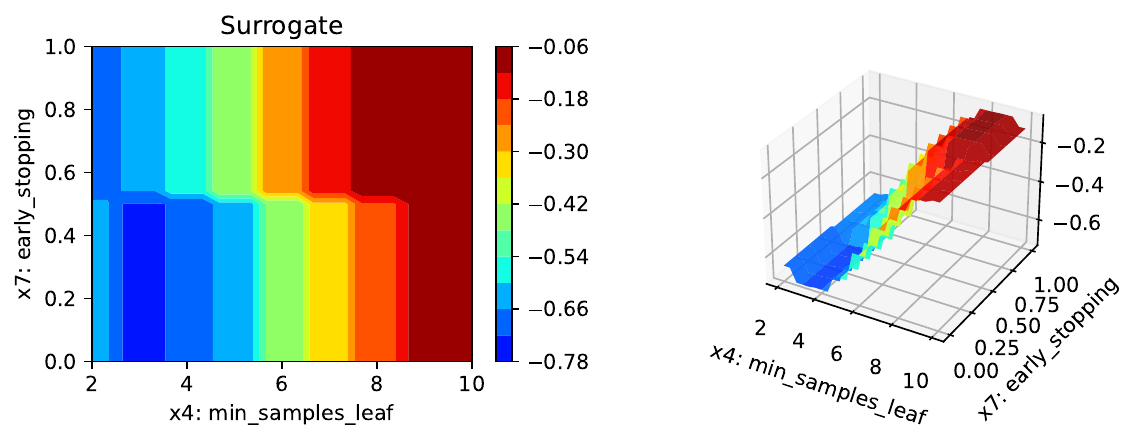}

}

\end{figure}

\hypertarget{parallel-coordinates-plot-2}{%
\subsection{Parallel Coordinates
Plot}\label{parallel-coordinates-plot-2}}

\begin{Shaded}
\begin{Highlighting}[]
\NormalTok{spot\_tuner.parallel\_plot()}
\end{Highlighting}
\end{Shaded}

\begin{verbatim}
Unable to display output for mime type(s): text/html
\end{verbatim}

\begin{verbatim}
Unable to display output for mime type(s): text/html
\end{verbatim}

\hypertarget{plot-all-combinations-of-hyperparameters-3}{%
\subsection{Plot all Combinations of
Hyperparameters}\label{plot-all-combinations-of-hyperparameters-3}}

\begin{itemize}
\tightlist
\item
  Warning: this may take a while.
\end{itemize}

\begin{Shaded}
\begin{Highlighting}[]
\NormalTok{PLOT\_ALL }\OperatorTok{=} \VariableTok{False}
\ControlFlowTok{if}\NormalTok{ PLOT\_ALL:}
\NormalTok{    n }\OperatorTok{=}\NormalTok{ spot\_tuner.k}
    \ControlFlowTok{for}\NormalTok{ i }\KeywordTok{in} \BuiltInTok{range}\NormalTok{(n}\OperatorTok{{-}}\DecValTok{1}\NormalTok{):}
        \ControlFlowTok{for}\NormalTok{ j }\KeywordTok{in} \BuiltInTok{range}\NormalTok{(i}\OperatorTok{+}\DecValTok{1}\NormalTok{, n):}
\NormalTok{            spot\_tuner.plot\_contour(i}\OperatorTok{=}\NormalTok{i, j}\OperatorTok{=}\NormalTok{j, min\_z}\OperatorTok{=}\NormalTok{min\_z, max\_z }\OperatorTok{=}\NormalTok{ max\_z)}
\end{Highlighting}
\end{Shaded}

\hypertarget{sec-hpt-sklearn-svc-vbdp-data}{%
\chapter{HPT: sklearn SVC VBDP
Data}\label{sec-hpt-sklearn-svc-vbdp-data}}

This chapter describes the hyperparameter tuning of a \texttt{SVC} on
the Vector Borne Disease Prediction (VBDP) data set.

\begin{tcolorbox}[enhanced jigsaw, left=2mm, title=\textcolor{quarto-callout-important-color}{\faExclamation}\hspace{0.5em}{Vector Borne Disease Prediction Data Set}, bottomrule=.15mm, titlerule=0mm, breakable, rightrule=.15mm, toprule=.15mm, coltitle=black, colbacktitle=quarto-callout-important-color!10!white, leftrule=.75mm, arc=.35mm, colframe=quarto-callout-important-color-frame, bottomtitle=1mm, colback=white, opacitybacktitle=0.6, toptitle=1mm, opacityback=0]

This chapter uses the Vector Borne Disease Prediction data set from
Kaggle. It is a categorical dataset for eleven Vector Borne Diseases
with associated symptoms.

\begin{quote}
The person who associated a work with this deed has dedicated the work
to the public domain by waiving all of his or her rights to the work
worldwide under copyright law, including all related and neighboring
rights, to the extent allowed by law.You can copy, modify, distribute
and perform the work, even for commercial purposes, all without asking
permission. See Other Information below, see
\url{https://creativecommons.org/publicdomain/zero/1.0/}.
\end{quote}

The data set is available at:
\url{https://www.kaggle.com/datasets/richardbernat/vector-borne-disease-prediction},

The data should be downloaded and stored in the \texttt{data/VBDP}
subfolder. The data set is not available as a part of the
\texttt{spotPython} package.

\end{tcolorbox}

\hypertarget{sec-setup-18}{%
\section{Step 1: Setup}\label{sec-setup-18}}

Before we consider the detailed experimental setup, we select the
parameters that affect run time and the initial design size.

\begin{Shaded}
\begin{Highlighting}[]
\NormalTok{MAX\_TIME }\OperatorTok{=} \DecValTok{1}
\NormalTok{INIT\_SIZE }\OperatorTok{=} \DecValTok{5}
\NormalTok{ORIGINAL }\OperatorTok{=} \VariableTok{True}
\NormalTok{PREFIX }\OperatorTok{=} \StringTok{"18"}
\end{Highlighting}
\end{Shaded}

\begin{Shaded}
\begin{Highlighting}[]
\ImportTok{import}\NormalTok{ warnings}
\NormalTok{warnings.filterwarnings(}\StringTok{"ignore"}\NormalTok{)}
\end{Highlighting}
\end{Shaded}

\hypertarget{step-2-initialization-of-the-empty-fun_control-dictionary-3}{%
\section{\texorpdfstring{Step 2: Initialization of the Empty
\texttt{fun\_control}
Dictionary}{Step 2: Initialization of the Empty fun\_control Dictionary}}\label{step-2-initialization-of-the-empty-fun_control-dictionary-3}}

\begin{Shaded}
\begin{Highlighting}[]
\ImportTok{from}\NormalTok{ spotPython.utils.init }\ImportTok{import}\NormalTok{ fun\_control\_init}
\ImportTok{from}\NormalTok{ spotPython.utils.}\BuiltInTok{file} \ImportTok{import}\NormalTok{ get\_experiment\_name, get\_spot\_tensorboard\_path}
\ImportTok{from}\NormalTok{ spotPython.utils.device }\ImportTok{import}\NormalTok{ getDevice}

\NormalTok{experiment\_name }\OperatorTok{=}\NormalTok{ get\_experiment\_name(prefix}\OperatorTok{=}\NormalTok{PREFIX)}

\NormalTok{fun\_control }\OperatorTok{=}\NormalTok{ fun\_control\_init(}
\NormalTok{    task}\OperatorTok{=}\StringTok{"classification"}\NormalTok{,}
\NormalTok{    spot\_tensorboard\_path}\OperatorTok{=}\NormalTok{get\_spot\_tensorboard\_path(experiment\_name))}
\end{Highlighting}
\end{Shaded}

\hypertarget{step-3-pytorch-data-loading-1}{%
\section{Step 3: PyTorch Data
Loading}\label{step-3-pytorch-data-loading-1}}

\hypertarget{load-data-classification-vbdp-2}{%
\subsection{1. Load Data: Classification
VBDP}\label{load-data-classification-vbdp-2}}

\begin{Shaded}
\begin{Highlighting}[]
\ImportTok{import}\NormalTok{ pandas }\ImportTok{as}\NormalTok{ pd}
\ControlFlowTok{if}\NormalTok{ ORIGINAL }\OperatorTok{==} \VariableTok{True}\NormalTok{:}
\NormalTok{    train\_df }\OperatorTok{=}\NormalTok{ pd.read\_csv(}\StringTok{\textquotesingle{}./data/VBDP/trainn.csv\textquotesingle{}}\NormalTok{)}
\NormalTok{    test\_df }\OperatorTok{=}\NormalTok{ pd.read\_csv(}\StringTok{\textquotesingle{}./data/VBDP/testt.csv\textquotesingle{}}\NormalTok{)}
\ControlFlowTok{else}\NormalTok{:}
\NormalTok{    train\_df }\OperatorTok{=}\NormalTok{ pd.read\_csv(}\StringTok{\textquotesingle{}./data/VBDP/train.csv\textquotesingle{}}\NormalTok{)}
    \CommentTok{\# remove the id column}
\NormalTok{    train\_df }\OperatorTok{=}\NormalTok{ train\_df.drop(columns}\OperatorTok{=}\NormalTok{[}\StringTok{\textquotesingle{}id\textquotesingle{}}\NormalTok{])}
\end{Highlighting}
\end{Shaded}

\begin{Shaded}
\begin{Highlighting}[]
\ImportTok{from}\NormalTok{ sklearn.preprocessing }\ImportTok{import}\NormalTok{ OrdinalEncoder}
\NormalTok{n\_samples }\OperatorTok{=}\NormalTok{ train\_df.shape[}\DecValTok{0}\NormalTok{]}
\NormalTok{n\_features }\OperatorTok{=}\NormalTok{ train\_df.shape[}\DecValTok{1}\NormalTok{] }\OperatorTok{{-}} \DecValTok{1}
\NormalTok{target\_column }\OperatorTok{=} \StringTok{"prognosis"}
\CommentTok{\# Encoder our prognosis labels as integers for easier decoding later}
\NormalTok{enc }\OperatorTok{=}\NormalTok{ OrdinalEncoder()}
\NormalTok{train\_df[target\_column] }\OperatorTok{=}\NormalTok{ enc.fit\_transform(train\_df[[target\_column]])}
\NormalTok{train\_df.columns }\OperatorTok{=}\NormalTok{ [}\SpecialStringTok{f"x}\SpecialCharTok{\{}\NormalTok{i}\SpecialCharTok{\}}\SpecialStringTok{"} \ControlFlowTok{for}\NormalTok{ i }\KeywordTok{in} \BuiltInTok{range}\NormalTok{(}\DecValTok{1}\NormalTok{, n\_features}\OperatorTok{+}\DecValTok{1}\NormalTok{)] }\OperatorTok{+}\NormalTok{ [target\_column]}
\BuiltInTok{print}\NormalTok{(train\_df.shape)}
\NormalTok{train\_df.head()}
\end{Highlighting}
\end{Shaded}

\begin{verbatim}
(252, 65)
\end{verbatim}

\begin{longtable}[]{@{}llllllllllllllllllllll@{}}
\toprule\noalign{}
& x1 & x2 & x3 & x4 & x5 & x6 & x7 & x8 & x9 & x10 & ... & x56 & x57 &
x58 & x59 & x60 & x61 & x62 & x63 & x64 & prognosis \\
\midrule\noalign{}
\endhead
\bottomrule\noalign{}
\endlastfoot
0 & 0 & 1 & 1 & 1 & 1 & 0 & 1 & 0 & 0 & 0 & ... & 0 & 0 & 0 & 0 & 0 & 0
& 0 & 0 & 0 & 0.0 \\
1 & 1 & 1 & 1 & 1 & 1 & 0 & 1 & 1 & 1 & 0 & ... & 0 & 0 & 0 & 0 & 0 & 0
& 0 & 0 & 0 & 0.0 \\
2 & 0 & 1 & 0 & 1 & 0 & 0 & 1 & 1 & 0 & 0 & ... & 0 & 0 & 0 & 0 & 0 & 0
& 0 & 0 & 0 & 0.0 \\
3 & 0 & 0 & 0 & 0 & 0 & 1 & 1 & 1 & 0 & 0 & ... & 0 & 0 & 0 & 0 & 0 & 0
& 0 & 0 & 0 & 0.0 \\
4 & 1 & 0 & 0 & 0 & 1 & 1 & 1 & 1 & 0 & 0 & ... & 0 & 0 & 0 & 0 & 0 & 0
& 0 & 0 & 0 & 0.0 \\
\end{longtable}

The full data set \texttt{train\_df} 64 features. The target column is
labeled as \texttt{prognosis}.

\hypertarget{holdout-train-and-test-data-2}{%
\subsection{Holdout Train and Test
Data}\label{holdout-train-and-test-data-2}}

We split out a hold-out test set (25\% of the data) so we can calculate
an example MAP@K

\begin{Shaded}
\begin{Highlighting}[]
\ImportTok{import}\NormalTok{ numpy }\ImportTok{as}\NormalTok{ np}
\ImportTok{from}\NormalTok{ sklearn.model\_selection }\ImportTok{import}\NormalTok{ train\_test\_split}
\NormalTok{X\_train, X\_test, y\_train, y\_test }\OperatorTok{=}\NormalTok{ train\_test\_split(train\_df.drop(target\_column, axis}\OperatorTok{=}\DecValTok{1}\NormalTok{), train\_df[target\_column],}
\NormalTok{                                                    random\_state}\OperatorTok{=}\DecValTok{42}\NormalTok{,}
\NormalTok{                                                    test\_size}\OperatorTok{=}\FloatTok{0.25}\NormalTok{,}
\NormalTok{                                                    stratify}\OperatorTok{=}\NormalTok{train\_df[target\_column])}
\NormalTok{train }\OperatorTok{=}\NormalTok{ pd.DataFrame(np.hstack((X\_train, np.array(y\_train).reshape(}\OperatorTok{{-}}\DecValTok{1}\NormalTok{, }\DecValTok{1}\NormalTok{))))}
\NormalTok{test }\OperatorTok{=}\NormalTok{ pd.DataFrame(np.hstack((X\_test, np.array(y\_test).reshape(}\OperatorTok{{-}}\DecValTok{1}\NormalTok{, }\DecValTok{1}\NormalTok{))))}
\NormalTok{train.columns }\OperatorTok{=}\NormalTok{ [}\SpecialStringTok{f"x}\SpecialCharTok{\{}\NormalTok{i}\SpecialCharTok{\}}\SpecialStringTok{"} \ControlFlowTok{for}\NormalTok{ i }\KeywordTok{in} \BuiltInTok{range}\NormalTok{(}\DecValTok{1}\NormalTok{, n\_features}\OperatorTok{+}\DecValTok{1}\NormalTok{)] }\OperatorTok{+}\NormalTok{ [target\_column]}
\NormalTok{test.columns }\OperatorTok{=}\NormalTok{ [}\SpecialStringTok{f"x}\SpecialCharTok{\{}\NormalTok{i}\SpecialCharTok{\}}\SpecialStringTok{"} \ControlFlowTok{for}\NormalTok{ i }\KeywordTok{in} \BuiltInTok{range}\NormalTok{(}\DecValTok{1}\NormalTok{, n\_features}\OperatorTok{+}\DecValTok{1}\NormalTok{)] }\OperatorTok{+}\NormalTok{ [target\_column]}
\BuiltInTok{print}\NormalTok{(train.shape)}
\BuiltInTok{print}\NormalTok{(test.shape)}
\NormalTok{train.head()}
\end{Highlighting}
\end{Shaded}

\begin{verbatim}
(189, 65)
(63, 65)
\end{verbatim}

\begin{longtable}[]{@{}llllllllllllllllllllll@{}}
\toprule\noalign{}
& x1 & x2 & x3 & x4 & x5 & x6 & x7 & x8 & x9 & x10 & ... & x56 & x57 &
x58 & x59 & x60 & x61 & x62 & x63 & x64 & prognosis \\
\midrule\noalign{}
\endhead
\bottomrule\noalign{}
\endlastfoot
0 & 1.0 & 0.0 & 0.0 & 1.0 & 0.0 & 1.0 & 0.0 & 0.0 & 0.0 & 1.0 & ... &
0.0 & 0.0 & 0.0 & 0.0 & 1.0 & 1.0 & 1.0 & 0.0 & 0.0 & 7.0 \\
1 & 1.0 & 0.0 & 1.0 & 1.0 & 1.0 & 1.0 & 1.0 & 0.0 & 1.0 & 1.0 & ... &
0.0 & 1.0 & 1.0 & 1.0 & 1.0 & 0.0 & 1.0 & 1.0 & 1.0 & 3.0 \\
2 & 0.0 & 0.0 & 1.0 & 0.0 & 1.0 & 0.0 & 0.0 & 0.0 & 0.0 & 0.0 & ... &
0.0 & 0.0 & 0.0 & 0.0 & 0.0 & 0.0 & 0.0 & 0.0 & 0.0 & 10.0 \\
3 & 1.0 & 1.0 & 1.0 & 1.0 & 1.0 & 1.0 & 0.0 & 0.0 & 1.0 & 1.0 & ... &
1.0 & 0.0 & 1.0 & 1.0 & 1.0 & 0.0 & 0.0 & 1.0 & 1.0 & 3.0 \\
4 & 1.0 & 1.0 & 1.0 & 0.0 & 1.0 & 1.0 & 0.0 & 1.0 & 1.0 & 0.0 & ... &
0.0 & 0.0 & 0.0 & 0.0 & 0.0 & 0.0 & 0.0 & 0.0 & 0.0 & 8.0 \\
\end{longtable}

\begin{Shaded}
\begin{Highlighting}[]
\CommentTok{\# add the dataset to the fun\_control}
\NormalTok{fun\_control.update(\{}\StringTok{"data"}\NormalTok{: train\_df, }\CommentTok{\# full dataset,}
               \StringTok{"train"}\NormalTok{: train,}
               \StringTok{"test"}\NormalTok{: test,}
               \StringTok{"n\_samples"}\NormalTok{: n\_samples,}
               \StringTok{"target\_column"}\NormalTok{: target\_column\})}
\end{Highlighting}
\end{Shaded}

\hypertarget{sec-specification-of-preprocessing-model-18}{%
\section{Step 4: Specification of the Preprocessing
Model}\label{sec-specification-of-preprocessing-model-18}}

Data preprocesssing can be very simple, e.g., you can ignore it. Then
you would choose the \texttt{prep\_model} ``None'':

\begin{Shaded}
\begin{Highlighting}[]
\NormalTok{prep\_model }\OperatorTok{=} \VariableTok{None}
\NormalTok{fun\_control.update(\{}\StringTok{"prep\_model"}\NormalTok{: prep\_model\})}
\end{Highlighting}
\end{Shaded}

A default approach for numerical data is the \texttt{StandardScaler}
(mean 0, variance 1). This can be selected as follows:

\begin{Shaded}
\begin{Highlighting}[]
\CommentTok{\# prep\_model = StandardScaler()}
\CommentTok{\# fun\_control.update(\{"prep\_model": prep\_model\})}
\end{Highlighting}
\end{Shaded}

Even more complicated pre-processing steps are possible, e.g., the
follwing pipeline:

\begin{Shaded}
\begin{Highlighting}[]
\CommentTok{\# categorical\_columns = []}
\CommentTok{\# one\_hot\_encoder = OneHotEncoder(handle\_unknown="ignore", sparse\_output=False)}
\CommentTok{\# prep\_model = ColumnTransformer(}
\CommentTok{\#         transformers=[}
\CommentTok{\#             ("categorical", one\_hot\_encoder, categorical\_columns),}
\CommentTok{\#         ],}
\CommentTok{\#         remainder=StandardScaler(),}
\CommentTok{\#     )}
\end{Highlighting}
\end{Shaded}

\hypertarget{step-5-select-model-algorithm-and-core_model_hyper_dict-3}{%
\section{\texorpdfstring{Step 5: Select Model (\texttt{algorithm}) and
\texttt{core\_model\_hyper\_dict}}{Step 5: Select Model (algorithm) and core\_model\_hyper\_dict}}\label{step-5-select-model-algorithm-and-core_model_hyper_dict-3}}

The selection of the algorithm (ML model) that should be tuned is done
by specifying the its name from the \texttt{sklearn} implementation. For
example, the \texttt{SVC} support vector machine classifier is selected
as follows:

\texttt{add\_core\_model\_to\_fun\_control(SVC,\ fun\_control,\ SklearnHyperDict)}

Other core\_models are, e.g.,:

\begin{itemize}
\tightlist
\item
  RidgeCV
\item
  GradientBoostingRegressor
\item
  ElasticNet
\item
  RandomForestClassifier
\item
  LogisticRegression
\item
  KNeighborsClassifier
\item
  RandomForestClassifier
\item
  GradientBoostingClassifier
\item
  HistGradientBoostingClassifier
\end{itemize}

We will use the \texttt{RandomForestClassifier} classifier in this
example.

\begin{Shaded}
\begin{Highlighting}[]
\ImportTok{from}\NormalTok{ sklearn.linear\_model }\ImportTok{import}\NormalTok{ RidgeCV}
\ImportTok{from}\NormalTok{ sklearn.ensemble }\ImportTok{import}\NormalTok{ RandomForestClassifier}
\ImportTok{from}\NormalTok{ sklearn.svm }\ImportTok{import}\NormalTok{ SVC}
\ImportTok{from}\NormalTok{ sklearn.linear\_model }\ImportTok{import}\NormalTok{ LogisticRegression}
\ImportTok{from}\NormalTok{ sklearn.neighbors }\ImportTok{import}\NormalTok{ KNeighborsClassifier}
\ImportTok{from}\NormalTok{ sklearn.ensemble }\ImportTok{import}\NormalTok{ GradientBoostingClassifier}
\ImportTok{from}\NormalTok{ sklearn.ensemble }\ImportTok{import}\NormalTok{ GradientBoostingRegressor}
\ImportTok{from}\NormalTok{ sklearn.ensemble }\ImportTok{import}\NormalTok{ HistGradientBoostingClassifier}
\ImportTok{from}\NormalTok{ sklearn.linear\_model }\ImportTok{import}\NormalTok{ ElasticNet}
\ImportTok{from}\NormalTok{ spotPython.hyperparameters.values }\ImportTok{import}\NormalTok{ add\_core\_model\_to\_fun\_control}
\ImportTok{from}\NormalTok{ spotPython.data.sklearn\_hyper\_dict }\ImportTok{import}\NormalTok{ SklearnHyperDict}
\ImportTok{from}\NormalTok{ spotPython.fun.hypersklearn }\ImportTok{import}\NormalTok{ HyperSklearn}
\end{Highlighting}
\end{Shaded}

\begin{Shaded}
\begin{Highlighting}[]
\CommentTok{\# core\_model  = RidgeCV}
\CommentTok{\# core\_model = GradientBoostingRegressor}
\CommentTok{\# core\_model = ElasticNet}
\CommentTok{\# core\_model = RandomForestClassifier}
\NormalTok{core\_model }\OperatorTok{=}\NormalTok{ SVC}
\CommentTok{\# core\_model = LogisticRegression}
\CommentTok{\# core\_model = KNeighborsClassifier}
\CommentTok{\# core\_model = GradientBoostingClassifier}
\CommentTok{\# core\_model = HistGradientBoostingClassifier}
\NormalTok{add\_core\_model\_to\_fun\_control(core\_model}\OperatorTok{=}\NormalTok{core\_model,}
\NormalTok{                              fun\_control}\OperatorTok{=}\NormalTok{fun\_control,}
\NormalTok{                              hyper\_dict}\OperatorTok{=}\NormalTok{SklearnHyperDict,}
\NormalTok{                              filename}\OperatorTok{=}\VariableTok{None}\NormalTok{)}
\end{Highlighting}
\end{Shaded}

Now \texttt{fun\_control} has the information from the JSON file. The
available hyperparameters are:

\begin{Shaded}
\begin{Highlighting}[]
\BuiltInTok{print}\NormalTok{(}\OperatorTok{*}\NormalTok{fun\_control[}\StringTok{"core\_model\_hyper\_dict"}\NormalTok{].keys(), sep}\OperatorTok{=}\StringTok{"}\CharTok{\textbackslash{}n}\StringTok{"}\NormalTok{)}
\end{Highlighting}
\end{Shaded}

\begin{verbatim}
C
kernel
degree
gamma
coef0
shrinking
probability
tol
cache_size
break_ties
\end{verbatim}

\hypertarget{step-6-modify-hyper_dict-hyperparameters-for-the-selected-algorithm-aka-core_model-3}{%
\section{\texorpdfstring{Step 6: Modify \texttt{hyper\_dict}
Hyperparameters for the Selected Algorithm aka
\texttt{core\_model}}{Step 6: Modify hyper\_dict Hyperparameters for the Selected Algorithm aka core\_model}}\label{step-6-modify-hyper_dict-hyperparameters-for-the-selected-algorithm-aka-core_model-3}}

\hypertarget{modify-hyperparameter-of-type-numeric-and-integer-boolean-3}{%
\subsection{Modify hyperparameter of type numeric and integer
(boolean)}\label{modify-hyperparameter-of-type-numeric-and-integer-boolean-3}}

Numeric and boolean values can be modified using the
\texttt{modify\_hyper\_parameter\_bounds} method. For example, to change
the \texttt{tol} hyperparameter of the \texttt{SVC} model to the
interval {[}1e-3, 1e-2{]}, the following code can be used:

\texttt{modify\_hyper\_parameter\_bounds(fun\_control,\ "tol",\ bounds={[}1e-3,\ 1e-2{]})}

\begin{Shaded}
\begin{Highlighting}[]
\ImportTok{from}\NormalTok{ spotPython.hyperparameters.values }\ImportTok{import}\NormalTok{ modify\_hyper\_parameter\_bounds}
\NormalTok{modify\_hyper\_parameter\_bounds(fun\_control, }\StringTok{"probability"}\NormalTok{, bounds}\OperatorTok{=}\NormalTok{[}\DecValTok{1}\NormalTok{, }\DecValTok{1}\NormalTok{])}
\end{Highlighting}
\end{Shaded}

\hypertarget{modify-hyperparameter-of-type-factor-4}{%
\subsection{Modify hyperparameter of type
factor}\label{modify-hyperparameter-of-type-factor-4}}

\texttt{spotPython} provides functions for modifying the
hyperparameters, their bounds and factors as well as for activating and
de-activating hyperparameters without re-compilation of the Python
source code. These functions were described in
Section~\ref{sec-modification-of-hyperparameters-14}.

Factors can be modified with the
\texttt{modify\_hyper\_parameter\_levels} function. For example, to
exclude the \texttt{sigmoid} kernel from the tuning, the \texttt{kernel}
hyperparameter of the \texttt{SVC} model can be modified as follows:

\texttt{modify\_hyper\_parameter\_levels(fun\_control,\ "kernel",\ {[}"linear",\ "rbf"{]})}

The new setting can be controlled via:

\texttt{fun\_control{[}"core\_model\_hyper\_dict"{]}{[}"kernel"{]}}

\begin{Shaded}
\begin{Highlighting}[]
\ImportTok{from}\NormalTok{ spotPython.hyperparameters.values }\ImportTok{import}\NormalTok{ modify\_hyper\_parameter\_levels}
\NormalTok{modify\_hyper\_parameter\_levels(fun\_control, }\StringTok{"kernel"}\NormalTok{, [}\StringTok{"rbf"}\NormalTok{])}
\end{Highlighting}
\end{Shaded}

\hypertarget{sec-optimizers-18}{%
\subsection{Optimizers}\label{sec-optimizers-18}}

Optimizers are described in Section~\ref{sec-optimizers-14}.

\hypertarget{selection-of-the-objective-metric-and-loss-functions-2}{%
\subsection{Selection of the Objective: Metric and Loss
Functions}\label{selection-of-the-objective-metric-and-loss-functions-2}}

\begin{itemize}
\tightlist
\item
  Machine learning models are optimized with respect to a metric, for
  example, the \texttt{accuracy} function.
\item
  Deep learning, e.g., neural networks are optimized with respect to a
  loss function, for example, the \texttt{cross\_entropy} function and
  evaluated with respect to a metric, for example, the \texttt{accuracy}
  function.
\end{itemize}

\hypertarget{step-7-selection-of-the-objective-loss-function-4}{%
\section{Step 7: Selection of the Objective (Loss)
Function}\label{step-7-selection-of-the-objective-loss-function-4}}

The loss function, that is usually used in deep learning for optimizing
the weights of the net, is stored in the \texttt{fun\_control}
dictionary as \texttt{"loss\_function"}.

\hypertarget{metric-function-2}{%
\subsection{Metric Function}\label{metric-function-2}}

There are two different types of metrics in \texttt{spotPython}:

\begin{enumerate}
\def\labelenumi{\arabic{enumi}.}
\tightlist
\item
  \texttt{"metric\_river"} is used for the river based evaluation via
  \texttt{eval\_oml\_iter\_progressive}.
\item
  \texttt{"metric\_sklearn"} is used for the sklearn based evaluation.
\end{enumerate}

We will consider multi-class classification metrics, e.g.,
\texttt{mapk\_score} and \texttt{top\_k\_accuracy\_score}.

\begin{tcolorbox}[enhanced jigsaw, left=2mm, title=\textcolor{quarto-callout-note-color}{\faInfo}\hspace{0.5em}{Predict Probabilities}, bottomrule=.15mm, titlerule=0mm, breakable, rightrule=.15mm, toprule=.15mm, coltitle=black, colbacktitle=quarto-callout-note-color!10!white, leftrule=.75mm, arc=.35mm, colframe=quarto-callout-note-color-frame, bottomtitle=1mm, colback=white, opacitybacktitle=0.6, toptitle=1mm, opacityback=0]

In this multi-class classification example the machine learning
algorithm should return the probabilities of the specific classes
(\texttt{"predict\_proba"}) instead of the predicted values.

\end{tcolorbox}

We set \texttt{"predict\_proba"} to \texttt{True} in the
\texttt{fun\_control} dictionary.

\hypertarget{the-mapk-metric-2}{%
\subsubsection{The MAPK Metric}\label{the-mapk-metric-2}}

To select the MAPK metric, the following two entries can be added to the
\texttt{fun\_control} dictionary:

\texttt{"metric\_sklearn":\ mapk\_score"}

\texttt{"metric\_params":\ \{"k":\ 3\}}.

\hypertarget{other-metrics-2}{%
\subsubsection{Other Metrics}\label{other-metrics-2}}

Alternatively, other metrics for multi-class classification can be used,
e.g.,: * top\_k\_accuracy\_score or * roc\_auc\_score

The metric \texttt{roc\_auc\_score} requires the parameter
\texttt{"multi\_class"}, e.g.,

\texttt{"multi\_class":\ "ovr"}.

This is set in the \texttt{fun\_control} dictionary.

\begin{tcolorbox}[enhanced jigsaw, left=2mm, title=\textcolor{quarto-callout-note-color}{\faInfo}\hspace{0.5em}{Weights}, bottomrule=.15mm, titlerule=0mm, breakable, rightrule=.15mm, toprule=.15mm, coltitle=black, colbacktitle=quarto-callout-note-color!10!white, leftrule=.75mm, arc=.35mm, colframe=quarto-callout-note-color-frame, bottomtitle=1mm, colback=white, opacitybacktitle=0.6, toptitle=1mm, opacityback=0]

\texttt{spotPython} performs a minimization, therefore, metrics that
should be maximized have to be multiplied by -1. This is done by setting
\texttt{"weights"} to \texttt{-1}.

\end{tcolorbox}

\begin{itemize}
\tightlist
\item
  The complete setup for the metric in our example is:
\end{itemize}

\begin{Shaded}
\begin{Highlighting}[]
\ImportTok{from}\NormalTok{ spotPython.utils.metrics }\ImportTok{import}\NormalTok{ mapk\_score}
\NormalTok{fun\_control.update(\{}
               \StringTok{"weights"}\NormalTok{: }\OperatorTok{{-}}\DecValTok{1}\NormalTok{,}
               \StringTok{"metric\_sklearn"}\NormalTok{: mapk\_score,}
               \StringTok{"predict\_proba"}\NormalTok{: }\VariableTok{True}\NormalTok{,}
               \StringTok{"metric\_params"}\NormalTok{: \{}\StringTok{"k"}\NormalTok{: }\DecValTok{3}\NormalTok{\},}
\NormalTok{               \})}
\end{Highlighting}
\end{Shaded}

\hypertarget{evaluation-on-hold-out-data-2}{%
\subsection{Evaluation on Hold-out
Data}\label{evaluation-on-hold-out-data-2}}

\begin{itemize}
\tightlist
\item
  The default method for computing the performance is
  \texttt{"eval\_holdout"}.
\item
  Alternatively, cross-validation can be used for every machine learning
  model.
\item
  Specifically for RandomForests, the OOB-score can be used.
\end{itemize}

\begin{Shaded}
\begin{Highlighting}[]
\NormalTok{fun\_control.update(\{}
    \StringTok{"eval"}\NormalTok{: }\StringTok{"train\_hold\_out"}\NormalTok{,}
\NormalTok{\})}
\end{Highlighting}
\end{Shaded}

\hypertarget{cross-validation-3}{%
\subsubsection{Cross Validation}\label{cross-validation-3}}

Instead of using the OOB-score, the classical cross validation can be
used. The number of folds is set by the key \texttt{"k\_folds"}. For
example, to use 5-fold cross validation, the key \texttt{"k\_folds"} is
set to \texttt{5}. Uncomment the following line to use cross validation:

\begin{Shaded}
\begin{Highlighting}[]
\CommentTok{\# fun\_control.update(\{}
\CommentTok{\#      "eval": "train\_cv",}
\CommentTok{\#      "k\_folds": 10,}
\CommentTok{\# \})}
\end{Highlighting}
\end{Shaded}

\hypertarget{step-8-calling-the-spot-function-4}{%
\section{Step 8: Calling the SPOT
Function}\label{step-8-calling-the-spot-function-4}}

\hypertarget{sec-prepare-spot-call-18}{%
\subsection{Preparing the SPOT Call}\label{sec-prepare-spot-call-18}}

\begin{itemize}
\tightlist
\item
  Get types and variable names as well as lower and upper bounds for the
  hyperparameters.
\end{itemize}

\begin{Shaded}
\begin{Highlighting}[]
\CommentTok{\# extract the variable types, names, and bounds}
\ImportTok{from}\NormalTok{ spotPython.hyperparameters.values }\ImportTok{import}\NormalTok{ (get\_bound\_values,}
\NormalTok{    get\_var\_name,}
\NormalTok{    get\_var\_type,)}
\NormalTok{var\_type }\OperatorTok{=}\NormalTok{ get\_var\_type(fun\_control)}
\NormalTok{var\_name }\OperatorTok{=}\NormalTok{ get\_var\_name(fun\_control)}
\NormalTok{lower }\OperatorTok{=}\NormalTok{ get\_bound\_values(fun\_control, }\StringTok{"lower"}\NormalTok{)}
\NormalTok{upper }\OperatorTok{=}\NormalTok{ get\_bound\_values(fun\_control, }\StringTok{"upper"}\NormalTok{)}
\end{Highlighting}
\end{Shaded}

\begin{Shaded}
\begin{Highlighting}[]
\ImportTok{from}\NormalTok{ spotPython.utils.eda }\ImportTok{import}\NormalTok{ gen\_design\_table}
\BuiltInTok{print}\NormalTok{(gen\_design\_table(fun\_control))}
\end{Highlighting}
\end{Shaded}

\begin{verbatim}
| name        | type   | default   |    lower |   upper | transform   |
|-------------|--------|-----------|----------|---------|-------------|
| C           | float  | 1.0       |   0.1    |   10    | None        |
| kernel      | factor | rbf       |   0      |    0    | None        |
| degree      | int    | 3         |   3      |    3    | None        |
| gamma       | factor | scale     |   0      |    1    | None        |
| coef0       | float  | 0.0       |   0      |    0    | None        |
| shrinking   | factor | 0         |   0      |    1    | None        |
| probability | factor | 0         |   1      |    1    | None        |
| tol         | float  | 0.001     |   0.0001 |    0.01 | None        |
| cache_size  | float  | 200.0     | 100      |  400    | None        |
| break_ties  | factor | 0         |   0      |    1    | None        |
\end{verbatim}

\hypertarget{sec-the-objective-function-18}{%
\subsection{The Objective
Function}\label{sec-the-objective-function-18}}

The objective function is selected next. It implements an interface from
\texttt{sklearn}'s training, validation, and testing methods to
\texttt{spotPython}.

\begin{Shaded}
\begin{Highlighting}[]
\ImportTok{from}\NormalTok{ spotPython.fun.hypersklearn }\ImportTok{import}\NormalTok{ HyperSklearn}
\NormalTok{fun }\OperatorTok{=}\NormalTok{ HyperSklearn().fun\_sklearn}
\end{Highlighting}
\end{Shaded}

\hypertarget{run-the-spot-optimizer-4}{%
\subsection{\texorpdfstring{Run the \texttt{Spot}
Optimizer}{Run the Spot Optimizer}}\label{run-the-spot-optimizer-4}}

\begin{itemize}
\tightlist
\item
  Run SPOT for approx. x mins (\texttt{max\_time}).
\item
  Note: the run takes longer, because the evaluation time of initial
  design (here: \texttt{initi\_size}, 20 points) is not considered.
\end{itemize}

\begin{Shaded}
\begin{Highlighting}[]
\ImportTok{from}\NormalTok{ spotPython.hyperparameters.values }\ImportTok{import}\NormalTok{ get\_default\_hyperparameters\_as\_array}
\NormalTok{X\_start }\OperatorTok{=}\NormalTok{ get\_default\_hyperparameters\_as\_array(fun\_control)}
\NormalTok{X\_start}
\end{Highlighting}
\end{Shaded}

\begin{verbatim}
array([[1.e+00, 0.e+00, 3.e+00, 0.e+00, 0.e+00, 0.e+00, 0.e+00, 1.e-03,
        2.e+02, 0.e+00]])
\end{verbatim}

\begin{Shaded}
\begin{Highlighting}[]
\ImportTok{import}\NormalTok{ numpy }\ImportTok{as}\NormalTok{ np}
\ImportTok{from}\NormalTok{ spotPython.spot }\ImportTok{import}\NormalTok{ spot}
\ImportTok{from}\NormalTok{ math }\ImportTok{import}\NormalTok{ inf}
\NormalTok{spot\_tuner }\OperatorTok{=}\NormalTok{ spot.Spot(fun}\OperatorTok{=}\NormalTok{fun,}
\NormalTok{                   lower }\OperatorTok{=}\NormalTok{ lower,}
\NormalTok{                   upper }\OperatorTok{=}\NormalTok{ upper,}
\NormalTok{                   fun\_evals }\OperatorTok{=}\NormalTok{ inf,}
\NormalTok{                   fun\_repeats }\OperatorTok{=} \DecValTok{1}\NormalTok{,}
\NormalTok{                   max\_time }\OperatorTok{=}\NormalTok{ MAX\_TIME,}
\NormalTok{                   noise }\OperatorTok{=} \VariableTok{False}\NormalTok{,}
\NormalTok{                   tolerance\_x }\OperatorTok{=}\NormalTok{ np.sqrt(np.spacing(}\DecValTok{1}\NormalTok{)),}
\NormalTok{                   var\_type }\OperatorTok{=}\NormalTok{ var\_type,}
\NormalTok{                   var\_name }\OperatorTok{=}\NormalTok{ var\_name,}
\NormalTok{                   infill\_criterion }\OperatorTok{=} \StringTok{"y"}\NormalTok{,}
\NormalTok{                   n\_points }\OperatorTok{=} \DecValTok{1}\NormalTok{,}
\NormalTok{                   seed}\OperatorTok{=}\DecValTok{123}\NormalTok{,}
\NormalTok{                   log\_level }\OperatorTok{=} \DecValTok{50}\NormalTok{,}
\NormalTok{                   show\_models}\OperatorTok{=} \VariableTok{False}\NormalTok{,}
\NormalTok{                   show\_progress}\OperatorTok{=} \VariableTok{True}\NormalTok{,}
\NormalTok{                   fun\_control }\OperatorTok{=}\NormalTok{ fun\_control,}
\NormalTok{                   design\_control}\OperatorTok{=}\NormalTok{\{}\StringTok{"init\_size"}\NormalTok{: INIT\_SIZE,}
                                   \StringTok{"repeats"}\NormalTok{: }\DecValTok{1}\NormalTok{\},}
\NormalTok{                   surrogate\_control}\OperatorTok{=}\NormalTok{\{}\StringTok{"noise"}\NormalTok{: }\VariableTok{True}\NormalTok{,}
                                      \StringTok{"cod\_type"}\NormalTok{: }\StringTok{"norm"}\NormalTok{,}
                                      \StringTok{"min\_theta"}\NormalTok{: }\OperatorTok{{-}}\DecValTok{4}\NormalTok{,}
                                      \StringTok{"max\_theta"}\NormalTok{: }\DecValTok{3}\NormalTok{,}
                                      \StringTok{"n\_theta"}\NormalTok{: }\BuiltInTok{len}\NormalTok{(var\_name),}
                                      \StringTok{"model\_fun\_evals"}\NormalTok{: }\DecValTok{10\_000}\NormalTok{,}
                                      \StringTok{"log\_level"}\NormalTok{: }\DecValTok{50}
\NormalTok{                                      \})}
\NormalTok{spot\_tuner.run(X\_start}\OperatorTok{=}\NormalTok{X\_start)}
\end{Highlighting}
\end{Shaded}

\begin{verbatim}
spotPython tuning: -0.875 [----------] 0.69% 
\end{verbatim}

\begin{verbatim}
spotPython tuning: -0.875 [----------] 1.39% 
\end{verbatim}

\begin{verbatim}
spotPython tuning: -0.875 [----------] 1.95% 
\end{verbatim}

\begin{verbatim}
spotPython tuning: -0.875 [----------] 2.44% 
\end{verbatim}

\begin{verbatim}
spotPython tuning: -0.875 [----------] 2.92% 
\end{verbatim}

\begin{verbatim}
spotPython tuning: -0.875 [----------] 3.37% 
\end{verbatim}

\begin{verbatim}
spotPython tuning: -0.875 [----------] 4.03% 
\end{verbatim}

\begin{verbatim}
spotPython tuning: -0.875 [----------] 4.71% 
\end{verbatim}

\begin{verbatim}
spotPython tuning: -0.875 [#---------] 5.37% 
\end{verbatim}

\begin{verbatim}
spotPython tuning: -0.875 [#---------] 6.59% 
\end{verbatim}

\begin{verbatim}
spotPython tuning: -0.875 [#---------] 7.91% 
\end{verbatim}

\begin{verbatim}
spotPython tuning: -0.875 [#---------] 8.57% 
\end{verbatim}

\begin{verbatim}
spotPython tuning: -0.875 [#---------] 11.02% 
\end{verbatim}

\begin{verbatim}
spotPython tuning: -0.875 [#---------] 13.56% 
\end{verbatim}

\begin{verbatim}
spotPython tuning: -0.875 [##--------] 15.81% 
\end{verbatim}

\begin{verbatim}
spotPython tuning: -0.875 [##--------] 17.87% 
\end{verbatim}

\begin{verbatim}
spotPython tuning: -0.875 [##--------] 19.96% 
\end{verbatim}

\begin{verbatim}
spotPython tuning: -0.875 [##--------] 21.91% 
\end{verbatim}

\begin{verbatim}
spotPython tuning: -0.875 [##--------] 24.33% 
\end{verbatim}

\begin{verbatim}
spotPython tuning: -0.875 [###-------] 28.28% 
\end{verbatim}

\begin{verbatim}
spotPython tuning: -0.875 [###-------] 31.75% 
\end{verbatim}

\begin{verbatim}
spotPython tuning: -0.875 [####------] 35.60% 
\end{verbatim}

\begin{verbatim}
spotPython tuning: -0.875 [####------] 38.92% 
\end{verbatim}

\begin{verbatim}
spotPython tuning: -0.875 [####------] 43.60% 
\end{verbatim}

\begin{verbatim}
spotPython tuning: -0.875 [#####-----] 47.09% 
\end{verbatim}

\begin{verbatim}
spotPython tuning: -0.8854166666666666 [#####-----] 50.64% 
\end{verbatim}

\begin{verbatim}
spotPython tuning: -0.8854166666666666 [#####-----] 54.78% 
\end{verbatim}

\begin{verbatim}
spotPython tuning: -0.8854166666666666 [######----] 60.03% 
\end{verbatim}

\begin{verbatim}
spotPython tuning: -0.8854166666666666 [#######---] 65.57% 
\end{verbatim}

\begin{verbatim}
spotPython tuning: -0.8854166666666666 [#######---] 70.76% 
\end{verbatim}

\begin{verbatim}
spotPython tuning: -0.8854166666666666 [########--] 76.44% 
\end{verbatim}

\begin{verbatim}
spotPython tuning: -0.8854166666666666 [########--] 82.52% 
\end{verbatim}

\begin{verbatim}
spotPython tuning: -0.8854166666666666 [#########-] 88.58% 
\end{verbatim}

\begin{verbatim}
spotPython tuning: -0.8854166666666666 [##########] 95.42% 
\end{verbatim}

\begin{verbatim}
spotPython tuning: -0.8854166666666666 [##########] 100.00% Done...
\end{verbatim}

\begin{verbatim}
<spotPython.spot.spot.Spot at 0x17ff57c40>
\end{verbatim}

\hypertarget{sec-tensorboard-18}{%
\section{Step 9: Tensorboard}\label{sec-tensorboard-18}}

The textual output shown in the console (or code cell) can be visualized
with Tensorboard as described in Section~\ref{sec-tensorboard-14}, see
also the description in the documentation:
\href{https://sequential-parameter-optimization.github.io/spotPython/14_spot_ray_hpt_torch_cifar10.html\#sec-tensorboard-14}{Tensorboard.}

\hypertarget{sec-results-tuning-18}{%
\section{Step 10: Results}\label{sec-results-tuning-18}}

After the hyperparameter tuning run is finished, the progress of the
hyperparameter tuning can be visualized. The following code generates
the progress plot from \textbf{?@fig-progress}.

\begin{Shaded}
\begin{Highlighting}[]
\NormalTok{spot\_tuner.plot\_progress(log\_y}\OperatorTok{=}\VariableTok{False}\NormalTok{,}
\NormalTok{    filename}\OperatorTok{=}\StringTok{"./figures/"} \OperatorTok{+}\NormalTok{ experiment\_name}\OperatorTok{+}\StringTok{"\_progress.png"}\NormalTok{)}
\end{Highlighting}
\end{Shaded}

\begin{figure}[H]

{\centering \includegraphics{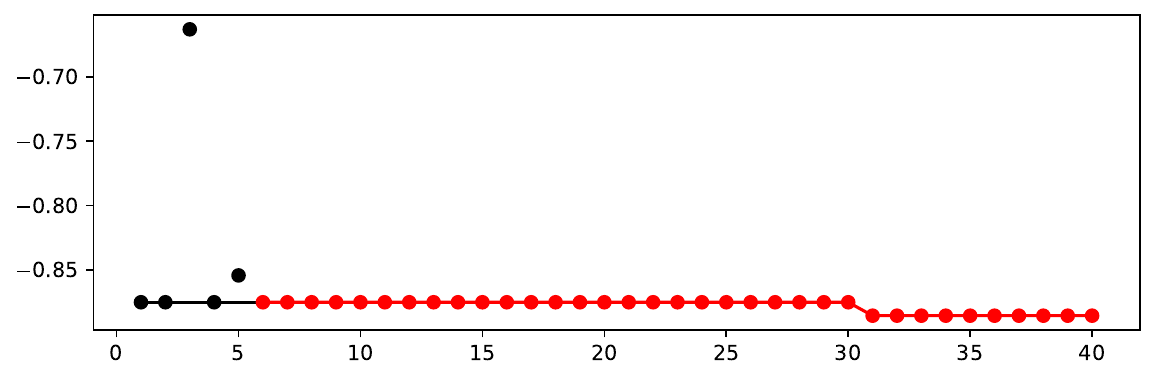}

}

\caption{Progress plot. \emph{Black} dots denote results from the
initial design. \emph{Red} dots illustrate the improvement found by the
surrogate model based optimization.}

\end{figure}

\begin{itemize}
\tightlist
\item
  Print the results
\end{itemize}

\begin{Shaded}
\begin{Highlighting}[]
\BuiltInTok{print}\NormalTok{(gen\_design\_table(fun\_control}\OperatorTok{=}\NormalTok{fun\_control,}
\NormalTok{    spot}\OperatorTok{=}\NormalTok{spot\_tuner))}
\end{Highlighting}
\end{Shaded}

\begin{verbatim}
| name        | type   | default   |   lower |   upper |                tuned | transform   |   importance | stars   |
|-------------|--------|-----------|---------|---------|----------------------|-------------|--------------|---------|
| C           | float  | 1.0       |     0.1 |    10.0 |    4.809957939164208 | None        |         2.09 | *       |
| kernel      | factor | rbf       |     0.0 |     0.0 |                  0.0 | None        |         0.00 |         |
| degree      | int    | 3         |     3.0 |     3.0 |                  3.0 | None        |         0.00 |         |
| gamma       | factor | scale     |     0.0 |     1.0 |                  0.0 | None        |         5.52 | *       |
| coef0       | float  | 0.0       |     0.0 |     0.0 |                  0.0 | None        |         0.00 |         |
| shrinking   | factor | 0         |     0.0 |     1.0 |                  0.0 | None        |         0.00 |         |
| probability | factor | 0         |     1.0 |     1.0 |                  1.0 | None        |         0.00 |         |
| tol         | float  | 0.001     |  0.0001 |    0.01 | 0.003969298209225212 | None        |         0.00 |         |
| cache_size  | float  | 200.0     |   100.0 |   400.0 |    145.9365804877652 | None        |         0.00 |         |
| break_ties  | factor | 0         |     0.0 |     1.0 |                  0.0 | None        |       100.00 | ***     |
\end{verbatim}

\hypertarget{show-variable-importance-3}{%
\subsection{Show variable importance}\label{show-variable-importance-3}}

\begin{Shaded}
\begin{Highlighting}[]
\NormalTok{spot\_tuner.plot\_importance(threshold}\OperatorTok{=}\FloatTok{0.025}\NormalTok{, filename}\OperatorTok{=}\StringTok{"./figures/"} \OperatorTok{+}\NormalTok{ experiment\_name}\OperatorTok{+}\StringTok{"\_importance.png"}\NormalTok{)}
\end{Highlighting}
\end{Shaded}

\begin{figure}[H]

{\centering \includegraphics{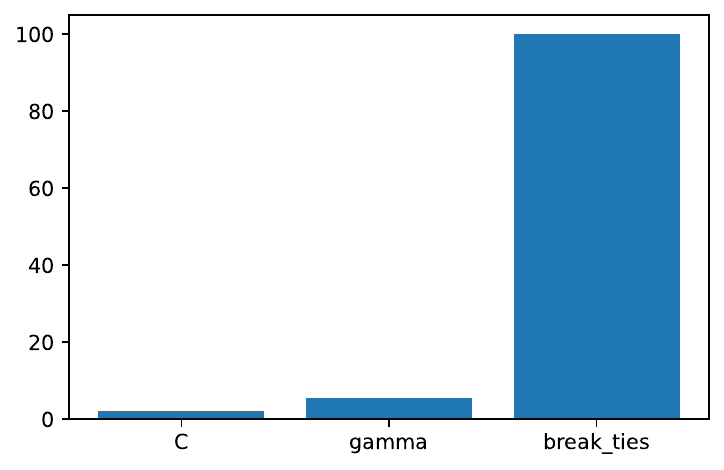}

}

\caption{Variable importance plot, threshold 0.025.}

\end{figure}

\hypertarget{get-default-hyperparameters-5}{%
\subsection{Get Default
Hyperparameters}\label{get-default-hyperparameters-5}}

\begin{Shaded}
\begin{Highlighting}[]
\ImportTok{from}\NormalTok{ spotPython.hyperparameters.values }\ImportTok{import}\NormalTok{ get\_default\_values, transform\_hyper\_parameter\_values}
\NormalTok{values\_default }\OperatorTok{=}\NormalTok{ get\_default\_values(fun\_control)}
\NormalTok{values\_default }\OperatorTok{=}\NormalTok{ transform\_hyper\_parameter\_values(fun\_control}\OperatorTok{=}\NormalTok{fun\_control, hyper\_parameter\_values}\OperatorTok{=}\NormalTok{values\_default)}
\NormalTok{values\_default}
\end{Highlighting}
\end{Shaded}

\begin{verbatim}
{'C': 1.0,
 'kernel': 'rbf',
 'degree': 3,
 'gamma': 'scale',
 'coef0': 0.0,
 'shrinking': 0,
 'probability': 0,
 'tol': 0.001,
 'cache_size': 200.0,
 'break_ties': 0}
\end{verbatim}

\begin{Shaded}
\begin{Highlighting}[]
\ImportTok{from}\NormalTok{ sklearn.pipeline }\ImportTok{import}\NormalTok{ make\_pipeline}
\NormalTok{model\_default }\OperatorTok{=}\NormalTok{ make\_pipeline(fun\_control[}\StringTok{"prep\_model"}\NormalTok{], fun\_control[}\StringTok{"core\_model"}\NormalTok{](}\OperatorTok{**}\NormalTok{values\_default))}
\NormalTok{model\_default}
\end{Highlighting}
\end{Shaded}

\begin{verbatim}
Pipeline(steps=[('nonetype', None),
                ('svc',
                 SVC(break_ties=0, cache_size=200.0, probability=0,
                     shrinking=0))])
\end{verbatim}

\begin{tcolorbox}[enhanced jigsaw, left=2mm, title=\textcolor{quarto-callout-note-color}{\faInfo}\hspace{0.5em}{Note}, bottomrule=.15mm, titlerule=0mm, breakable, rightrule=.15mm, toprule=.15mm, coltitle=black, colbacktitle=quarto-callout-note-color!10!white, leftrule=.75mm, arc=.35mm, colframe=quarto-callout-note-color-frame, bottomtitle=1mm, colback=white, opacitybacktitle=0.6, toptitle=1mm, opacityback=0]

\begin{itemize}
\tightlist
\item
  Default value for ``probability'' is False, but we need it to be True
  for the metric ``mapk\_score''.
\end{itemize}

\begin{Shaded}
\begin{Highlighting}[]
\NormalTok{values\_default.update(\{}\StringTok{"probability"}\NormalTok{: }\DecValTok{1}\NormalTok{\})}
\end{Highlighting}
\end{Shaded}

\end{tcolorbox}

\hypertarget{get-spot-results-4}{%
\subsection{Get SPOT Results}\label{get-spot-results-4}}

\begin{Shaded}
\begin{Highlighting}[]
\NormalTok{X }\OperatorTok{=}\NormalTok{ spot\_tuner.to\_all\_dim(spot\_tuner.min\_X.reshape(}\DecValTok{1}\NormalTok{,}\OperatorTok{{-}}\DecValTok{1}\NormalTok{))}
\BuiltInTok{print}\NormalTok{(X)}
\end{Highlighting}
\end{Shaded}

\begin{verbatim}
[[4.80995794e+00 0.00000000e+00 3.00000000e+00 0.00000000e+00
  0.00000000e+00 0.00000000e+00 1.00000000e+00 3.96929821e-03
  1.45936580e+02 0.00000000e+00]]
\end{verbatim}

\begin{Shaded}
\begin{Highlighting}[]
\ImportTok{from}\NormalTok{ spotPython.hyperparameters.values }\ImportTok{import}\NormalTok{ assign\_values, return\_conf\_list\_from\_var\_dict}
\NormalTok{v\_dict }\OperatorTok{=}\NormalTok{ assign\_values(X, fun\_control[}\StringTok{"var\_name"}\NormalTok{])}
\NormalTok{return\_conf\_list\_from\_var\_dict(var\_dict}\OperatorTok{=}\NormalTok{v\_dict, fun\_control}\OperatorTok{=}\NormalTok{fun\_control)}
\end{Highlighting}
\end{Shaded}

\begin{verbatim}
[{'C': 4.809957939164208,
  'kernel': 'rbf',
  'degree': 3,
  'gamma': 'scale',
  'coef0': 0.0,
  'shrinking': 0,
  'probability': 1,
  'tol': 0.003969298209225212,
  'cache_size': 145.9365804877652,
  'break_ties': 0}]
\end{verbatim}

\begin{Shaded}
\begin{Highlighting}[]
\ImportTok{from}\NormalTok{ spotPython.hyperparameters.values }\ImportTok{import}\NormalTok{ get\_one\_sklearn\_model\_from\_X}
\NormalTok{model\_spot }\OperatorTok{=}\NormalTok{ get\_one\_sklearn\_model\_from\_X(X, fun\_control)}
\NormalTok{model\_spot}
\end{Highlighting}
\end{Shaded}

\begin{verbatim}
SVC(C=4.809957939164208, break_ties=0, cache_size=145.9365804877652,
    probability=1, shrinking=0, tol=0.003969298209225212)
\end{verbatim}

\hypertarget{evaluate-spot-results-2}{%
\subsection{Evaluate SPOT Results}\label{evaluate-spot-results-2}}

\begin{itemize}
\tightlist
\item
  Fetch the data.
\end{itemize}

\begin{Shaded}
\begin{Highlighting}[]
\ImportTok{from}\NormalTok{ spotPython.utils.convert }\ImportTok{import}\NormalTok{ get\_Xy\_from\_df}
\NormalTok{X\_train, y\_train }\OperatorTok{=}\NormalTok{ get\_Xy\_from\_df(fun\_control[}\StringTok{"train"}\NormalTok{], fun\_control[}\StringTok{"target\_column"}\NormalTok{])}
\NormalTok{X\_test, y\_test }\OperatorTok{=}\NormalTok{ get\_Xy\_from\_df(fun\_control[}\StringTok{"test"}\NormalTok{], fun\_control[}\StringTok{"target\_column"}\NormalTok{])}
\NormalTok{X\_test.shape, y\_test.shape}
\end{Highlighting}
\end{Shaded}

\begin{verbatim}
((63, 64), (63,))
\end{verbatim}

\begin{itemize}
\tightlist
\item
  Fit the model with the tuned hyperparameters. This gives one result:
\end{itemize}

\begin{Shaded}
\begin{Highlighting}[]
\NormalTok{model\_spot.fit(X\_train, y\_train)}
\NormalTok{y\_pred }\OperatorTok{=}\NormalTok{ model\_spot.predict\_proba(X\_test)}
\NormalTok{res }\OperatorTok{=}\NormalTok{ mapk\_score(y\_true}\OperatorTok{=}\NormalTok{y\_test, y\_pred}\OperatorTok{=}\NormalTok{y\_pred, k}\OperatorTok{=}\DecValTok{3}\NormalTok{)}
\NormalTok{res}
\end{Highlighting}
\end{Shaded}

\begin{verbatim}
0.8571428571428571
\end{verbatim}

\begin{Shaded}
\begin{Highlighting}[]
\KeywordTok{def}\NormalTok{ repeated\_eval(n, model):}
\NormalTok{    res\_values }\OperatorTok{=}\NormalTok{ []}
    \ControlFlowTok{for}\NormalTok{ i }\KeywordTok{in} \BuiltInTok{range}\NormalTok{(n):}
\NormalTok{        model.fit(X\_train, y\_train)}
\NormalTok{        y\_pred }\OperatorTok{=}\NormalTok{ model.predict\_proba(X\_test)}
\NormalTok{        res }\OperatorTok{=}\NormalTok{ mapk\_score(y\_true}\OperatorTok{=}\NormalTok{y\_test, y\_pred}\OperatorTok{=}\NormalTok{y\_pred, k}\OperatorTok{=}\DecValTok{3}\NormalTok{)}
\NormalTok{        res\_values.append(res)}
\NormalTok{    mean\_res }\OperatorTok{=}\NormalTok{ np.mean(res\_values)}
    \BuiltInTok{print}\NormalTok{(}\SpecialStringTok{f"mean\_res: }\SpecialCharTok{\{}\NormalTok{mean\_res}\SpecialCharTok{\}}\SpecialStringTok{"}\NormalTok{)}
\NormalTok{    std\_res }\OperatorTok{=}\NormalTok{ np.std(res\_values)}
    \BuiltInTok{print}\NormalTok{(}\SpecialStringTok{f"std\_res: }\SpecialCharTok{\{}\NormalTok{std\_res}\SpecialCharTok{\}}\SpecialStringTok{"}\NormalTok{)}
\NormalTok{    min\_res }\OperatorTok{=}\NormalTok{ np.}\BuiltInTok{min}\NormalTok{(res\_values)}
    \BuiltInTok{print}\NormalTok{(}\SpecialStringTok{f"min\_res: }\SpecialCharTok{\{}\NormalTok{min\_res}\SpecialCharTok{\}}\SpecialStringTok{"}\NormalTok{)}
\NormalTok{    max\_res }\OperatorTok{=}\NormalTok{ np.}\BuiltInTok{max}\NormalTok{(res\_values)}
    \BuiltInTok{print}\NormalTok{(}\SpecialStringTok{f"max\_res: }\SpecialCharTok{\{}\NormalTok{max\_res}\SpecialCharTok{\}}\SpecialStringTok{"}\NormalTok{)}
\NormalTok{    median\_res }\OperatorTok{=}\NormalTok{ np.median(res\_values)}
    \BuiltInTok{print}\NormalTok{(}\SpecialStringTok{f"median\_res: }\SpecialCharTok{\{}\NormalTok{median\_res}\SpecialCharTok{\}}\SpecialStringTok{"}\NormalTok{)}
    \ControlFlowTok{return}\NormalTok{ mean\_res, std\_res, min\_res, max\_res, median\_res}
\end{Highlighting}
\end{Shaded}

\hypertarget{handling-non-deterministic-results-2}{%
\subsection{Handling Non-deterministic
Results}\label{handling-non-deterministic-results-2}}

\begin{itemize}
\tightlist
\item
  Because the model is non-determinstic, we perform \(n=30\) runs and
  calculate the mean and standard deviation of the performance metric.
\end{itemize}

\begin{Shaded}
\begin{Highlighting}[]
\NormalTok{\_ }\OperatorTok{=}\NormalTok{ repeated\_eval(}\DecValTok{30}\NormalTok{, model\_spot)}
\end{Highlighting}
\end{Shaded}

\begin{verbatim}
mean_res: 0.862522045855379
std_res: 0.003580941887874279
min_res: 0.8571428571428571
max_res: 0.8650793650793651
median_res: 0.8650793650793651
\end{verbatim}

\hypertarget{evalution-of-the-default-hyperparameters-2}{%
\subsection{Evalution of the Default
Hyperparameters}\label{evalution-of-the-default-hyperparameters-2}}

\begin{Shaded}
\begin{Highlighting}[]
\NormalTok{model\_default[}\StringTok{"svc"}\NormalTok{].probability }\OperatorTok{=} \VariableTok{True}
\NormalTok{model\_default.fit(X\_train, y\_train)[}\StringTok{"svc"}\NormalTok{]}
\end{Highlighting}
\end{Shaded}

\begin{verbatim}
SVC(break_ties=0, cache_size=200.0, probability=True, shrinking=0)
\end{verbatim}

\begin{itemize}
\tightlist
\item
  One evaluation of the default hyperparameters is performed on the
  hold-out test set.
\end{itemize}

\begin{Shaded}
\begin{Highlighting}[]
\NormalTok{y\_pred }\OperatorTok{=}\NormalTok{ model\_default.predict\_proba(X\_test)}
\NormalTok{mapk\_score(y\_true}\OperatorTok{=}\NormalTok{y\_test, y\_pred}\OperatorTok{=}\NormalTok{y\_pred, k}\OperatorTok{=}\DecValTok{3}\NormalTok{)}
\end{Highlighting}
\end{Shaded}

\begin{verbatim}
0.8571428571428571
\end{verbatim}

Since one single evaluation is not meaningful, we perform, similar to
the evaluation of the SPOT results, \(n=30\) runs of the default setting
and and calculate the mean and standard deviation of the performance
metric.

\begin{Shaded}
\begin{Highlighting}[]
\NormalTok{\_ }\OperatorTok{=}\NormalTok{ repeated\_eval(}\DecValTok{30}\NormalTok{, model\_default)}
\end{Highlighting}
\end{Shaded}

\begin{verbatim}
mean_res: 0.8545855379188712
std_res: 0.0041258157196788605
min_res: 0.8492063492063492
max_res: 0.8650793650793651
median_res: 0.8571428571428571
\end{verbatim}

\hypertarget{plot-compare-predictions-3}{%
\subsection{Plot: Compare
Predictions}\label{plot-compare-predictions-3}}

\begin{Shaded}
\begin{Highlighting}[]
\ImportTok{from}\NormalTok{ spotPython.plot.validation }\ImportTok{import}\NormalTok{ plot\_confusion\_matrix}
\NormalTok{plot\_confusion\_matrix(model\_default, fun\_control, title }\OperatorTok{=} \StringTok{"Default"}\NormalTok{)}
\end{Highlighting}
\end{Shaded}

\begin{figure}[H]

{\centering \includegraphics{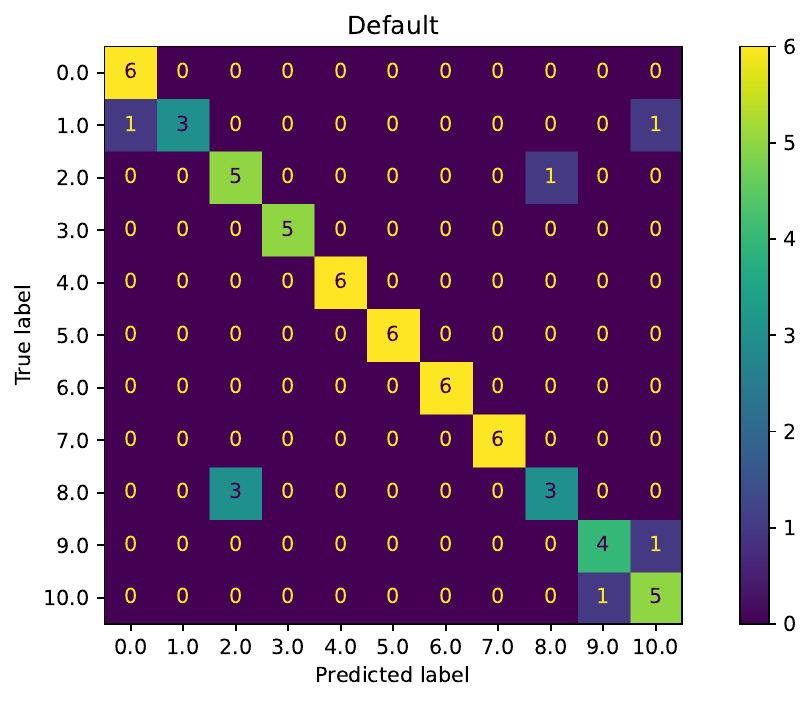}

}

\end{figure}

\begin{Shaded}
\begin{Highlighting}[]
\NormalTok{plot\_confusion\_matrix(model\_spot, fun\_control, title}\OperatorTok{=}\StringTok{"SPOT"}\NormalTok{)}
\end{Highlighting}
\end{Shaded}

\begin{figure}[H]

{\centering \includegraphics{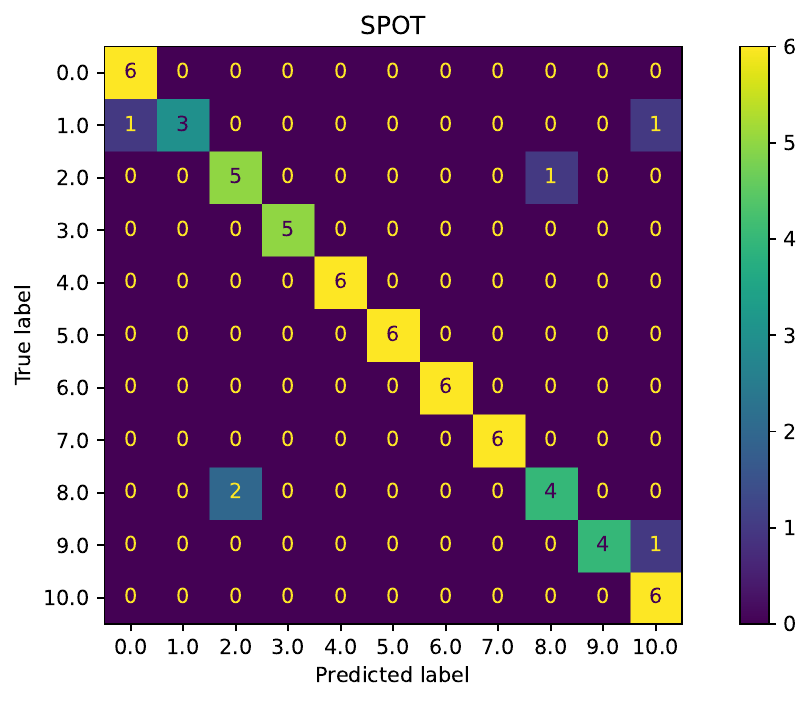}

}

\end{figure}

\begin{Shaded}
\begin{Highlighting}[]
\BuiltInTok{min}\NormalTok{(spot\_tuner.y), }\BuiltInTok{max}\NormalTok{(spot\_tuner.y)}
\end{Highlighting}
\end{Shaded}

\begin{verbatim}
(-0.8854166666666666, -0.041666666666666664)
\end{verbatim}

\hypertarget{cross-validated-evaluations-2}{%
\subsection{Cross-validated
Evaluations}\label{cross-validated-evaluations-2}}

\begin{Shaded}
\begin{Highlighting}[]
\ImportTok{from}\NormalTok{ spotPython.sklearn.traintest }\ImportTok{import}\NormalTok{ evaluate\_cv}
\NormalTok{fun\_control.update(\{}
     \StringTok{"eval"}\NormalTok{: }\StringTok{"train\_cv"}\NormalTok{,}
     \StringTok{"k\_folds"}\NormalTok{: }\DecValTok{10}\NormalTok{,}
\NormalTok{\})}
\NormalTok{evaluate\_cv(model}\OperatorTok{=}\NormalTok{model\_spot, fun\_control}\OperatorTok{=}\NormalTok{fun\_control, verbose}\OperatorTok{=}\DecValTok{0}\NormalTok{)}
\end{Highlighting}
\end{Shaded}

\begin{verbatim}
(0.8671539961013645, None)
\end{verbatim}

\begin{Shaded}
\begin{Highlighting}[]
\NormalTok{fun\_control.update(\{}
     \StringTok{"eval"}\NormalTok{: }\StringTok{"test\_cv"}\NormalTok{,}
     \StringTok{"k\_folds"}\NormalTok{: }\DecValTok{10}\NormalTok{,}
\NormalTok{\})}
\NormalTok{evaluate\_cv(model}\OperatorTok{=}\NormalTok{model\_spot, fun\_control}\OperatorTok{=}\NormalTok{fun\_control, verbose}\OperatorTok{=}\DecValTok{0}\NormalTok{)}
\end{Highlighting}
\end{Shaded}

\begin{verbatim}
Error in fun_sklearn(). Call to evaluate_cv failed. err=ValueError('n_splits=10 cannot be greater than the number of members in each class.'), type(err)=<class 'ValueError'>
\end{verbatim}

\begin{verbatim}
(nan, None)
\end{verbatim}

\begin{itemize}
\tightlist
\item
  This is the evaluation that will be used in the comparison:
\end{itemize}

\begin{Shaded}
\begin{Highlighting}[]
\NormalTok{fun\_control.update(\{}
     \StringTok{"eval"}\NormalTok{: }\StringTok{"data\_cv"}\NormalTok{,}
     \StringTok{"k\_folds"}\NormalTok{: }\DecValTok{10}\NormalTok{,}
\NormalTok{\})}
\NormalTok{evaluate\_cv(model}\OperatorTok{=}\NormalTok{model\_spot, fun\_control}\OperatorTok{=}\NormalTok{fun\_control, verbose}\OperatorTok{=}\DecValTok{0}\NormalTok{)}
\end{Highlighting}
\end{Shaded}

\begin{verbatim}
(0.882, None)
\end{verbatim}

\hypertarget{detailed-hyperparameter-plots-5}{%
\subsection{Detailed Hyperparameter
Plots}\label{detailed-hyperparameter-plots-5}}

\begin{Shaded}
\begin{Highlighting}[]
\NormalTok{filename }\OperatorTok{=} \StringTok{"./figures/"} \OperatorTok{+}\NormalTok{ experiment\_name}
\NormalTok{spot\_tuner.plot\_important\_hyperparameter\_contour(filename}\OperatorTok{=}\NormalTok{filename)}
\end{Highlighting}
\end{Shaded}

\begin{verbatim}
C:  2.085932206795447
gamma:  5.522956421657414
break_ties:  100.00000000000001
\end{verbatim}

\begin{figure}[H]

{\centering \includegraphics{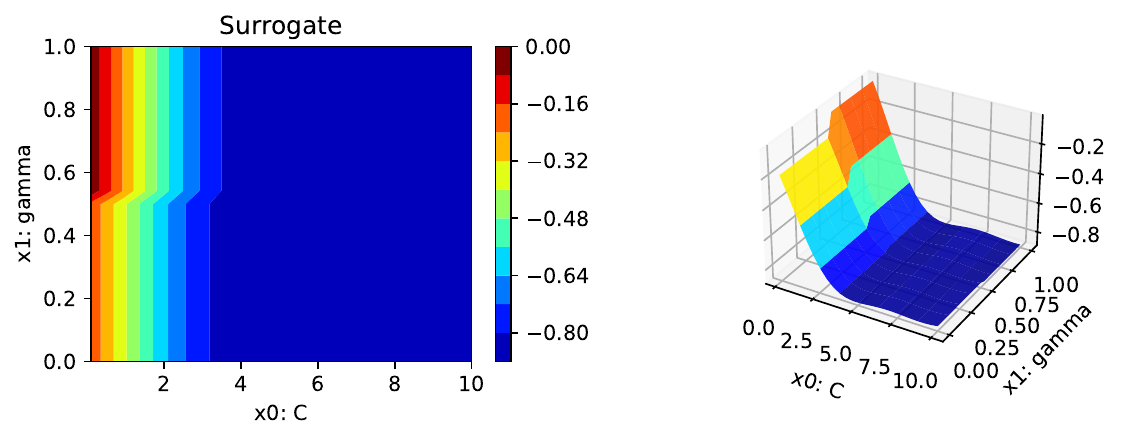}

}

\end{figure}

\begin{figure}[H]

{\centering \includegraphics{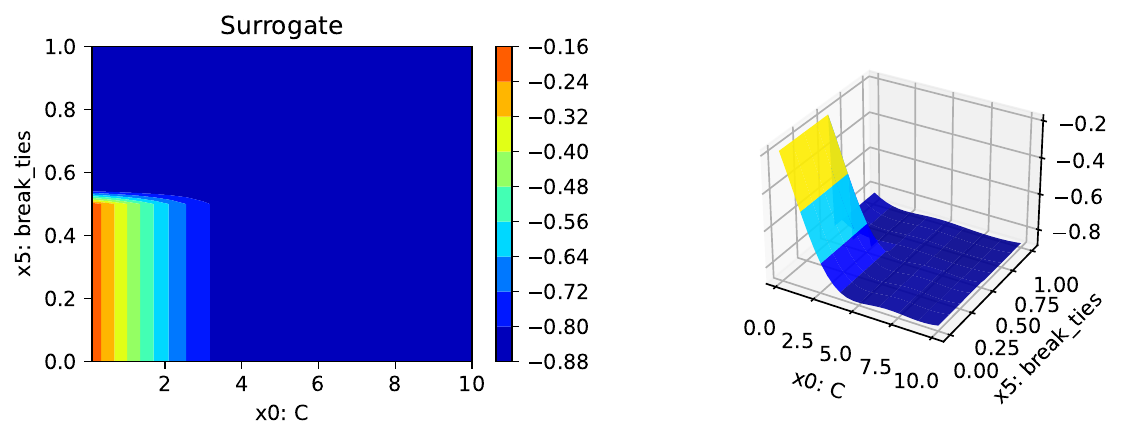}

}

\end{figure}

\begin{figure}[H]

{\centering \includegraphics{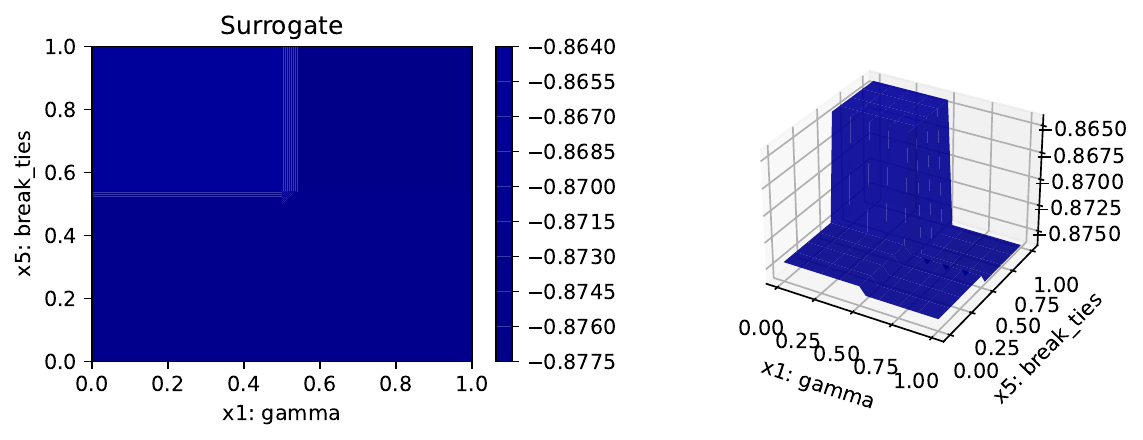}

}

\end{figure}

\hypertarget{parallel-coordinates-plot-3}{%
\subsection{Parallel Coordinates
Plot}\label{parallel-coordinates-plot-3}}

\begin{Shaded}
\begin{Highlighting}[]
\NormalTok{spot\_tuner.parallel\_plot()}
\end{Highlighting}
\end{Shaded}

\begin{verbatim}
Unable to display output for mime type(s): text/html
\end{verbatim}

\begin{verbatim}
Unable to display output for mime type(s): text/html
\end{verbatim}

\hypertarget{plot-all-combinations-of-hyperparameters-4}{%
\subsection{Plot all Combinations of
Hyperparameters}\label{plot-all-combinations-of-hyperparameters-4}}

\begin{itemize}
\tightlist
\item
  Warning: this may take a while.
\end{itemize}

\begin{Shaded}
\begin{Highlighting}[]
\NormalTok{PLOT\_ALL }\OperatorTok{=} \VariableTok{False}
\ControlFlowTok{if}\NormalTok{ PLOT\_ALL:}
\NormalTok{    n }\OperatorTok{=}\NormalTok{ spot\_tuner.k}
    \ControlFlowTok{for}\NormalTok{ i }\KeywordTok{in} \BuiltInTok{range}\NormalTok{(n}\OperatorTok{{-}}\DecValTok{1}\NormalTok{):}
        \ControlFlowTok{for}\NormalTok{ j }\KeywordTok{in} \BuiltInTok{range}\NormalTok{(i}\OperatorTok{+}\DecValTok{1}\NormalTok{, n):}
\NormalTok{            spot\_tuner.plot\_contour(i}\OperatorTok{=}\NormalTok{i, j}\OperatorTok{=}\NormalTok{j, min\_z}\OperatorTok{=}\NormalTok{min\_z, max\_z }\OperatorTok{=}\NormalTok{ max\_z)}
\end{Highlighting}
\end{Shaded}

\hypertarget{sec-hpt-sklearn-knn-classifier-vbdp-data}{%
\chapter{HPT: sklearn KNN Classifier VBDP
Data}\label{sec-hpt-sklearn-knn-classifier-vbdp-data}}

This chapter describes the hyperparameter tuning of a
\texttt{KNeighborsClassifier} on the Vector Borne Disease Prediction
(VBDP) data set.

\begin{tcolorbox}[enhanced jigsaw, left=2mm, title=\textcolor{quarto-callout-important-color}{\faExclamation}\hspace{0.5em}{Vector Borne Disease Prediction Data Set}, bottomrule=.15mm, titlerule=0mm, breakable, rightrule=.15mm, toprule=.15mm, coltitle=black, colbacktitle=quarto-callout-important-color!10!white, leftrule=.75mm, arc=.35mm, colframe=quarto-callout-important-color-frame, bottomtitle=1mm, colback=white, opacitybacktitle=0.6, toptitle=1mm, opacityback=0]

This chapter uses the Vector Borne Disease Prediction data set from
Kaggle. It is a categorical dataset for eleven Vector Borne Diseases
with associated symptoms.

\begin{quote}
The person who associated a work with this deed has dedicated the work
to the public domain by waiving all of his or her rights to the work
worldwide under copyright law, including all related and neighboring
rights, to the extent allowed by law.You can copy, modify, distribute
and perform the work, even for commercial purposes, all without asking
permission. See Other Information below, see
\url{https://creativecommons.org/publicdomain/zero/1.0/}.
\end{quote}

The data set is available at:
\url{https://www.kaggle.com/datasets/richardbernat/vector-borne-disease-prediction},

The data should be downloaded and stored in the \texttt{data/VBDP}
subfolder. The data set is not available as a part of the
\texttt{spotPython} package.

\end{tcolorbox}

\hypertarget{sec-setup-19}{%
\section{Step 1: Setup}\label{sec-setup-19}}

Before we consider the detailed experimental setup, we select the
parameters that affect run time and the initial design size.

\begin{Shaded}
\begin{Highlighting}[]
\NormalTok{MAX\_TIME }\OperatorTok{=} \DecValTok{1}
\NormalTok{INIT\_SIZE }\OperatorTok{=} \DecValTok{5}
\NormalTok{ORIGINAL }\OperatorTok{=} \VariableTok{True}
\NormalTok{PREFIX }\OperatorTok{=} \StringTok{"19"}
\end{Highlighting}
\end{Shaded}

\begin{Shaded}
\begin{Highlighting}[]
\ImportTok{import}\NormalTok{ warnings}
\NormalTok{warnings.filterwarnings(}\StringTok{"ignore"}\NormalTok{)}
\end{Highlighting}
\end{Shaded}

\hypertarget{step-2-initialization-of-the-empty-fun_control-dictionary-4}{%
\section{\texorpdfstring{Step 2: Initialization of the Empty
\texttt{fun\_control}
Dictionary}{Step 2: Initialization of the Empty fun\_control Dictionary}}\label{step-2-initialization-of-the-empty-fun_control-dictionary-4}}

\begin{Shaded}
\begin{Highlighting}[]
\ImportTok{from}\NormalTok{ spotPython.utils.init }\ImportTok{import}\NormalTok{ fun\_control\_init}
\ImportTok{from}\NormalTok{ spotPython.utils.}\BuiltInTok{file} \ImportTok{import}\NormalTok{ get\_experiment\_name, get\_spot\_tensorboard\_path}
\ImportTok{from}\NormalTok{ spotPython.utils.device }\ImportTok{import}\NormalTok{ getDevice}

\NormalTok{experiment\_name }\OperatorTok{=}\NormalTok{ get\_experiment\_name(prefix}\OperatorTok{=}\NormalTok{PREFIX)}

\NormalTok{fun\_control }\OperatorTok{=}\NormalTok{ fun\_control\_init(}
\NormalTok{    task}\OperatorTok{=}\StringTok{"classification"}\NormalTok{,}
\NormalTok{    spot\_tensorboard\_path}\OperatorTok{=}\NormalTok{get\_spot\_tensorboard\_path(experiment\_name))}
\end{Highlighting}
\end{Shaded}

\hypertarget{load-data-classification-vbdp-3}{%
\subsection{Load Data: Classification
VBDP}\label{load-data-classification-vbdp-3}}

\begin{Shaded}
\begin{Highlighting}[]
\ImportTok{import}\NormalTok{ pandas }\ImportTok{as}\NormalTok{ pd}
\ControlFlowTok{if}\NormalTok{ ORIGINAL }\OperatorTok{==} \VariableTok{True}\NormalTok{:}
\NormalTok{    train\_df }\OperatorTok{=}\NormalTok{ pd.read\_csv(}\StringTok{\textquotesingle{}./data/VBDP/trainn.csv\textquotesingle{}}\NormalTok{)}
\NormalTok{    test\_df }\OperatorTok{=}\NormalTok{ pd.read\_csv(}\StringTok{\textquotesingle{}./data/VBDP/testt.csv\textquotesingle{}}\NormalTok{)}
\ControlFlowTok{else}\NormalTok{:}
\NormalTok{    train\_df }\OperatorTok{=}\NormalTok{ pd.read\_csv(}\StringTok{\textquotesingle{}./data/VBDP/train.csv\textquotesingle{}}\NormalTok{)}
    \CommentTok{\# remove the id column}
\NormalTok{    train\_df }\OperatorTok{=}\NormalTok{ train\_df.drop(columns}\OperatorTok{=}\NormalTok{[}\StringTok{\textquotesingle{}id\textquotesingle{}}\NormalTok{])}
\end{Highlighting}
\end{Shaded}

\begin{Shaded}
\begin{Highlighting}[]
\ImportTok{from}\NormalTok{ sklearn.preprocessing }\ImportTok{import}\NormalTok{ OrdinalEncoder}
\NormalTok{n\_samples }\OperatorTok{=}\NormalTok{ train\_df.shape[}\DecValTok{0}\NormalTok{]}
\NormalTok{n\_features }\OperatorTok{=}\NormalTok{ train\_df.shape[}\DecValTok{1}\NormalTok{] }\OperatorTok{{-}} \DecValTok{1}
\NormalTok{target\_column }\OperatorTok{=} \StringTok{"prognosis"}
\CommentTok{\# Encoder our prognosis labels as integers for easier decoding later}
\NormalTok{enc }\OperatorTok{=}\NormalTok{ OrdinalEncoder()}
\NormalTok{train\_df[target\_column] }\OperatorTok{=}\NormalTok{ enc.fit\_transform(train\_df[[target\_column]])}
\NormalTok{train\_df.columns }\OperatorTok{=}\NormalTok{ [}\SpecialStringTok{f"x}\SpecialCharTok{\{}\NormalTok{i}\SpecialCharTok{\}}\SpecialStringTok{"} \ControlFlowTok{for}\NormalTok{ i }\KeywordTok{in} \BuiltInTok{range}\NormalTok{(}\DecValTok{1}\NormalTok{, n\_features}\OperatorTok{+}\DecValTok{1}\NormalTok{)] }\OperatorTok{+}\NormalTok{ [target\_column]}
\BuiltInTok{print}\NormalTok{(train\_df.shape)}
\NormalTok{train\_df.head()}
\end{Highlighting}
\end{Shaded}

\begin{verbatim}
(252, 65)
\end{verbatim}

\begin{longtable}[]{@{}llllllllllllllllllllll@{}}
\toprule\noalign{}
& x1 & x2 & x3 & x4 & x5 & x6 & x7 & x8 & x9 & x10 & ... & x56 & x57 &
x58 & x59 & x60 & x61 & x62 & x63 & x64 & prognosis \\
\midrule\noalign{}
\endhead
\bottomrule\noalign{}
\endlastfoot
0 & 0 & 1 & 1 & 1 & 1 & 0 & 1 & 0 & 0 & 0 & ... & 0 & 0 & 0 & 0 & 0 & 0
& 0 & 0 & 0 & 0.0 \\
1 & 1 & 1 & 1 & 1 & 1 & 0 & 1 & 1 & 1 & 0 & ... & 0 & 0 & 0 & 0 & 0 & 0
& 0 & 0 & 0 & 0.0 \\
2 & 0 & 1 & 0 & 1 & 0 & 0 & 1 & 1 & 0 & 0 & ... & 0 & 0 & 0 & 0 & 0 & 0
& 0 & 0 & 0 & 0.0 \\
3 & 0 & 0 & 0 & 0 & 0 & 1 & 1 & 1 & 0 & 0 & ... & 0 & 0 & 0 & 0 & 0 & 0
& 0 & 0 & 0 & 0.0 \\
4 & 1 & 0 & 0 & 0 & 1 & 1 & 1 & 1 & 0 & 0 & ... & 0 & 0 & 0 & 0 & 0 & 0
& 0 & 0 & 0 & 0.0 \\
\end{longtable}

The full data set \texttt{train\_df} 64 features. The target column is
labeled as \texttt{prognosis}.

\hypertarget{holdout-train-and-test-data-3}{%
\subsection{Holdout Train and Test
Data}\label{holdout-train-and-test-data-3}}

We split out a hold-out test set (25\% of the data) so we can calculate
an example MAP@K

\begin{Shaded}
\begin{Highlighting}[]
\ImportTok{import}\NormalTok{ numpy }\ImportTok{as}\NormalTok{ np}
\ImportTok{from}\NormalTok{ sklearn.model\_selection }\ImportTok{import}\NormalTok{ train\_test\_split}
\NormalTok{X\_train, X\_test, y\_train, y\_test }\OperatorTok{=}\NormalTok{ train\_test\_split(train\_df.drop(target\_column, axis}\OperatorTok{=}\DecValTok{1}\NormalTok{), train\_df[target\_column],}
\NormalTok{                                                    random\_state}\OperatorTok{=}\DecValTok{42}\NormalTok{,}
\NormalTok{                                                    test\_size}\OperatorTok{=}\FloatTok{0.25}\NormalTok{,}
\NormalTok{                                                    stratify}\OperatorTok{=}\NormalTok{train\_df[target\_column])}
\NormalTok{train }\OperatorTok{=}\NormalTok{ pd.DataFrame(np.hstack((X\_train, np.array(y\_train).reshape(}\OperatorTok{{-}}\DecValTok{1}\NormalTok{, }\DecValTok{1}\NormalTok{))))}
\NormalTok{test }\OperatorTok{=}\NormalTok{ pd.DataFrame(np.hstack((X\_test, np.array(y\_test).reshape(}\OperatorTok{{-}}\DecValTok{1}\NormalTok{, }\DecValTok{1}\NormalTok{))))}
\NormalTok{train.columns }\OperatorTok{=}\NormalTok{ [}\SpecialStringTok{f"x}\SpecialCharTok{\{}\NormalTok{i}\SpecialCharTok{\}}\SpecialStringTok{"} \ControlFlowTok{for}\NormalTok{ i }\KeywordTok{in} \BuiltInTok{range}\NormalTok{(}\DecValTok{1}\NormalTok{, n\_features}\OperatorTok{+}\DecValTok{1}\NormalTok{)] }\OperatorTok{+}\NormalTok{ [target\_column]}
\NormalTok{test.columns }\OperatorTok{=}\NormalTok{ [}\SpecialStringTok{f"x}\SpecialCharTok{\{}\NormalTok{i}\SpecialCharTok{\}}\SpecialStringTok{"} \ControlFlowTok{for}\NormalTok{ i }\KeywordTok{in} \BuiltInTok{range}\NormalTok{(}\DecValTok{1}\NormalTok{, n\_features}\OperatorTok{+}\DecValTok{1}\NormalTok{)] }\OperatorTok{+}\NormalTok{ [target\_column]}
\BuiltInTok{print}\NormalTok{(train.shape)}
\BuiltInTok{print}\NormalTok{(test.shape)}
\NormalTok{train.head()}
\end{Highlighting}
\end{Shaded}

\begin{verbatim}
(189, 65)
(63, 65)
\end{verbatim}

\begin{longtable}[]{@{}llllllllllllllllllllll@{}}
\toprule\noalign{}
& x1 & x2 & x3 & x4 & x5 & x6 & x7 & x8 & x9 & x10 & ... & x56 & x57 &
x58 & x59 & x60 & x61 & x62 & x63 & x64 & prognosis \\
\midrule\noalign{}
\endhead
\bottomrule\noalign{}
\endlastfoot
0 & 1.0 & 0.0 & 0.0 & 1.0 & 0.0 & 1.0 & 0.0 & 0.0 & 0.0 & 1.0 & ... &
0.0 & 0.0 & 0.0 & 0.0 & 1.0 & 1.0 & 1.0 & 0.0 & 0.0 & 7.0 \\
1 & 1.0 & 0.0 & 1.0 & 1.0 & 1.0 & 1.0 & 1.0 & 0.0 & 1.0 & 1.0 & ... &
0.0 & 1.0 & 1.0 & 1.0 & 1.0 & 0.0 & 1.0 & 1.0 & 1.0 & 3.0 \\
2 & 0.0 & 0.0 & 1.0 & 0.0 & 1.0 & 0.0 & 0.0 & 0.0 & 0.0 & 0.0 & ... &
0.0 & 0.0 & 0.0 & 0.0 & 0.0 & 0.0 & 0.0 & 0.0 & 0.0 & 10.0 \\
3 & 1.0 & 1.0 & 1.0 & 1.0 & 1.0 & 1.0 & 0.0 & 0.0 & 1.0 & 1.0 & ... &
1.0 & 0.0 & 1.0 & 1.0 & 1.0 & 0.0 & 0.0 & 1.0 & 1.0 & 3.0 \\
4 & 1.0 & 1.0 & 1.0 & 0.0 & 1.0 & 1.0 & 0.0 & 1.0 & 1.0 & 0.0 & ... &
0.0 & 0.0 & 0.0 & 0.0 & 0.0 & 0.0 & 0.0 & 0.0 & 0.0 & 8.0 \\
\end{longtable}

\begin{Shaded}
\begin{Highlighting}[]
\CommentTok{\# add the dataset to the fun\_control}
\NormalTok{fun\_control.update(\{}\StringTok{"data"}\NormalTok{: train\_df, }\CommentTok{\# full dataset,}
               \StringTok{"train"}\NormalTok{: train,}
               \StringTok{"test"}\NormalTok{: test,}
               \StringTok{"n\_samples"}\NormalTok{: n\_samples,}
               \StringTok{"target\_column"}\NormalTok{: target\_column\})}
\end{Highlighting}
\end{Shaded}

\hypertarget{sec-specification-of-preprocessing-model-19}{%
\section{Step 4: Specification of the Preprocessing
Model}\label{sec-specification-of-preprocessing-model-19}}

Data preprocesssing can be very simple, e.g., you can ignore it. Then
you would choose the \texttt{prep\_model} ``None'':

\begin{Shaded}
\begin{Highlighting}[]
\NormalTok{prep\_model }\OperatorTok{=} \VariableTok{None}
\NormalTok{fun\_control.update(\{}\StringTok{"prep\_model"}\NormalTok{: prep\_model\})}
\end{Highlighting}
\end{Shaded}

A default approach for numerical data is the \texttt{StandardScaler}
(mean 0, variance 1). This can be selected as follows:

\begin{Shaded}
\begin{Highlighting}[]
\CommentTok{\# prep\_model = StandardScaler()}
\CommentTok{\# fun\_control.update(\{"prep\_model": prep\_model\})}
\end{Highlighting}
\end{Shaded}

Even more complicated pre-processing steps are possible, e.g., the
follwing pipeline:

\begin{Shaded}
\begin{Highlighting}[]
\CommentTok{\# categorical\_columns = []}
\CommentTok{\# one\_hot\_encoder = OneHotEncoder(handle\_unknown="ignore", sparse\_output=False)}
\CommentTok{\# prep\_model = ColumnTransformer(}
\CommentTok{\#         transformers=[}
\CommentTok{\#             ("categorical", one\_hot\_encoder, categorical\_columns),}
\CommentTok{\#         ],}
\CommentTok{\#         remainder=StandardScaler(),}
\CommentTok{\#     )}
\end{Highlighting}
\end{Shaded}

\hypertarget{step-5-select-model-algorithm-and-core_model_hyper_dict-4}{%
\section{\texorpdfstring{Step 5: Select Model (\texttt{algorithm}) and
\texttt{core\_model\_hyper\_dict}}{Step 5: Select Model (algorithm) and core\_model\_hyper\_dict}}\label{step-5-select-model-algorithm-and-core_model_hyper_dict-4}}

The selection of the algorithm (ML model) that should be tuned is done
by specifying the its name from the \texttt{sklearn} implementation. For
example, the \texttt{SVC} support vector machine classifier is selected
as follows:

\texttt{add\_core\_model\_to\_fun\_control(SVC,\ fun\_control,\ SklearnHyperDict)}

Other core\_models are, e.g.,:

\begin{itemize}
\tightlist
\item
  RidgeCV
\item
  GradientBoostingRegressor
\item
  ElasticNet
\item
  RandomForestClassifier
\item
  LogisticRegression
\item
  KNeighborsClassifier
\item
  RandomForestClassifier
\item
  GradientBoostingClassifier
\item
  HistGradientBoostingClassifier
\end{itemize}

We will use the \texttt{RandomForestClassifier} classifier in this
example.

\begin{Shaded}
\begin{Highlighting}[]
\ImportTok{from}\NormalTok{ sklearn.linear\_model }\ImportTok{import}\NormalTok{ RidgeCV}
\ImportTok{from}\NormalTok{ sklearn.ensemble }\ImportTok{import}\NormalTok{ RandomForestClassifier}
\ImportTok{from}\NormalTok{ sklearn.svm }\ImportTok{import}\NormalTok{ SVC}
\ImportTok{from}\NormalTok{ sklearn.linear\_model }\ImportTok{import}\NormalTok{ LogisticRegression}
\ImportTok{from}\NormalTok{ sklearn.neighbors }\ImportTok{import}\NormalTok{ KNeighborsClassifier}
\ImportTok{from}\NormalTok{ sklearn.ensemble }\ImportTok{import}\NormalTok{ GradientBoostingClassifier}
\ImportTok{from}\NormalTok{ sklearn.ensemble }\ImportTok{import}\NormalTok{ GradientBoostingRegressor}
\ImportTok{from}\NormalTok{ sklearn.ensemble }\ImportTok{import}\NormalTok{ HistGradientBoostingClassifier}
\ImportTok{from}\NormalTok{ sklearn.linear\_model }\ImportTok{import}\NormalTok{ ElasticNet}
\ImportTok{from}\NormalTok{ spotPython.hyperparameters.values }\ImportTok{import}\NormalTok{ add\_core\_model\_to\_fun\_control}
\ImportTok{from}\NormalTok{ spotPython.data.sklearn\_hyper\_dict }\ImportTok{import}\NormalTok{ SklearnHyperDict}
\ImportTok{from}\NormalTok{ spotPython.fun.hypersklearn }\ImportTok{import}\NormalTok{ HyperSklearn}
\end{Highlighting}
\end{Shaded}

\begin{Shaded}
\begin{Highlighting}[]
\CommentTok{\# core\_model  = RidgeCV}
\CommentTok{\# core\_model = GradientBoostingRegressor}
\CommentTok{\# core\_model = ElasticNet}
\CommentTok{\# core\_model = RandomForestClassifier}
\NormalTok{core\_model }\OperatorTok{=}\NormalTok{ KNeighborsClassifier}
\CommentTok{\# core\_model = LogisticRegression}
\CommentTok{\# core\_model = KNeighborsClassifier}
\CommentTok{\# core\_model = GradientBoostingClassifier}
\CommentTok{\# core\_model = HistGradientBoostingClassifier}
\NormalTok{add\_core\_model\_to\_fun\_control(core\_model}\OperatorTok{=}\NormalTok{core\_model,}
\NormalTok{                              fun\_control}\OperatorTok{=}\NormalTok{fun\_control,}
\NormalTok{                              hyper\_dict}\OperatorTok{=}\NormalTok{SklearnHyperDict,}
\NormalTok{                              filename}\OperatorTok{=}\VariableTok{None}\NormalTok{)}
\end{Highlighting}
\end{Shaded}

Now \texttt{fun\_control} has the information from the JSON file. The
available hyperparameters are:

\begin{Shaded}
\begin{Highlighting}[]
\BuiltInTok{print}\NormalTok{(}\OperatorTok{*}\NormalTok{fun\_control[}\StringTok{"core\_model\_hyper\_dict"}\NormalTok{].keys(), sep}\OperatorTok{=}\StringTok{"}\CharTok{\textbackslash{}n}\StringTok{"}\NormalTok{)}
\end{Highlighting}
\end{Shaded}

\begin{verbatim}
n_neighbors
weights
algorithm
leaf_size
p
\end{verbatim}

\hypertarget{step-6-modify-hyper_dict-hyperparameters-for-the-selected-algorithm-aka-core_model-4}{%
\section{\texorpdfstring{Step 6: Modify \texttt{hyper\_dict}
Hyperparameters for the Selected Algorithm aka
\texttt{core\_model}}{Step 6: Modify hyper\_dict Hyperparameters for the Selected Algorithm aka core\_model}}\label{step-6-modify-hyper_dict-hyperparameters-for-the-selected-algorithm-aka-core_model-4}}

\hypertarget{modify-hyperparameter-of-type-numeric-and-integer-boolean-4}{%
\subsection{Modify hyperparameter of type numeric and integer
(boolean)}\label{modify-hyperparameter-of-type-numeric-and-integer-boolean-4}}

Numeric and boolean values can be modified using the
\texttt{modify\_hyper\_parameter\_bounds} method. For example, to change
the \texttt{tol} hyperparameter of the \texttt{SVC} model to the
interval {[}1e-3, 1e-2{]}, the following code can be used:

\texttt{modify\_hyper\_parameter\_bounds(fun\_control,\ "tol",\ bounds={[}1e-3,\ 1e-2{]})}

\begin{Shaded}
\begin{Highlighting}[]
\CommentTok{\# from spotPython.hyperparameters.values import modify\_hyper\_parameter\_bounds}
\CommentTok{\# modify\_hyper\_parameter\_bounds(fun\_control, "probability", bounds=[1, 1])}
\end{Highlighting}
\end{Shaded}

\hypertarget{modify-hyperparameter-of-type-factor-5}{%
\subsection{Modify hyperparameter of type
factor}\label{modify-hyperparameter-of-type-factor-5}}

\texttt{spotPython} provides functions for modifying the
hyperparameters, their bounds and factors as well as for activating and
de-activating hyperparameters without re-compilation of the Python
source code. These functions were described in
Section~\ref{sec-modification-of-hyperparameters-14}.

Factors can be modified with the
\texttt{modify\_hyper\_parameter\_levels} function. For example, to
exclude the \texttt{sigmoid} kernel from the tuning, the \texttt{kernel}
hyperparameter of the \texttt{SVC} model can be modified as follows:

\texttt{modify\_hyper\_parameter\_levels(fun\_control,\ "kernel",\ {[}"linear",\ "rbf"{]})}

The new setting can be controlled via:

\texttt{fun\_control{[}"core\_model\_hyper\_dict"{]}{[}"kernel"{]}}

\begin{Shaded}
\begin{Highlighting}[]
\CommentTok{\# from spotPython.hyperparameters.values import modify\_hyper\_parameter\_levels}
\CommentTok{\# modify\_hyper\_parameter\_levels(fun\_control, "kernel", ["rbf"])}
\end{Highlighting}
\end{Shaded}

\hypertarget{sec-optimizers-19}{%
\subsection{Optimizers}\label{sec-optimizers-19}}

Optimizers are described in Section~\ref{sec-optimizers-14}.

\hypertarget{selection-of-the-objective-metric-and-loss-functions-3}{%
\subsection{Selection of the Objective: Metric and Loss
Functions}\label{selection-of-the-objective-metric-and-loss-functions-3}}

\begin{itemize}
\tightlist
\item
  Machine learning models are optimized with respect to a metric, for
  example, the \texttt{accuracy} function.
\item
  Deep learning, e.g., neural networks are optimized with respect to a
  loss function, for example, the \texttt{cross\_entropy} function and
  evaluated with respect to a metric, for example, the \texttt{accuracy}
  function.
\end{itemize}

\hypertarget{step-7-selection-of-the-objective-loss-function-5}{%
\section{Step 7: Selection of the Objective (Loss)
Function}\label{step-7-selection-of-the-objective-loss-function-5}}

The loss function, that is usually used in deep learning for optimizing
the weights of the net, is stored in the \texttt{fun\_control}
dictionary as \texttt{"loss\_function"}.

\hypertarget{metric-function-3}{%
\subsection{Metric Function}\label{metric-function-3}}

There are two different types of metrics in \texttt{spotPython}:

\begin{enumerate}
\def\labelenumi{\arabic{enumi}.}
\tightlist
\item
  \texttt{"metric\_river"} is used for the river based evaluation via
  \texttt{eval\_oml\_iter\_progressive}.
\item
  \texttt{"metric\_sklearn"} is used for the sklearn based evaluation.
\end{enumerate}

We will consider multi-class classification metrics, e.g.,
\texttt{mapk\_score} and \texttt{top\_k\_accuracy\_score}.

\begin{tcolorbox}[enhanced jigsaw, left=2mm, title=\textcolor{quarto-callout-note-color}{\faInfo}\hspace{0.5em}{Predict Probabilities}, bottomrule=.15mm, titlerule=0mm, breakable, rightrule=.15mm, toprule=.15mm, coltitle=black, colbacktitle=quarto-callout-note-color!10!white, leftrule=.75mm, arc=.35mm, colframe=quarto-callout-note-color-frame, bottomtitle=1mm, colback=white, opacitybacktitle=0.6, toptitle=1mm, opacityback=0]

In this multi-class classification example the machine learning
algorithm should return the probabilities of the specific classes
(\texttt{"predict\_proba"}) instead of the predicted values.

\end{tcolorbox}

We set \texttt{"predict\_proba"} to \texttt{True} in the
\texttt{fun\_control} dictionary.

\hypertarget{the-mapk-metric-3}{%
\subsubsection{The MAPK Metric}\label{the-mapk-metric-3}}

To select the MAPK metric, the following two entries can be added to the
\texttt{fun\_control} dictionary:

\texttt{"metric\_sklearn":\ mapk\_score"}

\texttt{"metric\_params":\ \{"k":\ 3\}}.

\hypertarget{other-metrics-3}{%
\subsubsection{Other Metrics}\label{other-metrics-3}}

Alternatively, other metrics for multi-class classification can be used,
e.g.,: * top\_k\_accuracy\_score or * roc\_auc\_score

The metric \texttt{roc\_auc\_score} requires the parameter
\texttt{"multi\_class"}, e.g.,

\texttt{"multi\_class":\ "ovr"}.

This is set in the \texttt{fun\_control} dictionary.

\begin{tcolorbox}[enhanced jigsaw, left=2mm, title=\textcolor{quarto-callout-note-color}{\faInfo}\hspace{0.5em}{Weights}, bottomrule=.15mm, titlerule=0mm, breakable, rightrule=.15mm, toprule=.15mm, coltitle=black, colbacktitle=quarto-callout-note-color!10!white, leftrule=.75mm, arc=.35mm, colframe=quarto-callout-note-color-frame, bottomtitle=1mm, colback=white, opacitybacktitle=0.6, toptitle=1mm, opacityback=0]

\texttt{spotPython} performs a minimization, therefore, metrics that
should be maximized have to be multiplied by -1. This is done by setting
\texttt{"weights"} to \texttt{-1}.

\end{tcolorbox}

\begin{itemize}
\tightlist
\item
  The complete setup for the metric in our example is:
\end{itemize}

\begin{Shaded}
\begin{Highlighting}[]
\ImportTok{from}\NormalTok{ spotPython.utils.metrics }\ImportTok{import}\NormalTok{ mapk\_score}
\NormalTok{fun\_control.update(\{}
               \StringTok{"weights"}\NormalTok{: }\OperatorTok{{-}}\DecValTok{1}\NormalTok{,}
               \StringTok{"metric\_sklearn"}\NormalTok{: mapk\_score,}
               \StringTok{"predict\_proba"}\NormalTok{: }\VariableTok{True}\NormalTok{,}
               \StringTok{"metric\_params"}\NormalTok{: \{}\StringTok{"k"}\NormalTok{: }\DecValTok{3}\NormalTok{\},}
\NormalTok{               \})}
\end{Highlighting}
\end{Shaded}

\hypertarget{evaluation-on-hold-out-data-3}{%
\subsection{Evaluation on Hold-out
Data}\label{evaluation-on-hold-out-data-3}}

\begin{itemize}
\tightlist
\item
  The default method for computing the performance is
  \texttt{"eval\_holdout"}.
\item
  Alternatively, cross-validation can be used for every machine learning
  model.
\item
  Specifically for RandomForests, the OOB-score can be used.
\end{itemize}

\begin{Shaded}
\begin{Highlighting}[]
\NormalTok{fun\_control.update(\{}
    \StringTok{"eval"}\NormalTok{: }\StringTok{"train\_hold\_out"}\NormalTok{,}
\NormalTok{\})}
\end{Highlighting}
\end{Shaded}

\hypertarget{cross-validation-4}{%
\subsubsection{Cross Validation}\label{cross-validation-4}}

Instead of using the OOB-score, the classical cross validation can be
used. The number of folds is set by the key \texttt{"k\_folds"}. For
example, to use 5-fold cross validation, the key \texttt{"k\_folds"} is
set to \texttt{5}. Uncomment the following line to use cross validation:

\begin{Shaded}
\begin{Highlighting}[]
\CommentTok{\# fun\_control.update(\{}
\CommentTok{\#      "eval": "train\_cv",}
\CommentTok{\#      "k\_folds": 10,}
\CommentTok{\# \})}
\end{Highlighting}
\end{Shaded}

\hypertarget{step-8-calling-the-spot-function-5}{%
\section{Step 8: Calling the SPOT
Function}\label{step-8-calling-the-spot-function-5}}

\hypertarget{sec-prepare-spot-call-19}{%
\subsection{Preparing the SPOT Call}\label{sec-prepare-spot-call-19}}

\begin{itemize}
\tightlist
\item
  Get types and variable names as well as lower and upper bounds for the
  hyperparameters.
\end{itemize}

\begin{Shaded}
\begin{Highlighting}[]
\CommentTok{\# extract the variable types, names, and bounds}
\ImportTok{from}\NormalTok{ spotPython.hyperparameters.values }\ImportTok{import}\NormalTok{ (get\_bound\_values,}
\NormalTok{    get\_var\_name,}
\NormalTok{    get\_var\_type,)}
\NormalTok{var\_type }\OperatorTok{=}\NormalTok{ get\_var\_type(fun\_control)}
\NormalTok{var\_name }\OperatorTok{=}\NormalTok{ get\_var\_name(fun\_control)}
\NormalTok{lower }\OperatorTok{=}\NormalTok{ get\_bound\_values(fun\_control, }\StringTok{"lower"}\NormalTok{)}
\NormalTok{upper }\OperatorTok{=}\NormalTok{ get\_bound\_values(fun\_control, }\StringTok{"upper"}\NormalTok{)}
\end{Highlighting}
\end{Shaded}

\begin{Shaded}
\begin{Highlighting}[]
\ImportTok{from}\NormalTok{ spotPython.utils.eda }\ImportTok{import}\NormalTok{ gen\_design\_table}
\BuiltInTok{print}\NormalTok{(gen\_design\_table(fun\_control))}
\end{Highlighting}
\end{Shaded}

\begin{verbatim}
| name        | type   | default   |   lower |   upper | transform             |
|-------------|--------|-----------|---------|---------|-----------------------|
| n_neighbors | int    | 2         |       1 |       7 | transform_power_2_int |
| weights     | factor | uniform   |       0 |       1 | None                  |
| algorithm   | factor | auto      |       0 |       3 | None                  |
| leaf_size   | int    | 5         |       2 |       7 | transform_power_2_int |
| p           | int    | 2         |       1 |       2 | None                  |
\end{verbatim}

\hypertarget{sec-the-objective-function-19}{%
\subsection{The Objective
Function}\label{sec-the-objective-function-19}}

The objective function is selected next. It implements an interface from
\texttt{sklearn}'s training, validation, and testing methods to
\texttt{spotPython}.

\begin{Shaded}
\begin{Highlighting}[]
\ImportTok{from}\NormalTok{ spotPython.fun.hypersklearn }\ImportTok{import}\NormalTok{ HyperSklearn}
\NormalTok{fun }\OperatorTok{=}\NormalTok{ HyperSklearn().fun\_sklearn}
\end{Highlighting}
\end{Shaded}

\hypertarget{run-the-spot-optimizer-5}{%
\subsection{\texorpdfstring{Run the \texttt{Spot}
Optimizer}{Run the Spot Optimizer}}\label{run-the-spot-optimizer-5}}

\begin{itemize}
\tightlist
\item
  Run SPOT for approx. x mins (\texttt{max\_time}).
\item
  Note: the run takes longer, because the evaluation time of initial
  design (here: \texttt{initi\_size}, 20 points) is not considered.
\end{itemize}

\begin{Shaded}
\begin{Highlighting}[]
\ImportTok{from}\NormalTok{ spotPython.hyperparameters.values }\ImportTok{import}\NormalTok{ get\_default\_hyperparameters\_as\_array}
\NormalTok{X\_start }\OperatorTok{=}\NormalTok{ get\_default\_hyperparameters\_as\_array(fun\_control)}
\NormalTok{X\_start}
\end{Highlighting}
\end{Shaded}

\begin{verbatim}
array([[2, 0, 0, 5, 2]])
\end{verbatim}

\begin{Shaded}
\begin{Highlighting}[]
\ImportTok{import}\NormalTok{ numpy }\ImportTok{as}\NormalTok{ np}
\ImportTok{from}\NormalTok{ spotPython.spot }\ImportTok{import}\NormalTok{ spot}
\ImportTok{from}\NormalTok{ math }\ImportTok{import}\NormalTok{ inf}
\NormalTok{spot\_tuner }\OperatorTok{=}\NormalTok{ spot.Spot(fun}\OperatorTok{=}\NormalTok{fun,}
\NormalTok{                   lower }\OperatorTok{=}\NormalTok{ lower,}
\NormalTok{                   upper }\OperatorTok{=}\NormalTok{ upper,}
\NormalTok{                   fun\_evals }\OperatorTok{=}\NormalTok{ inf,}
\NormalTok{                   fun\_repeats }\OperatorTok{=} \DecValTok{1}\NormalTok{,}
\NormalTok{                   max\_time }\OperatorTok{=}\NormalTok{ MAX\_TIME,}
\NormalTok{                   noise }\OperatorTok{=} \VariableTok{False}\NormalTok{,}
\NormalTok{                   tolerance\_x }\OperatorTok{=}\NormalTok{ np.sqrt(np.spacing(}\DecValTok{1}\NormalTok{)),}
\NormalTok{                   var\_type }\OperatorTok{=}\NormalTok{ var\_type,}
\NormalTok{                   var\_name }\OperatorTok{=}\NormalTok{ var\_name,}
\NormalTok{                   infill\_criterion }\OperatorTok{=} \StringTok{"y"}\NormalTok{,}
\NormalTok{                   n\_points }\OperatorTok{=} \DecValTok{1}\NormalTok{,}
\NormalTok{                   seed}\OperatorTok{=}\DecValTok{123}\NormalTok{,}
\NormalTok{                   log\_level }\OperatorTok{=} \DecValTok{50}\NormalTok{,}
\NormalTok{                   show\_models}\OperatorTok{=} \VariableTok{False}\NormalTok{,}
\NormalTok{                   show\_progress}\OperatorTok{=} \VariableTok{True}\NormalTok{,}
\NormalTok{                   fun\_control }\OperatorTok{=}\NormalTok{ fun\_control,}
\NormalTok{                   design\_control}\OperatorTok{=}\NormalTok{\{}\StringTok{"init\_size"}\NormalTok{: INIT\_SIZE,}
                                   \StringTok{"repeats"}\NormalTok{: }\DecValTok{1}\NormalTok{\},}
\NormalTok{                   surrogate\_control}\OperatorTok{=}\NormalTok{\{}\StringTok{"noise"}\NormalTok{: }\VariableTok{True}\NormalTok{,}
                                      \StringTok{"cod\_type"}\NormalTok{: }\StringTok{"norm"}\NormalTok{,}
                                      \StringTok{"min\_theta"}\NormalTok{: }\OperatorTok{{-}}\DecValTok{4}\NormalTok{,}
                                      \StringTok{"max\_theta"}\NormalTok{: }\DecValTok{3}\NormalTok{,}
                                      \StringTok{"n\_theta"}\NormalTok{: }\BuiltInTok{len}\NormalTok{(var\_name),}
                                      \StringTok{"model\_fun\_evals"}\NormalTok{: }\DecValTok{10\_000}\NormalTok{,}
                                      \StringTok{"log\_level"}\NormalTok{: }\DecValTok{50}
\NormalTok{                                      \})}
\NormalTok{spot\_tuner.run(X\_start}\OperatorTok{=}\NormalTok{X\_start)}
\end{Highlighting}
\end{Shaded}

\begin{verbatim}
spotPython tuning: -0.71875 [----------] 0.91% 
\end{verbatim}

\begin{verbatim}
spotPython tuning: -0.71875 [----------] 1.85% 
\end{verbatim}

\begin{verbatim}
spotPython tuning: -0.7326388888888888 [----------] 2.89% 
\end{verbatim}

\begin{verbatim}
spotPython tuning: -0.7326388888888888 [----------] 3.30% 
\end{verbatim}

\begin{verbatim}
spotPython tuning: -0.7326388888888888 [----------] 3.71% 
\end{verbatim}

\begin{verbatim}
spotPython tuning: -0.7326388888888888 [----------] 4.15% 
\end{verbatim}

\begin{verbatim}
spotPython tuning: -0.7326388888888888 [----------] 4.58% 
\end{verbatim}

\begin{verbatim}
spotPython tuning: -0.7326388888888888 [#---------] 5.04% 
\end{verbatim}

\begin{verbatim}
spotPython tuning: -0.7326388888888888 [#---------] 5.44% 
\end{verbatim}

\begin{verbatim}
spotPython tuning: -0.7326388888888888 [#---------] 5.91% 
\end{verbatim}

\begin{verbatim}
spotPython tuning: -0.7326388888888888 [#---------] 6.33% 
\end{verbatim}

\begin{verbatim}
spotPython tuning: -0.7326388888888888 [#---------] 7.87% 
\end{verbatim}

\begin{verbatim}
spotPython tuning: -0.7326388888888888 [#---------] 9.47% 
\end{verbatim}

\begin{verbatim}
spotPython tuning: -0.7326388888888888 [#---------] 11.02% 
\end{verbatim}

\begin{verbatim}
spotPython tuning: -0.7326388888888888 [#---------] 12.33% 
\end{verbatim}

\begin{verbatim}
spotPython tuning: -0.7326388888888888 [#---------] 13.90% 
\end{verbatim}

\begin{verbatim}
spotPython tuning: -0.7326388888888888 [##--------] 15.76% 
\end{verbatim}

\begin{verbatim}
spotPython tuning: -0.7326388888888888 [##--------] 17.19% 
\end{verbatim}

\begin{verbatim}
spotPython tuning: -0.7326388888888888 [##--------] 19.91% 
\end{verbatim}

\begin{verbatim}
spotPython tuning: -0.7465277777777777 [##--------] 21.44% 
\end{verbatim}

\begin{verbatim}
spotPython tuning: -0.7465277777777777 [##--------] 23.02% 
\end{verbatim}

\begin{verbatim}
spotPython tuning: -0.7465277777777777 [##--------] 24.40% 
\end{verbatim}

\begin{verbatim}
spotPython tuning: -0.7465277777777777 [###-------] 25.99% 
\end{verbatim}

\begin{verbatim}
spotPython tuning: -0.7465277777777777 [###-------] 27.57% 
\end{verbatim}

\begin{verbatim}
spotPython tuning: -0.7465277777777777 [###-------] 29.11% 
\end{verbatim}

\begin{verbatim}
spotPython tuning: -0.7465277777777777 [###-------] 30.49% 
\end{verbatim}

\begin{verbatim}
spotPython tuning: -0.7465277777777777 [###-------] 32.31% 
\end{verbatim}

\begin{verbatim}
spotPython tuning: -0.7465277777777777 [###-------] 34.69% 
\end{verbatim}

\begin{verbatim}
spotPython tuning: -0.7465277777777777 [####------] 36.58% 
\end{verbatim}

\begin{verbatim}
spotPython tuning: -0.7465277777777777 [####------] 38.43% 
\end{verbatim}

\begin{verbatim}
spotPython tuning: -0.7465277777777777 [####------] 40.64% 
\end{verbatim}

\begin{verbatim}
spotPython tuning: -0.7465277777777777 [####------] 43.54% 
\end{verbatim}

\begin{verbatim}
spotPython tuning: -0.7465277777777777 [#####-----] 47.35% 
\end{verbatim}

\begin{verbatim}
spotPython tuning: -0.7465277777777777 [#####-----] 50.21% 
\end{verbatim}

\begin{verbatim}
spotPython tuning: -0.7465277777777777 [#####-----] 54.47% 
\end{verbatim}

\begin{verbatim}
spotPython tuning: -0.7465277777777777 [######----] 57.77% 
\end{verbatim}

\begin{verbatim}
spotPython tuning: -0.7465277777777777 [######----] 61.95% 
\end{verbatim}

\begin{verbatim}
spotPython tuning: -0.7465277777777777 [#######---] 65.40% 
\end{verbatim}

\begin{verbatim}
spotPython tuning: -0.7465277777777777 [#######---] 68.64% 
\end{verbatim}

\begin{verbatim}
spotPython tuning: -0.7465277777777777 [#######---] 71.76% 
\end{verbatim}

\begin{verbatim}
spotPython tuning: -0.7465277777777777 [#######---] 74.88% 
\end{verbatim}

\begin{verbatim}
spotPython tuning: -0.7465277777777777 [########--] 78.26% 
\end{verbatim}

\begin{verbatim}
spotPython tuning: -0.7465277777777777 [########--] 82.36% 
\end{verbatim}

\begin{verbatim}
spotPython tuning: -0.7465277777777777 [#########-] 85.23% 
\end{verbatim}

\begin{verbatim}
spotPython tuning: -0.7465277777777777 [#########-] 88.90% 
\end{verbatim}

\begin{verbatim}
spotPython tuning: -0.7465277777777777 [#########-] 93.99% 
\end{verbatim}

\begin{verbatim}
spotPython tuning: -0.7465277777777777 [##########] 100.00% Done...
\end{verbatim}

\begin{verbatim}
<spotPython.spot.spot.Spot at 0x28868a230>
\end{verbatim}

\hypertarget{sec-tensorboard-19}{%
\section{Step 9: Tensorboard}\label{sec-tensorboard-19}}

The textual output shown in the console (or code cell) can be visualized
with Tensorboard as described in Section~\ref{sec-tensorboard-14}, see
also the description in the documentation:
\href{https://sequential-parameter-optimization.github.io/spotPython/14_spot_ray_hpt_torch_cifar10.html\#sec-tensorboard-14}{Tensorboard.}

\hypertarget{sec-results-tuning-19}{%
\section{Step 10: Results}\label{sec-results-tuning-19}}

After the hyperparameter tuning run is finished, the progress of the
hyperparameter tuning can be visualized. The following code generates
the progress plot from \textbf{?@fig-progress}.

\begin{Shaded}
\begin{Highlighting}[]
\NormalTok{spot\_tuner.plot\_progress(log\_y}\OperatorTok{=}\VariableTok{False}\NormalTok{,}
\NormalTok{    filename}\OperatorTok{=}\StringTok{"./figures/"} \OperatorTok{+}\NormalTok{ experiment\_name}\OperatorTok{+}\StringTok{"\_progress.png"}\NormalTok{)}
\end{Highlighting}
\end{Shaded}

\begin{figure}[H]

{\centering \includegraphics{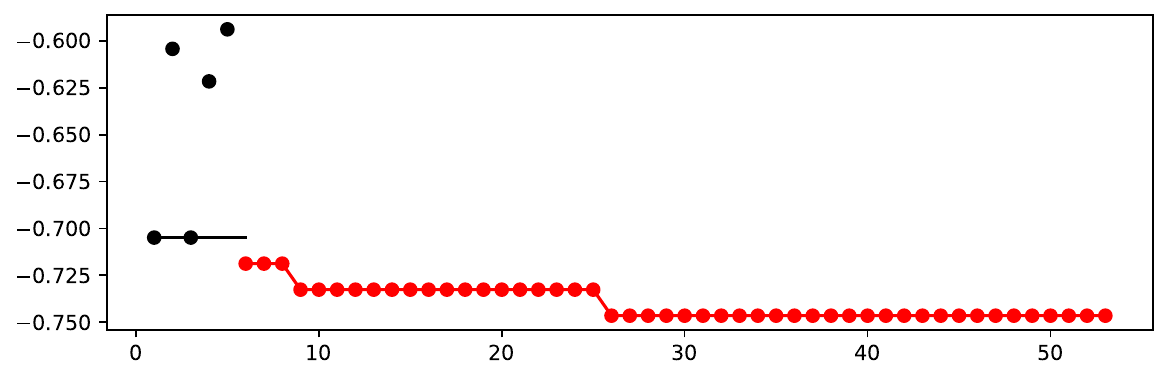}

}

\caption{Progress plot. \emph{Black} dots denote results from the
initial design. \emph{Red} dots illustrate the improvement found by the
surrogate model based optimization.}

\end{figure}

\begin{itemize}
\tightlist
\item
  Print the results
\end{itemize}

\begin{Shaded}
\begin{Highlighting}[]
\BuiltInTok{print}\NormalTok{(gen\_design\_table(fun\_control}\OperatorTok{=}\NormalTok{fun\_control,}
\NormalTok{    spot}\OperatorTok{=}\NormalTok{spot\_tuner))}
\end{Highlighting}
\end{Shaded}

\begin{verbatim}
| name        | type   | default   |   lower |   upper |   tuned | transform             |   importance | stars   |
|-------------|--------|-----------|---------|---------|---------|-----------------------|--------------|---------|
| n_neighbors | int    | 2         |       1 |       7 |     3.0 | transform_power_2_int |       100.00 | ***     |
| weights     | factor | uniform   |       0 |       1 |     0.0 | None                  |        63.21 | **      |
| algorithm   | factor | auto      |       0 |       3 |     1.0 | None                  |         0.00 |         |
| leaf_size   | int    | 5         |       2 |       7 |     4.0 | transform_power_2_int |         0.00 |         |
| p           | int    | 2         |       1 |       2 |     2.0 | None                  |         0.02 |         |
\end{verbatim}

\hypertarget{show-variable-importance-4}{%
\subsection{Show variable importance}\label{show-variable-importance-4}}

\begin{Shaded}
\begin{Highlighting}[]
\NormalTok{spot\_tuner.plot\_importance(threshold}\OperatorTok{=}\FloatTok{0.025}\NormalTok{, filename}\OperatorTok{=}\StringTok{"./figures/"} \OperatorTok{+}\NormalTok{ experiment\_name}\OperatorTok{+}\StringTok{"\_importance.png"}\NormalTok{)}
\end{Highlighting}
\end{Shaded}

\begin{figure}[H]

{\centering \includegraphics{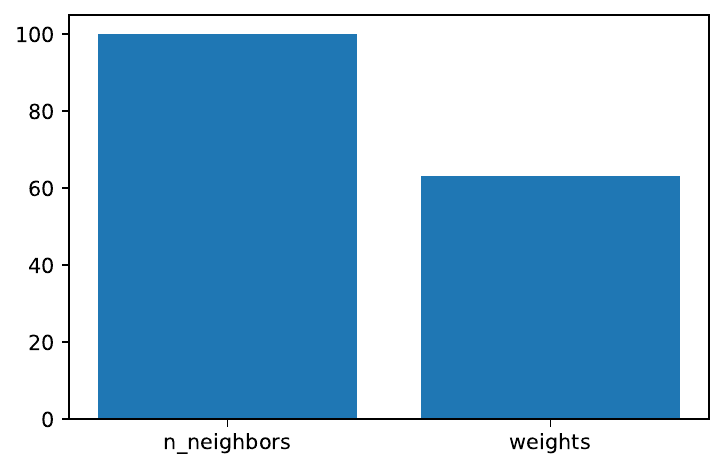}

}

\caption{Variable importance plot, threshold 0.025.}

\end{figure}

\hypertarget{get-default-hyperparameters-6}{%
\subsection{Get Default
Hyperparameters}\label{get-default-hyperparameters-6}}

\begin{Shaded}
\begin{Highlighting}[]
\ImportTok{from}\NormalTok{ spotPython.hyperparameters.values }\ImportTok{import}\NormalTok{ get\_default\_values, transform\_hyper\_parameter\_values}
\NormalTok{values\_default }\OperatorTok{=}\NormalTok{ get\_default\_values(fun\_control)}
\NormalTok{values\_default }\OperatorTok{=}\NormalTok{ transform\_hyper\_parameter\_values(fun\_control}\OperatorTok{=}\NormalTok{fun\_control, hyper\_parameter\_values}\OperatorTok{=}\NormalTok{values\_default)}
\NormalTok{values\_default}
\end{Highlighting}
\end{Shaded}

\begin{verbatim}
{'n_neighbors': 4,
 'weights': 'uniform',
 'algorithm': 'auto',
 'leaf_size': 32,
 'p': 2}
\end{verbatim}

\begin{Shaded}
\begin{Highlighting}[]
\ImportTok{from}\NormalTok{ sklearn.pipeline }\ImportTok{import}\NormalTok{ make\_pipeline}
\NormalTok{model\_default }\OperatorTok{=}\NormalTok{ make\_pipeline(fun\_control[}\StringTok{"prep\_model"}\NormalTok{], fun\_control[}\StringTok{"core\_model"}\NormalTok{](}\OperatorTok{**}\NormalTok{values\_default))}
\NormalTok{model\_default}
\end{Highlighting}
\end{Shaded}

\begin{verbatim}
Pipeline(steps=[('nonetype', None),
                ('kneighborsclassifier',
                 KNeighborsClassifier(leaf_size=32, n_neighbors=4))])
\end{verbatim}

\hypertarget{get-spot-results-5}{%
\subsection{Get SPOT Results}\label{get-spot-results-5}}

\begin{Shaded}
\begin{Highlighting}[]
\NormalTok{X }\OperatorTok{=}\NormalTok{ spot\_tuner.to\_all\_dim(spot\_tuner.min\_X.reshape(}\DecValTok{1}\NormalTok{,}\OperatorTok{{-}}\DecValTok{1}\NormalTok{))}
\BuiltInTok{print}\NormalTok{(X)}
\end{Highlighting}
\end{Shaded}

\begin{verbatim}
[[3. 0. 1. 4. 2.]]
\end{verbatim}

\begin{Shaded}
\begin{Highlighting}[]
\ImportTok{from}\NormalTok{ spotPython.hyperparameters.values }\ImportTok{import}\NormalTok{ assign\_values, return\_conf\_list\_from\_var\_dict}
\NormalTok{v\_dict }\OperatorTok{=}\NormalTok{ assign\_values(X, fun\_control[}\StringTok{"var\_name"}\NormalTok{])}
\NormalTok{return\_conf\_list\_from\_var\_dict(var\_dict}\OperatorTok{=}\NormalTok{v\_dict, fun\_control}\OperatorTok{=}\NormalTok{fun\_control)}
\end{Highlighting}
\end{Shaded}

\begin{verbatim}
[{'n_neighbors': 8,
  'weights': 'uniform',
  'algorithm': 'ball_tree',
  'leaf_size': 16,
  'p': 2}]
\end{verbatim}

\begin{Shaded}
\begin{Highlighting}[]
\ImportTok{from}\NormalTok{ spotPython.hyperparameters.values }\ImportTok{import}\NormalTok{ get\_one\_sklearn\_model\_from\_X}
\NormalTok{model\_spot }\OperatorTok{=}\NormalTok{ get\_one\_sklearn\_model\_from\_X(X, fun\_control)}
\NormalTok{model\_spot}
\end{Highlighting}
\end{Shaded}

\begin{verbatim}
KNeighborsClassifier(algorithm='ball_tree', leaf_size=16, n_neighbors=8)
\end{verbatim}

\hypertarget{evaluate-spot-results-3}{%
\subsection{Evaluate SPOT Results}\label{evaluate-spot-results-3}}

\begin{itemize}
\tightlist
\item
  Fetch the data.
\end{itemize}

\begin{Shaded}
\begin{Highlighting}[]
\ImportTok{from}\NormalTok{ spotPython.utils.convert }\ImportTok{import}\NormalTok{ get\_Xy\_from\_df}
\NormalTok{X\_train, y\_train }\OperatorTok{=}\NormalTok{ get\_Xy\_from\_df(fun\_control[}\StringTok{"train"}\NormalTok{], fun\_control[}\StringTok{"target\_column"}\NormalTok{])}
\NormalTok{X\_test, y\_test }\OperatorTok{=}\NormalTok{ get\_Xy\_from\_df(fun\_control[}\StringTok{"test"}\NormalTok{], fun\_control[}\StringTok{"target\_column"}\NormalTok{])}
\NormalTok{X\_test.shape, y\_test.shape}
\end{Highlighting}
\end{Shaded}

\begin{verbatim}
((63, 64), (63,))
\end{verbatim}

\begin{itemize}
\tightlist
\item
  Fit the model with the tuned hyperparameters. This gives one result:
\end{itemize}

\begin{Shaded}
\begin{Highlighting}[]
\NormalTok{model\_spot.fit(X\_train, y\_train)}
\NormalTok{y\_pred }\OperatorTok{=}\NormalTok{ model\_spot.predict\_proba(X\_test)}
\NormalTok{res }\OperatorTok{=}\NormalTok{ mapk\_score(y\_true}\OperatorTok{=}\NormalTok{y\_test, y\_pred}\OperatorTok{=}\NormalTok{y\_pred, k}\OperatorTok{=}\DecValTok{3}\NormalTok{)}
\NormalTok{res}
\end{Highlighting}
\end{Shaded}

\begin{verbatim}
0.7010582010582012
\end{verbatim}

\begin{Shaded}
\begin{Highlighting}[]
\KeywordTok{def}\NormalTok{ repeated\_eval(n, model):}
\NormalTok{    res\_values }\OperatorTok{=}\NormalTok{ []}
    \ControlFlowTok{for}\NormalTok{ i }\KeywordTok{in} \BuiltInTok{range}\NormalTok{(n):}
\NormalTok{        model.fit(X\_train, y\_train)}
\NormalTok{        y\_pred }\OperatorTok{=}\NormalTok{ model.predict\_proba(X\_test)}
\NormalTok{        res }\OperatorTok{=}\NormalTok{ mapk\_score(y\_true}\OperatorTok{=}\NormalTok{y\_test, y\_pred}\OperatorTok{=}\NormalTok{y\_pred, k}\OperatorTok{=}\DecValTok{3}\NormalTok{)}
\NormalTok{        res\_values.append(res)}
\NormalTok{    mean\_res }\OperatorTok{=}\NormalTok{ np.mean(res\_values)}
    \BuiltInTok{print}\NormalTok{(}\SpecialStringTok{f"mean\_res: }\SpecialCharTok{\{}\NormalTok{mean\_res}\SpecialCharTok{\}}\SpecialStringTok{"}\NormalTok{)}
\NormalTok{    std\_res }\OperatorTok{=}\NormalTok{ np.std(res\_values)}
    \BuiltInTok{print}\NormalTok{(}\SpecialStringTok{f"std\_res: }\SpecialCharTok{\{}\NormalTok{std\_res}\SpecialCharTok{\}}\SpecialStringTok{"}\NormalTok{)}
\NormalTok{    min\_res }\OperatorTok{=}\NormalTok{ np.}\BuiltInTok{min}\NormalTok{(res\_values)}
    \BuiltInTok{print}\NormalTok{(}\SpecialStringTok{f"min\_res: }\SpecialCharTok{\{}\NormalTok{min\_res}\SpecialCharTok{\}}\SpecialStringTok{"}\NormalTok{)}
\NormalTok{    max\_res }\OperatorTok{=}\NormalTok{ np.}\BuiltInTok{max}\NormalTok{(res\_values)}
    \BuiltInTok{print}\NormalTok{(}\SpecialStringTok{f"max\_res: }\SpecialCharTok{\{}\NormalTok{max\_res}\SpecialCharTok{\}}\SpecialStringTok{"}\NormalTok{)}
\NormalTok{    median\_res }\OperatorTok{=}\NormalTok{ np.median(res\_values)}
    \BuiltInTok{print}\NormalTok{(}\SpecialStringTok{f"median\_res: }\SpecialCharTok{\{}\NormalTok{median\_res}\SpecialCharTok{\}}\SpecialStringTok{"}\NormalTok{)}
    \ControlFlowTok{return}\NormalTok{ mean\_res, std\_res, min\_res, max\_res, median\_res}
\end{Highlighting}
\end{Shaded}

\hypertarget{handling-non-deterministic-results-3}{%
\subsection{Handling Non-deterministic
Results}\label{handling-non-deterministic-results-3}}

\begin{itemize}
\tightlist
\item
  Because the model is non-determinstic, we perform \(n=30\) runs and
  calculate the mean and standard deviation of the performance metric.
\end{itemize}

\begin{Shaded}
\begin{Highlighting}[]
\NormalTok{\_ }\OperatorTok{=}\NormalTok{ repeated\_eval(}\DecValTok{30}\NormalTok{, model\_spot)}
\end{Highlighting}
\end{Shaded}

\begin{verbatim}
mean_res: 0.7010582010582015
std_res: 3.3306690738754696e-16
min_res: 0.7010582010582012
max_res: 0.7010582010582012
median_res: 0.7010582010582012
\end{verbatim}

\hypertarget{evalution-of-the-default-hyperparameters-3}{%
\subsection{Evalution of the Default
Hyperparameters}\label{evalution-of-the-default-hyperparameters-3}}

\begin{Shaded}
\begin{Highlighting}[]
\NormalTok{model\_default.fit(X\_train, y\_train)[}\StringTok{"kneighborsclassifier"}\NormalTok{]}
\end{Highlighting}
\end{Shaded}

\begin{verbatim}
KNeighborsClassifier(leaf_size=32, n_neighbors=4)
\end{verbatim}

\begin{itemize}
\tightlist
\item
  One evaluation of the default hyperparameters is performed on the
  hold-out test set.
\end{itemize}

\begin{Shaded}
\begin{Highlighting}[]
\NormalTok{y\_pred }\OperatorTok{=}\NormalTok{ model\_default.predict\_proba(X\_test)}
\NormalTok{mapk\_score(y\_true}\OperatorTok{=}\NormalTok{y\_test, y\_pred}\OperatorTok{=}\NormalTok{y\_pred, k}\OperatorTok{=}\DecValTok{3}\NormalTok{)}
\end{Highlighting}
\end{Shaded}

\begin{verbatim}
0.6878306878306879
\end{verbatim}

Since one single evaluation is not meaningful, we perform, similar to
the evaluation of the SPOT results, \(n=30\) runs of the default setting
and and calculate the mean and standard deviation of the performance
metric.

\begin{Shaded}
\begin{Highlighting}[]
\NormalTok{\_ }\OperatorTok{=}\NormalTok{ repeated\_eval(}\DecValTok{30}\NormalTok{, model\_default)}
\end{Highlighting}
\end{Shaded}

\begin{verbatim}
mean_res: 0.6878306878306877
std_res: 2.220446049250313e-16
min_res: 0.6878306878306879
max_res: 0.6878306878306879
median_res: 0.6878306878306879
\end{verbatim}

\hypertarget{plot-compare-predictions-4}{%
\subsection{Plot: Compare
Predictions}\label{plot-compare-predictions-4}}

\begin{Shaded}
\begin{Highlighting}[]
\ImportTok{from}\NormalTok{ spotPython.plot.validation }\ImportTok{import}\NormalTok{ plot\_confusion\_matrix}
\NormalTok{plot\_confusion\_matrix(model\_default, fun\_control, title }\OperatorTok{=} \StringTok{"Default"}\NormalTok{)}
\end{Highlighting}
\end{Shaded}

\begin{figure}[H]

{\centering \includegraphics{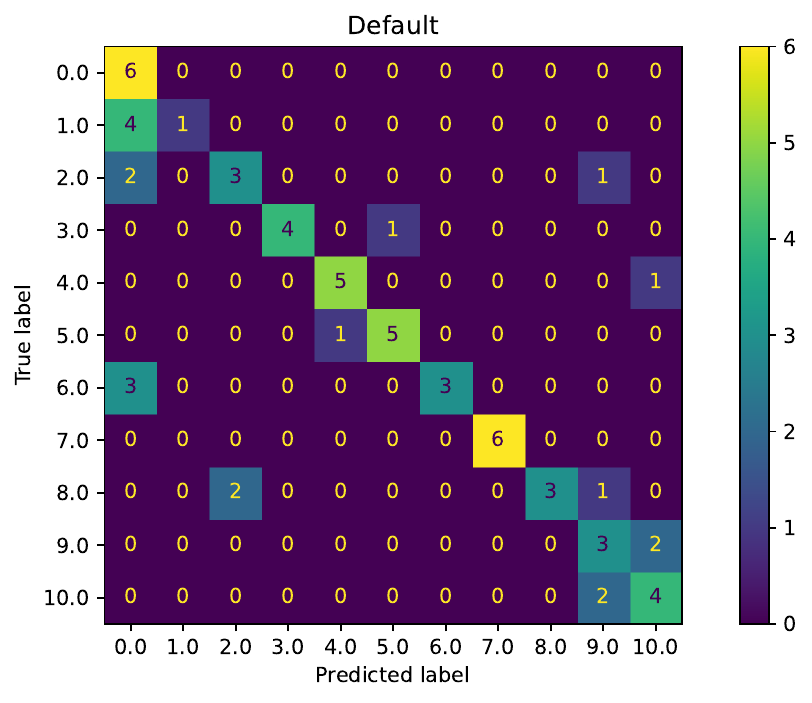}

}

\end{figure}

\begin{Shaded}
\begin{Highlighting}[]
\NormalTok{plot\_confusion\_matrix(model\_spot, fun\_control, title}\OperatorTok{=}\StringTok{"SPOT"}\NormalTok{)}
\end{Highlighting}
\end{Shaded}

\begin{figure}[H]

{\centering \includegraphics{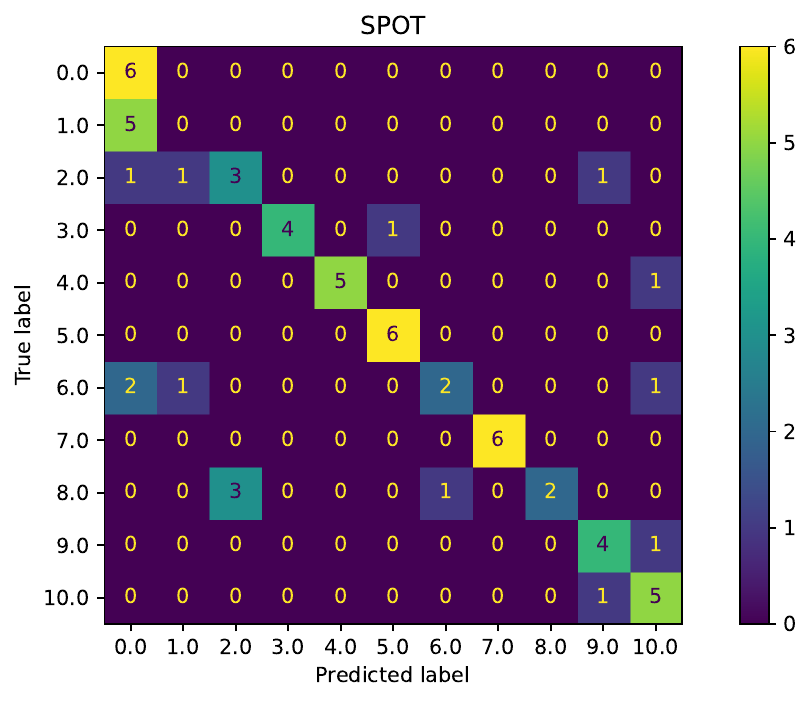}

}

\end{figure}

\begin{Shaded}
\begin{Highlighting}[]
\BuiltInTok{min}\NormalTok{(spot\_tuner.y), }\BuiltInTok{max}\NormalTok{(spot\_tuner.y)}
\end{Highlighting}
\end{Shaded}

\begin{verbatim}
(-0.7465277777777777, -0.16666666666666666)
\end{verbatim}

\hypertarget{cross-validated-evaluations-3}{%
\subsection{Cross-validated
Evaluations}\label{cross-validated-evaluations-3}}

\begin{Shaded}
\begin{Highlighting}[]
\ImportTok{from}\NormalTok{ spotPython.sklearn.traintest }\ImportTok{import}\NormalTok{ evaluate\_cv}
\NormalTok{fun\_control.update(\{}
     \StringTok{"eval"}\NormalTok{: }\StringTok{"train\_cv"}\NormalTok{,}
     \StringTok{"k\_folds"}\NormalTok{: }\DecValTok{10}\NormalTok{,}
\NormalTok{\})}
\NormalTok{evaluate\_cv(model}\OperatorTok{=}\NormalTok{model\_spot, fun\_control}\OperatorTok{=}\NormalTok{fun\_control, verbose}\OperatorTok{=}\DecValTok{0}\NormalTok{)}
\end{Highlighting}
\end{Shaded}

\begin{verbatim}
(0.7156920077972708, None)
\end{verbatim}

\begin{Shaded}
\begin{Highlighting}[]
\NormalTok{fun\_control.update(\{}
     \StringTok{"eval"}\NormalTok{: }\StringTok{"test\_cv"}\NormalTok{,}
     \StringTok{"k\_folds"}\NormalTok{: }\DecValTok{10}\NormalTok{,}
\NormalTok{\})}
\NormalTok{evaluate\_cv(model}\OperatorTok{=}\NormalTok{model\_spot, fun\_control}\OperatorTok{=}\NormalTok{fun\_control, verbose}\OperatorTok{=}\DecValTok{0}\NormalTok{)}
\end{Highlighting}
\end{Shaded}

\begin{verbatim}
Error in fun_sklearn(). Call to evaluate_cv failed. err=ValueError('n_splits=10 cannot be greater than the number of members in each class.'), type(err)=<class 'ValueError'>
\end{verbatim}

\begin{verbatim}
(nan, None)
\end{verbatim}

\begin{itemize}
\tightlist
\item
  This is the evaluation that will be used in the comparison:
\end{itemize}

\begin{Shaded}
\begin{Highlighting}[]
\NormalTok{fun\_control.update(\{}
     \StringTok{"eval"}\NormalTok{: }\StringTok{"data\_cv"}\NormalTok{,}
     \StringTok{"k\_folds"}\NormalTok{: }\DecValTok{10}\NormalTok{,}
\NormalTok{\})}
\NormalTok{evaluate\_cv(model}\OperatorTok{=}\NormalTok{model\_spot, fun\_control}\OperatorTok{=}\NormalTok{fun\_control, verbose}\OperatorTok{=}\DecValTok{0}\NormalTok{)}
\end{Highlighting}
\end{Shaded}

\begin{verbatim}
(0.7089487179487179, None)
\end{verbatim}

\hypertarget{detailed-hyperparameter-plots-6}{%
\subsection{Detailed Hyperparameter
Plots}\label{detailed-hyperparameter-plots-6}}

\begin{Shaded}
\begin{Highlighting}[]
\NormalTok{filename }\OperatorTok{=} \StringTok{"./figures/"} \OperatorTok{+}\NormalTok{ experiment\_name}
\NormalTok{spot\_tuner.plot\_important\_hyperparameter\_contour(filename}\OperatorTok{=}\NormalTok{filename)}
\end{Highlighting}
\end{Shaded}

\begin{verbatim}
n_neighbors:  100.0
weights:  63.20992884888382
\end{verbatim}

\begin{figure}[H]

{\centering \includegraphics{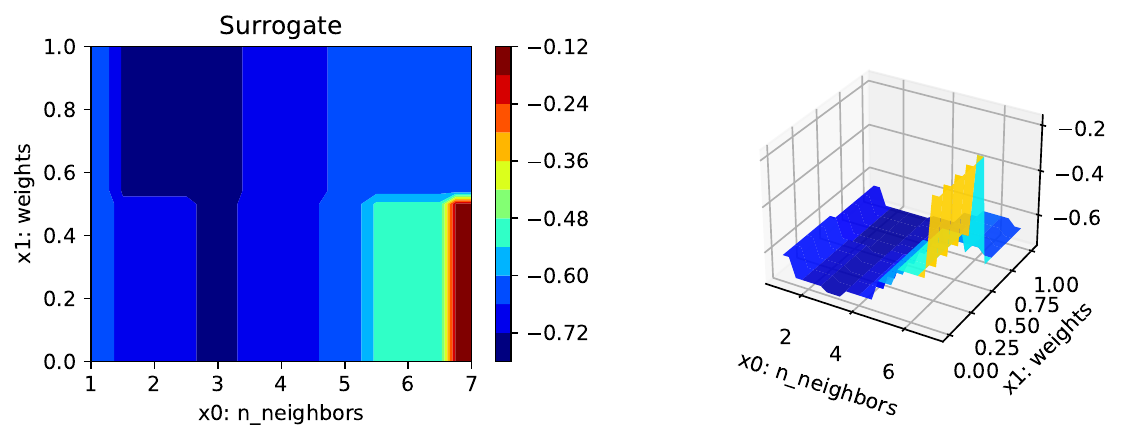}

}

\end{figure}

\hypertarget{parallel-coordinates-plot-4}{%
\subsection{Parallel Coordinates
Plot}\label{parallel-coordinates-plot-4}}

\begin{Shaded}
\begin{Highlighting}[]
\NormalTok{spot\_tuner.parallel\_plot()}
\end{Highlighting}
\end{Shaded}

\begin{verbatim}
Unable to display output for mime type(s): text/html
\end{verbatim}

\begin{verbatim}
Unable to display output for mime type(s): text/html
\end{verbatim}

\hypertarget{plot-all-combinations-of-hyperparameters-5}{%
\subsection{Plot all Combinations of
Hyperparameters}\label{plot-all-combinations-of-hyperparameters-5}}

\begin{itemize}
\tightlist
\item
  Warning: this may take a while.
\end{itemize}

\begin{Shaded}
\begin{Highlighting}[]
\NormalTok{PLOT\_ALL }\OperatorTok{=} \VariableTok{False}
\ControlFlowTok{if}\NormalTok{ PLOT\_ALL:}
\NormalTok{    n }\OperatorTok{=}\NormalTok{ spot\_tuner.k}
    \ControlFlowTok{for}\NormalTok{ i }\KeywordTok{in} \BuiltInTok{range}\NormalTok{(n}\OperatorTok{{-}}\DecValTok{1}\NormalTok{):}
        \ControlFlowTok{for}\NormalTok{ j }\KeywordTok{in} \BuiltInTok{range}\NormalTok{(i}\OperatorTok{+}\DecValTok{1}\NormalTok{, n):}
\NormalTok{            spot\_tuner.plot\_contour(i}\OperatorTok{=}\NormalTok{i, j}\OperatorTok{=}\NormalTok{j, min\_z}\OperatorTok{=}\NormalTok{min\_z, max\_z }\OperatorTok{=}\NormalTok{ max\_z)}
\end{Highlighting}
\end{Shaded}

\hypertarget{sec-hyperparameter-tuning-lightning-31}{%
\chapter{HPT PyTorch Lightning:
VBDP}\label{sec-hyperparameter-tuning-lightning-31}}

In this tutorial, we will show how \texttt{spotPython} can be integrated
into the \texttt{PyTorch} Lightning training workflow for a
classification task.

This chapter describes the hyperparameter tuning of a
\texttt{PyTorch\ Lightning} network on the Vector Borne Disease
Prediction (VBDP) data set.

\begin{tcolorbox}[enhanced jigsaw, left=2mm, title=\textcolor{quarto-callout-important-color}{\faExclamation}\hspace{0.5em}{Vector Borne Disease Prediction Data Set}, bottomrule=.15mm, titlerule=0mm, breakable, rightrule=.15mm, toprule=.15mm, coltitle=black, colbacktitle=quarto-callout-important-color!10!white, leftrule=.75mm, arc=.35mm, colframe=quarto-callout-important-color-frame, bottomtitle=1mm, colback=white, opacitybacktitle=0.6, toptitle=1mm, opacityback=0]

This chapter uses the Vector Borne Disease Prediction data set from
Kaggle. It is a categorical dataset for eleven Vector Borne Diseases
with associated symptoms.

\begin{quote}
The person who associated a work with this deed has dedicated the work
to the public domain by waiving all of his or her rights to the work
worldwide under copyright law, including all related and neighboring
rights, to the extent allowed by law.You can copy, modify, distribute
and perform the work, even for commercial purposes, all without asking
permission. See Other Information below, see
\url{https://creativecommons.org/publicdomain/zero/1.0/}.
\end{quote}

The data set is available at:
\url{https://www.kaggle.com/datasets/richardbernat/vector-borne-disease-prediction},

The data should be downloaded and stored in the \texttt{data/VBDP}
subfolder. The data set is not available as a part of the
\texttt{spotPython} package.

\end{tcolorbox}

This document refers to the latest \texttt{spotPython} version, which
can be installed via pip. Alternatively, the source code can be
downloaded from gitHub:
\url{https://github.com/sequential-parameter-optimization/spotPython}.

\begin{itemize}
\tightlist
\item
  Uncomment the following lines if you want to for (re-)installation the
  latest version of \texttt{spotPython} from GitHub.
\end{itemize}

\begin{Shaded}
\begin{Highlighting}[]
\CommentTok{\# import sys}
\CommentTok{\# !\{sys.executable\} {-}m pip install {-}{-}upgrade build}
\CommentTok{\# !\{sys.executable\} {-}m pip install {-}{-}upgrade {-}{-}force{-}reinstall spotPython}
\end{Highlighting}
\end{Shaded}

\hypertarget{sec-setup-31}{%
\section{Step 1: Setup}\label{sec-setup-31}}

\begin{itemize}
\tightlist
\item
  Before we consider the detailed experimental setup, we select the
  parameters that affect run time, initial design size, etc.
\item
  The parameter \texttt{MAX\_TIME} specifies the maximum run time in
  seconds.
\item
  The parameter \texttt{INIT\_SIZE} specifies the initial design size.
\item
  The parameter \texttt{WORKERS} specifies the number of workers.
\item
  The prefix \texttt{PREFIX} is used for the experiment name and the
  name of the log file.
\end{itemize}

\begin{Shaded}
\begin{Highlighting}[]
\NormalTok{MAX\_TIME }\OperatorTok{=} \DecValTok{1}
\NormalTok{INIT\_SIZE }\OperatorTok{=} \DecValTok{5}
\NormalTok{WORKERS }\OperatorTok{=} \DecValTok{0}
\NormalTok{PREFIX}\OperatorTok{=}\StringTok{"31"}
\end{Highlighting}
\end{Shaded}

\begin{tcolorbox}[enhanced jigsaw, left=2mm, title=\textcolor{quarto-callout-caution-color}{\faFire}\hspace{0.5em}{Caution: Run time and initial design size should be increased for real
experiments}, bottomrule=.15mm, titlerule=0mm, breakable, rightrule=.15mm, toprule=.15mm, coltitle=black, colbacktitle=quarto-callout-caution-color!10!white, leftrule=.75mm, arc=.35mm, colframe=quarto-callout-caution-color-frame, bottomtitle=1mm, colback=white, opacitybacktitle=0.6, toptitle=1mm, opacityback=0]

\begin{itemize}
\tightlist
\item
  \texttt{MAX\_TIME} is set to one minute for demonstration purposes.
  For real experiments, this should be increased to at least 1 hour.
\item
  \texttt{INIT\_SIZE} is set to 5 for demonstration purposes. For real
  experiments, this should be increased to at least 10.
\item
  \texttt{WORKERS} is set to 0 for demonstration purposes. For real
  experiments, this should be increased. See the warnings that are
  printed when the number of workers is set to 0.
\end{itemize}

\end{tcolorbox}

\begin{tcolorbox}[enhanced jigsaw, left=2mm, title=\textcolor{quarto-callout-note-color}{\faInfo}\hspace{0.5em}{Note: Device selection}, bottomrule=.15mm, titlerule=0mm, breakable, rightrule=.15mm, toprule=.15mm, coltitle=black, colbacktitle=quarto-callout-note-color!10!white, leftrule=.75mm, arc=.35mm, colframe=quarto-callout-note-color-frame, bottomtitle=1mm, colback=white, opacitybacktitle=0.6, toptitle=1mm, opacityback=0]

\begin{itemize}
\tightlist
\item
  Although there are no .cuda() or .to(device) calls required, because
  Lightning does these for you, see
  \href{https://lightning.ai/docs/pytorch/stable/common/lightning_module.html}{LIGHTNINGMODULE},
  we would like to know which device is used. Threrefore, we imitate the
  LightningModule behaviour which selects the highest device.
\item
  The method \texttt{spotPython.utils.device.getDevice()} returns the
  device that is used by Lightning.
\end{itemize}

\end{tcolorbox}

\hypertarget{step-2-initialization-of-the-fun_control-dictionary}{%
\section{\texorpdfstring{Step 2: Initialization of the
\texttt{fun\_control}
Dictionary}{Step 2: Initialization of the fun\_control Dictionary}}\label{step-2-initialization-of-the-fun_control-dictionary}}

\texttt{spotPython} uses a Python dictionary for storing the information
required for the hyperparameter tuning process, which was described in
Section~\ref{sec-initialization-fun-control-14}, see
\href{https://sequential-parameter-optimization.github.io/spotPython/14_spot_ray_hpt_torch_cifar10.html\#sec-initialization-fun-control-14}{Initialization
of the fun\_control Dictionary} in the documentation.

\begin{Shaded}
\begin{Highlighting}[]
\ImportTok{from}\NormalTok{ spotPython.utils.init }\ImportTok{import}\NormalTok{ fun\_control\_init}
\ImportTok{from}\NormalTok{ spotPython.utils.}\BuiltInTok{file} \ImportTok{import}\NormalTok{ get\_experiment\_name, get\_spot\_tensorboard\_path}
\ImportTok{from}\NormalTok{ spotPython.utils.device }\ImportTok{import}\NormalTok{ getDevice}

\NormalTok{experiment\_name }\OperatorTok{=}\NormalTok{ get\_experiment\_name(prefix}\OperatorTok{=}\NormalTok{PREFIX)}
\NormalTok{fun\_control }\OperatorTok{=}\NormalTok{ fun\_control\_init(}
\NormalTok{    spot\_tensorboard\_path}\OperatorTok{=}\NormalTok{get\_spot\_tensorboard\_path(experiment\_name),}
\NormalTok{    num\_workers}\OperatorTok{=}\NormalTok{WORKERS,}
\NormalTok{    device}\OperatorTok{=}\NormalTok{getDevice(),}
\NormalTok{    \_L\_in}\OperatorTok{=}\DecValTok{64}\NormalTok{,}
\NormalTok{    \_L\_out}\OperatorTok{=}\DecValTok{11}\NormalTok{,}
\NormalTok{    TENSORBOARD\_CLEAN}\OperatorTok{=}\VariableTok{True}\NormalTok{)}
\end{Highlighting}
\end{Shaded}

\begin{Shaded}
\begin{Highlighting}[]
\NormalTok{fun\_control[}\StringTok{"device"}\NormalTok{]}
\end{Highlighting}
\end{Shaded}

\begin{verbatim}
'mps'
\end{verbatim}

\hypertarget{sec-data-loading-31}{%
\section{Step 3: PyTorch Data Loading}\label{sec-data-loading-31}}

\hypertarget{lightning-dataset-and-datamodule}{%
\subsection{Lightning Dataset and
DataModule}\label{lightning-dataset-and-datamodule}}

The data loading and preprocessing is handled by \texttt{Lightning} and
\texttt{PyTorch}. It comprehends the following classes:

\begin{itemize}
\tightlist
\item
  \texttt{CSVDataset}: A class that loads the data from a CSV file.
  \href{https://github.com/sequential-parameter-optimization/spotPython/blob/main/src/spotPython/light/csvdataset.py}{{[}SOURCE{]}}
\item
  \texttt{CSVDataModule}: A class that prepares the data for training
  and testing.
  \href{https://github.com/sequential-parameter-optimization/spotPython/blob/main/src/spotPython/light/csvdatamodule.py}{{[}SOURCE{]}}
\end{itemize}

Section Section~\ref{sec-taking-a-look-at-the-data-31} illustrates how
to access the data.

\hypertarget{sec-preprocessing-31}{%
\section{Step 4: Preprocessing}\label{sec-preprocessing-31}}

Preprocessing is handled by \texttt{Lightning} and \texttt{PyTorch}. It
can be implemented in the \texttt{CSVDataModule} class
\href{https://github.com/sequential-parameter-optimization/spotPython/blob/main/src/spotPython/light/csvdatamodule.py}{{[}SOURCE{]}}
and is described in the
\href{https://lightning.ai/docs/pytorch/stable/data/datamodule.html}{LIGHTNINGDATAMODULE}
documentation. Here you can find information about the
\texttt{transforms} methods.

\hypertarget{sec-selection-of-the-algorithm-31}{%
\section{\texorpdfstring{Step 5: Select the NN Model
(\texttt{algorithm}) and
\texttt{core\_model\_hyper\_dict}}{Step 5: Select the NN Model (algorithm) and core\_model\_hyper\_dict}}\label{sec-selection-of-the-algorithm-31}}

\texttt{spotPython} includes the \texttt{NetLightBase} class
\href{https://github.com/sequential-parameter-optimization/spotPython/blob/main/src/spotPython/light/netlightbase.py}{{[}SOURCE{]}}
for configurable neural networks. The class is imported here. It
inherits from the class \texttt{Lightning.LightningModule}, which is the
base class for all models in \texttt{Lightning}.
\texttt{Lightning.LightningModule} is a subclass of
\texttt{torch.nn.Module} and provides additional functionality for the
training and testing of neural networks. The class
\texttt{Lightning.LightningModule} is described in the
\href{https://lightning.ai/docs/pytorch/stable/common/lightning_module.html}{Lightning
documentation}.

\begin{itemize}
\tightlist
\item
  Here we simply add the NN Model to the fun\_control dictionary by
  calling the function \texttt{add\_core\_model\_to\_fun\_control}:
\end{itemize}

\begin{Shaded}
\begin{Highlighting}[]
\ImportTok{from}\NormalTok{ spotPython.light.netlightbase }\ImportTok{import}\NormalTok{ NetLightBase }
\ImportTok{from}\NormalTok{ spotPython.data.light\_hyper\_dict }\ImportTok{import}\NormalTok{ LightHyperDict}
\ImportTok{from}\NormalTok{ spotPython.hyperparameters.values }\ImportTok{import}\NormalTok{ add\_core\_model\_to\_fun\_control}
\NormalTok{add\_core\_model\_to\_fun\_control(core\_model}\OperatorTok{=}\NormalTok{NetLightBase,}
\NormalTok{                              fun\_control}\OperatorTok{=}\NormalTok{fun\_control,}
\NormalTok{                              hyper\_dict}\OperatorTok{=}\NormalTok{ LightHyperDict)}
\end{Highlighting}
\end{Shaded}

The \texttt{NetLightBase} is a configurable neural network. The
hyperparameters of the model are specified in the
\texttt{core\_model\_hyper\_dict} dictionary
\href{https://github.com/sequential-parameter-optimization/spotPython/blob/main/src/spotPython/data/light_hyper_dict.json}{{[}SOURCE{]}}.

\hypertarget{sec-modification-of-hyperparameters-31}{%
\section{\texorpdfstring{Step 6: Modify \texttt{hyper\_dict}
Hyperparameters for the Selected Algorithm aka
\texttt{core\_model}}{Step 6: Modify hyper\_dict Hyperparameters for the Selected Algorithm aka core\_model}}\label{sec-modification-of-hyperparameters-31}}

\texttt{spotPython} provides functions for modifying the
hyperparameters, their bounds and factors as well as for activating and
de-activating hyperparameters without re-compilation of the Python
source code. These functions were described in
Section~\ref{sec-modification-of-hyperparameters-14}.

\begin{tcolorbox}[enhanced jigsaw, left=2mm, title=\textcolor{quarto-callout-caution-color}{\faFire}\hspace{0.5em}{Caution: Small number of epochs for demonstration purposes}, bottomrule=.15mm, titlerule=0mm, breakable, rightrule=.15mm, toprule=.15mm, coltitle=black, colbacktitle=quarto-callout-caution-color!10!white, leftrule=.75mm, arc=.35mm, colframe=quarto-callout-caution-color-frame, bottomtitle=1mm, colback=white, opacitybacktitle=0.6, toptitle=1mm, opacityback=0]

\begin{itemize}
\tightlist
\item
  \texttt{epochs} and \texttt{patience} are set to small values for
  demonstration purposes. These values are too small for a real
  application.
\item
  More resonable values are, e.g.:

  \begin{itemize}
  \tightlist
  \item
    \texttt{modify\_hyper\_parameter\_bounds(fun\_control,\ "epochs",\ bounds={[}7,\ 9{]})}
    and
  \item
    \texttt{modify\_hyper\_parameter\_bounds(fun\_control,\ "patience",\ bounds={[}2,\ 7{]})}
  \end{itemize}
\end{itemize}

\end{tcolorbox}

\begin{Shaded}
\begin{Highlighting}[]
\ImportTok{from}\NormalTok{ spotPython.hyperparameters.values }\ImportTok{import}\NormalTok{ modify\_hyper\_parameter\_bounds}

\NormalTok{modify\_hyper\_parameter\_bounds(fun\_control, }\StringTok{"l1"}\NormalTok{, bounds}\OperatorTok{=}\NormalTok{[}\DecValTok{5}\NormalTok{,}\DecValTok{8}\NormalTok{])}
\NormalTok{modify\_hyper\_parameter\_bounds(fun\_control, }\StringTok{"epochs"}\NormalTok{, bounds}\OperatorTok{=}\NormalTok{[}\DecValTok{6}\NormalTok{,}\DecValTok{13}\NormalTok{])}
\NormalTok{modify\_hyper\_parameter\_bounds(fun\_control, }\StringTok{"batch\_size"}\NormalTok{, bounds}\OperatorTok{=}\NormalTok{[}\DecValTok{2}\NormalTok{, }\DecValTok{8}\NormalTok{])}
\end{Highlighting}
\end{Shaded}

\begin{Shaded}
\begin{Highlighting}[]
\ImportTok{from}\NormalTok{ spotPython.hyperparameters.values }\ImportTok{import}\NormalTok{ modify\_hyper\_parameter\_levels}
\NormalTok{modify\_hyper\_parameter\_levels(fun\_control, }\StringTok{"optimizer"}\NormalTok{,[}\StringTok{"Adam"}\NormalTok{, }\StringTok{"AdamW"}\NormalTok{, }\StringTok{"Adamax"}\NormalTok{, }\StringTok{"NAdam"}\NormalTok{])}
\CommentTok{\# modify\_hyper\_parameter\_levels(fun\_control, "optimizer", ["Adam"])}
\end{Highlighting}
\end{Shaded}

Now, the dictionary \texttt{fun\_control} contains all information
needed for the hyperparameter tuning. Before the hyperparameter tuning
is started, it is recommended to take a look at the experimental design.
The method \texttt{gen\_design\_table}
\href{https://github.com/sequential-parameter-optimization/spotPython/blob/main/src/spotPython/utils/eda.py}{{[}SOURCE{]}}
generates a design table as follows:

\begin{Shaded}
\begin{Highlighting}[]
\ImportTok{from}\NormalTok{ spotPython.utils.eda }\ImportTok{import}\NormalTok{ gen\_design\_table}
\BuiltInTok{print}\NormalTok{(gen\_design\_table(fun\_control))}
\end{Highlighting}
\end{Shaded}

\begin{verbatim}
| name           | type   | default   |   lower |   upper | transform             |
|----------------|--------|-----------|---------|---------|-----------------------|
| l1             | int    | 3         |     5   |    8    | transform_power_2_int |
| epochs         | int    | 4         |     6   |   13    | transform_power_2_int |
| batch_size     | int    | 4         |     2   |    8    | transform_power_2_int |
| act_fn         | factor | ReLU      |     0   |    5    | None                  |
| optimizer      | factor | SGD       |     0   |    3    | None                  |
| dropout_prob   | float  | 0.01      |     0   |    0.25 | None                  |
| lr_mult        | float  | 1.0       |     0.1 |   10    | None                  |
| patience       | int    | 2         |     2   |    6    | transform_power_2_int |
| initialization | factor | Default   |     0   |    2    | None                  |
\end{verbatim}

This allows to check if all information is available and if the
information is correct.

\begin{tcolorbox}[enhanced jigsaw, left=2mm, title=\textcolor{quarto-callout-note-color}{\faInfo}\hspace{0.5em}{Note: Hyperparameters of the Tuned Model and the \texttt{fun\_control}
Dictionary}, bottomrule=.15mm, titlerule=0mm, breakable, rightrule=.15mm, toprule=.15mm, coltitle=black, colbacktitle=quarto-callout-note-color!10!white, leftrule=.75mm, arc=.35mm, colframe=quarto-callout-note-color-frame, bottomtitle=1mm, colback=white, opacitybacktitle=0.6, toptitle=1mm, opacityback=0]

The updated \texttt{fun\_control} dictionary can be shown with the
command \texttt{fun\_control{[}"core\_model\_hyper\_dict"{]}}.

\end{tcolorbox}

\hypertarget{step-7-data-splitting-the-objective-loss-function-and-the-metric}{%
\section{Step 7: Data Splitting, the Objective (Loss) Function and the
Metric}\label{step-7-data-splitting-the-objective-loss-function-and-the-metric}}

\hypertarget{sec-selection-of-target-function-31}{%
\subsection{Evaluation}\label{sec-selection-of-target-function-31}}

The evaluation procedure requires the specification of two elements:

\begin{enumerate}
\def\labelenumi{\arabic{enumi}.}
\tightlist
\item
  the way how the data is split into a train and a test set (see
  Section~\ref{sec-data-splitting-14})
\item
  the loss function (and a metric).
\end{enumerate}

\begin{tcolorbox}[enhanced jigsaw, left=2mm, title=\textcolor{quarto-callout-caution-color}{\faFire}\hspace{0.5em}{Caution: Data Splitting in Lightning}, bottomrule=.15mm, titlerule=0mm, breakable, rightrule=.15mm, toprule=.15mm, coltitle=black, colbacktitle=quarto-callout-caution-color!10!white, leftrule=.75mm, arc=.35mm, colframe=quarto-callout-caution-color-frame, bottomtitle=1mm, colback=white, opacitybacktitle=0.6, toptitle=1mm, opacityback=0]

\begin{itemize}
\tightlist
\item
  The data splitting is handled by \texttt{Lightning}.
\end{itemize}

\end{tcolorbox}

\hypertarget{sec-loss-functions-and-metrics-31}{%
\subsection{Loss Functions and
Metrics}\label{sec-loss-functions-and-metrics-31}}

The loss function is specified in the configurable network class
\href{https://github.com/sequential-parameter-optimization/spotPython/blob/main/src/spotPython/light/netlightbase.py}{{[}SOURCE{]}}
We will use CrossEntropy loss for the multiclass-classification task.

\hypertarget{sec-metric-31}{%
\subsection{Metric}\label{sec-metric-31}}

\begin{itemize}
\tightlist
\item
  We will use the MAP@k metric
  \href{https://github.com/sequential-parameter-optimization/spotPython/blob/main/src/spotPython/torch/mapk.py}{{[}SOURCE{]}}
  for the evaluation of the model.
\item
  An example, how this metric works, is shown in the Appendix, see
  Section \{Section~\ref{sec-the-mapk-metric-31}\}.
\end{itemize}

Similar to the loss function, the metric is specified in the
configurable network class
\href{https://github.com/sequential-parameter-optimization/spotPython/blob/main/src/spotPython/light/netlightbase.py}{{[}SOURCE{]}}.

\begin{tcolorbox}[enhanced jigsaw, left=2mm, title=\textcolor{quarto-callout-caution-color}{\faFire}\hspace{0.5em}{Caution: Loss Function and Metric in Lightning}, bottomrule=.15mm, titlerule=0mm, breakable, rightrule=.15mm, toprule=.15mm, coltitle=black, colbacktitle=quarto-callout-caution-color!10!white, leftrule=.75mm, arc=.35mm, colframe=quarto-callout-caution-color-frame, bottomtitle=1mm, colback=white, opacitybacktitle=0.6, toptitle=1mm, opacityback=0]

\begin{itemize}
\tightlist
\item
  The loss function and the metric are not hyperparameters that can be
  tuned with \texttt{spotPython}.
\item
  They are handled by \texttt{Lightning}.
\end{itemize}

\end{tcolorbox}

\hypertarget{step-8-calling-the-spot-function-6}{%
\section{Step 8: Calling the SPOT
Function}\label{step-8-calling-the-spot-function-6}}

\hypertarget{sec-prepare-spot-call-31}{%
\subsection{Preparing the SPOT Call}\label{sec-prepare-spot-call-31}}

The following code passes the information about the parameter ranges and
bounds to \texttt{spot}. It extracts the variable types, names, and
bounds

\begin{Shaded}
\begin{Highlighting}[]
\ImportTok{from}\NormalTok{ spotPython.hyperparameters.values }\ImportTok{import}\NormalTok{ (get\_bound\_values,}
\NormalTok{    get\_var\_name,}
\NormalTok{    get\_var\_type,)}
\NormalTok{var\_type }\OperatorTok{=}\NormalTok{ get\_var\_type(fun\_control)}
\NormalTok{var\_name }\OperatorTok{=}\NormalTok{ get\_var\_name(fun\_control)}
\NormalTok{lower }\OperatorTok{=}\NormalTok{ get\_bound\_values(fun\_control, }\StringTok{"lower"}\NormalTok{)}
\NormalTok{upper }\OperatorTok{=}\NormalTok{ get\_bound\_values(fun\_control, }\StringTok{"upper"}\NormalTok{)}
\end{Highlighting}
\end{Shaded}

\hypertarget{sec-the-objective-function-31}{%
\subsection{\texorpdfstring{The Objective Function
\texttt{fun}}{The Objective Function fun}}\label{sec-the-objective-function-31}}

The objective function \texttt{fun} from the class \texttt{HyperLight}
\href{https://github.com/sequential-parameter-optimization/spotPython/blob/main/src/spotPython/light/hyperlight.py}{{[}SOURCE{]}}
is selected next. It implements an interface from \texttt{PyTorch}'s
training, validation, and testing methods to \texttt{spotPython}.

\begin{Shaded}
\begin{Highlighting}[]
\ImportTok{from}\NormalTok{ spotPython.fun.hyperlight }\ImportTok{import}\NormalTok{ HyperLight}
\NormalTok{fun }\OperatorTok{=}\NormalTok{ HyperLight().fun}
\end{Highlighting}
\end{Shaded}

\hypertarget{sec-call-the-hyperparameter-tuner-31}{%
\subsection{Starting the Hyperparameter
Tuning}\label{sec-call-the-hyperparameter-tuner-31}}

The \texttt{spotPython} hyperparameter tuning is started by calling the
\texttt{Spot} function
\href{https://github.com/sequential-parameter-optimization/spotPython/blob/main/src/spotPython/spot/spot.py}{{[}SOURCE{]}}
as described in Section~\ref{sec-call-the-hyperparameter-tuner-14}.

\begin{Shaded}
\begin{Highlighting}[]
\ImportTok{import}\NormalTok{ numpy }\ImportTok{as}\NormalTok{ np}
\ImportTok{from}\NormalTok{ spotPython.spot }\ImportTok{import}\NormalTok{ spot}
\ImportTok{from}\NormalTok{ math }\ImportTok{import}\NormalTok{ inf}
\NormalTok{spot\_tuner }\OperatorTok{=}\NormalTok{ spot.Spot(fun}\OperatorTok{=}\NormalTok{fun,}
\NormalTok{                   lower }\OperatorTok{=}\NormalTok{ lower,}
\NormalTok{                   upper }\OperatorTok{=}\NormalTok{ upper,}
\NormalTok{                   fun\_evals }\OperatorTok{=}\NormalTok{ inf,}
\NormalTok{                   max\_time }\OperatorTok{=}\NormalTok{ MAX\_TIME,}
\NormalTok{                   tolerance\_x }\OperatorTok{=}\NormalTok{ np.sqrt(np.spacing(}\DecValTok{1}\NormalTok{)),}
\NormalTok{                   var\_type }\OperatorTok{=}\NormalTok{ var\_type,}
\NormalTok{                   var\_name }\OperatorTok{=}\NormalTok{ var\_name,}
\NormalTok{                   show\_progress}\OperatorTok{=} \VariableTok{True}\NormalTok{,}
\NormalTok{                   fun\_control }\OperatorTok{=}\NormalTok{ fun\_control,}
\NormalTok{                   design\_control}\OperatorTok{=}\NormalTok{\{}\StringTok{"init\_size"}\NormalTok{: INIT\_SIZE\},}
\NormalTok{                   surrogate\_control}\OperatorTok{=}\NormalTok{\{}\StringTok{"noise"}\NormalTok{: }\VariableTok{True}\NormalTok{,}
                                      \StringTok{"min\_theta"}\NormalTok{: }\OperatorTok{{-}}\DecValTok{4}\NormalTok{,}
                                      \StringTok{"max\_theta"}\NormalTok{: }\DecValTok{3}\NormalTok{,}
                                      \StringTok{"n\_theta"}\NormalTok{: }\BuiltInTok{len}\NormalTok{(var\_name),}
                                      \StringTok{"model\_fun\_evals"}\NormalTok{: }\DecValTok{10\_000}\NormalTok{,}
\NormalTok{                                      \})}
\NormalTok{spot\_tuner.run()}
\end{Highlighting}
\end{Shaded}

\begin{verbatim}

config: {'l1': 256, 'epochs': 4096, 'batch_size': 32, 'act_fn': ReLU(), 'optimizer': 'AdamW', 'dropout_prob': 0.10939527466721133, 'lr_mult': 4.211776903906428, 'patience': 16, 'initialization': 'Default'}
\end{verbatim}

\begin{verbatim}
┏━━━━━━━━━━━━━━━━━━━━━━━━━━━┳━━━━━━━━━━━━━━━━━━━━━━━━━━━┓
┃      Validate metric      ┃       DataLoader 0        ┃
┡━━━━━━━━━━━━━━━━━━━━━━━━━━━╇━━━━━━━━━━━━━━━━━━━━━━━━━━━┩
│         hp_metric         │     2.263709545135498     │
│          val_acc          │     0.268551230430603     │
│         val_loss          │     2.263709545135498     │
│        valid_mapk         │    0.3537808656692505     │
└───────────────────────────┴───────────────────────────┘
\end{verbatim}

\begin{verbatim}

config: {'l1': 32, 'epochs': 128, 'batch_size': 256, 'act_fn': LeakyReLU(), 'optimizer': 'Adamax', 'dropout_prob': 0.012926647388264517, 'lr_mult': 0.832718394912432, 'patience': 8, 'initialization': 'Kaiming'}
\end{verbatim}

\begin{verbatim}
┏━━━━━━━━━━━━━━━━━━━━━━━━━━━┳━━━━━━━━━━━━━━━━━━━━━━━━━━━┓
┃      Validate metric      ┃       DataLoader 0        ┃
┡━━━━━━━━━━━━━━━━━━━━━━━━━━━╇━━━━━━━━━━━━━━━━━━━━━━━━━━━┩
│         hp_metric         │    2.2617576122283936     │
│          val_acc          │    0.2720848023891449     │
│         val_loss          │    2.2617576122283936     │
│        valid_mapk         │    0.3213372826576233     │
└───────────────────────────┴───────────────────────────┘
\end{verbatim}

\begin{verbatim}

config: {'l1': 128, 'epochs': 256, 'batch_size': 8, 'act_fn': Swish(), 'optimizer': 'NAdam', 'dropout_prob': 0.22086376796923401, 'lr_mult': 7.65501078489161, 'patience': 64, 'initialization': 'Xavier'}
\end{verbatim}

\begin{verbatim}
┏━━━━━━━━━━━━━━━━━━━━━━━━━━━┳━━━━━━━━━━━━━━━━━━━━━━━━━━━┓
┃      Validate metric      ┃       DataLoader 0        ┃
┡━━━━━━━━━━━━━━━━━━━━━━━━━━━╇━━━━━━━━━━━━━━━━━━━━━━━━━━━┩
│         hp_metric         │     2.451167345046997     │
│          val_acc          │    0.09187278896570206    │
│         val_loss          │     2.451167345046997     │
│        valid_mapk         │    0.16377314925193787    │
└───────────────────────────┴───────────────────────────┘
\end{verbatim}

\begin{verbatim}

config: {'l1': 64, 'epochs': 512, 'batch_size': 16, 'act_fn': Sigmoid(), 'optimizer': 'Adam', 'dropout_prob': 0.1890928563375006, 'lr_mult': 2.3450676871382794, 'patience': 32, 'initialization': 'Kaiming'}
\end{verbatim}

\begin{verbatim}
┏━━━━━━━━━━━━━━━━━━━━━━━━━━━┳━━━━━━━━━━━━━━━━━━━━━━━━━━━┓
┃      Validate metric      ┃       DataLoader 0        ┃
┡━━━━━━━━━━━━━━━━━━━━━━━━━━━╇━━━━━━━━━━━━━━━━━━━━━━━━━━━┩
│         hp_metric         │    2.3177614212036133     │
│          val_acc          │    0.22968198359012604    │
│         val_loss          │    2.3177614212036133     │
│        valid_mapk         │    0.2924031913280487     │
└───────────────────────────┴───────────────────────────┘
\end{verbatim}

\begin{verbatim}

config: {'l1': 64, 'epochs': 4096, 'batch_size': 64, 'act_fn': ReLU(), 'optimizer': 'Adamax', 'dropout_prob': 0.0708380794924471, 'lr_mult': 9.528945328733357, 'patience': 4, 'initialization': 'Xavier'}
\end{verbatim}

\begin{verbatim}
┏━━━━━━━━━━━━━━━━━━━━━━━━━━━┳━━━━━━━━━━━━━━━━━━━━━━━━━━━┓
┃      Validate metric      ┃       DataLoader 0        ┃
┡━━━━━━━━━━━━━━━━━━━━━━━━━━━╇━━━━━━━━━━━━━━━━━━━━━━━━━━━┩
│         hp_metric         │     2.25834321975708      │
│          val_acc          │    0.2614840865135193     │
│         val_loss          │     2.25834321975708      │
│        valid_mapk         │    0.36971449851989746    │
└───────────────────────────┴───────────────────────────┘
\end{verbatim}

\begin{verbatim}

config: {'l1': 32, 'epochs': 4096, 'batch_size': 128, 'act_fn': ReLU(), 'optimizer': 'AdamW', 'dropout_prob': 0.0, 'lr_mult': 10.0, 'patience': 16, 'initialization': 'Default'}
\end{verbatim}

\begin{verbatim}
┏━━━━━━━━━━━━━━━━━━━━━━━━━━━┳━━━━━━━━━━━━━━━━━━━━━━━━━━━┓
┃      Validate metric      ┃       DataLoader 0        ┃
┡━━━━━━━━━━━━━━━━━━━━━━━━━━━╇━━━━━━━━━━━━━━━━━━━━━━━━━━━┩
│         hp_metric         │    2.2992091178894043     │
│          val_acc          │    0.23674911260604858    │
│         val_loss          │    2.2992091178894043     │
│        valid_mapk         │    0.3636349141597748     │
└───────────────────────────┴───────────────────────────┘
\end{verbatim}

\begin{verbatim}
spotPython tuning: 2.25834321975708 [----------] 4.20% 
\end{verbatim}

\begin{verbatim}

config: {'l1': 32, 'epochs': 4096, 'batch_size': 64, 'act_fn': ReLU(), 'optimizer': 'Adam', 'dropout_prob': 0.08021662034280548, 'lr_mult': 3.6585398760718895, 'patience': 64, 'initialization': 'Xavier'}
\end{verbatim}

\begin{verbatim}
┏━━━━━━━━━━━━━━━━━━━━━━━━━━━┳━━━━━━━━━━━━━━━━━━━━━━━━━━━┓
┃      Validate metric      ┃       DataLoader 0        ┃
┡━━━━━━━━━━━━━━━━━━━━━━━━━━━╇━━━━━━━━━━━━━━━━━━━━━━━━━━━┩
│         hp_metric         │    2.2508015632629395     │
│          val_acc          │    0.27915194630622864    │
│         val_loss          │    2.2508015632629395     │
│        valid_mapk         │    0.35104164481163025    │
└───────────────────────────┴───────────────────────────┘
\end{verbatim}

\begin{verbatim}
spotPython tuning: 2.2508015632629395 [###-------] 30.73% 
\end{verbatim}

\begin{verbatim}

config: {'l1': 32, 'epochs': 512, 'batch_size': 128, 'act_fn': ReLU(), 'optimizer': 'AdamW', 'dropout_prob': 0.11515978605575264, 'lr_mult': 1.4629919730875696, 'patience': 16, 'initialization': 'Kaiming'}
\end{verbatim}

\begin{verbatim}
┏━━━━━━━━━━━━━━━━━━━━━━━━━━━┳━━━━━━━━━━━━━━━━━━━━━━━━━━━┓
┃      Validate metric      ┃       DataLoader 0        ┃
┡━━━━━━━━━━━━━━━━━━━━━━━━━━━╇━━━━━━━━━━━━━━━━━━━━━━━━━━━┩
│         hp_metric         │     2.285112142562866     │
│          val_acc          │    0.2226148396730423     │
│         val_loss          │     2.285112142562866     │
│        valid_mapk         │    0.3453253507614136     │
└───────────────────────────┴───────────────────────────┘
\end{verbatim}

\begin{verbatim}
spotPython tuning: 2.2508015632629395 [####------] 42.49% 
\end{verbatim}

\begin{verbatim}

config: {'l1': 64, 'epochs': 256, 'batch_size': 64, 'act_fn': ReLU(), 'optimizer': 'Adam', 'dropout_prob': 0.25, 'lr_mult': 0.1, 'patience': 8, 'initialization': 'Xavier'}
\end{verbatim}

\begin{verbatim}
┏━━━━━━━━━━━━━━━━━━━━━━━━━━━┳━━━━━━━━━━━━━━━━━━━━━━━━━━━┓
┃      Validate metric      ┃       DataLoader 0        ┃
┡━━━━━━━━━━━━━━━━━━━━━━━━━━━╇━━━━━━━━━━━━━━━━━━━━━━━━━━━┩
│         hp_metric         │     2.387507200241089     │
│          val_acc          │    0.06713780760765076    │
│         val_loss          │     2.387507200241089     │
│        valid_mapk         │    0.16718751192092896    │
└───────────────────────────┴───────────────────────────┘
\end{verbatim}

\begin{verbatim}
spotPython tuning: 2.2508015632629395 [#####-----] 53.64% 
\end{verbatim}

\begin{verbatim}

config: {'l1': 32, 'epochs': 8192, 'batch_size': 128, 'act_fn': ReLU(), 'optimizer': 'NAdam', 'dropout_prob': 0.07912235457426961, 'lr_mult': 4.163993340890585, 'patience': 64, 'initialization': 'Xavier'}
\end{verbatim}

\begin{verbatim}
┏━━━━━━━━━━━━━━━━━━━━━━━━━━━┳━━━━━━━━━━━━━━━━━━━━━━━━━━━┓
┃      Validate metric      ┃       DataLoader 0        ┃
┡━━━━━━━━━━━━━━━━━━━━━━━━━━━╇━━━━━━━━━━━━━━━━━━━━━━━━━━━┩
│         hp_metric         │    2.2326242923736572     │
│          val_acc          │    0.3074204921722412     │
│         val_loss          │    2.2326242923736572     │
│        valid_mapk         │    0.39719972014427185    │
└───────────────────────────┴───────────────────────────┘
\end{verbatim}

\begin{verbatim}
spotPython tuning: 2.2326242923736572 [########--] 76.50% 
\end{verbatim}

\begin{verbatim}

config: {'l1': 32, 'epochs': 8192, 'batch_size': 256, 'act_fn': ReLU(), 'optimizer': 'NAdam', 'dropout_prob': 0.07805510131337337, 'lr_mult': 1.4621425750875474, 'patience': 32, 'initialization': 'Kaiming'}
\end{verbatim}

\begin{verbatim}
┏━━━━━━━━━━━━━━━━━━━━━━━━━━━┳━━━━━━━━━━━━━━━━━━━━━━━━━━━┓
┃      Validate metric      ┃       DataLoader 0        ┃
┡━━━━━━━━━━━━━━━━━━━━━━━━━━━╇━━━━━━━━━━━━━━━━━━━━━━━━━━━┩
│         hp_metric         │    2.3005590438842773     │
│          val_acc          │    0.23674911260604858    │
│         val_loss          │    2.3005590438842773     │
│        valid_mapk         │    0.3299093246459961     │
└───────────────────────────┴───────────────────────────┘
\end{verbatim}

\begin{verbatim}
spotPython tuning: 2.2326242923736572 [########--] 82.07% 
\end{verbatim}

\begin{verbatim}

config: {'l1': 64, 'epochs': 128, 'batch_size': 256, 'act_fn': LeakyReLU(), 'optimizer': 'AdamW', 'dropout_prob': 0.076487528469829, 'lr_mult': 3.074542097312815, 'patience': 16, 'initialization': 'Xavier'}
\end{verbatim}

\begin{verbatim}
┏━━━━━━━━━━━━━━━━━━━━━━━━━━━┳━━━━━━━━━━━━━━━━━━━━━━━━━━━┓
┃      Validate metric      ┃       DataLoader 0        ┃
┡━━━━━━━━━━━━━━━━━━━━━━━━━━━╇━━━━━━━━━━━━━━━━━━━━━━━━━━━┩
│         hp_metric         │    2.2523937225341797     │
│          val_acc          │    0.2862190902233124     │
│         val_loss          │    2.2523937225341797     │
│        valid_mapk         │    0.3655478358268738     │
└───────────────────────────┴───────────────────────────┘
\end{verbatim}

\begin{verbatim}
spotPython tuning: 2.2326242923736572 [#########-] 90.08% 
\end{verbatim}

\begin{verbatim}

config: {'l1': 128, 'epochs': 4096, 'batch_size': 64, 'act_fn': Tanh(), 'optimizer': 'AdamW', 'dropout_prob': 0.09458258775992502, 'lr_mult': 1.186247604300184, 'patience': 64, 'initialization': 'Xavier'}
\end{verbatim}

\begin{verbatim}
┏━━━━━━━━━━━━━━━━━━━━━━━━━━━┳━━━━━━━━━━━━━━━━━━━━━━━━━━━┓
┃      Validate metric      ┃       DataLoader 0        ┃
┡━━━━━━━━━━━━━━━━━━━━━━━━━━━╇━━━━━━━━━━━━━━━━━━━━━━━━━━━┩
│         hp_metric         │    2.2571563720703125     │
│          val_acc          │    0.27915194630622864    │
│         val_loss          │    2.2571563720703125     │
│        valid_mapk         │    0.38902392983436584    │
└───────────────────────────┴───────────────────────────┘
\end{verbatim}

\begin{verbatim}
spotPython tuning: 2.2326242923736572 [##########] 100.00% Done...
\end{verbatim}

\begin{verbatim}
<spotPython.spot.spot.Spot at 0x2854daf50>
\end{verbatim}

\hypertarget{sec-tensorboard-31}{%
\section{Step 9: Tensorboard}\label{sec-tensorboard-31}}

The textual output shown in the console (or code cell) can be visualized
with Tensorboard.

\begin{Shaded}
\begin{Highlighting}[]
\NormalTok{tensorboard {-}{-}logdir="runs/"}
\end{Highlighting}
\end{Shaded}

Further information can be found in the
\href{https://lightning.ai/docs/pytorch/stable/api/lightning.pytorch.loggers.tensorboard.html}{PyTorch
Lightning documentation} for Tensorboard.

\hypertarget{sec-results-31}{%
\section{Step 10: Results}\label{sec-results-31}}

After the hyperparameter tuning run is finished, the results can be
analyzed as described in Section~\ref{sec-results-14}.

\begin{Shaded}
\begin{Highlighting}[]
\NormalTok{spot\_tuner.plot\_progress(log\_y}\OperatorTok{=}\VariableTok{False}\NormalTok{,}
\NormalTok{    filename}\OperatorTok{=}\StringTok{"./figures/"} \OperatorTok{+}\NormalTok{ experiment\_name}\OperatorTok{+}\StringTok{"\_progress.png"}\NormalTok{)}
\end{Highlighting}
\end{Shaded}

\begin{figure}[H]

{\centering \includegraphics{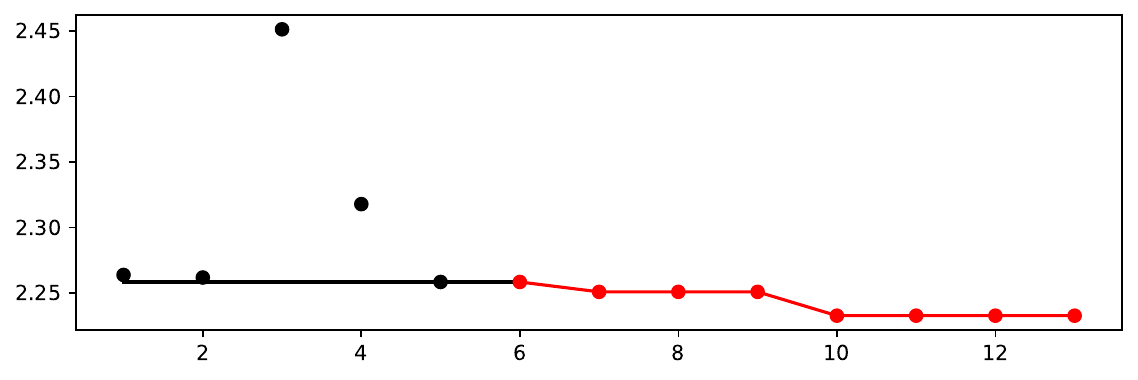}

}

\caption{Progress plot. \emph{Black} dots denote results from the
initial design. \emph{Red} dots illustrate the improvement found by the
surrogate model based optimization.}

\end{figure}

\begin{Shaded}
\begin{Highlighting}[]
\ImportTok{from}\NormalTok{ spotPython.utils.eda }\ImportTok{import}\NormalTok{ gen\_design\_table}
\BuiltInTok{print}\NormalTok{(gen\_design\_table(fun\_control}\OperatorTok{=}\NormalTok{fun\_control, spot}\OperatorTok{=}\NormalTok{spot\_tuner))}
\end{Highlighting}
\end{Shaded}

\begin{verbatim}
| name           | type   | default   |   lower |   upper |               tuned | transform             |   importance | stars   |
|----------------|--------|-----------|---------|---------|---------------------|-----------------------|--------------|---------|
| l1             | int    | 3         |     5.0 |     8.0 |                 5.0 | transform_power_2_int |         0.00 |         |
| epochs         | int    | 4         |     6.0 |    13.0 |                13.0 | transform_power_2_int |         0.00 |         |
| batch_size     | int    | 4         |     2.0 |     8.0 |                 7.0 | transform_power_2_int |         0.00 |         |
| act_fn         | factor | ReLU      |     0.0 |     5.0 |                 2.0 | None                  |         0.00 |         |
| optimizer      | factor | SGD       |     0.0 |     3.0 |                 3.0 | None                  |         0.00 |         |
| dropout_prob   | float  | 0.01      |     0.0 |    0.25 | 0.07912235457426961 | None                  |         1.62 | *       |
| lr_mult        | float  | 1.0       |     0.1 |    10.0 |   4.163993340890585 | None                  |         0.00 |         |
| patience       | int    | 2         |     2.0 |     6.0 |                 6.0 | transform_power_2_int |         0.00 |         |
| initialization | factor | Default   |     0.0 |     2.0 |                 2.0 | None                  |       100.00 | ***     |
\end{verbatim}

\begin{Shaded}
\begin{Highlighting}[]
\NormalTok{spot\_tuner.plot\_importance(threshold}\OperatorTok{=}\FloatTok{0.025}\NormalTok{,}
\NormalTok{    filename}\OperatorTok{=}\StringTok{"./figures/"} \OperatorTok{+}\NormalTok{ experiment\_name}\OperatorTok{+}\StringTok{"\_importance.png"}\NormalTok{)}
\end{Highlighting}
\end{Shaded}

\begin{figure}[H]

{\centering \includegraphics{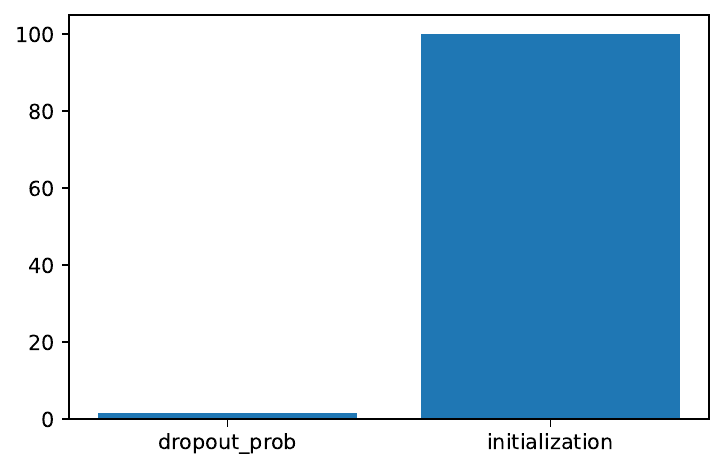}

}

\caption{Variable importance plot, threshold 0.025.}

\end{figure}

\hypertarget{sec-get-spot-results-31}{%
\subsection{Get the Tuned Architecture}\label{sec-get-spot-results-31}}

\begin{Shaded}
\begin{Highlighting}[]
\ImportTok{from}\NormalTok{ spotPython.light.utils }\ImportTok{import}\NormalTok{ get\_tuned\_architecture}
\NormalTok{config }\OperatorTok{=}\NormalTok{ get\_tuned\_architecture(spot\_tuner, fun\_control)}
\end{Highlighting}
\end{Shaded}

\begin{itemize}
\tightlist
\item
  Test on the full data set
\end{itemize}

\begin{Shaded}
\begin{Highlighting}[]
\ImportTok{from}\NormalTok{ spotPython.light.traintest }\ImportTok{import}\NormalTok{ test\_model}
\NormalTok{test\_model(config, fun\_control)}
\end{Highlighting}
\end{Shaded}

\begin{verbatim}
┏━━━━━━━━━━━━━━━━━━━━━━━━━━━┳━━━━━━━━━━━━━━━━━━━━━━━━━━━┓
┃        Test metric        ┃       DataLoader 0        ┃
┡━━━━━━━━━━━━━━━━━━━━━━━━━━━╇━━━━━━━━━━━━━━━━━━━━━━━━━━━┩
│         hp_metric         │    2.0217232704162598     │
│      test_mapk_epoch      │    0.5595039129257202     │
│          val_acc          │    0.5190947651863098     │
│         val_loss          │    2.0217232704162598     │
└───────────────────────────┴───────────────────────────┘
\end{verbatim}

\begin{verbatim}
(2.0217232704162598, 0.5190947651863098)
\end{verbatim}

\begin{Shaded}
\begin{Highlighting}[]
\ImportTok{from}\NormalTok{ spotPython.light.traintest }\ImportTok{import}\NormalTok{ load\_light\_from\_checkpoint}

\NormalTok{model\_loaded }\OperatorTok{=}\NormalTok{ load\_light\_from\_checkpoint(config, fun\_control)}
\end{Highlighting}
\end{Shaded}

\begin{verbatim}
Loading model from runs/lightning_logs/32_8192_128_ReLU()_NAdam_0.07912235457426961_4.163993340890585_64_Xavier_TEST/checkpoints/last.ckpt
\end{verbatim}

\hypertarget{cross-validation-with-lightning}{%
\subsection{Cross Validation With
Lightning}\label{cross-validation-with-lightning}}

\begin{itemize}
\tightlist
\item
  The \texttt{KFold} class from \texttt{sklearn.model\_selection} is
  used to generate the folds for cross-validation.
\item
  These mechanism is used to generate the folds for the final evaluation
  of the model.
\item
  The \texttt{CrossValidationDataModule} class
  \href{https://github.com/sequential-parameter-optimization/spotPython/blob/main/src/spotPython/light/crossvalidationdatamodule.py}{{[}SOURCE{]}}
  is used to generate the folds for the hyperparameter tuning process.
\item
  It is called from the \texttt{cv\_model} function
  \href{https://github.com/sequential-parameter-optimization/spotPython/blob/main/src/spotPython/light/traintest.py}{{[}SOURCE{]}}.
\end{itemize}

\begin{Shaded}
\begin{Highlighting}[]
\ImportTok{from}\NormalTok{ spotPython.light.traintest }\ImportTok{import}\NormalTok{ cv\_model}
\CommentTok{\# set the number of folds to 10}
\NormalTok{fun\_control[}\StringTok{"k\_folds"}\NormalTok{] }\OperatorTok{=} \DecValTok{10}
\NormalTok{cv\_model(config, fun\_control)}
\end{Highlighting}
\end{Shaded}

\begin{verbatim}
k: 0
Train Dataset Size: 636
Val Dataset Size: 71
\end{verbatim}

\begin{verbatim}
┏━━━━━━━━━━━━━━━━━━━━━━━━━━━┳━━━━━━━━━━━━━━━━━━━━━━━━━━━┓
┃      Validate metric      ┃       DataLoader 0        ┃
┡━━━━━━━━━━━━━━━━━━━━━━━━━━━╇━━━━━━━━━━━━━━━━━━━━━━━━━━━┩
│         hp_metric         │     2.169114351272583     │
│          val_acc          │    0.3661971688270569     │
│         val_loss          │     2.169114351272583     │
│        valid_mapk         │    0.43896713852882385    │
└───────────────────────────┴───────────────────────────┘
\end{verbatim}

\begin{verbatim}
train_model result: {'valid_mapk': 0.43896713852882385, 'val_loss': 2.169114351272583, 'val_acc': 0.3661971688270569, 'hp_metric': 2.169114351272583}
k: 1
Train Dataset Size: 636
Val Dataset Size: 71
\end{verbatim}

\begin{verbatim}
┏━━━━━━━━━━━━━━━━━━━━━━━━━━━┳━━━━━━━━━━━━━━━━━━━━━━━━━━━┓
┃      Validate metric      ┃       DataLoader 0        ┃
┡━━━━━━━━━━━━━━━━━━━━━━━━━━━╇━━━━━━━━━━━━━━━━━━━━━━━━━━━┩
│         hp_metric         │    2.2213735580444336     │
│          val_acc          │    0.3239436745643616     │
│         val_loss          │    2.2213735580444336     │
│        valid_mapk         │    0.4507042169570923     │
└───────────────────────────┴───────────────────────────┘
\end{verbatim}

\begin{verbatim}
train_model result: {'valid_mapk': 0.4507042169570923, 'val_loss': 2.2213735580444336, 'val_acc': 0.3239436745643616, 'hp_metric': 2.2213735580444336}
k: 2
Train Dataset Size: 636
Val Dataset Size: 71
\end{verbatim}

\begin{verbatim}
┏━━━━━━━━━━━━━━━━━━━━━━━━━━━┳━━━━━━━━━━━━━━━━━━━━━━━━━━━┓
┃      Validate metric      ┃       DataLoader 0        ┃
┡━━━━━━━━━━━━━━━━━━━━━━━━━━━╇━━━━━━━━━━━━━━━━━━━━━━━━━━━┩
│         hp_metric         │     2.305535078048706     │
│          val_acc          │    0.23943662643432617    │
│         val_loss          │     2.305535078048706     │
│        valid_mapk         │    0.2957746386528015     │
└───────────────────────────┴───────────────────────────┘
\end{verbatim}

\begin{verbatim}
train_model result: {'valid_mapk': 0.2957746386528015, 'val_loss': 2.305535078048706, 'val_acc': 0.23943662643432617, 'hp_metric': 2.305535078048706}
k: 3
Train Dataset Size: 636
Val Dataset Size: 71
\end{verbatim}

\begin{verbatim}
┏━━━━━━━━━━━━━━━━━━━━━━━━━━━┳━━━━━━━━━━━━━━━━━━━━━━━━━━━┓
┃      Validate metric      ┃       DataLoader 0        ┃
┡━━━━━━━━━━━━━━━━━━━━━━━━━━━╇━━━━━━━━━━━━━━━━━━━━━━━━━━━┩
│         hp_metric         │     2.282437801361084     │
│          val_acc          │    0.23943662643432617    │
│         val_loss          │     2.282437801361084     │
│        valid_mapk         │    0.34741783142089844    │
└───────────────────────────┴───────────────────────────┘
\end{verbatim}

\begin{verbatim}
train_model result: {'valid_mapk': 0.34741783142089844, 'val_loss': 2.282437801361084, 'val_acc': 0.23943662643432617, 'hp_metric': 2.282437801361084}
k: 4
Train Dataset Size: 636
Val Dataset Size: 71
\end{verbatim}

\begin{verbatim}
┏━━━━━━━━━━━━━━━━━━━━━━━━━━━┳━━━━━━━━━━━━━━━━━━━━━━━━━━━┓
┃      Validate metric      ┃       DataLoader 0        ┃
┡━━━━━━━━━━━━━━━━━━━━━━━━━━━╇━━━━━━━━━━━━━━━━━━━━━━━━━━━┩
│         hp_metric         │    2.3464431762695312     │
│          val_acc          │    0.18309858441352844    │
│         val_loss          │    2.3464431762695312     │
│        valid_mapk         │    0.28169015049934387    │
└───────────────────────────┴───────────────────────────┘
\end{verbatim}

\begin{verbatim}
train_model result: {'valid_mapk': 0.28169015049934387, 'val_loss': 2.3464431762695312, 'val_acc': 0.18309858441352844, 'hp_metric': 2.3464431762695312}
k: 5
Train Dataset Size: 636
Val Dataset Size: 71
\end{verbatim}

\begin{verbatim}
┏━━━━━━━━━━━━━━━━━━━━━━━━━━━┳━━━━━━━━━━━━━━━━━━━━━━━━━━━┓
┃      Validate metric      ┃       DataLoader 0        ┃
┡━━━━━━━━━━━━━━━━━━━━━━━━━━━╇━━━━━━━━━━━━━━━━━━━━━━━━━━━┩
│         hp_metric         │    2.2693288326263428     │
│          val_acc          │    0.26760563254356384    │
│         val_loss          │    2.2693288326263428     │
│        valid_mapk         │    0.3779342770576477     │
└───────────────────────────┴───────────────────────────┘
\end{verbatim}

\begin{verbatim}
train_model result: {'valid_mapk': 0.3779342770576477, 'val_loss': 2.2693288326263428, 'val_acc': 0.26760563254356384, 'hp_metric': 2.2693288326263428}
k: 6
Train Dataset Size: 636
Val Dataset Size: 71
\end{verbatim}

\begin{verbatim}
┏━━━━━━━━━━━━━━━━━━━━━━━━━━━┳━━━━━━━━━━━━━━━━━━━━━━━━━━━┓
┃      Validate metric      ┃       DataLoader 0        ┃
┡━━━━━━━━━━━━━━━━━━━━━━━━━━━╇━━━━━━━━━━━━━━━━━━━━━━━━━━━┩
│         hp_metric         │    2.3574538230895996     │
│          val_acc          │    0.18309858441352844    │
│         val_loss          │    2.3574538230895996     │
│        valid_mapk         │    0.26056337356567383    │
└───────────────────────────┴───────────────────────────┘
\end{verbatim}

\begin{verbatim}
train_model result: {'valid_mapk': 0.26056337356567383, 'val_loss': 2.3574538230895996, 'val_acc': 0.18309858441352844, 'hp_metric': 2.3574538230895996}
k: 7
Train Dataset Size: 637
Val Dataset Size: 70
\end{verbatim}

\begin{verbatim}
┏━━━━━━━━━━━━━━━━━━━━━━━━━━━┳━━━━━━━━━━━━━━━━━━━━━━━━━━━┓
┃      Validate metric      ┃       DataLoader 0        ┃
┡━━━━━━━━━━━━━━━━━━━━━━━━━━━╇━━━━━━━━━━━━━━━━━━━━━━━━━━━┩
│         hp_metric         │    2.2719171047210693     │
│          val_acc          │    0.2571428716182709     │
│         val_loss          │    2.2719171047210693     │
│        valid_mapk         │    0.3499999940395355     │
└───────────────────────────┴───────────────────────────┘
\end{verbatim}

\begin{verbatim}
train_model result: {'valid_mapk': 0.3499999940395355, 'val_loss': 2.2719171047210693, 'val_acc': 0.2571428716182709, 'hp_metric': 2.2719171047210693}
k: 8
Train Dataset Size: 637
Val Dataset Size: 70
\end{verbatim}

\begin{verbatim}
┏━━━━━━━━━━━━━━━━━━━━━━━━━━━┳━━━━━━━━━━━━━━━━━━━━━━━━━━━┓
┃      Validate metric      ┃       DataLoader 0        ┃
┡━━━━━━━━━━━━━━━━━━━━━━━━━━━╇━━━━━━━━━━━━━━━━━━━━━━━━━━━┩
│         hp_metric         │    2.3715577125549316     │
│          val_acc          │    0.17142857611179352    │
│         val_loss          │    2.3715577125549316     │
│        valid_mapk         │    0.3047619163990021     │
└───────────────────────────┴───────────────────────────┘
\end{verbatim}

\begin{verbatim}
train_model result: {'valid_mapk': 0.3047619163990021, 'val_loss': 2.3715577125549316, 'val_acc': 0.17142857611179352, 'hp_metric': 2.3715577125549316}
k: 9
Train Dataset Size: 637
Val Dataset Size: 70
\end{verbatim}

\begin{verbatim}
┏━━━━━━━━━━━━━━━━━━━━━━━━━━━┳━━━━━━━━━━━━━━━━━━━━━━━━━━━┓
┃      Validate metric      ┃       DataLoader 0        ┃
┡━━━━━━━━━━━━━━━━━━━━━━━━━━━╇━━━━━━━━━━━━━━━━━━━━━━━━━━━┩
│         hp_metric         │     2.270128011703491     │
│          val_acc          │    0.24285714328289032    │
│         val_loss          │     2.270128011703491     │
│        valid_mapk         │    0.3857142925262451     │
└───────────────────────────┴───────────────────────────┘
\end{verbatim}

\begin{verbatim}
train_model result: {'valid_mapk': 0.3857142925262451, 'val_loss': 2.270128011703491, 'val_acc': 0.24285714328289032, 'hp_metric': 2.270128011703491}
\end{verbatim}

\begin{verbatim}
0.3493527829647064
\end{verbatim}

\begin{tcolorbox}[enhanced jigsaw, left=2mm, title=\textcolor{quarto-callout-note-color}{\faInfo}\hspace{0.5em}{Note: Evaluation for the Final Comaprison}, bottomrule=.15mm, titlerule=0mm, breakable, rightrule=.15mm, toprule=.15mm, coltitle=black, colbacktitle=quarto-callout-note-color!10!white, leftrule=.75mm, arc=.35mm, colframe=quarto-callout-note-color-frame, bottomtitle=1mm, colback=white, opacitybacktitle=0.6, toptitle=1mm, opacityback=0]

\begin{itemize}
\tightlist
\item
  This is the evaluation that will be used in the comparison.
\end{itemize}

\end{tcolorbox}

\hypertarget{detailed-hyperparameter-plots-7}{%
\subsection{Detailed Hyperparameter
Plots}\label{detailed-hyperparameter-plots-7}}

\begin{Shaded}
\begin{Highlighting}[]
\NormalTok{filename }\OperatorTok{=} \StringTok{"./figures/"} \OperatorTok{+}\NormalTok{ experiment\_name}
\NormalTok{spot\_tuner.plot\_important\_hyperparameter\_contour(filename}\OperatorTok{=}\NormalTok{filename)}
\end{Highlighting}
\end{Shaded}

\begin{verbatim}
dropout_prob:  1.6176864689332775
initialization:  100.0
\end{verbatim}

\begin{figure}[H]

{\centering \includegraphics{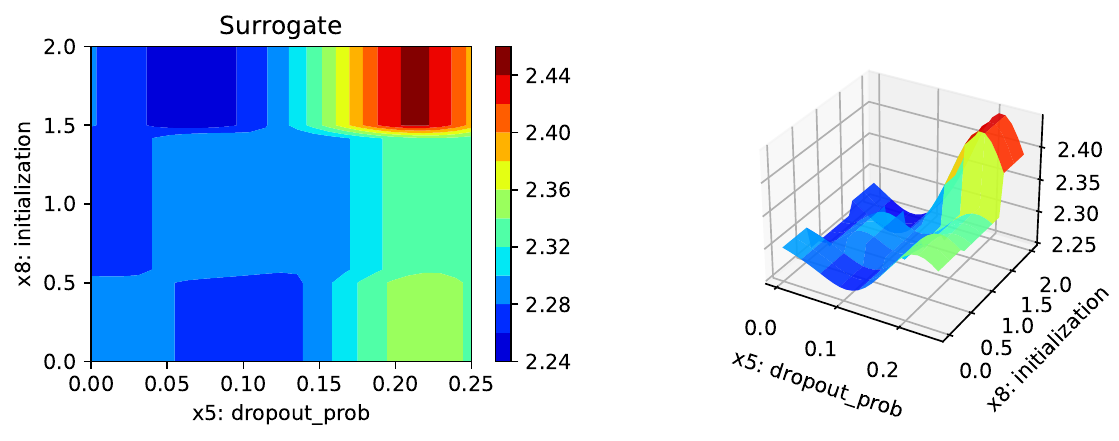}

}

\caption{Contour plots.}

\end{figure}

\hypertarget{parallel-coordinates-plot-5}{%
\subsection{Parallel Coordinates
Plot}\label{parallel-coordinates-plot-5}}

\begin{Shaded}
\begin{Highlighting}[]
\NormalTok{spot\_tuner.parallel\_plot()}
\end{Highlighting}
\end{Shaded}

\begin{verbatim}
Unable to display output for mime type(s): text/html
\end{verbatim}

Parallel coordinates plots

\begin{verbatim}
Unable to display output for mime type(s): text/html
\end{verbatim}

\hypertarget{plot-all-combinations-of-hyperparameters-6}{%
\subsection{Plot all Combinations of
Hyperparameters}\label{plot-all-combinations-of-hyperparameters-6}}

\begin{itemize}
\tightlist
\item
  Warning: this may take a while.
\end{itemize}

\begin{Shaded}
\begin{Highlighting}[]
\NormalTok{PLOT\_ALL }\OperatorTok{=} \VariableTok{False}
\ControlFlowTok{if}\NormalTok{ PLOT\_ALL:}
\NormalTok{    n }\OperatorTok{=}\NormalTok{ spot\_tuner.k}
    \ControlFlowTok{for}\NormalTok{ i }\KeywordTok{in} \BuiltInTok{range}\NormalTok{(n}\OperatorTok{{-}}\DecValTok{1}\NormalTok{):}
        \ControlFlowTok{for}\NormalTok{ j }\KeywordTok{in} \BuiltInTok{range}\NormalTok{(i}\OperatorTok{+}\DecValTok{1}\NormalTok{, n):}
\NormalTok{            spot\_tuner.plot\_contour(i}\OperatorTok{=}\NormalTok{i, j}\OperatorTok{=}\NormalTok{j, min\_z}\OperatorTok{=}\NormalTok{min\_z, max\_z }\OperatorTok{=}\NormalTok{ max\_z)}
\end{Highlighting}
\end{Shaded}

\hypertarget{visualizing-the-activation-distribution}{%
\subsection{Visualizing the Activation
Distribution}\label{visualizing-the-activation-distribution}}

\begin{tcolorbox}[enhanced jigsaw, left=2mm, title=\textcolor{quarto-callout-note-color}{\faInfo}\hspace{0.5em}{Reference:}, bottomrule=.15mm, titlerule=0mm, breakable, rightrule=.15mm, toprule=.15mm, coltitle=black, colbacktitle=quarto-callout-note-color!10!white, leftrule=.75mm, arc=.35mm, colframe=quarto-callout-note-color-frame, bottomtitle=1mm, colback=white, opacitybacktitle=0.6, toptitle=1mm, opacityback=0]

\begin{itemize}
\tightlist
\item
  The following code is based on
  \href{https://lightning.ai/docs/pytorch/stable/notebooks/course_UvA-DL/02-activation-functions.html}{{[}PyTorch
  Lightning TUTORIAL 2: ACTIVATION FUNCTIONS{]}}, Author: Phillip Lippe,
  License: \href{https://creativecommons.org/licenses/by-sa/3.0/}{{[}CC
  BY-SA{]}}, Generated: 2023-03-15T09:52:39.179933.
\end{itemize}

\end{tcolorbox}

After we have trained the models, we can look at the actual activation
values that find inside the model. For instance, how many neurons are
set to zero in ReLU? Where do we find most values in Tanh? To answer
these questions, we can write a simple function which takes a trained
model, applies it to a batch of images, and plots the histogram of the
activations inside the network:

\begin{Shaded}
\begin{Highlighting}[]
\ImportTok{from}\NormalTok{ spotPython.torch.activation }\ImportTok{import}\NormalTok{ Sigmoid, Tanh, ReLU, LeakyReLU, ELU, Swish}
\NormalTok{act\_fn\_by\_name }\OperatorTok{=}\NormalTok{ \{}\StringTok{"sigmoid"}\NormalTok{: Sigmoid, }\StringTok{"tanh"}\NormalTok{: Tanh, }\StringTok{"relu"}\NormalTok{: ReLU, }\StringTok{"leakyrelu"}\NormalTok{: LeakyReLU, }\StringTok{"elu"}\NormalTok{: ELU, }\StringTok{"swish"}\NormalTok{: Swish\}}
\end{Highlighting}
\end{Shaded}

\begin{Shaded}
\begin{Highlighting}[]
\ImportTok{from}\NormalTok{ spotPython.hyperparameters.values }\ImportTok{import}\NormalTok{ get\_one\_config\_from\_X}
\NormalTok{X }\OperatorTok{=}\NormalTok{ spot\_tuner.to\_all\_dim(spot\_tuner.min\_X.reshape(}\DecValTok{1}\NormalTok{,}\OperatorTok{{-}}\DecValTok{1}\NormalTok{))}
\NormalTok{config }\OperatorTok{=}\NormalTok{ get\_one\_config\_from\_X(X, fun\_control)}
\NormalTok{model }\OperatorTok{=}\NormalTok{ fun\_control[}\StringTok{"core\_model"}\NormalTok{](}\OperatorTok{**}\NormalTok{config, \_L\_in}\OperatorTok{=}\DecValTok{64}\NormalTok{, \_L\_out}\OperatorTok{=}\DecValTok{11}\NormalTok{)}
\NormalTok{model}
\end{Highlighting}
\end{Shaded}

\begin{verbatim}
NetLightBase(
  (train_mapk): MAPK()
  (valid_mapk): MAPK()
  (test_mapk): MAPK()
  (layers): Sequential(
    (0): Linear(in_features=64, out_features=32, bias=True)
    (1): ReLU()
    (2): Dropout(p=0.07912235457426961, inplace=False)
    (3): Linear(in_features=32, out_features=16, bias=True)
    (4): ReLU()
    (5): Dropout(p=0.07912235457426961, inplace=False)
    (6): Linear(in_features=16, out_features=16, bias=True)
    (7): ReLU()
    (8): Dropout(p=0.07912235457426961, inplace=False)
    (9): Linear(in_features=16, out_features=8, bias=True)
    (10): ReLU()
    (11): Dropout(p=0.07912235457426961, inplace=False)
    (12): Linear(in_features=8, out_features=11, bias=True)
  )
)
\end{verbatim}

\begin{Shaded}
\begin{Highlighting}[]
\ImportTok{from}\NormalTok{ spotPython.utils.eda }\ImportTok{import}\NormalTok{ visualize\_activations}
\NormalTok{visualize\_activations(model, color}\OperatorTok{=}\SpecialStringTok{f"C}\SpecialCharTok{\{}\DecValTok{0}\SpecialCharTok{\}}\SpecialStringTok{"}\NormalTok{)}
\end{Highlighting}
\end{Shaded}

\begin{figure}[H]

{\centering \includegraphics{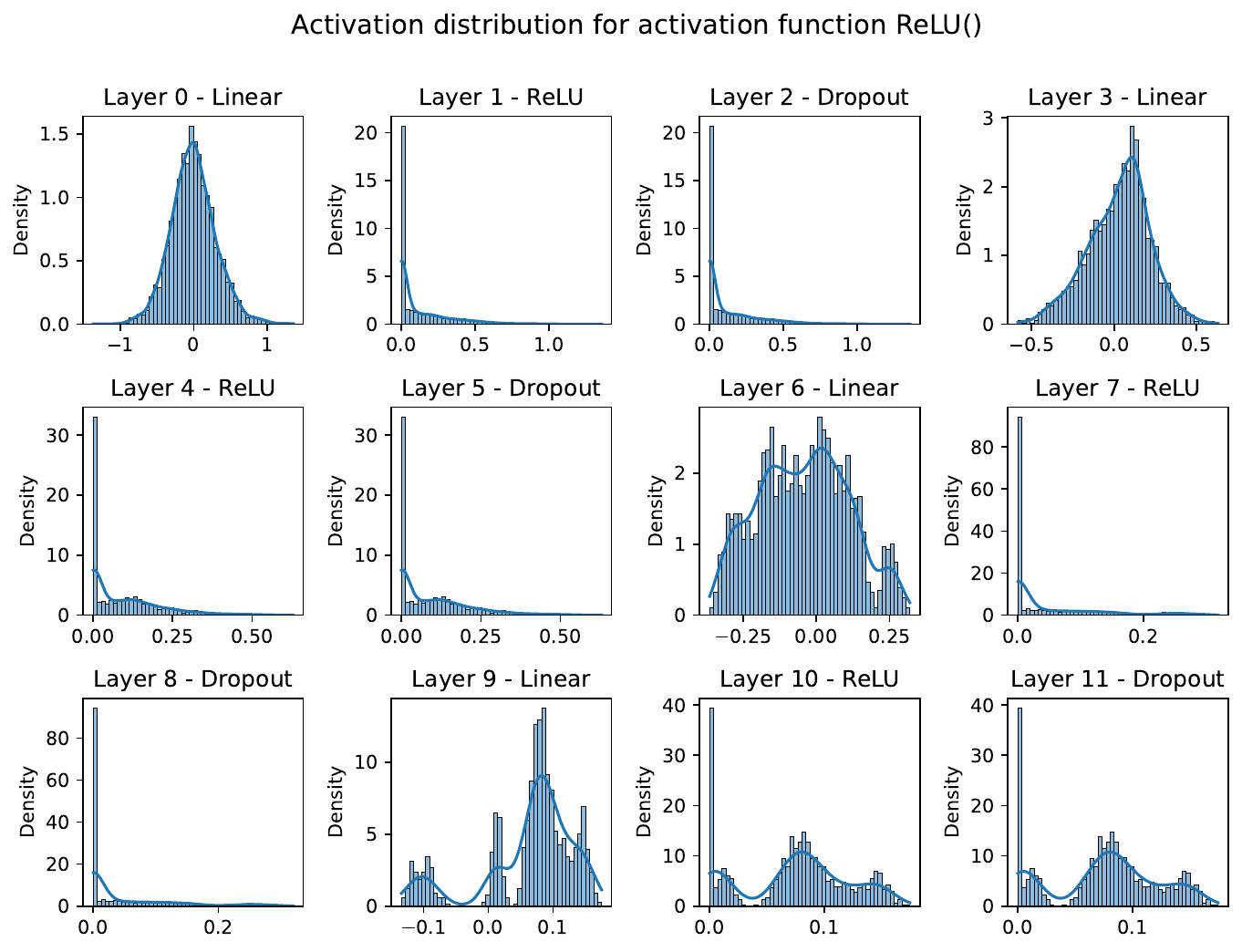}

}

\end{figure}

\hypertarget{submission}{%
\section{Submission}\label{submission}}

\begin{Shaded}
\begin{Highlighting}[]
\ImportTok{import}\NormalTok{ numpy }\ImportTok{as}\NormalTok{ np}
\ImportTok{import}\NormalTok{ pandas }\ImportTok{as}\NormalTok{ pd}
\ImportTok{from}\NormalTok{ sklearn.preprocessing }\ImportTok{import}\NormalTok{ OrdinalEncoder}
\end{Highlighting}
\end{Shaded}

\begin{Shaded}
\begin{Highlighting}[]
\ImportTok{import}\NormalTok{ pandas }\ImportTok{as}\NormalTok{ pd}
\ImportTok{from}\NormalTok{ sklearn.preprocessing }\ImportTok{import}\NormalTok{ OrdinalEncoder}
\NormalTok{train\_df }\OperatorTok{=}\NormalTok{ pd.read\_csv(}\StringTok{\textquotesingle{}./data/VBDP/train.csv\textquotesingle{}}\NormalTok{, index\_col}\OperatorTok{=}\DecValTok{0}\NormalTok{)}
\CommentTok{\# remove the id column}
\CommentTok{\# train\_df = train\_df.drop(columns=[\textquotesingle{}id\textquotesingle{}])}
\NormalTok{n\_samples }\OperatorTok{=}\NormalTok{ train\_df.shape[}\DecValTok{0}\NormalTok{]}
\NormalTok{n\_features }\OperatorTok{=}\NormalTok{ train\_df.shape[}\DecValTok{1}\NormalTok{] }\OperatorTok{{-}} \DecValTok{1}
\NormalTok{target\_column }\OperatorTok{=} \StringTok{"prognosis"}
\CommentTok{\# Encode our prognosis labels as integers for easier decoding later}
\NormalTok{enc }\OperatorTok{=}\NormalTok{ OrdinalEncoder()}
\NormalTok{y }\OperatorTok{=}\NormalTok{ enc.fit\_transform(train\_df[[target\_column]])}
\NormalTok{test\_df }\OperatorTok{=}\NormalTok{ pd.read\_csv(}\StringTok{\textquotesingle{}./data/VBDP/test.csv\textquotesingle{}}\NormalTok{, index\_col}\OperatorTok{=}\DecValTok{0}\NormalTok{)}
\NormalTok{test\_df}
\end{Highlighting}
\end{Shaded}

\begin{longtable}[]{@{}llllllllllllllllllllll@{}}
\toprule\noalign{}
& sudden\_fever & headache & mouth\_bleed & nose\_bleed & muscle\_pain &
joint\_pain & vomiting & rash & diarrhea & hypotension & ... &
lymph\_swells & breathing\_restriction & toe\_inflammation &
finger\_inflammation & lips\_irritation & itchiness & ulcers &
toenail\_loss & speech\_problem & bullseye\_rash \\
id & & & & & & & & & & & & & & & & & & & & & \\
\midrule\noalign{}
\endhead
\bottomrule\noalign{}
\endlastfoot
707 & 0.0 & 0.0 & 0.0 & 0.0 & 0.0 & 0.0 & 0.0 & 0.0 & 0.0 & 1.0 & ... &
0.0 & 0.0 & 0.0 & 0.0 & 0.0 & 0.0 & 0.0 & 0.0 & 0.0 & 0.0 \\
708 & 1.0 & 1.0 & 0.0 & 1.0 & 0.0 & 1.0 & 1.0 & 1.0 & 1.0 & 1.0 & ... &
0.0 & 0.0 & 0.0 & 0.0 & 0.0 & 0.0 & 0.0 & 0.0 & 0.0 & 0.0 \\
709 & 1.0 & 1.0 & 0.0 & 1.0 & 1.0 & 1.0 & 1.0 & 0.0 & 1.0 & 0.0 & ... &
0.0 & 0.0 & 0.0 & 0.0 & 0.0 & 1.0 & 0.0 & 0.0 & 0.0 & 0.0 \\
710 & 0.0 & 1.0 & 0.0 & 0.0 & 0.0 & 1.0 & 1.0 & 1.0 & 0.0 & 0.0 & ... &
0.0 & 0.0 & 0.0 & 0.0 & 0.0 & 0.0 & 0.0 & 0.0 & 0.0 & 0.0 \\
711 & 0.0 & 0.0 & 1.0 & 0.0 & 1.0 & 1.0 & 0.0 & 0.0 & 1.0 & 1.0 & ... &
0.0 & 0.0 & 0.0 & 0.0 & 0.0 & 0.0 & 0.0 & 0.0 & 0.0 & 0.0 \\
... & ... & ... & ... & ... & ... & ... & ... & ... & ... & ... & ... &
... & ... & ... & ... & ... & ... & ... & ... & ... & ... \\
1005 & 0.0 & 1.0 & 0.0 & 0.0 & 0.0 & 0.0 & 0.0 & 0.0 & 1.0 & 0.0 & ... &
0.0 & 0.0 & 0.0 & 0.0 & 0.0 & 0.0 & 0.0 & 0.0 & 0.0 & 0.0 \\
1006 & 1.0 & 0.0 & 1.0 & 0.0 & 1.0 & 1.0 & 0.0 & 1.0 & 1.0 & 1.0 & ... &
0.0 & 0.0 & 0.0 & 0.0 & 0.0 & 0.0 & 0.0 & 0.0 & 0.0 & 0.0 \\
1007 & 1.0 & 0.0 & 0.0 & 1.0 & 1.0 & 0.0 & 1.0 & 1.0 & 1.0 & 1.0 & ... &
1.0 & 1.0 & 1.0 & 1.0 & 1.0 & 0.0 & 0.0 & 0.0 & 0.0 & 0.0 \\
1008 & 1.0 & 0.0 & 1.0 & 1.0 & 1.0 & 0.0 & 1.0 & 0.0 & 0.0 & 0.0 & ... &
0.0 & 0.0 & 0.0 & 1.0 & 0.0 & 0.0 & 0.0 & 0.0 & 0.0 & 0.0 \\
1009 & 1.0 & 0.0 & 0.0 & 0.0 & 0.0 & 0.0 & 1.0 & 0.0 & 1.0 & 0.0 & ... &
0.0 & 0.0 & 0.0 & 0.0 & 0.0 & 0.0 & 0.0 & 0.0 & 0.0 & 0.0 \\
\end{longtable}

\begin{Shaded}
\begin{Highlighting}[]
\ImportTok{import}\NormalTok{ torch}
\NormalTok{X\_tensor }\OperatorTok{=}\NormalTok{ torch.Tensor(test\_df.values)}
\NormalTok{X\_tensor }\OperatorTok{=}\NormalTok{ X\_tensor.to(fun\_control[}\StringTok{"device"}\NormalTok{])}
\end{Highlighting}
\end{Shaded}

\begin{Shaded}
\begin{Highlighting}[]
\NormalTok{y }\OperatorTok{=}\NormalTok{ model\_loaded(X\_tensor)}
\NormalTok{y.shape}
\end{Highlighting}
\end{Shaded}

\begin{verbatim}
torch.Size([303, 11])
\end{verbatim}

\begin{Shaded}
\begin{Highlighting}[]
\CommentTok{\# convert the predictions to a numpy array}
\NormalTok{y }\OperatorTok{=}\NormalTok{ y.cpu().detach().numpy()}
\NormalTok{y}
\end{Highlighting}
\end{Shaded}

\begin{verbatim}
array([[5.41232845e-26, 1.27095045e-05, 0.00000000e+00, ...,
        0.00000000e+00, 1.54120984e-27, 1.93529959e-29],
       [9.99993920e-01, 6.03577928e-06, 3.03779888e-26, ...,
        6.81878993e-17, 5.46532127e-15, 3.95063810e-13],
       [1.46817777e-29, 1.97117316e-07, 9.99999762e-01, ...,
        4.56797316e-22, 1.94337737e-32, 8.91749244e-11],
       ...,
       [0.00000000e+00, 6.26950018e-16, 5.88089470e-16, ...,
        2.32772049e-17, 0.00000000e+00, 4.82207195e-23],
       [3.28134635e-15, 5.67918718e-02, 6.38429940e-01, ...,
        8.29307712e-04, 2.18537442e-21, 1.48154972e-02],
       [1.39392523e-25, 9.65945560e-07, 9.99998450e-01, ...,
        7.54383891e-15, 0.00000000e+00, 8.63872707e-09]], dtype=float32)
\end{verbatim}

\begin{Shaded}
\begin{Highlighting}[]
\NormalTok{test\_sorted\_prediction\_ids }\OperatorTok{=}\NormalTok{ np.argsort(}\OperatorTok{{-}}\NormalTok{y, axis}\OperatorTok{=}\DecValTok{1}\NormalTok{)}
\NormalTok{test\_top\_3\_prediction\_ids }\OperatorTok{=}\NormalTok{ test\_sorted\_prediction\_ids[:,:}\DecValTok{3}\NormalTok{]}
\NormalTok{original\_shape }\OperatorTok{=}\NormalTok{ test\_top\_3\_prediction\_ids.shape}
\NormalTok{test\_top\_3\_prediction }\OperatorTok{=}\NormalTok{ enc.inverse\_transform(test\_top\_3\_prediction\_ids.reshape(}\OperatorTok{{-}}\DecValTok{1}\NormalTok{, }\DecValTok{1}\NormalTok{))}
\NormalTok{test\_top\_3\_prediction }\OperatorTok{=}\NormalTok{ test\_top\_3\_prediction.reshape(original\_shape)}
\NormalTok{test\_df[}\StringTok{\textquotesingle{}prognosis\textquotesingle{}}\NormalTok{] }\OperatorTok{=}\NormalTok{ np.apply\_along\_axis(}\KeywordTok{lambda}\NormalTok{ x: np.array(}\StringTok{\textquotesingle{} \textquotesingle{}}\NormalTok{.join(x), dtype}\OperatorTok{=}\StringTok{"object"}\NormalTok{), }\DecValTok{1}\NormalTok{, test\_top\_3\_prediction)}
\NormalTok{test\_df[}\StringTok{\textquotesingle{}prognosis\textquotesingle{}}\NormalTok{].reset\_index().to\_csv(}\StringTok{\textquotesingle{}./data/VBDP/submission.csv\textquotesingle{}}\NormalTok{, index}\OperatorTok{=}\VariableTok{False}\NormalTok{)}
\end{Highlighting}
\end{Shaded}

\hypertarget{appendix}{%
\section{Appendix}\label{appendix}}

\hypertarget{differences-to-the-spotpython-approaches-for-torch-sklearn-and-river}{%
\subsection{\texorpdfstring{Differences to the spotPython Approaches for
\texttt{torch}, \texttt{sklearn} and
\texttt{river}}{Differences to the spotPython Approaches for torch, sklearn and river}}\label{differences-to-the-spotpython-approaches-for-torch-sklearn-and-river}}

\begin{tcolorbox}[enhanced jigsaw, left=2mm, title=\textcolor{quarto-callout-caution-color}{\faFire}\hspace{0.5em}{Caution: Data Loading in Lightning}, bottomrule=.15mm, titlerule=0mm, breakable, rightrule=.15mm, toprule=.15mm, coltitle=black, colbacktitle=quarto-callout-caution-color!10!white, leftrule=.75mm, arc=.35mm, colframe=quarto-callout-caution-color-frame, bottomtitle=1mm, colback=white, opacitybacktitle=0.6, toptitle=1mm, opacityback=0]

\begin{itemize}
\tightlist
\item
  Data loading is handled independently from the \texttt{fun\_control}
  dictionary by \texttt{Lightning} and \texttt{PyTorch}.
\item
  In contrast to \texttt{spotPython} with \texttt{torch}, \texttt{river}
  and \texttt{sklearn}, the data sets are not added to the
  \texttt{fun\_control} dictionary.
\end{itemize}

\end{tcolorbox}

\hypertarget{sec-specification-of-preprocessing-model-31}{%
\subsubsection{Specification of the Preprocessing
Model}\label{sec-specification-of-preprocessing-model-31}}

The \texttt{fun\_control} dictionary, the \texttt{torch},
\texttt{sklearn}and \texttt{river} versions of \texttt{spotPython} allow
the specification of a data preprocessing pipeline, e.g., for the
scaling of the data or for the one-hot encoding of categorical
variables, see
Section~\ref{sec-specification-of-preprocessing-model-14}. This feature
is not used in the \texttt{Lightning} version.

\begin{tcolorbox}[enhanced jigsaw, left=2mm, title=\textcolor{quarto-callout-caution-color}{\faFire}\hspace{0.5em}{Caution: Data preprocessing in Lightning}, bottomrule=.15mm, titlerule=0mm, breakable, rightrule=.15mm, toprule=.15mm, coltitle=black, colbacktitle=quarto-callout-caution-color!10!white, leftrule=.75mm, arc=.35mm, colframe=quarto-callout-caution-color-frame, bottomtitle=1mm, colback=white, opacitybacktitle=0.6, toptitle=1mm, opacityback=0]

Lightning allows the data preprocessing to be specified in the
\texttt{LightningDataModule} class. It is not considered here, because
it should be computed at one location only.

\end{tcolorbox}

\hypertarget{sec-taking-a-look-at-the-data-31}{%
\subsection{Taking a Look at the
Data}\label{sec-taking-a-look-at-the-data-31}}

\begin{Shaded}
\begin{Highlighting}[]
\ImportTok{import}\NormalTok{ torch}
\ImportTok{from}\NormalTok{ spotPython.light.csvdataset }\ImportTok{import}\NormalTok{ CSVDataset}
\ImportTok{from}\NormalTok{ torch.utils.data }\ImportTok{import}\NormalTok{ DataLoader}
\ImportTok{from}\NormalTok{ torchvision.transforms }\ImportTok{import}\NormalTok{ ToTensor}

\CommentTok{\# Create an instance of CSVDataset}
\NormalTok{dataset }\OperatorTok{=}\NormalTok{ CSVDataset(csv\_file}\OperatorTok{=}\StringTok{"./data/VBDP/train.csv"}\NormalTok{, train}\OperatorTok{=}\VariableTok{True}\NormalTok{)}
\CommentTok{\# show the dimensions of the input data}
\BuiltInTok{print}\NormalTok{(dataset[}\DecValTok{0}\NormalTok{][}\DecValTok{0}\NormalTok{].shape)}
\CommentTok{\# show the first element of the input data}
\BuiltInTok{print}\NormalTok{(dataset[}\DecValTok{0}\NormalTok{][}\DecValTok{0}\NormalTok{])}
\CommentTok{\# show the size of the dataset}
\BuiltInTok{print}\NormalTok{(}\SpecialStringTok{f"Dataset Size: }\SpecialCharTok{\{}\BuiltInTok{len}\NormalTok{(dataset)}\SpecialCharTok{\}}\SpecialStringTok{"}\NormalTok{)}
\end{Highlighting}
\end{Shaded}

\begin{verbatim}
torch.Size([64])
tensor([1., 1., 0., 1., 1., 1., 1., 0., 1., 1., 1., 1., 0., 0., 1., 1., 0., 0.,
        1., 0., 1., 0., 1., 1., 1., 1., 1., 1., 1., 0., 0., 1., 0., 0., 0., 0.,
        1., 0., 0., 0., 0., 0., 1., 0., 1., 0., 1., 0., 0., 0., 0., 1., 0., 1.,
        0., 0., 0., 0., 0., 0., 0., 0., 0., 0.])
Dataset Size: 707
\end{verbatim}

\begin{Shaded}
\begin{Highlighting}[]
\CommentTok{\# Set batch size for DataLoader}
\NormalTok{batch\_size }\OperatorTok{=} \DecValTok{3}
\CommentTok{\# Create DataLoader}
\NormalTok{dataloader }\OperatorTok{=}\NormalTok{ DataLoader(dataset, batch\_size}\OperatorTok{=}\NormalTok{batch\_size, shuffle}\OperatorTok{=}\VariableTok{True}\NormalTok{)}

\CommentTok{\# Iterate over the data in the DataLoader}
\ControlFlowTok{for}\NormalTok{ batch }\KeywordTok{in}\NormalTok{ dataloader:}
\NormalTok{    inputs, targets }\OperatorTok{=}\NormalTok{ batch}
    \BuiltInTok{print}\NormalTok{(}\SpecialStringTok{f"Batch Size: }\SpecialCharTok{\{}\NormalTok{inputs}\SpecialCharTok{.}\NormalTok{size(}\DecValTok{0}\NormalTok{)}\SpecialCharTok{\}}\SpecialStringTok{"}\NormalTok{)}
    \BuiltInTok{print}\NormalTok{(}\StringTok{"{-}{-}{-}{-}{-}{-}{-}{-}{-}{-}{-}{-}{-}{-}{-}"}\NormalTok{)}
    \BuiltInTok{print}\NormalTok{(}\SpecialStringTok{f"Inputs: }\SpecialCharTok{\{}\NormalTok{inputs}\SpecialCharTok{\}}\SpecialStringTok{"}\NormalTok{)}
    \BuiltInTok{print}\NormalTok{(}\SpecialStringTok{f"Targets: }\SpecialCharTok{\{}\NormalTok{targets}\SpecialCharTok{\}}\SpecialStringTok{"}\NormalTok{)}
    \ControlFlowTok{break}
\end{Highlighting}
\end{Shaded}

\begin{verbatim}
Batch Size: 3
---------------
Inputs: tensor([[1., 0., 1., 1., 1., 1., 1., 1., 0., 1., 1., 1., 1., 1., 1., 1., 1., 1.,
         1., 1., 1., 1., 1., 0., 0., 0., 0., 1., 0., 0., 0., 0., 0., 0., 0., 0.,
         1., 0., 1., 0., 0., 0., 1., 0., 1., 1., 0., 1., 0., 0., 1., 0., 1., 1.,
         1., 0., 0., 0., 0., 0., 0., 0., 0., 0.],
        [0., 0., 0., 0., 0., 0., 0., 1., 0., 1., 1., 0., 0., 0., 0., 0., 1., 0.,
         0., 1., 0., 0., 1., 0., 1., 1., 0., 0., 0., 0., 1., 1., 1., 0., 1., 1.,
         1., 1., 1., 0., 0., 0., 0., 0., 0., 0., 0., 0., 0., 0., 0., 0., 0., 0.,
         0., 0., 0., 0., 0., 0., 0., 0., 0., 0.],
        [1., 0., 1., 0., 1., 1., 0., 0., 0., 0., 0., 1., 1., 1., 1., 1., 1., 1.,
         0., 0., 1., 0., 0., 0., 1., 0., 1., 0., 1., 0., 0., 1., 1., 0., 0., 0.,
         0., 0., 1., 1., 1., 1., 1., 1., 1., 0., 0., 0., 0., 0., 1., 0., 0., 0.,
         1., 0., 0., 0., 0., 0., 0., 0., 0., 0.]])
Targets: tensor([6, 2, 2])
\end{verbatim}

\hypertarget{sec-the-mapk-metric-31}{%
\subsection{The MAPK Metric}\label{sec-the-mapk-metric-31}}

Here is an example how the MAPK metric is calculated.

\begin{Shaded}
\begin{Highlighting}[]
\ImportTok{from}\NormalTok{ spotPython.torch.mapk }\ImportTok{import}\NormalTok{ MAPK}
\ImportTok{import}\NormalTok{ torch}
\NormalTok{mapk }\OperatorTok{=}\NormalTok{ MAPK(k}\OperatorTok{=}\DecValTok{2}\NormalTok{)}
\NormalTok{target }\OperatorTok{=}\NormalTok{ torch.tensor([}\DecValTok{0}\NormalTok{, }\DecValTok{1}\NormalTok{, }\DecValTok{2}\NormalTok{, }\DecValTok{2}\NormalTok{])}
\NormalTok{preds }\OperatorTok{=}\NormalTok{ torch.tensor(}
\NormalTok{    [}
\NormalTok{        [}\FloatTok{0.5}\NormalTok{, }\FloatTok{0.2}\NormalTok{, }\FloatTok{0.2}\NormalTok{],  }\CommentTok{\# 0 is in top 2}
\NormalTok{        [}\FloatTok{0.3}\NormalTok{, }\FloatTok{0.4}\NormalTok{, }\FloatTok{0.2}\NormalTok{],  }\CommentTok{\# 1 is in top 2}
\NormalTok{        [}\FloatTok{0.2}\NormalTok{, }\FloatTok{0.4}\NormalTok{, }\FloatTok{0.3}\NormalTok{],  }\CommentTok{\# 2 is in top 2}
\NormalTok{        [}\FloatTok{0.7}\NormalTok{, }\FloatTok{0.2}\NormalTok{, }\FloatTok{0.1}\NormalTok{],  }\CommentTok{\# 2 isn\textquotesingle{}t in top 2}
\NormalTok{    ]}
\NormalTok{)}
\NormalTok{mapk.update(preds, target)}
\BuiltInTok{print}\NormalTok{(mapk.compute()) }\CommentTok{\# tensor(0.6250)}
\end{Highlighting}
\end{Shaded}

\begin{verbatim}
tensor(0.6250)
\end{verbatim}

\cleardoublepage
\phantomsection
\addcontentsline{toc}{part}{Appendices}
\appendix

\hypertarget{documentation-of-the-sequential-parameter-optimization}{%
\chapter{Documentation of the Sequential Parameter
Optimization}\label{documentation-of-the-sequential-parameter-optimization}}

This document describes the \texttt{Spot} features. The official
\texttt{spotPython} documentation can be found here:
\url{https://sequential-parameter-optimization.github.io/spotPython/}.

\hypertarget{example-spot}{%
\section{Example: spot}\label{example-spot}}

\begin{Shaded}
\begin{Highlighting}[]
\ImportTok{import}\NormalTok{ numpy }\ImportTok{as}\NormalTok{ np}
\ImportTok{from}\NormalTok{ math }\ImportTok{import}\NormalTok{ inf}
\ImportTok{from}\NormalTok{ spotPython.fun.objectivefunctions }\ImportTok{import}\NormalTok{ analytical}
\ImportTok{from}\NormalTok{ spotPython.spot }\ImportTok{import}\NormalTok{ spot}
\ImportTok{from}\NormalTok{ scipy.optimize }\ImportTok{import}\NormalTok{ shgo}
\ImportTok{from}\NormalTok{ scipy.optimize }\ImportTok{import}\NormalTok{ direct}
\ImportTok{from}\NormalTok{ scipy.optimize }\ImportTok{import}\NormalTok{ differential\_evolution}
\ImportTok{import}\NormalTok{ matplotlib.pyplot }\ImportTok{as}\NormalTok{ plt}
\end{Highlighting}
\end{Shaded}

\hypertarget{the-objective-function}{%
\subsection{The Objective Function}\label{the-objective-function}}

The \texttt{spotPython} package provides several classes of objective
functions. We will use an analytical objective function, i.e., a
function that can be described by a (closed) formula: \[f(x) = x^2\]

\begin{Shaded}
\begin{Highlighting}[]
\NormalTok{fun }\OperatorTok{=}\NormalTok{ analytical().fun\_sphere}
\end{Highlighting}
\end{Shaded}

\begin{Shaded}
\begin{Highlighting}[]
\NormalTok{x }\OperatorTok{=}\NormalTok{ np.linspace(}\OperatorTok{{-}}\DecValTok{1}\NormalTok{,}\DecValTok{1}\NormalTok{,}\DecValTok{100}\NormalTok{).reshape(}\OperatorTok{{-}}\DecValTok{1}\NormalTok{,}\DecValTok{1}\NormalTok{)}
\NormalTok{y }\OperatorTok{=}\NormalTok{ fun(x)}
\NormalTok{plt.figure()}
\NormalTok{plt.plot(x,y, }\StringTok{"k"}\NormalTok{)}
\NormalTok{plt.show()}
\end{Highlighting}
\end{Shaded}

\begin{figure}[H]

{\centering \includegraphics{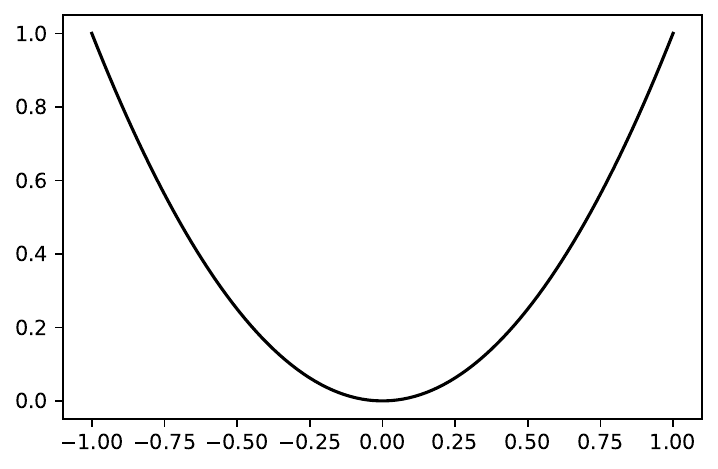}

}

\end{figure}

\begin{Shaded}
\begin{Highlighting}[]
\NormalTok{spot\_1 }\OperatorTok{=}\NormalTok{ spot.Spot(fun}\OperatorTok{=}\NormalTok{fun,}
\NormalTok{                   lower }\OperatorTok{=}\NormalTok{ np.array([}\OperatorTok{{-}}\DecValTok{10}\NormalTok{]),}
\NormalTok{                   upper }\OperatorTok{=}\NormalTok{ np.array([}\DecValTok{100}\NormalTok{]),}
\NormalTok{                   fun\_evals }\OperatorTok{=} \DecValTok{7}\NormalTok{,}
\NormalTok{                   fun\_repeats }\OperatorTok{=} \DecValTok{1}\NormalTok{,}
\NormalTok{                   max\_time }\OperatorTok{=}\NormalTok{ inf,}
\NormalTok{                   noise }\OperatorTok{=} \VariableTok{False}\NormalTok{,}
\NormalTok{                   tolerance\_x }\OperatorTok{=}\NormalTok{ np.sqrt(np.spacing(}\DecValTok{1}\NormalTok{)),}
\NormalTok{                   var\_type}\OperatorTok{=}\NormalTok{[}\StringTok{"num"}\NormalTok{],}
\NormalTok{                   infill\_criterion }\OperatorTok{=} \StringTok{"y"}\NormalTok{,}
\NormalTok{                   n\_points }\OperatorTok{=} \DecValTok{1}\NormalTok{,}
\NormalTok{                   seed}\OperatorTok{=}\DecValTok{123}\NormalTok{,}
\NormalTok{                   log\_level }\OperatorTok{=} \DecValTok{50}\NormalTok{,}
\NormalTok{                   show\_models}\OperatorTok{=}\VariableTok{True}\NormalTok{,}
\NormalTok{                   fun\_control }\OperatorTok{=}\NormalTok{ \{\},}
\NormalTok{                   design\_control}\OperatorTok{=}\NormalTok{\{}\StringTok{"init\_size"}\NormalTok{: }\DecValTok{5}\NormalTok{,}
                                   \StringTok{"repeats"}\NormalTok{: }\DecValTok{1}\NormalTok{\},}
\NormalTok{                   surrogate\_control}\OperatorTok{=}\NormalTok{\{}\StringTok{"noise"}\NormalTok{: }\VariableTok{False}\NormalTok{,}
                                      \StringTok{"cod\_type"}\NormalTok{: }\StringTok{"norm"}\NormalTok{,}
                                      \StringTok{"min\_theta"}\NormalTok{: }\OperatorTok{{-}}\DecValTok{4}\NormalTok{,}
                                      \StringTok{"max\_theta"}\NormalTok{: }\DecValTok{3}\NormalTok{,}
                                      \StringTok{"n\_theta"}\NormalTok{: }\DecValTok{1}\NormalTok{,}
                                      \StringTok{"model\_optimizer"}\NormalTok{: differential\_evolution,}
                                      \StringTok{"model\_fun\_evals"}\NormalTok{: }\DecValTok{1000}\NormalTok{,}
\NormalTok{                                      \})}
\end{Highlighting}
\end{Shaded}

\texttt{spot}'s \texttt{\_\_init\_\_} method sets the control
parameters. There are two parameter groups:

\begin{enumerate}
\def\labelenumi{\arabic{enumi}.}
\tightlist
\item
  external parameters can be specified by the user
\item
  internal parameters, which are handled by \texttt{spot}.
\end{enumerate}

\hypertarget{external-parameters}{%
\subsection{External Parameters}\label{external-parameters}}

\begin{longtable}[]{@{}
  >{\raggedright\arraybackslash}p{(\columnwidth - 8\tabcolsep) * \real{0.2000}}
  >{\raggedright\arraybackslash}p{(\columnwidth - 8\tabcolsep) * \real{0.2000}}
  >{\raggedright\arraybackslash}p{(\columnwidth - 8\tabcolsep) * \real{0.2000}}
  >{\raggedright\arraybackslash}p{(\columnwidth - 8\tabcolsep) * \real{0.2000}}
  >{\raggedright\arraybackslash}p{(\columnwidth - 8\tabcolsep) * \real{0.2000}}@{}}
\toprule\noalign{}
\begin{minipage}[b]{\linewidth}\raggedright
external parameter
\end{minipage} & \begin{minipage}[b]{\linewidth}\raggedright
type
\end{minipage} & \begin{minipage}[b]{\linewidth}\raggedright
description
\end{minipage} & \begin{minipage}[b]{\linewidth}\raggedright
default
\end{minipage} & \begin{minipage}[b]{\linewidth}\raggedright
mandatory
\end{minipage} \\
\midrule\noalign{}
\endhead
\bottomrule\noalign{}
\endlastfoot
\texttt{fun} & object & objective function & & yes \\
\texttt{lower} & array & lower bound & & yes \\
\texttt{upper} & array & upper bound & & yes \\
\texttt{fun\_evals} & int & number of function evaluations & 15 & no \\
\texttt{fun\_evals} & int & number of function evaluations & 15 & no \\
\texttt{fun\_control} & dict & noise etc. & \{\} & n \\
\texttt{max\_time} & int & max run time budget & \texttt{inf} & no \\
\texttt{noise} & bool & if repeated evaluations of \texttt{fun} results
in different values, then \texttt{noise} should be set to \texttt{True}.
& \texttt{False} & no \\
\texttt{tolerance\_x} & float & tolerance for new x solutions. Minimum
distance of new solutions, generated by \texttt{suggest\_new\_X}, to
already existing solutions. If zero (which is the default), every new
solution is accepted. & \texttt{0} & no \\
\texttt{var\_type} & list & list of type information, can be either
\texttt{"num"} or \texttt{"factor"} & \texttt{{[}"num"{]}} & no \\
\texttt{infill\_criterion} & string & Can be \texttt{"y"}, \texttt{"s"},
\texttt{"ei"} (negative expected improvement), or \texttt{"all"} &
\texttt{"y"} & no \\
\texttt{n\_points} & int & number of infill points & 1 & no \\
\texttt{seed} & int & initial seed. If \texttt{Spot.run()} is called
twice, different results will be generated. To reproduce results, the
\texttt{seed} can be used. & \texttt{123} & no \\
\texttt{log\_level} & int & log level with the following settings:
\texttt{NOTSET} (\texttt{0}), \texttt{DEBUG} (\texttt{10}: Detailed
information, typically of interest only when diagnosing problems.),
\texttt{INFO} (\texttt{20}: Confirmation that things are working as
expected.), \texttt{WARNING} (\texttt{30}: An indication that something
unexpected happened, or indicative of some problem in the near future
(e.g.~`disk space low'). The software is still working as expected.),
\texttt{ERROR} (\texttt{40}: Due to a more serious problem, the software
has not been able to perform some function.), and \texttt{CRITICAL}
(\texttt{50}: A serious error, indicating that the program itself may be
unable to continue running.) & \texttt{50} & no \\
\texttt{show\_models} & bool & Plot model. Currently only 1-dim
functions are supported & \texttt{False} & no \\
\texttt{design} & object & experimental design & \texttt{None} & no \\
\texttt{design\_control} & dict & control parameters & see below & no \\
\texttt{surrogate} & & surrogate model & \texttt{kriging} & no \\
\texttt{surrogate\_control} & dict & control parameters & see below &
no \\
\texttt{optimizer} & object & optimizer & see below & no \\
\texttt{optimizer\_control} & dict & control parameters & see below &
no \\
\end{longtable}

\begin{itemize}
\tightlist
\item
  Besides these single parameters, the following parameter dictionaries
  can be specified by the user:

  \begin{itemize}
  \tightlist
  \item
    \texttt{fun\_control}
  \item
    \texttt{design\_control}
  \item
    \texttt{surrogate\_control}
  \item
    \texttt{optimizer\_control}
  \end{itemize}
\end{itemize}

\hypertarget{the-fun_control-dictionary}{%
\section{\texorpdfstring{The \texttt{fun\_control}
Dictionary}{The fun\_control Dictionary}}\label{the-fun_control-dictionary}}

\begin{longtable}[]{@{}lllll@{}}
\toprule\noalign{}
external parameter & type & description & default & mandatory \\
\midrule\noalign{}
\endhead
\bottomrule\noalign{}
\endlastfoot
\texttt{sigma} & float & noise: standard deviation & \texttt{0} & yes \\
\texttt{seed} & int & seed for rng & \texttt{124} & yes \\
\end{longtable}

\hypertarget{the-design_control-dictionary}{%
\section{\texorpdfstring{The \texttt{design\_control}
Dictionary}{The design\_control Dictionary}}\label{the-design_control-dictionary}}

\begin{longtable}[]{@{}
  >{\raggedright\arraybackslash}p{(\columnwidth - 8\tabcolsep) * \real{0.2000}}
  >{\raggedright\arraybackslash}p{(\columnwidth - 8\tabcolsep) * \real{0.2000}}
  >{\raggedright\arraybackslash}p{(\columnwidth - 8\tabcolsep) * \real{0.2000}}
  >{\raggedright\arraybackslash}p{(\columnwidth - 8\tabcolsep) * \real{0.2000}}
  >{\raggedright\arraybackslash}p{(\columnwidth - 8\tabcolsep) * \real{0.2000}}@{}}
\toprule\noalign{}
\begin{minipage}[b]{\linewidth}\raggedright
external parameter
\end{minipage} & \begin{minipage}[b]{\linewidth}\raggedright
type
\end{minipage} & \begin{minipage}[b]{\linewidth}\raggedright
description
\end{minipage} & \begin{minipage}[b]{\linewidth}\raggedright
default
\end{minipage} & \begin{minipage}[b]{\linewidth}\raggedright
mandatory
\end{minipage} \\
\midrule\noalign{}
\endhead
\bottomrule\noalign{}
\endlastfoot
\texttt{init\_size} & int & initial sample size & \texttt{10} & yes \\
\texttt{repeats} & int & number of repeats of the initial sammples &
\texttt{1} & yes \\
\end{longtable}

\hypertarget{the-surrogate_control-dictionary}{%
\section{\texorpdfstring{The \texttt{surrogate\_control}
Dictionary}{The surrogate\_control Dictionary}}\label{the-surrogate_control-dictionary}}

\begin{longtable}[]{@{}
  >{\raggedright\arraybackslash}p{(\columnwidth - 8\tabcolsep) * \real{0.2000}}
  >{\raggedright\arraybackslash}p{(\columnwidth - 8\tabcolsep) * \real{0.2000}}
  >{\raggedright\arraybackslash}p{(\columnwidth - 8\tabcolsep) * \real{0.2000}}
  >{\raggedright\arraybackslash}p{(\columnwidth - 8\tabcolsep) * \real{0.2000}}
  >{\raggedright\arraybackslash}p{(\columnwidth - 8\tabcolsep) * \real{0.2000}}@{}}
\toprule\noalign{}
\begin{minipage}[b]{\linewidth}\raggedright
external parameter
\end{minipage} & \begin{minipage}[b]{\linewidth}\raggedright
type
\end{minipage} & \begin{minipage}[b]{\linewidth}\raggedright
description
\end{minipage} & \begin{minipage}[b]{\linewidth}\raggedright
default
\end{minipage} & \begin{minipage}[b]{\linewidth}\raggedright
mandatory
\end{minipage} \\
\midrule\noalign{}
\endhead
\bottomrule\noalign{}
\endlastfoot
\texttt{noise} & & & & \\
\texttt{model\_optimizer} & object & optimizer &
\texttt{differential\_evolution} & no \\
\texttt{model\_fun\_evals} & & & & \\
\texttt{min\_theta} & & & \texttt{-3.} & \\
\texttt{max\_theta} & & & \texttt{3.} & \\
\texttt{n\_theta} & & & \texttt{1} & \\
\texttt{n\_p} & & & \texttt{1} & \\
\texttt{optim\_p} & & & \texttt{False} & \\
\texttt{cod\_type} & & & \texttt{"norm"} & \\
\texttt{var\_type} & & & & \\
\texttt{use\_cod\_y} & bool & & \texttt{False} & \\
\end{longtable}

\hypertarget{the-optimizer_control-dictionary}{%
\section{\texorpdfstring{The \texttt{optimizer\_control}
Dictionary}{The optimizer\_control Dictionary}}\label{the-optimizer_control-dictionary}}

\begin{longtable}[]{@{}
  >{\raggedright\arraybackslash}p{(\columnwidth - 8\tabcolsep) * \real{0.2000}}
  >{\raggedright\arraybackslash}p{(\columnwidth - 8\tabcolsep) * \real{0.2000}}
  >{\raggedright\arraybackslash}p{(\columnwidth - 8\tabcolsep) * \real{0.2000}}
  >{\raggedright\arraybackslash}p{(\columnwidth - 8\tabcolsep) * \real{0.2000}}
  >{\raggedright\arraybackslash}p{(\columnwidth - 8\tabcolsep) * \real{0.2000}}@{}}
\toprule\noalign{}
\begin{minipage}[b]{\linewidth}\raggedright
external parameter
\end{minipage} & \begin{minipage}[b]{\linewidth}\raggedright
type
\end{minipage} & \begin{minipage}[b]{\linewidth}\raggedright
description
\end{minipage} & \begin{minipage}[b]{\linewidth}\raggedright
default
\end{minipage} & \begin{minipage}[b]{\linewidth}\raggedright
mandatory
\end{minipage} \\
\midrule\noalign{}
\endhead
\bottomrule\noalign{}
\endlastfoot
\texttt{max\_iter} & int & max number of iterations. Note: these are the
cheap evaluations on the surrogate. & \texttt{1000} & no \\
\end{longtable}

\hypertarget{run}{%
\section{Run}\label{run}}

\begin{Shaded}
\begin{Highlighting}[]
\NormalTok{spot\_1.run()}
\end{Highlighting}
\end{Shaded}

\begin{figure}[H]

{\centering \includegraphics{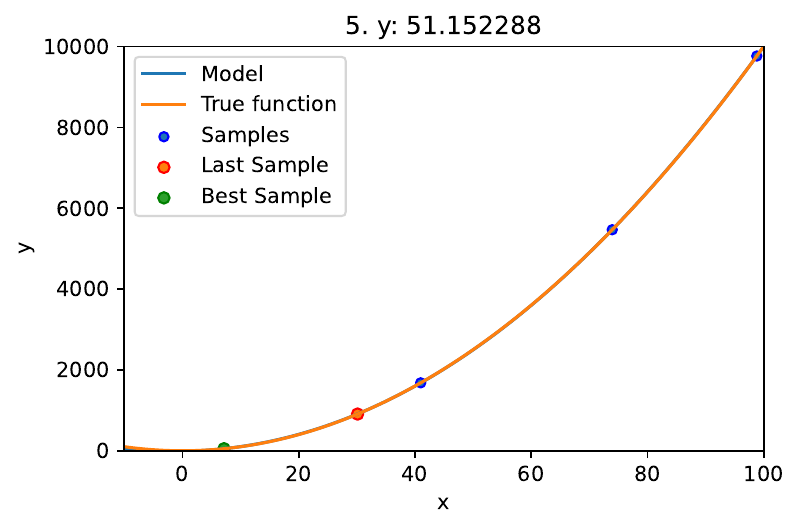}

}

\end{figure}

\begin{figure}[H]

{\centering \includegraphics{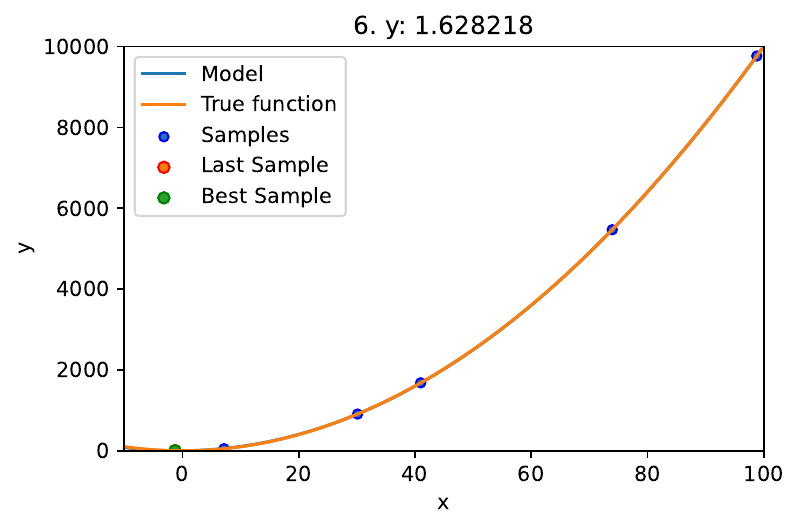}

}

\end{figure}

\begin{verbatim}
spotPython tuning: 1.6282181269484761 [#########-] 85.71% 
\end{verbatim}

\begin{figure}[H]

{\centering \includegraphics{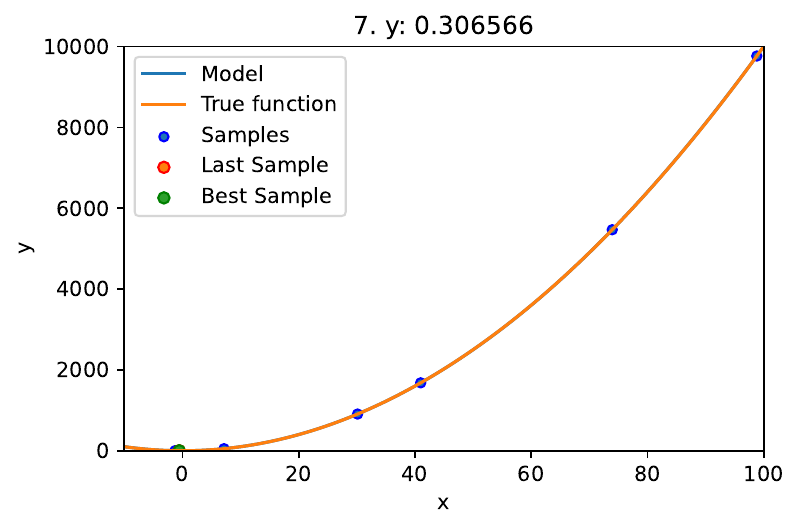}

}

\end{figure}

\begin{verbatim}
spotPython tuning: 0.30656551286610595 [##########] 100.00% Done...
\end{verbatim}

\begin{verbatim}
<spotPython.spot.spot.Spot at 0x165192fe0>
\end{verbatim}

\hypertarget{print-the-results-5}{%
\section{Print the Results}\label{print-the-results-5}}

\begin{Shaded}
\begin{Highlighting}[]
\NormalTok{spot\_1.print\_results()}
\end{Highlighting}
\end{Shaded}

\begin{verbatim}
min y: 0.30656551286610595
x0: -0.5536835855126157
\end{verbatim}

\begin{verbatim}
[['x0', -0.5536835855126157]]
\end{verbatim}

\hypertarget{show-the-progress-2}{%
\section{Show the Progress}\label{show-the-progress-2}}

\begin{Shaded}
\begin{Highlighting}[]
\NormalTok{spot\_1.plot\_progress()}
\end{Highlighting}
\end{Shaded}

\begin{figure}[H]

{\centering \includegraphics{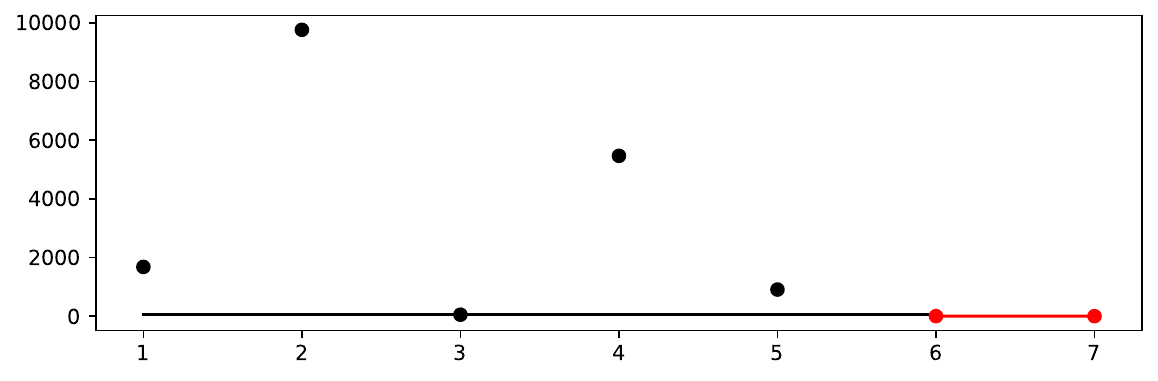}

}

\end{figure}

\hypertarget{visualize-the-surrogate}{%
\section{Visualize the Surrogate}\label{visualize-the-surrogate}}

\begin{itemize}
\tightlist
\item
  The plot method of the \texttt{kriging} surrogate is used.
\item
  Note: the plot uses the interval defined by the ranges of the natural
  variables.
\end{itemize}

\begin{Shaded}
\begin{Highlighting}[]
\NormalTok{spot\_1.surrogate.plot()}
\end{Highlighting}
\end{Shaded}

\begin{verbatim}
<Figure size 2700x1800 with 0 Axes>
\end{verbatim}

\begin{figure}[H]

{\centering \includegraphics{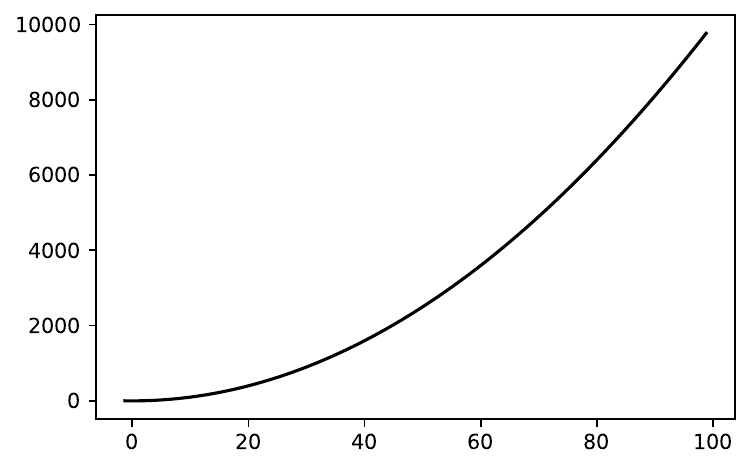}

}

\end{figure}

\hypertarget{init-build-initial-design-1}{%
\section{Init: Build Initial Design}\label{init-build-initial-design-1}}

\begin{Shaded}
\begin{Highlighting}[]
\ImportTok{from}\NormalTok{ spotPython.design.spacefilling }\ImportTok{import}\NormalTok{ spacefilling}
\ImportTok{from}\NormalTok{ spotPython.build.kriging }\ImportTok{import}\NormalTok{ Kriging}
\ImportTok{from}\NormalTok{ spotPython.fun.objectivefunctions }\ImportTok{import}\NormalTok{ analytical}
\NormalTok{gen }\OperatorTok{=}\NormalTok{ spacefilling(}\DecValTok{2}\NormalTok{)}
\NormalTok{rng }\OperatorTok{=}\NormalTok{ np.random.RandomState(}\DecValTok{1}\NormalTok{)}
\NormalTok{lower }\OperatorTok{=}\NormalTok{ np.array([}\OperatorTok{{-}}\DecValTok{5}\NormalTok{,}\OperatorTok{{-}}\DecValTok{0}\NormalTok{])}
\NormalTok{upper }\OperatorTok{=}\NormalTok{ np.array([}\DecValTok{10}\NormalTok{,}\DecValTok{15}\NormalTok{])}
\NormalTok{fun }\OperatorTok{=}\NormalTok{ analytical().fun\_branin}
\NormalTok{fun\_control }\OperatorTok{=}\NormalTok{ \{}\StringTok{"sigma"}\NormalTok{: }\DecValTok{0}\NormalTok{,}
               \StringTok{"seed"}\NormalTok{: }\DecValTok{123}\NormalTok{\}}

\NormalTok{X }\OperatorTok{=}\NormalTok{ gen.scipy\_lhd(}\DecValTok{10}\NormalTok{, lower}\OperatorTok{=}\NormalTok{lower, upper }\OperatorTok{=}\NormalTok{ upper)}
\BuiltInTok{print}\NormalTok{(X)}
\NormalTok{y }\OperatorTok{=}\NormalTok{ fun(X, fun\_control}\OperatorTok{=}\NormalTok{fun\_control)}
\BuiltInTok{print}\NormalTok{(y)}
\end{Highlighting}
\end{Shaded}

\begin{verbatim}
[[ 8.97647221 13.41926847]
 [ 0.66946019  1.22344228]
 [ 5.23614115 13.78185824]
 [ 5.6149825  11.5851384 ]
 [-1.72963184  1.66516096]
 [-4.26945568  7.1325531 ]
 [ 1.26363761 10.17935555]
 [ 2.88779942  8.05508969]
 [-3.39111089  4.15213772]
 [ 7.30131231  5.22275244]]
[128.95676449  31.73474356 172.89678121 126.71295908  64.34349975
  70.16178611  48.71407916  31.77322887  76.91788181  30.69410529]
\end{verbatim}

\hypertarget{replicability}{%
\section{Replicability}\label{replicability}}

Seed

\begin{Shaded}
\begin{Highlighting}[]
\NormalTok{gen }\OperatorTok{=}\NormalTok{ spacefilling(}\DecValTok{2}\NormalTok{, seed}\OperatorTok{=}\DecValTok{123}\NormalTok{)}
\NormalTok{X0 }\OperatorTok{=}\NormalTok{ gen.scipy\_lhd(}\DecValTok{3}\NormalTok{)}
\NormalTok{gen }\OperatorTok{=}\NormalTok{ spacefilling(}\DecValTok{2}\NormalTok{, seed}\OperatorTok{=}\DecValTok{345}\NormalTok{)}
\NormalTok{X1 }\OperatorTok{=}\NormalTok{ gen.scipy\_lhd(}\DecValTok{3}\NormalTok{)}
\NormalTok{X2 }\OperatorTok{=}\NormalTok{ gen.scipy\_lhd(}\DecValTok{3}\NormalTok{)}
\NormalTok{gen }\OperatorTok{=}\NormalTok{ spacefilling(}\DecValTok{2}\NormalTok{, seed}\OperatorTok{=}\DecValTok{123}\NormalTok{)}
\NormalTok{X3 }\OperatorTok{=}\NormalTok{ gen.scipy\_lhd(}\DecValTok{3}\NormalTok{)}
\NormalTok{X0, X1, X2, X3}
\end{Highlighting}
\end{Shaded}

\begin{verbatim}
(array([[0.77254938, 0.31539299],
        [0.59321338, 0.93854273],
        [0.27469803, 0.3959685 ]]),
 array([[0.78373509, 0.86811887],
        [0.06692621, 0.6058029 ],
        [0.41374778, 0.00525456]]),
 array([[0.121357  , 0.69043832],
        [0.41906219, 0.32838498],
        [0.86742658, 0.52910374]]),
 array([[0.77254938, 0.31539299],
        [0.59321338, 0.93854273],
        [0.27469803, 0.3959685 ]]))
\end{verbatim}

\hypertarget{surrogates}{%
\section{Surrogates}\label{surrogates}}

\hypertarget{a-simple-predictor-1}{%
\subsection{A Simple Predictor}\label{a-simple-predictor-1}}

The code below shows how to use a simple model for prediction. Assume
that only two (very costly) measurements are available:

\begin{enumerate}
\def\labelenumi{\arabic{enumi}.}
\tightlist
\item
  f(0) = 0.5
\item
  f(2) = 2.5
\end{enumerate}

We are interested in the value at \(x_0 = 1\), i.e., \(f(x_0 = 1)\), but
cannot run an additional, third experiment.

\begin{Shaded}
\begin{Highlighting}[]
\ImportTok{from}\NormalTok{ sklearn }\ImportTok{import}\NormalTok{ linear\_model}
\NormalTok{X }\OperatorTok{=}\NormalTok{ np.array([[}\DecValTok{0}\NormalTok{], [}\DecValTok{2}\NormalTok{]])}
\NormalTok{y }\OperatorTok{=}\NormalTok{ np.array([}\FloatTok{0.5}\NormalTok{, }\FloatTok{2.5}\NormalTok{])}
\NormalTok{S\_lm }\OperatorTok{=}\NormalTok{ linear\_model.LinearRegression()}
\NormalTok{S\_lm }\OperatorTok{=}\NormalTok{ S\_lm.fit(X, y)}
\NormalTok{X0 }\OperatorTok{=}\NormalTok{ np.array([[}\DecValTok{1}\NormalTok{]])}
\NormalTok{y0 }\OperatorTok{=}\NormalTok{ S\_lm.predict(X0)}
\BuiltInTok{print}\NormalTok{(y0)}
\end{Highlighting}
\end{Shaded}

\begin{verbatim}
[1.5]
\end{verbatim}

Central Idea: Evaluation of the surrogate model \texttt{S\_lm} is much
cheaper (or / and much faster) than running the real-world experiment
\(f\).

\hypertarget{demotest-objective-function-fails}{%
\section{Demo/Test: Objective Function
Fails}\label{demotest-objective-function-fails}}

SPOT expects \texttt{np.nan} values from failed objective function
values. These are handled. Note: SPOT's counter considers only
successful executions of the objective function.

\begin{Shaded}
\begin{Highlighting}[]
\ImportTok{import}\NormalTok{ numpy }\ImportTok{as}\NormalTok{ np}
\ImportTok{from}\NormalTok{ spotPython.fun.objectivefunctions }\ImportTok{import}\NormalTok{ analytical}
\ImportTok{from}\NormalTok{ spotPython.spot }\ImportTok{import}\NormalTok{ spot}
\ImportTok{import}\NormalTok{ numpy }\ImportTok{as}\NormalTok{ np}
\ImportTok{from}\NormalTok{ math }\ImportTok{import}\NormalTok{ inf}
\CommentTok{\# number of initial points:}
\NormalTok{ni }\OperatorTok{=} \DecValTok{20}
\CommentTok{\# number of points}
\NormalTok{n }\OperatorTok{=} \DecValTok{30}

\NormalTok{fun }\OperatorTok{=}\NormalTok{ analytical().fun\_random\_error}
\NormalTok{lower }\OperatorTok{=}\NormalTok{ np.array([}\OperatorTok{{-}}\DecValTok{1}\NormalTok{])}
\NormalTok{upper }\OperatorTok{=}\NormalTok{ np.array([}\DecValTok{1}\NormalTok{])}
\NormalTok{design\_control}\OperatorTok{=}\NormalTok{\{}\StringTok{"init\_size"}\NormalTok{: ni\}}

\NormalTok{spot\_1 }\OperatorTok{=}\NormalTok{ spot.Spot(fun}\OperatorTok{=}\NormalTok{fun,}
\NormalTok{            lower }\OperatorTok{=}\NormalTok{ lower,}
\NormalTok{            upper}\OperatorTok{=}\NormalTok{ upper,}
\NormalTok{            fun\_evals }\OperatorTok{=}\NormalTok{ n,}
\NormalTok{            show\_progress}\OperatorTok{=}\VariableTok{False}\NormalTok{,}
\NormalTok{            design\_control}\OperatorTok{=}\NormalTok{design\_control,)}
\NormalTok{spot\_1.run()}
\CommentTok{\# To check whether the run was successfully completed,}
\CommentTok{\# we compare the number of evaluated points to the specified}
\CommentTok{\# number of points.}
\ControlFlowTok{assert}\NormalTok{ spot\_1.y.shape[}\DecValTok{0}\NormalTok{] }\OperatorTok{==}\NormalTok{ n}
\end{Highlighting}
\end{Shaded}

\begin{verbatim}
[ 0.53176481 -0.9053821  -0.02203599 -0.21843718  0.78240941 -0.58120945
 -0.3923345   0.67234256         nan         nan -0.75129705  0.97550354
  0.41757584         nan  0.82585329  0.23700598 -0.49274073 -0.82319082
 -0.17991251  0.1481835 ]
[-1.]
[-0.47259301]
\end{verbatim}

\begin{verbatim}
[0.95541987]
\end{verbatim}

\begin{verbatim}
\end{verbatim}

\begin{verbatim}
[nan]
[0.17335968]
\end{verbatim}

\begin{verbatim}
[nan]
\end{verbatim}

\begin{verbatim}

[-0.58552368]
[-0.20126111]
\end{verbatim}

\begin{verbatim}
[-0.60100809]
[-0.97897336]
\end{verbatim}

\begin{verbatim}
[-0.2748985]
[0.8359486]
\end{verbatim}

\begin{verbatim}
[0.99035591]
[0.01641232]
\end{verbatim}

\begin{verbatim}
[nan]
[0.5629346]
\end{verbatim}

\newpage{}

\hypertarget{sec-detailed-data-splitting}{%
\section{PyTorch: Detailed Description of the Data
Splitting}\label{sec-detailed-data-splitting}}

\hypertarget{description-of-the-train_hold_out-setting}{%
\subsection{\texorpdfstring{Description of the
\texttt{"train\_hold\_out"}
Setting}{Description of the "train\_hold\_out" Setting}}\label{description-of-the-train_hold_out-setting}}

The \texttt{"train\_hold\_out"} setting is used by default. It uses the
loss function specfied in \texttt{fun\_control} and the metric specified
in \texttt{fun\_control}.

\begin{enumerate}
\def\labelenumi{\arabic{enumi}.}
\tightlist
\item
  First, the method \texttt{HyperTorch().fun\_torch} is called.
\item
  \texttt{fun\_torc()}, which is implemented in the file
  \texttt{hypertorch.py}, calls \texttt{evaluate\_hold\_out()} as
  follows:
\end{enumerate}

\begin{Shaded}
\begin{Highlighting}[]
\NormalTok{df\_eval, \_ = evaluate\_hold\_out(}
\NormalTok{    model,}
\NormalTok{    train\_dataset=fun\_control["train"],}
\NormalTok{    shuffle=self.fun\_control["shuffle"],}
\NormalTok{    loss\_function=self.fun\_control["loss\_function"],}
\NormalTok{    metric=self.fun\_control["metric\_torch"],}
\NormalTok{    device=self.fun\_control["device"],}
\NormalTok{    show\_batch\_interval=self.fun\_control["show\_batch\_interval"],}
\NormalTok{    path=self.fun\_control["path"],}
\NormalTok{    task=self.fun\_control["task"],}
\NormalTok{    writer=self.fun\_control["writer"],}
\NormalTok{    writerId=config\_id,}
\NormalTok{)}
\end{Highlighting}
\end{Shaded}

Note: Only the data set \texttt{fun\_control{[}"train"{]}} is used for
training and validation. It is used in \texttt{evaluate\_hold\_out} as
follows:

\begin{Shaded}
\begin{Highlighting}[]
\NormalTok{trainloader, valloader = create\_train\_val\_data\_loaders(}
\NormalTok{                dataset=train\_dataset, batch\_size=batch\_size\_instance, shuffle=shuffle}
\NormalTok{            )}
\end{Highlighting}
\end{Shaded}

\texttt{create\_train\_val\_data\_loaders()} splits the
\texttt{train\_dataset} into \texttt{trainloader} and \texttt{valloader}
using \texttt{torch.utils.data.random\_split()} as follows:

\begin{Shaded}
\begin{Highlighting}[]
\NormalTok{def create\_train\_val\_data\_loaders(dataset, batch\_size, shuffle, num\_workers=0):}
\NormalTok{    test\_abs = int(len(dataset) * 0.6)}
\NormalTok{    train\_subset, val\_subset = random\_split(dataset, [test\_abs, len(dataset) {-} test\_abs])}
\NormalTok{    trainloader = torch.utils.data.DataLoader(}
\NormalTok{        train\_subset, batch\_size=int(batch\_size), shuffle=shuffle, num\_workers=num\_workers}
\NormalTok{    )}
\NormalTok{    valloader = torch.utils.data.DataLoader(}
\NormalTok{        val\_subset, batch\_size=int(batch\_size), shuffle=shuffle, num\_workers=num\_workers}
\NormalTok{    )}
\NormalTok{    return trainloader, valloader}
\end{Highlighting}
\end{Shaded}

The optimizer is set up as follows:

\begin{Shaded}
\begin{Highlighting}[]
\NormalTok{optimizer\_instance = net.optimizer}
\NormalTok{lr\_mult\_instance = net.lr\_mult}
\NormalTok{sgd\_momentum\_instance = net.sgd\_momentum}
\NormalTok{optimizer = optimizer\_handler(}
\NormalTok{    optimizer\_name=optimizer\_instance,}
\NormalTok{    params=net.parameters(),}
\NormalTok{    lr\_mult=lr\_mult\_instance,}
\NormalTok{    sgd\_momentum=sgd\_momentum\_instance,}
\NormalTok{)}
\end{Highlighting}
\end{Shaded}

\begin{enumerate}
\def\labelenumi{\arabic{enumi}.}
\setcounter{enumi}{2}
\tightlist
\item
  \texttt{evaluate\_hold\_out()} sets the \texttt{net} attributes such
  as \texttt{epochs}, \texttt{batch\_size}, \texttt{optimizer}, and
  \texttt{patience}. For each epoch, the methods
  \texttt{train\_one\_epoch()} and \texttt{validate\_one\_epoch()} are
  called, the former for training and the latter for validation and
  early stopping. The validation loss from the last epoch (not the best
  validation loss) is returned from \texttt{evaluate\_hold\_out}.
\item
  The method \texttt{train\_one\_epoch()} is implemented as follows:
\end{enumerate}

\begin{Shaded}
\begin{Highlighting}[]
\NormalTok{def train\_one\_epoch(}
\NormalTok{    net,}
\NormalTok{    trainloader,}
\NormalTok{    batch\_size,}
\NormalTok{    loss\_function,}
\NormalTok{    optimizer,}
\NormalTok{    device,}
\NormalTok{    show\_batch\_interval=10\_000,}
\NormalTok{    task=None,}
\NormalTok{):}
\NormalTok{    running\_loss = 0.0}
\NormalTok{    epoch\_steps = 0}
\NormalTok{    for batch\_nr, data in enumerate(trainloader, 0):}
\NormalTok{        input, target = data}
\NormalTok{        input, target = input.to(device), target.to(device)}
\NormalTok{        optimizer.zero\_grad()}
\NormalTok{        output = net(input)}
\NormalTok{        if task == "regression":}
\NormalTok{            target = target.unsqueeze(1)}
\NormalTok{            if target.shape == output.shape:}
\NormalTok{                loss = loss\_function(output, target)}
\NormalTok{            else:}
\NormalTok{                raise ValueError(f"Shapes of target and output do not match:}
\NormalTok{                 \{target.shape\} vs \{output.shape\}")}
\NormalTok{        elif task == "classification":}
\NormalTok{            loss = loss\_function(output, target)}
\NormalTok{        else:}
\NormalTok{            raise ValueError(f"Unknown task: \{task\}")}
\NormalTok{        loss.backward()}
\NormalTok{        torch.nn.utils.clip\_grad\_norm\_(net.parameters(), max\_norm=1.0)}
\NormalTok{        optimizer.step()}
\NormalTok{        running\_loss += loss.item()}
\NormalTok{        epoch\_steps += 1}
\NormalTok{        if batch\_nr \% show\_batch\_interval == (show\_batch\_interval {-} 1):  }
\NormalTok{            print(}
\NormalTok{                "Batch: \%5d. Batch Size: \%d. Training Loss (running): \%.3f"}
\NormalTok{                \% (batch\_nr + 1, int(batch\_size), running\_loss / epoch\_steps)}
\NormalTok{            )}
\NormalTok{            running\_loss = 0.0}
\NormalTok{    return loss.item()}
\end{Highlighting}
\end{Shaded}

\begin{enumerate}
\def\labelenumi{\arabic{enumi}.}
\setcounter{enumi}{4}
\tightlist
\item
  The method \texttt{validate\_one\_epoch()} is implemented as follows:
\end{enumerate}

\begin{Shaded}
\begin{Highlighting}[]
\NormalTok{def validate\_one\_epoch(net, valloader, loss\_function, metric, device, task):}
\NormalTok{    val\_loss = 0.0}
\NormalTok{    val\_steps = 0}
\NormalTok{    total = 0}
\NormalTok{    correct = 0}
\NormalTok{    metric.reset()}
\NormalTok{    for i, data in enumerate(valloader, 0):}
\NormalTok{        \# get batches}
\NormalTok{        with torch.no\_grad():}
\NormalTok{            input, target = data}
\NormalTok{            input, target = input.to(device), target.to(device)}
\NormalTok{            output = net(input)}
\NormalTok{            \# print(f"target: \{target\}")}
\NormalTok{            \# print(f"output: \{output\}")}
\NormalTok{            if task == "regression":}
\NormalTok{                target = target.unsqueeze(1)}
\NormalTok{                if target.shape == output.shape:}
\NormalTok{                    loss = loss\_function(output, target)}
\NormalTok{                else:}
\NormalTok{                    raise ValueError(f"Shapes of target and output }
\NormalTok{                        do not match: \{target.shape\} vs \{output.shape\}")}
\NormalTok{                metric\_value = metric.update(output, target)}
\NormalTok{            elif task == "classification":}
\NormalTok{                loss = loss\_function(output, target)}
\NormalTok{                metric\_value = metric.update(output, target)}
\NormalTok{                \_, predicted = torch.max(output.data, 1)}
\NormalTok{                total += target.size(0)}
\NormalTok{                correct += (predicted == target).sum().item()}
\NormalTok{            else:}
\NormalTok{                raise ValueError(f"Unknown task: \{task\}")}
\NormalTok{            val\_loss += loss.cpu().numpy()}
\NormalTok{            val\_steps += 1}
\NormalTok{    loss = val\_loss / val\_steps}
\NormalTok{    print(f"Loss on hold{-}out set: \{loss\}")}
\NormalTok{    if task == "classification":}
\NormalTok{        accuracy = correct / total}
\NormalTok{        print(f"Accuracy on hold{-}out set: \{accuracy\}")}
\NormalTok{    \# metric on all batches using custom accumulation}
\NormalTok{    metric\_value = metric.compute()}
\NormalTok{    metric\_name = type(metric).\_\_name\_\_}
\NormalTok{    print(f"\{metric\_name\} value on hold{-}out data: \{metric\_value\}")}
\NormalTok{    return metric\_value, loss}
\end{Highlighting}
\end{Shaded}

\hypertarget{description-of-the-test_hold_out-setting}{%
\subsubsection{\texorpdfstring{Description of the
\texttt{"test\_hold\_out"}
Setting}{Description of the "test\_hold\_out" Setting}}\label{description-of-the-test_hold_out-setting}}

It uses the loss function specfied in \texttt{fun\_control} and the
metric specified in \texttt{fun\_control}.

\begin{enumerate}
\def\labelenumi{\arabic{enumi}.}
\tightlist
\item
  First, the method \texttt{HyperTorch().fun\_torch} is called.
\item
  \texttt{fun\_torc()} calls
  \texttt{spotPython.torch.traintest.evaluate\_hold\_out()} similar to
  the \texttt{"train\_hold\_out"} setting with one exception: It passes
  an additional \texttt{test} data set to \texttt{evaluate\_hold\_out()}
  as follows:
\end{enumerate}

\begin{Shaded}
\begin{Highlighting}[]
\NormalTok{test\_dataset=fun\_control["test"]}
\end{Highlighting}
\end{Shaded}

\texttt{evaluate\_hold\_out()} calls
\texttt{create\_train\_test\_data\_loaders} instead of
\texttt{create\_train\_val\_data\_loaders}: The two data sets are used
in \texttt{create\_train\_test\_data\_loaders} as follows:

\begin{Shaded}
\begin{Highlighting}[]
\NormalTok{def create\_train\_test\_data\_loaders(dataset, batch\_size, shuffle, test\_dataset, }
\NormalTok{        num\_workers=0):}
\NormalTok{    trainloader = torch.utils.data.DataLoader(}
\NormalTok{        dataset, batch\_size=int(batch\_size), shuffle=shuffle, }
\NormalTok{        num\_workers=num\_workers}
\NormalTok{    )}
\NormalTok{    testloader = torch.utils.data.DataLoader(}
\NormalTok{        test\_dataset, batch\_size=int(batch\_size), shuffle=shuffle, }
\NormalTok{        num\_workers=num\_workers}
\NormalTok{    )}
\NormalTok{    return trainloader, testloader}
\end{Highlighting}
\end{Shaded}

\begin{enumerate}
\def\labelenumi{\arabic{enumi}.}
\setcounter{enumi}{2}
\tightlist
\item
  The following steps are identical to the \texttt{"train\_hold\_out"}
  setting. Only a different data loader is used for testing.
\end{enumerate}

\hypertarget{detailed-description-of-the-train_cv-setting}{%
\subsubsection{\texorpdfstring{Detailed Description of the
\texttt{"train\_cv"}
Setting}{Detailed Description of the "train\_cv" Setting}}\label{detailed-description-of-the-train_cv-setting}}

It uses the loss function specfied in \texttt{fun\_control} and the
metric specified in \texttt{fun\_control}.

\begin{enumerate}
\def\labelenumi{\arabic{enumi}.}
\tightlist
\item
  First, the method \texttt{HyperTorch().fun\_torch} is called.
\item
  \texttt{fun\_torc()} calls
  \texttt{spotPython.torch.traintest.evaluate\_cv()} as follows (Note:
  Only the data set \texttt{fun\_control{[}"train"{]}} is used for CV.):
\end{enumerate}

\begin{Shaded}
\begin{Highlighting}[]
\NormalTok{df\_eval, \_ = evaluate\_cv(}
\NormalTok{    model,}
\NormalTok{    dataset=fun\_control["train"],}
\NormalTok{    shuffle=self.fun\_control["shuffle"],}
\NormalTok{    device=self.fun\_control["device"],}
\NormalTok{    show\_batch\_interval=self.fun\_control["show\_batch\_interval"],}
\NormalTok{    task=self.fun\_control["task"],}
\NormalTok{    writer=self.fun\_control["writer"],}
\NormalTok{    writerId=config\_id,}
\NormalTok{)}
\end{Highlighting}
\end{Shaded}

\begin{enumerate}
\def\labelenumi{\arabic{enumi}.}
\setcounter{enumi}{2}
\tightlist
\item
  In `evaluate\_cv(), the following steps are performed: The optimizer
  is set up as follows:
\end{enumerate}

\begin{Shaded}
\begin{Highlighting}[]
\NormalTok{optimizer\_instance = net.optimizer}
\NormalTok{lr\_instance = net.lr}
\NormalTok{sgd\_momentum\_instance = net.sgd\_momentum}
\NormalTok{optimizer = optimizer\_handler(optimizer\_name=optimizer\_instance,}
\NormalTok{     params=net.parameters(), lr\_mult=lr\_mult\_instance)}
\end{Highlighting}
\end{Shaded}

\texttt{evaluate\_cv()} sets the \texttt{net} attributes such as
\texttt{epochs}, \texttt{batch\_size}, \texttt{optimizer}, and
\texttt{patience}. CV is implemented as follows:

\begin{Shaded}
\begin{Highlighting}[]
\NormalTok{def evaluate\_cv(}
\NormalTok{    net,}
\NormalTok{    dataset,}
\NormalTok{    shuffle=False,}
\NormalTok{    loss\_function=None,}
\NormalTok{    num\_workers=0,}
\NormalTok{    device=None,}
\NormalTok{    show\_batch\_interval=10\_000,}
\NormalTok{    metric=None,}
\NormalTok{    path=None,}
\NormalTok{    task=None,}
\NormalTok{    writer=None,}
\NormalTok{    writerId=None,}
\NormalTok{):}
\NormalTok{    lr\_mult\_instance = net.lr\_mult}
\NormalTok{    epochs\_instance = net.epochs}
\NormalTok{    batch\_size\_instance = net.batch\_size}
\NormalTok{    k\_folds\_instance = net.k\_folds}
\NormalTok{    optimizer\_instance = net.optimizer}
\NormalTok{    patience\_instance = net.patience}
\NormalTok{    sgd\_momentum\_instance = net.sgd\_momentum}
\NormalTok{    removed\_attributes, net = get\_removed\_attributes\_and\_base\_net(net)}
\NormalTok{    metric\_values = \{\}}
\NormalTok{    loss\_values = \{\}}
\NormalTok{    try:}
\NormalTok{        device = getDevice(device=device)}
\NormalTok{        if torch.cuda.is\_available():}
\NormalTok{            device = "cuda:0"}
\NormalTok{            if torch.cuda.device\_count() \textgreater{} 1:}
\NormalTok{                print("We will use", torch.cuda.device\_count(), "GPUs!")}
\NormalTok{                net = nn.DataParallel(net)}
\NormalTok{        net.to(device)}
\NormalTok{        optimizer = optimizer\_handler(}
\NormalTok{            optimizer\_name=optimizer\_instance,}
\NormalTok{            params=net.parameters(),}
\NormalTok{            lr\_mult=lr\_mult\_instance,}
\NormalTok{            sgd\_momentum=sgd\_momentum\_instance,}
\NormalTok{        )}
\NormalTok{        kfold = KFold(n\_splits=k\_folds\_instance, shuffle=shuffle)}
\NormalTok{        for fold, (train\_ids, val\_ids) in enumerate(kfold.split(dataset)):}
\NormalTok{            print(f"Fold: \{fold + 1\}")}
\NormalTok{            train\_subsampler = torch.utils.data.SubsetRandomSampler(train\_ids)}
\NormalTok{            val\_subsampler = torch.utils.data.SubsetRandomSampler(val\_ids)}
\NormalTok{            trainloader = torch.utils.data.DataLoader(}
\NormalTok{                dataset, batch\_size=batch\_size\_instance, }
\NormalTok{                sampler=train\_subsampler, num\_workers=num\_workers}
\NormalTok{            )}
\NormalTok{            valloader = torch.utils.data.DataLoader(}
\NormalTok{                dataset, batch\_size=batch\_size\_instance, }
\NormalTok{                sampler=val\_subsampler, num\_workers=num\_workers}
\NormalTok{            )}
\NormalTok{            \# each fold starts with new weights:}
\NormalTok{            reset\_weights(net)}
\NormalTok{            \# Early stopping parameters}
\NormalTok{            best\_val\_loss = float("inf")}
\NormalTok{            counter = 0}
\NormalTok{            for epoch in range(epochs\_instance):}
\NormalTok{                print(f"Epoch: \{epoch + 1\}")}
\NormalTok{                \# training loss from one epoch:}
\NormalTok{                training\_loss = train\_one\_epoch(}
\NormalTok{                    net=net,}
\NormalTok{                    trainloader=trainloader,}
\NormalTok{                    batch\_size=batch\_size\_instance,}
\NormalTok{                    loss\_function=loss\_function,}
\NormalTok{                    optimizer=optimizer,}
\NormalTok{                    device=device,}
\NormalTok{                    show\_batch\_interval=show\_batch\_interval,}
\NormalTok{                    task=task,}
\NormalTok{                )}
\NormalTok{                \# Early stopping check. Calculate validation loss from one epoch:}
\NormalTok{                metric\_values[fold], loss\_values[fold] = validate\_one\_epoch(}
\NormalTok{                    net, valloader=valloader, loss\_function=loss\_function, }
\NormalTok{                    metric=metric, device=device, task=task}
\NormalTok{                )}
\NormalTok{                \# Log the running loss averaged per batch}
\NormalTok{                metric\_name = "Metric"}
\NormalTok{                if metric is None:}
\NormalTok{                    metric\_name = type(metric).\_\_name\_\_}
\NormalTok{                    print(f"\{metric\_name\} value on hold{-}out data: }
\NormalTok{                        \{metric\_values[fold]\}")}
\NormalTok{                if writer is not None:}
\NormalTok{                    writer.add\_scalars(}
\NormalTok{                        "evaluate\_cv fold:" + str(fold + 1) + }
\NormalTok{                        ". Train \& Val Loss and Val Metric" + writerId,}
\NormalTok{                        \{"Train loss": training\_loss, "Val loss": }
\NormalTok{                        loss\_values[fold], metric\_name: metric\_values[fold]\},}
\NormalTok{                        epoch + 1,}
\NormalTok{                    )}
\NormalTok{                    writer.flush()}
\NormalTok{                if loss\_values[fold] \textless{} best\_val\_loss:}
\NormalTok{                    best\_val\_loss = loss\_values[fold]}
\NormalTok{                    counter = 0}
\NormalTok{                    \# save model:}
\NormalTok{                    if path is not None:}
\NormalTok{                        torch.save(net.state\_dict(), path)}
\NormalTok{                else:}
\NormalTok{                    counter += 1}
\NormalTok{                    if counter \textgreater{}= patience\_instance:}
\NormalTok{                        print(f"Early stopping at epoch \{epoch\}")}
\NormalTok{                        break}
\NormalTok{        df\_eval = sum(loss\_values.values()) / len(loss\_values.values())}
\NormalTok{        df\_metrics = sum(metric\_values.values()) / len(metric\_values.values())}
\NormalTok{        df\_preds = np.nan}
\NormalTok{    except Exception as err:}
\NormalTok{        print(f"Error in Net\_Core. Call to evaluate\_cv() failed. \{err=\}, }
\NormalTok{            \{type(err)=\}")}
\NormalTok{        df\_eval = np.nan}
\NormalTok{        df\_preds = np.nan}
\NormalTok{    add\_attributes(net, removed\_attributes)}
\NormalTok{    if writer is not None:}
\NormalTok{        metric\_name = "Metric"}
\NormalTok{        if metric is None:}
\NormalTok{            metric\_name = type(metric).\_\_name\_\_}
\NormalTok{        writer.add\_scalars(}
\NormalTok{            "CV: Val Loss and Val Metric" + writerId,}
\NormalTok{            \{"CV{-}loss": df\_eval, metric\_name: df\_metrics\},}
\NormalTok{            epoch + 1,}
\NormalTok{        )}
\NormalTok{        writer.flush()}
\NormalTok{    return df\_eval, df\_preds, df\_metrics}
\end{Highlighting}
\end{Shaded}

\begin{enumerate}
\def\labelenumi{\arabic{enumi}.}
\setcounter{enumi}{3}
\item
  The method \texttt{train\_fold()} is implemented as shown above.
\item
  The method \texttt{validate\_one\_epoch()} is implemented as shown
  above. In contrast to the hold-out setting, it is called for each of
  the \(k\) folds. The results are stored in a dictionaries
  \texttt{metric\_values} and \texttt{loss\_values}. The results are
  averaged over the \(k\) folds and returned as \texttt{df\_eval}.
\end{enumerate}

\hypertarget{detailed-description-of-the-test_cv-setting}{%
\subsubsection{\texorpdfstring{Detailed Description of the
\texttt{"test\_cv"}
Setting}{Detailed Description of the "test\_cv" Setting}}\label{detailed-description-of-the-test_cv-setting}}

It uses the loss function specfied in \texttt{fun\_control} and the
metric specified in \texttt{fun\_control}.

\begin{enumerate}
\def\labelenumi{\arabic{enumi}.}
\tightlist
\item
  First, the method \texttt{HyperTorch().fun\_torch} is called.
\item
  \texttt{fun\_torc()} calls
  \texttt{spotPython.torch.traintest.evaluate\_cv()} as follows:
\end{enumerate}

\begin{Shaded}
\begin{Highlighting}[]
\NormalTok{df\_eval, \_ = evaluate\_cv(}
\NormalTok{    model,}
\NormalTok{    dataset=fun\_control["test"],}
\NormalTok{    shuffle=self.fun\_control["shuffle"],}
\NormalTok{    device=self.fun\_control["device"],}
\NormalTok{    show\_batch\_interval=self.fun\_control["show\_batch\_interval"],}
\NormalTok{    task=self.fun\_control["task"],}
\NormalTok{    writer=self.fun\_control["writer"],}
\NormalTok{    writerId=config\_id,}
\NormalTok{)}
\end{Highlighting}
\end{Shaded}

Note: The data set \texttt{fun\_control{[}"test"{]}} is used for CV. The
rest is the same as for the \texttt{"train\_cv"} setting.

\hypertarget{sec-final-model-evaluation}{%
\subsubsection{Detailed Description of the Final Model Training and
Evaluation}\label{sec-final-model-evaluation}}

There are two methods that can be used for the final evaluation of a
Pytorch model:

\begin{enumerate}
\def\labelenumi{\arabic{enumi}.}
\tightlist
\item
  \texttt{"train\_tuned} and
\item
  \texttt{"test\_tuned"}.
\end{enumerate}

\texttt{train\_tuned()} is just a wrapper to
\texttt{evaluate\_hold\_out} using the \texttt{train} data set. It is
implemented as follows:

\begin{Shaded}
\begin{Highlighting}[]
\NormalTok{def train\_tuned(}
\NormalTok{    net,}
\NormalTok{    train\_dataset,}
\NormalTok{    shuffle,}
\NormalTok{    loss\_function,}
\NormalTok{    metric,}
\NormalTok{    device=None,}
\NormalTok{    show\_batch\_interval=10\_000,}
\NormalTok{    path=None,}
\NormalTok{    task=None,}
\NormalTok{    writer=None,}
\NormalTok{):}
\NormalTok{    evaluate\_hold\_out(}
\NormalTok{        net=net,}
\NormalTok{        train\_dataset=train\_dataset,}
\NormalTok{        shuffle=shuffle,}
\NormalTok{        test\_dataset=None,}
\NormalTok{        loss\_function=loss\_function,}
\NormalTok{        metric=metric,}
\NormalTok{        device=device,}
\NormalTok{        show\_batch\_interval=show\_batch\_interval,}
\NormalTok{        path=path,}
\NormalTok{        task=task,}
\NormalTok{        writer=writer,}
\NormalTok{    )}
\end{Highlighting}
\end{Shaded}

The \texttt{test\_tuned()} procedure is implemented as follows:

\begin{Shaded}
\begin{Highlighting}[]
\NormalTok{def test\_tuned(net, shuffle, test\_dataset=None, loss\_function=None,}
\NormalTok{    metric=None, device=None, path=None, task=None):}
\NormalTok{    batch\_size\_instance = net.batch\_size}
\NormalTok{    removed\_attributes, net = get\_removed\_attributes\_and\_base\_net(net)}
\NormalTok{    if path is not None:}
\NormalTok{        net.load\_state\_dict(torch.load(path))}
\NormalTok{        net.eval()}
\NormalTok{    try:}
\NormalTok{        device = getDevice(device=device)}
\NormalTok{        if torch.cuda.is\_available():}
\NormalTok{            device = "cuda:0"}
\NormalTok{            if torch.cuda.device\_count() \textgreater{} 1:}
\NormalTok{                print("We will use", torch.cuda.device\_count(), "GPUs!")}
\NormalTok{                net = nn.DataParallel(net)}
\NormalTok{        net.to(device)}
\NormalTok{        valloader = torch.utils.data.DataLoader(}
\NormalTok{            test\_dataset, batch\_size=int(batch\_size\_instance),}
\NormalTok{            shuffle=shuffle, }
\NormalTok{            num\_workers=0}
\NormalTok{        )}
\NormalTok{        metric\_value, loss = validate\_one\_epoch(}
\NormalTok{            net, valloader=valloader, loss\_function=loss\_function,}
\NormalTok{            metric=metric, device=device, task=task}
\NormalTok{        )}
\NormalTok{        df\_eval = loss}
\NormalTok{        df\_metric = metric\_value}
\NormalTok{        df\_preds = np.nan}
\NormalTok{    except Exception as err:}
\NormalTok{        print(f"Error in Net\_Core. Call to test\_tuned() failed. \{err=\}, }
\NormalTok{            \{type(err)=\}")}
\NormalTok{        df\_eval = np.nan}
\NormalTok{        df\_metric = np.nan}
\NormalTok{        df\_preds = np.nan}
\NormalTok{    add\_attributes(net, removed\_attributes)}
\NormalTok{    print(f"Final evaluation: Validation loss: \{df\_eval\}")}
\NormalTok{    print(f"Final evaluation: Validation metric: \{df\_metric\}")}
\NormalTok{    print("{-}{-}{-}{-}{-}{-}{-}{-}{-}{-}{-}{-}{-}{-}{-}{-}{-}{-}{-}{-}{-}{-}{-}{-}{-}{-}{-}{-}{-}{-}{-}{-}{-}{-}{-}{-}{-}{-}{-}{-}{-}{-}{-}{-}{-}{-}")}
\NormalTok{    return df\_eval, df\_preds, df\_metric}
\end{Highlighting}
\end{Shaded}

\hypertarget{references}{%
\chapter*{References}\label{references}}
\addcontentsline{toc}{chapter}{References}

\markboth{References}{References}

\hypertarget{refs}{}
\begin{CSLReferences}{1}{0}
\leavevmode\vadjust pre{\hypertarget{ref-bart21i}{}}%
Bartz, Eva, Thomas Bartz-Beielstein, Martin Zaefferer, and Olaf
Mersmann, eds. 2022. \emph{{Hyperparameter Tuning for Machine and Deep
Learning with R - A Practical Guide}}. Springer.

\leavevmode\vadjust pre{\hypertarget{ref-bart23e}{}}%
Bartz-Beielstein, Thomas. 2023. {``{PyTorch} Hyperparameter Tuning with
{SPOT}: Comparison with {Ray Tuner} and Default Hyperparameters on
{CIFAR10}.''}
\url{https://github.com/sequential-parameter-optimization/spotPython/blob/main/notebooks/14_spot_ray_hpt_torch_cifar10.ipynb}.

\leavevmode\vadjust pre{\hypertarget{ref-Bart13j}{}}%
Bartz-Beielstein, Thomas, Jürgen Branke, Jörn Mehnen, and Olaf Mersmann.
2014. {``Evolutionary Algorithms.''} \emph{Wiley Interdisciplinary
Reviews: Data Mining and Knowledge Discovery} 4 (3): 178--95.

\leavevmode\vadjust pre{\hypertarget{ref-bart20gArxiv}{}}%
Bartz-Beielstein, Thomas, Carola Doerr, Jakob Bossek, Sowmya
Chandrasekaran, Tome Eftimov, Andreas Fischbach, Pascal Kerschke, et al.
2020. {``Benchmarking in Optimization: Best Practice and Open Issues.''}
arXiv. \url{https://arxiv.org/abs/2007.03488}.

\leavevmode\vadjust pre{\hypertarget{ref-BLP05}{}}%
Bartz-Beielstein, Thomas, Christian Lasarczyk, and Mike Preuss. 2005.
{``{Sequential Parameter Optimization}.''} In \emph{{Proceedings 2005
Congress on Evolutionary Computation (CEC'05), Edinburgh, Scotland}},
edited by B McKay et al., 773--80. Piscataway NJ: {IEEE Press}.

\leavevmode\vadjust pre{\hypertarget{ref-Torczon00}{}}%
Lewis, R M, V Torczon, and M W Trosset. 2000. {``{Direct search methods:
Then and now}.''} \emph{Journal of Computational and Applied
Mathematics} 124 (1--2): 191--207.

\leavevmode\vadjust pre{\hypertarget{ref-Li16a}{}}%
Li, Lisha, Kevin Jamieson, Giulia DeSalvo, Afshin Rostamizadeh, and
Ameet Talwalkar. 2016. {``{Hyperband: A Novel Bandit-Based Approach to
Hyperparameter Optimization}.''} \emph{arXiv e-Prints}, March,
arXiv:1603.06560.

\leavevmode\vadjust pre{\hypertarget{ref-Meignan:2015vp}{}}%
Meignan, David, Sigrid Knust, Jean-Marc Frayet, Gilles Pesant, and
Nicolas Gaud. 2015. {``{A Review and Taxonomy of Interactive
Optimization Methods in Operations Research}.''} \emph{ACM Transactions
on Interactive Intelligent Systems}, September.

\leavevmode\vadjust pre{\hypertarget{ref-mont20a}{}}%
Montiel, Jacob, Max Halford, Saulo Martiello Mastelini, Geoffrey
Bolmier, Raphael Sourty, Robin Vaysse, Adil Zouitine, et al. 2021.
{``River: Machine Learning for Streaming Data in Python.''}

\leavevmode\vadjust pre{\hypertarget{ref-pyto23a}{}}%
PyTorch. 2023a. {``Hyperparameter Tuning with Ray Tune.''}
\url{https://pytorch.org/tutorials/beginner/hyperparameter_tuning_tutorial.html}.

\leavevmode\vadjust pre{\hypertarget{ref-pyto23b}{}}%
---------. 2023b. {``Training a Classifier.''}
\url{https://pytorch.org/tutorials/beginner/blitz/cifar10_tutorial.html}.

\end{CSLReferences}

\end{document}